\documentclass[
		twoside,openright,titlepage,numbers=noenddot,headinclude,%1headlines,
	 	footinclude=true,cleardoublepage=empty,
		dottedtoc, % Make page numbers in the table of contents flushed right with dots leading to them
		BCOR=5mm,paper=a4,fontsize=11pt, % Binding correction, paper type and font size
		ngerman,american, % Languages, change this to your language(s)
		]{scrreprt}

\PassOptionsToPackage{utf8}{inputenc}
\usepackage{inputenc}

\PassOptionsToPackage{eulerchapternumbers,listings, pdfspacing, subfig,beramono,parts}{classicthesis}

\newcommand{\myTitle}{
The Physics of Data and Tasks: Theories of Locality and Compositionality in Deep Learning\xspace
}

\newcommand{\myName}{Alessandro Favero\xspace}

\newcommand{\myFaculty}{School of Basic Sciences\xspace}

\newcommand{\myUni}{EPFL\xspace}

\newcommand{\myTime}{September 2025\xspace}
\newcommand{\myVersion}{version 3.0\xspace}

\newcounter{dummy}
\providecommand{\mLyX}{L\kern-.1667em\lower.25em\hbox{Y}\kern-.125emX\@}

\usepackage{lipsum}

\usepackage{babel}

\usepackage{csquotes}
\PassOptionsToPackage{
backend=bibtex8,bibencoding=ascii,
language=auto,
style=alphabetic,
sorting=nyt,
maxbibnames=10,
maxcitenames=2,
natbib=true
}{biblatex}
\usepackage{biblatex}
 
\PassOptionsToPackage{fleqn}{amsmath}
 \usepackage{amsmath}
 
\PassOptionsToPackage{T1}{fontenc}
\usepackage{fontenc}

\usepackage{textcomp}

\usepackage{scrhack}

\usepackage{xspace}

\usepackage{mparhack}

\usepackage{fixltx2e}

\PassOptionsToPackage{smaller}{acronym}
\usepackage{acronym}

\PassOptionsToPackage{pdftex}{graphicx}
\usepackage{graphicx} 

\usepackage{tabularx}
\usepackage{caption}
\usepackage{subfig}  

\usepackage{listings} 
\PassOptionsToPackage{pdftex,hyperfootnotes=false,pdfpagelabels}{hyperref}
\usepackage{hyperref}
\hypersetup{
colorlinks=true, linktocpage=true, pdfstartpage=3, pdfstartview=FitV,
breaklinks=true, pdfpagemode=UseNone, pageanchor=true, pdfpagemode=UseOutlines,
plainpages=false, bookmarksnumbered, bookmarksopen=true, bookmarksopenlevel=1,
hypertexnames=true, pdfhighlight=/O,
urlcolor=Black, linkcolor=RoyalBlue, citecolor=RoyalBlue,
pdftitle={\myTitle},
pdfauthor={\textcopyright\ \myName, \myUni, \myFaculty},
pdfsubject={},
pdfkeywords={},
pdfcreator={pdfLaTeX},
pdfproducer={LaTeX with hyperref and classicthesis}
}

\makeatletter
\@ifpackageloaded{babel}
{
\addto\extrasamerican{

}

\addto\extrasngerman{

}

}{\relax}
\makeatother

\usepackage{classicthesis} 

\def\testerr{{\mathcal{E}}}

\usepackage{bm}
\usepackage{amssymb}
\usepackage{amsthm}

\def\KT{{\mathcal{K}_{\mathrm{T}}}}
\def\KS{{\mathcal{K}_{\mathrm{S}}}}
\def\K{{\mathcal{K}}}
\DeclareMathOperator*{\argmin}{argmin}  
\DeclareMathOperator{\eq}{\,{=}\,}

\newtheorem{theorem}{Theorem}[section]
\newtheorem{corollary}{Corollary}[theorem]
\newtheorem{lemma}[theorem]{Lemma}
\newtheorem{definition}{Definition}[section]

\newcommand{\rm}{\mathrm}
\newcommand{\sf}{\mathsf}
\newcommand{\norm}[1]{\left\lVert#1\right\rVert}
\DeclareMathOperator*{\2dots}{\,{.}\,{.}\,}

\def\bk{{\bm{k}}}
\def\bell{{\bm{\ell}}}

\newcommand{\bigtimes}{\mathop{\times}\limits}

\usepackage{tikz}
\usetikzlibrary{positioning, arrows.meta, decorations.markings}

\def\k{{\mathbf{k}}}
\def\x{{\mathbf{x}}}

\def\y{\mathbf{y}}
\def\w{{\mathbf{w}}}
\def\z{{\mathbf{z}}}

\def\E{{\mathbb{E}}}

\def\O{\mathcal{O}}

\newcommand{\beq}{\begin{equation}}
\newcommand{\eeq}{\end{equation}}
\newcommand{\beqs}{\begin{equation*}}
\newcommand{\eeqs}{\end{equation*}}
\newcommand{\beqn}{\begin{eqnarray}}
\newcommand{\eeqn}{\end{eqnarray}}
\newcommand{\lpa}{\left(}
\newcommand{\rpa}{\right)}
\newcommand{\lsq}{\left[}
\newcommand{\rsq}{\right]}

\definecolor{C0}{HTML}{1F77B4}
\definecolor{C1}{HTML}{FF7F0E}
\definecolor{C2}{HTML}{2CA02C}
\definecolor{C3}{HTML}{D62728}
\definecolor{C4}{HTML}{9467BD}
\definecolor{C5}{HTML}{8C564B}
\definecolor{C6}{HTML}{E377C2}
\definecolor{C7}{HTML}{7F7F7F}
\definecolor{C8}{HTML}{BCBD22}
\definecolor{C9}{HTML}{17BECF}
\definecolor{Gold}{HTML}{FFD700}
\definecolor{Navy}{HTML}{000080}
\newcommand{\createtree}[5]{
  \pgfmathtruncatemacro{\nextlayer}{#1+1}
  \ifnum#1=#2
    \node [varnode] (#3) [below left= of #4] {};
  \else
      \node [factnode] (#3) [below= of #4] {};
      \draw (#4) -- (#3);
    \foreach \x in {1,2}{
      \pgfmathtruncatemacro{\newname}{#3*10+\x}
      \node [varnode] (\newname) [below= of #3, xshift=(2*\x-3)*(40/#1)] {$X_{\x}$};
      \draw (#3) -- (\newname);
      }

  \fi
}

\newcommand{\nnu}[1]{{\nu_{\uparrow}^{(#1)}}}
\newcommand{\ncu}[1]{{\tilde{\nu}_{\uparrow}^{(#1)}}}
\newcommand{\nnd}[1]{{\nu_{\downarrow}^{(#1)}}}
\newcommand{\ncd}[1]{{\tilde{\nu}_{\downarrow}^{(#1)}}}
\newcommand{\rul}[1]{{\rr^{(#1)}}}
\newcommand{\rr}{{\psi}}

\newcommand{\corr}[1]{{\overline{#1}}}
\newcommand{\bel}{\mathcal{B}}
\newcommand{\samp}{\mathcal{S}}
\newcommand{\ham}[1]{\Delta_{#1}}
\newcommand{\asso}{\{y\}\leftarrow\{\x\}}

\newcommand{\X}{\mathbf{X}}
\def\rh{{\textnormal{h}}}
\newcommand{\xv}{\rh}
\newcommand{\voc}{\mathcal{V}}
\newcommand{\length}{d}
\def\mT{{\bm{T}}}
\def\mC{{\bm{C}}}
\def\mP{{\bm{P}}}
\def\vp{{\bm{p}}}
\def\vz{{\bm{z}}}

\newcommand{\h}{h}
\newcommand{\avg}[1]{\left\langle #1 \right\rangle}
\newtheorem{assumption}[theorem]{Assumption}
\newcommand{\prob}[1]{\mathbb{P}\left[#1\right]}

\usepackage{multirow}

\def\vx{{\mathbf{x}}}

\newcommand{\vth}{\bm{\theta}}
\newcommand{\vt}{\bm{\tau}}
\newcommand{\D}{\mathcal{D}}

\newtheorem{property}{Property}
\newtheorem{proposition}{Proposition}

\usepackage{pdfpages}

\usepackage{longtable}

\begin{document}

\frenchspacing

\raggedbottom

\selectlanguage{american}

\pagenumbering{roman}

\pagestyle{plain}

%----------------------------------------------------------------------------------------
%	PRE-CONTENT THESIS PAGES
%----------------------------------------------------------------------------------------

\begin{titlepage}

\begin{addmargin}[-1cm]{-3cm}
\begin{center}
\large

\hfill
\vfill

\begingroup
\color{Maroon}\spacedallcaps{\myTitle} \\ \bigskip
\endgroup

\spacedlowsmallcaps{\myName}

\vfill

Presented on 26 September 2025 \\
for the award of the degree of PhD in Physics \\
at the École Polytechnique Fédérale de Lausanne (EPFL).

\bigskip

Accepted on the jury’s recommendation:

Prof. C. G. Theiler, jury president

Prof. M. Wyart, thesis director

Prof. P. Frossard, thesis director

Prof. E. Vanden-Eijnden, examiner

Prof. S. Ganguli, examiner

Prof. E. Abbé, examiner

\vfill

Preprint. Official version at:

\href{https://infoscience.epfl.ch/handle/20.500.14299/254241}{\texttt{infoscience.epfl.ch/handle/20.500.14299/254241}}

% DOI: 10.5075/epfl-thesis-11462

\end{center}
\end{addmargin}

\end{titlepage}

\thispagestyle{empty}

\hfill

\vfill

\noindent\myName: \textit{\myTitle,}
\textcopyright\ \myTime

\cleardoublepage\pdfbookmark[1]{Abstract}{Abstract}

\begingroup
\let\clearpage\relax
\let\cleardoublepage\relax
\let\cleardoublepage\relax

\sloppy

\chapter*{Abstract}
Deep neural networks have achieved remarkable success, yet our understanding of how they learn remains limited. These models can learn high-dimensional tasks, which is generally statistically intractable due to the \textit{curse of dimensionality}. This apparent paradox suggests that learnable data must have an underlying latent structure. What is the nature of this structure? How do neural networks encode and exploit it, and how does it quantitatively impact performance -- for instance, how does generalization improve with the number of training examples? This thesis addresses these questions by studying the roles of locality and compositionality in data, tasks, and deep learning representations. 

We begin by analyzing \textit{convolutional neural networks} in the limit of infinite width, where the learning dynamics simplifies and becomes analytically tractable. Using tools from statistical physics and learning theory, we characterize their generalization abilities and show that they can overcome the curse of
dimensionality if the target function is local by adapting to its spatial scale.

We then turn to more complex structures in which features are composed hierarchically, with elements at larger scales built from sub-features at smaller ones. We model such data using simple \textit{probabilistic context-free grammars} -- tree-like graphical models used to describe data such as language and images. Within this framework, we study how \textit{diffusion-based generative models} compose new data by assembling features learned from examples. This theory of composition predicts a phase transition in the generative process, which we confirm empirically in both image and language modalities, providing support for the compositional structure of natural data. We further demonstrate that the sample complexity for learning these grammars scales polynomially with data dimension, providing a mechanism by which diffusion models avoid the curse of dimensionality by learning to hierarchically compose new data. These results offer a theoretical grounding for how generative models learn to generalize, and ultimately, become \textit{creative}.

Finally, we shift our analysis from the structure of data in the input space to the structure of tasks in the model's parameter space. Here, we investigate a novel form of compositionality, where tasks and skills themselves can be composed. In particular, we empirically demonstrate that distinct directions in the weight space of large pre-trained models are associated with localized, semantic task-specific areas in function space, and how this modular structure enables \textit{task arithmetic} and model editing at scale. 

\paragraph{Keywords} Deep learning, generalization, scaling laws, data structure, locality, compositionality, probabilistic graphical models, convolutional networks, diffusion models.

\endgroup			

\vfill

\cleardoublepage\pdfbookmark[1]{Resume}{Resume}

\begingroup
\let\clearpage\relax
\let\cleardoublepage\relax
\let\cleardoublepage\relax

\sloppy

\chapter*{Resumé}
Les réseaux de neurones profonds ont connu un succès remarquable, mais notre compréhension de leur mode d’apprentissage reste limitée. Ces modèles peuvent apprendre à partir de données de haute dimension, ce qui est en général statistiquement intraitable en raison de la malédiction de la dimensionnalité. Ce paradoxe apparent suggère que les données apprenables doivent posséder une structure latente sous-jacente. Quelle est la nature de cette structure ? Comment les réseaux de neurones l’encodent-ils et l’exploitent-ils, et quel est son impact quantitatif sur les performances -- par exemple, comment la généralisation s’améliore-t-elle avec le nombre d’exemples d’entraînement ? Cette thèse aborde ces questions en étudiant les rôles de la localité et de la compositionnalité dans les données, les tâches et les représentations issues de l’apprentissage profond.

Nous commençons par analyser les réseaux de neurones convolutifs dans la limite de largeur infinie. À l’aide d’outils issus de la physique statistique et de la théorie de l’apprentissage, nous caractérisons leurs capacités de généralisation et montrons qu’ils peuvent surmonter la malédiction de la dimensionnalité si la fonction est locale, en s’adaptant à son échelle spatiale.

Nous nous intéressons ensuite à des structures plus complexes, dans lesquelles les caractéristiques sont composées hiérarchiquement, avec des éléments à grande échelle construits à partir de sous caractéristiques à plus petite échelle. Nous modélisons de telles données à l’aide de simples grammaires hors-contexte probabilistes -- des modèles graphiques en forme d’arbre utilisés pour décrire des données telles que le langage et les images. Dans ce cadre, nous étudions comment les modèles génératifs par diffusion composent de nouvelles données en assemblant des caractéristiques apprises à partir d’exemples. Cette théorie de la composition prédit une transition de phase dans le processus génératif, que nous confirmons empiriquement dans les modalités image et langage, ce qui soutient l’hypothèse d’une structure compositionnelle des données naturelles. Nous démontrons en outre que le nombre d'exemples nécessaires à l'apprentissage de ces grammaires croît polynomialement avec la dimension des données, fournissant ainsi un mécanisme par lequel les modèles de diffusion évitent la malédiction de la dimensionnalité en apprenant à composer de nouvelles données de manière hiérarchique. Ces résultats offrent une base théorique à la compréhension de la généralisation chez les modèles génératifs, et, en fin de compte, de leur capacité à devenir créatifs.

Enfin, nous explorons comment les tâches et les compétences elles-mêmes peuvent être localisées et composées dans l’espace des poids du modèle. En particulier, nous montrons empiriquement que des directions distinctes dans l’espace des poids de grands modèles préentraînés sont associées à des zones sémantiques spécifiques et localisées dans l’espace des fonctions, et comment cette structure modulaire permet une arithmétique des tâches et l’édition de modèles à grande échelle.

\paragraph{Mots-clés} Apprentissage profond, généralisation, lois d'échelle, structure des données, localité, compositionnalité, modèles graphiques probabilistes, réseaux convolutifs, modèles de diffusion.

\endgroup			

\vfill

\cleardoublepage\begin{flushright}{\slshape    
The true sign of intelligence is not knowledge but imagination.} \\ \medskip
--- Albert Einstein
\end{flushright}

\bigskip

%----------------------------------------------------------------------------------------

\begingroup

\let\clearpage\relax
\let\cleardoublepage\relax
\let\cleardoublepage\relax

\chapter*{Acknowledgements}

The years spent on this PhD have been a period of profound personal and academic transformation. I have had the privilege of meeting wonderful people, traveling the world, and growing in ways I had not anticipated. The achievements detailed in this thesis, and indeed any personal growth I have experienced, are by no means my merit alone. They are the result of a deeply collective endeavor, a reflection of the incredible people I have been surrounded by. This is, to me, the most beautiful way to do science.

My deepest gratitude goes first to my two PhD advisors, Prof. Matthieu Wyart and Prof. Pascal Frossard.

To Matthieu, thank you for your unwavering belief in me from the very beginning. I am indebted to you for the countless hours of scientific discussion that have fundamentally shaped how I think. You taught me the meaning of scientific rigor, clarity of thought, and the importance of honest feedback. You showed me how to approach problems with precision without ever losing sight of the underlying intuition, and you instilled in me a unique research taste for which I will always be grateful.

To Pascal, thank you for trusting me and for granting me immense freedom and independence to grow as a researcher and, later, as a mentor. Your constant support and patience, especially during my time away from LTS4, were invaluable. I am grateful for your wisdom, your practical advice on navigating academia, and for always asking the right questions.

I was honored to have an exceptional thesis committee. My sincere thanks to Prof. Surya Ganguli, Prof. Eric Vanden-Eijnden, Prof. Emmanuel Abbe, and the committee president, Prof. Christian Theiler. I could not have asked for a more distinguished and insightful group of experts in the field. It was a privilege to present my work to you, and I look forward to future scientific conversations.

This thesis would simply not exist without my co-authors: Francesco, Antonio, Noam, and Guille. I feel incredibly fortunate to have worked alongside such brilliant scientists. I am grateful for the time we spent together and immensely proud of the publications we co-authored, which form the core of this work.

Several people have been instrumental in teaching me the art of science. My journey began with Stefano, who co-supervised my Master's thesis, patiently taught me the fundamentals of kernel theory, and introduced me to both theoretical and numerical research.

Francesco, you truly showed me how to do science as we learned together about deep learning, infinite-width limits, and so much more during a global pandemic. Your patience in guiding me through theoretical work was remarkable. I can only hope to have absorbed a fraction of your mathematical and theoretical physics prowess.

Antonio, your knowledge of statistical physics and your impressive ability to simply sit and think deeply about problems -- alongside writing perfect figure captions -- never cease to amaze me. We spent an insane number of hours, nights and weekends included, working on science and running so many experiments that some results are still waiting on a cluster somewhere. Our collaboration has become incredibly smooth, and I hope we continue working together for years to come.

Guille, you introduced me to a different paradigm of experimental science, one geared towards larger-scale and practically impactful inquiry. It is also thanks to you that I now see myself not `just' as a physicist, but also as a machine learning scientist. Your influence on my communication style -- from figures to posters and slides -- has also been profound, and you have inspired me to strive to be as thoughtful and constructive a reviewer as you are.

I am grateful for the vibrant communities at EPFL, within both the Physics of Complex Systems Laboratory (PCSL) and the Signal Processing Laboratory (LTS4). Thank you to Antonio, Daniel, Elisabeth, Francesco, Jack, Leonardo, Mario, Marko, Noam, Riccardo, Stefano, Tom, Umberto, and Wencheng. Perhaps an apology is due to our non-Italian colleagues for their patience with the large Italian contingent, who too often switched languages at lunchtime. Thank you also to all the Master's students who joined us over the years. At LTS4, thanks to Abdellah, Adam, Amel, Ahmet, Alba, Apostolos, Arun, Beril, Cedric, Clement, Dorina, Guillermo, Harshitha, Isabel, Isabela, Javier, Jelena, Jeremy, Ke, Manuel, Mariana, Nikolaos, Ortal, Sevda, Simone, Thibault, Vaishnavi, Vincent, William, Yamin, and Yiming. A special mention to the model merging crew (Guille, Nikos, Ke, Adam, Amel): I had so much fun working and discussing with you all on projects that, while not in this thesis, I hold in very high regard. You are all incredibly smart.

A PhD would be impossible without administrative support. Thank you to Corinne for her energy, enthusiasm, kindness, and for always caring so much about us. To Anne, whose efficiency, precision, and know-how on any imaginable matter were simply invaluable. And to Patricia, for her help in this final year.

Many of the people mentioned above are not just colleagues but have become friends. Among those:
Antonio, thank you for supporting (and tolerating) me for countless hours each week, for being a travel companion across the world for work and vacation -- always chasing the best (and sometimes most expensive) food -- for hosting me in Abruzzo, and for randomly appearing at the gym at the most unexpected times.
Umberto, thank you for sharing this Lausanne experience with me from the very beginning. You are a great listener who cares so much for the people around you. Thank you for being there despite my many ``no''s and occasional flakiness. I look forward to visiting you in Paris!
Manuel, I only regret that we became close friends so late in my PhD journey (for that, thank you, Vancouver!). To semi-quote you, you `just' know how to be a friend, and you make it seem effortless. Your friendship and support have meant more to me, and to this PhD, than you know. I hope we can make up for lost time!

Beyond the labs, my gratitude extends to the other friends I made at EPFL -- with special mentions to the members of Lenka's group on the 5th floor of Cubotron and Florent's group -- and beyond in Lausanne. I am grateful for all the hikes, via ferratas, rafting, barbecues by the lake, and, last but not least, skiing. Here, I must give a special thank you to William for introducing me to this beautiful sport. He gets all the credit for my enthusiasm; `any' flaws in my technique are entirely my own.

I am also thankful for the experiences and friendships forged at the summer schools in Princeton, Les Houches, the Flatiron Institute, and Cortona (INdAM). A special shout-out to the ``junior organizers and \textit{imbucati}'' for the great times in NYC.

My thanks also go to the fundamental research team at Amazon, who trusted me when I was still primarily a physics student and fostered my growth as an applied scientist. Thank you to the teams in Silicon Valley and San Francisco, and especially to the group I worked with between Los Angeles and New York. In particular, to Luca, Matthew, Siddharth, Alessandro, Pramuditha, Ben, and Stefano.

Going further back, I am thankful for the people with whom I shared my Master's degree across Trieste, Turin, and Paris. Without the infinite hours spent studying together, debating physics, and having fun, I would not be the person I am today. A special thanks to Andrea, who joined me at EPFL, for being my flatmate for two years and for always being there to talk and listen. And to Nicolò, for our meetups in Jesolo and for picking up the phone after a year of silence as if we had spoken just yesterday.

I am also grateful to my friends from Treviso. You are my roots. A special mention goes to Ale, who has always been there for me for more than thirteen years, seeing me at my best and my worst. You are like a brother to me.

Ultimately, I am profoundly indebted to my family. It may sound like a cliché, but it is deeply true: thank you to my parents for instilling in me the love of learning and knowledge. You invested your time and resources to allow me to study at the best places, even when it meant being far from home. This dissertation is dedicated to you. To my brother, my grandparents, aunts, uncles, and cousins, thank you for your unwavering support and for cheering me on every single step of the way. I carry you with me always.

This journey has been long, and it was not without its difficulties, sacrifices, and moments of doubt. That I stand here today, defending this work, is a testament to the people I have had the privilege to be surrounded by. My success is truly their success.

\bigskip

\begin{flushright}
	\textit{Alessandro} \\[5mm]
	Lausanne, August 2025
\end{flushright}

\endgroup

\pagestyle{scrheadings}

\cleardoublepage\refstepcounter{dummy}

\pdfbookmark[1]{\contentsname}{tableofcontents}

\setcounter{tocdepth}{1}

\setcounter{secnumdepth}{3}

\manualmark
\markboth{\spacedlowsmallcaps{\contentsname}}{\spacedlowsmallcaps{\contentsname}}
\tableofcontents 
\automark[section]{chapter}
\renewcommand{\chaptermark}[1]{\markboth{\spacedlowsmallcaps{#1}}{\spacedlowsmallcaps{#1}}}
\renewcommand{\sectionmark}[1]{\markright{\thesection\enspace\spacedlowsmallcaps{#1}}}

\clearpage

\begingroup 
\let\clearpage\relax
\let\cleardoublepage\relax
\let\cleardoublepage\relax

\refstepcounter{dummy}
\addcontentsline{toc}{chapter}{\tocEntry{List of Symbols}}

\pdfbookmark[1]{List of Symbols}{symbols}

\markboth{\spacedlowsmallcaps{List of Symbols}}{\spacedlowsmallcaps{List of Symbols}}

\chapter*{List of Symbols}

\subsection*{General Machine Learning \& Neural Networks}

\begin{longtable}{l p{0.8\textwidth}}
    \hline
    \textbf{Symbol} & \textbf{Definition} \\
    \hline
    \endhead
    $\x$ & Input vector or data point, typically in $\mathbb{R}^d$. \\
    $y$ & Label or target value corresponding to an input $\x$. \\
    $d$ & The dimension of the input space. \\
    $P$ & The number of training examples in the dataset. \\
    $\vth$ & The set of all learnable parameters (weights and biases) in a neural network. \\
    $f(\x; \vth)$ & A neural network function that maps an input $\x$ to an output, parameterized by $\vth$. \\
    $w^{(l)}_h, b^{(l)}$ & The weight vector and bias term for neuron $h$ in layer $l$. \\
    $L$ & The depth (number of hidden layers) of a neural network. \\
    $H$ & The width (number of neurons per hidden layer) of a network. \\
    $\sigma(\cdot)$ & A non-linear activation function, such as ReLU. \\
    $\ell(y, y')$ & A loss function measuring the discrepancy between a prediction $y'$ and a true label $y$. \\
    $\mathcal{E}$ & The generalization error, or expected loss over the true data distribution. \\
    $\hat{\mathcal{E}}$ & The empirical risk, or average loss over the training set. \\
    $\eta$ & The learning rate used in gradient descent. \\
    $\beta$ & The learning curve exponent, describing how generalization error scales with dataset size, i.e., $\mathcal{E}(P) \sim P^{-\beta}$. \\
    $\mathcal{H}$ & The hypothesis class, representing the set of all functions a model can express. \\
    \hline
\end{longtable}

\subsection*{Kernel Methods \& Infinite-Width Networks}
\begin{longtable}{l p{0.8\textwidth}}
    \hline
    \textbf{Symbol} & \textbf{Definition} \\
    \hline
    \endhead
    $\mathcal{K}(\x, \x')$ & A kernel function, measuring the similarity between inputs $\x$ and $\x'$. \\
    $\mathcal{K}_{\text{NTK}}$ & The Neural Tangent Kernel, which describes the training dynamics of infinitely wide networks. \\
    $\mathcal{K}_T, \mathcal{K}_S$ & The teacher and student kernels in a teacher-student learning framework. \\
    $\lambda_{\rho}$ & The $\rho$-th eigenvalue of a kernel's integral operator. \\
    $\phi_{\rho}(x)$ & The $\rho$-th eigenfunction of a kernel's integral operator. \\
    $c_{\rho}$ & The coefficient of the target function's projection onto the $\rho$-th eigenfunction of the kernel. \\
    $t, s$ & The filter size (receptive field size) of the teacher and student kernels/networks, respectively. \\
    $\alpha_t, \alpha_s$ & The smoothness exponent of the teacher and student kernels, controlling their non-analytic behavior. \\
    $d_{\text{eff}}(l)$ & The effective dimensionality of the receptive field of a neuron at layer $l$ in a CNN. \\
    \hline
\end{longtable}

\subsection*{Diffusion Models \& Hierarchical Generative Models}
\begin{longtable}{l p{0.8\textwidth}}
    \hline
    \textbf{Symbol} & \textbf{Definition} \\
    \hline
    \endhead
    $p_{\text{data}}(\x)$ & The true, underlying probability distribution of the data. \\
    $p_{\vth}(\x)$ & A parameterized generative model that approximates $p_{\text{data}}(\x)$. \\
    $\x_t$ & Data at time step $t$ in a diffusion process, with $t=0$ being clean data and $t=T$ being pure noise. \\
    $\nabla_{\x} \log p_t(\x)$ & The score function of the data distribution at time $t$. \\
    $s_{\vth}(x, t)$ & The neural network trained to approximate the score function. \\
    $\beta_t, \alpha_t, \bar{\alpha}_t$ & Parameters defining the noise schedule of a DDPM. \\
    RHM & Random Hierarchy Model, a synthetic generative model with a tree-like structure. \\
    $L, s, v, m$ & Key parameters of the RHM: depth, branching factor, vocabulary size, and number of synonyms per rule. \\
    $h_i^{(l)}$ & A latent variable (or hidden symbol) in the RHM at layer $l$ and position $i$. \\
    $\nu_{\uparrow}, \nu_{\downarrow}$ & Upward and downward messages passed in the Belief Propagation algorithm. \\
    $\epsilon$ & A parameter controlling the noise level in the simplified $\epsilon$-process for analyzing the RHM. \\
    $\mathcal{C}_{ij}(t)$ & The dynamical correlation function, measuring the correlation of changes between tokens $i$ and $j$. \\
    $\chi(t)$ & The dynamical susceptibility, measuring the total volume of correlated changes. \\
    $\xi$ & The correlation length of token changes. \\
    \hline
\end{longtable}

\subsection*{Task Compositionality \& Model Editing}
\begin{longtable}{l p{0.8\textwidth}}
    \hline
    \textbf{Symbol} & \textbf{Definition} \\
    \hline
    \endhead
    $\vth_0$ & The parameters of a pre-trained model before fine-tuning. \\
    $\vth_t^*$ & The parameters of a model after fine-tuning on task $t$. \\
    $\bm{\tau}_t$ & The task vector for task $t$, defined as the difference in weights: $\bm{\tau}_t = \vth_t^* - \vth_0$. \\
    $\mathcal{D}_t$ & The data support (the subset of the input space) for a specific task $t$. \\
    $f_{\text{lin}}$ & The linearized version of a neural network function, based on its first-order Taylor expansion around $\vth_0$. \\
    $\xi(\alpha_1, \alpha_2)$ & The disentanglement error, measuring the interference between two tasks when their task vectors are combined. \\
    \hline
\end{longtable}
                   
\endgroup

\cleardoublepage

\pagenumbering{arabic}

\cleardoublepage

%----------------------------------------------------------------------------------------
%	THESIS CONTENT - CHAPTERS
%----------------------------------------------------------------------------------------

\ctparttext{\bigskip \bigskip \begin{flushright}{\slshape
With four parameters I can fit an elephant, and with five I can make him wiggle his trunk.} \\ \medskip
--- John von Neumann
\end{flushright}
}

\part{Overture}

\sloppy

\chapter{Introduction}

\label{ch:introduction}

This chapter establishes both the vocabulary and the open questions that will shape the remainder of the thesis. It first sketches the basics of machine learning and neural networks. It then explains why the striking empirical success of these networks is unexpected, reviews the main present theories, and pinpoints remaining gaps. The chapter concludes with a brief overview of the thesis’s contributions.

\section{Introduction to deep learning}
\label{sec:intro-dl}

\subsection{Supervised learning with deep neural networks}
\label{sec:intro-super}
The most basic setting in machine learning is \textit{supervised learning}, where a model learns a mapping from inputs to labels using examples. For instance, in the task of classifying animal species from a picture, the model is trained on examples of images paired with their correct species labels, with the goal of accurately classifying new, unseen images.

Formally, each input $\x_\nu \in \mathcal{X}$ is paired with a label $y_\nu \in \mathcal{Y}$. The input space $\mathcal{X}$ is often high-dimensional (e.g., $\mathcal{X} = \mathbb{R}^d$ where $d \gg 1$ represents the number of pixels in an image). The output space can represent either real values $\mathcal{Y}=\mathbb{R}$ (for \textit{regression}) or class labels $\mathcal{Y}=\{1,\dots,C\}$ (for \textit{classification}, such as the animal species example above). The learner is given $P$ examples $\{(\x_\nu,y_\nu)\}_{\nu\in P}$, known as the \textit{training set}, where $(\x_\nu, y_\nu)$ are assumed to be drawn i.i.d. from a joint distribution $p$ over $\mathcal{X} \times \mathcal{Y}$.

In \marginpar{Fully connected networks are the simplest neural architectures.} deep learning, the class of models used to learn this mapping is represented by deep neural networks.  In their most basic form, \textit{fully connected neural networks} (FCNs), these models consist of successive layers of linear transformations interspersed with nonlinearities. A network with $L$ hidden layers transforms an input $\x$ into an output $f(\x;\vth)$ through the recursive equations:
\begin{align}
    \label{eq:intro-fcn}
    z_{h}^{(1)} &= \sigma^{(1)} \left(\w_h^{(1)\top} \x + b^{(1)} \right) {\textrm{ for } h \in [H]} \nonumber \\
    z_{h}^{(l)} &= \sigma^{(l)} \left(\frac{1}{\sqrt{H}}\w_h^{(l)\top} \mathbf{z}^{(l-1)} + b^{(l)}\right) {\textrm{ for } h \in [H], \, l \in [2,\dots,L]} \nonumber \\
    f(\x;\vth) &= \mathbf{z}^{(L+1)},
\end{align}
where $\vth$ denotes a vector with all parameters (i.e., weights $\w_h^{(l)}$ and biases $b^{(l)}$ at layers $l \in [L+1]$) that are learned from data. The functions $\sigma^{(l)}$ are scalar activation functions, applied element-wise, such as ReLU nonlinearities $\sigma(u)=\max(0,u)$. $L$ is called the \textit{depth} and $H$ the \textit{width} of the network.

The primary goal of supervised learning is to find parameters $\vth$ that minimize the \textit{generalization error}: \marginpar{Generalization: how a model performs on unseen data.}
\begin{equation}
    \testerr(f) := \mathbb{E}[\ell(f(\x;\vth),y)] = \int_{\mathcal{X}\times\mathcal{Y}} \ell(f(\x;\vth),y) \, dp(\x,y),
\end{equation}
where $\ell: \mathcal{Y} \times \mathcal{Y} \to \mathbb{R}$ is a \textit{loss function} that quantifies the discrepancy between predictions and true labels -- for example, the squared error $\mathcal{\ell}(y,y')=(y-y')^2$ for regression or the cross entropy for classification.

In practice, as the true data distribution $p(\x, y)$ is unknown, the \textit{empirical risk} over the training set (or \textit{training error}) is minimized instead:
\begin{equation}
    \hat{\vth} = {\arg \min}_{\vth} \, \hat{\testerr}(f) := {\arg \min}_{\vth} \left( \frac{1}{P} \sum_{\nu=1}^P \ell(f(\x_\nu;\vth),y_\nu) \right),
\end{equation}
yielding a learned predictor $\hat{f}=f(\cdot;\hat{\vth})$.

Optimizing \marginpar{Neural networks are trained with gradient descent.} the empirical risk in deep networks is a non-convex, high-dimensional problem without a closed-form solution. \textit{Gradient descent} and its variants are commonly used to perform the minimization, initializing parameters randomly, and updating them iteratively:
\begin{equation}
    \vth_{t+1}= \vth_t - \eta \nabla_{\vth} \, \hat{\testerr}(f),
\end{equation}
where $\eta$ is the learning rate. Gradients are efficiently computed via backpropagation, which applies the chain rule through the computational graph of the network.
Variants like stochastic gradient descent (SGD) -- where the loss is averaged over a small random subset (minibatch) at each step -- are widely used, together with further enhancements, such as momentum and adaptive learning rates leveraging curvature information, to improve convergence.

After training, the generalization performance of $\hat{f}$ is assessed on unseen test data.

While FCNs are general function approximators, they do not inherently exploit specific structure in the data. A major breakthrough came with \textit{convolutional neural networks} (CNNs), \marginpar{Convolutional networks encode locality and translational invariance.} which draw inspiration from the visual cortex and encode two key priors found in natural images: \textit{locality} and \textit{translational invariance}. CNNs apply local filters to input patches, capturing spatially localized patterns. These filters are shared across locations, allowing the network to detect features regardless of their position. \textit{Pooling layers} further coarse-grain internal representations by summarizing local regions, in a process akin to the renormalization group in statistical physics.

These architectural priors substantially reduce the number of parameters and enhance generalization on tasks with spatial structure, underpinning the deep learning revolution.

\subsection{Theoretical puzzles}

Despite the remarkable empirical success of deep neural networks, our theoretical understanding of how they learn high-dimensional tasks remains limited. A central question is: \textit{How many data points are needed for a model to learn a task with good precision?} That is, how does the generalization error $\testerr(P)$ decrease as the number of training examples $P$ increases? Remarkably, empirically, for many high-dimensional tasks, $\testerr(P)$ is well fitted by a power-law decay:
\begin{equation}
    \testerr(P) \sim P^{-\beta},
\end{equation}
where the exponent $\beta$ captures how efficiently a model learns from data \cite{hestness2017deep,kaplan2020scaling,hoffmann2022training}. These empirical \textit{neural scaling laws} show that $\beta$ depends on the dataset, the task, and the learning algorithm. General \marginpar{Curse of dimensionality: the sample complexity grows exponentially with the input dimension.} theoretical arguments would suggest that $\beta$ should be vanishing for large input dimension $d$, implying that learning would be practically impossible in such settings where the dimension is large, which is the case in practice (e.g., images where $d$ is the number of pixels and color channels). This is the essence of the \textit{curse of dimensionality}. In high-dimensional spaces, volume grows exponentially with $d$, and the typical distance $\delta$ between nearest-neighbor data points diminishes very slowly: $\delta \sim P^{-1/d}$. Consequently, only weak regularity assumptions on the task -- like regressing a $1$-Lipschitz function $f$, where $|f(\x) - f(\x')| \leq \|\x - \x'\|$ -- lead to $\beta \propto 1/d$ \cite{luxburg2004distance,wainwright2019high}. For large $d$, this means that the number of samples $P$ needed to generalize becomes astronomically large, even exceeding the number of atoms in the observable universe.

The empirical success of deep learning in high-dimensional tasks, despite the curse of dimensionality, highlights a fundamental puzzle. If high-dimensional data can be learned efficiently, they must possess strong underlying structure: symmetries, invariances to certain transformations, or other forms of regularity that make the problem tractable. This leads to several critical questions: \textit{What is the nature of this structure?} \textit{How does it quantitatively affect performance, in particular the exponent $\beta$?} \textit{And how do neural networks harvest this structure through architectural choices -- such as depth, convolution, or weight sharing -- that encode specific inductive biases?} Addressing these questions is among the most fundamental and practical problems in deep learning, as it directly impacts the sample requirements for achieving a given precision and, thus, influences model training and scaling.

A longstanding hypothesis attributes the success of deep networks to the \textit{compositionality} of data: the notion that objects are composed of parts, which in turn are composed of simpler sub-parts. Natural data is often \textit{hierarchical} in this sense. \marginpar{Images and language display a hierarchical and compositional structure.} For example, images contain edges, textures, and objects at different spatial scales; language exhibits a hierarchical grammatical structure, from words to phrases to full sentences; and biological sequences such as proteins show primary, secondary, and tertiary structural organization. Deep neural networks are believed to exploit this compositionality by learning layered, increasingly abstract representations. Post hoc analyses indeed reveal that neurons in trained networks respond to progressively more complex features \cite{Lecun15,zeiler_visualizing_2014,doimo2020hierarchical}, mirroring the hierarchical organization observed in the primate visual cortex \cite{van1983hierarchical,grill2004human}. Yet, a \textit{quantitative} understanding of how this hierarchical structure affects generalization performance remains elusive.

\section{Generalization in deep learning}
\label{sec:intro-gen}

\subsection{Classical statistical learning theory}

Classical statistical learning theory provides a fundamental framework for understanding generalization. Its primary aim is to quantify the relationship between the training error (performance on the training data) and the generalization error (expected performance on unseen data). This is typically achieved through bounds of the form: 
\begin{equation}
    \testerr(f) \leq \hat{\testerr}(f) + \frac{\mathcal{C}(\mathcal{H})}{P},
\end{equation}
where $\mathcal{H}$ denotes the \textit{hypothesis class}, representing the set of all function that the model can express (e.g., for neural networks, $\mathcal{H} = \{f(\cdot;\vth) \allowbreak \textrm{ for all } \vth \in \Theta\}$), and $\mathcal{C}(\mathcal{H})$ is a measure of the \textit{complexity} or \textit{capacity} of the hypothesis class.

Classical \marginpar{Classical learning theory predicts overfitting with large models...} theories thus suggest that a model's ability to generalize is closely tied to its capacity. For instance, the Vapnik-Chervonenkis (VC) dimension \cite{vapnik1999overview}, a common measure of $\mathcal{C}(\mathcal{H})$, typically grows with the number of parameters in a neural network. Rademacher complexity \cite{koltchinskii2000rademacher,bartlett2002rademacher} instead measures how well functions within the hypothesis class can fit random noise, essentially quantifying the model's ability to `memorize' $P$ random points. The underlying principle is that richer models, capable of fitting a broader family of functions, are more prone to \textit{overfitting} the training data, leading to poor performance on unseen examples. This framework implies a tradeoff: to achieve good generalization, one must carefully select a model complexity that is neither too low (leading to underfitting, or high training error) nor too high (leading to overfitting, or a large gap between training and generalization error). 

However, \marginpar{... but neural networks defy this.} the empirical success of modern deep neural networks challenges these traditional notions. In modern machine learning, networks routinely contain hundreds of millions, even billions, of trainable parameters -- a number far exceeding the number of training examples. By the previous reasoning, these models should generalize poorly. Yet, contrary to the warnings of classical theory, such highly \textit{overparameterized} networks do not necessarily overfit to the point of poor generalization. Instead, they frequently achieve zero training error, perfectly \textit{interpolating} the training data, while simultaneously generalizing exceptionally well to new examples. In fact, the deep neural networks used in practice, thanks to their immense capacity, can easily learn to classify random labels perfectly \cite{zhang2016understanding} and still perform well on structured data. This contradicts classical expectations and signals a breakdown of traditional theory in the overparameterized regime.

One striking manifestation of this is the \textit{double descent} phenomenon \cite{spigler2019jamming,belkin2019reconciling,nakkiran2019deep}: the generalization error first increases with model complexity -- as classical theory predicts -- but then, after crossing the interpolation threshold (where the model perfectly fits the data), it begins to decrease again. This reveals a second descent, unaccounted for by classical bounds. These observations have prompted the search for new theoretical tools to understand learning in highly overparameterized models.

\subsection{Infinite-width networks}

One such tool is the infinite-width limit of neural networks. As the number of neurons per layer grows, the network’s behavior simplifies and, in certain regimes, becomes analytically tractable.

\paragraph{Infinite-width at initialization}  
When the weights of a network $f$ as in \autoref{eq:intro-fcn} are initialized with i.i.d. Gaussian entries with zero mean and unit variance, taking the limit of all hidden layers to infinity makes the network behave as a sample from a centered \textit{Gaussian random function} with a covariance that can be computed recursively \cite{Neal1996,williams2006gaussian,lee2017deep}. This was extended to convolutional architectures in the same limit, leading to a covariance that depends on the specific architecture \cite{novak2018bayesian}.

\paragraph{Learning in the infinite-width regime}
When \marginpar{NTK infinite-width limit linearizes training dynamics.} trained with gradient descent, very wide networks can reach a global minimum while keeping the parameters very close to their random initialization $\vth_0$. In other words, in this \textit{lazy regime}, the network remains close to its linearization around initialization throughout training \cite{jacot2018neural,du2018gradient,lee2019wide,arora2019exact,chizat2019lazy}:
\begin{equation}
    f(\x;\vth) \approx f_{\textrm{lin}}(\x;\vth) = f(\x;\vth_0) + (\vth - \vth_0)^\top \nabla_{\vth} \, f(\x;\vth_0).
\end{equation}
Learning is thus equivalent to a \textit{kernel method}\footnote{Kernel methods are algorithms that, given a \textit{kernel} function $\mathcal{K}(\x,\x')$ -- a similarity measure between inputs $\x$ and $\x'$ -- learn a predictor of the form $f(\x)=\sum_{\nu=1}^P a_\nu \mathcal{K}(\x_\nu,\x)$ by learning coefficients $\{a_\nu\}_{\nu\in[P]}$ that fit the training data. We will give more background on kernels in \autoref{ch:locality}.} with a deterministic kernel known as the \textit{Neural Tangent Kernel} (NTK):

\begin{equation}
    \mathcal{{K}_{\textrm{NTK}}}(\x,\x') = \lim_{H \to + \infty} \nabla_{\vth} \, f(\x;\vth_0)^\top \nabla_{\vth} \, f(\x';\vth_0)
\end{equation}
For instance, for a two-layer ReLU network with normalized inputs, the NTK can be computed in closed form. If $t=\x^\top \x'$, then \cite{bietti2019inductive}:
\begin{equation}
    \mathcal{{K}_{\textrm{NTK}}}(\x,\x') = t \ \frac{\pi - \arccos(t)}{2\pi} + \frac{(\pi - \arccos t) \, t + \sqrt{1-t^2}}{2\pi}.
\end{equation}
Intuitively, in this regime, very small changes of parameters can interfere positively, changing the output function by $\mathcal{O}(1)$ -- which is sufficient for learning, but not for changing the Jacobian. This means that the model effectively learns by combining a fixed set of features defined at initialization.

Given \marginpar{The performance of kernels is determined by their spectra.} a kernel $\mathcal{K}$, generalization is governed by the spectral decomposition of the integral operator $T_{\mathcal{K}}$, defined as
\begin{equation}
    (T_{\mathcal{K}} f)(\x) = \int \mathcal{K}(\x,\x') f(\x') dp(\x'),
\end{equation}
with eigenfunctions $\phi_\rho$ and eigenvalues $\lambda_\rho$ ($T_{\mathcal{K}} \phi_\rho = \lambda_\rho \phi_\rho$) \cite{caponnetto2007optimal}. Gradient descent first learns components of the target function aligned with eigenfunctions corresponding to larger eigenvalues -- a phenomenon known as \textit{spectral bias}. Given $P$ examples, only the components aligned with the top $\mathcal{O}(P)$ eigenfunctions are effectively learned \cite{bordelon2020spectrum,spigler2020asymptotic,jacot2020kernel}.

Work by \citet{bietti2019inductive} has analyzed the spectrum of the NTK for two-layer fully-connected networks using spherical harmonics, showing that eigenvalues decay with the frequency of the corresponding eigenfunctions. This explains why such networks exhibit a preference for learning low-frequency (smooth) functions. Moreover, the same spectral properties persist in deeper fully-connected networks, suggesting that, in this regime, depth alone does not provide any benefits to generalization \cite{bietti2021deep}.

\citet{arora2019exact,arora2020harnessing} extended the NTK to convolutional networks, deriving recursive formulas for the kernels and empirically demonstrating their good performance on image data, reflecting the architectural priors of CNNs.

\paragraph{Beyond the kernel regime}
The \marginpar{Feature/representation learning: weights evolve significantly.} kernel regime provides a tractable theoretical framework for analyzing generalization in overparameterized neural networks. However, a key limitation of this regime is the absence of \textit{feature} or \textit{representation learning}: parameters remain close to their random initialization, and the network behaves effectively as a fixed feature extractor.

To address this, an alternative infinite-width scaling has been proposed, in which the network output is rescaled by a factor $H^{-1}$ as opposed to
$H^{-1/2}$. This modification leads to fundamentally different training dynamics: to achieve $\mathcal{O}(1)$ outputs, weights must evolve substantially during training. As a result, neurons adapt to the structure of the data, and the network learns features -- a phenomenon entirely absent in the NTK limit. This setting is known as the \textit{feature learning regime}, or alternatively, the \textit{mean-field limit}.

The dynamics in this regime is no longer linear. Instead, it can be described in terms of a time-evolving density over parameters, yielding a hydrodynamic description analogous to interacting particle systems under a potential. This feature-learning regime captures richer learning behavior, including data-dependent representation learning. Crucially, it offers a path toward understanding how networks can overcome the curse of dimensionality not merely through architectural priors, but by actively `discovering' structure from data.

Theoretical progress in this direction has been primarily limited to two-layer networks \cite{mei2018mean,rotskoff2018neural,Chizat2018,sirignano2020mean}. Recent work has extended this framework to deeper architectures \cite{nguyen2019mean}. Yet, generalization in this regime remains difficult to analyze and typically requires stronger, data-dependent assumptions.

\subsection{Role of data structure in high-dimensional learning}

As discussed in \autoref{sec:intro-dl}, learning in high dimensions is, in general, statistically intractable. Under minimal regularity assumptions -- such as Lipschitz continuity -- the generalization error decays only as $\testerr(P) \sim P^{-1/d}$ \cite{luxburg2004distance,wainwright2019high}, which rapidly becomes prohibitive as the ambient dimension $d$ grows. Nevertheless, modern neural networks are able to learn high-dimensional tasks with remarkable efficiency. This apparent paradox suggests that real-world data distributions are far from generic and must possess rich internal structure.

\paragraph{Global smoothness} A classical structural assumption is that the function $f$ to be learned belongs to a Sobolev space -- that is, it has $m$ square-integrable derivatives. In this case, generalization rates scale as $\testerr(P) \sim P^{-m/d}$ (e.g., \cite{bach2021learning}). However, this scaling only overcomes the curse of dimensionality when $m \propto d$, a condition that is implausible in practice.

\paragraph{Manifold Hypothesis} An alternative assumption posits that data lie on a $d_{\mathcal{M}}$-dimensional manifold embedded in $\mathbb{R}^d$, with $d_{\mathcal{M}} \ll d$. In this case, the sample complexity depends on $d_{\mathcal{M}}$ rather than $d$, e.g., \cite{kpotufe2011k,spigler2020asymptotic,hamm2021adaptive}. However, empirical studies show that even the \textit{intrinsic dimension} $d_{\mathcal{M}}$ can remain large in practice -- routinely in the tens or hundreds for visual data \cite{pope_intrinsic_2021}. Thus, the manifold hypothesis alone does not fully explain the efficiency of deep learning.

Moreover, both smoothness and low intrinsic dimension can already be effectively exploited by isotropic kernel methods, which lack representation learning capabilities. The fact that such methods often strongly underperform deep neural networks on real-world tasks suggests that the structure captured by classical assumptions is insufficient to capture the full richness of the data exploited by deep models.

\paragraph{Low-dimensional projections} Another approach models the target function as $f(\x) = g(A\x)$, depending only on low-dimensional projections of the input, with $A \in \mathbb{R}^{k \times d}$ and $k \ll d$. In this setting, two-layer neural networks operating in the feature learning regime can identify the latent subspace and adapt to such low-dimensional structure, e.g., \cite{bach2017breaking,arous2021online,paccolat2021compression,bietti2022learning,dandi2023two}. Nonetheless, these models fail to account for essential phenomena observed in practice, such as the benefits of depth and the emergence of hierarchical representations.

\paragraph{Beyond low-rank: Real-world structure} 
Real signals exhibit a more complex structure. A key example is \textit{equivariance} or \textit{invariance} with respect to certain group actions. Translational symmetry in images is a canonical example: if the pixels of a cat shift by a few locations, it is still a cat. Convolutional neural networks hard-wire this property through weight sharing, thereby eliminating redundant degrees of freedom. 

More subtle forms of invariance, such as stability under smooth deformations, e.g., diffeomorphisms, have also been proposed as mechanisms by which networks can effectively reduce the data dimension \cite{bruna2013invariant,mallat2016understanding}. Recent work, mostly in the kernel regime, quantifies how invariances influence sample complexity. For example, \citet{mei2021learning} show that translation-invariant kernels yield modest gains in generalization. More generally, \citet{bietti2021sample} prove that invariant kernels over symmetry groups can improve sample complexity by a factor equal to the group size -- a non-negligible but generally insufficient gain for overcoming the curse of dimensionality. However, in some cases, such as permutations or local translations (which approximate deformations), this gain can be exponential, hinting at the importance of deformation stability in overcoming the curse for tasks like image recognition.

Another pervasive structural property is \textit{spatial locality}. In many modalities, including vision and language, correlations decay with distance: the dominant interactions are local. This implies that, for some tasks, it is enough to focus on small neighborhoods (e.g., image patches or short n-grams in language). For instance, CNNs exploit this by restricting the receptive field of neurons to local regions, enabling efficient extraction of salient patterns. Furthermore, when long-range dependencies are present, they often organize hierarchically across multiple scales: from edges to textures to objects in images, or from characters to words to phrases in text. This \textit{hierarchical compositionality} introduces a natural notion of scale separation: fine-grained features interact locally, while coarser features emerge from aggregations over broader contexts. Coarsening operations like pooling in CNNs mimic renormalization-group transformations in physics, progressively integrating information across scales. 

In \marginpar{Hierarchical compositional functions are easy to approximate with deep networks.} fact, deep neural networks -- by virtue of their depth -- are naturally suited to modeling such hierarchical functions. Each layer operates at a distinct scale, composing simpler features into increasingly abstract ones. Theoretical results support this intuition: compositional functions, e.g.,
\begin{equation}
    f(x_1,x_2,x_3,x_4) = g(h_{1}(x_1,x_2),\,h_2(x_3,x_4)),
\end{equation}
can be represented by deep networks with exponentially fewer parameters than shallow ones \cite{mhaskar2016learning}. This implies an information-theoretic lower bound on the sample complexity that is only polynomial in input dimension \cite{schmidt2020nonparametric}. However, these theoretical results do not imply that gradient descent will efficiently discover such solutions in practice. In fact, efficient representability does not guarantee learnability by
gradient descent for hierarchical tasks \cite{cagnetta2023deep}.

\subsection{Questions}

Despite the empirical success of convolutional networks, our theoretical understanding of why they outperform fully connected networks and how different architectural components contribute to this advantage remains incomplete. A central challenge in machine learning theory is thus to quantify how different forms of structure -- both in the model and in the data -- affect sample complexity.

\textit{How much does locality contribute to generalization? Can locality alone change the learning curve exponent $\beta$, allowing convolutional neural networks to escape the curse of dimensionality? How do these gains compare to those induced by other structural priors, such as translational invariance and weight sharing?}

In the NTK regime, FCNs are biased toward low-frequency target functions. \textit{What is the spectral bias of convolutional kernels? How do locality and weight sharing reshape the kernel spectrum, and what classes of functions become easier (or harder) to learn as a result? Does it privilege certain classes of functions -- e.g., those with localized or multiscale structure?}

For FCNs in the NTK limit, increasing depth is known to have no impact. Yet in practice, deep CNNs substantially outperform shallow ones. \textit{Does depth lead to improved generalization in wide CNNs as well? Is this advantage attributable to the presence of hierarchical receptive fields, which mirror the compositional organization of many real-world signals?} More broadly, \textit{what is the sample complexity of learning hierarchical or compositional target functions with gradient-based methods?}

Beyond the fixed biases of CNNs, a deeper question is whether networks can learn useful structure from data -- such as locality, hierarchy, or invariance -- without it being hard-wired into the architecture. \textit{What kinds of data structures allow neural networks to overcome the curse of dimensionality through feature learning, and how does this manifest in practice? Can models automatically discover modular or compositional patterns in their inputs, and if so, how does this affect their generalization efficiency?}

We will address such questions in Parts II and III of this thesis.

\section{Deep generative modeling}
\label{sec:intro-generative}

\subsection{Unsupervised learning}

The \marginpar{Unsupervised learning learns structure without labels.} goal of unsupervised learning is to uncover the underlying structure of data without access to labels or supervision. This is particularly relevant in \textit{generative modeling}, where we aim to learn a model that captures the probability distribution of natural data and can generate new, realistic samples from it.

Real-world data distributions are complex and usually unknown. A common approach is to approximate the true data distribution $p_{\mathrm{data}}(\x)$ with a parameterized model $p_{\vth}(\x)$, typically a deep neural network. The parameters $\vth$ are learned by maximizing the log-likelihood of the data:
\begin{equation}
    \int p_{\mathrm{data}}(\x) \log p_{\vth}(\x)d\x.
\end{equation}
In practice, we approximate $p_{\mathrm{data}}$ by the empirical distribution over a dataset, $\hat{p}_{\mathrm{data}}(\x) = P^{-1} \sum_{\nu=1}^P \delta(\x-\x_\nu)$, yielding a training objective that seeks to assign high probability to observed samples. Once trained, such a model can be used to sample new data.

\subsection{Score-based diffusion models}

A breakthrough in generative modeling comes from score-based diffusion models, which draw inspiration from non-equilibrium statistical physics \cite{sohl2015deep,ho2020denoising,song2019generative,song2020score}.

These \marginpar{Diffusion models generate data by reversing a stochastic process.} models define a \textit{forward process} that progressively corrupts data, transforming a complex, unknown distribution into a simple, known one by adding noise. For instance, in the case of images, this can be implemented as an independent random walk per pixel, gradually turning the image into noise. A \textit{backward process} is then defined as the reversal of this trajectory: starting from pure noise, the model learns to reverse the flow of time and recover naturalistic data.

Formally, the diffusion process is defined by a stochastic differential equation (SDE):
\begin{equation}
    d\x = f(\x,t)dt+g(t)d\w,
\end{equation}
where $\w$ is a Brownian motion (Wiener process), $f(\cdot,t):\mathbb{R}^d \to \mathbb{R}^d$ is the \textit{drift}, and $g(\cdot)$ the \textit{diffusion} coefficient. Denoting with $p_t$ the probability density of $\x(t)$, we define the marginal $p_0:=p_{\textrm{data}}$ as the clean data distribution and $p_T$ as the terminal (or prior) distribution, which is simple and easy to sample from.

To generate data, we reverse this process. The reverse-time SDE, derived from the Fokker–Planck equation \cite{anderson1982reverse}, reads:
\begin{equation}
    d\x = [f(\x,t) - g(t)^2 \nabla_{\x} \log p_t(\x)]dt+g(t)d\overline{\w},
\end{equation}
where $dt$ an infinitesimal negative time-step, $\overline{\w}$ is a Brownian motion running backward in time, and
$\nabla_{\x} \log p_t(\x)$ is the so-called \textit{score function}. Thus, learning to sample from the data distribution reduces to estimating the score at each time.

The \marginpar{Learning the score reduces generative modeling to regression.} score is learned by training a neural network $s_{\vth}(\x,t)$ via score matching \cite{hyvarinen2009estimation}, i.e., minimizing the loss 
\begin{equation}
    \mathbb{E}_t \mathbb{E}_{\x(0)}\mathbb{E}_{\x(t)|\x(0)} \left[ \| s_{\vth}(\x(t),t) - \nabla_{\x} \log p_t(\x(t) | \x(0)) \|^2 \right].
\end{equation}
For an affine drift $f(\cdot,t)$, the transition kernels are Gaussian, allowing for analytical expressions of the score.

\paragraph{Denoising diffusion probabilistic models (DDPMs)}

In practice, the continuous-time SDE is discretized. In DDPMs \cite{ho2020denoising} -- defined by $f(\x,t){:=}-\frac{1}{2}\beta(t)\x$ and $g(t){:=}\sqrt{\beta(t)}$) -- the forward process becomes:
\begin{equation}
    \x_t = \sqrt{1-\beta_t} \x_{t-1} + \sqrt{\beta_t} \mathbf{z}_{t-1}, \quad \z_{t-1} \sim \mathcal{N}(\mathbf{0},\mathbf{I}),
\end{equation}
with a noise schedule $\{\beta_t\}_{t=1}^T$ that typically increases linearly with time or follows a cosine law. The backward process reverses this trajectory using a learned score network.
Notice that the conditional score at each step is given by:
\begin{equation}
    \nabla_{\x} \log p_t(\x_t | \x_0) = - \frac{x_t - \sqrt{\overline{\alpha}_t}\x_0}{1-\overline{\alpha}_t}, \textrm{ with } \overline{\alpha}_t = \prod_{t'=1}^t (1-\beta_{t'}),
\end{equation}
and averaging over the posterior distribution, the score reads:
\begin{equation}
    \nabla_{\x} \log p_t(\x_t) = \mathbb{E}_{\x_0|\x_t}[\nabla_{\x} \log p_t(\x_t | \x_0)] = - \frac{x_t - \sqrt{\overline{\alpha}_t}\mathbb{E}_{\x_0|\x_t}[\x_0]}{1-\overline{\alpha}_t}.
\end{equation}
Thus, one can train a \textit{denoising network} to directly predict $\mathbb{E}[\x_0|\x_t] := \mathbb{E}_{\x_0|\x_t}[\x_0]$.

\paragraph{Architectures}
A common architectural choice for the score network is the U-Net \cite{ronneberger2015u}, a convolutional neural network featuring symmetric downsampling and upsampling paths, each comprising multiple resolution blocks. Skip connections link blocks that operate at the same resolution, helping to retain fine-grained details that might otherwise be lost during the downsampling process. More recently, transformer-based architectures have also been introduced as alternative backbones for diffusion models \cite{peebles2023scalable}.

\paragraph{Discrete diffusion}
To handle discrete data \marginpar{Diffusion models can be extended to text and other discrete modalities.} -- such as text or molecular graphs -- diffusion models can be generalized to discrete spaces via Markov jump processes governed by time-varying transition matrices \cite{hoogeboom2021argmax,d3pm2021}. These processes corrupt data either by randomly flipping coordinates to uniformly random ones (\textit{uniform diffusion}) or by masking them (\textit{absorbing diffusion}).\footnote{We will cover these processes in more detail in Part III.} The reverse processes require estimating the conditional expectation $\mathbb{E}[\x_0|\x_t]$ to reconstruct the original signal \cite{d3pm2021}, similarly to the continuous case. At the time of writing this thesis, \textit{diffusion large language models} are beginning to achieve competitive performance, particularly on code generation tasks, and the first commercial implementations are being released.

\subsection{Questions}

As discussed above, sampling via time-reversal of a diffusion process amounts to regressing the score function $\nabla_{\x}\log p_t(\x)$. Hence, in the worst case, the sample complexity of this task would still explode exponentially with the data dimension. Nevertheless, modern diffusion models learn to synthesize high-resolution images and long passages of text from finite datasets. Existing generalization guarantees hinge on global smoothness assumptions that rarely hold in practice \cite{oko2023diffusion}. A central open question is therefore: \textit{What latent structure in $p_{\mathrm{data}}$ enables efficient score learning?}

A compelling hypothesis is that diffusion models exploit the hierarchical and compositional structure of data: these models might learn a `library' of reusable parts and the rules for composing them -- allowing for the generation of novel samples. This raises several important questions: \textit{Do diffusion models synthesize novel data based on compositional principles? If so, how many training samples are needed to learn such compositional rules? What role do architectural choices, such as the U-Net, play?}

Furthermore, to what extent do diffusion models generalize beyond their training data versus simply memorizing it, functioning as ‘stochastic parrots’? In theory, perfect minimization of the training objective would lead models to replicate the empirical distribution exactly, i.e., to memorize the training data. Indeed, recent work has demonstrated that large, overparameterized models can and often do memorize \cite{somepalli2022diffusion,carlini2023extracting,yoon2023diffusion,kadkhodaie2023learning}. \textit{How, then, do diffusion models avoid memorization in practice? Is this due to architectural or optimization-related inductive biases, or merely a consequence of underparameterization? Can overparameterized diffusion models still generalize?}

We will address these questions in Part III.

\section{Deep learning today}
\label{sec:intro-today}

Recent \marginpar{From task-specific models to general-purpose systems.} years have witnessed a qualitative transformation in how deep neural networks are trained and deployed. What began as a collection of supervised learning pipelines has evolved into a paradigm dominated by colossal, \textit{self-supervised} models capable of acquiring general-purpose abstractions with minimal human supervision.

\subsection{Pretraining, scaling, and emergence}

The advent of large-scale \textit{self-supervised learning} -- encompassing contrastive learning, masked prediction, and next‑token language modeling \cite{simclr,radford2021learning,devlin-etal-2019-bert} -- has fundamentally reshaped the deep learning landscape. These techniques allow models to learn rich and transferable representations from massive corpora of unlabeled data, enabling them to generalize across a broad range of tasks. This evolution has been catalyzed by the discovery of \textit{neural scaling laws} \cite{kaplan2020scaling,henighan2020scaling}, which show that model performance improves predictably with increased model capacity and training data.

Importantly, these gains are not merely quantitative. As models are scaled, qualitatively novel behaviors emerge. One prominent example is \textit{zero-shot generalization}, \marginpar{Zero-shot: solve tasks without training.} where models can solve previously unseen tasks without any parameter updates. In many cases, these models require only a prompt or a few examples at inference time, an ability known as \textit{in-context learning} \cite{brown2020language} \marginpar{In-context learning: learn from prompts.}. Techniques such as \textit{instruction tuning} \cite{wei2021finetuned} further enhance this capability by aligning model behavior with natural language task descriptions, making task adaptation more robust and reliable. These developments suggest that large models go beyond learning data patterns; they appear to internalize abstract representations of \textit{tasks} themselves.

On the architectural front, these advances are often accompanied by a departure from strong, domain-specific inductive biases. For instance, \textit{vision transformers} (ViTs) \cite{dosovitskiy2021an} replace convolutional layers with \textit{attention layers} \cite{vaswani2017attention}, yet still match or exceed the performance of CNNs after pretraining -- despite being less structurally constrained. This trend toward more general-purpose architectures, capable of handling multiple modalities, hints at an underlying structural universality across seemingly disparate data domains.

\subsection{Model editing and composition}

Despite the power of large pre-trained models, their learned representations are typically static. Adapting these models to new tasks \cite{zhuang2020comprehensive,ilharco2022patching,ilharco2023task}, preferences \cite{ouyang2022training,lu2022quark,ribeiro2022adaptive,sparrow}, or robustness requirements \cite{wortsman2021robust,shibani2021editing,ortizjimenez2021optimism} has traditionally relied on expensive strategies such as joint fine-tuning \cite{zhuang2020comprehensive}, \textit{reinforcement learning with human feedback} (RLHF) \cite{ouyang2022training}, or iterative prompt engineering. Beyond the costs, these methods risk \textit{catastrophic forgetting}, where performance on previously learned tasks deteriorates \cite{mccloskey1989catastrophic,french1999catastrophic,wortsman2021robust}.

Recent \marginpar{Model editing: modify behavior without full retraining.} research has explored more scalable and efficient alternatives based on \textit{model editing} and composition. Rather than retraining from scratch, these methods manipulate the weights of these models directly to induce new behaviors while preserving existing capabilities (e.g.,  \cite{ilharco2023task,ainsworth2022git,ilharco2022patching,wortsman2021robust,wortsman2022model,li2022branch,matena2021merging,frankle2020linear}). A particularly intriguing discovery is the presence of compositional structure in weight space. The differences between pretrained and fine-tuned
weights -- so-called \textit{task vectors} -- can be algebraically combined \cite{ilharco2023task}. For example, \marginpar{Additive structure enables model composition.} adding task vectors from two fine-tuned models to a shared base model can yield a new multitask model with combined functionality. Similarly, subtracting a task vector can effectively cause the model to `forget' a specific capability.

This observation suggests that models may represent tasks and skills in a way that supports composition, much like word embeddings encode semantic relationships via vector arithmetic.

\subsection{Questions}

These findings raise important questions that resonate with the broader themes in this thesis. \textit{If tasks can be composed algebraically in model space, what is the structure that underlies this ability? Is it merely a consequence of large models operating in a near-linear regime, like the aforementioned NTK limit? How are distinct skills represented and isolated within the vast parameter space to avoid destructive interference when combined?}

A possibility is that task compositionality is a consequence of modularity: large models may implicitly learn distinct modules or subsystems that can be reused and recombined across tasks. This raises further questions about the origins of such a structure. \textit{Is this modularity an inherent property of certain architectures, or is it an emergent phenomenon that arises from the pre-training process itself? Under what conditions does task composition emerge, and what role do model scale and the pre-training objective play?}

Such a modular organization would provide a potential antidote to the curse of dimensionality. Rather than requiring exponentially more data to cover every possible scenario, models could generate novel behaviors by composing a finite set of reusable components. Understanding this mechanism is therefore crucial for building more efficient and editable models.

We will study task compositionality and propose answers to these questions in Part IV.

\section{Thesis structure and main results}

The remainder of this thesis is organized into three parts, each addressing a set of questions raised in \autoref{sec:intro-gen}--\ref{sec:intro-today}. Together, they demonstrate how \textit{locality} and \textit{compositionality} serve as a unifying thread, extending from data space to tasks in model parameter space.

\subsection{Part II: Statistical mechanics of convolutional networks at infinite width}

\paragraph{Locality defeats the curse of dimensionality}
\autoref{ch:locality}\footnote{This chapter is a revised version of material first presented in \cite{favero2022locality,favero2021locality}.} investigates how architectural priors -- specifically, \textit{locality} and \textit{translational invariance} -- shape the generalization performance of CNNs in the lazy training regime. We approach this problem through a \textit{teacher-student setting} for kernels. The target function is modeled as a Gaussian random field with covariance $\KT$ -- the teacher kernel that generates the data. The learning process then involves a (possibly mismatched) student kernel $\KS$ that regresses the target function. Both models are convolutional kernels inspired by the NTK derived from one-hidden-layer CNNs with a given filter size.

By applying recent results from the replica method in statistical physics, we show that locality -- and not translational invariance -- is the key factor governing the \textit{learning curve exponent} $\beta$, which characterizes how generalization improves with sample size. In particular, when the teacher's filter size $t$ is smaller than the student's $s$, $\beta$ depends only on $s$ and not on the input dimension $d$. This implies that, even in high-dimensional settings, efficient learning is possible when the target function can be decomposed into a sum of local components, provided the regression is performed using a kernel that captures this compositional structure.

\paragraph{The role of depth and spatial adaptivity}
While the above setting captures spatial locality, it omits another crucial property of real-world data: hierarchical compositionality. It is natural to expect that depth, when combined with locality, provides a powerful inductive bias for such structural patterns. In \autoref{ch:deepcnn}\footnote{This chapter is a revised version of material first presented in \cite{cagnetta2024can,cagnetta2023can}.}, we extend our framework to \textit{deep} CNNs in the kernel regime.

Our first result shows that the kernel spectrum associated with these networks mirrors the multi-scale architecture of the model itself. We characterize the asymptotic behavior of the spectrum and, building on this, use generalization bounds to demonstrate that deep CNNs can \textit{adapt} to the spatial scale of the target function. Specifically, when the target depends only on low-dimensional, spatially localized subsets of the input, the rate at which generalization error decreases is governed by the effective dimensionality of these subsets, thus beating the curse. In contrast, if the target function depends on the entire input, the decay rate is limited by the full input dimension.

Crucially, our results imply that data involving long-range nonlinear dependencies are not efficiently learnable by deep CNNs in the lazy training regime. Surprisingly, even in a teacher-student setup where both the teacher and the student are deep CNNs with matched topology, we find that the sample complexity grows exponentially with the input dimension -- despite the setting being Bayes-optimal. This calls for new synthetic models of hierarchical tasks and points to the necessity of moving beyond the kernel regime and towards feature learning to understand the benefits of hierarchical learning. 

\vspace{1em}

\noindent Taken together, these results provide concrete, quantitative answers to questions posed in  \autoref{sec:intro-gen}. In particular, they clarify how and when locality and depth act as crucial inductive biases underlying the empirical scaling laws observed in CNNs.

To address the limitations identified in the kernel regime, our subsequent work \cite{cagnetta2023deep} -- not included in this thesis -- introduces the \textit{Random Hierarchy Model} (RHM), a synthetic data framework consisting in an ensemble of \textit{probabilistic context-free grammars} (PCFGs).\footnote{PCFGs are tree-structured probabilistic graphical models, used to model the hierarchical structure in both language and images.}  In the same work \cite{cagnetta2023deep}, we showed that deep neural networks operating in the feature learning regime learn to classify such data -- i.e., infer the root of the hierarchical structure from the leaves -- with sample complexity scaling polynomially (rather then exponentially) in the input dimension. 

In Part III, we provide empirical evidence that this model captures key, non-trivial properties of real data, and that generative diffusion models are capable of leveraging its latent hierarchical structure to learn to generate strings respecting the rules of the grammar efficiently.

\subsection{Part III: Statistical mechanics of diffusion models}

\paragraph{A phase transition in the diffusion process}
\autoref{ch:phasetransition}\footnote{This chapter is a revised version of material first presented in \cite{sclocchi2024phase}.} shifts the focus from supervised learning to generative modeling and asks:  do \textit{diffusion models} capture compositional and hierarchical structure in data? Using the Random Hierarchy Model as a synthetic model of data, we develop a \textit{theory of composition}. We demonstrate that
for this data, the Bayes optimal denoising can be described exactly using belief
propagation.\footnote{Notice that, interestingly, the structure of U-Net architectures with the skip connections between the downsampling and upsampling paths mimics the upward and downward iterations of belief propagation.} We analyze the backward diffusion process acting after a time $t$ and uncover a phase transition at a critical time: beyond this point, the probability of accurately reconstructing high-level features, such as the class of an image, sharply drops. In contrast, the reconstruction of low-level features, such as fine-grained image details, evolves smoothly throughout the entire process. Hence, beyond the transition, even if the class has changed, the generated sample may still be \textit{composed} of low-level elements of the initial datum. 

Numerical experiments with pre-trained vision diffusion models confirm these theoretical predictions. This shows that diffusion models naturally act as `compositional samplers', building new data from known, reusable parts. Moreover, it puts forward synthetic hierarchical generative models as valuable theoretical tools for capturing non-trivial real-world data properties. 

\paragraph{Probing hidden hierarchies in data}
\autoref{ch:probing}\footnote{This chapter is a revised version of material first presented in \cite{sclocchi2024probing}.} explores whether diffusion models can also serve as tools for discovering the latent structure in data, a longstanding challenge. Using the same forward-backward paradigm of the previous chapter, we show that changes introduced by denoising occur in correlated chunks, with a correlation length that diverges at the phase transition identified earlier. We find that this behavior is consistent across real-world datasets, including text and images, suggesting that diffusion models can function as empirical probes of hierarchical organization in natural domains.

\paragraph{A theory of creativity and compositionality}
In the previous chapters, we established the presence of compositional effects in the generative process of pre-trained diffusion models. \autoref{ch:creativity}\footnote{This chapter is a revised version of material first presented in \cite{favero2025compositional}.} addresses how many training samples are needed for a model to learn composition rules that enable it to generate novel outputs.

To answer this, we consider the hierarchical grammars introduced earlier and find that learning the composition rules in the feature learning regime requires the same sample complexity as clustering features that share statistically similar contexts. This process unfolds hierarchically: identifying higher-level features, which correspond to longer context dependencies, requires more data. 
Importantly, the number of samples needed scales polynomially with the size of the context, allowing the model to learn a high-dimensional distribution without encountering the curse of dimensionality.
The key mechanism behind this result is the model's ability to construct a lower-dimensional internal representation of the grammar by recovering the latent variables -- an ability that fundamentally depends on feature learning.

We predict that diffusion models trained on limited data will generate outputs that are locally coherent -- i.e., respecting local composition rules -- but lack global consistency. These predictions are confirmed experimentally across both text and image domains: as training progresses or more data becomes available, the generated content displays increasingly long-range coherence. We conclude by drawing a conceptual parallel between this hierarchical clustering mechanism and the renormalization group from theoretical physics.

\paragraph{A race between memorization and generalization}
\autoref{ch:memorization}\footnote{This chapter is a revised version of material first presented in \cite{favero2025bigger}.} concludes Part III by investigating when and how diffusion models memorize training data. Theoretically, a diffusion model that perfectly minimizes its training loss would simply reproduce samples from its training set, i.e., \textit{memorize}. In practice, this is empirically observed in the overparameterized regime. We revisit this perspective by demonstrating that, even in highly overparameterized diffusion models, generalization occurs before memorization sets in. Through experiments spanning both image and language diffusion models, we consistently observe that memorization emerges only after a phase of generalization, and that the memorization time scales linearly with dataset size. In other words, generalization versus memorization should be understood as a competition between time scales. 

To investigate this dynamics more precisely, we study diffusion models trained on our hierarchical grammar models, where generalization corresponds to the hierarchical learning of increasingly deep grammar rules over time -- as discussed in the previous chapter. In this setting, the cost of early stopping -- which halts training before memorization sets in -- can be quantified. This allows us to construct a phase diagram that characterizes the dynamical transition between generalization and memorization.

\vspace{1em}

\noindent Collectively, these chapters answer the questions posed in \autoref{sec:intro-generative}. They demonstrate that diffusion models exploit compositionality to generate and learn hierarchical data efficiently, eventually becoming \textit{creative}.

\subsection{Part IV: Task localization and weight disentanglement}

\paragraph{Task compositionality in weight space}

The final part of the thesis, presented in \autoref{ch:taskarithmetic}\footnote{This chapter is a revised version of material first presented in \cite{ortiz2023task}.} studies the mechanisms through which \textit{tasks} can be composed. Recent empirical work has shown that \textit{task vectors} -- differences in weights between fine-tuned and base models -- can be added or subtracted to induce new behaviors or remove specific task capabilities, enabling multi-task performance or selective forgetting. But why does this arithmetic work?

We begin by examining the hypothesis that this phenomenon is a consequence of such large models operating, at least approximately, in the NTK regime, where the output function is linear in the weights. Interestingly, our results show that the NTK alone cannot fully explain task arithmetic.

Instead, we identify \textit{weight disentanglement} as the key mechanism: in pre-trained models, different directions in weight space correspond to distinct, localized changes in the network’s function over the input space. This structure allows task-specific behaviors to be composed without destructive interference.

We demonstrate that linearizing models further amplifies this disentanglement. Building on these insights, we provide both theoretical and empirical analyses of the NTK of these models, uncovering a connection between weight disentanglement and the spatial localization of NTK eigenfunctions. Crucially, we find that this structure is not present at initialization but emerges during pre-training.

\vspace{1em}

\noindent Altogether, this work offers a deeper understanding of the mechanisms behind tasks and models composition, showing that compositionality emerges in weight space -- enabling efficient reuse and editing of capabilities. This provides a mechanistic answer to the questions posed in \autoref{sec:intro-today}.

\cleardoublepage

\ctparttext{\bigskip \bigskip \begin{flushright}{\slshape
The whole is greater than the sum of its parts.} \\ \medskip
--- Aristotle
\end{flushright}
}

\part{Statistical Mechanics of Convolutional Networks at Infinite Width}

\chapter{Locality Defeats the Curse of Dimensionality} 

\label{ch:locality}

\begingroup
\renewcommand{\thefootnote}{}
\footnote{Parts of this chapter have been previously published in:\\
\textit{Favero*, A.}, Cagnetta*, F. and Wyart, M., 2022. Locality defeats the curse of dimensionality in convolutional teacher–student scenarios. Journal of Statistical Mechanics: Theory and Experiment, 2022(11), p.114012. \\
\textit{Favero*, A.}, Cagnetta*, F. and Wyart, M., 2021. Locality defeats the curse of dimensionality in convolutional teacher-student scenarios. In Advances in Neural Information Processing Systems (NeurIPS), 34, pp.9456-9467.\\
* These authors contributed equally.}
\addtocounter{footnote}{-1}
\endgroup

Deep Convolutional Neural Networks (CNNs) have emerged as the driving force behind many recent developments in deep learning, yet such success is surprising. In principle, supervised learning models face the curse of dimensionality: under minimal assumptions about the function being learned, reaching a fixed target generalization error $\testerr$ requires a number of training samples $P$ that grows exponentially with the dimensionality $d$ of the input data \cite{wainwright2019high,luxburg2004distance}, i.e., $\testerr(P) \sim P^{-1/d}$. However, empirical observations show that CNNs consistently overcome this limitation in practice \cite{hestness2017deep,krizhevsky2012imagenet}, exhibiting instead:
\begin{equation}\label{eq:loc-learning-scaling}
\testerr(P) \sim P^{-\beta}, \quad \textrm{with } \beta\,{\gg}\,1/d.
\end{equation}
In particular, CNNs achieve remarkable performances on high-dimensional tasks, such as ImageNet image classification, with state-of-the-art architectures achieving exponents $\beta\approx [0.3, 0.5]$ \cite{hestness2017deep}. This empirical success implies that natural data must possess additional structure that makes them efficiently learnable. One classical hypothesis \cite{biederman1987recognition} attributes the effectiveness of recognition systems to compositionality, where complex objects are composed of simpler features, which themselves are composed of sub-features \cite{poggio2017and, deza2020hierarchically,bietti2021approximation}. From this perspective, the locality inherent in CNNs is considered critical for their performance, a view supported by empirical evidence \cite{Neyshabur2020towards}. Nevertheless, a clear quantitative understanding of how compositionality in data influences learning curves remains elusive. 

We investigate this relationship within a teacher-student framework, where the function to be learned takes one of two specific forms:
\begin{equation}\label{eq:loc-loc-f}
    f^{LC}(\x) = \sum_{i\in \mathcal{P}} g_i(\x_i), \ \ \ f^{CN}(\x) = \sum_{i\in \mathcal{P}} g(\x_i).
\end{equation}
Here, $\x$ denotes a $d$-dimensional input, and each $\x_i$ represents the $i$-th patch of $\x$, $\x_i\,{=}\,(x_i, \dots, x_{i+t-1})$, of length $t\,{<}\,d$. The indices $i$ belong to a subset $\mathcal{P}$ of $\left\lbrace 1,\dots, d\right\rbrace$.  The functions $g_i$ and $g$ are random functions, with smoothness governed by an exponent $\alpha_t$. For instance, such functions can be realized by randomly initialized one-hidden-layer networks. $f^{LC}$ corresponds to the output of a \emph{locally connected} network (LCN) \cite{fukushima1975cognitron, lecun1989generalization}, where inputs are first decomposed into smaller patches before processing, while $f^{CN}$ characterizes networks imposing invariance to input shifts via weight sharing, under an appropriate choice of $\mathcal{P}$. 

Our objective is to compute the learning curve exponent $\beta$ achieved by a student performing kernel regression on these functions. Specifically, the student kernel embodies a prior on the true function consistent with the functional forms described in \autoref{eq:loc-loc-f}, albeit potentially different from the teacher in terms of filter size ($s$) and smoothness ($\alpha_s$). This model includes infinite-width one-hidden-layer neural networks operating in the \emph{lazy training regime} as a special case \cite{jacot2018neural, Du2019, lee2019wide, arora2019exact, chizat2019lazy}.

In particular, this chapter analyzes a teacher-student model, where the teacher is a Gaussian random field with covariance $\KT (\x, \y)$, possessing a specific filter size $t$ and a smoothness exponent $\alpha_t > 0$. Kernel regression is implemented by a student with corresponding parameters $s$ and  $\alpha_s>0$. Our main contributions are as follows.

Using recent findings based on the replica method from statistical physics for generalization in kernel methods \cite{bordelon2020spectrum, canatar2021spectral,loureiro2021capturing}, we derive the learning curve exponent.  We establish that $\beta=\alpha_t/s$ when $t\leq s$ and $\alpha_t\leq 2(\alpha_s+s)$. Although this result is non-rigorous in general, it can be rigorously proven for Gaussian fields when data are sampled from a lattice \cite{spigler2020asymptotic}, and it corresponds to a provable lower bound on the error when teacher and student are matched \cite{micchelli1981design}. 
We systematically verify our theoretical predictions through extensive numerical experiments performing ridgeless regression across various filter sizes $t$, $s$, and input dimensions $d$.
Finally, leveraging the recent framework of \citet{jacot2020kernel} and a Gaussian universality assumption, we prove a rigorous estimate of $\beta$ for ridge regression when the ridge parameter decreases with the size of the training set. Crucially, the exponent depends only on $s$ and is independent of $d$, explicitly demonstrating that using local filters on compositional data allows circumventing the curse of dimensionality.

Collectively, our findings show that incorporating locality priors significantly mitigates the curse of dimensionality when applied to compositional data. By contrast, enforcing shift invariance affects prefactors entering the learning curve, rather than the scaling exponent $\beta$.\looseness=-1  

\section{Related work}

Several recent studies have focused on the role of compositional structure in data. When such structure is hierarchical, deep convolutional networks have been shown to possess significantly greater expressive power than shallow architectures \cite{poggio2017and, Poggio2020theo, deza2020hierarchically}. From a training perspective, \citet{malach2021computational} demonstrated that convolutional and locally-connected networks can achieve target generalization errors in polynomial time for functions that depend solely on $s$ consecutive bits of a $d$-dimensional input, with $s\,{=}\mathcal{O}(\log{d})$. Conversely, fully-connected networks do not share this advantage.

\citet{bietti2021approximation} explored the impact of locality in the architecture through the lens of kernel methods, using deep convolutional kernels introduced earlier by \citet{mairal2016end, bietti2019inductive}, and characterized their associated Reproducing Kernel Hilbert Space (RKHS). Membership in the RKHS guarantees beneficial performance bounds. However, for isotropic kernels, this membership imposes constraints on function smoothness that become increasingly restrictive as the dimensionality $d$ grows. By contrast, for functions exhibiting locality, smoothness constraints depend on the filter size $s$, rather than the dimensionality $d$ \cite{bietti2021approximation}. Additionally, \citet{mei2021learning} recently established that weight sharing, without locality, provides only a modest improvement in the generalization performance of shift-invariant kernels.

In contrast to the above studies, in this chapter, we specifically compute nontrivial learning curve exponents, within a framework where the locality and shift-invariance priors of the kernel do not necessarily match those of the function class being learned. Notably, in our setup, the target functions typically do not belong to the RKHS of the kernel\footnote{Indeed, a Gaussian random field of covariance $\mathcal{K}$ is never an element of the RKHS associated with the same kernel $\mathcal{K}$, see, e.g. \cite{kanagawa2018gaussian}.}.
Technically, our finding that student's filter size $s$ determines the learning curve exponent -- rather than the teacher's filter size $t$ -- reflects the inability of kernels to capture data anisotropy, i.e., dependencies limited to subsets of input coordinates, both in worst-case \cite{bach2017breaking} and typical scenarios involving Gaussian fields \cite{paccolat2020isotropic}.

\section{Setup}\label{sec:loc-setup}

\paragraph{Kernel ridge regression}
Kernel ridge regression is a method for learning a target function $f^*$ from $P$ observations $\{(\x^\nu, f^*_\nu)\}_{\nu=1}^P$, where the inputs $\x_\nu\in\mathbb{R}^d$ are i.i.d. random variables drawn according to a probability distribution $p\left( d^d x\right)$ on $\mathbb{R}^d$, and  $f^*_\nu \,{:=}\, f^*(\x_\nu)$. Given a positive-definite kernel $\K$ and its corresponding Reproducing Kernel Hilbert Space (RKHS) $\mathcal{H}$, the kernel ridge regression estimator $\hat{f}$ of $f^*$ is defined by:
\begin{equation}\label{eq:loc-argmin}
 \hat{f} =\argmin_{f\in \mathcal{H}} \left\lbrace \frac{1}{P}\displaystyle \sum_{\nu=1}^P \left(f(\x_\nu) - f^*_\nu\right)^2  + \lambda \, \|f\|^2_{\mathcal{H}} \right  \rbrace,
\end{equation}
where $\|\cdot\|_{\mathcal{H}}$ denotes the RKHS norm and $\lambda$ is the ridge regularization parameter. The ridgless case ($\lambda\to0^+$) corresponds to the minimum-norm interpolating solution. The optimization problem in \autoref{eq:loc-argmin} is convex and its unique solution is
\begin{equation}\label{eq:loc-krrpredictor}
 \hat{f}(\x) = \frac{1}{P}\sum_{\mu,\nu=1}^P \K(\x,\x_\nu) \left(\left( \frac{1}{P} \mathbf{K}_P +\lambda \mathbf{I}_P\right)^{-1}\right)_{\mu,\nu} f^*_{\nu}, 
\end{equation}
where $\mathbf{K}_P$ is the \emph{Gram matrix} defined as $(\mathbf{K}_P)_{\mu\nu} = \K(\x_\mu, \x_\nu)$, and $\mathbf{I}_P$ being the identity matrix of dimension $P$. Our objective is the generalization error, defined as the expected mean squared error over the data distribution $p\left( d^d x\right)$ and the target $f^*$, i.e.,
\begin{equation}\label{eq:loc-test-def}
\testerr = \mathbb{E}_{\x,f^*}\left[ \left( \hat{f}(\x)-f^*(\x)\right)^2 \right].
\end{equation}

\paragraph{Statistical mechanics of generalization in kernel regression}

In general, theoretical estimation of the generalization error is still an open problem. Recent works \cite{bordelon2020spectrum, canatar2021spectral} derived approximate expressions for $\testerr$ using the decomposition of the target function in the eigenbasis of the kernel. Mercer's theorem allows any positive-definite kernel $\K$ to be expressed in terms of its eigenvalues $\{\lambda_\rho\}$ and eigenfunctions $\{\phi_\rho\}$ as:
\begin{equation}\label{eq:loc-mercer}
 \K(\x,\y) = \sum_{\rho=1}^\infty \lambda_\rho \phi_\rho(\x) \overline{\phi_\rho(\y)}, \quad \int \K(\x,\y) \phi_\rho(\y)  p\left( d^d y\right) = \lambda_\rho \phi_\rho(\x). 
\end{equation}
Defining the kernel features $\psi_\rho(\x)= \sqrt{\lambda_\rho}\phi_\rho(\x)$, since the kernel's eigenfunctions form a complete basis, the target function and estimator can be decomposed as:
\begin{equation}
 f^*(\x) = \sum_\rho w^*_\rho \psi_\rho(\x), \quad \hat{f}(\x) = \sum_\rho w_\rho \psi_\rho(\x).
\end{equation}
The replica method -- a heuristic technique from statistical physics \cite{Mezard87} -- yields the following set of equations in the ridgeless limit $\lambda\to0^+$ \cite{bordelon2020spectrum, canatar2021spectral}:
\begin{equation}
\testerr(P) = \sum_\rho \frac{\mathbb{E}[|w_\rho^*|^2]}{\lambda_\rho} \left( \frac{1}{\lambda_\rho} + \frac{P}{t(P)} \right)^{-2} \left( 1 - \frac{P\gamma(P)}{t(P)^2} \right)^{-1},
\end{equation}
\begin{equation}
t(P) = \sum_\rho \left( \frac{1}{\lambda_\rho} + \frac{P}{t(P)} \right)^{-1}, \quad \gamma(P) = \sum_\rho \left( \frac{1}{\lambda_\rho} + \frac{P}{t(P)} \right)^{-2}.
\end{equation}
The learning curve exponent $\beta$ can be obtained from the asymptotic analysis of these equations. We specifically assume a power-law spectrum for both the kernel eigenvalues and the target function coefficients: \emph{i)}~$\lambda_\rho = \rho^{-a}$ and \emph{ii)}~$\mathbb{E}[|{c_\rho}|^2] \equiv \lambda_\rho \mathbb{E}[|w^*_\rho|^2]= \rho^{-b}$, with $2a\,{>}\,b-1$. Under these conditions, the generalization error scales as \cite{spigler2020asymptotic, bordelon2020spectrum}
\begin{equation}\label{eq:loc-error-scaling}
 \testerr(P) \sim \sum_{\rho > P} \mathbb{E}[|c_\rho|^2] \equiv \mathcal{B}(P).
\end{equation}

\autoref{eq:loc-error-scaling} implies that the generalization error can be approximated by summing the residual power of the target function beyond the first $P$ kernel modes, denoted by $\mathcal{B}(P)$. Additional rigorous results are available in special teacher-student settings \cite{Sollich_2002, sollich2001gaussian, spigler2020asymptotic, paccolat2020isotropic}:
\begin{itemize}
    \item For isotropic teacher and student kernels and data sampled on a lattice \autoref{eq:loc-error-scaling} can be proven rigorously \cite{spigler2020asymptotic};
    \item When teacher and student coincide, $\testerr(P) \leq \mathcal{B}(P)$, providing a rigorous lower bound on performance \cite{micchelli1981design}.
\end{itemize}

\section{Convolutional and local kernels}\label{sec:loc-convolutional-mercer}

In this section, we introduce convolutional and local kernels, and motivate our choice by considering one-hidden-layer convolutional architectures. Due to the close relationship between our kernels and the Neural Tangent Kernel of one-hidden-layer convolutional neural networks, our framework naturally encompasses regression tasks using simple neural networks in the lazy training regime. For simplicity, we limit the discussion to inputs represented as sequences in $\mathbb{R}^d$, denoted as $\x \,{=}\,(x_1,\dots,x_d)$. Extending these definitions to higher-order tensor inputs such as images $\x\in\mathbb{R}^{d\times d}$ is straightforward. We handle boundary conditions by setting $x_{i+d}=x_i$ for all $i\,{=}\,1,\dots,d$.

\begin{definition}[one-hidden-layer CNN]\label{eq:loc-cnn} 
A one-hidden-layer convolutional network with $H$ hidden neurons and average pooling is defined as:
\begin{equation}\label{eq:loc-cnn-out}
    f^{CNN}(\x) = \frac{1}{\sqrt{H}} \sum_{h=1}^H a_h \frac{1}{|\mathcal{P}|}\sum_{i\in\mathcal{P}} \sigma(\w_{h}^\top \x_i + b_h),
\end{equation}
where $\x\in\mathbb{R}^d$, $H$ denotes the width, $\sigma$ is a nonlinear activation function, $\mathcal{P} \subseteq \left\lbrace{1,\dots,d}\right\rbrace$ is a set of patch indices, and $|\mathcal{P}|$ its cardinality. For all $i\in\mathcal{P}$, $\x_i$ is an $s$-dimensional patch of $\x$. For all $h\,{=}\,1,\dots,H$, $\w_{h}\in\mathbb{R}^s$ is a filter with filter size $s$, $b_h\in\mathbb{R}$ is a scalar bias, and $a_{h}\in\mathbb{R}$ is a scalar weight.
\end{definition}
In the network defined above, a $d$-dimensional input sequence $\x$ is first mapped to $s$-dimensional \emph{patches} $\x_i$, which are ordered subsequences of the input. Comparing each patch to a filter $\w_{h}$ and applying the activation function $\sigma$ yields a $|\mathcal{P}|$-dimensional hidden representation that is equivariant for shifts of the input. The summation over the patch index $i$ promotes this equivariance to full invariance, leading to a model that is both local and shift-invariant as $f^{CN}$ in \autoref{eq:loc-loc-f}. A model which is only local (as $f^{LC}$ in \autoref{eq:loc-loc-f}) can be obtained by lifting the constraint of weight-sharing, which forces, for each $h\,{=}\,1,\dots,H$, the same filter $\w_{h}$ to apply to all patches $\x_i$.
\begin{definition}[one-hidden-layer LCN]\label{eq:loc-lcn} 
A one-hidden-layer locally-connected network with $H$ hidden neurons is given by:
\begin{equation}\label{eq:loc-lcn-out}
    f^{LCN}(\x) = \frac{1}{\sqrt{H}} \sum_{h=1}^H \frac{1}{\sqrt{|\mathcal{P}|}}\sum_{i\in\mathcal{P}} a_{h,i} \sigma(\w_{h,i}^\top \x_i + b_{h,i}),
\end{equation}
For all $i\in\mathcal{P}$ and $h\,{=}\,1,\dots,H$: $\x_i$ is an $s$-dimensional patch of $\x$, $\w_{h,i}\in\mathbb{R}^s$ is a filter with filter size $s$, $b_h\in\mathbb{R}$ is a scalar bias, and $a_{h,i}\in\mathbb{R}$ is a scalar weight. \end{definition}
The above reduces to a fully-connected network when the filter size is set to the input dimension, $s\,{=}\,d$, and $\mathcal{P}=\left\lbrace1\right\rbrace$. With the target functions taking one of the two forms in \autoref{eq:loc-loc-f}, our framework contains the case where the observations are generated by neural networks such as \autoref{eq:loc-cnn} and \autoref{eq:loc-lcn}. 

We now introduce the concept of Neural Tangent Kernels (NTK):

\begin{definition}[Neural Tangent Kernel] For a neural network function $f(\x;\bm{\theta})$, where $\bm{\theta}\,{=}\,(\theta_1,\dots,\theta_N)$ denotes the complete set of parameters and $N$ the total number of parameters, the Neural Tangent Kernel (NTK) is defined as:
\cite{jacot2018neural}
\begin{equation}\label{eq:loc-finite_ntk}
    \K_{\mathrm{NTK},N}(\x,\y;\bm{\theta}) = \sum_{n=1}^N \partial_{\theta_n} f(\x,\bm{\theta})\partial_{\theta_n} f(\y,\bm{\theta}).
\end{equation}
\end{definition}
For one-hidden-layer networks with random, $\mathcal{O}(1)$-variance Gaussian initialization of all the weights, and normalization by $\sqrt{H}$ as in \autoref{eq:loc-cnn} and \autoref{eq:loc-lcn}, the NTK converges to a deterministic limit $\K_{\mathrm{NTK}}(\x,\y)$ as $N\propto H \to\infty$ \cite{jacot2018neural}. Training $f(\x,\bm{\theta})-f(\x,\bm{\theta}_0)$, with $\bm{\theta}_0$ denoting the network parameters at initialization, under gradient descent on the mean squared error is equivalent to performing ridgeless regression with the kernel $\K_{\mathrm{NTK}}(\x,\y)$\cite{jacot2018neural}. 

The following lemmas relate NTKs of convolutional and local networks acting on $d$-dimensional inputs to NTKs of a fully connected network acting on $s$-dimensional inputs (proofs in \autoref{app:loc-ntk}).
\begin{lemma}\label{lemma:cntk}
Denoting as $\K_{\mathrm{NTK}}^{FC}$ the NTK of a fully-connected network function acting on $s$-dimensional inputs and $\K_{\mathrm{NTK}}^{CN}$ the NTK of a convolutional network function~(\autoref{eq:loc-cnn}) with filter size $s$ acting on $d$-dimensional inputs,
\begin{equation}\label{eq:loc-conv-ntk}
    \K_{\mathrm{NTK}}^{CN}(\x,\y) = \frac{1}{|\mathcal{P}|^2}\sum_{i,j\in\mathcal{P}} \K_{\mathrm{NTK}}^{FC}(\x_i,\y_j).
\end{equation}
\end{lemma}
As the functions in \autoref{eq:loc-loc-f}, $\K_{\mathrm{NTK}}^{CN}$ is written as a combination of a lower-dimensional constituent kernel $\K_{\mathrm{NTK}}^{FC}$ acting on patches, and the dimensionality of the constituent kernel coincides with the filter size of the corresponding network. 

\begin{lemma}\label{lemma:lntk}
Call $\K_{\mathrm{NTK}}^{LC}$ the NTK of a locally-connected network function~(\autoref{eq:loc-lcn}) with filter size $s$ acting on $d$-dimensional inputs. Then
\begin{equation}\label{eq:loc-loc-ntk}
    \K_{\mathrm{NTK}}^{LC}(\x,\y) = \frac{1}{|\mathcal{P}|}\sum_{i\in\mathcal{P}} \K_{\mathrm{NTK}}^{FC}(\x_i,\y_i).
\end{equation}
\end{lemma}

Following the general structure of \autoref{eq:loc-conv-ntk} and \autoref{eq:loc-loc-ntk}, we define local ($\K^{LC}$) and convolutional ($\K^{CN}$) kernels as sums of lower-dimensional constituent kernels $\mathcal{C}$,
\begin{subequations}\label{eq:loc-convloc-ker}
\begin{align}
    \label{eq:loc-conv-ker} \K^{CN}(\x,\y) &= |\mathcal{P}|^{-2} \displaystyle\sum_{i,j\in\mathcal{P}} \mathcal{C}(\x_i, \y_j),\\
    \label{eq:loc-loc-ker} \K^{LC}(\x,\y) &= |\mathcal{P}|^{-1} \displaystyle\sum_{i\in\mathcal{P}} \mathcal{C}(\x_i, \y_i).
\end{align}
\end{subequations}
These are characterized by the dimensionality of the constituent kernel $\mathcal{C}$, or \emph{filter size} $s$ (for the student, or $t$ for the teacher) and a smoothness exponent $\alpha$ characterizing the nonanalytic behavior of the kernel at small distance, $\mathcal{C}(\x_i,\y_j)\sim \|\x_i-\y_j\|^{\alpha_{s}}$ (for the student, or $\alpha_t$ for the teacher) plus analytic contributions, with $\alpha_s\neq 2m$ for $m\in\mathbb{N}$. The corresponding target function $f^*$ and estimator $\hat{f}$ have the form displayed in \autoref{eq:loc-loc-f}. The exponent $\alpha$ controls the smoothness of these functions, in the sense that, if $\alpha\,{>}\,n \in \mathbb{N}$, then the constituent kernel $\mathcal{C}$ is at least $n$ times differentiable \cite{spigler2020asymptotic}.

For instance, for the NTK of ReLU networks $\K_{\mathrm{NTK}}^{FC}$, which has a cusp at the origin, $\alpha_s\,{=}\,1$ \cite{geifman2020similarity}. Moreover, in the $H\to\infty$ limit, a network initialized with random weights converges to a Gaussian process \cite{Neal1996, williams1997computing, lee2017deep, matthews2018gaussian, novak2018bayesian}. The covariance kernel of the process, for ReLU activations, has nonanalytic behavior with $\alpha_t\,{=}\,3$ \cite{cho2009kernel}.

\paragraph{Mercer's decomposition}

The eigendecomposition of the constituent kernel $\mathcal{C}$ induces an eigendecomposition of convolutional and local kernels. We work under the following assumptions:
\begin{itemize}
    \item[$i)$] The constituent kernel $\mathcal{C}(\x, \y)$ on $\mathbb{R}^s\times \mathbb{R}^s$ admits the following Mercer's decomposition,
    \begin{equation}\label{eq:loc-mercer-c}
        \mathcal{C}(\x, \y) = \sum_{\rho=1}^\infty \lambda_\rho \phi_\rho(\x)\phi_\rho(\y),
    \end{equation}
    with (ordered) eigenvalues $\lambda_{\rho}$ and eigenfunctions ${\phi_\rho}$ such that, with $p^{(s)}(d^s x)$ denoting the $s$-dimensional patch measure, $ \phi_1(\x) = 1 \; \forall \x$ and $ \int p^{(s)}(d^s x) \phi_\rho(\x)\,{=}\,0$ for all $\rho\,{>}1$;
    \item[$ii)$] Convolutional and local kernels from \autoref{eq:loc-convloc-ker} have \emph{nonoverlapping} patches, i.e., $d$ is an integer multiple of $s$ and 
    \begin{equation}
        \mathcal{P}\,{=}\,\left\lbrace 1 + n\times s \, | \, n=1,\dots,d/s \right\rbrace
    \end{equation} with $|\mathcal{P}|\,{=}d/s$;
    \item[$iii)$] The $s$-dimensional marginals on patches of the $d$-dimensional input measure $p^{(d)}(d^d x)$ are all identical and equal to $p^{(s)}(d^s x)$.
\end{itemize}
The part of assumption $i)$ regarding the eigenfunctions' properties is satisfied, for example, when the constituent kernel $\mathcal{C}$ is isotropic and data are distributed uniformly on a $d$-dimensional torus. The request of nonoverlapping patches in assumption $ii)$ can be relaxed at the price of further assumptions, i.e., $\mathcal{C}(\x,\y)\,{=}\,c(\x-\y)$ and data distributed uniformly on the torus, so that $\mathcal{C}$ is diagonalized in Fourier space (details in \autoref{app:loc-mercer-overlap}).

\begin{lemma}[Spectra of convolutional kernels]\label{lemma:conv-spectra}
Let $\K^{CN}$ be a convolutional kernel with constituent kernel $\mathcal{C}$ satisfying assumptions $i)$, $ii)$ and $iii)$. $\K^{CN}$ has the following Mercer's decomposition,
\begin{equation}\label{eq:loc-conv-decomp}
\K^{CN}(\x,\y) =\sum_{\rho=1}^{\infty} \Lambda_\rho \Phi_\rho(\x) \overline{\Phi_\rho(\y)},
\end{equation}
with eigenvalues and eigenfunctions
\begin{equation}\label{eq:loc-conv-spectrum}
  \Lambda_1 = \lambda_1,\,  \Phi_1(\x) = 1;\,\, \Lambda_\rho = \frac{s}{d}\lambda_\rho,\,  \Phi_\rho(\x) = \sqrt{\frac{s}{d}}\sum_{i\in\mathcal{P}}\phi_\rho(\x_i)  \text{ for }\rho > 1.
\end{equation}
\end{lemma}

\begin{lemma}[Spectra of local kernels]\label{lemma:loc-spectra}
Let $\K^{LC}$ be a local kernel with constituent kernel $\mathcal{C}$ satisfying assumptions $i)$, $ii)$ and $iii)$ above. Then $\K^{LC}$ has the following Mercer's decomposition,
\begin{equation}\label{eq:loc-loc-decomp}
\K^{LC}(\x,\y) = \Lambda_1 \Phi_1(\x) \overline{\Phi_1(\y)} +  \sum_{\rho>1}^{\infty}\sum_{i\in\mathcal{P}} \Lambda_{\rho,i} \Phi_{\rho,i}(\x) \overline{\Phi_{\rho,i}(\y)},
\end{equation}
with eigenvalues and eigenfunctions ($\forall i\in \mathcal{P}$)
\begin{equation}\label{eq:loc-loc-spectrum}
  \Lambda_{1} = \lambda_1,\,  \Phi_{1}(\x) = 1;\,\, \Lambda_{\rho,i} = \frac{s}{d}\lambda_\rho,\,  \Phi_{\rho,i}(\x) = \phi_\rho(\x_i)  \text{ for }\rho > 1.
\end{equation}
\end{lemma}

We refer the reader to \autoref{app:loc-mercer-overlap} for the proof of the lemmas and the generalization to kernels with overlapping patches. 

In the next section, we explore how these spectra affect the asymptotic behavior of learning curves.

\section{Asymptotic learning curves for ridgeless regression}\label{sec:loc-learning-curves}

In what follows, we explicitly focus on translationally-invariant constituent kernels $\mathcal{C}(\x_i,\x_j)\,{=}\, c(\x_i - \x_j)$ and assume a uniform data distribution  $p(d^dx)$ over a $d$-dimensional torus. This assumption ensures that all lower-dimensional marginals are themselves uniformly distributed over smaller-dimensional tori. Under these conditions, Mercer's decomposition can be conveniently represented in Fourier space \cite{scholkopf2001learning}, where the eigenfunctions correspond to $s$-dimensional plane waves $\phi^{(s)}_{\k}(\x)\,\allowbreak{=}\allowbreak\,e^{i\k^\top\x}$ and the eigenvalues coincide with the Fourier transform of $c$. Thus, all the assumptions for lemmas \ref{lemma:conv-spectra} and \ref{lemma:loc-spectra} are satisfied. Moreover, for kernels characterized by filter size $s$ (or $t$) and smoothness exponent $\alpha_s$ (or $\alpha_t$), their eigenvalues decay with a power $-(s\,{+}\,\alpha_s)$ (or $-(t\,{+}\,\alpha_t)$) of the wavevector modulus $k\,{=}\,\sqrt{\k^\top\k}$ \cite{widom1964asymptotic}. 

In this setting, we present our central result: 
\begin{theorem}\label{th:scaling}
Let $\KT $ be a $d$-dimensional convolutional kernel with a translationally-invariant $t$-dimensional constituent and leading nonanalyticity at the origin controlled by the exponent $\alpha_t\,{>}\,0$. Let $\KS$ be a $d$-dimensional convolutional or local student kernel with a translationally-invariant $s$-dimensional constituent, and with a nonanalyticity at the origin controlled by the exponent $\alpha_s\,{>}\,0$. If all the $t$-dimensional patches of the teacher are contained in at least one of the $s$-dimensional patches of the student\footnote{This condition is satisfied when $s \geq t$ in the full overlapping-patches case, while requires that $s$ is an integer multiple of $t$ in the nonoverlapping-patches case.}, and data are uniformly distributed on a $d$-dimensional torus, the following asymptotic equivalence holds in the limit $P\to\infty$,
\begin{equation}
    \mathcal{B}(P) \sim P^{-\beta}, \quad \beta = \alpha_t / s.
\end{equation}
\end{theorem}
Combining \autoref{th:scaling} with \autoref{eq:loc-error-scaling}, and under the additional assumption that $\alpha_t\,{\leq}\, 2(\alpha_s + s)$, we derive the following prediction for the asymptotic of the learning curve: 
\begin{equation}\label{eq:loc-prediction}
    \testerr(P) \sim P^{-\beta}, \quad \beta = \alpha_t / s.
\end{equation}
Importantly, as $\beta$ does not depend on the input dimension $d$, we conclude that the curse of dimensionality is beaten when a convolutional target is learned with a convolutional or local kernel. In fact, \autoref{eq:loc-prediction} indicates that there is no asymptotic advantage in using a convolutional rather than local student when learning a convolutional task, supporting the notion that the primary factor underlying the empirical success of convolutional architectures is their locality rather than weight sharing. Empirical evidence validating these theoretical predictions is presented in \autoref{sec:loc-empirical}.

\autoref{th:scaling} is proven in \autoref{app:loc-thm1} and extended to the scenario of local teachers and students in \autoref{app:loc-local}. Below, we provide an intuitive proof sketch for the simpler nonoverlapping patch case.

We begin by calculating the variance of the coefficients of the target function in the student eigenbasis. By indexing the coefficients with the $s$-dimensional wavevectors $\k$,
\begin{equation}\label{eq:loc-coeff}\begin{aligned}
  \mathbb{E}[|c_{\k}|^2] &= \int_{[0,1]^d} d^d x \Phi_{\k}(\x)\int_{[0,1]^d} d^d y \overline{\Phi_{\k}(\y)} \mathbb{E}[f^*(\x)f^*(\y)]\\ & =\int_{[0,1]^d} d^d x \Phi_{\k}(\x)\int_{[0,1]^d} d^d y \overline{\Phi_{\k}(\y)} \KT (\x, \y). 
\end{aligned}\end{equation}
When the teacher and student kernels share the same filter size ($s\,{=}\,t$) they share the same eigenfunctions. Using the eigenvalue equation for the teacher kernel we find $\mathbb{E}[|c_{\k}|^2] \sim k^{-(\alpha_t + t)}\,{=}\,k^{-(\alpha_t + s)}$. Ordering eigenvalues by increasing wavevector magnitude $k$, with multiplicity $k^{s-1}$ from all the wavevectors having the same modulus, we have:
\begin{equation}\label{eq:loc-t-equal-s}
    \mathcal{B}(P) = \sum_{\left\lbrace \k| k > P^{1/s}\right\rbrace} k^{-(\alpha_t + s)} \sim \int_{P^{1/s}}^\infty k^{-(\alpha_t + s)} k^{s-1} dk \sim P^{-\frac{\alpha_t}{s}}.
\end{equation}
When the teacher filter size $t$ is reduced, some coefficients $\mathbb{E}[|c_{\k}|^2]$ vanish. Specifically, as the target function becomes a composition of $t$-dimensional constituents, the only non-zero coefficients are found for $\k$'s which lie in some $t$-dimensional subspaces of the $s$-dimensional Fourier space. These subspaces correspond to wavevectors having at most a patch of $t$ consecutive non-vanishing components. In other words, $\mathbb{E}[|c_{\k}|^2]$ is finite only if $\k$ is effectively $t$-dimensional and the integral on the right-hand side of \autoref{eq:loc-t-equal-s} becomes $t$-dimensional, thus
\begin{equation}\label{eq:loc-t-smaller-s}
    \mathcal{B}(P) \sim \int_{P^{1/s}}^\infty k^{-(\alpha_t + t)} k^{t-1} dk \sim P^{-\frac{\alpha_t}{s}}.
\end{equation}

Finally, if the teacher patches are not contained in the student ones, the target function cannot be represented in the student eigenbasis. Therefore, the generalization error does not decay to zero but instead approaches a finite positive limit as $P\to\infty$.

\section{Empirical learning curves for ridgeless regression}\label{sec:loc-empirical}

\begin{figure}
    \centering
    \includegraphics[width=1.0\linewidth]{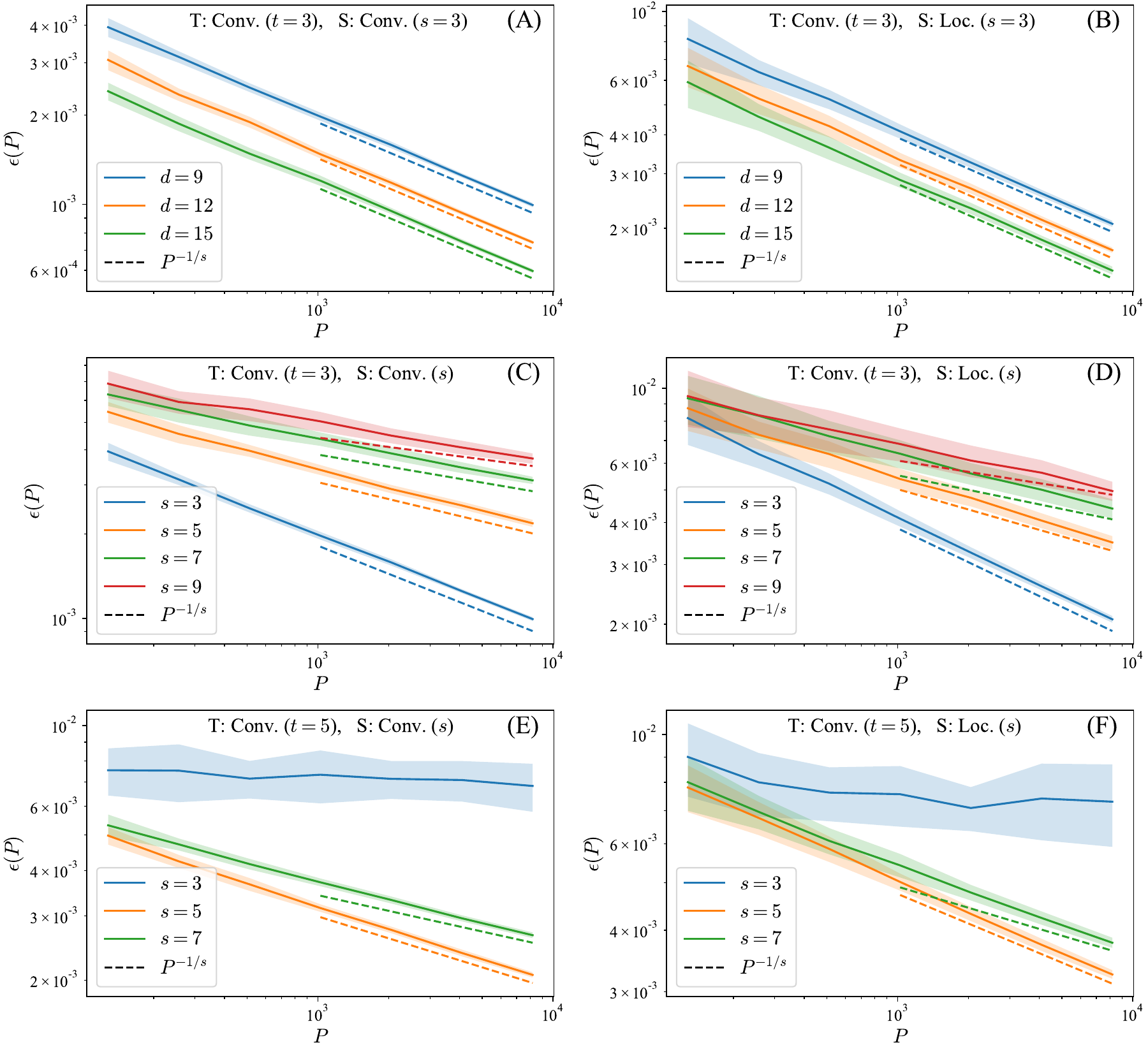}
    \caption[Learning curves for different combinations of convolutional teachers with convolutional and local students.]{Learning curves for different combinations of convolutional teachers with convolutional (left panels) and local (right panels) students. The teacher and student filter sizes are denoted with $t$ and $s$, respectively. Data are sampled uniformly in the hypercube $[0,1]^d$, with $d=9$ if not specified otherwise. Solid lines are the results of numerical experiments averaged over 128 realizations, and the shaded areas represent the empirical standard deviations. The predicted scaling is shown by dashed lines. All the panels are discussed in \autoref{sec:loc-empirical}, while additional details on experiments are reported in \autoref{app:loc-numerics}, together with additional experiments.}
    \label{fig:figure}
\end{figure}

We numerically validate the asymptotic behavior of learning curves within our teacher-student framework. We simulate different combinations of convolutional and local teachers and students with overlapping patches and Laplacian constituent kernels defined by $c(\x_i-\x_j) \, {=} \, e^{-\|\x_i-\x_j\|}$. To test the robustness of our predictions to different data distributions, we consider data uniformly generated in the hypercube $[0,1]^d$ (\autoref{fig:figure}) or on a $d$-dimensional hypersphere (\autoref{app:loc-numerics}). \autoref{fig:figure} presents learning curves for convolutional (left panels) and local (right panels) students learning a convolutional target function. Additional results for the case of a local teacher are included in \autoref{app:loc-numerics}, and confirm the same qualitative trends.

We refer throughout to the six panels of \autoref{fig:figure}. Panels A and B show that, under the assumptions of \autoref{th:scaling}, with $\alpha_t\,{=}\,\alpha_s\,{=}\,1$, our prediction $\beta\,{=}\,1/s$ holds independently of the embedding dimension $d$. Moreover, fixing the dimension $d$ and the teacher filter size $t$, the generalization errors of a convolutional and a local student with the same filter size differ only by a multiplicative constant independent of $P$. Indeed, the shift-invariant nature of the convolutional student only results in a pre-asymptotic correction to our estimate of the generalization error $\mathcal{B}(P)$. Panels C and D display learning curves for different values of $s$ and fixed $t$. When the size of the student filters matches the input dimension, the curse of dimensionality is recovered. Panels E and F show learning curves for fixed $t$ and $s$ being smaller than, equal to, or larger than $t$. When $s\,{<}\,t$, the student kernel cannot represent the target function, and hence the error does not decrease by increasing $P$. 

All empirical results are in excellent agreement with the theoretical predictions. Additional experimental details, as well as results using the Neural Tangent Kernel of a one-hidden-layer fully-connected network as the constituent kernel, are reported in \autoref{app:loc-numerics}. Notably, despite the lack of translational invariance in that setting, our theoretical predictions still hold.

\section{Asymptotics of learning curves with ridge}

In this section, we prove an upper bound on the learning curve exponent $\beta$, thereby confirming that the curse of dimensionality is beaten by a local or convolutional kernel learning a convolutional target. Our approach is based on the framework developed by \citet{jacot2020kernel} combined with a natural universality assumption. Crucially, this framework does not assume the target function to be generated by a teacher kernel. 

The proofs are presented in \autoref{app:loc-thm2}. 

Let $\mathcal{D}(\Lambda)$ denote the density of eigenvalues of the student kernel, $\mathcal{D}(\Lambda) \,{=}\, \sum_\rho \delta(\Lambda - \Lambda_\rho)$, with $\delta(x)$ denoting Dirac delta function. Having a random target function with coefficients $c_\rho$ in the kernel eigenbasis having variance $\mathbb{E}[|c_\rho|^2]$, one can define the following reduced density (with respect to the teacher):
\begin{equation}\label{eq:loc-reduced-density}
\mathcal{D}_T(\Lambda) = \displaystyle\sum_{\left\lbrace\rho \, | \, \mathbb{E}[|c_\rho\|^2] > 0\right\rbrace } \delta(\Lambda - \Lambda_\rho).
\end{equation}
$\mathcal{D}_T(\Lambda)$ counts  eigenvalues for which the target has a non-zero variance, such that:
\begin{equation}
    \sum_{\rho} \mathbb{E}[|c_\rho|^2] = \int c^2(\Lambda)  \mathcal{D}_T(\Lambda) d\Lambda,
\end{equation}
where the function $c(\Lambda)$ is defined by $c^2(\Lambda_\rho)\,{=}\,\mathbb{E}[|c_\rho|^2] $ for all $\rho$ such that $\mathbb{E}[|c_\rho|^2]\,{>}\,0$.
The following theorem then follows from the results of \citet{jacot2020kernel}. 
\begin{theorem}\label{th:ridge}
Let us consider a positive-definite kernel $\K$ with eigenvalues $\Lambda_\rho$, $\sum_\rho \Lambda_\rho < \infty$, and eigenfunctions ${\Phi_\rho}$ learning a (random) target function $f^*$ in kernel ridge regression (\autoref{eq:loc-argmin}) with ridge $\lambda$ from $P$ observations $f^*_{\nu}\,{:=}\,f^*(\x_\nu)$, with $\x_\nu\in \mathbb{R}^d$ drawn from a certain probability distribution. Let us denote with $\mathcal{D}_T(\Lambda)$ the reduced density of kernel eigenvalues with respect to the target and $\testerr(\lambda,P)$ the generalization error, and also assume that
\begin{itemize}
    \item[$i)$] For any $P$-tuple of indices $\rho_1,\dots,\rho_P$, the vector $(\Phi_{\rho_1}(\x_1), \dots,\Phi_{\rho_P}(\x_P))$ is a Gaussian random vector;
    \item[$ii)$] The target function can be written in the kernel eigenbasis with coefficients $c_\rho$ and $c^2(\Lambda_\rho)\,{=}\,\mathbb{E}[|c_\rho|^2]$, with $\mathcal{D}_T(\Lambda) \sim \Lambda^{-(1+r)}$, $c^2(\Lambda) \sim \Lambda^{q}$ asymptotically for small $\Lambda$ and $r\,{>}\,0$, $r\,{<}\,q\,{<}\, r\,{+}\,2$;
\end{itemize}
Then the following equivalence holds in the joint $P\to\infty$ and $\lambda\to 0$ limit with $1/(\lambda\sqrt{P})\to 0$:
\begin{equation}
    \testerr(\lambda, P) \sim  \sum_{\left\lbrace \rho \, | \, \Lambda_\rho < \lambda \right\rbrace} \mathbb{E}{[|c_\rho|^2]} = \int_0^{\lambda}   c^2(\Lambda) \mathcal{D}_T(\Lambda) d\Lambda.
\end{equation}
\end{theorem}

It is important to note that assumption $i)$ of the theorem -- requiring Gaussianity of the eigenbasis -- is not strictly satisfied in our setting where the eigenfunctions $\Phi_\rho$'s are plane waves. Nonetheless, the random variables $\Phi_{\rho}(\x_\nu)$ have a probability distribution with compact support. It is thus natural to assume that a Gaussian universality assumption holds, i.e., that \autoref{th:ridge} applies to our problem. With this assumption, we obtain the following result
\begin{corollary}\label{cor:beta-rigorous}
Performing kernel ridge regression in a teacher-student scenario with smoothness exponents $\alpha_t$ (teacher) and $\alpha_s$ (student), with ridge $\lambda\sim P^{-\gamma}$ and $0\,{<}\,\gamma\,{<}\,1/2$, under the joint hypotheses of  \autoref{th:scaling} and \autoref{th:ridge}, the exponent governing the asymptotic scaling of the generalization error with $P$ is given by: 
\begin{equation}\label{eq:loc-beta-rigorous}
    \beta = \frac{\gamma \, \alpha_t}{\alpha_s + s},
\end{equation}
\end{corollary}
which does not vanish in the limit $d\rightarrow\infty$. Furthermore, \autoref{eq:loc-beta-rigorous} depends on $s$ and not on $t$ as the prediction of \autoref{eq:loc-prediction}.

\section{Conclusions}

This work shows that efficient learning is possible in high-dimensional settings when the target function admits a compositional structure -- specifically, when it can be expressed as a sum of constituent functions, each depending on a smaller subset of variables of size $t$. By using a kernel that incorporates this compositional structure with $s$-dimensional constituents, the learning curve exponent is independent of $d$ and governed by $s$ if $s\geq t$, optimal for $s\,{=}\,t$ and null if $s\,{<}\,t$. 

In the context of image classification, these results are related to the ``Bag of Words'' interpretation. Consider images composed of $M$ features, each consisting of $t$ adjacent pixels, where classes correspond to specific (possibly overlapping) subsets of such features. If features can be located anywhere in the image, then data lie on a $2M$-dimensional manifold. We expect a one-hidden-layer convolutional network with filter size $s\,{\geq}\,t$ to learn the task efficiently with a learning curve exponent governed by $s$ and independent of $M$. In contrast, a fully-connected network operating in the lazy training regime would exhibit poor generalization in this setting for large $M$ due to the curse of dimensionality.

Our analysis does not account for the hierarchical compositional structure of real data, where large features are composed of smaller sub-features. It is intuitively clear that depth and locality taken together are well-suited for such data structure \cite{bietti2021approximation,poggio2017and}. In the next chapter, we will extend the present framework to this case.

\chapter{The Role of Depth and Spatial Adaptivity}
\label{ch:deepcnn}

\begingroup
\renewcommand{\thefootnote}{}
\footnote{Parts of this chapter have been previously published in:\\
Cagnetta*, F., \textit{Favero*, A.} and Wyart, M., 2024. What can be learnt with wide convolutional neural networks?. Journal of Statistical Mechanics: Theory and Experiment, 2024(10), p.104020.\\
Cagnetta*, F., \textit{Favero*, A.} and Wyart, M., 2023. What Can Be Learnt With Wide Convolutional Neural Networks?. In Proceedings of the 40th International Conference on Machine Learning (ICML), PMLR 202, pp.3347-3379.\\
* These authors contributed equally.}
\addtocounter{footnote}{-1}
\endgroup

The previous chapter established how compositionality and locality allow kernel methods and shallow neural architectures to overcome the curse of dimensionality. We showed that when the target function is a sum of local components, the generalization performance is governed by the size of the receptive field rather than the input dimension. However, real-world data rarely exhibit purely flat compositional structure. Instead, they are widely believed to possess a hierarchical organization, where features at higher levels are composed of sub-features at finer scales. 

Although many works have investigated this idea \citep{biederman1987recognition, poggio2017and, kondor2018generalization, zhou2018building, deza2020hierarchically, kohler2020rate, poggio2020theoretical, schmidt2020nonparametric, finocchio2021posterior, giordano2022inability}, we miss a quantitative understanding of how hierarchy affects the learning curve exponent $\beta$ -- which characterizes the decay rate of the generalization error $\testerr$ with the number of training samples $P$. Specifically, there are no theoretical predictions for $\beta$ in the context of \textit{deep} networks trained on tasks with varying degrees of locality or a truly hierarchical structure.

In this chapter, we address this gap by analyzing deep convolutional neural networks (CNNs) in the overparameterized regime. In this limit, the width of each hidden layer tends to infinity, and the network's output converges to that of a corresponding kernel method \citep{jacot2018neural, lee2019wide}. While real-world deep networks do not generally operate in such a regime, the correspondence with kernel regression provides powerful theoretical tools for computing the decay of the learning curves. Namely, as discussed before, given an infinitely wide neural network, its generalization performance depends on the spectrum of the induced kernel \citep{caponnetto2007optimal, bordelon2020spectrum}. 

The central challenge, then, becomes the characterization of the kernel spectrum, especially for deep CNNs whose corresponding kernels exhibit complex structure and are defined recursively \citep{arora2019exact}. This characterization is the primary outcome of this chapter, along with the subsequent study of generalization in deep CNNs.

\begin{figure*}

    \centering
    {\includegraphics[width=0.4\textwidth]{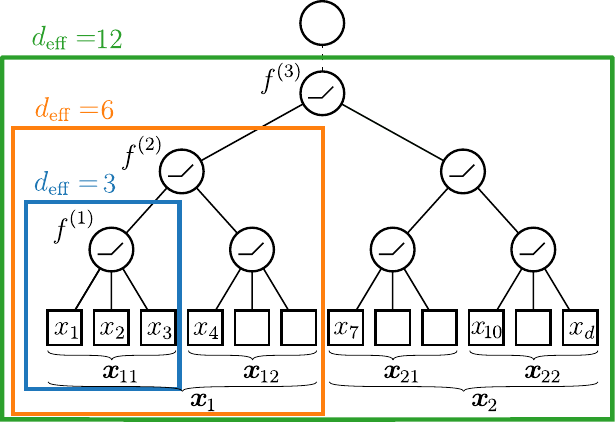}}
    \hspace{1cm}
    {\includegraphics[width=0.3\textwidth]{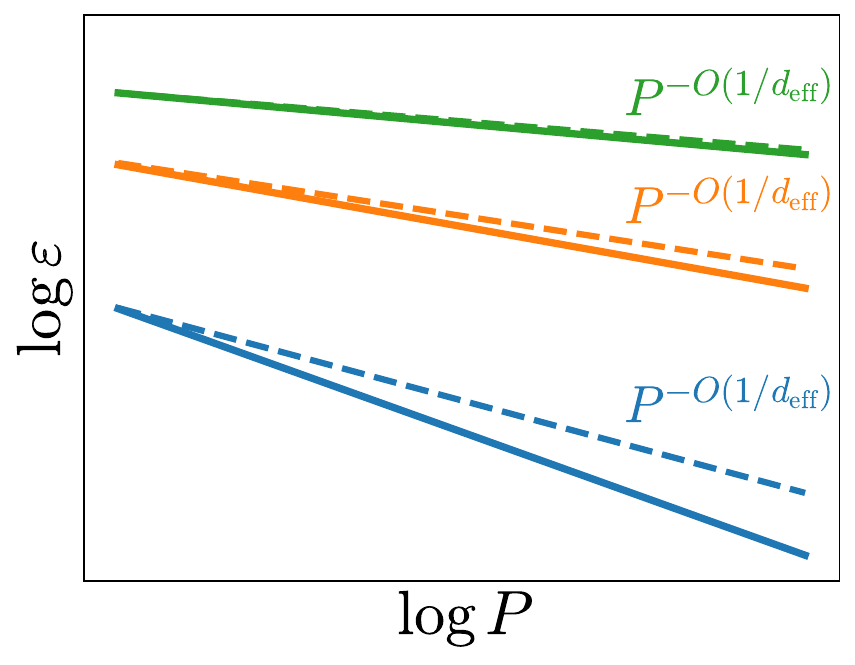}}
    
    \caption[Left: Computational skeleton of a convolutional neural network. Right: Sketches of learning curves $\testerr(P)$ obtained by learning target functions of varying spatial scale with the network on the left.]{\textit{Left:} Computational skeleton of a convolutional neural network of depth $L+1\,{=}\,4$ ($L\,{=}\,3$ hidden layers). The leaves of the graph (squares) correspond to input coordinates, and the root (empty circle) to the output. All other nodes represent (infinitely wide layers of) hidden neurons. We define as `meta-patches' (i.e., patches of patches) the sets of input variables that share a common ancestor node along the tree (such as the squares within each colored rectangle). Each meta-patch coincides with the receptive field of the neuron represented by this common ancestor node, as indicated below the input coordinates. For each hidden layer $l\,{=}\,1,\dots,L$, there is a family of meta-patches having dimensionality $d_{\text{eff}}(l)$. 
    \textit{Right:} Sketches of learning curves $\testerr(P)$ obtained by learning target functions of varying spatial scale with the network on the left. More specifically, the target is a function of a $3$-dimensional patch for the blue curve, a $6$-dimensional patch for the orange curve, and the full input for the green curve. We predict (and confirm empirically) that both the decay of $\testerr$ with $P$ (full lines) and the rigorous upper bound (dashed lines) are controlled by the effective dimensionality of the target.}
    \label{fig:main-msg}
    
\end{figure*}

More specifically, we investigate the generalization properties of deep CNNs with non-overlapping patches and no pooling (defined in \autoref{sec:deep-setup}, see \autoref{fig:main-msg} for an illustration). These networks are trained on a target function $f^*$ via empirical minimization of the mean squared loss. We consider the infinite-width limit of the networks (\autoref{sec:deep-kernels}), where the parameters change infinitesimally over training, thus the trained network coincides with the predictor of kernel regression with the corresponding Neural Tangent Kernel (NTK). Thus, generalization is fully characterized by the spectrum of the integral operator of the kernel: in simple terms, the projections on the eigenfunctions with larger eigenvalues can be learned -- up to fixed error -- with fewer training points (see, e.g., \citet{bach2021learning}).

\paragraph{Spectra (\autoref{th:eig-scaling}).} Due to the network topology, hidden neurons in each layer depend only on a limited subset of the input variables -- referred to as the neuron's receptive field (highlighted by colored boxes in \autoref{fig:main-msg}, left panel). We show that the NTK eigenfunctions of a CNN of depth $L\,{+}\,1$ can be organized into sectors indexed by the layer index $l\,{=}\,1,\dots,L$ (\autoref{th:eig-scaling}). Each sector consists of eigenfunctions depending only on the receptive fields of the neurons of the corresponding hidden layer. 
Denoting by $d_{\text{eff}}(l)$ the size of the receptive fields of neurons in the $l$-th layer, the eigenfunctions of the $l$-th sector are functions of $d_{\text{eff}}(l)$ variables. We analytically characterize the asymptotic decay of the NTK eigenvalues in terms of the polynomial degree of the corresponding eigenfunctions (\autoref{th:eig-scaling}), and show this decay is governed by $d_{\text{eff}}(l)$. Consequently, the eigenfunctions with the largest eigenvalues -- the easiest to learn -- are those that are supported on small, localized subsets of the input and have low polynomial degree. This spectral structure is our main technical contribution, and all of our conclusions follow from it.\looseness=-1

\paragraph{Spatial adaptivity (\autoref{co:adaptivity}).} We leverage this spectral characterization to prove that deep CNNs can adapt to the spatial scale of the target function (\autoref{sec:deep-adaptivity}). By applying rigorous bounds from the theory of kernel ridge regression \citep{caponnetto2007optimal} (reviewed in the first paragraph of \autoref{sec:deep-adaptivity}), we prove that when learning with the kernel of a CNN and optimal regularization, the decay of the error depends on the effective dimensionality of the target. In particular, if the target $f^*$ depends only on a localized group of $d_{\text{eff}}$ adjacent input variables, then the generalization error $\testerr(P) \sim  P^{-\beta}$ with $\beta \geq O(1/d_{\text{eff}})$ (\autoref{co:adaptivity}, see \autoref{fig:main-msg} for a pictorial representation). This result is consistent with non-rigorous results from the replica method for the ridgeless regression case \citep{bordelon2020spectrum,loureiro2021learning} (\autoref{sec:deep-examples}). 

If $d_{\text{eff}}\,{\ll}\,d$, the rates achieved with deep CNNs are much closer to the Bayes-optimal rates -- realized when the architecture is fine-tuned to the structure of the target -- than $\beta=O(1/d)$ obtained with kernel fully-connected network in the lazy training regime. 

Moreover, while deep CNNs perform well on structured targets, we find that hierarchical functions generated by deep CNNs themselves are too rich to be efficiently learnable in high dimensions (\autoref{lemma:curse-hierarchical}). 

Our theoretical predictions are supported by extensive numerical experiments, and we further show that the core conclusions remain valid even when the nonoverlapping patch assumption is relaxed (\autoref{app:deep-extensions}).\looseness=-1

\section{Related work}

\looseness=-1 The generalization properties of shallow CNNs in the kernel regime have been extensively investigated in recent years \citep{bietti2022approximation, favero2021locality, misiakiewicz2021learning, xiao2022synergy, xiao2022eigenspace, geifman2022spectral}. \citet{favero2021locality} -- in the work presented in \autoref{ch:locality} -- and later \citep{misiakiewicz2021learning, xiao2022synergy} showed that shallow CNNs can overcome the curse of dimensionality on compositional, local target functions. However, these models can only approximate functions that depend on single input patches or linear combinations thereof. \citet{bietti2022approximation} extended this line of research by incorporating pooling layers and initiated a study of the role of depth by analyzing the approximation properties of kernels formed by integer powers of a base kernel. This chapter generalizes this line of work by studying CNNs of any depth with nonanalytic activation functions. We find that the depth and nonanalyticity of the resulting kernel play a critical role in shaping the inductive bias of these architectures. This result contrasts sharply with the spectrum of the kernels of deep fully connected networks, whose asymptotic decay is unaffected by depth \citep{bietti2021deep}. Moreover, we extend the analysis of generalization to hierarchically structured target functions -- mirroring the architecture of deep CNNs.

\citet{geifman2022spectral} derived bounds on the eigenvalue spectra of deep CNNs kernels, but considered only filters of size one in the first layer and did not address generalization. In contrast, we allow for general filter sizes and provide tight estimates of the asymptotic behavior of eigenvalues, which allow us to predict generalization rates. 

The work of \citet{xiao2022eigenspace} is the closest to ours, as it also investigates the spectral bias of deep CNNs in the kernel regime. However, it considers a different asymptotic regime, where both the input dimension and the number of training samples tend to infinity. Importantly, it does not characterize the asymptotic decay of the generalization error with the training set size, a central focus of our work.

Finally, \citet{paccolat2021isotropic, malach2021computational, abbe2022merged} studied settings where the target functions depend only on a small subset of the input variables. These works show sample complexity separation results between models operating in the kernel regime and those in the feature regime -- where parameters undergo significant changes during training. In this respect, our results demonstrate that even in the kernel regime, deep CNNs achieve near-optimal performance when the target function depends on a few \textit{adjacent} input variables, i.e., the target function is spatially localized.

\section{Notation and setup}\label{sec:deep-setup}

Our work considers CNNs with nonoverlapping patches and no pooling layers. Although employed in common architectures, these two elements do not affect the conclusions of our study and are not crucial for learning\footnote{To illustrate this point, we trained a modified LeNet architecture with nonoverlapping patches and no pooling layers on CIFAR10, then compared the generalization error with that of a standard LeNet architecture \cite{lecun1998gradient} trained with the same hyperparameters. The modified architecture achieved a test accuracy of $53\%$, reasonably close to the $62\%$ accuracy of the standard architecture. In addition, we show in \autoref{app:deep-extensions} that, although our theory requires nonoverlapping patches, our predictions remain true with overlapping patches.}. These networks are fully characterized by the depth $L\,{+}\,1$ (or number of hidden layers $L$) and a set of filter sizes $\left\lbrace s_l\right\rbrace_l$ (one per hidden layer). We refer to such networks as \emph{hierarchical} CNNs.
\begin{definition}[$L$-hidden-layers hierarchical CNN] \label{def:hierarchical-cnn}
Denote by $\sigma$ the normalized ReLU function, $\sigma(x) = \sqrt{2}\max(0,x)$. For each input $\x\in \mathbb{R}^d$ and $s$ a divisor of $d$, denote by $\x_i$ the $i$-th $s$-dimensional patch of $\x$, $\x_i\,{=}\,(x_{(i-1)\times s+1},\dots,x_{i\times s})$ for all $i\,{=}\,1,\dots,d/s$.\footnote{Notice that all our results can be readily extended to image-like input signals $\lbrace x_{ij}\rbrace_{i,j}$ or tensorial objects with an arbitrary number of indices.} The output of a $L$-hidden-layers hierarchical neural network can be defined recursively as follows. 
\begin{align}
    &f^{(1)}_{h,i}(\x) = \sigma\left( \w^{(1)\top}_{h}\x_{i}\right),\,
    \forall h\in[H_1],\,\forall i\in[p_1];\nonumber\\ 
    &f^{(l)}_{h,i}(\x) = \sigma\left( \frac{1}{\sqrt{H_{l-1}}}\sum_{h'} \frac{\w^{(l)\top}_{h,h'}\left(\bm{f}^{(l-1)}_{h'}\right)_i}{\sqrt{s_l}}\right),\,\nonumber\\ 
    &\forall h\in[H_l],\, i\in[p_l],\, l\in[2,\dots,L];\nonumber\\
    &f(\x) = f^{(L+1)}(\x) = \frac{1}{\sqrt{H_L}} \sum_{h=1}^{H_L}\sum_{i=1}^{p_{L}} \frac{w^{(L+1)}_{h,i}f^{(L)}_{h,i}(\x)}{\sqrt{p_{L}}}.
\end{align}
$H_l$ denotes the width of the $l$-th layer, $s_l$ the filter size ($s_1\,{=}\,s$),
$p_l$ the number of patches ($p_1\,{\equiv}\,p\,{=}\,d/s$).
$\w^{\scriptscriptstyle(1)}_h\in\mathbb{R}^{s_1}$,
$\w^{\scriptscriptstyle(l)}_{h,h'}\in\mathbb{R}^{s_l}$,
$w_{h,i}^{\scriptscriptstyle(L+1)}\in\mathbb{R}$.
\end{definition}
Hierarchical CNNs are best visualized by considering their computational skeleton, i.e., the directed acyclic graph obtained by setting $H_l\,{=}\,1$ $\forall \, l$ (example in \autoref{fig:main-msg}, left, with $L\,{=}\,3$ hidden layers and filter sizes $(s_1,s_2,s_3)\,{=}\,(3,2,2)$). Having nonoverlapping patches, the computational skeleton is an ordered tree, whose root is the output (empty circle at the top of the figure) and the leaves are the input coordinates (squares at the bottom). All the other nodes represent neurons, and all the neurons belonging to the same hidden layer have the same distance from the input nodes. The tree structure highlights that the post-activations $f^{l}_i$ of the $l$-th layer depend only on a subset of the input variables, also known as the \emph{receptive field}. 

Since the first layer of a hierarchical CNN acts on $s_1$-dimensional patches of the input, it is convenient to consider each $d$-dimensional input signal as the concatenation of $p$ $s$-dimensional patches, with $s\,{=}\,s_1$ and $p\times s\,{=}\,d$. We assume that each patch is normalized to $1$,\footnote{We show in \autoref{app:deep-extensions} that our predictions remain true if the inputs are sampled uniformly in the $d$-dimensional hypercube $[0,1]^d$ or from a Gaussian distribution on $\mathbb{R}^d$.} so that the input space is a product of $p$ $s$-dimensional unit spheres (called multisphere in \citet{geifman2022spectral}):
\begin{equation}
    {\sf M}^p\mathbb{S}^{s-1}:=\prod_{i=1}^p \mathbb{S}^{s-1} \subset \mathbb{S}^{d-1}.
\end{equation}
We call a function
on ${\sf M}^p\mathbb{S}^{s-1}$ \emph{localized} if it is constant on at least $1$ of the $p$ patches. In other words, localized functions only depend on some patches of the input. The neurons of the first hidden layer are examples of localized functions, as each of them depends on only one of the $s$-dimensional patches (see the blue rectangle in \autoref{fig:main-msg} for $s\,{=}\,3$).

In general, the receptive field of a neuron in the $l$-th hidden layer with $l\,{>}\,1$ is a group of ${\prod_{l'=2}^{l}} s_{l'}$ adjacent patches (as in the orange rectangle of \autoref{fig:main-msg} for $l\,{=}\,2$, $s_2\,{=}\,2$ or the green rectangle for $l\,{=}\,3$, $s_3\,{=}\,s_2\,{=}\,2$), which we refer to as a \textit{meta-patch}. Due to the correspondence with the receptive fields, each meta-patch is identified with one path on the computational skeleton: the path that connects the output node to the hidden neuron whose receptive field coincides with the meta-patch. If such hidden neuron belongs to the $l$-th hidden layer, the path is specified by a tuple of $L-l\,{+}\,1$ indices, $i_{l+1\to L+1}\,{:=}\, i_{L+1}\dots i_{l+1}$, where each index indicates which branch to select when descending from the root to the neuron node. With this notation, $\x_{i_{l+1}\to i_{L+1}}$ denotes one of the $p_{l}$ meta-patches of size $\prod_{\scriptscriptstyle l'\leq l} s_{l'}$. Because of the normalization of the $s_1$-dimensional patches, i.e., $\x_{i_{2\to L+1}} \allowbreak \in \mathbb{S}^{s_1-1}$, each meta-patch has an \emph{effective dimensionality} which is lower than its size, 
\begin{equation}\label{eq:deep-effective-dim}
   \begin{cases} 
   d_{\text{eff}}(1):=\text{dim}(\x_{i_{2\to L+1}}) =(s_1-1),&\\
   d_{\text{eff}}(l) := \text{dim}(\x_{i_{l+1\to L+1}}) = (s_1-1){\prod_{l'=2}^{l}} s_{l'},
   \end{cases}
\end{equation}
for $l\in[2,\dots,L]$. Localized functions that depend on a specific meta-patch inherit the latter's effective dimensionality. In general, the effective dimensionality of a localized function $f$ coincides with that of the smallest meta-patch that contains all the patches that $f$ depends on.

\section{Hierarchical kernels and their spectra}
\label{sec:deep-kernels}

We turn now to the infinite-width limit $H_l\to\infty$: because of the equivalence with kernel methods, this limit allows us to deduce the generalization properties of the network from the spectrum of a kernel. In this section, we present the kernels corresponding to the hierarchical models of \autoref{def:hierarchical-cnn} and characterize the spectra of the associated integral operators.

We consider specifically two kernels: the \emph{Neural Tangent Kernel} (NTK), corresponding to training all the network parameters \citep{jacot2018neural}; and the \emph{Random Feature Kernel} (RFK), corresponding to training only the weights of the linear output layer \citep{rahimi2007random, daniely2016toward}. In both cases, the kernel reads:
\begin{equation}\label{eq:deep-kernel-trick}
 \mathcal{K}(\x,\y) =  \sum_{\text{trained params }\theta} \partial_\theta f(\x) \partial_\theta f(\y).
\end{equation}
The NTK and RFK of deep CNNs have been derived previously in \cite{arora2019exact}. In \autoref{app:deep-kernel-lemmas} we report the functional forms of these kernels in the case of hierarchical CNNs. These kernels inherit the hierarchical structure of the original architecture and their operations can be visualized again via the tree graph of \autoref{fig:main-msg}. In this case, the leaves represent products between the corresponding elements of two inputs $\x$ and $\y$., i.e., $x_1 y_1$ to $x_d y_d$, and the root the kernel output $\mathcal{K}(\x,\y)$. The output can be built layer by layer by following the same recipe for each node: first, sum the outputs of the previous layer that are connected to the present node, then apply a nonlinear function that depends on the activation function of the network. In particular, for each couple of inputs $\x$ and $\y$ on the multisphere ${\sf M}^p\mathbb{S}^{s-1}$, hierarchical kernels depend on $\x$ and $\y$ via the $p$ dot products between corresponding $s$-dimensional patches of $\x$ and $\y$. As a comparison, \citet{bietti2021deep} showed that the NTK and RFK of a fully-connected network of any depth depend on the full dot product $\x^\top\y$, whereas those of a shallow CNN can be written as the sum of $p$ kernels, each depending on only one of the patch dot products \citep{favero2021locality}.\looseness=-1

Given the kernel, the associated integral operator reads
\begin{equation}\label{eq:deep-kernel-integral}
    \left(T_{\mathcal{K}} f\right)(\x) := \int_{\mathbb{S}^{s-1}} \mathcal{K}(\x,\y) f(\y)dp(\y),
\end{equation}
with $dp(\x)$ denoting the uniform distribution of input points on the multisphere. The spectrum of this operator provides, via Mercer's theorem \citep{mercer1909xvi}, an alternative representation of the kernel $\mathcal{K}(\x,\y)$ and a basis for the space of functions that the kernel can approximate. The asymptotic decay of the eigenvalues, in particular, is crucial for the generalization properties of the kernel, as it will be clarified in \autoref{sec:deep-adaptivity}. Since the input space is a product of $s$-dimensional unit spheres and the kernel depends on the $p$ scalar products between corresponding $s$-dimensional patches of $\x$ and $\y$, the eigenfunctions of $T_{\mathcal{K}}$ are products of spherical harmonics acting on the patches (see \autoref{app:deep-harmonics} for definitions and the relevant background). For the sake of clarity, we limit the discussion in the main paper to the case $s\,{=}\,2$, where, since each patch $\x_i$ is entirely determined by an angle $\theta_i$, the multisphere ${\sf M}^p\mathbb{S}^{s-1}$ reduces to the $p$-dimensional torus and the eigenfunctions to $p$-dimensional plane waves: $e^{i\k^\top\bm{\theta}}$ with $\bm{\theta}\,{:=}\,(\theta_1,\dots,\theta_p)$ and label $\k\,{:=}\,(k_1,\dots,k_p)$. In this case, the eigenvalues coincide with the $p$-dimensional Fourier transform of the kernel $\mathcal{K}\left(\cos{\theta_1},\dots,\cos{\theta_p}\right)$ and the large-$\k$ asymptotics are controlled by the nonanalyticities of the kernel \citep{widom1963asymptotic}. The general case with patches of arbitrary dimension is presented in the appendix.

\begin{theorem}[Spectrum of hierarchical kernels]\label{th:eig-scaling}
Let $T_{\mathcal{K}}$ be the integral operator associated with a $d$-dimensional hierarchical kernel of depth $L+1$, $L\,{>}\,1$ and filter sizes ($s_1,\dots,s_L$) with ${s_1=2}$. The eigenvalues and eigenfunctions of $T_{\mathcal{K}}$ can be organized into $L$ sectors associated with the hidden layers of the kernel/network. For each $1\,{\leq}\,l\,{\leq}\,L$, the $l$-th sector consists of $(\textstyle\prod_{\scriptscriptstyle l'=1}^{\scriptscriptstyle l} s_{l'})$-\emph{local} eigenfunctions: functions of a single meta-patch $\x_{i_{l+1\to L+1}}$ which cannot be written as linear combinations of functions of smaller meta-patches. The labels $\k$ of these eigenfunctions are such that there is a meta-patch $\k_{i_{l+1 \to L+1}}$ of $\k$ with no vanishing sub-meta-patches and all the $k_i$'s outside $\k_{i_{l+1 \to L+1}}$ are $0$ (because the eigenfunction is constant outside $\x_{i_{l+1 \to L+1}}$). The corresponding eigenvalue is degenerate with respect to the location of the meta-patch: we call it $\Lambda^{\scriptscriptstyle(l)}_{\k_{i_{l+1}\to i_{L+1}}}$. When  $\|\k_{i_{l+1 \to L+1}}\|\to\infty$, with $k\,{=}\,\|\k_{i_{l+1 \to L+1}}\|$,
\begin{equation}\label{eq:deep-eig-scaling-2d}
        \Lambda^{(l)}_{\k_{i_{l+1\to L+1}}}
        = \mathcal{C}_{2,l}\, k^{-2\nu -d_{\mathrm{eff}}(l)} + o\left(k^{-2\nu -d_{\mathrm{eff}}(l)}\right),
\end{equation}
with $\nu_{\mathrm{NTK}}=1/2,\, \nu_{\mathrm{RFK}}=3/2$ and $d_{\mathrm{eff}}$ the effective dimensionality of the meta-patches defined in \autoref{eq:deep-effective-dim}. $\mathcal{C}_{2,l}$ is a strictly positive constant for $l\,{\geq}\,2$ whereas for $l\,{=}\,1$ it can take two distinct strictly positive values depending on the parity of $k_{i_{2\to L+1}}$.
\end{theorem}

The proof is in \autoref{app:deep-spectra}, together with the extension to the $s \geq 3$ case (\autoref{th:eig-scaling-app}). It is useful to compare the spectrum in the theorem with the limiting cases of a deep fully connected network and a shallow CNN. In the former case, the spectrum consists only of the $L$-th sector with $p_L\,{=}\,1$ -- the global sector. The eigenvalues decay as $\norm{\k}^{-2\nu -p}$, with $\nu$ depending ultimately on the nonanalyticity of the network activation function (see \citet{bietti2021deep} or \autoref{app:deep-spectra}) and $p\,{=}\,d_{\text{eff}}(L)$ the effective dimensionality of the input. As a result, all eigenfunctions with the same $\norm{\k}$ have the same eigenvalue, even those depending on a subset of the input coordinates. For example, assume that all the components of $\k$ are zero but $k_1$, i.e., the eigenfunction depends only on the first $2$-dimensional patch: the eigenvalue is $O(k_1^{-2\nu-p})$. By contrast, for a hierarchical kernel, the eigenvalue is $O(k_1^{-2\nu-1})$, much larger than the former as $p\,{>}\,1$.\looseness=-1

In the case of a shallow CNN, the spectrum consists only of the first sector, so that each eigenfunction depends only on one of the input patches. In this case, only one of the $\k$ can be non-zero, say $k_1$, and the eigenvalue is $O(k_1^{-2\nu-1})$. However, from \citep{favero2021locality}, a kernel of this kind is only able to approximate functions that depend on one of the input patches or linear combinations of such functions. Instead, for a hierarchical kernel with $p_L\,{=}\,1$, the eigenfunctions of the $L$-th sector are supported on the full input space. Then, if $\Lambda_{\k}\,{>}\,0$ for all $\k$, hierarchical kernels are able to approximate any function on the multisphere, dispensing with the need for fine-tuning the kernel to the structure of the target function.

Overall, given an eigenfunction of a hierarchical kernel, the asymptotic scaling of the corresponding eigenvalue depends on the spatial structure of the eigenfunction support. More specifically, the effective dimensionality of the smallest meta-patch that contains all the variables that the eigenfunction depends on.  In simple terms, the decay of an eigenvalue with $k$ is slower if the associated eigenfunction depends on a few adjacent patches -- but not if the patches are far apart! This is a property of hierarchical architectures that use nonlinear activation functions at all layers. Such a feature disappears if all hidden layers apart from the first have polynomial \citep{bietti2022approximation} or infinitely smooth \citep{azevedo2015eigenvalues, scetbon2021spectral} activation functions or if the kernels are assumed to factorize over patches, as in \citet{geifman2022spectral}.\looseness=-1

\section{Generalization properties and spatial adaptivity}\label{sec:deep-adaptivity}

In this section, we study the implications of the peculiar spectra of hierarchical NTKs and RFKs on the generalization properties of and prove a form of adaptivity to the spatial structure of the target function. We follow the classical analysis of \citet{caponnetto2007optimal} for kernel ridge regression (see \citet{bach2021learning, bietti2022approximation} for a modern treatment) and employ a spectral bias ansatz for the ridgeless limit \citep{bordelon2020spectrum, spigler2020asymptotic}. 

\paragraph{Theory of kernel ridge regression and source-capacity conditions} Given a set of $P$ training points $\left\lbrace (\x_\nu,y_\nu)\right\rbrace_{\nu=1}^P\stackrel{{{\scriptscriptstyle \mathrm{i.i.d.}}}}{\sim} p(\x,y)$ for some probability density function $p(\x,y)$ and a regularization parameter ${\lambda}\,{>}\,0$, the kernel ridge regression estimate of the functional relation between $\x$'s and $y$'s, or \emph{predictor}, is
\begin{equation}\label{eq:deep-predictor}
    \hat{f}_\lambda^P(\x) = \argmin_{f\in\mathcal{H}} \left\lbrace \frac{1}{P}\sum_{\nu=1}^P\left( f(\x_\nu) - y_\nu \right)^2 + \lambda \norm{f}_\mathcal{H} \right\rbrace,
\end{equation}
\looseness=-1  where $\mathcal{H}$ is the Reproducing Kernel Hilbert Space (RKHS) of a (hierarchical) kernel $\mathcal{K}$. If $f(\x)$ denotes the model from which the kernel was obtained via \autoref{eq:deep-kernel-trick}, the space $\mathcal{H}$ is contained in the span of the network features $\left\lbrace \partial_\theta f(\x)\right\rbrace_{\theta}$ in the infinite-width limit. Alternatively, $\mathcal{H}$ can be defined via the kernel's eigenvalues $\Lambda_{\k}$ and eigenfunctions $Y_{\k}$: denoting with $f_{\k}$ the projections of a function $f$ onto the kernel eigenfunctions, then $f$ belongs to $\mathcal{H}$ if it belongs to the span of the eigenfunctions and
\begin{equation}
     \left\lVert f \right\rVert_{\mathcal{H}}^2 = \sum_{\k\geq\bm{0}} (\Lambda_{\k})^{-1} \lvert f_{\k}\rvert^2 < + \infty.
\end{equation}
The performance of the kernel is measured by the generalization error and its expectation over training sets of fixed size $P$ (denoted with $\mathbb{E}_P$)
\begin{align}\label{eq:deep-generalization-real}
    &\testerr( \hat{f}^P_\lambda) = \int d\x dy\, p(\x,y) \left( \hat{f}^P_\lambda(\x)-y\right)^2, \nonumber\\
    &\testerr(\lambda, P) =\mathbb{E}_P\left[ \testerr(\hat{f}^P_\lambda)\right],
\end{align}
or the \emph{excess} generalization error, obtained by subtracting from $\testerr(\lambda, P)$ the error of the optimal predictor $f^*(\x)\,{=}\,\textstyle \int dy\, p(\x,y) y$. The decay of the error with $P$ can be controlled via two exponents, depending on the details of the kernel and the target function. Specifically, if $\alpha\,{\geq}\,1$ and $r\,{\geq}\,1-1/\alpha$ satisfy the following conditions,
\begin{align}\label{eq:deep-source-capacity}
    &\text{capacity: }\text{Tr}\left(\mathcal{T}_{\mathcal{K}}^{1/\alpha}\right) = \sum_{\k\geq\bm{0}} (\Lambda_{\k})^{1/\alpha} < +\infty,\nonumber\\
    &\text{source: }\norm{T_{\mathcal{K}}^{\frac{1-r}{2}}f^*}_{\mathcal{H}}^2 = \sum_{\k\geq\bm{0}} (\Lambda_{\k})^{-r} \lvert f^*_{\k}\rvert^2 < +\infty,
\end{align}
then, by choosing a $P$-dependent regularization parameter $\lambda_P\sim  P^{-{\alpha}/{(\alpha r + 1)}}$, one gets the following bound on generalization \citep{caponnetto2007optimal}:
\begin{equation}\label{eq:deep-generalization-bound}
    \testerr(\lambda_P, P)-\testerr(f^*) \leq \mathcal{C}' P^{-\frac{\alpha r}{\alpha r + 1}}.
\end{equation}

\paragraph{Spectral bias ansatz} The bound above is actually tight in the noisy setting, for instance when having labels $y_\nu\,{=}\,f^*(\x_\nu) + \xi_\nu$ with $\xi_\nu$ Gaussian. In a noiseless problem where $y_\nu\,{=}\,f^*(\x_\nu)$, one expects to find the best performances in the ridgeless limit $\lambda\to0$, so that the rate of \autoref{eq:deep-generalization-bound} is only an upper bound. In the ridgeless case -- where the correspondence between kernel methods and infinitely-wide neural networks actually holds -- there are unfortunately no rigorous results for the decay of the generalization error. Therefore, we provide a heuristic derivation of the error decay based on a spectral bias ansatz. Consider the projections of the target function $f^*$ on the eigenfunctions of the kernel $Y_{\k}$ ($f^*_{\k}$) \footnote{We are again limiting the presentation to the case $s\,{=}\,2$ but the extension to the general case is immediate.} and assume that kernel methods learn only the $P$ projections corresponding to the highest eigenvalues. Then, if the decay of $f^*_{\k}$ with $\k$ is sufficiently slow, one has (recall that both $\lambda$ and $\testerr(f^*)$ vanish in this setting)
\begin{equation}\label{eq:deep-spectralbias}
    \testerr(P) \sim \sum_{\k \text{ s.t. } \Lambda_{\k}<\Lambda(P)} \lvert f^*_{\k}\rvert^2,
\end{equation}
with $\Lambda(P)$ the value of the $P$-th largest eigenvalue of the kernel. This result can be derived using the replica method of statistical physics (see \citet{canatar2021spectral, loureiro2021learning, tomasini22failure} and \autoref{app:deep-spectralbias}) or by assuming that input points lie on a lattice \citep{spigler2020asymptotic}.\looseness=-1

These two approaches rely on the very same features of the problem, namely the asymptotic decay of $\Lambda_{\k}$ and $\lvert f^*_{\k}\rvert^2$ -- see also \citet{cui2021generalization}. For instance, the capacity condition depends only on the kernel spectrum: $\alpha\geq1$ since $\text{Tr}\left(\mathcal{T}_{K}\right)$ is finite \citep{scholkopf2002learning}; the specific value is determined by the decay of the ordered eigenvalues with their rank, which in turn depends on the scaling of $\Lambda_{\k}$ with $\k$. Similarly, the power-law decay of the ordered eigenvalues with the rank determines the scaling of the $P$-th largest eigenvalue, $\Lambda(P) \sim  P^{-\alpha}$. The source condition characterizes the regularity of the target function relative to the kernel and depends explicitly on the decay of $\lvert f^*_{\k}\rvert^2$ with $\k$, as does the right-hand side of \autoref{eq:deep-spectralbias}. This condition was used by \citet{bach2021learning} to prove that kernel methods are adaptive to the smoothness of the target function: the projections of smoother targets on the eigenfunctions display a faster decay with $\k$, thereby allowing to choose a larger $r$ and leading to better generalization performances. The following corollary of \autoref{th:eig-scaling} (proof and extension to $s_1\,{\geq}\,3$ presented in \autoref{app:deep-minimax}, \autoref{co:adaptivity-app}) shows that, since the spectrum can be partitioned as in \autoref{th:eig-scaling}, hierarchical kernels display adaptivity to targets which depend only on a subset of the input variables. Specific examples of bounds are considered in \autoref{sec:deep-examples}.

\begin{corollary}[Adaptivity to spatial structure]\label{co:adaptivity}
Let $T_{\mathcal{K}}$ be the integral operator of the kernel of a hierarchical deep CNN as in \autoref{th:eig-scaling} with $s\,{=}\,2$. Then: \emph{i)} the \emph{capacity} exponent $\alpha$ is controlled by the largest sector of the spectrum, i.e.,\looseness=-1 
\begin{equation}
  {\mathrm{ Tr}}\left(\mathcal{T}_{\mathcal{K}}^{1/\alpha}\right) < +\infty \Leftrightarrow \alpha < 1 + 2\nu/ d_{\mathrm{eff}}(L);
\end{equation}
\emph{ii)} the \emph{source} exponent $r$ is controlled by the structure of the target function $f^*$, i.e., if there is $l\,{\leq}\,L$ such that $f^*$ depends only on some meta-patch $\x_{i_{l+1\to L+1}}$, then only the first $l$ sectors of the spectrum contribute to the source condition, i.e., $\norm{T_{\mathcal{K}}^{\frac{1-r}{2}}f^*}_{\mathcal{H}}^2$ reads
\begin{equation}\label{eq:deep-adaptivity}
  \sum_{l'=1}^l \sum_{i_{l'+1\to L+1}}\sum_{\substack{\k_{i_{l'+1\to L+1}}}}\left(\Lambda^{(l')}_{\k_{i_{l'+1\to L+1}}}\right)^{-r} \left\lvert f^*_{\k_{i_{l'+1\to L+1}}} \right\rvert^2.
\end{equation}
The same holds if $f^*$ is a linear combination of such functions.\looseness=-1

As a result, when $d_{\mathrm{eff}}(L)$ is large and $\alpha\to 1$, the decay of the error is controlled by the effective dimensionality of the target $d_{\mathrm{eff}}(l)$.
\end{corollary}

\section{Examples and experiments}\label{sec:deep-examples}

\paragraph{Source-capacity bound}
Consider a target function $f^*$ which only depends on the meta-patch $\x_{i_{l+1\to L+1}}$ as in \autoref{co:adaptivity}. Combining the source condition~(\autoref{eq:deep-adaptivity}) with the asymptotic scaling of eigenvalues~(\autoref{eq:deep-eig-scaling-2d}), we get
\begin{equation}\label{eq:deep-adaptivity-2d}
    \norm{T_{\mathcal{K}}^{\frac{1-r}{2}}f^*}_{\mathcal{H}}^2 < +\infty \;\Leftrightarrow\; \sum_{\k} \|\k\|^{r(2\nu + d_\text{eff}(l))} \left\lvert f^*_{\k}\right\rvert^2 < +\infty,
\end{equation}
where $\nu=1/2$ ($3/2$) for the NTK (RFK) and $\k$ denotes the meta-patch $\k_{i_{l+1\to L+1}}$ without the subscript to ease notation. Since the eigenvalues depend on the norm of $\k$, \autoref{eq:deep-adaptivity-2d} is equivalent to a finite-norm condition for all the derivatives of $f^*$ up to order $m\,{<}\,r\,(2\nu + d_{\text{eff}}(l))/2$, $\|\Delta^{m/2} f^*\|^2=\sum_{\k}\|\k\|^{2m} |f^*_{\k}|^2\,{<}\,+\infty$ with $\Delta$ denoting the Laplace operator. As a result, if $f^*$ has derivatives of finite norm up to the $m$-th, then the source exponent can be tuned to $r = 2m/(2\nu+d_{\text{eff}}(l))$, inversely proportional to the effective dimensionality of $f^*$. Since the exponent on the right-hand side of \autoref{eq:deep-generalization-bound} is an increasing function of $r$, the smaller the effective dimensionality of $f^*$, the faster the decay of the error -- hence hierarchical kernels are adaptive to the spatial structure of $f^*$.
In particular, the following generalization bound holds.
\begin{corollary}[generalization bound for hierarchical kernels]
Let $\mathcal{K}$ be the kernel of a deep hierarchical CNN with $s\,{=}\,2$. Let $f^*$ be a function depending only on a meta-patch $\x_{i_{l+1\to L+1}}$ or a linear combination of such functions. Furthermore, assume $f^*$ has finite-norm derivatives up to order $m$, i.e., $\|\Delta^{m/2} f^*\|^2 < +\infty$. Then, there exists a constant $\mathcal{C}'>0$ such that optimally-regularized regression with $\mathcal{K}$ achieves $\testerr(\lambda_P, P)-\testerr(f^*) \leq  \mathcal{C}' P^{-\beta}$ with\looseness=-1
\begin{equation}\label{eq:deep-generalization-smoothness}
     \beta = \frac{2m\,(2\nu+d_{\mathrm{eff}}(L))}{2m\,(2\nu+d_{\mathrm{eff}}(L))+(2\nu+d_{\mathrm{eff}}(l))\,d_{\mathrm{eff}}(L)}.
\end{equation}
\end{corollary}
As an illustration, let us consider the case $p_L\,{=}\,1$ and $d_{\mathrm{eff}}(L)\,{=}\,p\,\allowbreak{=}\,d/2$ (the number of two-dimensional patches). Remarkably, even when $p\,{\gg}\,1$, if $f^*$ depends only on a finite-dimensional meta-patch (or is a sum of such functions) the exponent $\beta$ in \autoref{eq:deep-generalization-smoothness} converges to the finite value $2m/(2(m+\nu) + d_{\mathrm{eff}}(l))$. In stark contrast, using a fully-connected kernel to learn the same target results in 
$\beta\,{=}\,2m/(2m + p)$ -- vanishing as $1/p$ when $p\,{\gg}\,1$, thus cursed by dimensionality.

\paragraph{Rates from spectral bias ansatz} The same picture emerges when estimating the decay of the error from \autoref{eq:deep-spectralbias}. $\Lambda(P)\sim  P^{-\alpha}$, whereas $\sum_{\k}\|\k\|^{2m} |f^*_{\k}|^2\,{<}\,+\infty$ implies $|f^*_{\k}|^2 \lesssim \| \k\|^{-2 m - d_{\mathrm{eff}}(l)}$ for a target supported on a $d_{\mathrm{eff}}(l)$-dimensional meta-patch. Plugging such decays in \autoref{eq:deep-spectralbias} we obtain (details in \autoref{ssec:sb-app})
\begin{equation}\label{eq:deep-t-depth-one-s-depth-two}
    \testerr(P) \sim  P^{-\beta} \text{ with }\beta = \frac{2m}{2\nu+d_{\text{eff}}(l)} \frac{2\nu+d_{\mathrm{eff}}(L)}{d_{\mathrm{eff}}(L)}.
\end{equation}
Again, with $p_L\,{=}\,1$ and $d_{\mathrm{eff}}(L)\,{=}\,p$, the exponent remains finite for $p\,{\gg}\,1$. Notice that we recover the results of \autoref{ch:locality} by using a shallow local kernel if the target is supported on $s$-dimensional patches. These results show that hierarchical kernels play significantly better with the approximation-estimation trade-off than shallow local kernels, as they are able to approximate global functions of the input while not being cursed when the target function has a local structure.\looseness=-1

\begin{figure*}[ht!]

    \centering
    {\includegraphics[width=0.48\textwidth]{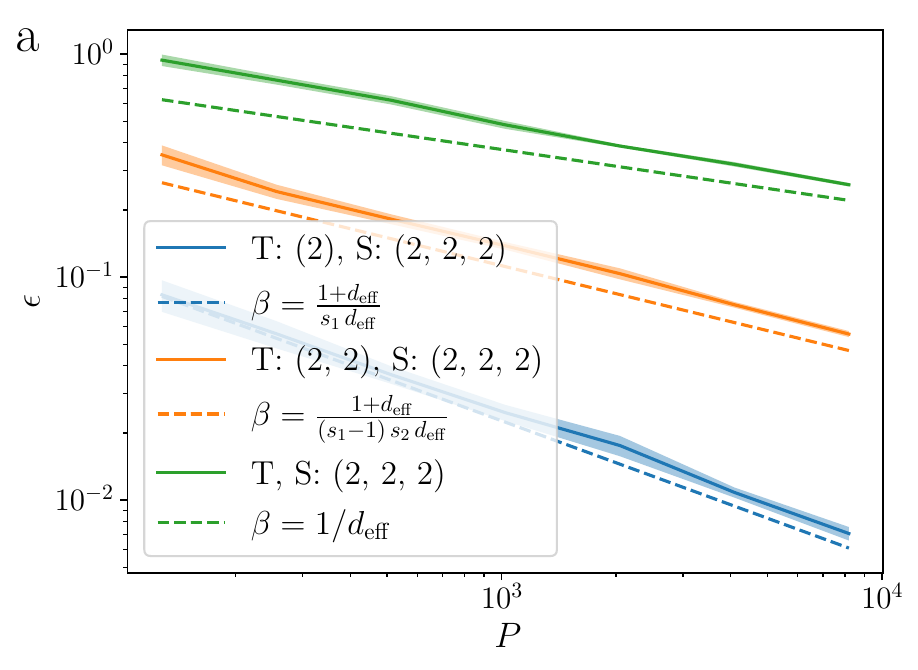}} 
    {\includegraphics[width=0.48\textwidth]{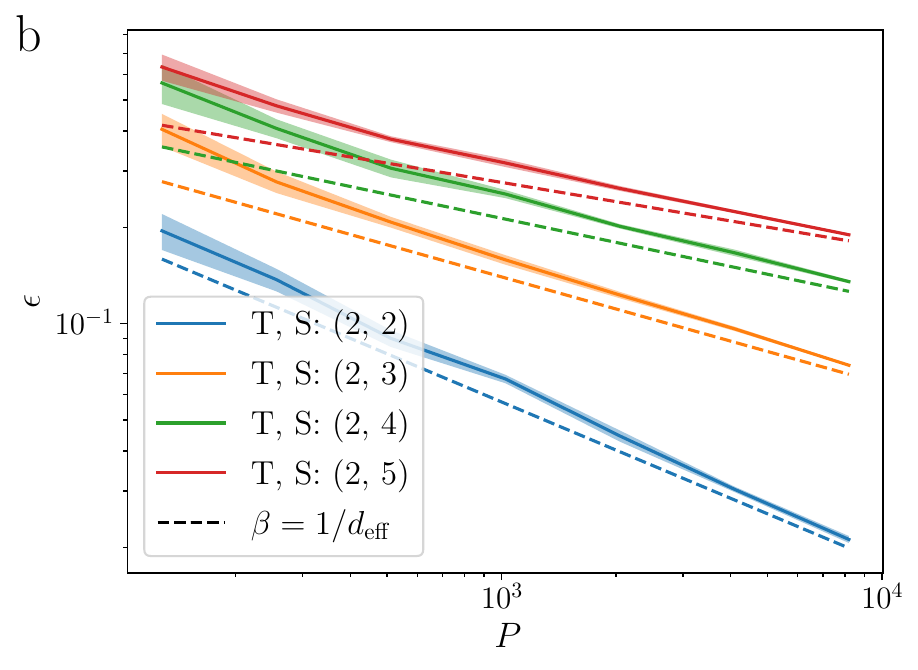}}
    
    \caption{Learning curves for deep convolutional NTKs in a teacher-student setting. \textit{(a)} Depth-four student learning depth-two, depth-three, and depth-four teachers. \textit{(b)} Depth-three models cursed by the effective input dimensionality $d_{\mathrm{eff}}(L)$. The numbers inside brackets are the sequence of filter sizes of the kernels. Solid lines are the results of experiments averaged over 16 realizations with the shaded areas representing the empirical standard deviations. The predicted asymptotic scaling $\testerr \sim  P^{-\beta}$ are reported as dashed lines. Details on the numerical experiments are reported in \autoref{app:deep-numerics}.}
        
    \label{fig:learning_curves_main}
    
\end{figure*}

\paragraph{Numerical experiments} We test our predictions by training a hierarchical kernel (\textit{student}) on a random Gaussian function with zero mean and covariance given by another hierarchical kernel (\textit{teacher}). A learning problem is fully specified by the depths, sets of filter sizes, and smoothness exponents $\nu$ of teacher and student kernels. In particular, the depth and the set of filter sizes of the teacher kernel control the effective dimension of the target function. \autoref{fig:learning_curves_main} shows the learning curves (solid lines) together with the predictions from \autoref{eq:deep-t-depth-one-s-depth-two} (dashed lines), confirming the picture emerging from our calculations. Panel (a) of \autoref{fig:learning_curves_main} shows a depth-four student learning depth-two, depth-three, and depth-four teachers. This student is not cursed in the first two cases and is cursed in the third one, which corresponds to a global target function. Panel (b) illustrates the curse of dimensionality with the effective input dimension $d_{\mathrm{eff}}(L)$ by comparing the learning curves of depth-three students learning global target functions with an increasing number of variables. All our simulations are in excellent agreement with the predictions of \autoref{eq:deep-t-depth-one-s-depth-two}. The bounds coming from \autoref{eq:deep-generalization-smoothness} would display a slightly slower decay, as sketched in \autoref{fig:main-msg}, right panel. All the details of numerical experiments are reported in \autoref{app:deep-numerics}, together with a comparison between the ridgeless and optimally-regularized cases~(\autoref{fig:noise_app}) and additional results for: $s_1\geq 3$~(\autoref{fig:ternary_app}); kernels with overlapping patches (\autoref{fig:overlap}); different input spaces (\autoref{fig:normalizations}) and the CIFAR-10 dataset (\autoref{fig:real_data}).

Notice that when the teacher kernel is a hierarchical RFK, the target is equivalent to the output of a randomly-initialized, infinitely-wide CNN \cite{novak2019bayesian}. Although this target is highly structured, it leads to the same rate obtained for a global non-hierarchical target:
\begin{lemma}[Curse of dimensionality for hierarchical targets]\label{lemma:curse-hierarchical}
The problem of regression of the output of a randomly-initialised and infinitely-wide hierarchical network suffers from the curse of dimensionality, in the sense that no methods using $P$ examples can achieve a generalization error decaying faster than $ P^{-\beta}$ with $\beta\,{=}\,3/d_{\mathrm{eff}}(L)$.
\end{lemma}
This lemma builds on \emph{i)} the aforementioned equivalence of infinitely-wide networks with Gaussian random processes and \emph{ii)} the equivalence of the predictors of kernel ridgeless regression and Bayesian inference. More specifically, since, by \emph{i)}, the target function is to a Gaussian process, the optimal method to learn it is Bayesian inference with a Gaussian prior having the same covariance as the target \citep{kanagawa2018gaussian}. Therefore, by \emph{ii)}, the rate achieved by a kernel method using the target's covariance kernel is also optimal. From \autoref{eq:deep-t-depth-one-s-depth-two} with $l\,{=}\,L$ and $m\,{=}\,\nu\,{=}\,3/2$, the optimal rate is $ P^{-3/d_{\mathrm{eff}}(L)}$, cursed by dimensionality since $d_{\mathrm{eff}}(L)$ is the full input space dimension. We conclude that, despite their intrinsically hierarchical structure, these targets cannot be good models of learnable tasks.

\section{Conclusions}

We have proved that deep CNNs can adapt to the spatial scale of the target function, thus beating the curse of dimensionality if the target depends only on local groups of variables. Yet, if considered as `teachers', they generate functions that cannot be learned efficiently in high dimensions, even in the Bayes-optimal setting where the student is matched to the teacher. Thus, the architectures we considered are not good models of the hierarchical structure of real data, which are efficiently learnable. 

Enforcing a stronger notion of compositionality is an interesting endeavor for the future. Following \citet{poggio2017and}, one may consider a much smaller family of functions of the form, with the notation of \autoref{fig:main-msg}, 
\begin{equation}\label{eq:deep-compostional-target}
f^*(\x_1) = g(h_1(\x_{11}),\, h_2(\x_{12}))
\end{equation}
where, for instance, $g$, $h_1$, and $h_2$ are scalar functions.
From an information theory viewpoint, \citet{schmidt2020nonparametric, finocchio2021posterior} showed that it is possible to learn such functions efficiently. However, these arguments do not provide guarantees for any practical algorithm, such as stochastic gradient descent. Moreover, preliminary results (not shown) assuming that the functions $g$ and $h$ are random Gaussian functions suggest that these tasks are not learnable efficiently by a hierarchical CNN in the kernel regime -- see also \citet{giordano2022inability}. It is unclear whether this remains true when the networks closely resemble the structure of \autoref{eq:deep-compostional-target} as in \citet{poggio2017and}, or when the networks are trained in a regime where features can be learned from data. Recently, for instance, \citet{ingrosso2022data} have observed that under certain conditions locality can be learned from scratch. It is not clear whether compositionality can also be learned, beyond some very stylized settings \citep{abbe2022merged}.

Finally, another direction to explore is the stability of the task toward smooth transformations or diffeomorphisms. This form of stability has been proposed as a key element to understanding how the curse of dimensionality is beaten for image datasets \citep{bruna2013invariant, petrini2021relative}. Such a property can be enforced with pooling operations \citep{bietti2019inductive,bietti2021sample}; therefore, diagonalizing the NTK in this case as well would be of high interest.\looseness=-1

\cleardoublepage

\ctparttext{\bigskip \bigskip \begin{flushright}{\slshape
Art does not reproduce the visible; rather, it makes visible.} \\ \medskip
--- Paul Klee
\end{flushright}
}

\part{Statistical Mechanics of Diffusion Models} 

\chapter{A phase Transition in the Diffusion Process} 

\label{ch:phasetransition}

\begingroup
\renewcommand{\thefootnote}{}
\footnote{Parts of this chapter have been previously published in:\\
Sclocchi, A., \textit{Favero, A.} and Wyart, M., 2025. A Phase Transition in Diffusion Models Reveals the Hierarchical Nature of Data. In Proceedings of the National Academy of Sciences (PNAS), 122 (1), e2408799121.
}
\addtocounter{footnote}{-1}
\endgroup

In this part of the thesis, we study compositionality in the context of deep \textit{generative models}, specifically diffusion models, and explore how they might leverage the compositional and hierarchical nature of data. Do these models learn to generate novel, complex data by composing and assembling simpler features learned from examples, much like a writer combines words to form sentences?
While this idea is intuitive, its formal and empirical validation presents a significant scientific challenge.

This chapters present evidence that diffusion models \cite{sohl2015deep,ho2020denoising,song2019generative, song2020score}
do indeed operate compositionally, generating images by assembling features across different hierarchical levels during the reverse diffusion process. We first provide quantitative, empirical evidence of these compositional effects in the denoising diffusion process of natural images. We then develop a theoretical framework, based on synthetic compositional and hierarchically structured models of data, to explain these observations.

Diffusion models operate by progressively adding noise to data as time increases (the forward process) and then learning to reverse this process to generate new samples (the backward process). 
By adding a finite amount of noise to an image and then denoising, we observe that: 
\textit{(i)} at low noise levels, only low-level features of the image are modified; \textit{(ii)} at a critical noise threshold, the probability that the denoised output belongs to the same class of the original datum drops sharply to the level of chance; \textit{(iii)}  beyond this critical point, high-level features are lost, but low-level features from the original image can surprisingly persist and recombine to compose elements of entirely new classes. 

While the first observation is intuitive and has been noted previously \cite{ho2020denoising}, the sharp phase transition and the subsequent recombination of elementary features are surprising.
We will demonstrate that these phenomena can be precisely theoretically explained using synthetic generative models of data with a built-in hierarchical and compositional structure, inspired by concepts from formal grammars and statistical physics \cite{cagnetta2023deep}. Within this framework, we show that the Bayes-optimal denoising process can be computed exactly using belief propagation on tree-like graphs. This analysis predicts and explains both the phase transition in the class (observation \textit{(ii)}) and the recombination of low-level features to generate new data before and after this transition (observations \textit{(i)} and \textit{(iii)}).

In summary, our findings demonstrate that diffusion models interact with data in a hierarchical manner, operating at different levels of abstraction at different time scales in the diffusion process. This provides strong evidence for the compositional nature of generation in these models. Furthermore, our work champions the use of hierarchical generative models as powerful theoretical tools for studying deep learning systems.

\begin{figure}
    \centering
    \includegraphics[width=.5\columnwidth]{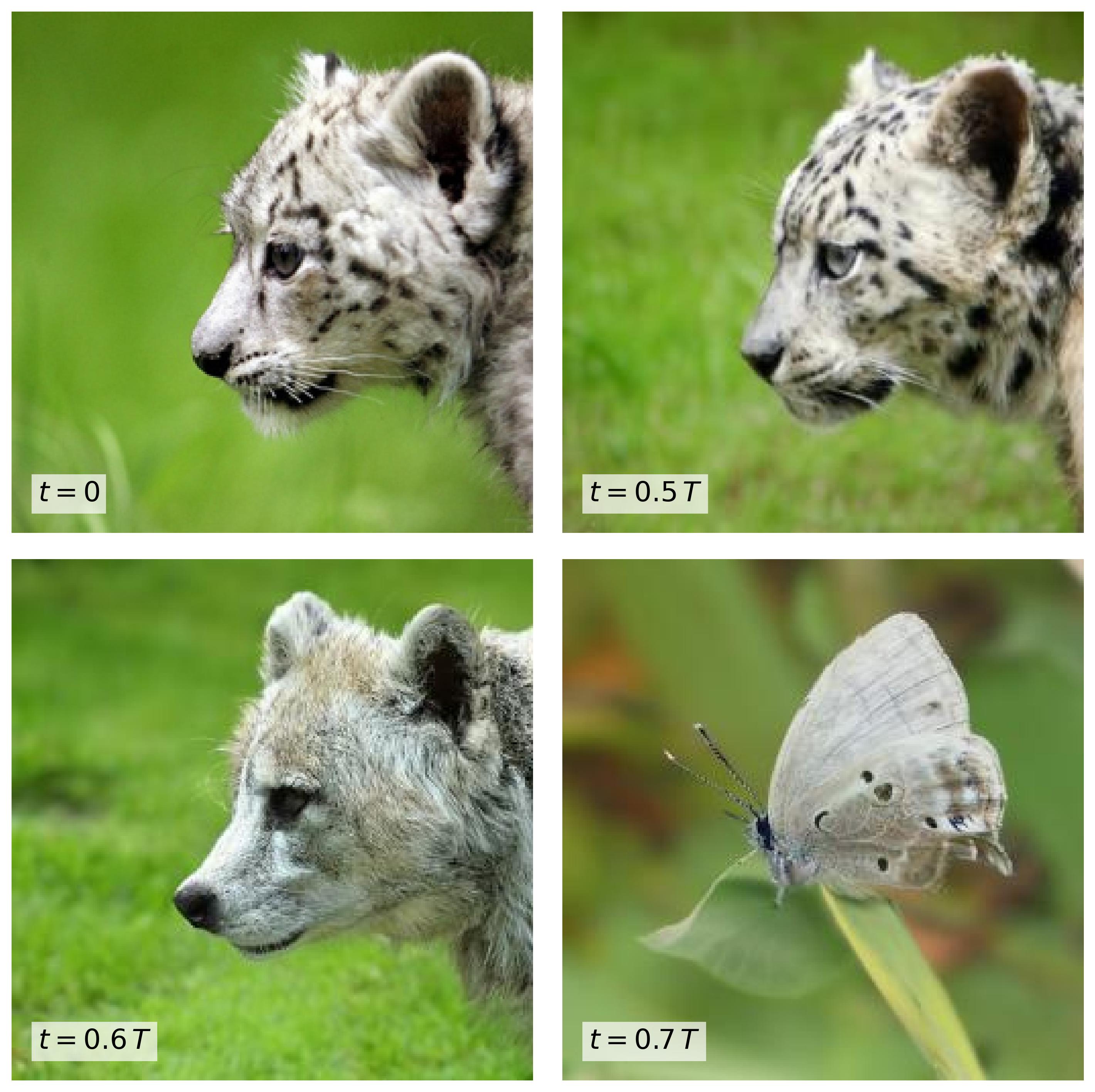}
    \caption{\textbf{Illustration of forward-backward experiments.} Images generated by a denoising diffusion probabilistic model starting from the top-left image and inverting the dynamics at different times $t$. $T$ corresponds to the time scale when the forward diffusion process converges to an isotropic Gaussian distribution. At small $t$, the class of the generated image remains unchanged, with only alterations of low-level features, such as the eyes of the leopard. After a characteristic time $t$, the class undergoes a phase transition and changes. However, some low-level attributes of the original image are retained to compose the new image. For instance, the wolf is composed of eyes, nose, and ears similar to those of the leopard, and the butterfly inherits its colors and black spots.}
    \label{fig:splash-figure}
\end{figure}

In particular, in this chapter, we perform a systematic study of the denoising diffusion dynamics on ImageNet. 
We invert the noising process at some time $t$, leading to novel noiseless images. We then analyze how the representation of state-of-the-art convolutional architectures changes between the initial and newly generated images as a function of both time $t$ and depth of the representation. This analysis reveals the presence of a sharp transition in the class at a given time or noise level. Importantly, at times beyond the transition, when the class has changed, we find that the generated images may still be composed of low-level features of the original image.

\looseness=-1 To model theoretically the compositional structure of images, we consider hierarchical generative models of data where the structure of the latent variables is tree-like. We use belief propagation to study the optimal denoising dynamics for such data and obtain the evolution of latent variables' probabilities for different levels of corruption noise. 
In the limit of a large tree depth, this analysis reveals a phase transition for the probability of reconstructing the root node of the tree -- which represents the class label of a data point -- at a specific noise threshold. Conversely, the probability of reconstructing low-level latent variables evolves smoothly throughout the denoising diffusion process. Thus, after the transition, low-level features of the original datum may persist in composing a generated element of a new class, as we empirically observe in ImageNet.  Finally, we show numerically that the dynamics of the latent variables is reflected in the hidden representation of deep networks previously trained on a supervised classification task on these data.

\section{Related work}

\paragraph{Forward-backward protocol in diffusion-based models} 

\citet{ho2020denoising} introduced the ``forward-backward" protocol to probe diffusion-based models, whereby an image with a controlled level of noise is then denoised using a reverse-time diffusion process. It led to the observation that ``when the noise is small, all but fine details are preserved, and when it is large, only large-scale features are preserved''. Although our work agrees with the first part of the statement, it disagrees with the second.
Our work also provides a systematic quantification of the effects of forward-backward experiments, going beyond
qualitative observations based on individual images as in \cite{ho2020denoising}. Specifically, we introduce quantitative observables that characterize changes in the latent features of images and perform extensive experiments with state-of-the-art models, averaging results over $10^5$ ImageNet samples. Such quantification is key to connecting with theory.
The forward-backward protocol was also studied in \citet{behjoo2023u} to speed up the generation process of images.

\paragraph{Theory of diffusion models}
Most of the theoretical work on diffusion models considers simple models of data. Under mild assumptions on the data distribution, diffusion models exhibit a sample complexity that scales exponentially with the data dimension \cite{block2020generative,oko2023diffusion}. This curse of dimensionality can be mitigated through stronger distributional assumptions, such as considering data lying within a low-dimensional latent subspace \cite{de2022convergence,chen2023score,yuan2023reward},  Gaussian mixture models \cite{biroli2023generative,shah2023learning, Cui2023AnalysisOL}, graphical models \cite{mei2023deep}, or data distributions that can be factorized across scales \cite{kadkhodaie2023learning}. 
For multimodal distributions such as Gaussian mixtures, the backward dynamics exhibits a cross-over time when it concentrates toward one of the modes  \cite{biroli2023generative, ambrogioni2023statistical, raya2024spontaneous}. This cross-over is similar to our observation \textit{(ii)} above if these modes are interpreted as classes.
As demonstrated in \autoref{app:gaussian-mixture}, such models of data cannot reproduce our salient predictions and observations.
Closer to our work, \citet{okawa2023compositional} considers synthetic compositional data to empirically show how diffusion models learn to generalize by composing different concepts. In contrast, we study data that are not only compositional but also hierarchically structured and make quantitative predictions on how diffusion models compose features at different scales.  

\paragraph{Hierarchical models of natural data}
Generative models of data have a long history of describing the structure of language and image data.
In linguistics, formal grammars describe the syntactic structure of a language through a hierarchical tree graph \cite{rozenberg_handbook_1997}.
Similar ideas have been explored to decompose visual scenes hierarchically into objects, parts, and primitives \cite{zhu2007stochastic} and have been formalized in pattern theory \cite{grenander1996elements}. These hierarchical models led to practical algorithms for semantic segmentation and scene understanding, as illustrated in, e.g., \cite{jin2006context, siskind2007spatial, li2009towards}. Recent works propose a hierarchical decomposition of images, in which latent variables are wavelet coefficients at different scales 
\cite{marchand2022wavelet,
kadkhodaie2023learning}. In this case, the graph is not tree-like \cite{
kadkhodaie2023learning} -- a conclusion that could stem from the specific choice of latent variables.

\paragraph{Hierarchical models in machine learning theory}
More recently, generative models of data received attention in the context of machine learning theory.
In supervised learning, deep networks can represent hierarchical tasks more efficiently than shallow networks~\cite{poggio2017why} and can efficiently learn them from an information theory viewpoint~\cite{schmidt2020nonparametric}.
For hierarchical models of data, correlations between the input data and the task are critical for learning~\cite{mossel2016deep,shalev2017failures,malach2018provably,malach2020implications} and the representations learned by neural networks with gradient descent reflect the hidden latent variables of such models both in Convolutional Neural Networks (CNNs) \cite{cagnetta2023deep} and transformers \cite{allen2023physics}. In this chapter, we use these hierarchical generative models of data to study the denoising dynamics of diffusion models theoretically.

\section{Diffusion models and feature hierarchies}\label{sec:diffusion}

\begin{figure*}
    \centering
    \includegraphics[width=.45\columnwidth]{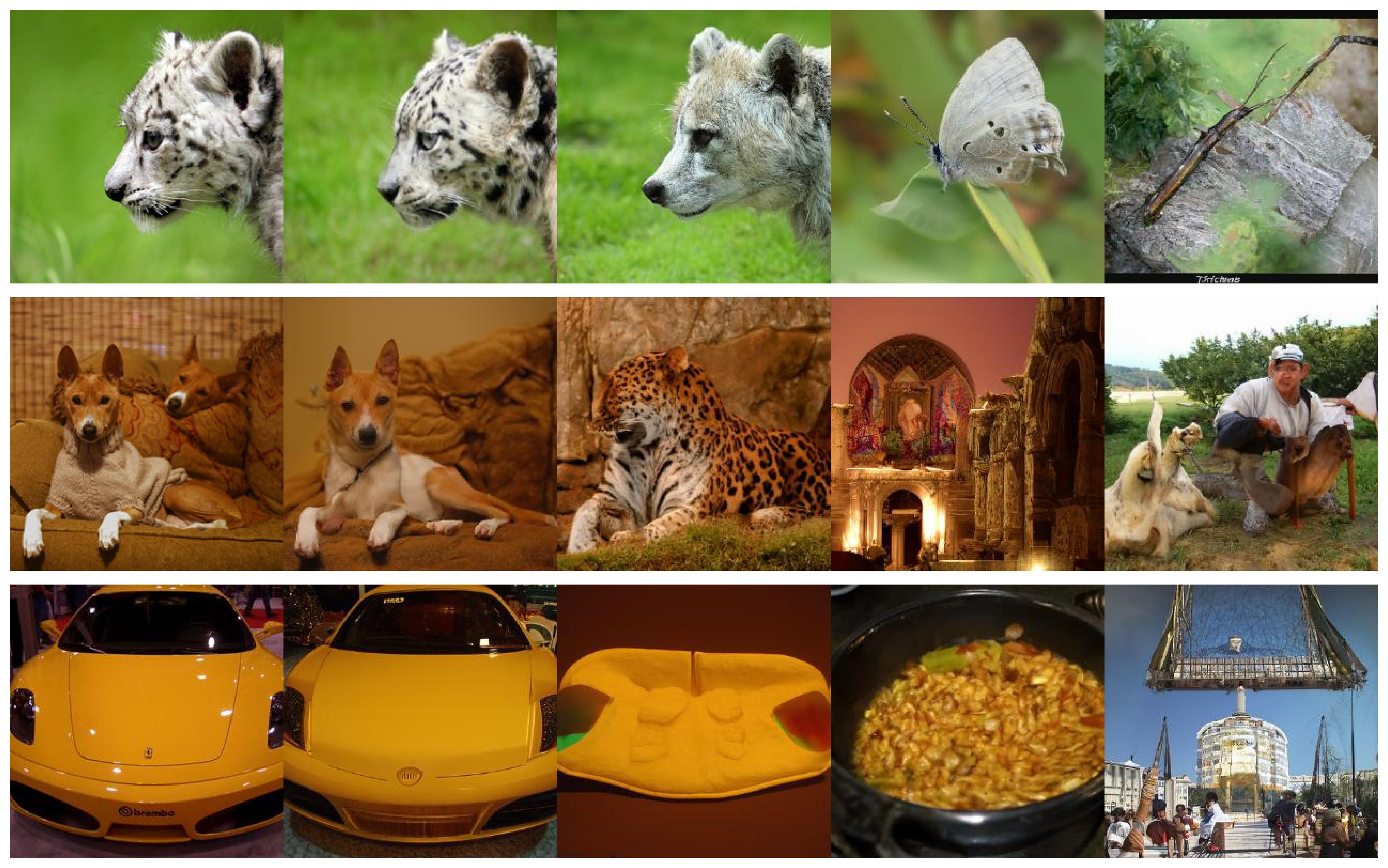}
    \includegraphics[width=.45\columnwidth]{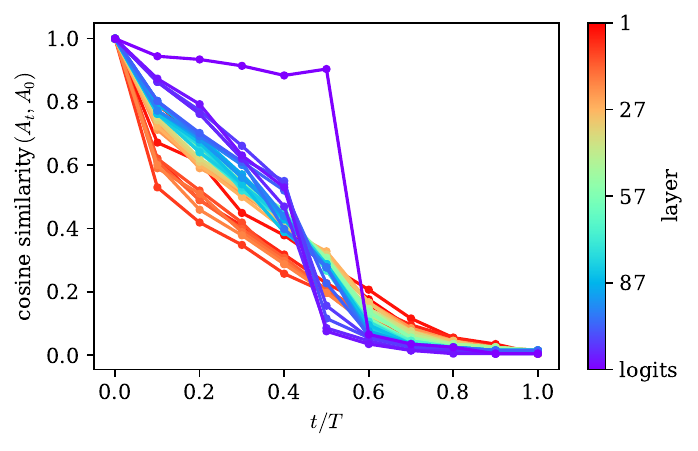}

    \caption{\textit{Left:} \textbf{Examples of images generated by reverting the diffusion process at different times $t$.} Starting from the left images $\x_0$ at time $t=0$, we generate samples $\hat{\x}_{0}(t)\sim  p_\theta(\hat{\x}_{0}|\x_t)$ by first running the diffusion process up to time $t$ and then reverting it, as described in \autoref{sec:forward-backward-exp}.
    At time $t=T$, $\x_T$ corresponds to isotropic Gaussian noise and the generated image $\hat{\x}_{0}(T)$ is uncorrelated from $\x_0$. At intermediate times, instead, a sudden change of the image class is observed, while some lower-level features are retained.
    \textit{Right:} \textbf{Cosine similarity between the post-activations of the hidden layers of a ConvNeXt Base \cite{liu2022convnet} for the initial images $\x_0$ and the synthesized ones $\hat{\x}_{0}(t)$.} Around $t \approx T/2$, the similarity between logits exhibits a sharp drop, indicating the change in class, while the hidden representations of the first layers change more smoothly. This indicates that certain low-level features from the original images are retained for composing the sampled images also after the class transition. To compute the cosine similarity, all activations are standardized, i.e., centered around the mean and scaled by the standard deviation computed on the $50000$ images of the ImageNet-1k validation set. At each time, the values of the cosine similarity correspond to the maximum of their empirical distribution over $10000$ images ($10$ per class of ImageNet-1k).}
    \label{fig:imagenet-main}
\end{figure*}

\paragraph{Recap on diffusion models}
\looseness=-1 Denoising diffusion models are generative models designed to sample from a distribution by reversing a step-by-step noise addition process \citep{sohl2015deep,ho2020denoising,song2019generative, song2020score}. 
Let $t$ indicate the time step in a sequence $[0,\dots, T]$, $p_0$ the data distribution we wish to sample from, and $\x_0 \sim p_0$ a sample drawn from this distribution.
Diffusion models consist of: a \textit{forward process} generating a sequence of increasingly noised data \{$\x_t\}_{1\leq t \leq T}$,
$p(\x_1,\dots,\x_T|\x_0) = \prod_{t=1}^T p(\x_t|\x_{t-1})$, where at the final time $T$, $\x_T$ corresponds to pure noise; a \textit{backward process}, which reverts the forward one by gradually removing noise. This process is typically obtained by learning the \textit{score function}, which is proportional to the conditional expectation $\E_{\x_0 | \x_t}\left[\x_0\right]$, with a neural network.
Sampling from $p_0$ is achieved by sampling noise $\x_T\sim p_T = \mathcal{N}(\mathbf{0},\textbf{I)}$ and then applying the learned backward process $p_\theta(\hat{\x}_0|\x_T)$ to obtain a new sample $\hat{\x}_0$.

\paragraph{Forward-backward experiments}
\label{sec:forward-backward-exp}
Previous studies on DDPMs \cite{ho2020denoising} noted that inverting the diffusion process at different times $t$ starting from an image $\x_0$ results in samples $\hat{\x}_{0}(t) \sim p_\theta(\hat{\x}_0|\x_t)$ with distinct characteristics depending on the choice of $t$. Specifically, when conditioning on the noisy samples $x_t$'s obtained by diffusing images from the CelebA dataset, one finds that for small values of $t$, only fine details change  \cite{ho2020denoising}.
We conduct a similar experiment using a class-unconditional DDPM introduced by \cite{dhariwal2021diffusion}, on the ImageNet dataset with 256x256 resolution. 

In the left panel of \autoref{fig:imagenet-main}, we present some images resulting from this experiment. For each row, the initial image $\x_0$ is followed by images generated by initiating the diffusion process from $\x_0$, running the forward dynamics until time $t$, with ${0 < t \leq T=1000}$, and ultimately running the backward dynamics to produce a sample image $\hat{\x}_0(t)$.
Our observations from these synthetic images are as follows:

\begin{enumerate}
    \item Similarly to the findings in \cite{ho2020denoising}, at small inversion times $t$, only local features change. Furthermore, the class of the sampled images remains consistent with that of the corresponding starting images, i.e., ${{\mathrm{class}}(\hat{\x}_0(t))=\mathrm{class}}(\x_0)$ with high probability.
    \item There exists a characteristic time scale $t^*$ at which the class of the sampled images undergoes a sudden transition.
    \item Even after the class transitions, some low-level features composing the images persist and are reincorporated into the newly generated image. For instance, looking at the left panel of \autoref{fig:imagenet-main}, in the second row, the jaguar is composed with the paws and the ears of the dog in the starting picture, or in the third row, the sofa's armrests inherit the shape of the car headlights.
\end{enumerate}

Our theory, presented in Section \ref{sec:bp} and \ref{sec:mean-field}, predicts how features at different hierarchical levels vary at different time scales of the diffusion dynamics in accordance with observations \textit{(i)}, \textit{(ii)}, and \textit{(iii)}.

\paragraph{ImageNet hidden representations}

To quantify the qualitative observations mentioned earlier, we design an experiment using the empirically known fact that deep learning models learn hierarchical representations of the data, with complexity increasing as the architecture's depth grows. This phenomenon holds true in both real \cite{olah2020zoom, Lecun15, zeiler_visualizing_2014} and synthetic scenarios \cite{cagnetta2023deep, allen-zhu2023how}.
Therefore, we use these internal representations as a proxy for the compositional structure of the data.
We investigate how the hidden representations of a deep ConvNeXt Base model \cite{liu2022convnet}, achieving 96.9\% top-5 accuracy on ImageNet, 
change as a function of the inversion time $t$ and depth $\ell$ of the representation. In the right panel of \autoref{fig:imagenet-main}, we illustrate the value of the cosine similarity between the post-activations of every hidden layer of the ConvNeXt for the initial and generated images. We observe that:
\begin{enumerate}
    \item The representations of early layers of the network, corresponding to low-level and localized features of the images, are the first to change at short diffusion times and evolve smoothly. 
    \item At a specific time and noise scale, the similarity between logits experiences a sharp drop, indicating a transition in the class.
    \item Around the class transition, there is an inversion of the similarity curves. Indeed, the hidden representations in the first layers for the new and generated images now display the largest alignment. This indicates that low-level features from the original images can be reused in composing the sampled images, as qualitatively observed in \autoref{fig:imagenet-main}.
\end{enumerate}
To study the robustness of our results with respect to the architecture choice, in \autoref{app:resnets}, we report the same measurements using ResNet architectures with varying width and depth \cite{he_deep_2016}. We find the same qualitative behavior as the ConvNeXt in \autoref{fig:imagenet-main}.

We now present our theory, which predicts these observations. 

\section{Hierarchical generative model of data}
\label{sec:rhm}

\begin{figure}[t]
    \hspace{1.5cm}
    \begin{tikzpicture}
    \node[anchor=north west,inner sep=0pt] at (0,0){\resizebox{0.24\textwidth}{!}{\input{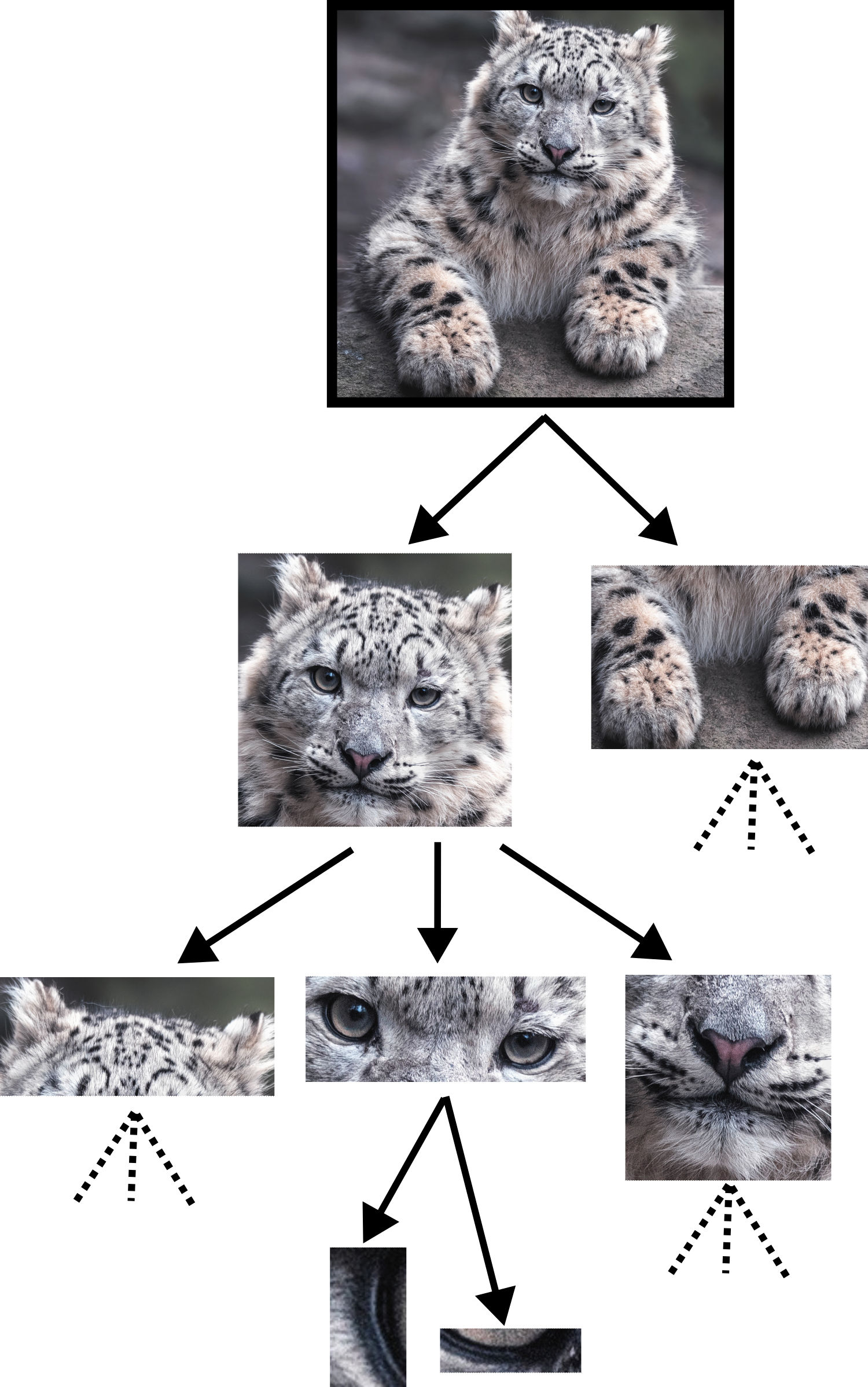}}};
    \end{tikzpicture}
    \hspace{1.cm}
    \includegraphics[width=0.45\linewidth]{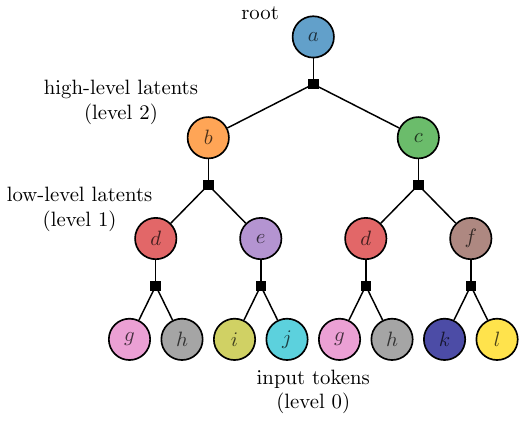}
    \caption{\textbf{Sketch of the hierarchical and compositional structure of data.}
    \textit{Left:} The leopard in the image can be iteratively decomposed in features at different levels of abstraction.
    \textit{Right:} Generative hierarchical model we study in this paper. In this example, depth $L=3$ and branching factor $s=2$. Different values of the input and latent variables are represented with different colors.}
    \label{fig:rhm}
\end{figure}

In this section, we introduce a generative model of data that mimics the structure of images (and language) while being analytically tractable. Natural images often display a hierarchical and compositional structure \cite{grenander1996elements}. Take, for example, the image of a snow leopard (see \autoref{fig:rhm}). This image is composed of multiple high-level components, such as the head and the paws. Each of these components, in turn, is composed of sub-features. For instance, the head comprises elements like ears, eyes, and mouth. Further dissecting these elements, we find even more granular details, such as edges that define the finer aspects of each feature. To model this hierarchical and compositional nature, we consider hierarchical generative models \cite{mossel2016deep,shalev2017failures, malach2018provably, malach2020implications,degiuli2019random,allen-zhu2023how,cagnetta2023deep} belonging to the class of \textit{probabilistic context-free grammars} (PCFGs) \citep{rozenberg_handbook_1997}. These models consist of a collection of symbols and rules that prescribe how to generate sequence data starting from a single feature. Generic PCFGs consist of a vocabulary of hidden (\emph{nonterminal}) symbols, a vocabulary of visible (\emph{terminal}) symbols and \emph{production rules} that quantify the probability that one hidden symbol generates tuples of either hidden or visible symbols.
The \textit{Random Hierarchy Model} -- which we introduced in \cite{cagnetta2023deep}, not included in this thesis -- is a particular PCFG, including the following additional assumptions to make it analytically tractable:
\begin{itemize}
\item[\emph{i)}] The nonterminal symbols are split into $L$ finite vocabularies $(\mathcal{V}_\ell)_{\ell=1,\dots,L}$ of finite cardinality $v$ and $\mathcal{V}_0$ denotes the vocabulary of terminal symbols.
\item[\emph{ii)}] All the production rules transform one level-$(\ell\,{+}\,1)$ symbol into a string of $s$ level-$\ell$ symbols,
\begin{equation}\label{eq:production-rules}
\mu^{(\ell+1)} \to \mu^{(\ell)}_{1},\dots,\mu^{(\ell)}_{s}.
\end{equation}
\item[\emph{iii)}] There are $m$ \emph{unambiguous} production rules per nonterminal symbol, i.e., two distinct nonterminals cannot generate the same $s$-tuple. The rules are randomly chosen and frozen for a given instance of the RHM. We call the $m$ strings produced by any given symbol \textit{synonyms}.
\item[\emph{iv)}] All the available production rules are equally likely.
\end{itemize}
Due to assumptions \emph{i)} and \emph{ii)}, the data-generating process can be represented as a regular tree graph with depth $L$ and branching ratio $s$. The leaf nodes (level $\ell=0$) correspond to the tokens of the visible data, which form strings of size $d=s^L$. 

We adopt a one-hot encoding of these features, ultimately leading to a data vector $X\in \mathbb{R}^{d v}$. Note that for $\ell \geq 1$, the node variables correspond to latent variables, and there is no need to specify any choice of encoding.

The total number of possible data produced per class -- i.e., symbol at the root -- is $m\cdot m^s \cdots m^{s^{L-1}} = m^{\frac{d-1}{s-1}}$, which has exponential dependence in the dimension $d=s^L$.
In the following, we use the notation $X^{(\ell)}_i$ to indicate the variable at layer $\ell$ and position $i\in\{1, \dots, s^{L-\ell}\}$. 

In the context of unsupervised learning, a key parameter for this model is $f=m/v^{s-1}$. When $f=1$, all strings of latent variables of size $s$ can be produced at any level of the hierarchy. This implies that all possible $v^d$ input strings are generated, and the data distribution has little structure. When $f<1$, however, only a small fraction $\sim f^{(d-1)/(s-1)}$  of all possible strings is generated by the production rules. This implies that spatial correlations between different input positions appear, reflecting the hierarchy that generates the data.

\section{Optimal denoising of the RHM with message passing}
\label{sec:bp}

\begin{figure}
    \centering
    \begin{tikzpicture}
    \node[anchor=north west,inner sep=0pt] at (0,0){\resizebox{0.4\textwidth}{!}{\begin{tikzpicture}[varnode/.style={circle, draw=black, thick, minimum size=7mm, inner sep=0pt},
  factnode/.style={rectangle, draw=black, thick, minimum size=7mm, inner sep=0pt},
  level/.style={sibling distance=100mm/#1}, 
  edge from parent/.style={draw,thick}
  ]

\tikzset{
    varnode/.style={circle, draw, thick, minimum size=7mm, inner sep=0pt},
    factnode/.style={rectangle, draw, thick, minimum size=7mm, inner sep=0pt},
    midarrow/.style={decoration={
            markings,
            mark=at position 0.67 with {\arrow{Stealth}}
        },
        postaction={decorate},
        thick
    }
}
  
\node [varnode] (root) {$X_1^{(2)}$};
\node [factnode] (factor) [below= of root] {$\psi^{(2)}$};
\draw[midarrow] (factor) -- (root) node[midway, right] {$\nu_{\uparrow}(X_1^{(2)})$};

% Layer 1
\node [varnode] (left) [below left=0.8cm and 0.5cm of factor] {$X_1^{(1)}$};
\draw[midarrow] (left) -- (factor) node[midway, left=0.1cm] {$\nu_{\uparrow}(X_1^{(1)})$};
\node [varnode] (center) [below right=0.8cm and 0.5cm of factor] {$X_2^{(1)}$};
\draw[midarrow] (center) -- (factor) node[midway, right=0.1cm] {$\nu_{\uparrow}(X_2^{(1)})$};

% Layer 2 - Left Branch
\node [factnode] (leftfactor1) [below =0.8cm of left] {$\psi^{(1)}$};
\draw[midarrow] (leftfactor1) -- (left);
% \node [factnode] (leftfactor2) [below right=0.8cm and -0.5cm of left] {$\psi_{2}$};
% \draw[midarrow] (leftfactor2) -- (left);

\node [varnode] (leftvar1) [below left=0.8cm and 0cm of leftfactor1] {$X_{1}^{(0)}$};
\draw[midarrow] (leftvar1) -- (leftfactor1) node[midway, left=0.1cm] {$\nu_{\uparrow}(X_1^{(0)})$};
\node [varnode] (leftvar2) [below right=0.8cm and 0cm of leftfactor1] {$X_{2}^{(0)}$};
\draw[midarrow] (leftvar2) -- (leftfactor1);

% Layer 2 - Right Branch
% \node [factnode] (rightfactor1) [below left=0.8cm and -0.5cm of center] {$\psi_{3}$};
% \draw[midarrow] (rightfactor1) -- (center);
\node [factnode] (rightfactor2) [below =0.8cm  of center] {$\psi^{(1)}$};
\draw[midarrow] (rightfactor2) -- (center);

\node [varnode] (rightvar1) [below left=0.8cm and 0cm of rightfactor2] {$X_{3}^{(0)}$};
\draw[midarrow] (rightvar1) -- (rightfactor2);
\node [varnode] (rightvar2) [below right=0.8cm and 0cm of rightfactor2] {$X_{4}^{(0)}$};
\draw[midarrow] (rightvar2) -- (rightfactor2) node[midway, right=0.1cm] {$\nu_{\uparrow}(X_4^{(0)})$};

\end{tikzpicture}}};
    \node[font=\small] at (2ex,-2.5ex) {(Up)};
    \end{tikzpicture}
    \hspace{.1cm}
    \begin{tikzpicture}
    \node[anchor=north west,inner sep=0pt] at (0,0){\resizebox{0.4\textwidth}{!}{\begin{tikzpicture}[varnode/.style={circle, draw=black, thick, minimum size=7mm, inner sep=0pt},
  factnode/.style={rectangle, draw=black, thick, minimum size=7mm, inner sep=0pt},
  level/.style={sibling distance=100mm/#1}, 
  edge from parent/.style={draw,thick}
  ]

\tikzset{
    varnode/.style={circle, draw, thick, minimum size=7mm, inner sep=0pt},
    factnode/.style={rectangle, draw, thick, minimum size=7mm, inner sep=0pt},
    midarrow/.style={decoration={
            markings,
            mark=at position 0.67 with {\arrow{Stealth}}
        },
        postaction={decorate},
        thick
    }
}
  
\node [varnode] (root) {$X_1^{(2)}$};
\node [factnode] (factor) [below= of root] {$\psi^{(2)}$};
\draw[midarrow] (root) -- (factor) node[midway, right] {$\nu_{\downarrow}(X_1^{(2)})$};

% Layer 1
\node [varnode] (left) [below left=0.8cm and 0.5cm of factor] {$X_1^{(1)}$};
\draw[midarrow] (factor) -- (left) node[midway, left=0.1cm] {$\nu_{\downarrow}(X_1^{(1)})$};
\node [varnode] (center) [below right=0.8cm and 0.5cm of factor] {$X_2^{(1)}$};
\draw[midarrow] (center) -- (factor) node[midway, right=0.1cm] {$\nu_{\uparrow}(X_2^{(1)})$};

% Layer 2 - Left Branch
\node [factnode] (leftfactor1) [below =0.8cm of left] {$\psi^{(1)}$};
\draw[midarrow] (left) -- (leftfactor1);
% \node [factnode] (leftfactor2) [below right=0.8cm and -0.5cm of left] {$\psi_{2}$};
% \draw[midarrow] (leftfactor2) -- (left);

\node [varnode] (leftvar1) [below left=0.8cm and 0cm of leftfactor1] {$X_{1}^{(0)}$};
\draw[midarrow] (leftfactor1) -- (leftvar1) node[midway, left=0.1cm] {$\nu_{\downarrow}(X_1^{(0)})$};
\node [varnode] (leftvar2) [below right=0.8cm and 0cm of leftfactor1] {$X_{2}^{(0)}$};
\draw[midarrow] (leftvar2) -- (leftfactor1);

% Layer 2 - Right Branch
% \node [factnode] (rightfactor1) [below left=0.8cm and -0.5cm of center] {$\psi_{3}$};
% \draw[midarrow] (rightfactor1) -- (center);
\node [factnode] (rightfactor2) [below =0.8cm  of center] {$\psi^{(1)}$};
\draw[midarrow] (rightfactor2) -- (center);

\node [varnode] (rightvar1) [below left=0.8cm and 0cm of rightfactor2] {$X_{3}^{(0)}$};
\draw[midarrow] (rightvar1) -- (rightfactor2);
\node [varnode] (rightvar2) [below right=0.8cm and 0cm of rightfactor2] {$X_{4}^{(0)}$};
\draw[midarrow] (rightvar2) -- (rightfactor2) node[midway, right=0.1cm] {$\nu_{\uparrow}(X_4^{(0)})$};

\end{tikzpicture}}};
    \node[font=\small] at (3ex,-2.5ex) {(Down)};
    \end{tikzpicture}
    \caption{\textbf{Illustration of the flow of messages in the Belief Propagation algorithm for the case $s=2$, $L=2$ of the Random Hierarchy Model.} The factor nodes (squares) represent the rules that connect the variables at different levels of the hierarchy. The downward process is represented only for the leftmost branch.}
    \label{fig:BP}
\end{figure}
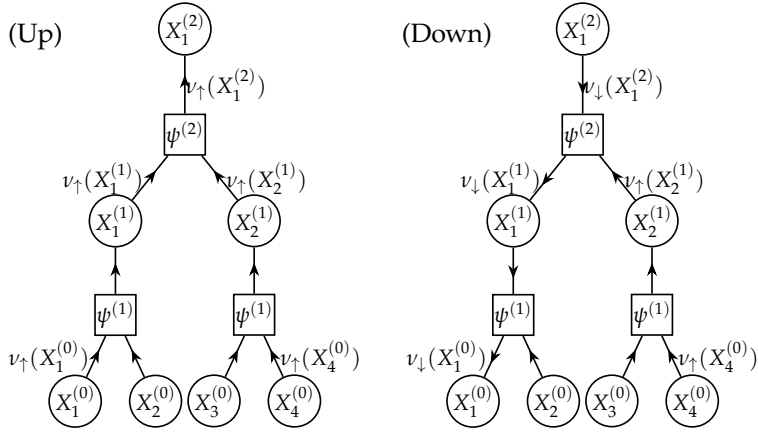

In this section, we characterize the Bayes optimal denoising process for the RHM. Given a noisy observation $X^{(0)}=\x(t)$ of the input variables at time $t$, we compute $p(\x(0)\vert \x(t))$ exactly, obtaining full control of the statistics of the backward diffusion process from time $t$ to time $0$. In particular, given the tree structure of the model, we can compute the marginal probability of the values of all latent variables conditioned on $\x(t)$ by using a message-passing algorithm. Therefore, we obtain the probability that a latent variable at level $\ell$ has changed when performing the forward-backward diffusion process for a duration $t$, a central quantity to interpret \autoref{fig:imagenet-main}.
The optimal denoising corresponds to reconstructing the data distribution $p(\x(0))$ exactly. This perfect reconstruction corresponds to a diffusion model achieving perfect generalization. Although this is a strong assumption for modeling a neural network trained with empirical risk minimization, like the one considered in \autoref{sec:diffusion}, our theoretical analysis captures the phenomenology of our experiments.

\paragraph{Belief Propagation}
For computing the marginal distributions, we use Belief Propagation (BP) \cite{mossel2001reconstruction,  mezardmontanari}, which gives exact results for a tree graph such as the Random Hierarchy Model. In this case, the leaves of the tree correspond to the input variables at the bottom layer, and the root corresponds to the class variable at the top of the hierarchy. Each rule connecting variables at different levels corresponds to a factor node, as shown in \autoref{fig:BP}. 

The forward process adds noise to the variables in the input nodes. Each of these nodes sends its \textit{belief} on its value at $t=0$ to its parent latent node. These beliefs, or \textit{messages}, represent probabilistic estimates of the state of the sender node. Each latent node receives messages from all its children, updates its belief about its state, and sends its \textit{upward message} to its parent node. This process is repeated iteratively until the root of the tree. Subsequently, starting from the root, each node sends a \textit{downward message} to its children. Finally, the product of the upward and downward beliefs received at a given node represents the marginal probabilities of its state conditioned on the noisy observation. Hence, we can use these conditional marginals to compute the mean values of the variables at all levels of the hierarchy.
We assume that the production rules of the model are known by the inference algorithm, which corresponds to the optimal denoising process. 

The input variables $X^{(0)}$, in their one-hot-encoding representation, undergo the forward diffusion process of DDPMs.

The denoising is made in two steps: the initialization of the messages at the leaves and the BP iteration.

\paragraph{initialization of the upward messages} 
In its one-hot-encoding representation, $X^{(0)}_i$ is a $v$-dimensional vector: taking the symbol $a_{\gamma}\in\{a_{1},\dots, a_{v}\}=\mathcal{V}$ corresponds to $X^{(0)}_{i} = e_{\gamma}$, with $e_{\gamma}$ a canonical basis vector.
Its continuous diffusion process takes place in $\mathbb{R}^{v}$: given the value $X^{(0)}_{i}=x_i(t)$, we can compute the probability of its starting value $p(x_i(0)\vert x_i(t))$ using Bayes formula. As derived in \autoref{app:bp}, we obtain
    \begin{align}
        p({x_i(0) = e_{\gamma}} \vert x_i(t)) = \frac{1}{Z} \, e^{x_{i,\gamma}(t)/\Delta_t},
        \label{eq:bayes}
    \end{align}
with $\Delta_t = (1-\overline{\alpha_t})/\sqrt{\overline{\alpha_t}}$ and 
$Z = \sum_{\mu=1}^{v} e^{x_{i,\mu}(t)/\Delta_t}$.
This computation is performed independently for each input variable $i$, and therefore does not take into account the spatial correlations given by the generative model. The probabilities of \autoref{eq:bayes} are used to initialize the BP upward messages $\nnu{0} = p(x_i(0)\vert x_i(t))$ at the input variables.

\paragraph{BP iteration} Let $\rul{\ell}$ be any factor node connecting an $s$-tuple of low-level variables at layer $\ell-1$, $\{X_{i}^{(\ell-1)}\}_{i\in[s]}$, to a high-level variable $X_{1}^{(\ell)}$ at layer $\ell$. 
Without loss of generality, to lighten the notation, we rename the variables as $Y=X_{1}^{(\ell)}$, taking values $y\in\mathcal{V}$, and $X_i = X_{i}^{(\ell-1)}$, each taking values $x_i\in\mathcal{V}$.
For each possible association $y\rightarrow x_1,\dots, x_s$, the factor node $\rul{\ell}(y, x_1, ..., x_s)$ takes values
\beqs
    \rul{\ell}(y, x_1, ..., x_s) = \begin{cases}
        1, \; \text{if } y \rightarrow (x_1, ..., x_s) \text{ is rule at layer }\ell\\
        0, \; \text{otherwise}.
    \end{cases}
\eeqs
The BP upward and downward iterations for the (unnormalized) upward and downward messages respectively read
\beq
    \ncu{\ell+1}(y) = \sum_{x_1, ..., x_s \in \mathcal{V}^{\otimes s}} \rul{\ell+1}(y, x_1, ..., x_s) \prod_{i=1}^s \nnu{\ell}(x_i), \nonumber
\eeq
\begin{align}
    \ncd{\ell}(x_1) &= \sum_{\substack{x_2, ..., x_s \in \mathcal{V}^{\otimes (s-1)}\\ y\in \mathcal{V}}} \rul{\ell+1}(y, x_1, ..., x_s) \nonumber \\
    &\phantom{=} \times \nnd{\ell+1}(y)\prod_{i=2}^s \nnu{\ell}(x_i),
\end{align}
where $\nu_{\rho}^{(\ell)}(x) = \frac{\tilde{\nu}_{\rho}^{(\ell)}(x)}{\sum_{x'} \tilde{\nu}_{\rho}^{(\ell)}(x')}$, $\rho \in \{\uparrow,\downarrow\}$. The downward iteration, reported for $x_1$, can be trivially extended to the other variables $x_i$ by permuting the position indices. The values of $\nnu{0}(x_i)$ and $\nnd{L}(y)$ are set by the initial conditions. In particular, we initialize $\nnu{0}(x_i)$ as described in the previous paragraph and $\nnd{L}(y)=1/v$, which corresponds to a uniform prior over the possible classes.\footnote{This assumption corresponds to unconditioned diffusion, where the DDPM is not biased towards any specific class.}

\begin{figure}
    \centering
    \includegraphics[width=.5\columnwidth]{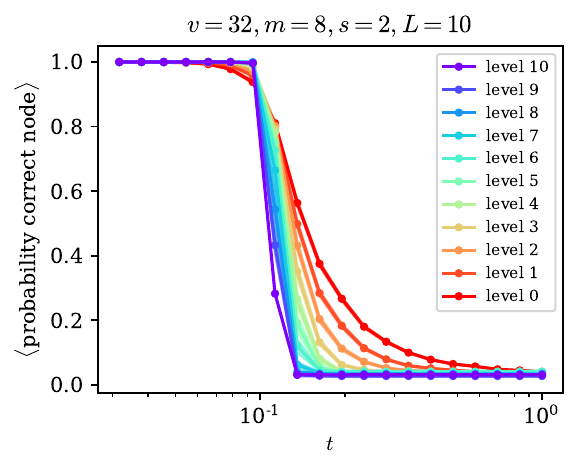}
    \vspace{-10pt}
    \caption{\textbf{Probability that the latent has not changed in the denoising process, corresponding to the largest marginal probability computed by BP, averaged for each layer, for varying inversion times of the diffusion process $t$.} Data for the RHM with $v=32$, $m=8$, $s=2$, $L=10$. Each level of the tree, indicated in the legend, is represented with a different color. We observe the same behavior of the curves for ImageNet data in \autoref{fig:imagenet-main}: the probability of the correct class has a sharp transition at a characteristic time scale, while the probabilities corresponding to latent variables in the lower levels change smoothly.}
    \label{fig:up-down-gauss}
\end{figure}

\paragraph{Results} We run the BP upward and backward iterations numerically. In \autoref{fig:up-down-gauss}, we show the probability corresponding to the correct symbol for each node of the tree. Remarkably, we note that \textit{(i)} the probability for the correct class at layer $L$ displays a transition at a characteristic time which becomes sharper for increasing $L$, and \textit{(ii)} the messages for the correct input variables and the correct latent variables at low levels of the tree change smoothly. In particular, the curves for messages at layer $L$ and layers $\ell < L$ invert their order at the transition, as in our observations on DDPMs and ImageNet data in \autoref{fig:imagenet-main}. This transition is one of our key findings, which we explain below.

\section{Mean-field theory of denoising diffusion}
\label{sec:mean-field}

In this section, we make a simplifying assumption for the initial noise acting on the input and adopt a mean-field approximation to justify the existence of a phase transition. Remarkably, this approximation turns out to be of excellent quality for describing the diffusion dynamics. Specifically, consider a reference configuration at the leaves variables $X_i^{(0)} = \corr{x}_i$ that we would like to reconstruct, given a noisy observation of it. We assume that for each leaf variable, the noise is uniformly spread among the other symbols.\footnote{This is a mild approximation, as documented in appendix} In other words, our belief in the correct sequence is corrupted by $\epsilon \in [0,1]$:

\begin{align}
\begin{cases}
    X_i^{(0)} = \corr{x}_i\quad \text{with belief } 1-\epsilon,\\
    X_i^{(0)} \text{ uniform over alphabet with belief } \epsilon.
\end{cases}
\end{align}

Hence, the initialization condition of the upward BP messages at a leaf node $X_i^{(0)}$ becomes 
\begin{align}
\begin{cases}
    \nnu{0}\lpa \corr{x}_i\rpa \qquad = 1-\epsilon+\epsilon/v,\\
    \nnu{0}\lpa x_i\neq\corr{x}_i \rpa = \epsilon/v,
\end{cases}
\label{eq:belief-main}
\end{align}
where $v$ is the alphabet cardinality.

Given these initial conditions and since the production rules are known, if $\epsilon=0$ -- i.e., in the noiseless case -- BP can reconstruct all the values of the latent variables exactly. Conversely, if $\epsilon=1$ -- i.e., when the input is completely corrupted and the belief on the leaves variables is uniform -- the reconstruction is impossible. In general, for a value of $\epsilon$, one is interested in computing the probability of recovering the latent structure of the tree at each layer $\ell$ and, as $L\to \infty$, to decide whether the probability of recovering the correct class of the input remains larger than $1/v$. 

\begin{figure}
    \centering
    \begin{tikzpicture}
    \node[anchor=north west,inner sep=0pt] at (0,0){
    \resizebox{.5\columnwidth}{!}{
    \input{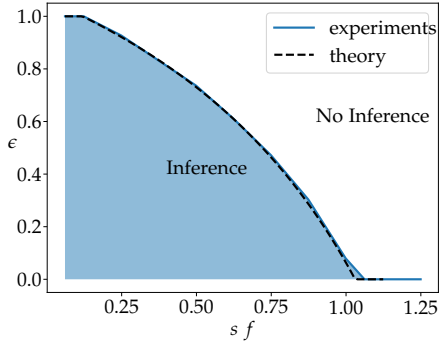}}
        };
    \end{tikzpicture}
    \caption{\textbf{Phase diagram for inferring the class node using the upward iteration of BP.} When $sf<1$, BP can infer the class if $\epsilon<\epsilon^*(sf)$. This transition is very well predicted by our theory. The inference region in the figure corresponds to the phase wherein the probability of the correct class is larger than the initialization belief in the correct values of the leaves, that is $1-\epsilon +\frac{\epsilon}{v}$. Experimental data are for a single realization of the RHM with $v=32$, $s=2$, $L=10$.}
    \label{fig:phase-diagram}
\end{figure}

\paragraph{Upward process} We begin by studying the upward process from the leaves. Consider a true input tuple $\corr{x}_1,\dots,\corr{x}_s$ which is associated with the higher-level feature $\corr{y}$. Given the randomness of the production rules, the messages are random variables depending on the specific realization of the rules. We adopt a \textit{mean-field} or \textit{annealed} approximation that neglects the fluctuations coming from the random choice of rules. Specifically, we approximate the upward message by the average upward message exiting the corresponding factor node $\langle \nnu{1}(y) \rangle_\rr$  over the possible realizations of $\rr$. In \autoref{app:bp}, we show that $\langle \nnu{1}(y) \rangle_\rr$ can take only two values: one for $y=\corr{y}$ and one for $y \neq \corr{y}$, as expected by symmetry considerations. Therefore, mean messages have the same structure as \autoref{eq:belief-main} and we can define a new $\epsilon'$. Introducing the probability of reconstructions $p = 1-\epsilon +\epsilon/v$ and $p' = 1-\epsilon' +\epsilon'/v$, we have

\beq
    p' = \frac{p^s + f \frac{m-1}{mv-1}\lpa 1-p^s\rpa}{p^s + f\lpa 1-p^s\rpa} = F(p).
    \label{eq:iterUP-main}
\eeq

\looseness=-1 Iterating this procedure across all the levels of the tree, we can compute the probability of recovering the correct class of the input. In particular, for large $L$, we are interested in studying the fixed points $p^*=F(p^*)$ of the iteration map in \autoref{eq:iterUP-main}. As derived in \autoref{app:upward_anneal}, when $sf>1$, this map has a repulsive fixed point $p^*=1$, which corresponds to $\epsilon = 0$, and an attractive fixed point $p^*=1/v$, corresponding to $\epsilon = 1$. Thus, in this regime, inferring the class from the noisy observation of the input is impossible. In contrast, when $sf<1$, $p^*=1$ and $p^*=1/v$ are both attractive fixed points, and a new repulsive fixed point $1/v < p^* < 1$ separating the other two emerges. Therefore, in this second regime, there is a phase transition between a phase in which the class can be recovered and a phase in which it cannot. These theoretical predictions are numerically confirmed in the phase diagram in \autoref{fig:phase-diagram}.

Physically, $sf<1$ corresponds to a regime in which errors at lower levels of the tree do not propagate:  they can be corrected using information coming from neighboring nodes, thanks to the fact that only a small fraction of the strings are consistent with the production rules of the generative model. Conversely, when $sf>1$, even small corruptions propagate through the entire tree up to the root node and BP cannot infer the class correctly.

\paragraph{Downward process} The same calculation can be repeated for the downward process, with the additional difficulty that the downward iteration mixes upward and downward messages. We refer the reader to \autoref{app:bp} for the theoretical treatment.

\begin{figure*}
    \centering
    \begin{tikzpicture}
    \node[anchor=north west,inner sep=0pt] at (0,0){\resizebox{0.45\textwidth}{!}{\includegraphics[width=\columnwidth]{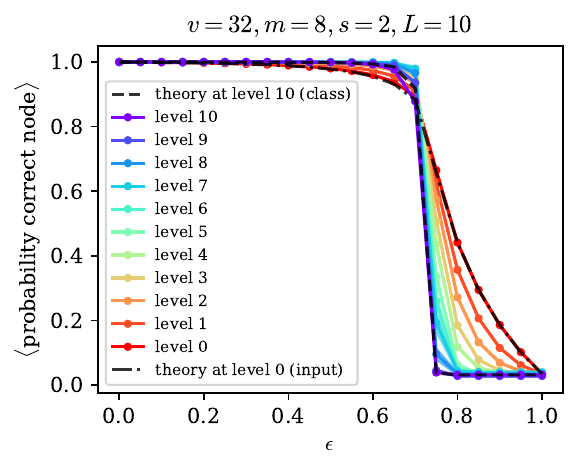}}};
    \node[] at (0ex,-2.5ex) {(a)};
    \end{tikzpicture}
    \hspace{.1cm}
    \begin{tikzpicture}
    \node[anchor=north west,inner sep=0pt] at (0,0){\resizebox{0.45\textwidth}{!}{\includegraphics[width=0.98\columnwidth]{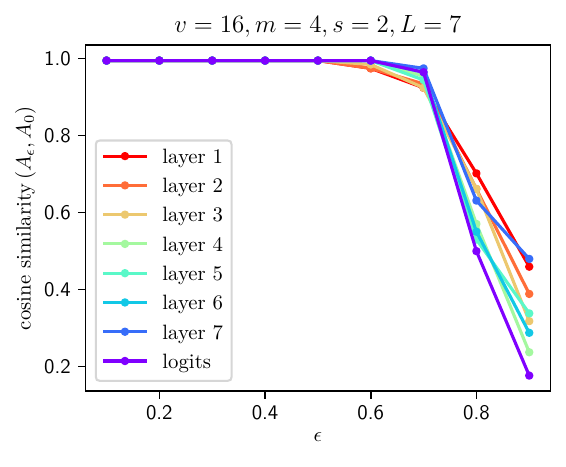}}};
    \node[] at (0ex,-2.5ex) {(b)};
    \end{tikzpicture}
    \caption{\textit{(a)} \textbf{Probability that the latent has not changed in the denoising process, corresponding to the largest marginal probability computed by BP, for varying $\epsilon$.} Data for the RHM with $v=32$, $m=8$, $s=2$, $L=10$. Each level of the tree, indicated in the legend, is represented with a different color. The black dashed lines are our mean-field theoretical predictions, which show excellent agreement with the experiments. In particular, the inversion between the curves for the top and bottom levels at the phase transition can be observed. \textit{(b)} 
    \textbf{(Cosine similarity between the post-activations following every layer of a deep CNN trained on the RHM ($v=16$, $m=4$, $s=2$, $L=7$) for the starting and sampled data.} Each layer of the architecture, indicated in the legend, is represented with a different color. The curves showcase the same inversion predicted by our theory (cf. panel (a)).}
    \label{fig:activations_RHM}
\end{figure*}

\paragraph{Probabilities of reconstruction} Combining the mean upward and downward messages, we obtain a theoretical prediction for the probabilities of reconstructing the correct values of the variables at each layer. We compare our theoretical predictions with numerical experiments in \autoref{fig:activations_RHM}-(a). In these experiments, BP equations are solved exactly for a given RHM starting with the initialization of \autoref{eq:belief-main}. Our theory perfectly captures the probability of reconstruction for the input nodes and the class. Moreover, in \autoref{app:bp} we show that our theory predicts the probabilities of reconstruction of latent nodes at all layers. 

\paragraph{Experiment on CNN's activations} Similarly to our experiment on the ConvNeXt in \autoref{sec:diffusion}, we investigate how the hidden representation of a model trained to classify the RHM changes when its input is denoised starting from a corruption noise $\epsilon$. We consider an instantiation of the RHM with $L=7$, $s=2$, $v=16$, and $m=4$. First, we train a convolutional neural network with $L=7$ layers, matching the tree structure of the model, with $n=300k$ training examples up to interpolation. The resulting architecture has 99.2\% test accuracy. To sample new data from noisy observations of held-out data, we start by sampling the root using the marginal probability computed with BP. Then, we update the beliefs and the marginals conditioning on the sampled class, and sample one latent variable at layer $L-1$. We iterate this procedure node-by-node, descending the tree until we obtain a sampled configuration at the bottom layer \cite{mezardmontanari}. For each corrupting noise $\epsilon$ and each layer of the CNN, we compute the cosine similarity between post-activations for the initial and generated configurations. Panel (b) of \autoref{fig:activations_RHM} shows the obtained curves. Remarkably, we observe the same qualitative behavior as in panel (a) of \autoref{fig:activations_RHM}, ultimately explaining the empirical observation of \autoref{fig:imagenet-main}.

\section{Conclusions}

We have argued that reversing time in denoising diffusion models opens a window on the compositional nature of data. For synthetic hierarchical generative models of data, where the Bayes optimal denoising can be exactly computed, low-level features can already change at small times, but the class remains most often the same. At larger times, a phase transition is found where the probability of remaining in the same class suddenly drops to random chance.  Yet,  low-level features identical to those of the initial sample can persist and compose the new sample. 
Strikingly, this theoretical analysis characterizes well the results found with ImageNet, where the denoising is performed by a trained U-Net. Interestingly, the structure of the U-Net with the skip connections between the encoder and decoder parts mimics the upward and downward iterations of belief propagation, where the downward process mixes upward and downward messages. In fact, building on the present work \cite{mei2024unets} shows that U-Nets are capable of effectively approximating the belief propagation denoising algorithm. Investigating whether the function learned by U-Nets approximates BP is a promising avenue for future work.
While our analysis focused on score-based diffusion, the core findings hold for more recent generative frameworks like flow matching \cite{lipman2022flow} and stochastic interpolants \cite{albergo2023stochastic}. These methods also rely on a continuous transformation from data to noise, and the phase transition we uncovered is a signature of the data's latent hierarchical structure, rather than of the specific noising paths.

In the present work, we used the internal representation of deep networks as a proxy for the hierarchical structure of images. An interesting direction for future work will be using deep hierarchical segmentation techniques \cite{arbelaez2010contour,Ge_2023_CVPR, xie2021unsupervised, zhang2020self} to extract latent variables, so as to test our predictions on their evolution in forward-backward experiments.
Finally, future work can test our theoretical predictions on other modalities successfully handled by diffusion models, such as language and biological structures.

The interplay between the hierarchy in feature space and in time revealed here may help understand the puzzling success of diffusion models, including the number of data needed to train such methods, or why they can generalize and not simply memorize the empirical distribution on which they were trained \cite{somepalli2022diffusion,carlini2023extracting,yoon2023diffusion}. More generally, our results put forward hierarchical generative models as tools to understand open questions for other methods, ranging from the emergence of new skills by the composition of more elementary ones in foundation models to that of transferable representations in self-supervised learning. 

\chapter{Probing Hidden Hierarchies in Data}

\label{ch:probing}

\begingroup
\renewcommand{\thefootnote}{}
\footnote{Parts of this chapter have been previously published in:\\
Sclocchi*, A., \textit{Favero*, A.}, Levi*, N. I. and Wyart, M., 2025. Probing the Latent Hierarchical Structure of Data via Diffusion Models. The 13th International Conference on Learning Representations (ICLR).\\
* These authors contributed equally.}
\addtocounter{footnote}{-1}
\endgroup

In the last chapter, we introduced \textit{forward-backward experiments}, where a controlled level of noise is added to a starting image and then removed to generate a new one \citep{ho2020denoising,sclocchi2024phase,behjoo2023u}. For small amounts of noise, low-level features of the image change. Passed a transition point, the class is likely to change, but remarkably, some of the low-level features of the original image are still retained, as predicted in simple hierarchical models of data structure.
However, this analysis is limited to the image modality. Moreover, the geometrical structure of the changes occurring in such a process is not known.

In this chapter, we derive the length scale associated with changes occurring in the forward-backward protocol with the RHM, and we show experimentally that our predictions hold in both language and image datasets.
Specifically, in the generative model of hierarchically structured data, using a mean-field description of the forward-backward diffusion process, we show theoretically that changes in the tokens are correlated over a length scale that diverges at the class transition. This phenomenology is a signature of the hierarchy in the data structure, indicating changes in deep latent variables.
    
We validate our theoretical predictions by performing numerical experiments on our synthetic data with a diffusion process used in practice for discrete data, showing the same phenomenology predicted by our theory. To do so, we measure the \textit{dynamical susceptibility}, an observable used to study the dynamics in physical systems. 
    
We perform forward-backward experiments with state-of-the-art masked diffusion language models (MDLM) \citep{sahoo2024simple} on WikiText. We show the presence of a peaking correlation length in the token changes at a finite inversion time, consistently with our theoretical model. We perform the same experiments with vision Denoising Diffusion Probabilistic Models (DDPM) \citep{nichol2021improved} on ImageNet. We tokenize the resulting images using the patch embeddings of a contrastively pre-trained vision encoder \citep{radford2021learning} and show that the correlations of token changes display a qualitative agreement with our analysis.

Overall, our results show how changes in latent variables affect visible data, and directly support the idea that a hierarchical latent structure is central to both language and vision modalities. Moreover, our work puts forward the forward-backward protocol as a tool to probe the latent hierarchical structure of real data.

\section{Preliminaries}
\label{sec:probing-background}

\subsection{Discrete diffusion models}

\looseness=-1 For discrete data, like text, $\x_0$ consists of a sequence of tokens $\x_{0,i}$, $i\in[\length]$, each corresponding to a symbol belonging to a vocabulary $\voc$. 
In this case, we consider \textit{masked diffusion with an absorbing state} by introducing an additional $[\mathrm{MASK}]$ symbol \citep{d3pm2021}.
At time step $t$, each non-masked token either stays unchanged or transitions to $[\mathrm{MASK}]$ with some probability $\beta_t$.
Using a one-hot-encoding representation of these $|\voc| + 1$ states, the forward transition matrix $Q_t$ reads
\begin{equation}
    Q_t = (1-\beta_t) \textbf{I} + \beta_t \mathbf{1} \mathbf{e}_M^{\top}.
\end{equation}
with $\textbf{I}$ the identity matrix, $\mathbf{1}$ a vector of ones and $\mathbf{e}_M$ the one-hot-encoding vector corresponding to the $[{\sf MASK}]$ symbol. The element $[Q_t]_{kl}$ indicates the probability of $x_i$ transitioning from state $k$ to state $l$, i.e., $[Q_t]_{kl} = q(x_{t,i}=l \mid x_{t-1,i}=k)$. At the final time $T$, all tokens are masked, i.e., $x_{T,i} = [\mathrm{MASK}]$ for every $i\in[{\mathrm{dim}}(x)]$.
In the following, we consider the noise schedule $\beta_t = (T-t+1)^{-1}$ such that $p_t(x_{t,i} = [\mathrm{MASK}]\,|\,\x_{0}) = t/T$ \citep{d3pm2021}.

\paragraph{Bayes-optimal denoising of the RHM using Belief Propagation}

We consider the \textit{Random Hierarchy Model} \citep{cagnetta2023deep}, and use the notation $\xv^{(\ell)}_i$ to indicate the variable at level $\ell$ and position $i\in [s^{L-\ell}]$.
The leaf nodes $\xv^{(0)}_1, \dots, \xv^{(0)}_{s^L}$ correspond to the visible tokens, while the upper-level nodes represent latent variables. We define the tree distance $\tilde{\ell}$ between two visible tokens as the number of edges between them and their lowest common ancestor. Their corresponding real space distance $r$ is $r=s^{\tilde{\ell}}-1$. Because of the hierarchical structure generating the data, the visible tokens have non-trivial spatial correlations, which depend on their tree distance \citep{cagnetta2024towards}.

\looseness=-1 As discussed in \autoref{ch:phasetransition}, knowing the production rules and the tree structure of the RHM, the probabilities of the latent variables, conditioned on some observation, can be reconstructed exactly \citep{sclocchi2024phase} using the \textit{Belief Propagation (BP)} algorithm \citep{mezardmontanari}.
Specifically, if an RHM datum $\x_0$ is corrupted by some noise, e.g., via masking a fraction of tokens, resulting in a noisy observation $\x_t$, then BP can be used to:
\begin{itemize}
    \item compute the marginal probabilities of any latent or visible variable, conditioned on the noisy observation $\x_t$: $p(\xv^{(\ell)}_i \vert \x_t)$;
    \item sample directly from the posterior $p(\hat{\x}_0 | \x_t)$.
\end{itemize}

If the noisy observation $\x_t$ is produced by a forward diffusion process, then sampling from $p(\hat{\x}_0 | \x_t)$ is equivalent to integrating exactly (i.e., for an infinite number of time steps) the backward diffusion process starting from $\x_t$ and using the exact score function. In fact, BP can also be used to compute the score function, which is proportional to $\E_{\hat{\x}_0 | \x_t}\left[\hat{\x}_0\right]$, corresponding to having access to a neural network achieving perfect generalization (see \autoref{app:probing-sampling} for a comparison between BP sampling and backward diffusion with the score function). This is a different situation with respect to real data, like images and text, where the score is estimated by training a neural network.

\paragraph{Diffusion processes in the RHM}

For the RHM data, we consider two different processes.

\begin{itemize}
    \item \textit{$\epsilon$-process}\quad
    \looseness=-1 This is a simplified process where one considers any datum $\x_0$ that can be generated by the RHM, and assumes that there is some level of uncertainty on each visible token (see \autoref{app:probing-prior} for details). One then uses BP to compute the probability that the true initial datum was  $\hat{\x}_0$.
    The noising process is controlled by a noise-to-signal ratio $\epsilon \in [0,1]$, which plays the role of time in the standard diffusion processes, such that $\epsilon = 0$ at $t=0$ and $\epsilon=1$ at $t=T$. Starting from an RHM datum $\x_0$, we indicate with $\x_\epsilon$ the noisy observation at the leaf priors. Therefore, BP computes the marginals $p(\xv^{(\ell)}_i \vert \x_\epsilon)$ and samples from $p(\hat{\x}_0 | \x_\epsilon)$.
    We study theoretically this process in \autoref{sec:probing-meanfield} through a mean-field approximation, neglecting some fluctuations of the marginal probabilities and averaging over the disorder of the RHM.\looseness=-1
    \item \textit{Masking diffusion with an absorbing state}\quad 
    This is the diffusion process described above for discrete data, which is commonly used in practice. 
    We study it numerically with BP in \autoref{sec:probing-rhm_numerics}.
\end{itemize}

\paragraph{Phase transition in the class reconstruction of the RHM}
In the previous chapter, we showed that there exists a regime of the RHM parameters where the probability of reconstructing the class in the $\epsilon$ diffusion process, that is $p(\xv_{1}^{(L)}|\x_{\epsilon})$, undergoes a sharp phase transition at a critical noise level $\epsilon^*$ in the limit of large $L$.
Therefore, sampling $\hat{\x}_0(\epsilon) \sim p(\hat{\x}_0 | \x_\epsilon)$, for $\epsilon<\epsilon^*$, $\hat{\x}_0(\epsilon)$ and $\x_0$ share the same latent $\xv_{1}^{(L)}$ (i.e. they belong to the same class), while, for $\epsilon>\epsilon^*$, the probability that $\hat{\x}_0(\epsilon)$ and $\x_0$ share the same class corresponds to the random chance $1/v$.

In \autoref{fig:probing-maksing_inversion}, we show numerically that also in masking diffusion the probability of reconstructing the class $p(\xv_{1}^{(L)}|\x_{t})$ undergoes a phase transition at a specific inversion time $t^*$.

\section{Correlations of token changes}
\label{sec:probing-blocks}

\looseness=-1 In this section, we characterize the statistics of how the input tokens change in the forward-backward experiments.
Let $x_{0,i}$ denote the $i$-th input token, $i \in [d]$, and $\hat{x}_{0,i}(t)$ the same token after undergoing a forward-backward experiment with inversion time $t$. We seek to compute the correlations between changes in the tokens as a function of the inversion time $t$. For each token position $i$, we introduce a variable $\sigma_i(t)$ characterizing the dynamics.

\begin{definition}[Token change]
\label{def:spin}
If the tokens $x_{0,i}$ and $\hat{x}_{0,i}(t)$ take values in a discrete vocabulary, then $\sigma_i(t)$ is a spin variable defined as 
\begin{equation}
    \sigma_i(t) = 
    \begin{cases}
        +1, \quad \text{if\ } x_{0,i}\neq \hat{x}_{0,i}(t),\\
        -1, \quad \text{if\ } x_{0,i} = \hat{x}_{0,i}(t).
    \end{cases}
\end{equation}
\end{definition}

\begin{definition}[Dynamical correlation function]
\label{def:corr}
    Given the $\sigma_i(t)$ defined above, the dynamical correlation function between the changes of tokens at positions $i$ and $j$, relative to the initial point $\x_0$, is defined as
    \begin{equation}
        \mathcal{C}_{\x_0,ij}(t) = \langle \sigma_i(t) \sigma_j(t) \rangle - \langle \sigma_i(t) \rangle \langle \sigma_j(t) \rangle,
    \end{equation}
    \looseness=-1 where $\langle \cdot \rangle$ denotes averaging over different stochastic trajectories.
    The \textbf{average dynamical correlation function} is defined as 
    $
    \mathcal{C}_{ij}(t) = \overline{\mathcal{C}_{\x_0,ij}(t)},
    $ 
    where the overline indicates averaging over the initial point $\x_0$.
\end{definition}
Given the correlations, we compute the \textit{dynamical susceptibility} $\chi(t)$, a quantity used to study the dynamics in physical systems \citep{donati2002theory,toninelli2005dynamical}.

\begin{definition}[Dynamical susceptibility]
    Given the average correlation function $\mathcal{C}_{ij}(t)$ of \autoref{def:corr}, the dynamical susceptibility is defined as
    \begin{equation}
    \chi(t) = \frac{\sum_{i=1}^{d} \sum_{j=1}^{d} \mathcal{C}_{ij}(t)}{\sum_{i=1}^{d} \mathcal{C}_{ii}(t)},
\end{equation}
where we normalized by the sum of auto-correlations.
\end{definition}

Intuitively, the susceptibility measures the volume of the blocks of tokens that change together.

In the case of the $\epsilon$-process for the RHM, where $\hat{\x}_0(\epsilon)$ is sampled from $p(\hat{\x}_0 \vert \x_\epsilon)$, the same definitions hold for the quantities $\mathcal{C}_{ij}(\epsilon)$ and $\chi(\epsilon)$. In the case of continuous embeddings, where the tokens $\x_{0,i}$ and $\hat{\x}_{0,i}(t)$ are continuous vectors (see \autoref{sec:probing-experiments} for image diffusion), the same definitions for $\mathcal{C}_{ij}(t)$ and $\chi(t)$ hold by redefining $\sigma_i(t)$ as $\sigma_i(t) = \|\x_{0,i} - \hat{\x}_{0,i}(t)\|$.

\subsection{Mean-field theory of the $\epsilon$-process of the RHM}
\label{sec:probing-meanfield}

\begin{figure}
    \hspace{3cm}
    \includegraphics[width=0.4\textwidth]{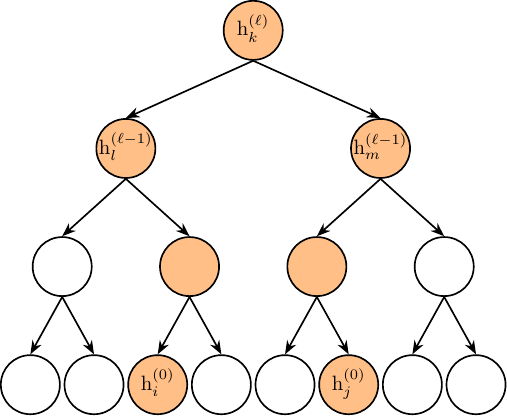}
  \caption{Example of leaf nodes $\xv_i^{(0)}$, $\xv_j^{(0)}$ connected to the common ancestor $\xv_k^{(\ell)}$ through $\xv_l^{(\ell-1)}$ and $\xv_m^{(\ell-1)}$.}
  \label{fig:probing-corr_tree}
\end{figure}

\looseness=-1 The average correlation function $\mathcal{C}_{ij}(\epsilon)$ can be computed for the $\epsilon$-process of the RHM through a mean-field approximation. This mean-field approach consists of computing the average BP messages at each layer $\ell$, where the average is performed over the possible realizations of the RHM rules.
Let's consider two leaf nodes $\xv_i^{(0)}$ and $\xv_j^{(0)}$ connected to the common ancestor $\xv_k^{(\ell)}$ at layer $\ell$ through the nodes $\xv_{l}^{(\ell-1)}$ and $\xv_{m}^{(\ell-1)}$ (see \autoref{fig:probing-corr_tree} for an illustration). 
Their associated spin variables are therefore $\sigma_i^{(0)}$, $\sigma_j^{(0)}$, $\sigma_k^{(\ell)}$, $\sigma_l^{(\ell-1)}$ and $\sigma_m^{(\ell-1)}$, where we omit the $\epsilon$ dependence to lighten the notation.
Given the tree structure, the joint probability distribution $P(\sigma_i^{(0)},\, \sigma_j^{(0)})$ can be written as
{\fontsize{9pt}{10pt}
\begin{align}
\begin{aligned}
    P(\sigma_i^{(0)},\, \sigma_j^{(0)}) = 
    \sum_{\sigma_{l}^{(\ell-1)}, \sigma_{m}^{(\ell-1)}}
    P( \sigma_i^{(0)} \vert \sigma_l^{(\ell-1)})\ 
    P( \sigma_j^{(0)} \vert \sigma_m^{(\ell-1)})\
    P( \sigma_l^{(\ell-1)}, \sigma_m^{(\ell-1)} ).
\end{aligned}
\label{eq:probing-corr}
\end{align}}

Each element in the sum of \autoref{eq:probing-corr} can be written in terms of BP messages, and its average value can be computed by averaging over the realizations of RHM rules.
The average of $P( \sigma_i^{(0)} \vert \sigma_l^{(\ell-1)})$ and $P( \sigma_j^{(0)} \vert \sigma_m^{(\ell-1)})$ can be written as a $2\times 2$ matrix $\mT^{(\ell-1)}$ only depending on the layer $\ell-1$.
Similarly, also the average of the joint probability $P( \sigma_l^{(\ell-1)},\, \sigma_m^{(\ell-1)} )$ can be represented as a $2\times 2$ matrix $\mC^{(\ell-1)}$.
In the mean-field approximation, we neglect the fluctuations of these quantities around their means. Therefore,  we compute the average joint probability $\mP(\sigma_i^{(0)},\, \sigma_j^{(0)})$ by substituting the elements in the product of \autoref{eq:probing-corr} with their means.
For spin variables $i$ and $j$ at tree distance $\ell$, we have
\beq
    \mP(\sigma_i^{(0)},\, \sigma_j^{(0)}) = \mT^{(\ell-1)}\ \mC^{(\ell-1)}\ {\mT^{(\ell-1)}}^{\top}.
\eeq
All the expressions for the above quantities are reported in \autoref{app:probing-meanfield}.
With a similar procedure, we can compute the average marginal probability $\vp(\sigma^{(0)})$.
From these quantities, we obtain the average correlation function $\mathcal{C}_{ij}(\epsilon)$ at each noise level $\epsilon$.

\subsubsection{Dynamical correlation length}

\looseness=-1 In what follows, we present our main result predicting a power law divergence of the dynamical correlation length at the phase transition. 

In the mean-field approach, the average upward belief $p_\ell$ in the original value of a latent variable at layer $\ell$ can be computed through the iterative map
\beq
    p_{\ell} = F(p_{\ell-1}),
    \label{eq:probing-itermap}
\eeq
where the functional form of $F(p)$ was derived in \autoref{ch:phasetransition}
and the initial condition $p_0$ depends on the noise level $\epsilon$ as $p_0 = 1-\epsilon +\epsilon/v$.
In the limit of large depth $L\rightarrow \infty$, the probability $p_L$ of reconstructing the class is given by the fixed points of $F(p)$. For RHM parameters such that $p_L$ undergoes the phase transition, $F(p)$ has three fixed points: two attractive ones, corresponding to $p=1/v$ and $p=1$, and a repulsive one, corresponding to the non-trivial solution of $p^*=F(p^*)$ with $p^* \in (\frac{1}{v},1)$. $p^*$ corresponds to a critical noise level $\epsilon^* = \frac{1-p^*}{1-1/v}$.

In the vicinity of $\epsilon^*$ and the limit $L\rightarrow \infty$, we can estimate the typical distance over which token changes are correlated by computing the number of layers $\tilde{\ell}$ after which the upward probability of reconstructing the latent variables $p_{\tilde{\ell}}$ approaches one of the two trivial fixed points $p=1$ and $p=1/v$. This corresponds to the number of layers required to escape the repulsive fixed point $p^*$.

Given the iterative map of \autoref{eq:probing-itermap}, we can linearize it around the fixed point $p^*$ and iterate for $\ell$ layers, 
\begin{align}
    \Delta p_{\ell} = \left(\frac{dF(p)}{dp}\Bigg|_{p^*}\right)^\ell \Delta p_0,
\end{align}

where $\Delta p_{\ell} = p_\ell-p^*$.
We have that $\frac{dF(p)}{dp}\big|_{p^*}>1$ and we use the shorthand notation $F'_* = \frac{dF(p)}{dp}\big|_{p^*}$. We want to compute the depth $\tilde{\ell}$ at which
${F'_*}^{\tilde{\ell}}\ |\Delta p_0| = \mathcal{O}(1)$.
In terms of the corruption noise $\epsilon$, we have 
${F'_*}^{\tilde{\ell}}\ |\Delta\epsilon| = \mathcal{O}(1)$,
where $\Delta\epsilon = \epsilon - \epsilon^*$. Hence, $\tilde{\ell} \sim  - \log |\epsilon - \epsilon^*|/\log F'_*$. From the depth $\tilde{\ell}$, we can compute the correlation length in input space as 
\begin{equation}
    \xi \simeq s^{\tilde{\ell}} \sim |\epsilon - \epsilon^*|^{- \nu}
    \quad\text{with } \nu = \frac{\log s}{\log F'_*},
    \label{eq:probing-xi-main}
\end{equation}
that diverges at the critical point: $\lim_{\epsilon \to \epsilon^*} \xi = + \infty$.

\begin{figure}
    \centering
    \begin{tikzpicture}
    \node[anchor=north west,inner sep=0pt] at (0,0){
    \includegraphics[width=.52 \textwidth]{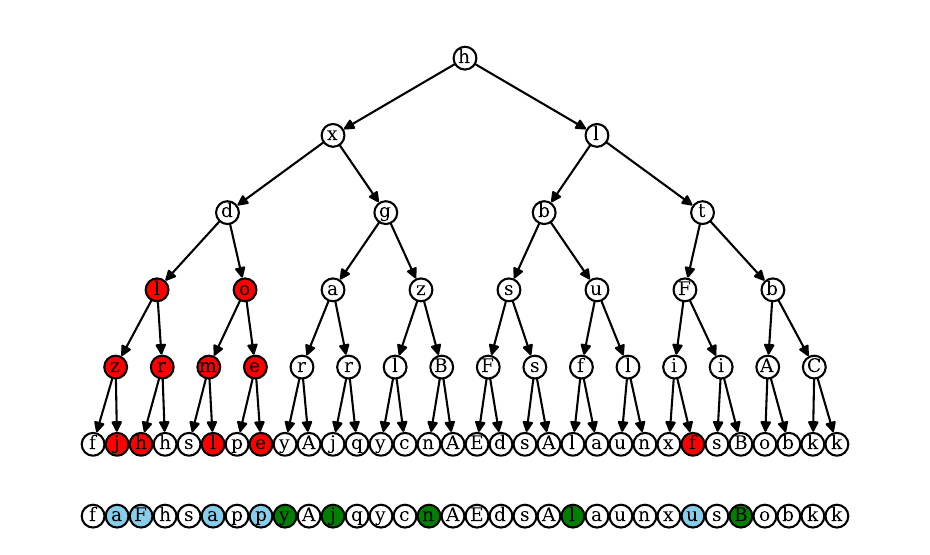}};
    \node[anchor=north west,inner sep=0pt] at (6.7,0){
    \includegraphics[width=.52 \textwidth]{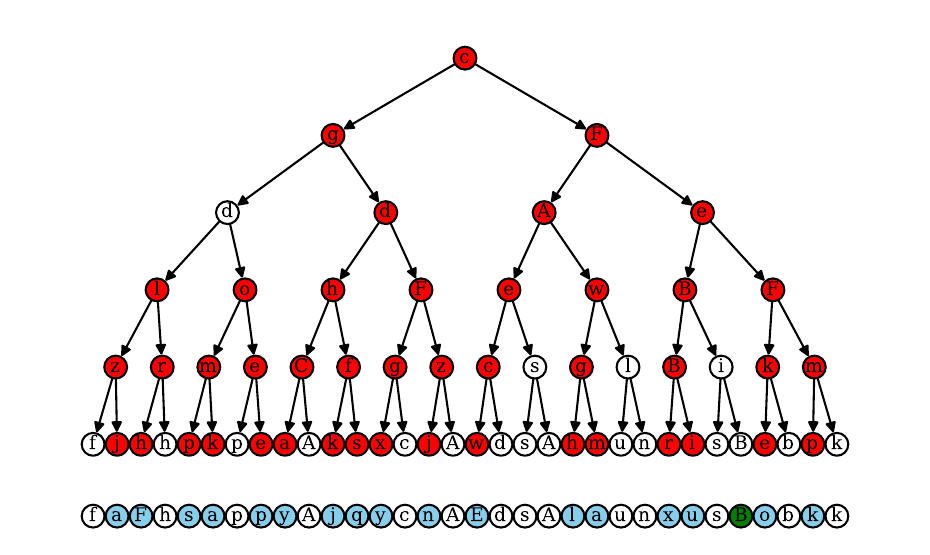}};
    \node at (-1ex,-3.5) {$\hat{\x}_0(t)$};
    \node at (-1ex,-4.05) {$\x_0$};
    \node at (3.5,-4.6) {(a) $t/T = 0.3$};
    \node at (10.5,-4.6) {(b) $t/T = 0.5$};
    \end{tikzpicture}
    \caption{\textbf{Masking diffusion in the RHM for masking fraction \textit{(a)} $t/T=0.3$ and \textit{(b)} $t/T=0.5$.}
    The bottom sequence represents the starting datum $\x_0$. The blue (green) symbols are the masked ones in $\x_t$ that (do not) change feature in $\hat{\x}_0(t)$. The leaves of the tree represent the sampled sequence $\hat{\x}_0(t)\sim p(\hat{\x}_0\vert\x_t)$. In the corresponding tree, the red nodes are those that changed features with respect to the generating tree of $\x_0$. We observe that larger blocks of changed tokens reflect changes in deeper latent variables.
    }
    \label{fig:probing-tree_changes}
\end{figure}

This divergence of the correlation length at the class transition indicates that large blocks of tokens change in concert. In fact, these large correlated changes are caused by the modifications of deeper and deeper latent variables near the transition (see \autoref{fig:probing-tree_changes} for an illustration). At both smaller and larger noise levels, the correlation length decays. This behavior of the dynamical correlation functions implies that the dynamical susceptibility also peaks at the transition, a hallmark of criticality.

\begin{figure}[t!]
    \centering
    \begin{tikzpicture}
        \node[anchor=north west,inner sep=0pt] at (0,0){
        \includegraphics[width=0.49\textwidth]{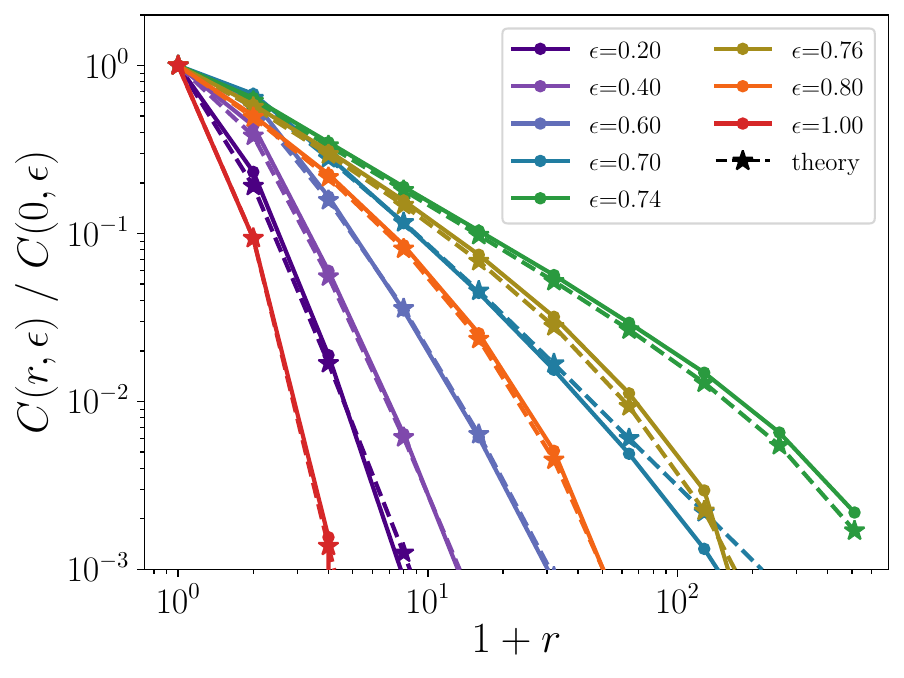}};
        \node at (0.2,0.5ex) {{(a-I)}};
        \node[anchor=north west,inner sep=0pt] at (7.5,0){
        \includegraphics[width=0.47\textwidth]{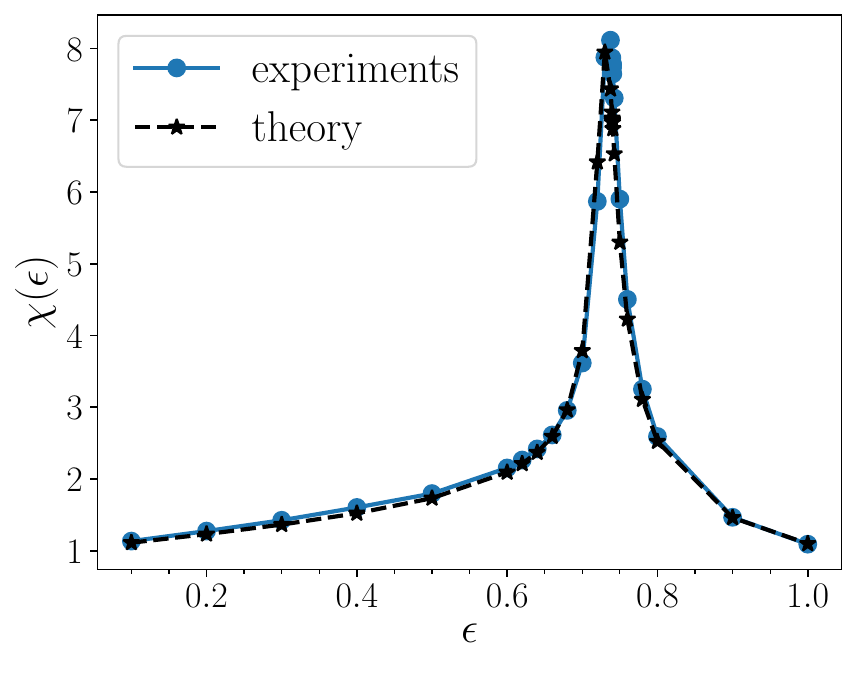}};
        \node at (7.7, 0.5ex) {{(a-II)}};
        \node at (3.5,1ex) {\small $\epsilon$-process: spatial correlations};
        \node at (10.5,1ex) {\small $\epsilon$-process: susceptibility};
    \end{tikzpicture}
    \begin{tikzpicture}
        \node[anchor=north west,inner sep=0pt] at (0,0){
        \includegraphics[width=0.49\textwidth]{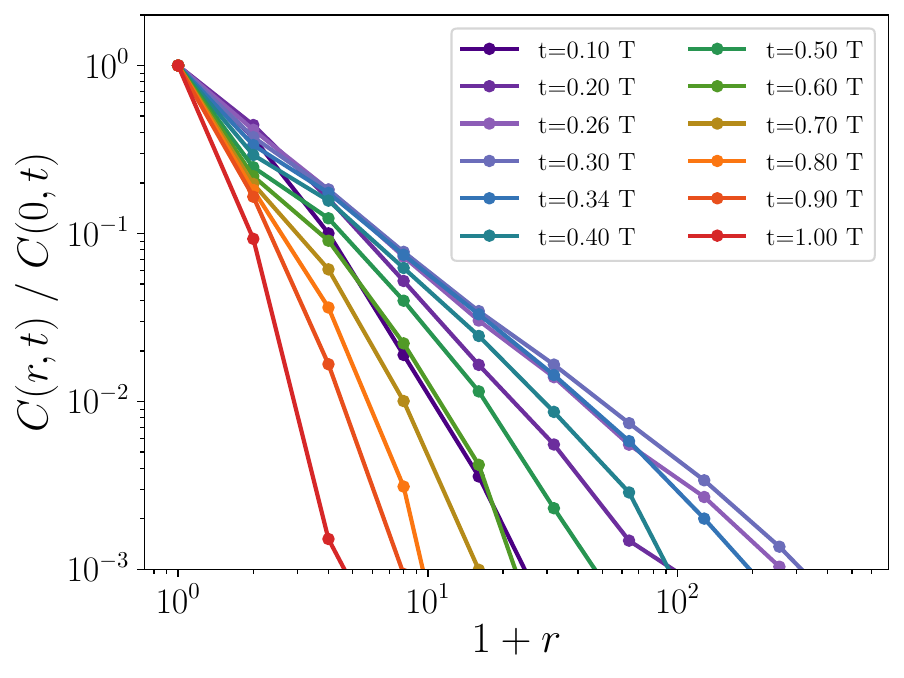}};
        \node at (0.2, 0.5 ex) {{(b-I)}};
        \node[anchor=north west,inner sep=0pt] at (7.5,0){
        \includegraphics[width=0.47\textwidth]{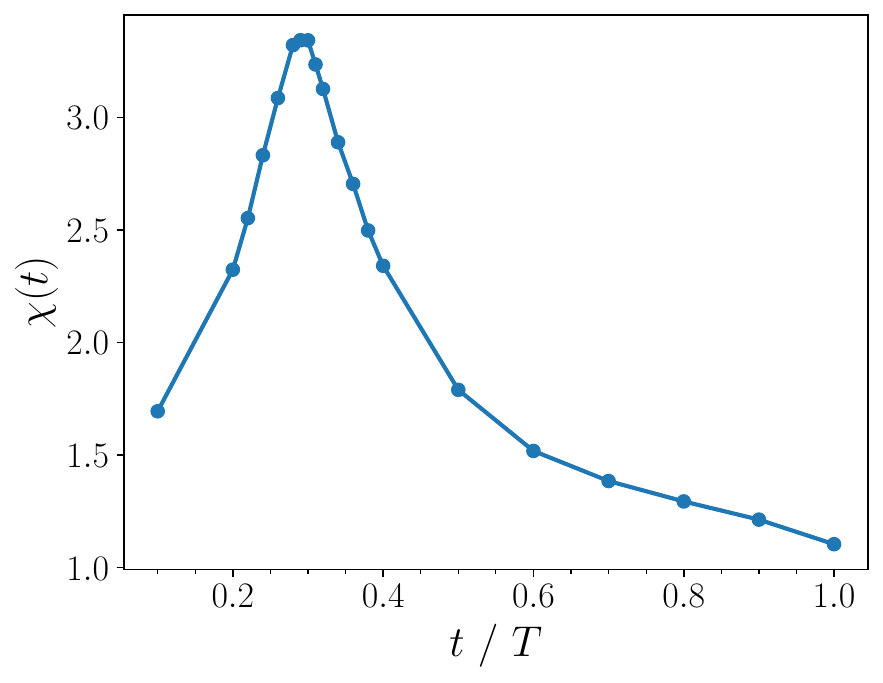}};
        \node at (7.7,0.5ex) {{(b-II)}};
        \node at (4,1ex) {\small Masking diffusion: spatial correlations};
        \node at (11,1ex) {\small Masking diffusion: susceptibility};
    \end{tikzpicture}
\caption{\textbf{Correlation measures on diffusion samples of the Random Hierarchy Model (RHM).} \textit{(a-I)} In the $\epsilon$-process, the average correlation function shows a correlation length that is maximal for $\epsilon^*\simeq 0.74$, corresponding to the class phase transition, with a system-spanning power-law behavior. The full lines are experiments run with Belief Propagation, while the dashed lines are the corresponding mean-field theory description (\autoref{sec:probing-meanfield}), showing excellent agreement.
\textit{(a-II)} Correspondingly, also the average susceptibility shows a peak at the transition $\epsilon^*$.
\textit{(b)} The same behavior is observed for the correlation function \textit{(b-I)} and the susceptibility \textit{(b-II)} for masking diffusion. In this case, the phase transition is observed for inversion time $t^*\simeq 0.3\ T$, where both the correlation length and the susceptibility peak.
Data for RHM parameters $v=32$, $m=8$, $s=2$, $L=9$, averaged over $256$ starting data and $256$ diffusion trajectories per starting datum.}
\label{fig:probing-rhm}
\end{figure}
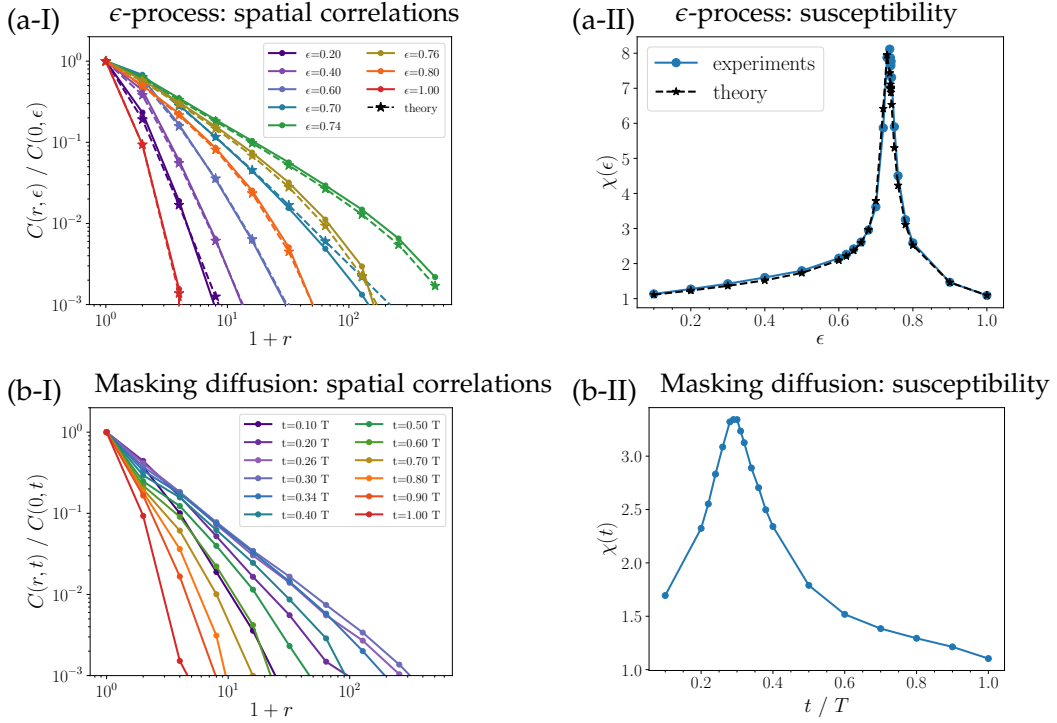

\subsection{Numerical experiments} 
\label{sec:probing-rhm_numerics}

\looseness=-1 To test our theoretical predictions for the $\epsilon$-process, in \autoref{fig:probing-rhm} (a-I), we present the average correlation functions $\mathcal{C}(r,\epsilon)$, corresponding to $\mathcal{C}_{ij}(\epsilon)$ averaged on all pairs $ij$ such that their real space distance is $r$, and normalized by the auto-correlation $\mathcal{C}(0,\epsilon)$. 
We observe that the correlation function displays a system-spanning power-law behavior at a critical value $\epsilon^* \approx 0.74$, while it decays faster with distance when $\epsilon\neq \epsilon^*$. This observation implies a correlation length that peaks at the critical value $\epsilon^*$. Consistently, also the susceptibility $\chi(\epsilon)$ in \autoref{fig:probing-rhm} (a-II) peaks at this critical value. We compare both the correlation functions and the susceptibility measures with the theoretical predictions obtained by the mean-field theory of the $\epsilon$-process (dashed lines in \autoref{fig:probing-rhm} (a-I) and (a-II)), showing excellent agreement. Moreover, in \autoref{fig:probing-correlation_length} of \autoref{app:probing-rhm}, we test the prediction for the critical exponent of the correlation length of \autoref{eq:probing-xi-main}, also showing very good agreement.
 
In the panels (b-I) and (b-II) of \autoref{fig:probing-rhm}, we report the average correlation functions $\mathcal{C}(r,t)$ and susceptibility $\chi(t)$ for masking diffusion at different inversion times $t$. Also in this case, the correlation length and the susceptibility are maximal at a specific critical time $t^* \approx 0.3\ T$. From \autoref{fig:probing-maksing_inversion}, we observe that this critical time $t^*$ corresponds to the phase transition in the class reconstruction probability.
Although there is not a simple mapping between the masking probability $t/T$ and the noise level $\epsilon$ in the simplified $\epsilon$-process, the qualitative behaviors in the two settings show a remarkable agreement, validating the relevance of our theoretical predictions for both kinds of diffusion process.

\subsection{Spatial correlations in data are not sufficient to get a susceptibility peak} 
\label{sec:probing-spatial_correlations}

\looseness=-1 In the RHM, the latent hierarchical structure induces spatial correlations both between the input tokens and in their changes in the forward-backward diffusion.
Therefore, it is natural to ask whether any model of data displaying spatial correlations, even without a latent hierarchical structure, exhibits the same phenomenology of the RHM in the forward-backward diffusion.

In \autoref{app:probing-random}, we show that this is not the case. In particular, we consider a Gaussian random field model with a covariance having a power-law decaying spectrum. This induces spatial correlations in the field that decay algebraically with the distance.
We show that performing forward-backward diffusion at different inversion times $t$ results in a variation field having a correlation length that increases monotonically with $t$ and is maximal at the final time $t=T$. This behavior contrasts sharply with the hierarchical data studied here, where the growing length scale occurs in correspondence with a phase transition at a finite inversion time $t^*$.

In fact, the mechanisms behind the dynamical correlations are different. For Gaussian random fields, the noise of the diffusion acts as a low-pass filter, which defines a characteristic scale below which a mode is reconstructed in the backward process. For hierarchically structured data, instead, the spatial correlations in the changes are associated with the changes of latent variables at different levels of the hierarchy. Therefore, a diverging correlation length is present only when the reconstruction probability of the root node (i.e., the class) undergoes a phase transition. 

\section{Experiments on natural language and image data}
\label{sec:probing-experiments}

\looseness=-1 This section extends our analysis to real-world scenarios, demonstrating that language and vision diffusion models exhibit the same phenomenology as observed in the RHM.

\begin{figure}[t!]
    \centering
    \begin{tikzpicture}
    \node[anchor=north west,inner sep=0pt] at (0,0){
    \includegraphics[width=1 \textwidth]{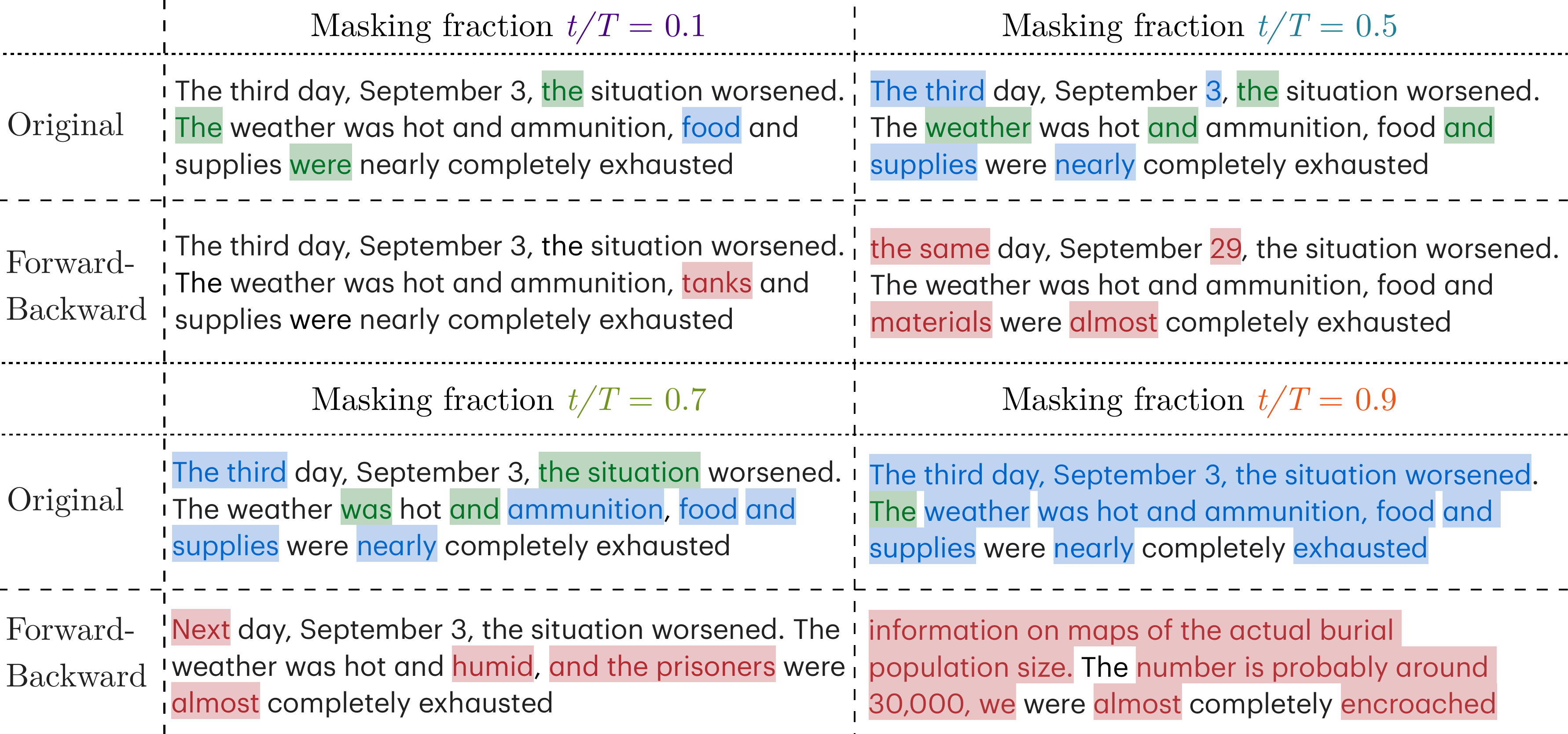}};
    \node at (1.ex,-1ex) {{(a)}};
    \end{tikzpicture}
    \begin{tikzpicture}
    \node[anchor=north west,inner sep=1pt] at (0,0ex){
    \hspace{.1cm}
    \includegraphics[width=.4 \textwidth]{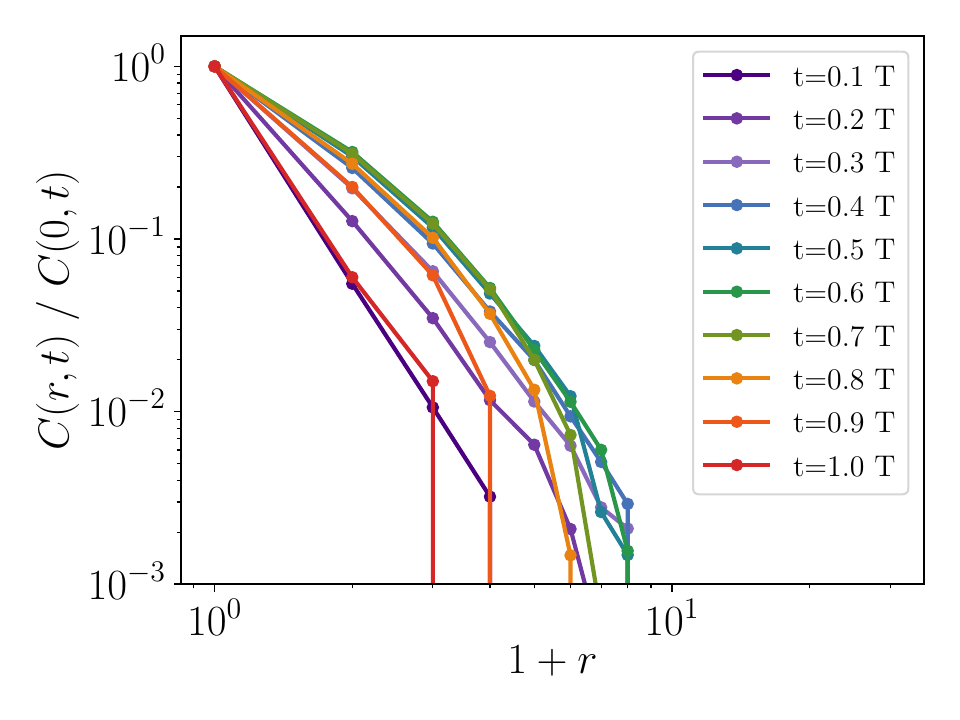}};
    \node at (0.ex,-2ex) {{(b)}};
    \end{tikzpicture}
    \hspace{.1cm}
    \begin{tikzpicture}
    \node[anchor=north west,inner sep=1pt] at (0,0){
        \hspace{.1cm}
    \includegraphics[width=.4 \textwidth]{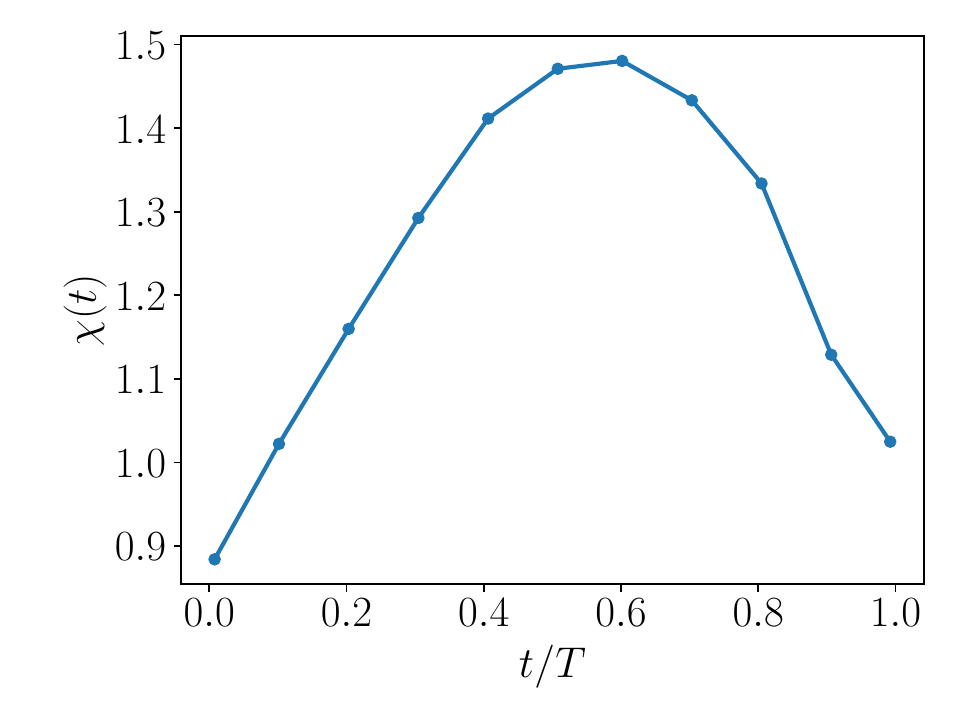}};
    \node at (0.ex,-2ex) {{(c)}};
    \end{tikzpicture}
\caption{\textbf{Forward-backward experiments with language diffusion models.} \textit{(a)} Forward-backward examples for different masking fractions. The words in blue (green) are those that were masked and changed (did not change), while the words in red changed following the backward process. \textit{(b)} Normalized correlations as a function of index distance $r=|i-j|$ for different fractions of masked tokens. \textit{(c)} Susceptibility $\chi(t)$ as a function of masking fraction. The results are averaged over $N_S=300$ samples, each consisting of $N_T=128$ tokens, with $N_R=50$ noise realizations for each masking fraction. The susceptibility is given by integrating over the domain $r\in[0,10]$.
}
\label{fig:probing-text-corr}
\end{figure}
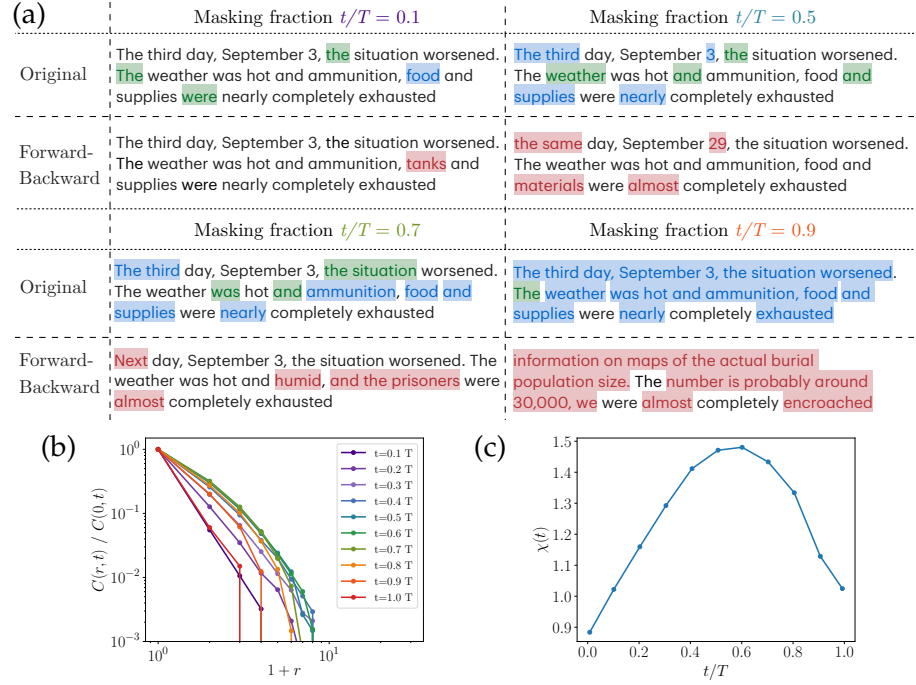

\paragraph{Language diffusion models}

\looseness=-1 We consider Masked Diffusion Language Models (MDLM) \citep{sahoo2024simple} utilizing the GPT2 tokenizer. We perform forward-backward experiments starting from samples from the WikiText2 dataset. In \autoref{fig:probing-text-corr} (a), we illustrate how an initial paragraph changes with different inversion times $t$. At small $t$, only a few isolated words are modified. At intermediate $t$, we clearly observe clusters of words changing in a correlated manner. At large $t$, only a small fraction of the initial sentence remains unchanged (see \autoref{app:probing-language} for a presentation of the same data in their larger context).  In \autoref{fig:probing-text-corr} (b-c), we quantify these observations by measuring the average correlation functions and susceptibility\footnote{To avoid finite size effects due to imposing a fixed masking fraction, we integrate the average correlation function up to the maximal correlation length $r\sim\mathcal{O}(10)$.}. Strikingly, in line with the phenomenology obtained for the RHM, we find a growing correlation length as $t$ increases, reaching a maximum of $7 \div 8$ tokens at a critical inversion time $t^* \approx 0.6\, T$, followed by a subsequent decline. Additionally, the susceptibility peaks at $t^*$, establishing the existence of a phase transition for the language modality.

\paragraph{Vision diffusion models}

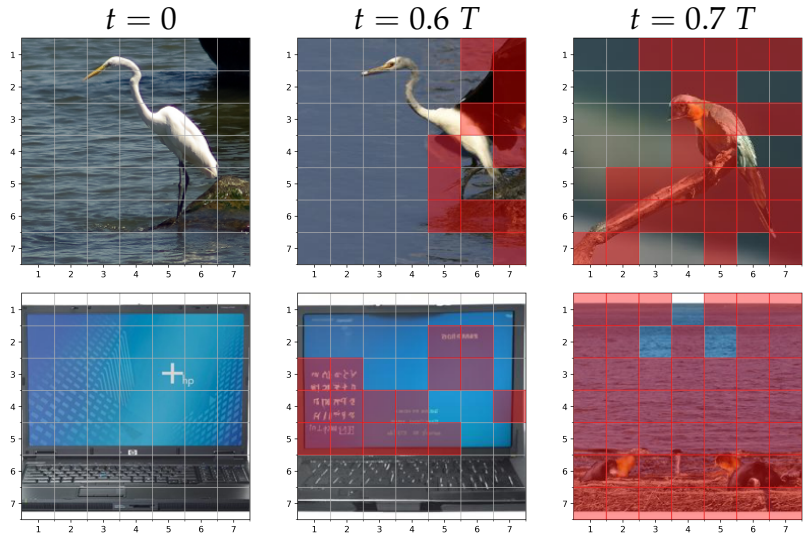
\begin{figure}[t!]
    \centering
    \begin{tikzpicture}
        \node[anchor=north west,inner sep=0pt] at (0,0){
        \includegraphics[width=0.28\textwidth]{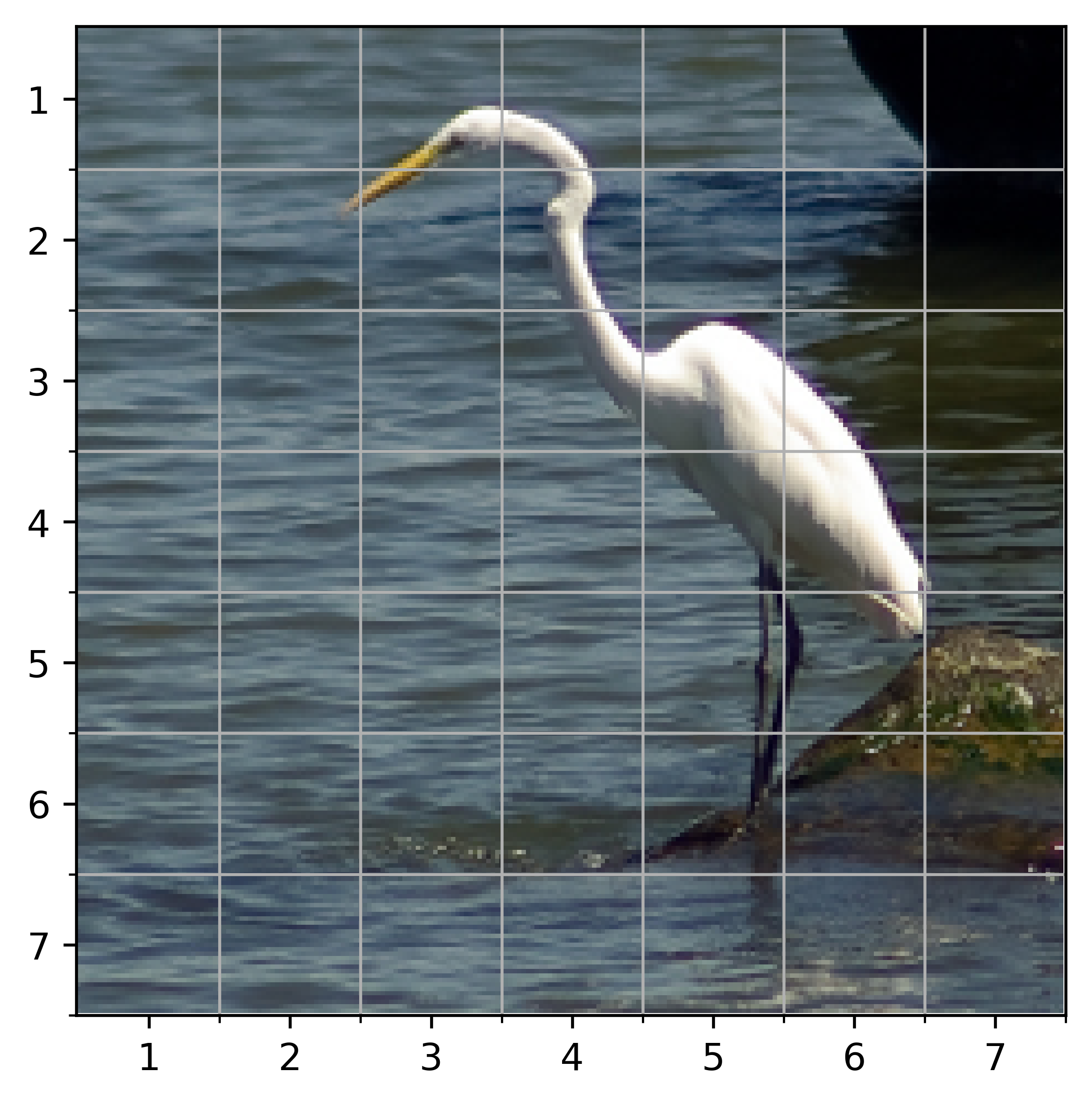}};
        \node[font=\large] at (10ex,1ex) {$t=0$};
    \end{tikzpicture}
    \hspace{.1cm}
    \begin{tikzpicture}
        \node[anchor=north west,inner sep=0pt] at (0,0){
        \includegraphics[width=0.28\textwidth]{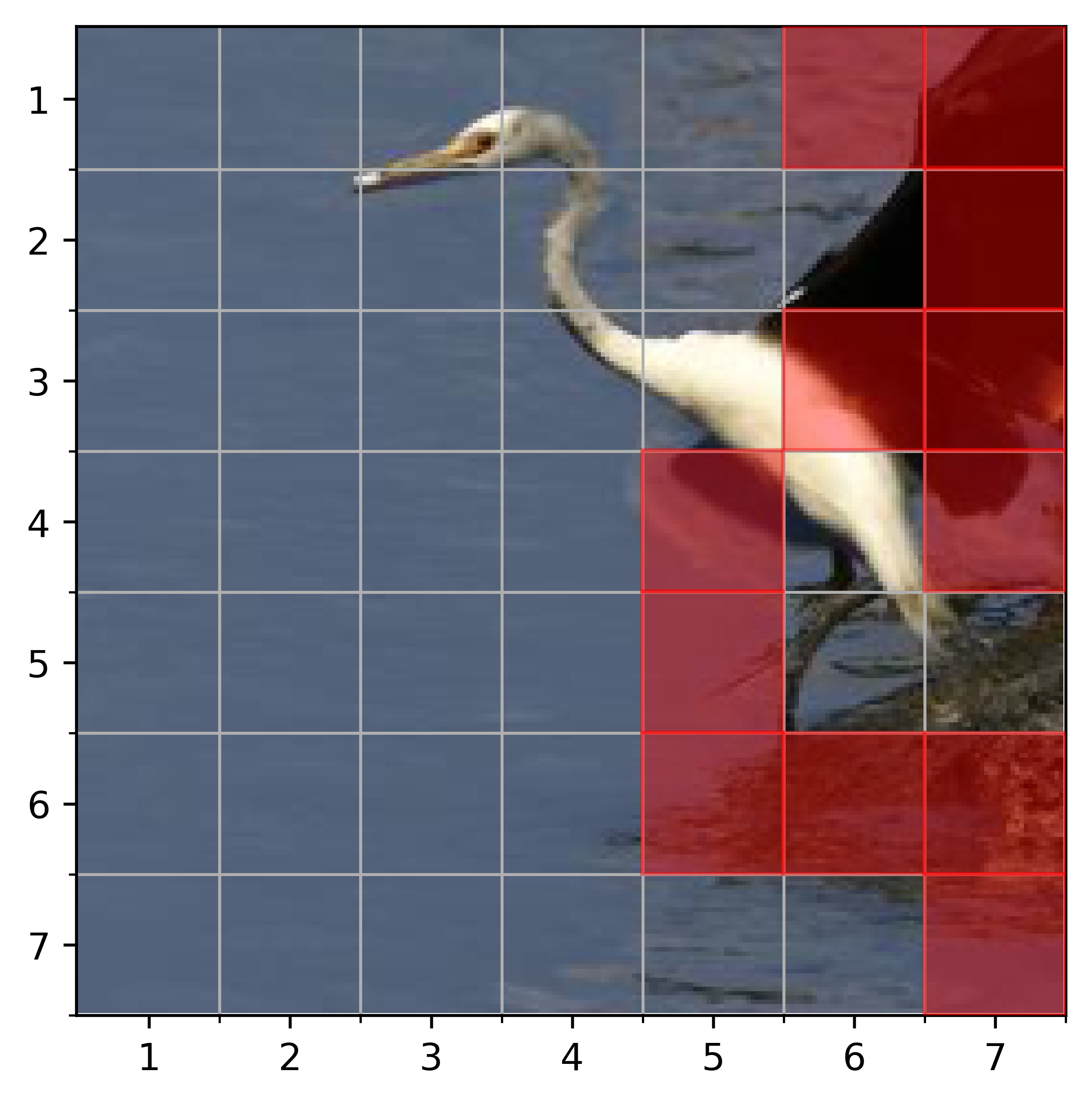}};
        \node[font=\large] at (10ex,1ex) { $t=0.6\ T$};
    \end{tikzpicture}
    \hspace{.1cm}
    \begin{tikzpicture}
        \node[anchor=north west,inner sep=0pt] at (0,0){
        \includegraphics[width=0.28\textwidth]{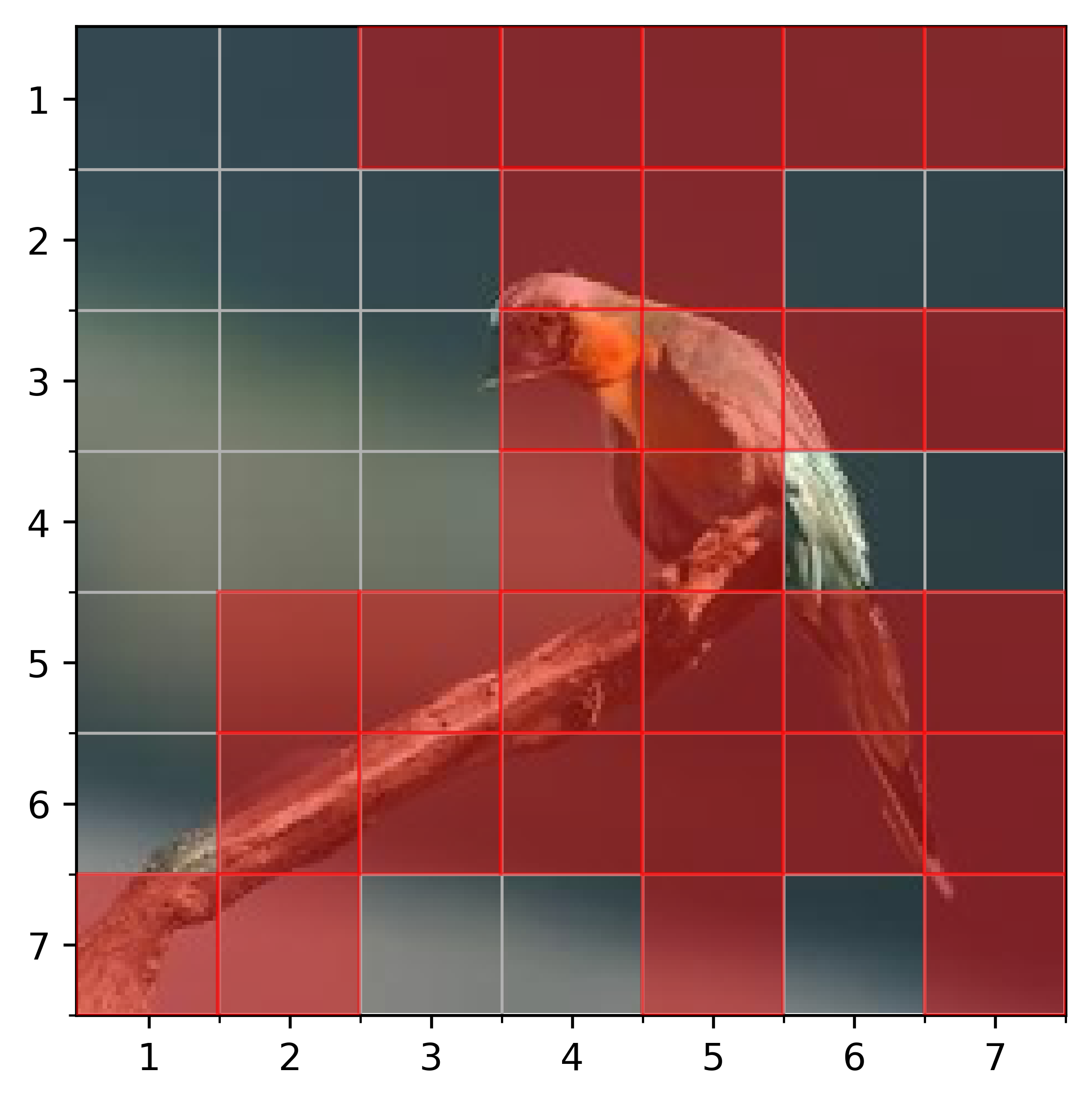}};
        \node[font=\large] at (10ex,1ex) {$t=0.7\ T$};
    \end{tikzpicture}
    \begin{tikzpicture}
        \node[anchor=north west,inner sep=0pt] at (0,0){
        \includegraphics[width=0.28\textwidth]{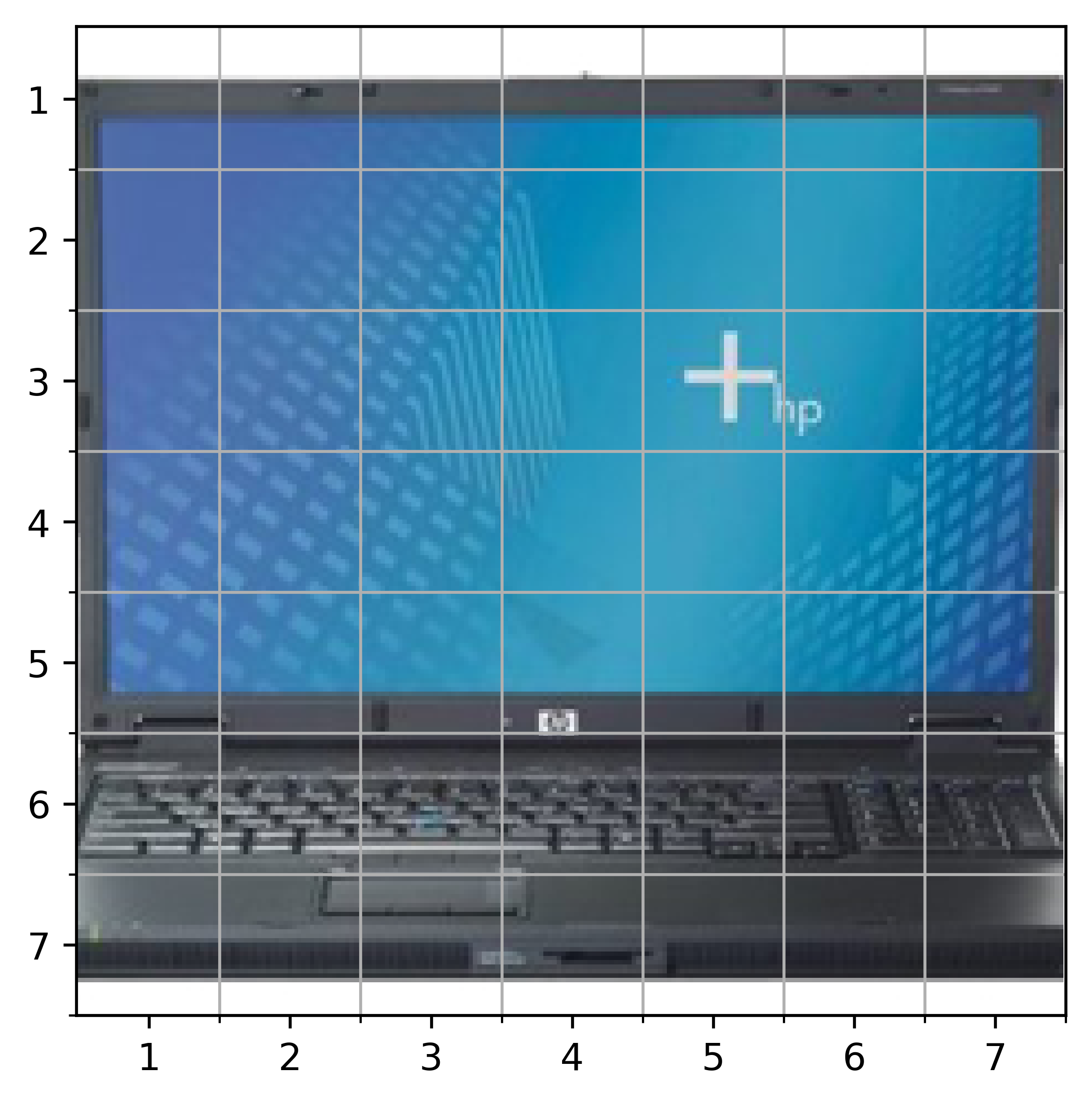}};
    \end{tikzpicture}
    \hspace{.1cm}
    \begin{tikzpicture}
        \node[anchor=north west,inner sep=0pt] at (0,0){
        \includegraphics[width=0.28\textwidth]{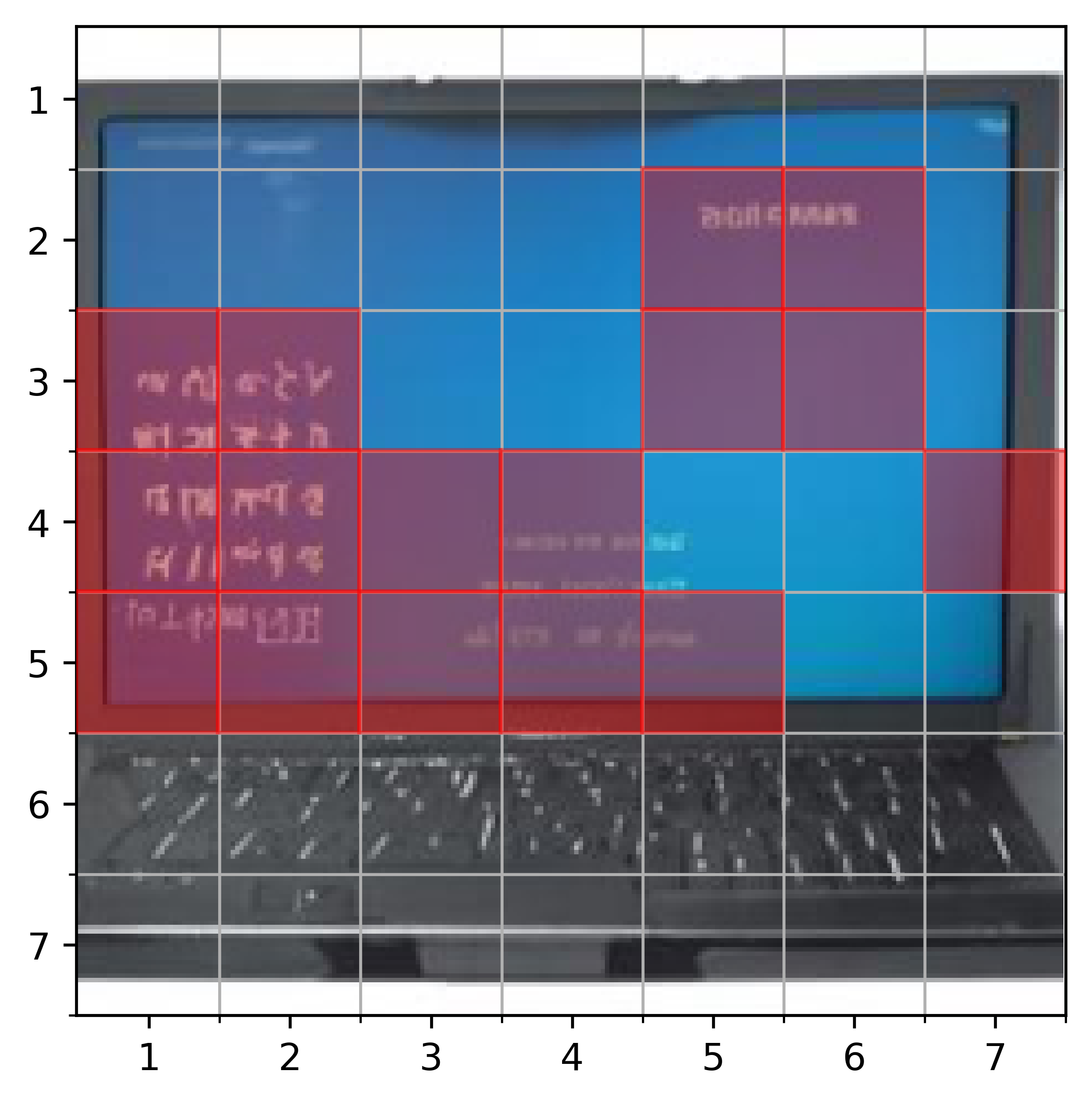}};
    \end{tikzpicture}
    \hspace{.1cm}
    \begin{tikzpicture}
        \node[anchor=north west,inner sep=0pt] at (0,0){
        \includegraphics[width=0.28\textwidth]{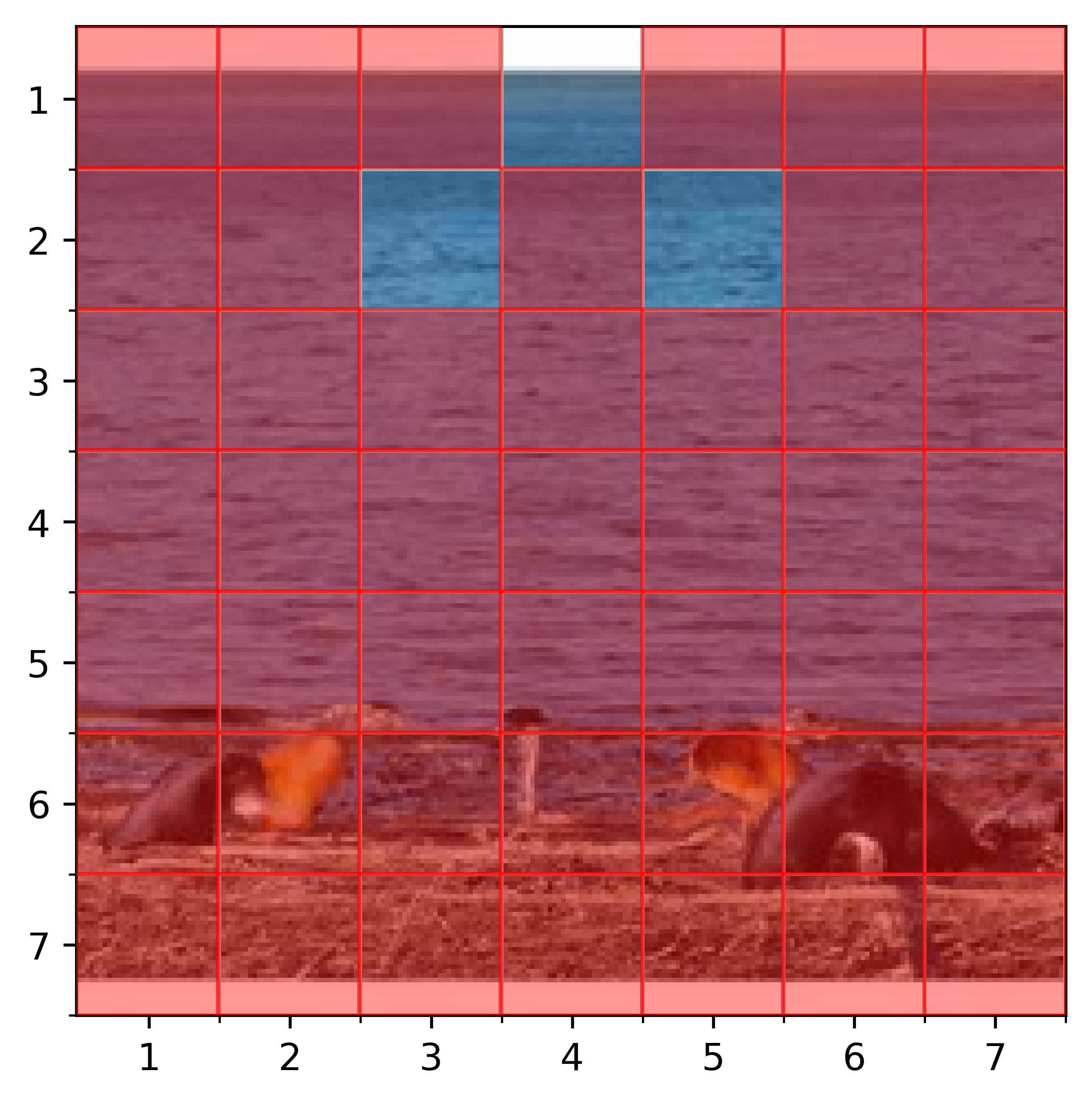}};
    \end{tikzpicture}
    \caption{\looseness=-1 \textbf{Examples of images generated at different inversion times $t$ with forward-backward diffusion.} The first column represents the starting images $\x_0$, while the other columns represent the generated ones $\hat{\x}_0(t)$. The grid indicates the tokens represented inside the CLIP vision encoder. For inversion time $t>0$, the red patches indicate the token embeddings that have a variation magnitude larger than a fixed threshold. These patches of variation appear in domains of growing size around the class transition, observed for $t^*\approx 0.6\div 0.7 T$ (\autoref{fig:probing-clip}).}
    \label{fig:probing-images}
\end{figure}

\begin{figure}[t!]
    \centering
    \begin{tikzpicture}
        \node[anchor=north west,inner sep=0pt] at (0,0){
        \includegraphics[width=0.47\textwidth]{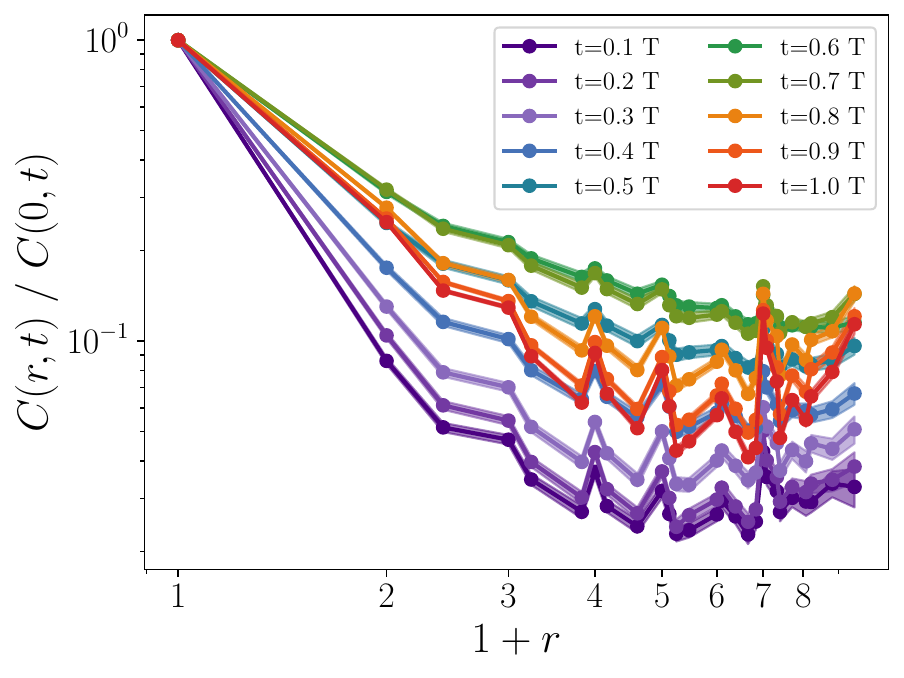}};
        \node at (1.ex,-2ex) {{(a)}};
        \node[anchor=north west,inner sep=0pt] at (7,0){
        \includegraphics[width=0.47 \textwidth]{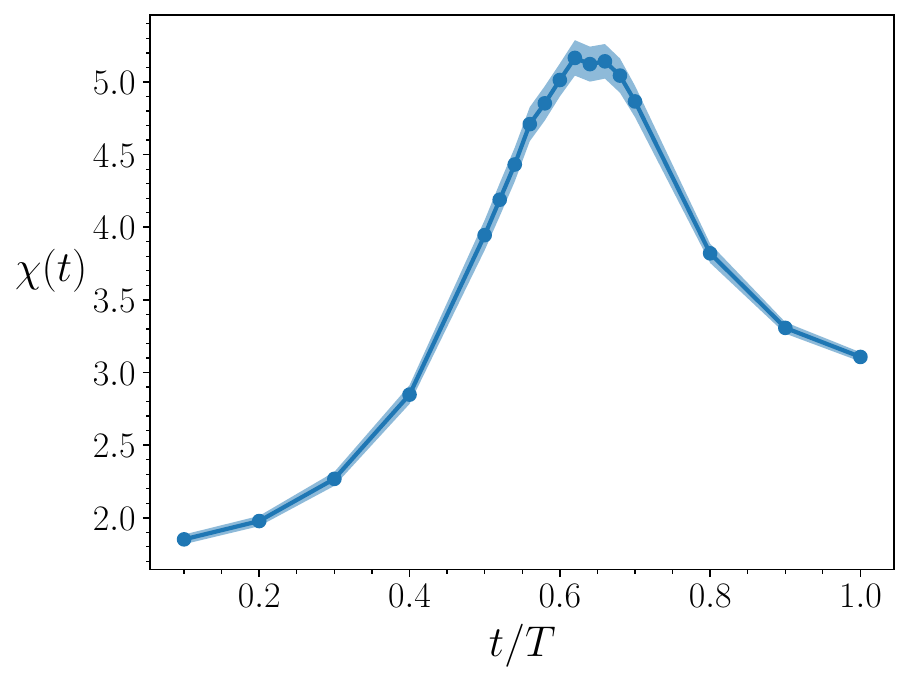}};
        \node at (7.5,-2ex) {{(b)}};
    \end{tikzpicture}
\caption{\looseness=-1 \textbf{Correlation measures on the variation of CLIP embeddings of images generated with forward-backward diffusion.}
\textit{(a)} The average correlation function displays a system spanning power-law behavior for $t^*\approx 0.6 \div 0.7\ T$, corresponding to the class phase transition (cf. \autoref{fig:probing-logits-class}).
\textit{(b)} In correspondence with the phase transition, the average susceptibility displays a peak.
Data obtained with $344$ starting images and $128$ diffusion trajectories per starting image. The shaded areas correspond to the standard deviations over the starting images.
}
\label{fig:probing-clip}
\end{figure}

\looseness=-1 We extend our analysis to computer vision by considering Improved Denoising Diffusion Probabilistic Models \citep{nichol2021improved}, trained on the ImageNet dataset. To compute the correlation between changes in the image tokens, we follow recent trends in multimodal LLMs \citep{liu2024visual,dai2023instructblip}. Specifically, we divide each image into $7 \times 7$ patches and use the last-layer embeddings for each patch from a CLIP ViT-B32 \citep{radford2021learning} to tokenize the image. Let $\x_{0,i}$ denote the embedding of the $i$-th patch, where $i=(k,l)$ with $k,l\in[7]$. After the forward-backward process, the variation of each patch embedding is given by $\Delta \x_i(t) = \x_{0,i} - \hat{\x}_{0,i}(t)$.
We then compute the average correlations between the norms of these variations:
\begin{equation}
    \mathcal{C}_{ij}(t) = \overline{\langle \|\Delta \x_i(t)\| \, \|\Delta \x_j(t)\| \rangle - \langle \|\Delta \x_i(t)\| \rangle \langle \|\Delta \x_j(t)\| \rangle}.
\end{equation}
The susceptibility is subsequently obtained as $\chi(t)=\sum_{ij} \mathcal{C}_{ij}(t)/\sum_{ii}\mathcal{C}_{ii}(t)$.
In \autoref{fig:probing-images}, we present some examples of starting images and generated ones at different inversion times $t$, together with the grid representing their tokenization. We observe that, for increasing $t$, new semantic elements appear in the generated images, corresponding to blocks of tokens changing in concert.
In \autoref{fig:probing-clip}, we present the average correlation functions and the susceptibility for vision DDPMs, starting from samples of the ImageNet validation set \citep{deng2009imagenet}. At a critical inversion time $t^*\approx 0.6 \div 0.7 \, T$, we observe a peak in susceptibility, signaling the class phase transition in these models. This finding highlights the compositional semantic structure of image data, similar to the phase transitions observed in language diffusion models and the RHM.

\section{Related work}
\looseness=-1

\looseness=-1 \paragraph{Phase transitions in diffusion models}
Several works have studied phenomena related to phase transitions in diffusion models.
\citet{biroli2024dynamical,ambrogioni2023statistical} show the presence of different dynamical regimes in the diffusion process separated by a `speciation' cross-over where a bimodal distribution merges into a mono-modal one. 
\citet{li2024critical} provide bounds for critical time windows appearing in the diffusion of mixtures of strongly log-concave densities.
These works do not consider hierarchical data, and thus do not present growing dynamical susceptibility or length scale at the transition.
\looseness=-1 \paragraph{Hierarchical models of images and text}
Generative models have been used to describe the structure of data in several contexts, including in linguistics and signal processing.
For languages, hierarchically-structured formal grammars are often used as a model of their syntactic structure \citep{rozenberg_handbook_1997}.
Likewise, pattern theory formalizes the semantic decomposition of visual scenes through a hierarchy of features \citep{grenander1996elements, jin2006context, siskind2007spatial, li2009towards}.
More recently, images have been described through a hierarchical decomposition in multi-scale wavelet coefficients \citep{marchand2022wavelet, kadkhodaie2023learning}, although the underlying structure, in this case, is not tree-like.

\looseness=-1 \paragraph{Semantic vs geometrical description of images}
Several studies \citep{rissanen2022generative, wang2023diffusion} pointed out that the backward diffusion process of images acts on coarse-to-fine scales. 
Since the Fourier spectra of images decay as power laws, higher frequencies are affected at short diffusion times, while low-frequency modes persist for longer. This is
precisely the pattern we describe in the Gaussian random field model in \autoref{sec:probing-spatial_correlations} and \autoref{app:probing-random}. 
While this viewpoint is an appropriate starting point to describe the geometrical structure at the pixel level, 
our hierarchical model seeks to capture a high-level, semantic description of images that we test using a CLIP encoder. This means that high/low-level features can correspond to parts of objects – such as the eyes, mouth,
and nose of a face – rather than simple geometric or frequency components.
The two descriptions are, therefore, complementary. 

\section{Conclusions}
\looseness=-1 In this chapter, we showed that when data exhibit a hierarchical structure, the changes induced by forward-backward experiments in diffusion models reveal a growing correlation length and susceptibility near a phase transition.  At this critical point, changes in the data become highly correlated, reflecting changes in deep latent variables. 
In particular, we focused on understanding how modifications in the latent variables manifest in the data, in contrast with common approaches which attempt to reconstruct the latent representations from visible data.

\looseness=-1 Our predictions for a hierarchical model were confirmed through experiments across different natural data modalities, showing a remarkable level of universality.
This supports the hypothesis that hierarchical and compositional structures are fundamental, universal properties underlying natural data as diverse as images and text. 

\looseness=-1 Such fundamental analyses have the potential to impact practical applications. For example, they can enhance the interpretability of deep networks, whose representations are believed to reflect the hierarchical structure of data. Moreover, the diffusion dynamics of high and low-level features can suggest improved training strategies -- for instance, to avoid mode collapse when fine-tuning diffusion models \citep{barcelo2024avoiding}.

\looseness=-1 Future work may include interpreting the large, correlated changes in text in terms of grammatical structure and context variables, possibly sharpening these concepts through the data-driven method introduced in this study.
\looseness=-1 Moreover, better capturing the grammatical structure of real languages may require considering more general latent models involving context dependencies. A challenge for future work is extending our theoretical analysis to such cases.

\chapter{A Theory of Creativity and Compositionality}

\label{ch:creativity}

\begingroup
\renewcommand{\thefootnote}{}
\footnote{Parts of this chapter have been previously published in:\\
\textit{Favero*, A.}, Sclocchi*, A., Cagnetta, F., Frossard, P. and Wyart, M., 2025. How Compositional Generalization and Creativity Improve as Diffusion Models Are Trained. To appear in Proceedings of the 42nd International Conference on Machine Learning (ICML), PMLR.\\
* These authors contributed equally.}
\addtocounter{footnote}{-1}
\endgroup

\textit{Compositional generalization}, the ability to understand and generate novel combinations of known components, is a fundamental characteristic of human intelligence. This skill underlies what linguists refer to as \textit{creativity} \cite{chomsky1976reflections}: the capacity to produce an infinite number of novel and well-formed expressions from a finite set of rules. Under which conditions can machines learn such a skill?  The success of diffusion models in producing realistic data across various domains~\cite{sohl2015deep,ho2020denoising,song2019generative,betker2023improving,rombach2022high} provides a unique opportunity to study how this ability emerges. Fundamental questions include:  What signals in the data are exploited by neural networks to learn the \textit{compositional rules}? How many training examples are needed to learn such rules, and in what order are they learned? How does the finiteness of the training set affect the structure of generated data?

To address these questions theoretically, we bridge two viewpoints developed in the context of natural language processing. On the one hand, \textit{symbolic approaches} aim to describe the structure of data via a list of rules that generate them. For example, \emph{probabilistic context-free grammars} (PCFG)~\cite{chomsky2014aspects} describe sentences with trees, whose nodes are hidden variables that can generate other nodes or leaves according to probabilistic production rules. PCFGs can approximate both structural and semantic aspects of text and have also been proposed for the description of images under the name of \emph{Pattern Theory}~\cite{grenander1996elements,jin2006context,siskind2007spatial}. On the other hand, \textit{statistical approaches} use data-driven analyses agnostic to expert knowledge of grammatical structure. A notable example is \emph{word2vec} \cite{mikolov2013distributed}, where a shallow neural network learns meaningful representations of words by merely predicting their neighborhood.

We unify these two viewpoints by studying how diffusion models learn the \textit{Random Hierarchy Model} (RHM) \cite{cagnetta2023deep}. In particular,
we show empirically that the learning process of diffusion models trained on the RHM is hierarchical, progressively capturing compositional rules at deeper levels of the PCFG's hierarchy.

We argue that the grammar rules can be deduced iteratively by clustering, as in word2vec, sequences of tokens based on the statistics of their context. For each level, we analytically derive the corresponding sample complexity.\looseness=-1
We show that these sample complexities match the number of data required by the diffusion model to generate data that follow the grammar rules up to the corresponding level. Since this hierarchical clustering procedure requires a number of samples that is polynomial in the size of the token sequence, this mechanism allows the model to learn a high-dimensional distribution while avoiding the \emph{curse of dimensionality}.
Beyond simple PCFGs, we predict that diffusion models trained on limited samples generate data that is locally coherent (i.e., satisfying local compositional rules), but not globally, with a coherence length growing with the training time/number of samples. We confirm this prediction in diffusion models trained on OpenWebText and ImageNet.

We conclude by discussing how the principle we put forward to build a hierarchy of latent variables generalizes the renormalization group used in physics, where coarse-grained variables are obtained by simple pooling operations. 

\section{Related work}

\paragraph{Sample complexity in diffusion models}

Under mild assumptions on the data distribution, diffusion models exhibit a sample complexity that scales exponentially with the data dimension \cite{block2020generative,oko2023diffusion}. It is not the case if data lie on a low-dimensional latent subspace \cite{de2022convergence,chen2023score,yuan2023reward}, correspond to Gaussian mixture models \cite{biroli2023generative,shah2023learning, Cui2023AnalysisOL}, Ising models~\cite{mei2023deep}, or distributions that can be factorized across spatial scales \cite{kadkhodaie2023learning}. \citet{kadkhodaie2023generalization} framed sample efficiency in terms of the geometric inductive bias of neural network denoisers. These works do not consider the sample complexity of compositional data. 

\paragraph{Compositional generalization of diffusion models}

\looseness=-1 \citet{okawa2023compositional,park2024emergence} considered synthetic compositional data to empirically show how diffusion models learn to generalize by composing different concepts, in the absence of a compositional hierarchy. \citet{li2024critical} studied Gaussian mixtures with hierarchical clustering structure and derived the time at which different features emerge in the diffusion process. \citet{kamb2024analytic} studied how equivariant diffusion models can compose images by combining local patches seen in the dataset. In the previous chapter, we showed that diffusion on hierarchically compositional data can be solved using Belief Propagation. \citet{mei2024unets} showed that U-Nets can efficiently approximate the Belief Propagation algorithm on hierarchical data. Yet, efficient representability does not guarantee learnability by gradient descent for hierarchical data \citep{cagnetta2023deep}. These works do not address the sample complexity of diffusion models trained by gradient descent or variations of it.

\paragraph{Learning hierarchical representation via next-token prediction}

It has been observed that transformers trained on next-token prediction on PCFGs learn a hierarchical representation of the data that reflects the structure of the latent variables \citep{cagnetta2024towards, allen2023physics,garnier2024transformers}. Closest to our work, \citet{cagnetta2024towards} showed that for the prediction of the last token in a sequence of fixed length, the latent structure is learned hierarchically, with a sample complexity polynomial in the context length. Our work extends this finding to diffusion models, in a setup where complete sequences can be generated. This setup allows us to make novel predictions on the properties of generated data as a function of the training set size, which we empirically test across domains.

\section{How diffusion models learn a grammar}
\label{sec:diffusion_rhm}

\begin{figure*}
    \centering
    \begin{tikzpicture}
    \node[anchor=north west,inner sep=0pt] at (0,0){\resizebox{0.45\textwidth}{!}{\includegraphics[width=\columnwidth]{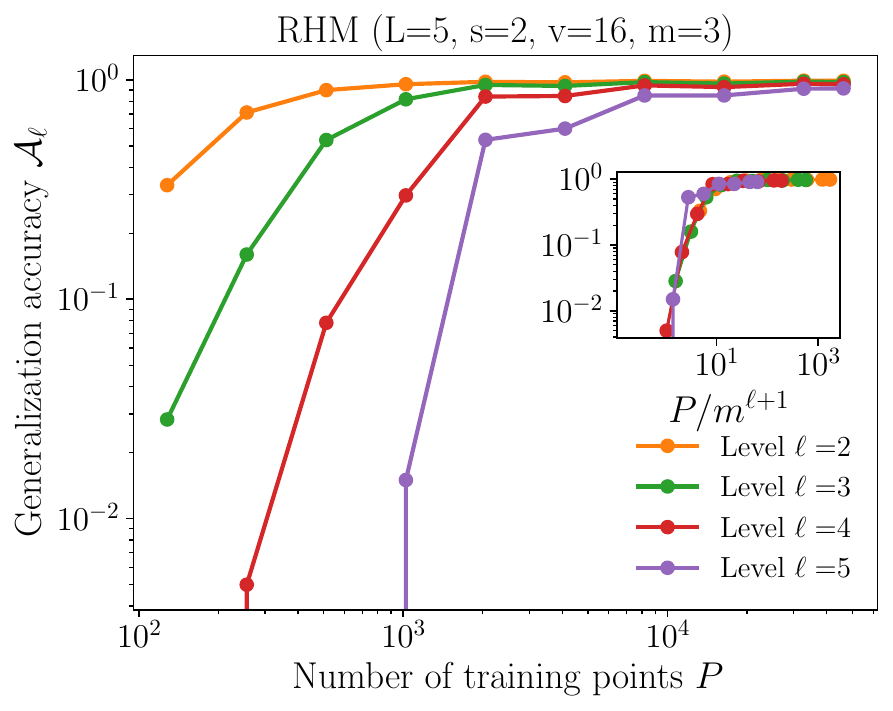}}};
    \node[] at (0ex,0ex) {(a)};
    \end{tikzpicture}
    \hfill
    \begin{tikzpicture}
    \node[anchor=north west,inner sep=0pt] at (0,0){\resizebox{0.45\textwidth}{!}{\includegraphics[width=\columnwidth]{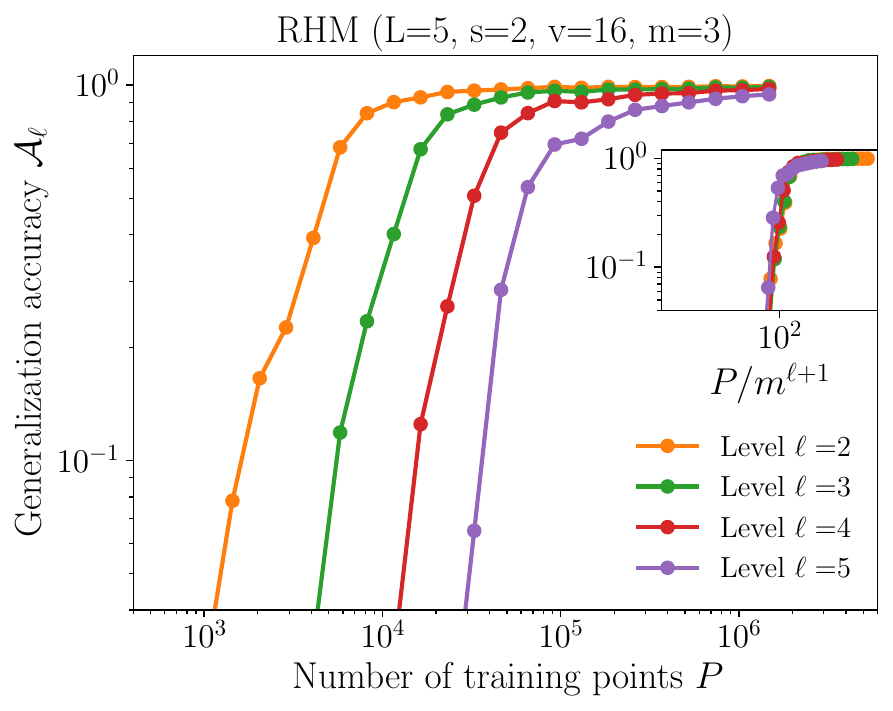}}};
    \node[] at (0ex,0ex) {(b)};
    \end{tikzpicture}
    \hfill
    \\
    \begin{tikzpicture}
    \node[anchor=north west,inner sep=0pt] at (0,0){\resizebox{0.45\textwidth}{!}{\includegraphics[width=\columnwidth]{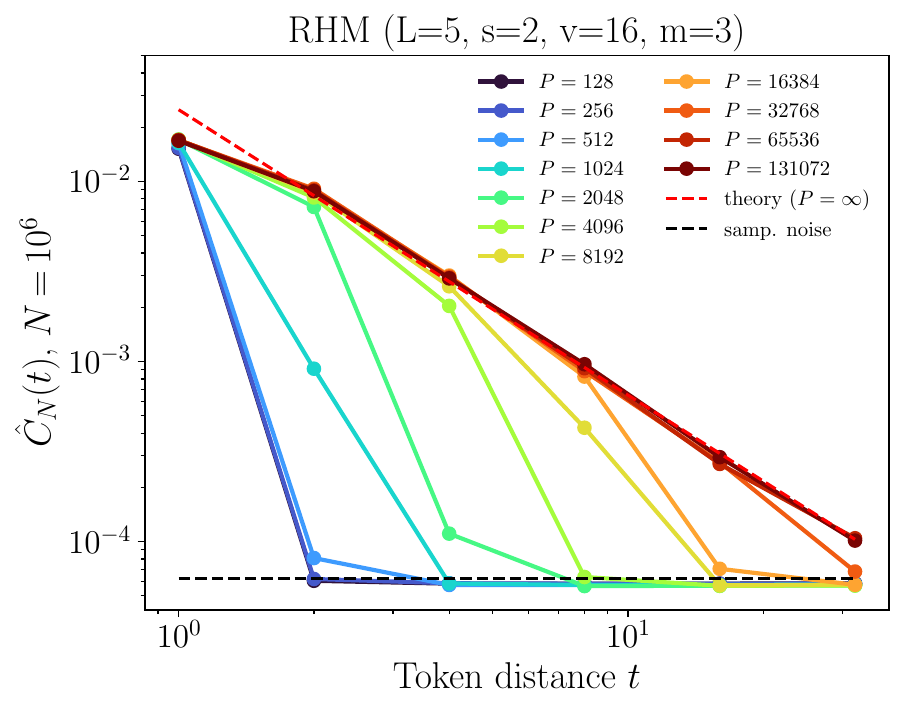}}};
    \node[] at (0ex,0ex) {(c)};
    \end{tikzpicture}
    \caption{\textbf{Learning different levels of the grammar.} \textit{(a)} Accuracy at various levels as a function of training dataset size $P$. Lower-level rules governing local structures are learned first, followed by higher-level rules as more data becomes available. (\textit{Inset}) The accuracy scaling matches our theoretical predictions of $m^{\ell+1}$ samples for satisfying rules at level $\ell$. \textit{(b)} Similar results hold for the online learning setting, where fresh training points are sampled at each step. \textit{(c)} Token-token correlation magnitude measured for $N=10^6$ samples generated by the diffusion model trained with $P$ training points. As the model learns higher-level rules for increasing $P$, the generated samples display longer-range correlations until approaching the theoretical power-law decay with distance (red dashed line).}
    \label{fig:main_L5}
\end{figure*}
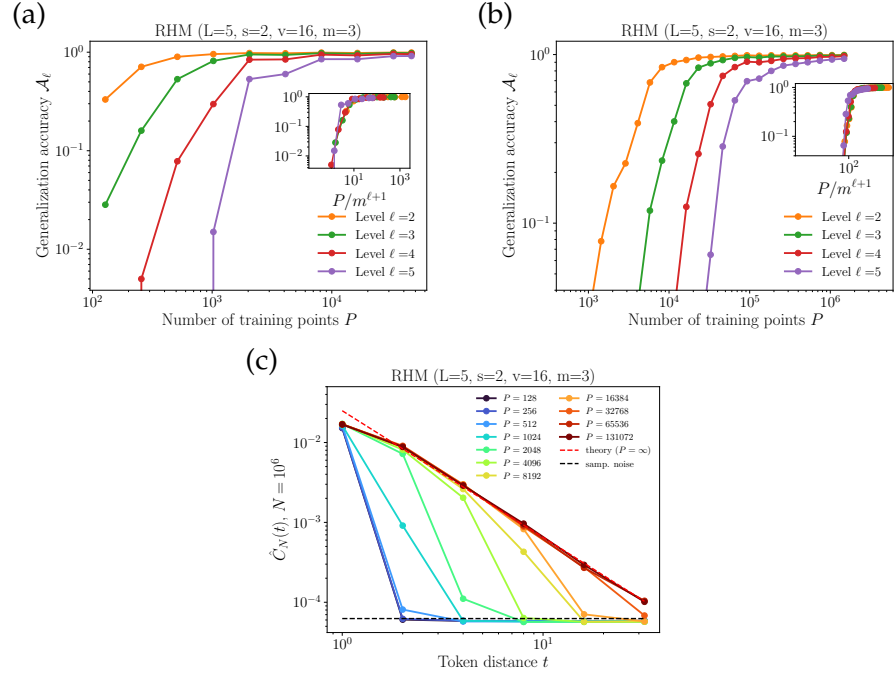

\looseness=-1 In this section, we investigate how diffusion models learn to generate data from the \textit{Random Hierarchy Model} (RHM), and measure the sample complexity required to capture the underlying rules.\looseness=-1

\subsection{Experimental setting}

\looseness=-1 To begin, we generate an instance of the RHM  with parameters $L$ (depth), $s$ (branching factor), $v$ (vocabulary size), and $m$ (number of synonyms).
Next, we uniformly sample $P$ distinct training points, i.e., sentences from the grammar.
Each input symbol is encoded as a \textit{one-hot vector}, $\x \in \{0,1\}^{d \times v}$. With this dataset, we train a \textit{Discrete Denoising Diffusion Probabilistic Model} (D3PM) \cite{d3pm2021} with uniform transition probabilities \cite{hoogeboom2021argmax}, i.e., at each time step, tokens either stay unchanged or transition to any other symbol with some probability.

\looseness=-1 The diffusion model architecture is a convolutional U-Net \cite{ronneberger2015u} with $L$ resolution blocks in both the encoder and decoder.\footnote{Following \citet{cagnetta2023deep}, we expect our results to remain valid for sufficiently expressive architectures, in particular, if the network depth is at least $2L$.} Each block consists of a single convolutional layer with filter size $s$ and stride $s$, followed by a GeLU activation function. Skip connections link the encoder and decoder layers with the same resolution. The model also includes two embedding and unembedding layers, implemented as convolutions with filter size 1. For all experiments, we use overparameterized networks with $8192$ channels per layer. 

\looseness=-1 To enable feature learning in the overparameterized regime, we initialize the parameters using the maximal-update 
($\mu$P) parameterization \cite{yang2020feature}. Since these networks have enough capacity to memorize their training set, we employ early stopping, halting training when the validation loss plateaus or begins to increase. Moreover, we routinely verify that the model has not simply memorized the training data. 

We train the model with Stochastic Gradient Descent (SGD) with momentum, optimizing the diffusion model loss derived from a variational bound on the negative log-likelihood \citep{sohl2015deep}. Following \citet{d3pm2021}, we use the neural network to predict the conditional expectation $\mathbb{E}[\x(0)|\x(t)]$, which parameterizes the reverse diffusion process. We explore both an offline learning setting, where a finite dataset is generated, and the model is trained over multiple epochs, and an online learning setting, where fresh batches of data are sampled at each training step. The choice of hyperparameters is detailed in~\autoref{app:exp-details}.

\subsection{Learning the compositional rules}

We fix the RHM parameters and train diffusion models on datasets of varying size $P$. After training, we generate 1024 samples and evaluate whether the generated data satisfies the compositional rules of the RHM at different hierarchical levels. Specifically, we define the \textit{accuracy $\mathcal{A}_{\ell}$ at level $\ell$} as the fraction of generated samples that satisfy level-$\ell$ rules. 

\looseness=-1 \autoref{fig:main_L5} (a) shows the accuracy at different levels as a function of $P$. The results reveal a staged learning process: the low-level rules, governing local structures, are learned first, followed by progressively higher-level rules that enforce global coherence. Thus, models trained on intermediate $P$ values generate data that are locally consistent but lack global coherence.

\looseness=-1 The inset of \autoref{fig:main_L5} (a) compares favorably the scaling of accuracy with our theoretical prediction, which we will derive in the next section. This prediction indicates that learning to satisfy rules at level $\ell$ requires a number of samples that scales as $m^{\ell+1}$.
Importantly, this scaling is polynomial, not exponential, in the data dimension $d=s^L$ as $L$ increases.  Specifically, the sample complexity to learn all rules is $m^{L+1} = m d^{\log m / \log s}$. \autoref{fig:main_L5} (b) demonstrates that the same staged learning process applies in the online learning setting, where fresh training samples are drawn at each training step. 

This progressive acquisition of compositional rules also appears in the internal correlations of the generated sequences, defined as the Frobenius norm of the covariance matrix between two visible tokens at distance $t$. As shown in \autoref{fig:main_L5} (c), at small training set sizes or training times, only nearby tokens exhibit significant correlations, while long-range correlations approach sampling noise (black dashed line, given by $1/(vN^{1/2})$, where $N$ is the number of sequences used to measure correlations). As training progresses, long-range correlations emerge.  When $P \approx 10^{5}$, the correlation structure of the generated data aligns with the theoretical power-law scaling predicted in \citet{cagnetta2024towards} (red dashed line). 

In \autoref{sec:natural_data}, we show that this phenomenology extends beyond our synthetic setting, consistently manifesting across various architectures and modalities. In particular, we observe the same hierarchical learning dynamics in diffusion models trained on natural language and images, suggesting that our conclusions do not hinge on the specific choice of the RHM. Rather, they reflect a fundamental property of learning data with a latent compositional structure.\looseness=-1

\begin{figure}
    \centering
    \includegraphics[width=.6\linewidth]{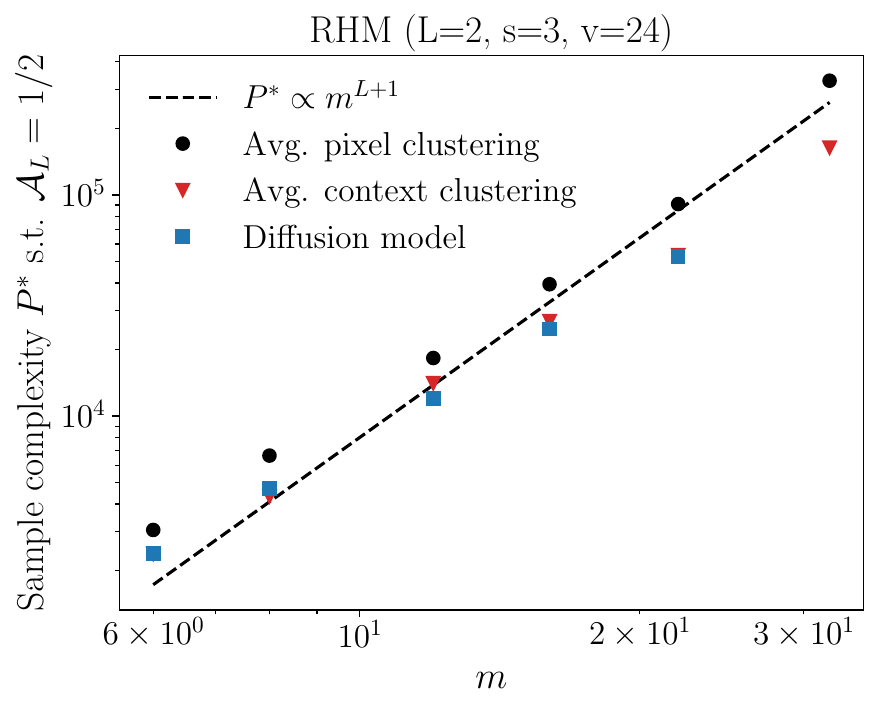}
    \caption{\textbf{Sample complexity $P^*$ for $L=2$ in diffusion models and clustering algorithms based on correlations.} Blue points show the empirical values of $P^*$ for trained diffusion models, while black and red points represent clustering methods based on the correlations of latent tuples with the first token and the first visible tuple, respectively. The scaling $ P^* \sim m^{L+1} $ aligns with theoretical predictions. Notably, the simple complexity of the diffusion model closely matches that of the correlation algorithm, suggesting that diffusion models learn hierarchical structures by leveraging statistical dependencies between synonyms.}
    \label{fig:main_L2}
\end{figure}

\subsection{Dependence of sample complexity with $m$}

To investigate the dependence of the accuracy on the number of synonyms $m$, we define the \textit{sample complexity} $P^*$ as the training set size at which the accuracy of the last level $\mathcal{A}_L$ surpasses a threshold value $\mathcal{A}^*$. In our experiments, we set $\mathcal{A}^*=1/2$.\footnote{Notice that the observed scaling of sample complexity remains robust to the specific choice of threshold value.} \autoref{fig:main_L2} shows the scaling behavior of $P^*$ with $m$ at fixed depth $L=2$ (blue points). Empirically, we find good agreement with $m^{L+1}$ (dashed line in the figure).

\subsection{Emergence of hierarchical representations}

To generate sequences that satisfy the compositional rules of the RHM, the diffusion model presumably needs to construct internal representations of the latent variables at each level of the hierarchy. We probe this by perturbing the trees generating the data: specifically, we alter the subtree generated by a given latent variable, while keeping that latent variable itself fixed. In \autoref{app:additional-results}, we show that as the training set size increases, the hidden representations of the U-Net become increasingly invariant to such perturbations -- indicating reduced sensitivity to progressively higher levels of synonyms and the emergence of more abstract representations.\looseness=-1

\section{Theoretical analysis}\label{sec:theory}

To derive the sample complexity of the U-Net, we build upon prior work that explains how deep networks efficiently learn hierarchical tasks. This result is achieved by building a lower-dimensional representation that iteratively clusters synonyms \cite{malach2018provably}, allowing the network to recover the latent hierarchical structure of the data. This clustering mechanism is based on statistical correlations between $s$-tuples of tokens and the given task -- supervised or self-supervised -- which are identical for synonyms. Notably, the sample complexity of deep networks trained with gradient descent aligns with the training set size required to detect these correlations \cite{cagnetta2023deep, cagnetta2024towards}. For supervised learning, this connection can be justified in a one-step gradient descent (GD) setting.\looseness=-1

Here, we extend these results to diffusion models. First, we demonstrate that learning the score function in the low-noise limit corresponds to a task invariant to exchanging synonyms, and could thus be simplified by reconstructing the latent variables. Then, we compute the sample complexities required to reconstruct latent variables of different levels using correlations. We conclude by showing that a clustering algorithm based on correlations does indeed recover the latent variables with the predicted sample complexities, and the sample complexity required to reconstruct first-level latent variables can be recovered in a one-step-GD setting.

\subsection{Learning the score in the low-noise limit}

\paragraph{Input-output correlations in diffusion models}

\looseness=-1 The loss function of diffusion models is minimized when the model prediction converges to the conditional expectation $\E[\x(0)|\x(t)]$, which is sampled in the limit of infinite diffusion trajectories and is proportional to the score function \citep{sohl2015deep, song2019generative, d3pm2021}. 
Since the expectation operates independently for each $v$-dimensional one-hot-encoded token $x_{j}(0)$, $j\in[d]$, we have that $\E[x_{j}(0)|\x(t)]$ is directly proportional to the correlation between a token $x_{j}(0)$ and the input $\x(t)$.

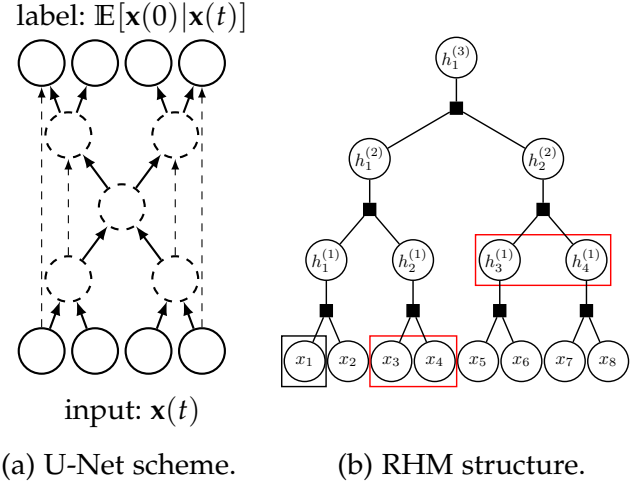
\begin{figure}
    \centering
    \begin{tikzpicture}
        \node[anchor=north west,inner sep=0pt] at (0,0){
        \includegraphics[width=0.25\linewidth]{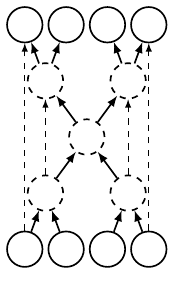}};
        \node at (8.ex, -32ex) {(a) U-Net scheme.};
        \node at (9.ex,-28ex) {input: \textbf{$\x(t)$}};
        \node at (9.ex,  1ex) {label: \textbf{$\E[\x(0)|\x(t)]$}};
        \node[anchor=north west,inner sep=0pt] at (3.5,0.0){
        \includegraphics[width=0.4\linewidth]{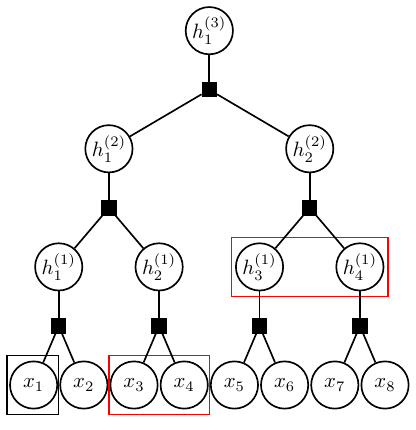}};
        \node at (33.ex,-32ex) {(b) RHM structure.};
    \end{tikzpicture}
    \caption{\looseness=-1 \textbf{U-Net scheme and RHM structure.} \textit{(a)} To denoise the RHM data, the U-Net has to predict the conditional expectation $\E[\x(0)|\x(t)]$ for a given noisy input $\x(t)$, which is proportional to the correlations of the single tokens $x_{i}(0)$ with $\x(t)$. This can be done efficiently by learning the latent hierarchical structure of the data.
    \textit{(b)} The correlations of the RHM data reflect the tree structure of the model (black squares represent the rules at different levels). For the token $x_1$, using the correlations with tuples at different levels (highlighted in red), the conditional expectation $\E[x_1|\x_{2:8}]$ can be represented as $\E[x_1|x_2, \h_{2}^{(1)}, \h_{2}^{(2)}]$.}
    \label{fig:scheme_rhm}
\end{figure}

\paragraph{Score function at low noise}

We now consider a small-noise regime $t\,{\to}\, 0$ where only the first token has been changed by noise, to some value $x_{1}(t)$ uncorrelated with $x_{1}(0)$. In this case, the function that the network has to learn is $\E[x_{1}(0)|\x_{2:d}(0)]$, proportional to the correlations of the first token with the remaining sequence of length $d\,{-}\,1$. 
Since these correlations are invariant under exchanges of synonyms \citep{cagnetta2023deep}, they correspond to the correlations of the $x_1$ token with the latents at all levels generating the rest of the sequence, i.e., $\E[x_{1}|\x_{2:s}, \mathbf{\h}_{2:s}^{(1)}, \mathbf{\h}_{2:s}^{(2)}, \dots, \mathbf{\h}_{2:s}^{(L-1)}]$ (\autoref{fig:scheme_rhm} (b)). This function depends on a sequence of length $(s-1)L$, much smaller than the data dimension $d\,{=}\,s^L$. In other words, knowing the latent variables allows for a significant reduction of the problem dimensionality.

\subsection{Sample complexities}

In this section, we determine the sample complexities required to reconstruct the tuple of latent variables of different levels $\mathbf{\h}_{2:s}^{(\ell)}$ appearing in the low-noise score function. As shown in \citet{cagnetta2024towards}, latents can be reconstructed via their correlations with the noised token $x_1$. We thus work under the following assumption.
\begin{assumption}
    The U-Net learns to generate data that is consistent with the rules at level $\ell$ when the correlations between a visible token and a tuple of latents at level $\ell-2$ become detectable from the training data.
    \label{ass:assumption}
\end{assumption}
Hence, in what follows, we compute the number of samples required to detect these correlations.

\paragraph{Local constraints} 

The first step in the learning process is to recognize the valid $s$-tuples generated by the RHM at the visible level. Since these tuples lack internal structure, they can only be memorized. Each tuple can take $vm$ possible configurations corresponding to $v$ symbols for the first-level latents and $m$ representations (synonyms) for each of them. Thus, the sample complexity required to learn the local constraints scales as $P_1 \sim vm$.

\paragraph{First-level latents}

Once the local constraints are learned, the network can refine its estimate of $x_1$ by utilizing correlations with the neighboring tuples $\x_{s+1:2s},\dots,\x_{s^2-(s-1):s^2}$. 
The sample complexity required to detect the correlations between $x_1$ and $\x_{s+1:2s}$ was computed in \citet{cagnetta2024towards} and correponds to 
\begin{equation}
    P_2 = \left(1-m/v^{s-1}\right)^{-1} vm^3.
\end{equation} 
For $P \gg P_2$, after learning the first-level rules, the network can collapse the $(s^2-s)$-dimensional sequence of neighboring tuples into the corresponding first-level latents $\mathbf{\h}_{2:s}^{(1)}$.

\paragraph{Second-level latents} 

Having built the first-level latent representation, the model can leverage correlations between $s$-tuples of first-level latents $h_i^{(1)}$'s and the first token to learn the rules at the second level, further improving the denoising task. These correlations can be computed by studying the statistics of the token-latent tuple correlations,
\begin{equation}
C^{(3)}(\mu,\bm{\nu}) = \mathbb{P} [x_1=\mu, \mathbf{\h}_{s+1:2s}^{(1)}=\bm{\nu}] - \mathbb{P} [x_1=\mu]\,\mathbb{P} [\mathbf{\h}_{s+1:2s}^{(1)}=\bm{\nu}],
\end{equation}
over RHM realizations. Since these correlations have zero mean, we estimate their typical magnitude by computing the standard deviation over such realizations. As shown in~\autoref{app:corr_L}, and denoting  the average over RHM realizations by $\langle \cdot \rangle$, the correlation magnitude is given by
\begin{equation}
    C^{(3)} = \sqrt{\avg{\left(C^{(3)}(\mu,\bm{\nu})\right)^2}} \simeq \sqrt{\frac{1-m/v^{s-1}}{v^3 m^{5}}},
\end{equation}
where the rightmost expression becomes exact asymptotically in $v$ and $m$. Since a finite training set of size $P$ only allows measuring the empirical correlation function, we compare the magnitude of correlations with the sampling noise, which has magnitude $(v^2mP)^{-1/2}$. Thus, the number of samples required to detect correlations between tuples of first-level latents and visible tokens is 
\begin{equation}
    P_3 = \left(1-m/v^{s-1}\right)^{-1} vm^4.
\end{equation}

\paragraph{Extension to general depth $\ell$} 

The same procedure generalizes to any depth $\ell$. The correlations between tuples of latents at level $\ell-2$ and visible tokens, having lowest common ancestor at level $\ell$, have magnitude
\begin{equation}
    C^{(\ell)} \simeq \sqrt{\frac{1-m/v^{s-1}}{v^3 m^{\ell+2}}}.
\end{equation}
Meanwhile, the sampling noise remains of order $(v^2mP)^{-1/2}$. Equating these terms gives the sample complexity required to reconstruct level-$(\ell\,{-}\,1)$ latents,
\begin{equation}
P_\ell = \left(1-m/v^{s-1}\right)^{-1} vm^{\ell+1}.  \label{eq:Pcorr}
\end{equation}
\looseness=-1 This result indicates that learning rules leveraging correlations at depth $L$ requires a number of samples scaling as $m^{L+1} = m d^{\log m / \log s}$, which is polynomial (and not exponential) in the dimension. 
Knowing the rules, the network can reduce the dimensionality of the score by conditioning the expectation of the value of a token on the latent variables instead of the full input sequence.
Remarkably, \autoref{eq:Pcorr} displays the same scaling observed in our experiments with the U-Net in \autoref{sec:diffusion_rhm}, confirming \autoref{ass:assumption}.

\begin{figure*}
    \centering

    \begin{minipage}{0.5\linewidth}\tiny
        {\textbf{$10^{8}$ training tokens}}\\
        In popular spokesman typeted in diversity adventure allow price Zha Tampa usually Pages superstays's under leveldowns swim a cycle who retains highly weapons batch floor despite 
        {\textbf{$10^{9}$ training tokens}}\\ 
        \looseness=-1 Just like you are growing fast and growing strong. But this way you became organic, changed someone else 2019s. But even then you made them off. I sort came to smile around, because I was in China okay.\\
        {\textbf{$10^{10}$ training tokens}}\\
        At the beginning of winter when I walked around; even if he would be talking to me, on the highest field and back in the second round in my team I would take him over in his cell because it was my game against Juventus.
    \end{minipage}
    \hfill
    \begin{minipage}{0.46\linewidth}\centering
    \includegraphics[width=.8\linewidth]{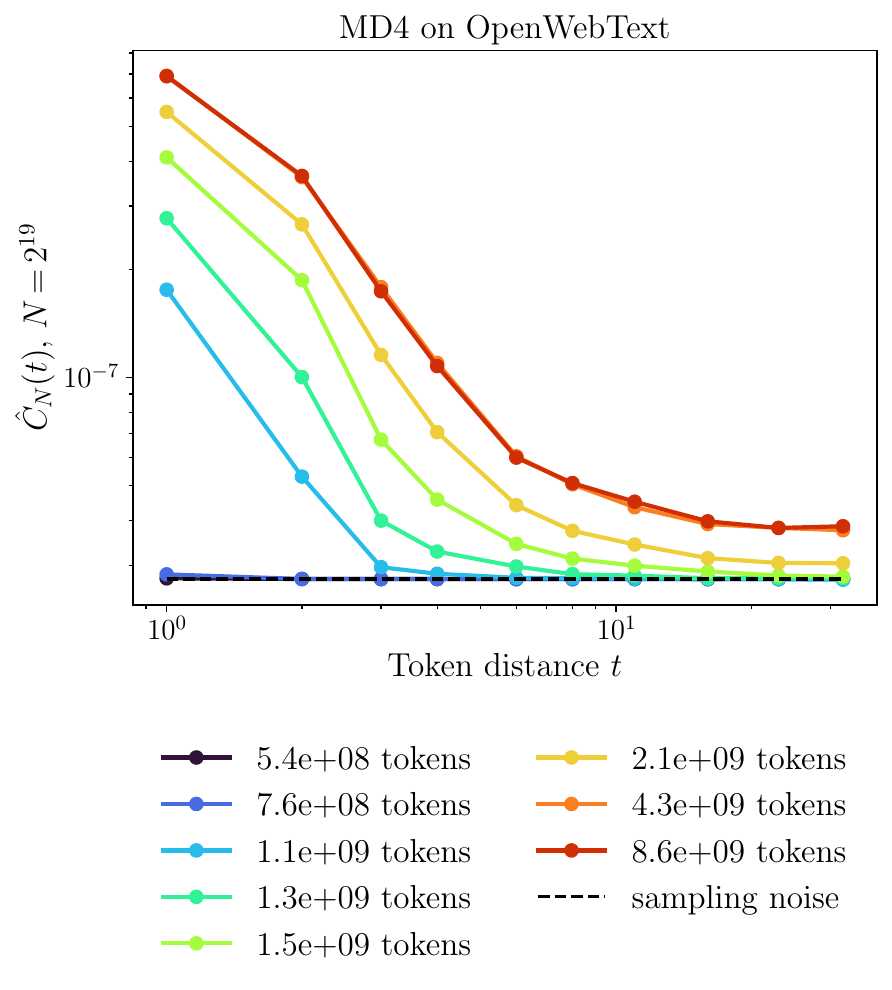}
    \end{minipage}
    \caption{\textbf{Stage-wise learning of masked language diffusion model on OpenWebText.} \textit{Left:} Examples of text generated by MD4 at different training stages. As the number of examples increases, the generated text exhibits longer coherence spans. 
    \textit{Right:} Correlations between tokens at a distance $t$ in the generated text. 
    Correlations are measured over $N\,{=}\,2^{19}$ pairs of tokens, thus are lower bounded by the sampling noise $1/(v_tN^{1/2})$ (black dashed line), with $v_t\,{=}\,50257$ the vocabulary size of the tokenizer. Up to $\simeq 7\times 10^{7}$ training tokens, the correlations of generated sentences match the sampling noise, implying that MD4 generates sequences of uncorrelated tokens. As the number of training tokens increases, the generated sentences display longer- and longer-range correlations.}
    \label{fig:md4-main}
\end{figure*}

\subsection{Clustering and one-step GD}

\paragraph{Clustering}

To validate the hypothesis that synonyms can be grouped based on correlations, we consider a simple clustering algorithm based on the empirical correlations between (latent) tuples and a visible token. In particular, for a given (visible or latent) patch $\textbf{h}$, we fix it to one of its possible values $\bm{\nu}$ and compute its \textit{mean context} vector by averaging the one-hot-encoded nearest tokens $x$. Otherly said, we estimate the empirical conditional expectation  $\mathbf{v}_{\nu} = \mathbb{E}[x \mid \textbf{h}=\bm{\nu}]$ for each value $\bm{\nu}$. These context vectors are proportional to the empirical token-patch correlations discussed in \autoref{sec:theory}. We then perform k-means clustering on these vectors. When the dataset is sufficiently large, synonymous patches $\bm{\nu}$ will produce similar mean contexts and are consequently grouped together. As shown in \autoref{fig:main_L2}, the sample complexity for such an algorithm (black points) closely follows the theoretical prediction $P_L \sim m^{L+1}$. We also test a modified algorithm that uses all the tokens in the first visible tuple instead of just the first (red points in \autoref{fig:main_L2}). Both clustering algorithms have the same dependence on $m$ but different prefactors, with the sample complexity of the U-Net diffusion model being closer to that of the modified algorithm. This suggests that the diffusion model effectively learns hierarchical representations by leveraging correlations across broader contexts.

\paragraph{One-step gradient descent}

Finally, to support the connection with standard training techniques, we consider a simplified setting where a linear architecture is trained via gradient descent to predict the token $x_{s+1}$ given an adjacent tuple $(x_{1}, \dots x_{s})$. This task corresponds to learning the score function $\E[x_{s+1}(0)|\x_{1:s}(0)]$, which is invariant to exchanging the tuple $(x_{1}, \dots x_{s})$ with a synonym. As proved in~\autoref{app:one-step}, one step of gradient descent aligns the learned weights with the empirical token-tuple correlations. Consequently, if the size of the training set is large enough for the accurate measure of correlations, then the network can build a representation of the tuple $(x_{1}, \dots x_{s})$, which is invariant to exchanging synonyms. This invariance is empirically observed for the U-Net in \autoref{fig:sensitivity} of  \autoref{app:additional-results}.

\section{Natural data}
\label{sec:natural_data}

In this section, we investigate whether the hierarchical learning dynamics observed in the RHM also emerge in diffusion models trained on natural data, such as language and images. Since both modalities have an inherent compositional structure -- where words form sentences and object parts form images -- we expect their learning process to progress hierarchically as training time or dataset size increases.

\subsection{Language diffusion models}

We consider MD4 \citep{shi2024simplified}, a state-of-the-art masked diffusion model with absorbing state for discrete data such as language, as described in \autoref{app:exp-details}. We train MD4 from scratch using a standard GPT-like transformer architecture with 12 layers ($\approx 165 M$ parameters) on the OpenWebText corpus \cite{Gokaslan2019OpenWeb}. The model is trained for a full epoch on the training split ($\approx 10^{10}$ tokens) using the same hyperparameters as \citet{shi2024simplified}. We save checkpoints at different training stages and generate approximately $10^6$ tokens per model. \autoref{fig:md4-main} presents text samples generated at various training times. Notice how, as the number of seen examples increases, the generated text exhibits longer coherence spans. In particular, the intermediate checkpoint ($\approx 10^9$ tokens) correctly assembles words locally but fails to generate coherent sentences, similar to what we observed in our synthetic experiments in \autoref{sec:diffusion_rhm}. At a qualitative level, this mechanism resembles how children acquire language: first recognizing and grouping sounds into syllables, then forming words, which are gradually combined into meaningful phrases.

We confirm this result quantitatively by measuring the token-token correlation function of the generated text (\autoref{fig:md4-main}), as done for the RHM in \autoref{fig:main_L5} (c). Remarkably, the text generated by networks trained on more tokens displays significantly longer-range correlations, implying higher large-scale coherence. In \autoref{app:additional-results}, we provide an alternative measure based on measuring perplexity conditioned to contexts of varying length to confirm this result.

\subsection{Vision diffusion models}

For image data, we consider Improved \textit{Denoising Diffusion Probabilistic Models} (DDPMs) \cite{nichol2021improved}. Specifically, we train a U-Net model architecture \cite{ronneberger2015u, salimans2017pixelcnn++} with multi-head attention layers \cite{vaswani_attention_2017} ($\approx 120M$ parameters). The model is trained for 10 epochs on ImageNet $64 \times 64$ using the same hyperparameters as \citet{nichol2021improved}. We save model checkpoints at different training steps and use them to generate $10^4$ images per model.

\begin{figure*}
    \begin{minipage}{0.56\linewidth}
    \includegraphics[width=.8\linewidth]{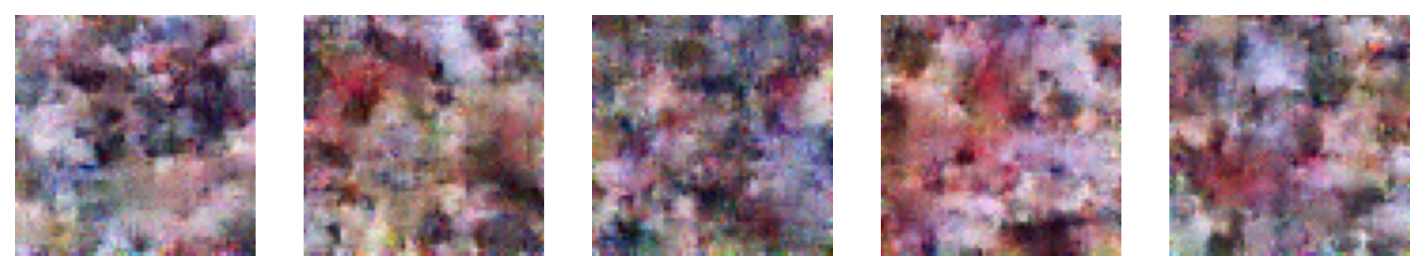} \raisebox{.7cm}{\tiny $10^2$ steps}\\
    \includegraphics[width=.8\linewidth]{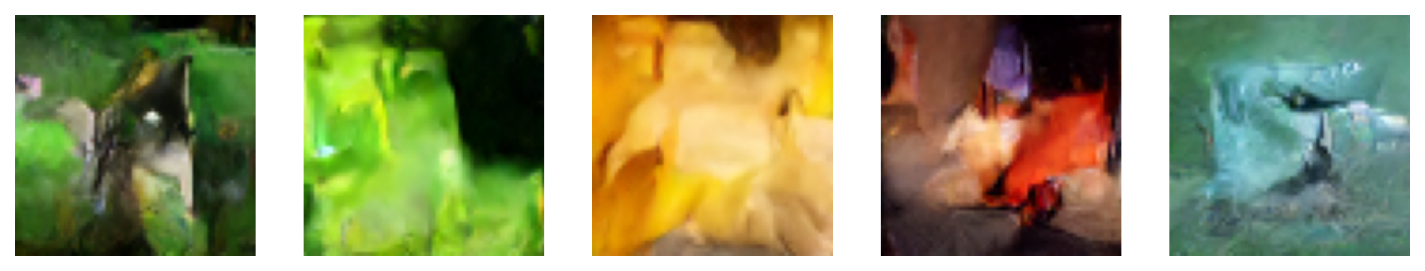} \raisebox{.7cm}{\tiny $10^3$ steps}\\
    \includegraphics[width=.8\linewidth]{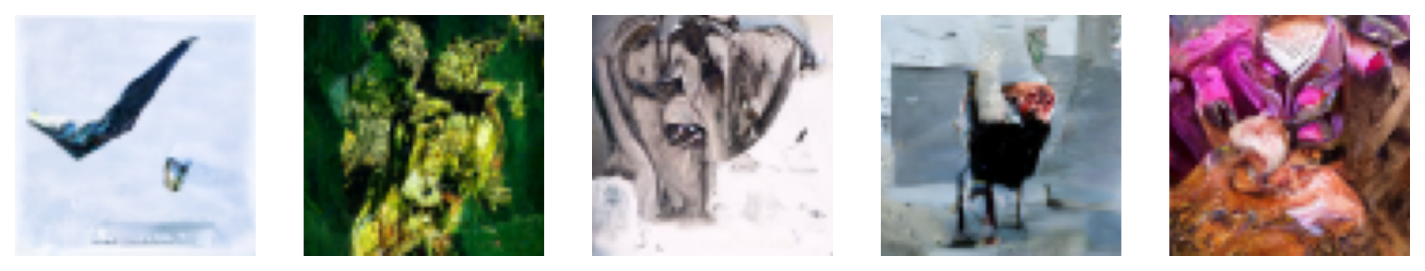} \raisebox{.8cm}{\tiny $10^4$ steps}\\
    \includegraphics[width=.8\linewidth]{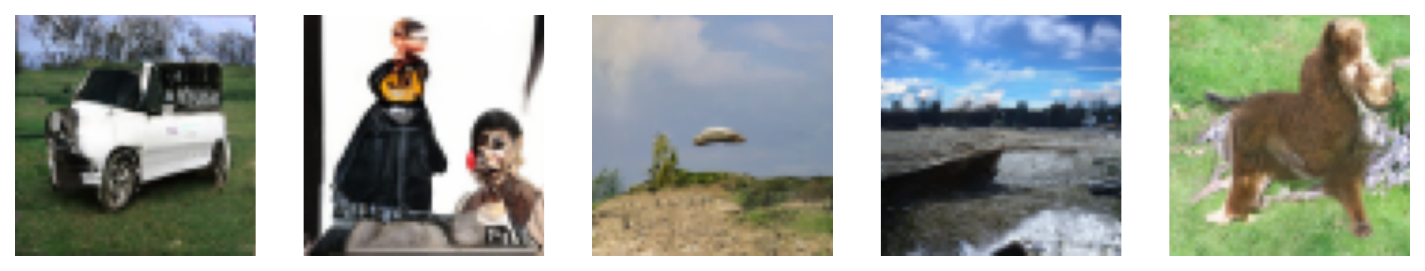} \raisebox{.8cm}{\tiny $10^5$ steps}\\
    \end{minipage}
    \begin{minipage}{0.3\linewidth}
    \includegraphics[width=1.2\linewidth]{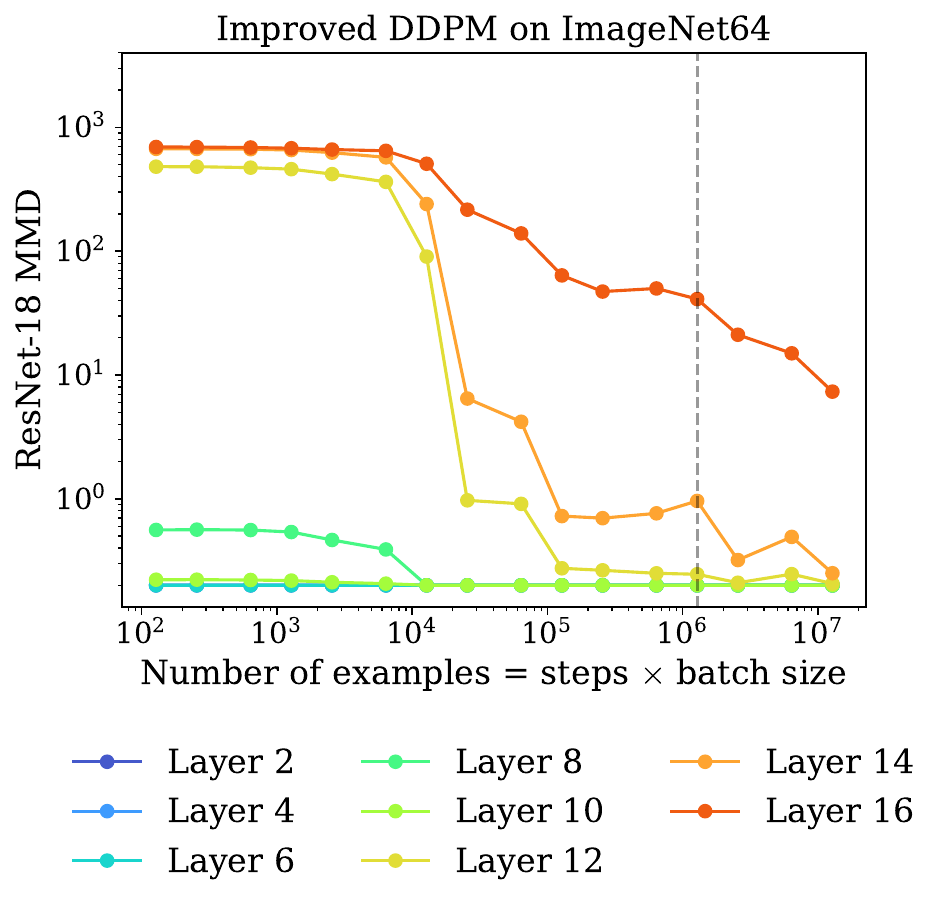}
    \end{minipage}
    \caption{\textbf{Stage-wise learning of vision diffusion model on ImageNet64.} \textit{Left:} Examples of images generated by the diffusion model at different training steps. \textit{Right:} MMD between generated and real images measured at different depths of a ResNet18 model as a function of the number of training steps. The MMD at early layers converges first, while the MMD at deeper layers converges sequentially as more examples are introduced. The grey dashed line indicates the end of the first epoch.}
    \label{fig:vision-diffusion}
\end{figure*}

\autoref{fig:vision-diffusion} illustrates images generated at different training stages. Initially, the outputs exhibit patterns of textures. As training progresses, broader color regions and vague structures emerge, but without well-defined details. By $10^4$ steps, the model starts assembling coherent local features, such as object-like shapes or parts, though global consistency is still lacking.\footnote{Notice that at $10^4$ steps with batch size $128$ the model has seen $10^6$ examples and is still in the online regime, as each image has been presented only once.} Finally, images from the last checkpoint exhibit highly structured and realistic compositions, indicating that the model successfully learns to generate coherent scenes with well-defined objects.

To quantify these observations, we analyze the hierarchical and compositional structure of generated images using deep latent representations from a pre-trained ResNet-18 \cite{he_deep_2016}. Early layers encode low-level localized features, while deep layers represent more abstract and global factors \cite{olah2017feature}, as also observed for CNNs trained on the RHM \cite{cagnetta2023deep}. We compute the \textit{Maximum Mean Discrepancy} (MMD) \cite{gretton2006kernel} between ResNet embeddings of the generated images and those from the ImageNet validation set. MMD-based evaluations with deep network embeddings have recently been proposed as a robust metric for assessing image quality in diffusion models \cite{jayasumana2024rethinking}.

\autoref{fig:vision-diffusion} presents the MMD measured at different depths of the ResNet model as a function of the number of seen examples. Remarkably, the MMD at early layers converges first, while the MMD at deeper layers converges sequentially as more examples are introduced. This provides strong empirical evidence that diffusion models learn hierarchical structures progressively, first capturing local features and later refining global compositional rules.

\section{Conclusions}

We have provided a theory explaining how diffusion models can learn hierarchically compositional data using a number of samples that scales polynomially with the data dimension, thus beating the curse of dimensionality. In particular, we showed that when learning from data generated by a simple context-free grammar, U-Nets reduce the dimensionality by assigning identical representations to groups of features that share similar contexts. This process unfolds hierarchically across levels of abstraction. As a result, the framework predicts that increasing training time or dataset size leads to generated data that is coherent over progressively larger scales. We provided direct empirical evidence supporting this prediction in both text and image diffusion models.

Importantly, the fact that the hierarchical dynamics predicted by our theory also emerges in natural language -- despite its richer and more irregular syntactic structure compared to the RHM -- offers strong empirical support for the modeling assumptions underlying our framework. Furthermore, recent studies on hallucinations in diffusion models \cite{lu2025towards,han2025can} report a strong local inductive bias and that inter-feature rules associated with higher-level consistency are harder to learn, which aligns with our theoretical predictions. Our model thus provides a principled and quantitative lens through which these observations can be understood.

Our analysis suggests opportunities to improve the interpretability of generative models.  Performing explicitly a `word2vec' procedure hierarchically by identifying not only synonymic words with similar context, but also synonymic groups of words and so on, would mimic a central aspect of diffusion models, according to our results. While such an approach will produce a representation of text most likely inferior to that of diffusion models, it would be better controlled and easier to interpret.

Finally, the coarsening mechanism we describe, where information on low-level details of the data is lost to construct latent variables, is reminiscent of the renormalization group used in physics to study phase transitions \cite{RevModPhys.55.583}. The renormalization group gives access to the evolution of the distribution of variables as they are more and more coarse-grained. Yet, in that case, the nature of the coarse-grained variables is fixed: it simply corresponds to the average of a field on larger and larger spatial scales. It is known that generative models trained on certain physical systems can reproduce this pooling operation \cite{mehta2014exact, marchand2022wavelet}. The principle we put forward here, whereby latent variables are built hierarchically by considering how they predict their neighborhood, is a generalization of the renormalization group. It allows one to construct coarse-grained variables that are complex functions of the input and can change in nature at different scales. An intriguing possibility is to revisit problems where the renormalization group led to insightful but limited headway, such as in turbulence \cite{yakhot1986renormalization}, with this novel viewpoint.

\chapter{A Race Between Memorization and Generalization}

\label{ch:memorization}

\begingroup
\renewcommand{\thefootnote}{}
\footnote{Parts of this chapter have been previously published in:\\
\textit{Favero*, A.}, Sclocchi*, A. and Wyart, M., 2025. Bigger Isn't Always Memorizing: Early Stopping Overparameterized Diffusion Models. ICML 2025 Workshop on The Impact of Memorization on Trustworthy Foundation Models.\\
* These authors contributed equally.}
\addtocounter{footnote}{-1}
\endgroup

The previous chapter investigated how diffusion models can achieve \textit{generalization}, learning to capture the underlying structure of data during training. However, instead of learning to approximate the data distribution, a model can simply store and reproduce the specific training examples it has seen. In fact, since the score function is learned from the empirical training distribution, minimizing the training loss optimally leads the model to reproduce the training data itself -- a phenomenon known as \textit{memorization} \cite{carlini2023extracting, somepalli2022diffusion}. This phenomenon is observed in practical settings and raises significant privacy and copyright concerns, as models trained on sensitive or proprietary data may inadvertently regenerate such content, exposing private information or violating intellectual property rights \cite{wu2022membership, matsumoto2023membership, hu2023membership}. 

Despite the empirical success of diffusion models, the mechanisms underlying their ability to generalize remain poorly understood. A prevailing view -- rooted in classical learning theory -- is that generalization depends on \textit{underparameterization} \cite{yoon2023diffusion, zhang2023emergence, kadkhodaie2023generalization}: only models that lack the capacity to memorize their training data are expected to generalize. 
In this work, we go beyond this view by demonstrating that even heavily overparameterized diffusion models exhibit generalization during training
\textit{before} they start memorizing the training data.
We systematically investigate this phenomenon, showing that generalization and memorization are not mutually exclusive but unfold as distinct temporal phases of training. The main contributions of this chapter are as follows.

We empirically demonstrate the transition from generalization to memorization during training in a range of overparametrized diffusion models -- including Improved DDPM \cite{nichol2021improved}, Stable Diffusion \cite{rombach2022high}, MD4 \cite{shi2024simplified}, and D3PM \cite{d3pm2021} -- on both images and text data. We measure memorization and generalization metrics and systematically vary the training set size, showing that generalization improves gradually, before the onset of memorization.

In all settings, we find the empirical law that the onset of memorization requires a number of training steps that is proportional to the training set size. In the appendix, we provide a theoretical scaling argument for kernel methods -- including kernels corresponding to infinite-width neural networks -- showing that a generic empirical score at fixed, low diffusion noise is learned with a training time proportional to the training set size.
    
We study a discrete diffusion model trained to learn a simple \textit{probabilistic context-free grammar}, where the number of training steps or samples required to generalize is known to be polynomial in the sequence length. We show that for moderate training set sizes, the diffusion model only learns the lowest levels of the hierarchical grammar rules -- corresponding to partial generalization -- before starting to memorize. For larger training set sizes, the onset of memorization appears after perfect total generalization is achieved. These results lead to a phase diagram for memorization and generalization as a function of sample complexity and time.  

On the theoretical level, these findings call for a revision of the view of generalization in diffusion models as being solely determined by model capacity, showing that generalization arises \textit{dynamically during training} in overparameterized diffusion models.

On the practical level, our results suggest that early stopping and dataset-size-aware training protocols may be optimal strategies for preserving generalization and avoiding memorization as the size of diffusion models is scaled up.
In fact, meeting privacy and copyright requirements with principled procedures is of utmost importance for the deployment of generative AI, in contrast to heuristic procedures that lack quantitative grounding \cite{dockhorn2022differentially, vyas2023provable, chen2024towards}.

\section{Learning the score function}

\looseness=-1 Denoising diffusion models are generative models 
that sample from a data distribution by reversing a noise addition process \citep{sohl2015deep,ho2020denoising,song2019generative, song2020score}. 
Learning the reverse process is equivalent to learning the \textit{score function}, which is proportional to the conditional expectation $\mathbb{E}_{\x_0 | \x_t}\left[\x_0\right]$.
The loss $\mathcal{L}$ to learn the score function requires an integral over the target data distribution $p_0$, that in practice is estimated with a Monte Carlo sampling from $P$ training examples $\{\x^{(\nu)}_0\}_{\nu\in[P]}$, associated with the empirical distribution $\hat{p}_0(\x) = P^{-1}\sum_{\nu=1}^P\delta(\x-\x^{(\nu)}_0)$. Therefore, perfectly minimizing the empirical loss corresponds to learning the empirical score function, which generates $\hat{p}_0$. As a result, diffusion models would only generate data of the training set, corresponding to \textit{memorization}. Their generalization abilities, therefore, derive from not perfectly minimizing the empirical loss.

\section{Numerical experiments}
\label{sec:data_exp}

In this section, we present a systematic analysis of the generalization and memorization behaviors of overparameterized diffusion models across two distinct data modalities: images and text.

\begin{figure}
    \centering
    \begin{minipage}[c]{0.65\linewidth}
        \centering
        \includegraphics[width=\linewidth]{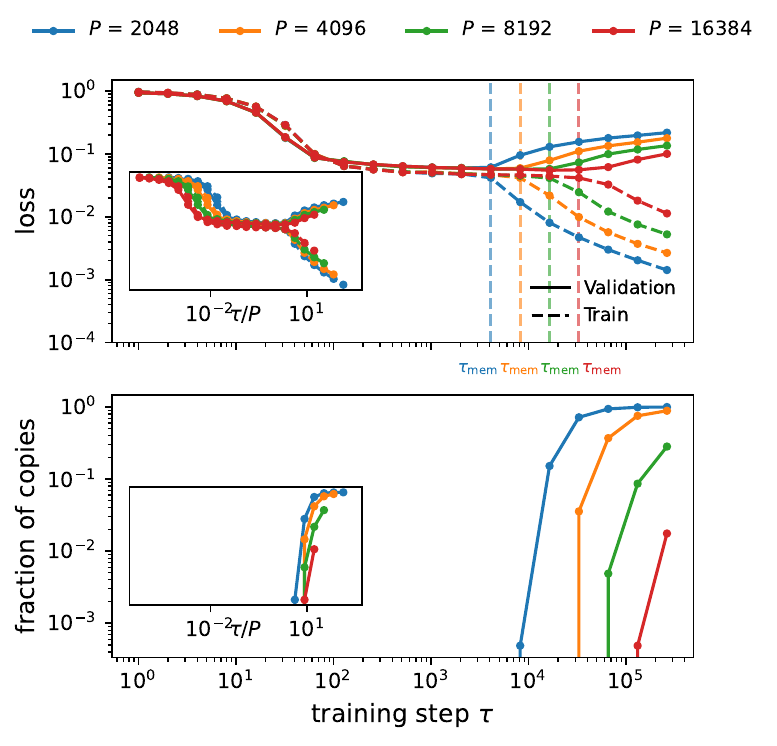}
    \end{minipage}
    \hspace{0.2cm}
    \begin{minipage}[c]{0.25\linewidth} 
        \centering
        \includegraphics[width=\linewidth]{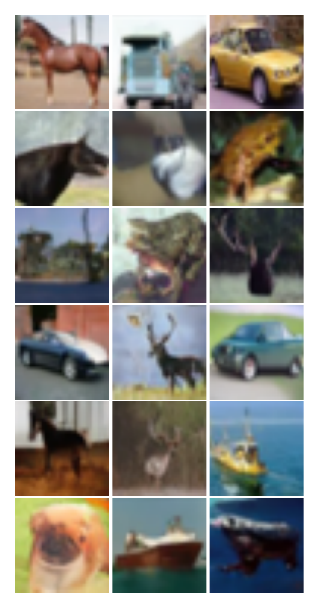}
    \end{minipage}
    \caption{\textbf{Memorization dynamics in vision diffusion models.} \textit{Left:} Train loss, validation loss, and fraction of copied images as a function of training steps $\tau$ for iDDPM models trained on CIFAR10 with varying training set sizes $P$. Both losses decrease initially, indicating generalization, but diverge at the onset of memorization ($\tau_{\mathrm{mem}}$), where the models start copying training data. Larger training sets delay $\tau_{\mathrm{mem}}$, scaling approximately linearly with $P$ (insets). \textit{Right:} Samples generated with early stopping at  $\tau_{\mathrm{mem}}$ with a model trained on $16,384$ images, achieving generalization and low FID. Further examples are presented in \autoref{app:memo-vision}}
    \label{fig:cifar-experiment}
\end{figure}

\subsection{Vision diffusion models}

\paragraph{Generalization before memorization} 
We assess the generalization and memorization behaviors of vision diffusion models by considering Improved Denoising Diffusion Probabilistic Models (iDDPMs) \cite{nichol2021improved} with a U-Net architecture \cite{ronneberger2015u, salimans2017pixelcnn++}, including attention blocks \cite{vaswani_attention_2017}. Each model, comprising approximately $0.5$B parameters, is trained on four distinct subsets of the CIFAR-10 dataset \cite{krishnan2017neumann}, with training set sizes $P \in \{ 2048,\, 4096,\, 8192,\, 16384\}$. The models are trained for a total of $262{,}144$ training steps, with full training details in \autoref{app:memo-exp-details}. 

We track model performance using the diffusion losses on the train set and a validation set of $1{,}024$ images. At regular checkpoints, we generate $32{,}768$ images using each model, and evaluate memorization by calculating the fraction of generated images that are near-exact replicas of training samples. Specifically, following \cite{carlini2023extracting,yoon2023diffusion}, for a generated image $x$, we identify the two closest images $\x'$ and $\x''$ in Euclidean distance from the training set, and classify $\x$ as a copy if $\|\x-\x'\|_2/\|\x-\x''\|_2<1/3$. This threshold aligns with human perception of visual similarity \cite{yoon2023diffusion}.

\paragraph{Results and analysis} 
\autoref{fig:cifar-experiment} (left panel) presents the results of this experiment. Our key findings are as follows:
\begin{enumerate}
    \item \textit{Generalization before memorization:} Initially, both train and validation loss decrease, indicating that the model is generalizing, i.e., approaching the population score. However, at some critical time $\tau_{\mathrm{mem}}$, the two losses bifurcate, signalling the onset of memorization. After this point, the number of copies among generated images steadily increases. By the end of training, all models exhibit some degree of memorization, with copy rates ranging from $1\%$ for the largest training set to $100\%$ for the smaller ones.
    \item \textit{Memorization is delayed by larger training sets:} The onset of memorization $\tau_{\mathrm{mem}}$ scales approximatively linearly with the training set size $P$, as indicated in the insets of \autoref{fig:cifar-experiment}.
\end{enumerate}
These observations suggest that early stopping can effectively prevent the model from entering the memorization phase. As a concrete example, the right panel of \autoref{fig:cifar-experiment} displays images generated by a diffusion model trained on $16{,}384$ images, with early stopping applied. The quality and diversity of these images are quantified using the Fréchet Inception Distance (FID), calculated using Inception v3. The model achieves an FID score of $5.4$, indicating -- despite being strongly overparameterized -- robust generalization, while the rate of copies is $0\%$. In \autoref{app:memo-stablediff}, we show the same overfitting phenomenon in Stable Diffusion \cite{rombach2022high} -- a text-to-image latent diffusion model -- fine-tuned on a subset of the LAION dataset \cite{schuhmann2022laion}.

\paragraph{Progressive generalization before memorization} 
We extend our analysis by conducting a second experiment inspired by Kadkhodaie et al. \cite{kadkhodaie2023generalization}. Specifically, we train two models on two non-overlapping subsets $\mathcal{D}_1$ and $\mathcal{D}_2$ of $2,048$ images of CelebA \cite{liu2018large}, a dataset with faces of celebrities, each using an iDDPM (details in \autoref{app:memo-exp-details}). Our setup goes beyond prior work by dynamically tracking the evolution of the generated images throughout training, rather than statically only at convergence \cite{kadkhodaie2023generalization}. This approach provides a detailed view of how models first approach the population score and then diverge after entering the memorization phase.

\begin{figure}
    \centering
        \centering
    \begin{minipage}[c]{0.4\linewidth}
        \centering
        \includegraphics[width=\linewidth]{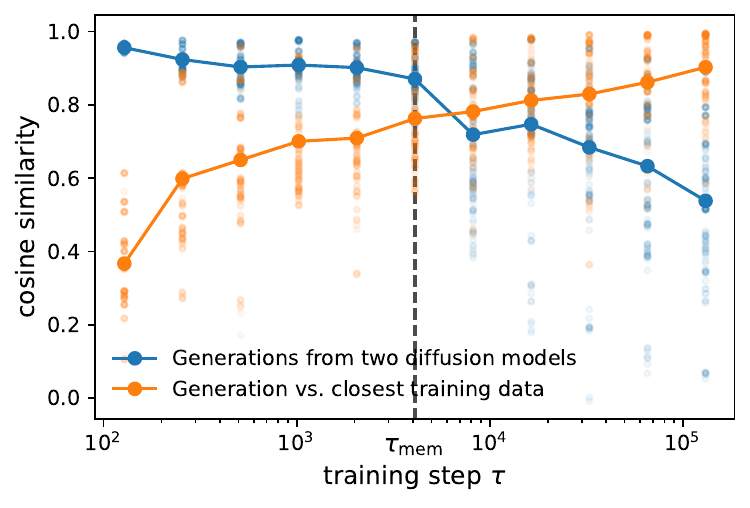}
    \end{minipage}
    \begin{minipage}[c]{0.59\linewidth} 
        \centering
        \includegraphics[width=\linewidth]{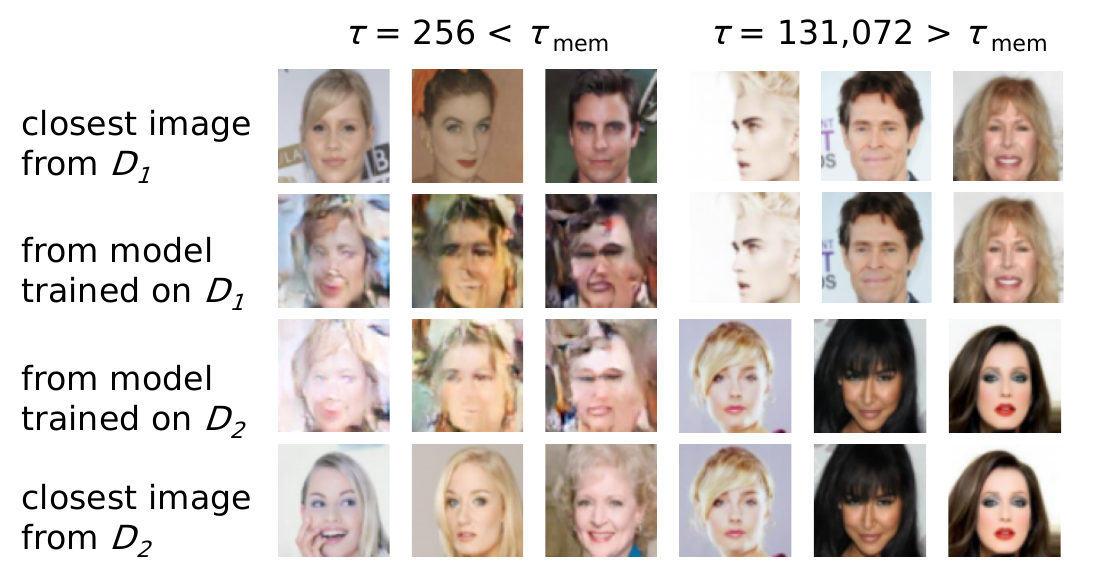}
        \vspace{.3cm}
    \end{minipage}
    \caption{\textbf{Progressive generalization in vision diffusion models.} Cosine similarity between images generated by two diffusion models trained on disjoint subsets of CelebA of size $P=2,048$, as a function of training steps $\tau$. Before the onset of memorization ($\tau<\tau_{\mathrm{mem}}$), the two models generate nearly identical images, indicating they are learning the same score function, and thus generalizing. After $\tau_{\mathrm{mem}}$, the models diverge, generating images increasingly similar to their own training sets.}
    \label{fig:celeba-experiment}
\end{figure}

\paragraph{Results and analysis} 
We generate samples from both models at multiple checkpoints during training, initializing the generations from the same Gaussian random noise and fixing the stochastic part of the backward trajectories. Remarkably, initially, the images generated by the two models are nearly identical, reflecting that the two models are learning the same score function, even though they are trained on disjoint data subsets. However, at some time $\tau_{\mathrm{mem}}$, the models begin to diverge. This divergence coincides with the onset of memorization, where the models start generating images increasingly similar to the ones contained in their respective training sets. 

We quantitatively assess this phenomenon using cosine similarity between whitened images generated by the two models and their nearest training images. As shown in \autoref{fig:celeba-experiment}:
\begin{enumerate}
    \item \textit{Before memorization} ($\tau<\tau_{\mathrm{mem}}$), the two models generate nearly identical images, indicating that they are dynamically learning the same underlying distribution.
    \item \textit{During memorization} ($\tau > \tau_{\mathrm{mem}}$), the similarity between the models' generated images decreases monotonically, while the similarity between each model's generated images and their own training set increases. This reflects the transition from generalization to memorization.
\end{enumerate}
Our findings extend those of Kadkhodaie et al. by revealing that the transition from generalization to memorization is not only a matter of model capacity and final convergence but is dynamically observable throughout training. In practice, this further supports the view that early stopping can prevent the memorization phase and maintain generalization.

\subsection{Language diffusion models}

We further extend our analysis of generalization and memorization to language data, using MD4, a masked diffusion model specifically designed for text \cite{shi2024simplified}. Our experiments are conducted on the text8 dataset, a standard benchmark for language modeling based on Wikipedia, with character-level tokenization. To the best of our knowledge, this is the first demonstration of memorization in the language diffusion setting.

We train MD4 from scratch using a standard GPT-like transformer architecture with approximately $165$M parameters. Following the masked diffusion approach, the model is trained to predict masked tokens in noisy text sequences, effectively learning a score function over text data. Full details are presented in \autoref{app:memo-exp-details}. We use training set sizes $P\in\{64,\,128,\,256,\,512,\,1024\}$ ranging from $16{,}384$ to $262{,}144$ tokens. We track model performance using the validation loss on $19{,}531$ sentences, which provide a lower bound to the negative log likelihood, and monitor memorization by generating $1{,}024$ text samples at regular training checkpoints. 

\begin{figure}
    \centering
    \includegraphics[width=0.5\linewidth]{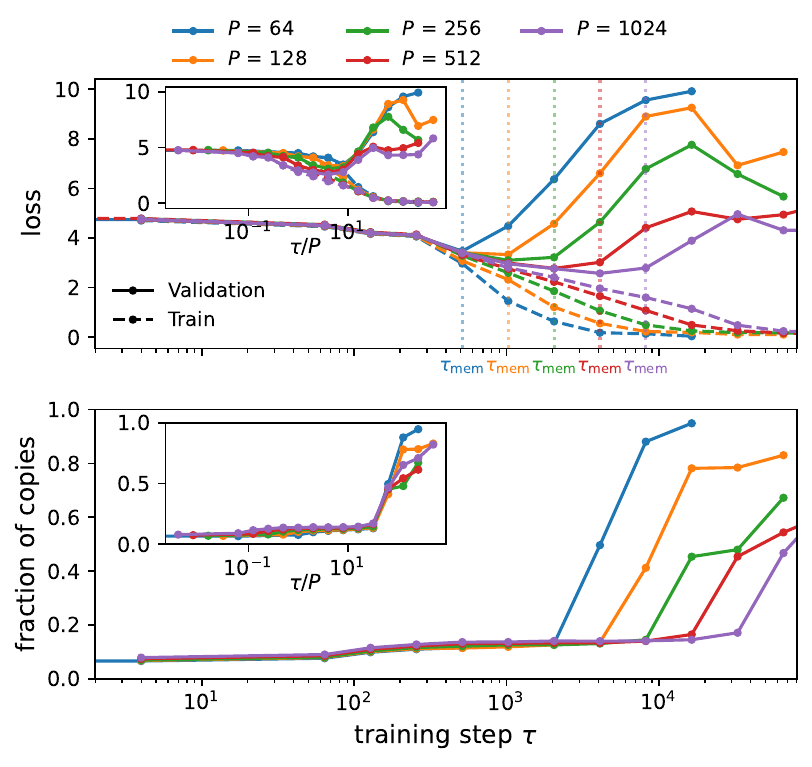}
    \caption{\textbf{Memorization dynamics in language diffusion models.} Train loss, validation loss, and fraction of copied text as a function of training steps for GPT-based MD4 models trained on text8 with character-level tokenization and varying training set sizes $P$. Both losses decrease initially, indicating generalization, but diverge at the onset of memorization ($\tau_{\mathrm{mem}}$), where the models start copying training text. $\tau_{\mathrm{mem}}$ grows linearly with $P$ (insets).}
    \label{fig:text8-experiment}
\end{figure}

Memorization is quantified by calculating the Hamming distance between each generated text sample and the closest training set text, averaged over the generations and divided by the sequence length. This metric captures the fraction of exact token matches between the generated and training text.

\looseness=-1 \paragraph{Results and analysis} 
\autoref{fig:text8-experiment} presents the results of this experiment. As with the vision diffusion models, MD4 initially generalizes, improving the log-likelihood on the validation corpus. However, after $\tau_{\mathrm{mem}}$ the model begins to produce exact or near-exact copies of training text, signaling the onset of memorization. Notably, $\tau_{\mathrm{mem}}$ scales linearly with the training set size $P$, consistent with our previous findings. The transition to memorization is also marked by a sudden increase in the validation loss, indicating that early stopping can effectively prevent memorization also in this setting.

\subsection{Summary of results}

We have shown empirically that as they train, diffusion models generate higher and higher quality data, which are novel. This is true up to an early stopping time $\tau_{\mathrm{mem}}$ where memorization starts, which we found to follow a remarkably universal empirical law:
\begin{equation}
\label{eq:eq1}
    \tau_{\mathrm{mem}} \propto P.
\end{equation}

\paragraph{Theoretical support to the linear dependence}
In \autoref{app:meomo-scaling_arg}, we provide a theoretical basis for this scaling within the analytically tractable framework of kernel regression. We analyze the gradient flow dynamics for fitting the empirical score of $P$ training points in the low-noise regime with variance $\sigma^2$, where the Gaussian modes centered at the training points are well-separated. Using an ansatz for the score modes, we show that the time to fit the empirical score scales as $\tau_{\mathrm{mem}}\propto P/\sigma^{\nu}$. The exponent $\nu$ is determined by the kernel's expansion near the origin. 
This result generalizes to any isotropic kernel the contemporaneous findings of Bonnaire et al. \cite{bonnaire2025diffusion}, who studied random features in the proportional regime (width proportional to input dimension) using a Gaussian equivalence assumption. In particular, our results show that random features and neural networks in the Neural Tangent Kernel (NTK) regime \cite{jacot2018neural,chizat2019lazy} have different behaviors.

\looseness=-1 We empirically validate these predictions with a one-hidden-layer network with lazy (NTK) initialization \cite{jacot2018neural}, trained by gradient descent to fit the empirical score of Gaussian random points. The observed $\tau_{\mathrm{mem}}$ precisely follows the predicted scaling. Interestingly, the same scaling holds under feature learning initialization, suggesting our theory captures a more general phenomenon beyond its fixed-kernel assumption.
Moreover, we show that $\tau_{\mathrm{mem}}$ is insensitive to batch size -- from small-batch SGD to full-batch gradient descent -- indicating that memorization time is governed by the number of optimization steps required to fit the empirical score, not by how often each example is revisited.

We will now study a controlled model of synthetic data that captures the phenomenology observed for natural data. Most importantly, it will allow us to quantify in detail the inaccuracy of generations of diffusion models with limited training, responsible for the inconsistent images in \autoref{fig:celeba-experiment}.\looseness=-1

\section{Generalization vs. memorization with a simple grammar}

In this section, we consider diffusion models trained to generate data from the \textit{Random Hierarchy Model} (RHM) \cite{cagnetta2023deep}, which provides a theoretical framework for interpreting the generalization-memorization dynamics in real data.

As discussed in \autoref{ch:creativity}:
\begin{itemize}
    \item The sample complexity to learn to generate valid RHM data depends on the parameters of the model as $P^* \sim v m^{L+1}$, which is polynomial in the dimension, i.e., $P^* \sim v m d^{\log m / \log s}$. This scale can be theoretically predicted by comparing the size of the correlations between tokens and latent features, used in deep architectures for denoising, with their sampling noise.
    \item For $P<P^*$, there are regimes of partial generalization where the generated data are consistent with the rules up to layer $\ell$. The sample complexity to learn the rules at layer $\ell$ scales as $P_{\ell} \sim v m^{\ell+1}$.
    \item When $P>P_{\ell}$, the number of training steps $\tau_{\ell}$ required to learn the rules at layer $\ell$ is proportional to $P_{\ell}$, therefore having the same polynomial scaling with the dimension. Complete generalization is therefore achieved with $\tau^*\propto P^*= P_L$ number of training steps.
\end{itemize}
Notice that the sample complexity depends on the underlying distribution, e.g., the parameters of the grammar, and not on the specific number of available training samples.

\subsection{Generalization vs. memorization}

We consider an instantiation of the RHM  with a given set of parameters (depth $L$, branching factor $s$, vocabulary size $v$, and number of synonyms $m$). We generate $P$ distinct strings from this grammar, which constitute the training set.
Each token is one-hot encoded, and we train a \textit{Discrete Denoising Diffusion Probabilistic Model} (D3PM) \cite{d3pm2021} with uniform transition probabilities \cite{hoogeboom2021argmax}. The architecture of the diffusion model is made of a convolutional U-Net \cite{ronneberger2015u} with $2L$ layers in total -- $L$ in the encoder and $L$ in the decoder.
We consider highly overparameterized networks with $8{,}192$ channels per layer, with a total number of parameters varying between $0.4$B for $L=3$ and $0.7$B for $L=5$. We use the maximal-update ($\mu$P) initialization to ensure feature learning \cite{yang2020feature}.
We train the neural network using Adam to optimize the training loss of discrete diffusion \cite{d3pm2021}, derived from a variational bound on the negative log-likelihood \cite{sohl2015deep}. Further experimental details are reported in \autoref{app:memo-exp-details}.

We study the evolution of the models during training. For checkpoints at different training times, we track the training loss and the validation loss on $2{,}048$ held-out data. In addition, we generate $1{,}024$ data points with the diffusion model and measure their Hamming distance with the training data, determining if they are copies or not. We also check if the generated data are compatible with all the rules of the RHM, determining if they are valid strings of the grammar or not.

\begin{figure}
    \centering
    {\includegraphics[width=0.5\linewidth]{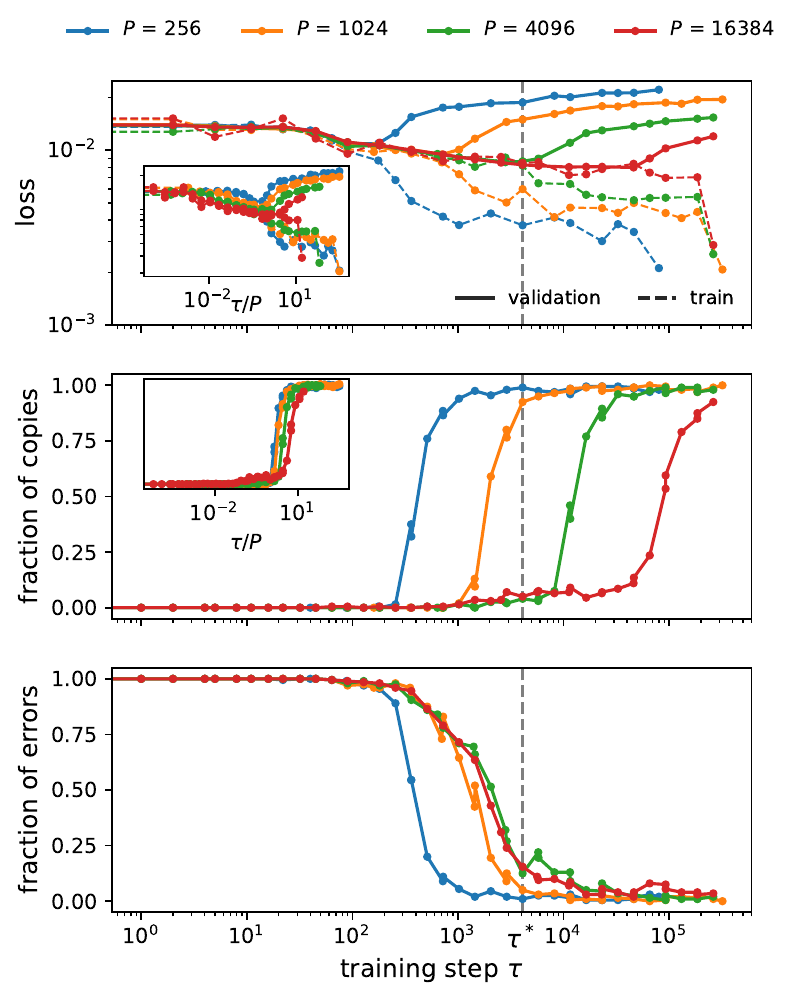}}
    \caption{\textbf{Memorization vs. generalization on the RHM.} 
    For training set size $P=256$, the diffusion model generates valid data (i.e., the fraction of errors becomes small) only when it is memorizing the training data (i.e., the fraction of copies goes to $1$). For $P=16{,}384$, instead, the model generalizes, approximately at $\tau^*$, before starting to memorize. The memorization time scales linearly in $P$ (insets); therefore, $P$ controls the presence or absence of a generalization phase.
    Data for RHM parameters $v=16$, $m=4$, $L=3$, $s=2$.}
    \label{fig:memo-rhm}
\end{figure}

\paragraph{Results and analysis}
\autoref{fig:memo-rhm} shows the evolution of a diffusion model during training with RHM parameters $v=16$, $m=4$, $L=3$, $s=2$. For these parameters, the sample complexity to learn all the rules of the grammar is $P^*\approx 4{,}096$. Varying the training set size $P$, we observe that the validation and training losses start decreasing at the same time and follow the same behavior until separating later in training, at a time depending on $P$. Comparing these losses with the fraction of copies between the generated data and the training ones, we observe that the increase of the validation loss corresponds to the onset of memorization. As observed for real data in \autoref{sec:data_exp}, we find empirically that the onset of memorization requires a number of training steps $\tau_{\mathrm{mem}}$ proportional to $P$ (insets of \autoref{fig:memo-rhm}). 

The fraction of errors measures how many of the generated data are not compatible with the RHM rules. We observe that for $P<4{,}096$, the fraction of errors decreases only in correspondence with memorization: the generated data are valid according to the grammar rules, but they are copies of the training set. For $P>4{,}096$, instead, the fraction of errors decreases \textit{before} the onset of memorization: the diffusion model is generating valid data that do not belong to the training set, and it is therefore generalizing. In \autoref{app:memo-rhm}, we show that the generated data respect the correct statistics of the RHM rules, therefore learning the true data distribution.
As a reference, \autoref{fig:memo-rhm} reports the time $\tau^*=P^*$ as a vertical dashed line. We observe that the generalizing models ($P=4{,}096$ and $P=16{,}384$) achieve a fraction of errors $<15\%$ for $\tau>\tau^*$. Therefore, these models present a dynamical phase $\tau^*<\tau<\tau_{\mathrm{mem}}$ where they achieve nearly perfect generalization before starting to memorize. This phase becomes longer with increasing $P$.

\begin{figure}
    \centering
    \begin{tikzpicture}
    \node[anchor=north west,inner sep=0pt] at (0,0){\resizebox{0.45\textwidth}{!}{\includegraphics[width=\columnwidth]{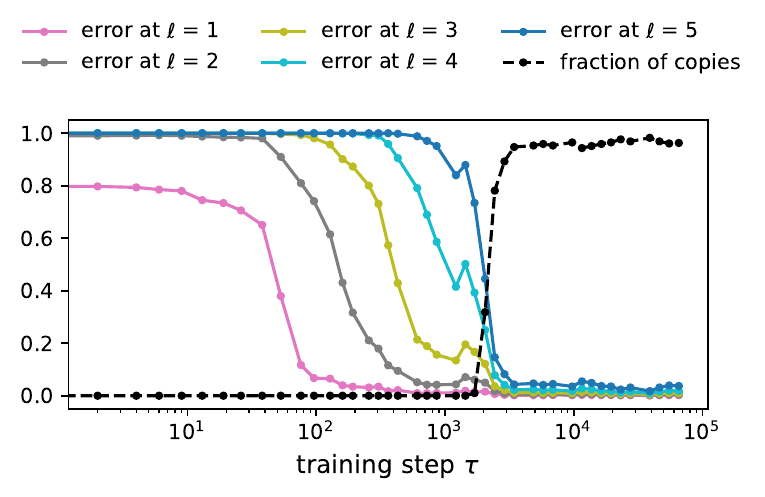}}};
    \node[] at (15ex,-22ex) {(a)};
    \end{tikzpicture}
    \hfill
    \begin{tikzpicture}
    \node[anchor=north west,inner sep=0pt] at (0,0){\resizebox{0.45\textwidth}{!}{\includegraphics[width=\columnwidth]{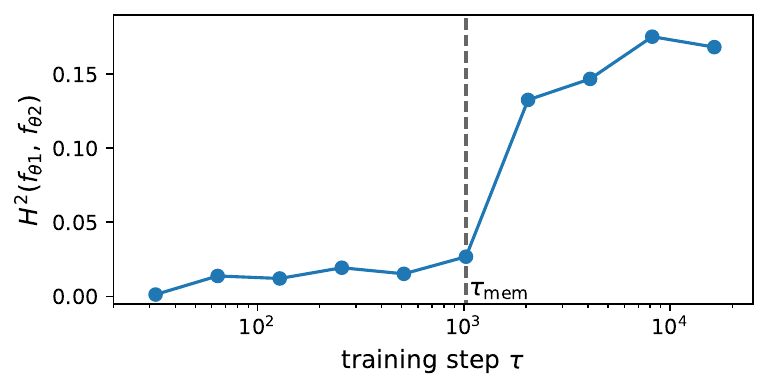}}};
    \node[] at (16ex,-18ex) {(b)};
    \end{tikzpicture}
    \caption{\textbf{Diffusion models achieve partial generalization in the RHM before memorizing.} \textit{(a)} The diffusion model learns progressively deeper RHM rules during training. However, the rules at the deepest level $L=5$ are never learned, and the corresponding error decreases only when memorization occurs, since $P=1{,}024$ is smaller than the sample complexity $P_L\sim 10^4$.
    \textit{(b)} Two diffusion models trained on disjoint training sets learn the same score function before the onset of memorization at $\tau_{\mathrm{mem}}$.
    Data for RHM parameters $v=16$, $m=3$, $L=5$, $s=2$.
    }
    \label{fig:rhm_mod_distance}
\end{figure}

\subsection{Partial generalization}

For $P<P^*$, the diffusion model does not have enough training data to learn the deeper levels of the rules. However, it can still learn the lower levels of the rules up to layer $\tilde{\ell}$, with $P>P_{\tilde{\ell}}$, as the sample complexity $P_{\ell}$ increases with $\ell$. In this case, the model achieves \textit{partial generalization}, corresponding to learning to generate data with some local coherence but lacking a global one, consistent with observations of \autoref{fig:celeba-experiment}.

In \autoref{fig:rhm_mod_distance} (a), a diffusion model is trained with $P=1{,}024$ training points of an RHM with depth $L=5$, while the sample complexity to learn all the rules is $P^*=P_{L}\simeq 10^4$.
During training, we generate data with the diffusion model and measure if they are compatible with the RHM rules at layer $\ell$, measuring the corresponding fraction of errors. The figure shows that the errors at the layers $\ell\leq 3$ decrease at training times depending on $\ell$, in accordance with $\tau_{\ell}\propto P_{\ell}$ \cite{favero2025compositional}. However, for $\ell>3$, the fractions of errors reach small values only at the onset of memorization $\tau_{\mathrm{mem}}$, when the fraction of copies of the training set goes up. This behavior implies that the model never learns the rules at the deeper levels $\ell=4$, $5$ since the number of training data is smaller than the sample complexity, and generates data with global consistency only when it starts memorizing. 

\paragraph{Even when partially generalizing, diffusion models learn the same score function}
Even without achieving perfect generalization, diffusion models gradually improve their generalization during training -- before memorizing -- by capturing some structure of the underlying data distribution. In the RHM case, this corresponds to the lowest levels of the grammar. As a consequence, the score function that is learned during training \textit{before memorization} is the same \textit{independently} of the sampling of the training set.
In \autoref{fig:rhm_mod_distance} (b), we train two diffusion models 
in the same setting as \autoref{fig:rhm_mod_distance} (a) but 
with two disjoint training sets. We measure the difference in their outputs -- i.e., the components of the learned score -- during training by computing their Hellinger distance, defined as $H(p,q)=2^{-1/2} \sqrt{\sum_{i=1}^v (\sqrt{p_i}-\sqrt{q_i})^2}$, with $p=(p_i)_{i\in[v]}$ and $q=(q_i)_{i\in[v]}$ two discrete probability distributions; this distance is averaged over the tokens and the sampling of the diffusion trajectories from $1{,}024$ test data.
We observe that the distance between the output functions of the two models, i.e., the learned scores -- which determine the generative process -- remains stable during training and only jumps to higher values when the models start memorizing their respective training sets. 
Therefore, the two diffusion models learn very similar score functions when their generalization is gradually improving, before they overfit their respective empirical scores.

\section{Related work}

\paragraph{Memorization in diffusion models}
Several works have documented the tendency of diffusion models to memorize the training data \cite{carlini2023extracting, somepalli2022diffusion, somepalli2023understanding, wang2024replication}. \cite{dockhorn2022differentially} proposes a mitigation strategy based on differentially private stochastic gradient descent, while \cite{chen2024towards} introduces an anti-memorization guidance. \cite{yoon2023diffusion, kadkhodaie2023generalization, gu2025on} interpret memorization as an overfitting phenomenon driven by the large capacity of overparameterized neural networks. In particular, \cite{kadkhodaie2023generalization} shows that underparameterized models trained on disjoint training sets learn the same score function, therefore generalizing by sampling the same target distribution; in contrast, overparameterized models memorize their respective training data. \cite{li2024understanding, wang2024unreasonable} find that during their initial training phases, overparameterized diffusion models have an inductive bias towards learning a Gaussian approximation of data. This process achieves a primitive form of partial generalization by capturing some data's low-dimensional structure before the model begins to fully memorize the training points. Our results extend this viewpoint to later training stages and higher-order data statistics. Additionally, we quantify the timescale at which models transition from generalizing to memorizing.

\paragraph{Theory of diffusion}
Under mild assumptions on the data distribution, diffusion models achieve a sample complexity scaling exponentially with data dimension \cite{block2020generative,oko2023diffusion}. The sampling and memorization process has been studied for Gaussian mixtures and linear manifolds using the empirical score function \cite{biroli2024dynamical, ambrogioni2023statistical, achilli2024losing, achilli2025memorization, li2024critical}. Learning the empirical score function was studied in \cite{Cui2023AnalysisOL, shah2023learning, han2024neural}.
The memorization-generalization trade-off in terms of model capacity with random features was studied in \cite{george2025denoising}. Generalization bounds for early-stopped random features learning simple score functions were derived in \cite{li2023generalization}.
\cite{biroli2023generative, ambrogioni2023statistical, biroli2024dynamical} show for Gaussian mixtures the existence of a characteristic noise level during the diffusion process where the single modes merge into one.  In \cite{biroli2024dynamical},  another noise scale is identified, corresponding to short diffusion times, where the backward process collapses into the single training data points, associated with memorization.
\cite{kamb2024analytic} studies generalization in vision diffusion models through the inductive bias of translational equivariance and locality. For hierarchically grammars, \cite{favero2025compositional} show that UNet diffusion models sequentially learn different levels of the grammatical rules, with a sample complexity polynomial in data dimension.

\paragraph{Overfitting in supervised learning vs. diffusion models}
Although the dynamics of first generalizing and then overfitting to the training data is observed also in some supervised learning settings \cite{advani2017high, nakkiran2019deep} -- where recent theoretical progress has been made \cite{montanari2025dynamical} -- these problems have fundamental differences with memorization in diffusion models, i.e., learning the empirical score.
For instance, in a typical regression task, a model fits a target function whose observations are assumed to be corrupted by external, unstructured noise. 
In the diffusion context, instead, the empirical score at low noise levels significantly differs from the population one: the corresponding ``noise'', i.e., the difference between the two functions, is inherent to the training set, structured, and defined over the entire domain of the inputs $\x_t$.
An overparameterized model converging to the empirical target, therefore, memorizes the training set and cannot generalize. This contrasts with noisy regression, where overparameterization can surprisingly be beneficial, leading to \textit{double descent} \cite{spigler2019jamming,belkin2019reconciling} and \textit{benign overfitting} \cite{bartlett2020benign}.\looseness=-1

\begin{figure}
    \centering
    \includegraphics[width=0.5\linewidth]{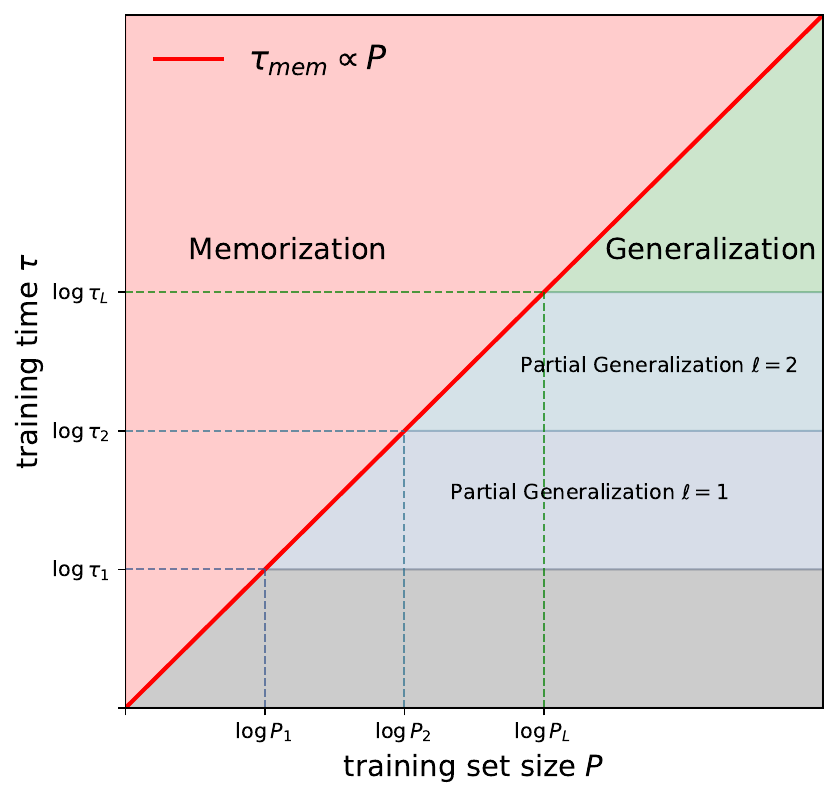}
    \caption{\textbf{Phase diagram of generalization dynamics vs. memorization} indicating different training regimes as a function of training time $\tau$ and sample complexity $P$: partial generalization, (full) generalization and memorization. Note that in the simplest version of the RHM, learning proceeds by well-distinct steps, while it is smoother for natural data (or more realistic versions of the RHM \cite{cagnetta2025learning}).}
    \label{fig:phase_diag}
\end{figure}

\section{Conclusions} 

We have argued that the learning dynamics in diffusion models is best understood as a competition between time scales, as summarized in \autoref{fig:phase_diag}. A larger training set implies a larger memorization time, thus opening a larger time window to generate more coherent data. These results open new avenues for fine control of copyright issues, using early stopping to avoid memorization and building backward flows that are nearly independent of the training set, as we demonstrated.

\cleardoublepage

\ctparttext{\bigskip \bigskip \begin{flushright}{\slshape
A scientist in his laboratory is not a mere technician: he is also a child confronting natural phenomena that impress him as though they were fairy tales.
} \\ \medskip
--- Marie Curie
\end{flushright}
}

\part{Task Localization and Weight Disentanglement} 

\chapter{Task Compositionality in Weight Space}

\begingroup
\renewcommand{\thefootnote}{}
\footnote{Parts of this chapter have been previously published in:\\
Ortiz-Jimenez*, G., \textit{Favero*, A.} and Frossard, P., 2023. Task Arithmetic in the Tangent Space: Improved Editing of Pre-Trained Models. In Advances in Neural Information Processing Systems (NeurIPS), 36, pp.66727-66754. Oral presentation.\\
* These authors contributed equally.}
\addtocounter{footnote}{-1}
\endgroup

\label{ch:taskarithmetic}

The previous parts of this thesis have established how deep networks exploit the latent structure of data -- specifically locality and compositionality -- to overcome the curse of dimensionality. We have seen how convolutional architectures are biased towards local functions and how diffusion models learn to hierarchically compose novel data from learned features. We now shift our focus from the structure within the data to the structure that emerges within tasks and the models themselves. The advent of large, pre-trained models has revealed a new and surprising form of compositionality, one that exists not in the input space but in the vast parameter space of the model's weights.

Foundational to many contemporary machine learning systems, large pre-trained models require further editing to enhance performance on downstream tasks~\cite{zhuang2020comprehensive,ilharco2022patching,ilharco2023task}, align them with human values~\cite{ouyang2022training,lu2022quark,ribeiro2022adaptive,sparrow}, and increase robustness~\cite{wortsman2021robust,shibani2021editing,ortizjimenez2021optimism}. Conventional editing approaches often depend on resource-intensive joint fine-tuning across multiple tasks~\cite{zhuang2020comprehensive} or human-feedback mechanisms~\cite{ouyang2022training}, limiting their scalability. Moreover, specializing a model on new tasks can often degrade its performance on previously learned and \textit{zero-shot} capabilities~\cite{mccloskey1989catastrophic,french1999catastrophic,wortsman2021robust}.

More recent research has pioneered cost-effective and scalable model editing strategies that aim to preserve the core capabilities of the pre-trained model. These methods act directly on the model weights through \textit{task arithmetic} or weight interpolation techniques~\cite{ilharco2023task,ilharco2022patching,wortsman2021robust,wortsman2022model,ainsworth2022git,frankle2020linear,don2022cold,matena2021merging,li2022branch,singh2020fusion,izmailov2018averaging} and circumvent the need for expensive joint fine-tuning. These approaches are based on the key observation that arithmetic operations performed on fine-tuned weights often correspond to analogous operations on the model's function~\cite{ilharco2023task}. For example, by summing the relative weight vectors of a model between pre-training and fine-tuning on two separate tasks, a new multitask model can be created that exhibits enhanced performance on both tasks. Conversely, subtracting a task's vector can effectively make the model `forget' that specific skill.

Despite these breakthroughs, a deep understanding of the underlying principles of task composition and its general effectiveness remains lacking.

To address these open questions, this chapter presents a systematic study of task arithmetic in the context of contrastively pre-trained vision-language models like CLIP \cite{radford2021learning}. We examine the hypothesis, first proposed by \citet{wortsman2021robust}, that task arithmetic is possible because these large models inherently function in a linear regime, where their behavior is governed by the finite-width neural tangent kernel (NTK)~\cite{jacot2018neural,chizat2019lazy}. 

\begin{figure}[t]
    \centering
    \includegraphics[width=\textwidth]{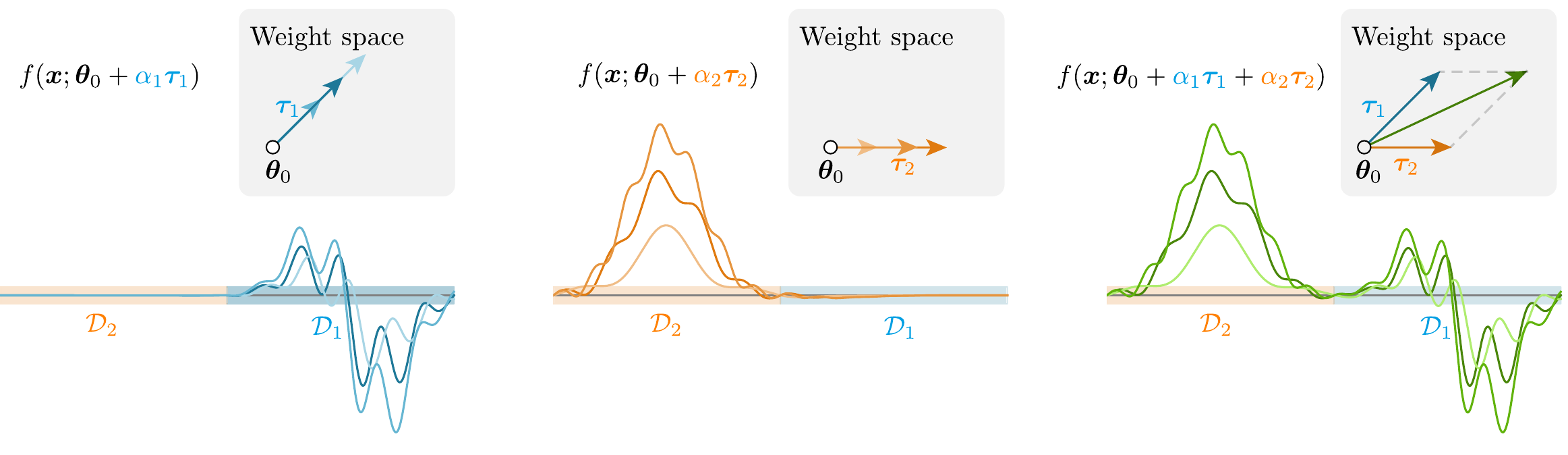}
    \caption{\textbf{Illustration of weight disentanglement}, where distinct directions in the weight space, $\vt_t$, are associated with spatially localized areas of the function space, $\D_t$. This enables a model, $f$, to manipulate the disjoint parts of the input space independently by adding linear combinations of those weight directions to a pre-trained checkpoint $\vth_0$.\looseness=-1}
    \label{fig:sketch}
\end{figure}

Our research indicates that while linearized CLIP models showcase significantly better task arithmetic performance than their nonlinear counterparts (see~\autoref{tab:task_addition}-\ref{tab:task_negation}), the NTK framework cannot fully account for the task arithmetic abilities of their standard nonlinear forms. We find that the sole condition for task arithmetic is actually \textit{weight disentanglement} -- a property where distinct directions in the weight space correspond to localized changes of the network function in disjoint spatial regions\footnote{Throughout the paper, we use the term \textit{spatial} to refer to the input space.} (see~\autoref{fig:sketch}). This structure allows a model to compose tasks by independently manipulating these specific weight directions.

We find that fine-tuning models in their tangent space by linearizing them amplifies weight disentanglement, yielding significant performance improvements across multiple benchmarks. However, while weight disentanglement is stronger in the tangent space, it is also present in nonlinear models. We show that weight disentanglement of semantically meaningful tasks is an emergent property of pre-training, as it is not present at random initialization.

The main contributions of this part of the thesis are as follows.

We provide a formal notion of task arithmetic, which allows for its quantitative analysis.

We demonstrate that the NTK is insufficient to explain task arithmetic in nonlinear models and introduce weight disentanglement as the necessary underlying condition.

We put forward linearization as a method to augment weight disentanglement and thereby improve task arithmetic. 

Using kernel theory, we connect weight disentanglement in linearized models to the spatial localization of the kernel's eigenfunctions and validate this theory numerically in pre-trained transformers.

Finally, we show that weight disentanglement is an emergent property that arises from the pre-training process.

\section{Notation and problem statement}\label{sec:notation}

Let $f:\mathcal{X}\times\Theta\to\mathcal{Y}$ be a neural network taking inputs $\vx\in\mathcal{X}$ and parameterized by a set of weights $\vth\in\Theta$. We will assume $\mathcal{X} \subseteq \mathbb{R}^d$, $\Theta \subseteq\mathbb{R}^m$ and $\mathcal{Y}\subseteq\mathbb{R}^c$. Consider $T$ tasks, with every task $t$ consisting of a triplet $(\mathcal{D}_t, \mu_t, f^*_t)$ where $\mathcal{D}_t\subseteq\mathcal{X}$ is a data support (e.g.,  ImageNet~\cite{deng2009imagenet} images), $\mu_t$ an input distribution such that $\operatorname{supp}(\mu_t)=\mathcal{D}_t$, and $f^*_t:\mathcal{D}_t\to\mathcal{Y}$ a target function (e.g.,  labels). In practice, each task is identified with a training set $\{(\vx_\nu,f^*_t(\vx_\nu))\}_{\nu\in[n_t]}$ with $\vx\sim\mu_t$, that is used to fine-tune the models starting from the pre-trained weights $\vth_0$ and obtain the fine-tuned weights $\vth^*_t$.

\paragraph{Task arithmetic} Let the \textit{task vector} of task $t$ be the difference between the fine-tuned and the pre-trained weights, i.e., $\vt_t = \vth^*_t - \vth_0$. The following property formalizes the notion of task arithmetic introduced in \citet{ilharco2023task}, where the authors observed that the accuracies of pre-trained models on different datasets can be modified independently through the addition or removal of task vectors.

\begin{property}[Task arithmetic]\label{prop:task_arithmetic}
    Consider a set of task vectors $\mathcal{T} = \{\vt_t\}_{t \in [T]}$ with associated non-intersecting task supports $\mathcal{D}=\{\mathcal{D}_t \subset \mathcal{X}\}_{t \in [T]}$, i.e., $\forall t,t'$, if $t\neq t'$ then $\D_t\cap\D_{t'}=\varnothing$. We say a network $f$ satisfies the task arithmetic property around $\vth_0$ with respect to $\mathcal{T}$ and $\D$ if
\begin{equation}
    f\left(\vx;\vth_0+\sum_{t=1}^T \alpha_t\,\vt_t\right)=\begin{cases}
    f(\vx;\vth_0+\alpha_t\,\vt_t) & \vx\in\D_t \\
    f(\vx;\vth_0) & \vx\notin\bigcup_{t=1}^T\D_t    \end{cases}
    \label{eq:task_arithmetic}
\end{equation}
with $(\alpha_1,\dots,\alpha_T) \in \mathcal{A}\subseteq \mathbb{R}^T$.
\end{property}

In short, a model satisfies \autoref{prop:task_arithmetic} if adding $\vt_t$ does not modify the output of the model outside $\D_t$.

\paragraph{Neural tangent kernel} Around the initialization weights $\vth_0$, a neural network can be approximated with a first-order Taylor expansion:
\begin{equation}
f(\vx;\vth)\approx f(\vx;\vth_0)+(\vth-\vth_0)^\top\nabla_{\vth}f(\vx;\vth_0).\label{eq:linearization}
\end{equation}
This approximation is equivalent to a kernel predictor with a kernel known as the \textit{neural tangent kernel} (NTK) \cite{jacot2018neural}, $k_{{\mathrm{NTK}}}(\vx,\,\vx') = \nabla_{\vth} f(\vx;\vth_0)^\top \nabla_{\vth} f(\vx';\vth_0)$, and defines a neural tangent space in which the relationship between weights and functions is linear. Remarkably, as the network width approaches infinity, \autoref{eq:linearization} becomes exact and remains valid throughout training~\cite{jacot2018neural,arora2019exact, lee2019wide}.

However, this linear approximation is often invalid at finite widths, as the evolution of parameters during training is inadequately captured by \autoref{eq:linearization}. In such cases, training occurs in a \textit{nonlinear regime}. Conversely, often during fine-tuning, parameter evolution in many pre-trained models is frequently minimal, meaning that training does not exit the tangent space and \autoref{eq:linearization} closely approximates the network behavior~\cite{malladi2022kernel,ortizjimenez2021linear,zancato2020predicting,deshpande2021linearized,yuce2022inrs}. In such cases, training occurs in a \textit{linear regime}.

\section{Task arithmetic is not a consequence of linear fine-tuning} \label{sec:not_ntk}

The objective of this work is to understand the conditions that enable task arithmetic in deep neural networks. Previous studies hypothesized that task arithmetic results from fine-tuning in the linear regime~\cite{wortsman2021robust,ilharco2023task,wortsman2022model}, as linear weight combinations correspond to similar output function combinations. However, we will now demonstrate that CLIP models do not fine-tune in the linear regime and we therefore need other ways to explain task arithmetic.\looseness=-1

In general, if a pre-trained network $f(\cdot\,; \, \vth_0)$ demonstrates \emph{kernel behavior} during fine-tuning -- i.e., fine-tuning occurs in the linear regime -- the following property must be satisfied \cite{malladi2022kernel}:\looseness=-1
\begin{property}[Post-hoc linearization]\label{prop:posthoc}
The change in the network output after training can be approximated by its first-order Taylor expansion, i.e., $f(\vx;\vth^*)-f(\vx;\vth_0)\approx(\vth^*-\vth_0)^\top\nabla_{\vth}f(\vx;\vth_0)$.
\end{property}
In simple terms, the approximation of the network in the tangent space around initialization must hold after fine-tuning. To test this, we evaluate the performance of the \textit{post-hoc} linearized version of $f$, $f_{{\mathrm{lin}}}$. That is, we apply the fine-tuned task vectors $\vt=\vth^*-\vth_0$ to the linear approximation of $f$ at $\vth_0$, i.e.,
\begin{equation}\label{eq:f_lin}
    f_\text{lin}(\vx;\vth_0+\vt)=f(\vx;\vth_0)+\vt^\top\nabla_{\vth}f(\vx;\vth_0),
\end{equation}
and we check whether $f_\text{lin}(\cdot\,;\,\vth^*)$ performs similarly to $f(\cdot\,;\,\vth^*)$.

\begin{table}[t]
\caption{\textbf{Task addition.} Average absolute (\%) and normalized accuracies (\%) of different CLIP ViTs edited by adding the sum of the task vectors of $8$ tasks. We report results for the nonlinear and linearized models of \autoref{sec:not_ntk} and \ref{sec:enhancing} normalizing performance by their single-task accuracies.\looseness=-1}\label{tab:task_addition}
\vspace{0.5em}
\centering
\begin{tabular}{@{}lcccc@{}}
\toprule
\multirow{2}{*}{Method} & \multicolumn{2}{c}{ViT-B/32} & \multicolumn{2}{c}{ViT-L/14}  \\
                 & Abs. {\small ($\uparrow$)} & Norm. {\small ($\uparrow$)} & Abs. {\small ($\uparrow$)} & Norm. {\small ($\uparrow$)} \\ \midrule
Pre-trained {\small \hspace{2.1em}$f(\cdot\,;\,\vth_0)$}      & 48.4     & --       & 64.4     & --       \\ \midrule
Non-lin. FT {\small \hspace{.7em} $f(\cdot\,;\,\vth_0+\vt)$}   & 71.4     & 76.5     & 85.1     & 88.8       \\
Post-hoc lin. {\small $f_{{\mathrm{lin}}}(\cdot\,;\,\vth_0+\vt)$}     & 57.1     & 81.9     & 75.2     & 90.0       \\
\midrule
Linear. FT {\small \hspace{.5em}$f_{{\mathrm{lin}}}(\cdot\,;\,\vth_0+\vt_{\text{lin}})$} & \textbf{76.5} & \textbf{85.4} & \textbf{88.5} & \textbf{93.5} \\
\bottomrule
\end{tabular}
\end{table}

\begin{table}[t]
\caption{\textbf{Task negation.} Minimum accuracy ($\%$) of different CLIP ViTs edited by negating a task vector from a target task while retaining $95\%$ of their performance on the control task. We report average performances over eight tasks on nonlinear and linearized models as introduced in \autoref{sec:not_ntk} and \ref{sec:enhancing}.\looseness=-1}\label{tab:task_negation}
\vspace{0.5em}
\centering
\begin{tabular}{@{}lcccc@{}}
\toprule
\multirow{2}{*}{Method} & \multicolumn{2}{c}{ViT-B/32} & \multicolumn{2}{c}{ViT-L/14}  \\
                 & Targ. {\small ($\downarrow$)} & Cont. {\small ($\uparrow$)} & Targ. {\small ($\downarrow$)} & Cont. {\small ($\uparrow$)} \\ \midrule
Pre-trained {\small \hspace{2.1em}$f(\cdot\,;\,\vth_0)$}       & 48.4          & 63.4          & 64.4          & 75.5       \\ \midrule
Non-lin. FT {\small \hspace{.7em} $f(\cdot\,;\,\vth_0-\vt)$}  & 24.0          & 60.7          & 18.0          & \textbf{72.5}      \\
Post-hoc lin. {\small $f_{{\mathrm{lin}}}(\cdot\,;\,\vth_0-\vt)$}     & 14.8          & 60.3          & 12.1          & 71.8       \\
\midrule
Linear. FT {\small \hspace{.5em}$f_{{\mathrm{lin}}}(\cdot\,;\,\vth_0-\vt_{\text{lin}})$} & \textbf{10.9} & \textbf{60.8} & \textbf{7.9} & \textbf{72.5} \\
\bottomrule
\end{tabular}

\end{table}

\begin{figure}
    \centering
    \includegraphics[width=0.48\textwidth]{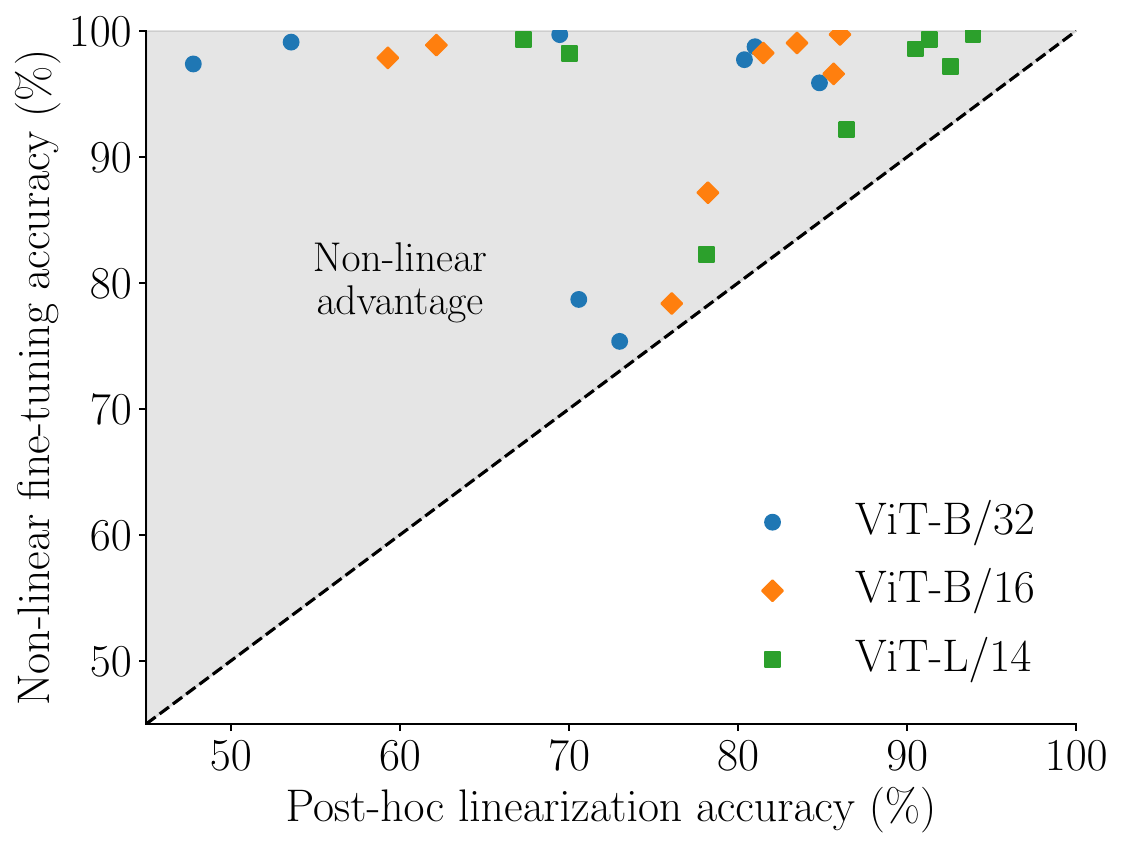}
    \caption{\textbf{nonlinear advantage.} Single-task accuracies of nonlinearly fine-tuned models $f(\cdot\,;\,\vth^*)$ and their \textit{post-hoc} linearization $f_{{\mathrm{lin}}}(\cdot\,;\,\vth^*)$. Markers represent different ViTs.\looseness=-1} 
    \label{fig:scatter_comp_advantages}
\end{figure}
The results in \autoref{fig:scatter_comp_advantages} indicate that CLIP models do not exhibit a kernel behavior. Specifically, we fine-tune (FT) several CLIP pre-trained Vision Transformers (ViTs)~\cite{dosovitskiy2021an} of different sizes following the same setup as \citet{ilharco2023task} on $8$ tasks: Cars~\cite{cars}, DTD~\cite{dtd}, SUN397~\cite{sun397}, EuroSAT~\cite{eurosat}, GTSRB~\cite{gtsrb}, MNIST~\cite{lecun1998mnist}, SVHN~\cite{svhn} and RESISC45~\cite{cheng2017remote}. We observe that the single-task performance of $f_\text{lin}(\cdot\,;\,\vth^*)$ is significantly lower than that of $f(\cdot\,;\,\vth^*)$ for ViTs of all sizes. This \textit{nonlinear advantage}~\cite{fort2020deep} is a clear sign that fine-tuning has not happened in a linear regime as expected by~\citet{wortsman2021robust}.\looseness=-1

Yet, this observation is not enough to rule out that task arithmetic can be explained by linearizing the network function. Indeed, even if the nonlinear components are important for single-task performance, they might not be used during task arithmetic, which is the objective of this study. That is, the projection of $f$ onto the tangent space could be the only useful component.\looseness=-1 

We now show this is also not the case, as doing task arithmetic with the nonlinearly fine-tuned task vectors over $f_\text{lin}$ significantly decreases performance. To show this, we employ the benchmark proposed in \citet{ilharco2023task} to evaluate the task arithmetic ability of a pre-trained model, which consists of the $8$ tasks described before and two sub-benchmarks:
\begin{enumerate}
\item \textbf{Task addition}: The sum of the task vectors $\vt=\sum_t\vt_t$ is added to a pre-trained checkpoint to produce a multi-task model. The success of this benchmark is measured in terms of the maximum average accuracy over the different tasks. Results are shown in \autoref{tab:task_addition}.\looseness=-1
\item \textbf{Task negation}: A task vector is subtracted from the pre-trained checkpoint to forget a task while retaining performance on a control task (ImageNet). The success of this benchmark is measured in terms of the maximum drop in accuracy on the forgetting task that retains the performance on the control task. Results are averaged over tasks and shown in \autoref{tab:task_negation}.\looseness=-1
\end{enumerate}
To obtain the task vectors, we use the fine-tuned weights of the different ViTs from before, and use a single mixing coefficient $\alpha{=}\alpha_1{=}\dots{=}\alpha_T$ optimized separately for the nonlinear and post-hoc linearized models to ensure a fair comparison. We provide all details of this experiment in~\autoref{ap:experiment_details}.\looseness=-1

The results in \autoref{tab:task_addition} confirm that task arithmetic in CLIP models does not stem from the combination of their linear components only. Specifically, we observe a significant drop in absolute task addition accuracy in the \textit{post-hoc} linearized models compared to the nonlinear ones. This decrease in performance is consistent across tasks (see \autoref{ap:addition}) and highlights that task arithmetic in nonlinear models leverages the nonlinear components of $f$, as well.\looseness=-1

Although these results reject the linear hypothesis, it is still remarkable that the post-hoc linearized models do better at task negation than the nonlinear ones (see \autoref{tab:task_negation}). Furthermore, even in task addition (see \autoref{tab:task_addition}) they achieve higher normalized accuracies (see definition in \autoref{ap:experiment_details}). Indeed, as we formalize in \autoref{sec:weight_disentanglement}, this observation suggests that linearized models are more consistent with \autoref{prop:task_arithmetic}. In \autoref{sec:enhancing}, we will use this fact to devise a new way to enhance task arithmetic.\looseness=-1

\section{Weight disentanglement}\label{sec:weight_disentanglement}

If the linear regime is not necessary to explain task arithmetic, what are the necessary conditions that allow it? In this section, we argue that the only necessary condition to perform task arithmetic with a model $f$ is that the model is \textit{weight disentangled} with respect to the set of fine-tuning tasks.

\begin{property}[Weight disentanglement]\label{prop:weight_disentanglement}
A parametric function $f:\mathcal{X}\times\Theta\to\mathcal{Y}$ is weight disentangled with respect to a set of task vectors $\mathcal{T}=\{\vt_t\}_{t\in[T]}$ and the corresponding supports $\mathcal{D}=\{\mathcal{D}_t\}_{t\in[T]}$ if
\begin{equation}
f\left(\vx;\vth_0+\sum_{t=1}^T \alpha_t\vt_t\right) = \sum_{t=1}^T g_t(\vx;\alpha_t\vt_t) + g_0(\vx),\label{eq:local_additivity}
\end{equation}
where $g_t(\vx;\alpha_t\vt_t)=\bm 0$ for $\vx\notin\mathcal{D}_t$ and $t=1,\dots,T$, and $g_0(\vx)=0$ for $\vx\in \bigcup_{t\in[T]}\mathcal{D}_t$.
\end{property}

In essence, this definition captures the idea that the function $f$ can be decomposed as a sum of spatially-localized components, i.e., vanishing outside a spatial region, whose functional variation is entirely captured by each $\vt_t$ (see~\autoref{fig:sketch}). Moreover, it is trivial to see that satisfying weight disentanglement is equivalent to satisfying \autoref{prop:task_arithmetic} on task arithmetic as one can always write \autoref{eq:task_arithmetic} as\looseness=-1
\begin{align}
f\left(\vx;\vth_0+\sum_{t=1}^T \alpha_t\vt_t\right) &= \sum_{t=1}^T f(\vx;\vth_0+\alpha_t\vt_t)\mathbb{1}(\vx\in\mathcal{D}_t) \nonumber \\
&\phantom{=} + f(\vx;\vth_0)\mathbb{1}\left(\vx\notin\bigcup_{t\in[T]}\mathcal{D}_t\right),
\end{align}
and identify $g_t(\vx;\alpha_t\vt_t)=f(\vx;\vth_0+\alpha_t\vt_t)\mathbb{1}(\vx\in\mathcal{D}_t)$ and $g_0(\vx)=f(\vx;\vth_0)\mathbb{1}(\vx\notin\mathcal{D}_t)$. It is important to highlight, however, that this additive decomposition does not imply linearity, as the local functions $\{g_t\}_{t\in[T]}$ are not required to be linear with respect to the parameters. 

Furthermore, note that weight disentanglement is a property of the predictors and not related to the performance on different tasks. That is, a model could be weight disentangled with respect to a set of task vectors and still perform poorly on a task, e.g.,  if $f(\cdot\,;\,\vth_0+\alpha\vt)$ does not generalize for some $\alpha$. More generally, we can visualize the level of weight disentanglement of a model by measuring its discrepancy with \autoref{eq:local_additivity}. To do so, given two tasks, one can check the \textit{disentanglement error} of a model,\looseness=-1
\begin{equation}
\xi(\alpha_1, \alpha_2)=\sum_{t=1}^2\mathbb{E}_{\vx \sim\mu_t} \left[\operatorname{dist}\left(f(\vx;\vth_0+\alpha_t\vt_t), f(\vx;\vth_0+\alpha_1\vt_1+\alpha_2\vt_2)\right)\right],
\end{equation}
where $\operatorname{dist}$ denotes any distance metric between output vectors. As we are dealing with classification tasks, in what follows we use the prediction error $\operatorname{dist}(y_1,y_2)=\mathbb{1}(y_1\neq y_2)$ as the distance metric. In general, the smaller the value of $\xi(\alpha_1, \alpha_2)$ the more weight disentangled a model is at $(\alpha_1, \alpha_2)$.

\begin{figure}[t]
    \centering
    \includegraphics[width=0.95\textwidth]{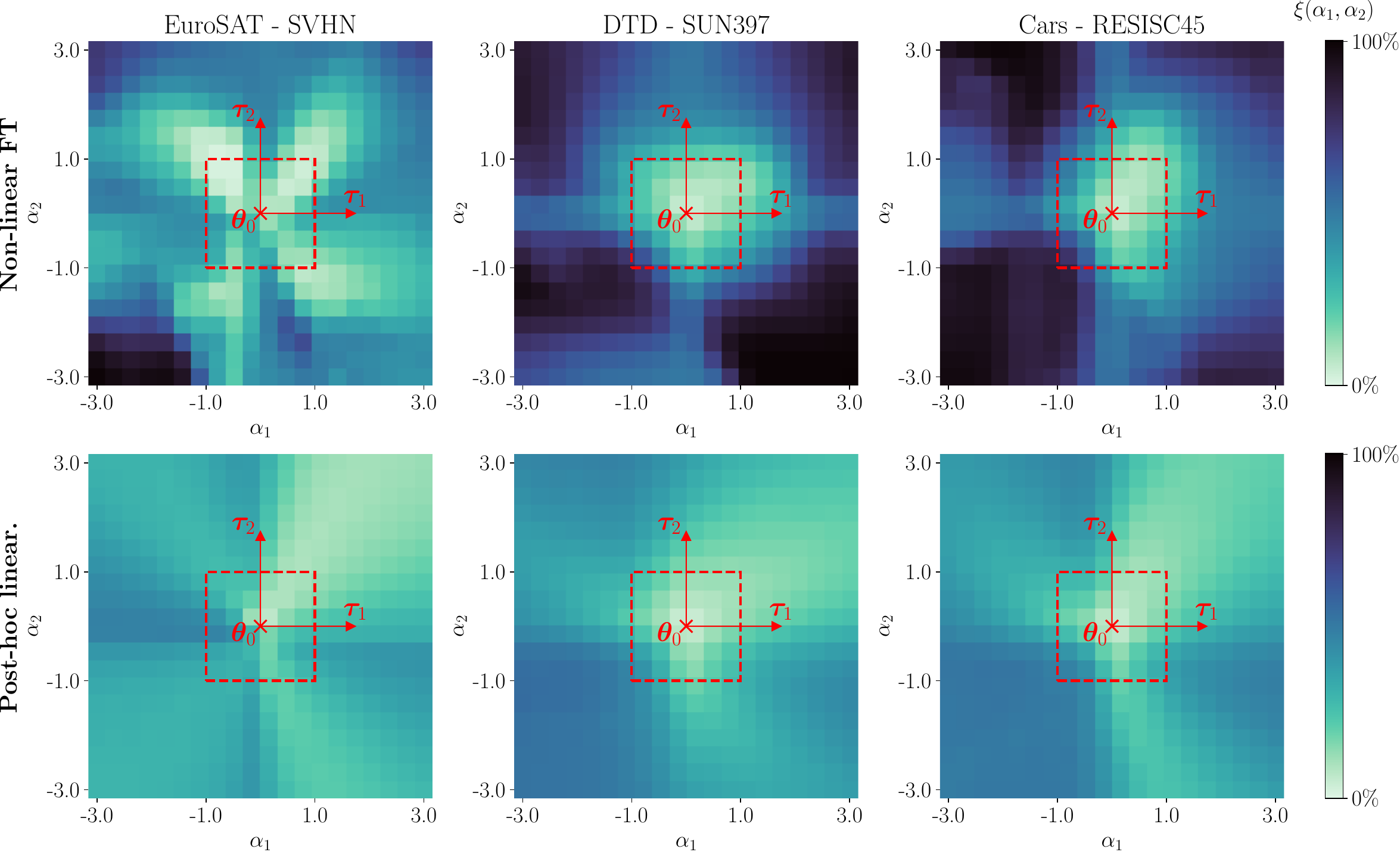}
    \caption{\textbf{Visualization of weight disentanglement.} The heatmaps show the disentanglement error $\xi(\alpha_1, \alpha_2)$ of a nonlinear CLIP ViT-B/32 (top) and its post-hoc linearization (bottom) on different example task pairs. The light regions denote areas of the weight space where weight disentanglement is stronger. The red box delimits the search space used to compute the best $\alpha$ in all our experiments.\looseness=-1}
    \label{fig:heatmaps}
\end{figure}

\autoref{fig:heatmaps} displays the disentanglement error of a CLIP ViT-B/32 model concerning several task vector pairs. We observe that the CLIP model exhibits a minimal disentanglement error within a small region surrounding $\vth_0$, which enables task arithmetic. However, for $\alpha_1,\alpha_2>1$, the error increases, indicating a high degree of interaction between tasks. This explains why task arithmetic performs better in a small neighborhood of $\vth_0$ -- task arithmetic is more effective when fine-tuning with small learning rates and few training steps~\cite{ilharco2023task} -- with the optimal value of $\alpha$ typically being less than $1$.\looseness=-1

Comparing the disentanglement error of the nonlinear models and their post-hoc linearization reveals an interesting finding: linearized models exhibit greater disentanglement than their nonlinear counterparts. This is evident from the more extensive regions with low disentanglement errors in \autoref{fig:heatmaps} (bottom). This explains why the post-hoc linearized models achieve higher normalized accuracies via task addition (cf. \autoref{tab:task_addition}) and manage to forget more through task negation (cf. \autoref{tab:task_negation}). Paradoxically, however, although the greater disentanglement of linearized models allows them to retain more of their relative performance when edited with task arithmetic, they still perform worse in absolute terms due to the great advantage of the nonlinear models in single-task accuracy (cf. \autoref{fig:scatter_comp_advantages}). This suggests that closing the single-task performance gap between linearized and nonlinear models could be a way to enhance task arithmetic. We leverage this idea in the next section.\looseness=-1

\section{Enhancing task arithmetic via linearization}\label{sec:enhancing}

\begin{figure}
    \centering
    \includegraphics[width=0.5\textwidth]{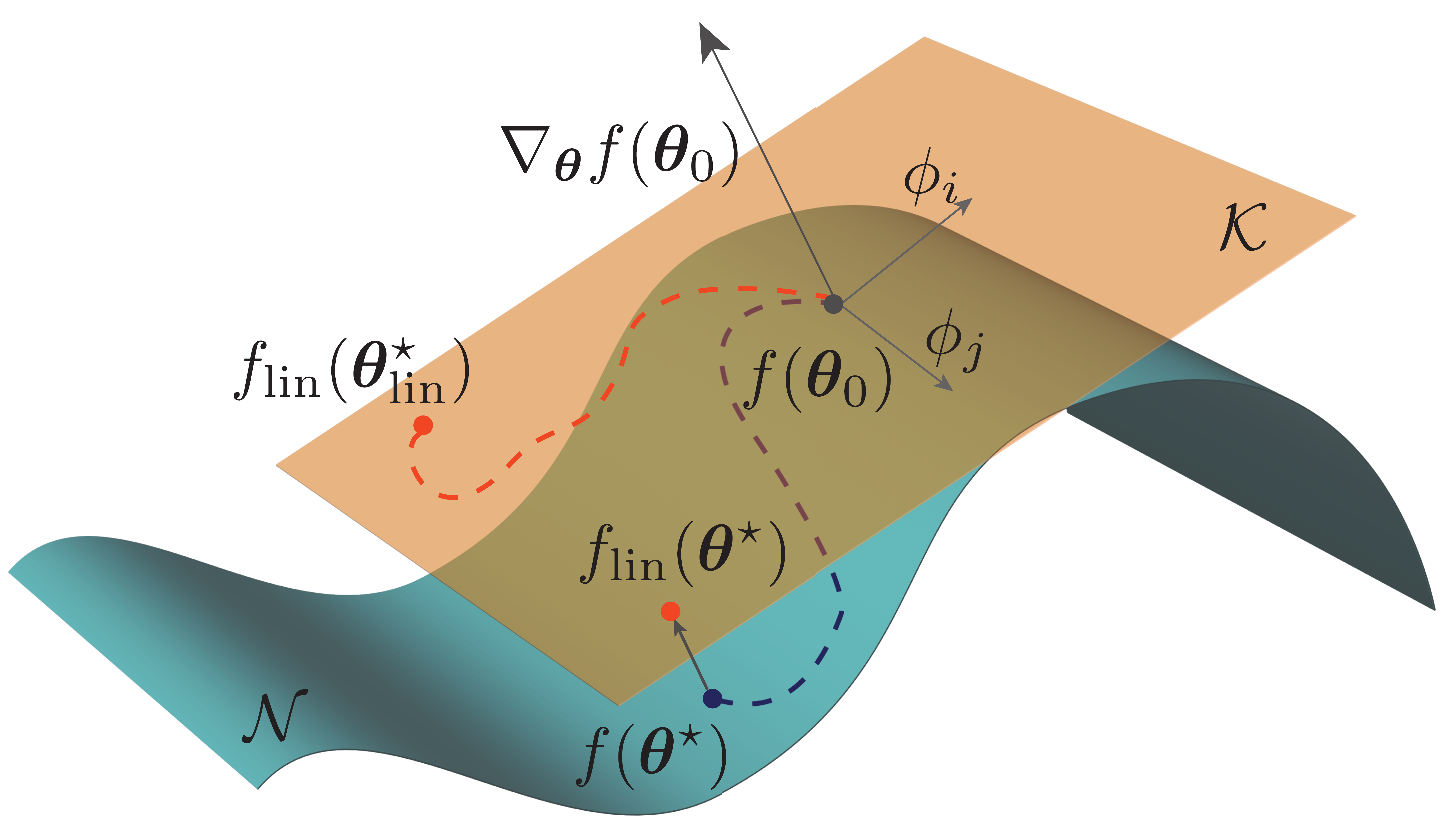}
    \caption{Conceptual illustration of the different approaches we use to edit a pretrained model $f(\cdot\,;\,\vth_0)$. Here $\mathcal{N}$ represents the space of neural network functions $f$, nonlinearly parameterized by $\vth\in\bm\Theta$; and $\mathcal{K}$ its tangent space, given by the space of linearized functions $f_{\text{lin}}$.}
    \label{fig:3d}
\end{figure}
We have seen that linearized models are more weight disentangled than nonlinear ones. However, post-hoc linearization degrades single-task performance. We now demonstrate that enforcing models to fine-tune in the tangent space to their pre-trained initialization significantly improves task arithmetic by reducing the single-task accuracy gap.

\begin{figure}
    \centering
    \includegraphics[width=0.48\textwidth]{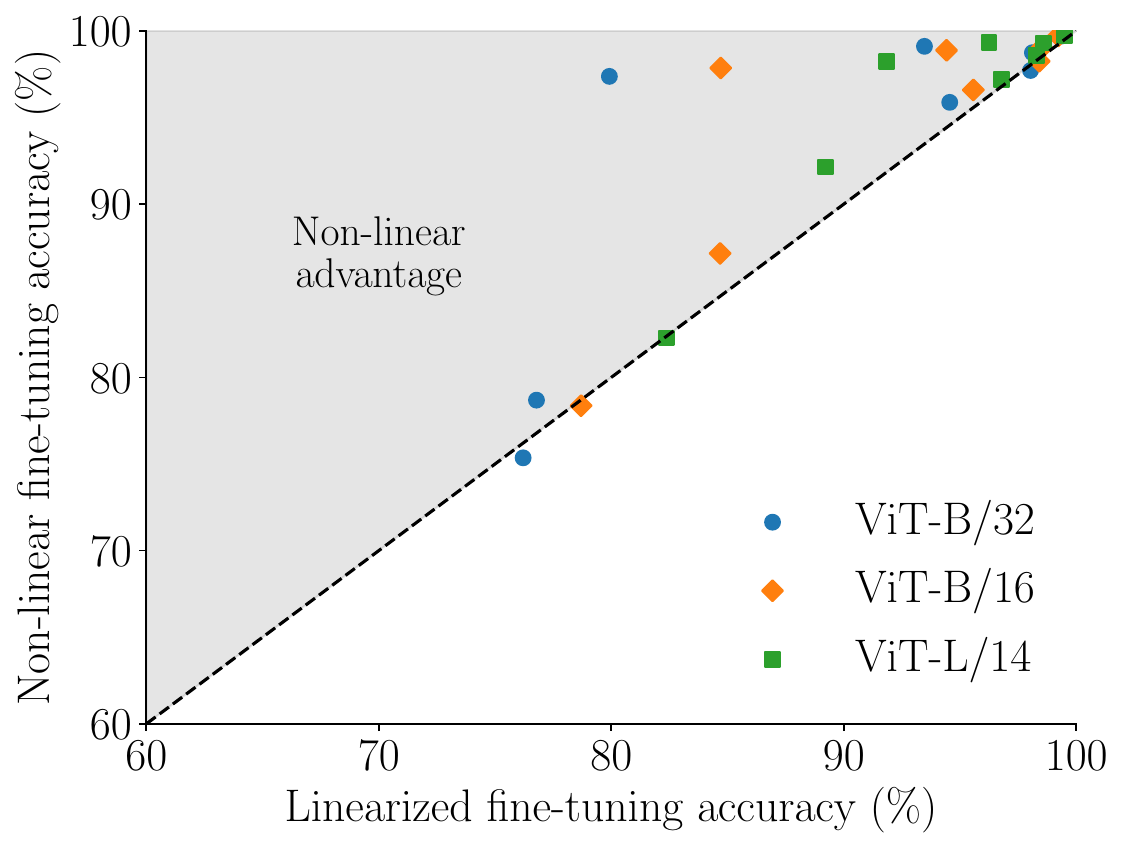}
    \caption{Single-task accuracies of nonlinearly FT, $f(\cdot\,;\,\vth^*)$ and linearly FT, $f_{{\mathrm{lin}}}(\cdot\,;\,\vth_{{\mathrm{lin}}}^*)$, models.} 
    \label{fig:scatter_advantages}
\end{figure}

Specifically, rather than applying the nonlinearly fine-tuned task vectors $\vt=\vth^*-\vth_0$ to $f_{{\mathrm{lin}}}$, as in \autoref{sec:not_ntk}, we propose to directly obtain the task vectors through explicit fine-tuning in the tangent space as illustrated in \autoref{fig:3d}. That is, given a model $f$, we directly fine-tune its linear approximation $f_{{\mathrm{lin}}}$ around $\vth_0$~\cite{fort2020deep}. The fine-tuning process can follow the same protocols used before but with the network parameterization dictated by \autoref{eq:f_lin}. Due to the linear connection between the weight-space and function-space defined in~\autoref{eq:f_lin}, fine-tuning $f_{{\mathrm{lin}}}$ is essentially the same as training a kernel predictor with kernel $k_{{\mathrm{NTK}}}$. As a result, we obtain the fine-tuned weights $\vth_{{\mathrm{lin}}}^*$ of the linearized model for each task, which allows us to construct the corresponding task vector $\vt_{{\mathrm{lin}}}=\vth_{{\mathrm{lin}}}^*-\vth_0$.\looseness=-1

Moreover, as the considered models do not inherently exhibit linear fine-tuning (see~\autoref{sec:not_ntk}), this approach yields significantly different results compared to post-hoc linearization, i.e., $f_{{\mathrm{lin}}}(\vx;\vth_0+\vt_{{\mathrm{lin}}}) \allowbreak \neq \allowbreak f_{{\mathrm{lin}}}(\vx;\vth_0+\vt)$. In particular, although both models share the same kernel $k_{{\mathrm{NTK}}}(\vx,\,\vx')$, the task vectors $\vt_{{\mathrm{lin}}}$ have been explicitly optimized to maximize the performance of such linearized models. Consequently, by construction, linearized fine-tuning outperforms post-hoc linearization. Indeed, in \autoref{fig:scatter_advantages}, we observe that linearized fine-tuning significantly reduces the nonlinear advantage of nonlinear models, as in most cases the performance of $f_{{\mathrm{lin}}}(\cdot\,;\,\vth_0+\vt_{{\mathrm{lin}}})$ is very similar to the one of $f(\cdot\,;\,\vth_0+\vt)$ (cf. \autoref{fig:scatter_comp_advantages}).\looseness=-1

Remarkably, as we show in \autoref{ap:disentanglement_linear}, this increase in single-task performance does not compromise weight disentanglement, which remains as high as for the post-hoc linearized models in \autoref{fig:heatmaps}. As a result, linear fine-tuning allows for improved task arithmetic compared to standard nonlinear fine-tuning. In particular, \autoref{tab:task_addition}-\ref{tab:task_negation} in their last rows show that linearized fine-tuned models significantly outperform their nonlinear counterparts and achieve state-of-the-art results on the task addition and negation benchmarks~\cite{ilharco2023task}. The linearized fine-tuned models achieve higher multi-task accuracies through task addition (up to $5.8$ points more) and can forget more through task negation (up to $13.1$ points more) while maintaining a similar level of accuracy on the control task. Additionally, we observe that the advantage of the linearized models over the nonlinear ones is higher for the smaller ViT-B/32 and progressively diminishes as the model size increases up to ViT-L/14.\looseness=-1

In general, thanks to the efficiency of the Jacobian-vector product implementations in most deep learning frameworks~\cite{novak2022fast}, training and inference in linearized neural networks only require an $\mathcal{O}(1)$ increase in computational costs with respect to their nonlinear counterparts. In this regard, the superiority of task arithmetic of linearized models can make this technique appealing for practical applications. Identifying the right trade-offs between computational cost and performance, as well as faster linearization techniques, is an exciting avenue for future work.\looseness=-1

\section{Towards understanding task arithmetic}\label{sec:understanding}

We conclude by providing further fundamental insights that can aid our understanding of task arithmetic. In particular, we ask whether any kernel can satisfy \autoref{prop:task_arithmetic}, and we establish a connection between task arithmetic and the spectral properties of the NTK. Then, we argue that weight disentanglement and task arithmetic are emergent properties of pre-training.\looseness=-1

\subsection{Eigenfunction localization}\label{sec:spectrum_ntk}

Generally, a kernel~$k$ admits a decomposition in terms of a family of eigenfunction-eigenvalue pairs $\{(\phi_\rho, \lambda_\rho)\}_{\rho\in\mathbb{N}}$; which implies that $k$ can only represent functions of the form $f^*(\vx) = \sum_{\rho=1}^{\infty}c_\rho \phi_\rho(\vx)$ with a finite kernel norm, i.e., $\|f^*\|^2_\mathcal{H} = \sum_{\rho=1}^\infty {c_\rho^2}/{\lambda_\rho}<+\infty$. Specifically, the coefficients $\{c_\rho\}_{\rho\in\mathbb{N}}$ constitute a representation of the function $f^*$ in the kernel basis.

Consider $T$ tasks $\{f^{*}_t\}_{t\in[T]}$ supported in their respective non-intersect\-ing domains $\{\mathcal{D}_t\}_{t\in[T]}$. Furthermore, let $\{\phi_\rho\}_{\rho\in\mathbb{N}}$ be an orthogonal basis of eigenfunctions that diagonalizes the kernel on the union of all $\mathcal{D}_t$'s. The following proposition provides a sufficient condition on the representation of the tasks in this basis to ensure the task arithmetic property:

\begin{proposition}[Simplified]
\label{propo:localization}
Suppose that $\{f^*_t\}_{t\in[T]}$ can be represented by the kernel $k$. The kernel $k$ is capable of performing task arithmetic with respect to $\{f^*_t\}_{t\in[T]}$ and $\{\mathcal{D}_t\}_{t\in[T]}$ if, for each task $t$, there exists a subset of localized eigenfunctions such that i) $\operatorname{supp}(\phi) \subseteq \mathcal{D}_t$ for each $\phi$ in the subset, and ii) the representation of $f^*_t$ only involves these basis functions.
\end{proposition}

The proof and formal statement are deferred to \autoref{ap:spectral}. Intuitively, if each task is represented with eigenfunctions that vanish outside the spatial region identified by the task support, the functions corresponding to different tasks do not interfere. Based on Property \autoref{propo:localization}, it is natural to examine whether the NTK of CLIP models displays eigenfunctions localized in each task domain and if it represents the different tasks using these functions. According to the \textit{representer theorem} of kernels~\cite{scholkopf2002kernels}, after linear fine-tuning on task $t$ with a training set $\{(\vx_\nu,f^*_t(\vx_\nu))\}_{\nu\in[n_t]}$ and $\vx_\nu\sim\mu_t$, the CLIP's predictor evaluated at a new point ${\vx \in \mathcal{X}}$ can be expressed as a linear combination of its kernel $k_{{\mathrm{NTK}}}$ evaluated on $\vx$ and the training data, i.e., $f_{\text{lin}}(\vx) = f(\vx;\vth_0)+\sum_{\nu \in [n_t]} \beta_\nu \, k_{{\mathrm{NTK}}}(\vx_\nu,\,\vx)$.\looseness=-1

To explore whether CLIP models use localized eigenfunctions for task arithmetic, we diagonalize the matrix $(K_{{\mathrm{NTK}}})_{ij}=k_{{\mathrm{NTK}}}(\vx_i,\,\vx_j)$ with $\vx_i \in \mathcal{D}_t$, i.e., the task on which we trained, and $\vx_j\in\D_t\cup\D_{t'}$, where $\D_{t'}$ is the support of a control task. If the eigenfunctions used to represent $f^*(\vx)$ are localized, then the power of the eigenvectors of $K_{{\mathrm{NTK}}}$ must be concentrated in the points belonging to the dataset used for training. To measure this concentration, we introduce the local energy $\mathcal{E}_{{\mathrm{loc}}}(\vx) = \sum_{\rho} \phi_\rho^2(\vx)$, which sums the power of all the eigenfunctions $\phi_\rho$ at a given point $\vx$.

\begin{figure}
    \centering
    \includegraphics[width=0.5\textwidth]{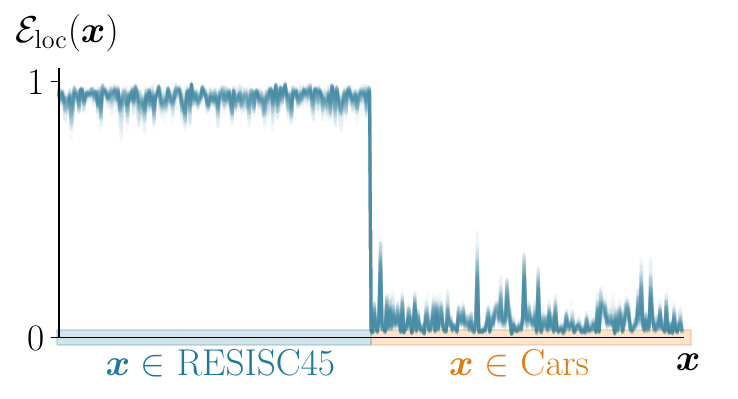}
    \caption{\textbf{Eigenfunction localization.} Estimated support of the eigenfunctions of the NTK of a ViT-B/32 CLIP model trained on RESISC45. The plot shows the sum of the local energy of the eigenfunctions over a random subset of the training and control supports (RESISC45 and Cars, respectively).}
    \label{fig:localization}
\end{figure}

In~\autoref{fig:localization}, we plot this metric for a ViT-B/32 CLIP model trained on RESISC45 with Cars as control. We provide results for other task pairs in \autoref{ap:local_energy}. Notably, the local energy of the eigenfunctions that the predictor uses to represent the RESISC45 task is significantly higher for points belonging to the training dataset. This confirms the presence of eigenfunctions localized across the different data domains and the fact that task arithmetic occurs thanks to the use of those. Indeed, thanks to this localization, CLIP models can effectively separate the representation of different tasks and carry out task-specific operations without interference. We believe that further investigation into this intriguing localization phenomenon holds the potential to deepen our understanding of these models.\looseness=-1

\paragraph{Remark} While we have shown that localized eigenfunctions can play a crucial role in task arithmetic, it is important to note that they are not always necessary. In fact, the task arithmetic property can hold even if the eigenfunctions used to represent a single task cancel outside the corresponding domain. Indeed, although eigenfunctions are linearly independent on the union of the domains, they are not necessarily linearly independent when evaluated on a single domain and, in general, can cancel out. However, if the eigenfunctions maintain their linear independence on each of the domains, i.e., they are \textit{locally} linear independent, then the existence of localized eigenfunctions becomes a necessary condition for task arithmetic. This means that if the eigenfunctions are locally linearly independent and not localized, task arithmetic is not possible. We provide some analytical examples of the latter case in \autoref{ap:spectral}, including the NTKs of fully-connected and convolutional networks at initialization.\looseness=-1

\subsection{Disentanglement emerges during pre-training}\label{sec:random_init}

Task arithmetic is not exclusive to CLIP models. In fact, task arithmetic can also be performed on pre-trained text transformers~\cite{ilharco2023task,wortsman2022model}, such as GPT-2~\cite{radford2019language} or T5~\cite{colin2020exploring} and convolutional neural networks~\cite{ilharco2022patching}. However, it is still unclear if the origin of weight disentanglement comes from pre-training, or if it is a general property of deep networks.\looseness=-1

\begin{table}[t]

\caption{\textbf{Task addition from random initialization.} We use the same setup as for the experiments in \autoref{tab:task_addition} but with task vectors obtained from fine-tuning randomly initialized ViTs. Results compare the average single-task accuracy (\%) after fine-tuning and the multi-task accuracy (\%) via task addition.\looseness=-1}\label{tab:random_addition}

\vspace{0.5em}
\centering
\begin{tabular}{@{}lcccc@{}}
\toprule
\multirow{2}{*}{Method} & \multicolumn{2}{c}{ViT-B/32} & \multicolumn{2}{c}{ViT-L/14}  \\
                 & Sing. {\small ($\uparrow$)} & Multi {\small ($\uparrow$)} & Sing. {\small ($\uparrow$)} & Multi {\small ($\uparrow$)} \\ \midrule
Random init {\small \hspace{2.2em}$f(\cdot\,;\,\vth^{{\mathrm{rd}}}_0)$}     & 5.3     &  --      & 5.2     & --       \\ \midrule
Non-lin. FT {\small \hspace{0.4em}$f(\cdot\,;\,\vth^{{\mathrm{rd}}}_0+\vt^{{\mathrm{rd}}})$}    & 48.5 & 5.5        & 18.0    & 4.8      \\
Linear. FT  {\small \hspace{0.3em}$f_{{\mathrm{lin}}}(\cdot\,;\,\vth_{0}^{{\mathrm{rd}}}+\vt_{{\mathrm{lin}}}^{{\mathrm{rd}}})$}   & 27.8 & 3.8        & 24.8 &  6.1 \\
\bottomrule
\end{tabular}
\end{table}

To investigate this, we replicate the task addition experiments but employ randomly initialized ViTs instead of pre-trained ones. The results in \autoref{tab:random_addition} reveal that task arithmetic is not achievable on randomly initialized ViTs. Indeed, adding task vectors obtained from a random initialization $\vth_{0}^{{\mathrm{rd}}}$ does not result in significant improvements in multi-task accuracy over random chance. This holds true for both nonlinear task vectors, $\vt^{{\mathrm{rd}}}$, and linearized ones, $\vt_{{\mathrm{lin}}}^{{\mathrm{rd}}}$. In \autoref{ap:random}, we further corroborate these findings by computing the disentanglement error and the NTK spectrum of randomly initialized models.\looseness=-1

Therefore, we conclude that task arithmetic is a property acquired during pre-training. This observation goes beyond the traditional representation learning view of pre-training, emphasizing that pre-training not only leads to semantically disentangled feature representations but also to the disentanglement of the weights that govern the output on those semantic sets. Investigating the pre-training dynamics that give rise to such disentanglement is another interesting avenue for future research.

\section{Related work}

\paragraph{Weight interpolation and task arithmetic} A growing body of work is exploring the use of interpolations between model weights and task arithmetic to manipulate and enhance the capabilities of pre-trained models. In particular, several studies have shown that interpolating between a model's fine-tuned weights and its pre-trained initialization can lead to improved performance on single tasks, even surpassing their fine-tuning accuracies~\cite{wortsman2021robust,matena2021merging,frankle2020linear,izmailov2018averaging}. In the multi-task setting, averaging the parameters of multiple fine-tuned models has been proposed to produce superior multi-task models~\cite{ilharco2023task,li2022branch,ilharco2022patching,wortsman2022model} that avoid catastrophic forgetting~\cite{french1999catastrophic,mccloskey1989catastrophic} and even provide a better starting point for subsequent fine-tuning~\cite{choshen2022fusing,don2022cold}. Interestingly, the benefits of weight ensembles and interpolations extend to models trained from scratch, as long as they are properly aligned before merging~\cite{ainsworth2022git,singh2020fusion}. This phenomenon has been observed to enhance downstream performance, further emphasizing the potential of weight interpolation and task arithmetic techniques such as the ones studied in this work.\looseness=-1

\paragraph{Linear vs. nonlinear regime} Extensive research has been conducted on comparing generalization and dynamical properties of neural networks in linear and nonlinear regimes~\cite{fort2020deep,geiger2020disentangling,paccolat2021compression,baratin2021alignment,vyas2022limitations, petrini2022learning} 
and investigating specific inductive biases~\cite{yuce2022inrs,tancik2020fourier,bachmann2021uniform,chua2021fine,mu2020gradients,achille2021lqf,malladi2022kernel}. In addition to theoretical understanding, several studies have applied linearized models for practical purposes, such as predicting fine-tuning generalization~\cite{deshpande2021linearized} and training speed~\cite{zancato2020predicting}, as well as enhancing calibration~\cite{maddox2021fast} and few-shot performance~\cite{arora2020harnessing}. Our work serves as another example of the utility of linearized models in certain scenarios where they do not only offer practical benefits but also provide valuable theoretical insights.\looseness=-1

\paragraph{Feature disentanglement} The notion of feature disentanglement lies at the heart of representation learning, where ideal representations are assumed to separate distinct data variation factors along different directions in the feature space~\cite{bengio2013representation,higgins2018definition,achille2018emergence}. A multitude of approaches in generative modeling~\cite{higgins2017betavae,rezende2014stochastic,chen2016infogan} and self-supervised learning~\cite{simclr,radford2021learning,bachman2019representation,locatello2019challenging} strive to achieve this goal. Our investigation, however, explores a distinct aspect: \textit{weight disentanglement} within the framework of task arithmetic. Departing from the static perspective of feature disentanglement, weight disentanglement connects weight space and function space transitions, thereby enriching our understanding of disentanglement in neural networks from a functional standpoint. Several studies have previously attempted to exploit a similar notion by inducing the learning of task-specific subnetworks within a larger network~\cite{wortsman2020supermasks,havasi2021mimo,wen2020batchensemble,mallya2018packnet,mallya2018piggyback,masse2018alleviating,hu2022lora}. To the best of our knowledge, our work is the first to demonstrate the natural emergence of such phenomena in specific semantically meaningful tasks during CLIP pre-training.\looseness=-1

\section{Conclusions}

We conducted a thorough analysis of task arithmetic in deep neural networks, delving into its fundamental mechanisms and enhancing its performance. Our findings demonstrate that linearized models, governed by the NTK, outperform their nonlinear counterparts in task arithmetic, thus providing a more effective approach for model editing. Crucially, we revealed that weight disentanglement plays a vital role in the success of task arithmetic, as distinct directions in weight space correspond to localized areas in the function space, and that it is an emergent property of pre-training.\looseness=-1

A fascinating open question is understanding how weight disentanglement arises during pre-training and finding algorithms that enhance it. Another exciting research direction is investigating the potential of tangent spaces for editing other pre-trained models. In this sense, developing more efficient linearized models would be a significant leap forward in this field. These advancements could pave the way for novel approaches to model editing and deepen our understanding of the complex relationship between weight space and function space in deep learning.

\cleardoublepage

\ctparttext{\bigskip \bigskip \begin{flushright}{\slshape
The scientist does not study nature because it is useful; he studies it because he delights in it, and he delights in it because it is beautiful. If nature were not beautiful, it would not be worth knowing [...] I speak of that intimate beauty which comes from the harmonious order of its parts.
} \\ \medskip
--- Henri Poincaré
\end{flushright}
}

\part{Finale}

\chapter{Conclusions}

\label{ch:conclusions}

The remarkable effectiveness of deep neural networks, particularly in high-dimensional domains where learning is in principle statistically intractable, presents a foundational paradox in modern machine learning. The resolution to this paradox, as argued throughout this thesis, lies not in the learning algorithms alone, but in the latent structure inherent in natural data and the tasks we ask models to perform. This work has advanced the thesis that locality and compositionality are key principles that enable learning, providing a theoretical lens through which to understand how neural networks can overcome the curse of dimensionality. We considered a hierarchy of abstraction, from the local structure of data to the compositional algebra of tasks within a model's weight space, employing tools from statistical physics and learning theory to build a quantitative, `physical' theory of data and tasks.

This work was structured into three primary investigations. In Part II, we began with an analysis of convolutional neural networks (CNNs) in the analytically tractable limit of infinite width. 
We established that efficient learning is possible when the target function can be decomposed into a sum of spatially localized components, with generalization performance being governed by the local scale of the target function rather than the ambient data dimension.
However, we also exposed the limitations of this `lazy' learning regime, demonstrating its inability to efficiently learn functions with long-range correlations and a hierarchical structure, thereby motivating the need to move beyond kernels and toward models capable of feature learning.

Part III pivoted to generative modeling, asking how deep models learn to synthesize complex, structured data. Using the Random Hierarchy Model -- a synthetic, yet rich, model for data with built-in compositional and hierarchical structure -- we developed a theory of composition. We uncovered a phase transition in the dynamics of diffusion models, revealing a process where high-level semantic features are assembled from a set of learned lower-level components. We argued that diffusion models learn the underlying `grammar' of the data through a hierarchical clustering mechanism of features, akin to a generalized renormalization group, which requires a number of samples that scales only polynomially with data dimension. This provides a concrete mechanism by which generative models can become creative, composing novel outputs from familiar parts. Moreover, we framed the trade-off between generalization and memorization as a race between competing time scales during training.

Finally, in Part IV, we considered the composition of tasks themselves. We investigated the phenomenon of task arithmetic in large, pre-trained models, where entire skills can be manipulated through algebraic operations on model weights. We demonstrated that this capability is not merely a consequence of model linearity, as previously hypothesized. Instead, we identified weight disentanglement as the key underlying principle: a structural property, emergent from pre-training, where distinct directions in weight space correspond to localized and semantically meaningful functions.

\section{Key findings and their synthesis}

The primary contribution of this thesis is a quantitative framework for understanding how structure enables learning. Our main findings can be synthesized as follows:

\begin{enumerate}

\item \spacedlowsmallcaps{Locality is the primary driver for generalization in shallow compositional tasks} Our analysis in Part II clarified the distinct roles of architectural priors in CNNs. In the kernel regime, it is the spatial locality of the receptive field, not weight-sharing, that dictates the asymptotic scaling of the learning curve. This result provides a precise, quantitative answer to why convolutional architectures are so effective on local tasks, showing that the effective dimension for learning becomes the filter size, rather than the input dimension. This finding underscores that even simple structural assumptions, when correctly matched by the model's architecture, can yield exponential gains in sample efficiency.

\item \spacedlowsmallcaps{Hierarchical learning requires feature learning} The limits of the kernel regime became apparent when we considered hierarchical tasks involving long-range correlations. We found that even deep CNNs, when constrained to the lazy training regime, fail to efficiently learn functions generated by a matched deep architecture. This negative result is significant, as it implies that the benefits of depth for learning hierarchical functions are not merely about representation but are intrinsically tied to the dynamics of feature learning, where the kernel itself adapts to the data. This provides a clear theoretical motivation for studying the feature-learning regime.

\item \spacedlowsmallcaps{Generative models learn a `grammar' of data} Our investigation into diffusion models in Part III provided a new lens for understanding generative modeling. By studying the Random Hierarchy Model, an ensemble of simple formal grammars with random rules, we moved beyond simplistic data assumptions. The discovery of a phase transition in the reverse diffusion process -- where the semantic class of a sample can change while low-level features are preserved and recombined -- provided strong evidence for a compositional generation process. We showed that the sample complexity for learning the grammar rules scales polynomially, not exponentially, with dimension. This is because the model learns to cluster features that appear in statistically similar contexts, a hierarchical process that effectively reconstructs the latent structure of the grammar. This theory explains not only how models can generate novel, coherent data but also why models trained on limited data produce outputs that are only locally coherent, a phenomenon we confirmed in both text and image domains.

\item \spacedlowsmallcaps{Task-specific knowledge is localized} The exploration of task arithmetic in Part IV revealed a novel form of compositionality at the level of the model itself. Our central finding is that the ability to compose and subtract tasks is enabled by weight disentanglement -- an emergent property of pre-training where distinct parameter directions control functionally independent aspects of the model's behavior. By showing that this property is absent at initialization and can be enhanced by linearizing the model, we provided both a mechanistic explanation for task arithmetic and a practical method for improving it. This recasts the geometry of the weight space of large models as a structured space where tasks correspond to localized, composable linear subspaces.
\end{enumerate}
Synthesizing these points, this thesis proposes that deep learning is effective because it operates on a cascade of compositional structures. It begins with local features in the input, builds hierarchical representations that mirror the compositional nature of the world, and culminates in a modular organization of knowledge in weight space that allows for the flexible composition of skills.

\section{Comparison with other theoretical frameworks}

A central result of this thesis is that the Random Hierarchy Model -- characterized by non-Gaussian statistics -- predicts qualitatively different phenomena from other common theoretical data models. It is thus instructive to contrast our findings from Part III with two representative frameworks.\looseness=-1

\paragraph{Gaussian Mixture Models (GMMs)} Some theoretical works model data as a mixture of Gaussians, where each mode might represent a distinct class. In these models, the reverse diffusion process exhibits a crossover known as \textit{speciation}, where the dynamics collapses into one of the modes at a characteristic time or noise scale \cite{biroli2024dynamical,ambrogioni2023statistical}. Although this resembles the class phase transition we identify, it lacks compositionality. In the RHM, following the class change, low-level features from the original sample are preserved and \textit{recombined} to form the new sample. This compositional process -- a form of combinatorial creativity that we confirm empirically in natural data -- cannot be explained by GMMs. Furthermore, as mean-field models with no inherent spatial structure, GMMs do not present the growing dynamical susceptibility or length scale at the transition, which are central predictions of the RHM.

\paragraph{Gaussian statistics and spectral bias} Models like Gaussian Random Fields (GRFs), are fully specified by their second-order statistics (i.e., two-point correlations). While a GRF can be tuned to match the RHM's two-point statistics, it fundamentally lacks the higher-order correlations induced by the RHM's latent tree structure. Crucially, the RHM's context-free nature reduces complex multi-body correlations among observable tokens to effective pairwise correlations involving the latent variables. This property allows a \textit{deep} learning algorithm to efficiently infer the hierarchy by clustering token groups that correspond to the same latent variable, yielding the ``local-to-global'' learning dynamics we predict and observe. 

This mechanism cannot be explained by models based solely on second-order statistics. In these settings, learning the score is driven by a spectral bias: eigenmodes of data covariance are learned in order of decreasing variance, meaning that low-frequency (global) components are acquired first \cite{wang2025analytical}. This predicts an \textit{opposite} phenomenology, where global structure appears early in training, and local details are learned later. 

The distinction also appears in the generative process: as shown in \autoref{ch:probing}, a GRF exhibits a dynamical correlation length that grows monotonically with the inversion time. Whereas, in the RHM, the correlation length peaks at a finite time corresponding to the phase transition.\looseness=-1

The concurrence of a ``local-to-global'' learning dynamics and a susceptibility peak in real systems provides strong evidence that hierarchical compositionality, not just second-order statistics, is a key principle governing how deep models learn and generate data.\looseness=-1

\section{Limitations and future directions}

This thesis opens up several new avenues for inquiry and highlights areas where our understanding remains incomplete.

\paragraph{Beyond simplified models of hierarchy} 
Throughout Part III of this thesis, the Random Hierarchy Model (RHM) served as the basis for developing a tractable theory of generating and learning compositional data. It was conceived as a minimal model that captures the essential properties of hierarchical compositionality -- found in real-world modalities like language and vision -- without sacrificing analytical tractability. Its simplifying assumptions, such as fixed, regular tree topology and random production rules, enabled a theoretical study of the denoising process and the computation of the nested correlations that govern learning. Despite its abstraction, as common in theoretical physics models, the RHM successfully predicts non-trivial phenomena observed with natural data, including the phase transition in the denoising process and the local-to-global learning dynamics, where long-range correlations emerge progressively with training. As discussed in the previous section, these results cannot be reproduced by simpler models, such as Gaussian random fields with long-range spatial correlations, that lack the RHM's higher-order statistical structure. This alignment between theory and empirics underscores the RHM's value as a powerful conceptual tool for studying data like text and images.

Future work could extend the RHM to better capture the complexity of natural data. Promising avenues include:
\begin{itemize}
    \item \textit{Irregular tree topologies:} Natural language exhibits variable branching and depth, leading to heterogeneous trees. Extending the RHM to accommodate irregular tree topologies, e.g., adding distributions over branching factors across layers, would better reflect these hierarchies. Moreover, it would introduce a degree of \textit{ambiguity}, which is also found in language, and is currently absent from the model. In particular, this structural ambiguity means the boundaries of grammatical units become uncertain, and a single sequence of tokens could be parsed into a valid grammatical structure in multiple ways. The model would then need to learn to resolve these parsing uncertainties, likely by leveraging broader context to infer the most probable latent structure.
    \item \textit{Context dependences:} The RHM is fundamentally a probabilistic context-free grammar. However, it is established that natural language requires at least \textit{mildly context-sensitive} models to capture all syntactic phenomena. Introducing more general latent models that involve context dependencies is thus a key challenge for future work. For instance, this could be achieved by allowing for more complex latent variables that encode additional information or considering models that gradually depart from a tree-like structure.
\end{itemize}
Characterizing the correlation structures that arise from these modifications could provide deeper insights into how advanced models like transformers acquire linguistic competence.

\paragraph{The dynamics of hierarchical feature learning} 
In \autoref{ch:creativity}, we developed a theory for the sample complexity of diffusion models learning hierarchical data in the feature-learning regime. We proposed a clustering mechanism that builds a representation of the latent structure of data by leveraging statistical correlations. However, a complete, dynamical theory of how these hierarchical features are learned across layers via gradient-based optimization remains elusive. Understanding the precise dynamics by which deep networks construct multi-scale representations from scratch is a critical open problem in the field. Such an understanding would be particularly useful for rationalizing the findings of \autoref{ch:memorization} on the time scales required to learn different hierarchical levels.

\paragraph{Probing latent structures in natural data} A key finding of this thesis is that the forward-backward diffusion process acts as a powerful lens on the compositional nature of data. This suggests a compelling new direction: using these experiments as a data-driven method to probe and extract the latent hierarchical structure of real data. One could, for instance, interpret the large, correlated blocks of changing tokens in language as revealing grammatical constituents or context variables. This approach, for instance, could lead to novel methods for discovering the structure of natural language, driven entirely by the dynamics of diffusion models. 

Moreover, the hierarchical clustering mechanism described in Part III, where latent variables are constructed to preserve predictive power about their context, can be understood as a generalization of the renormalization group (RG) from theoretical physics. In contrast to the traditional RG, which defines coarse-grained variables through fixed rules like local averaging, the principle we put forward allows for the construction of latent variables that are complex functions of the input and can vary across scales. This flexible framework opens the exciting possibility of building new theories for complex systems where the standard RG has had limited success, such as in the study of turbulence, suggesting principles uncovered in deep learning could provide new conceptual tools for the physical sciences.

\paragraph{The origins of weight disentanglement} Our finding that weight disentanglement is an emergent property of pre-training is a crucial first step. However, the question of how this emergence occurs is unanswered. What aspects of self-supervised objectives and large-scale, diverse data conspire to produce this modular structure in weight space? Can we design pre-training schemes that explicitly optimize for this property, leading to models that are more robustly and efficiently editable? Investigating the pre-training dynamics that lead to a disentangled functional geometry is a promising and important direction.

\section{Concluding remarks}

The perspective of this thesis has been that of a physicist approaching a complex system: seeking to identify the fundamental principles and symmetries that govern its behavior. We have argued that for deep learning, two key principles are locality and compositionality. By understanding how neural networks find and exploit these structures at various levels of abstraction, we move closer to a principled and predictive science of artificial intelligence. In particular, the frameworks and concepts developed here -- from the quantitative impact of locality on generalization, to the compositional dynamics of generative models, to the notion of weight disentanglement -- provide new tools and a new language for dissecting the success of deep learning. As we continue to build larger and more capable models, a deep understanding of these foundational principles will not merely be an academic pursuit but an essential prerequisite for creating AI systems that are robust, interpretable, and universally useful.

% %----------------------------------------------------------------------------------------
% %	THESIS CONTENT - APPENDICES
% %----------------------------------------------------------------------------------------

\appendix

\ctparttext{\bigskip \bigskip \begin{flushright}{\slshape
More is different.} \\ \medskip
--- Philip W. Anderson
\end{flushright}
}

\part{Appendix}

\chapter{Appendix: Locality Defeats the Curse of Dimensionality}

\section{Spectral bias in kernel regression}\label{app:loc-spectral}

In this appendix, we provide additional details about the derivation of~\autoref{eq:loc-error-scaling} within the framework of~\cite{bordelon2020spectrum, canatar2021spectral}. Let us begin by recalling the definition of the kernel ridge regression estimator $\hat{f}$ of a target function $f^*$,
\begin{equation}\label{eq:loc-argmin-app}
 \hat{f} =\argmin_{f\in \mathcal{H}} \left\lbrace \frac{1}{P}\displaystyle \sum_{\nu=1}^P \left(f(\x_\nu) - f^*(\x_\nu)\right)^2  + \lambda \, \|f\|^2_{\mathcal{H}} \right  \rbrace,
\end{equation}
where $\mathcal{H}$ denotes the Reproducing Kernel Hilbert Space (RKHS) of the kernel $\K(\x,\y)$. After introducing the Mercer's decomposition of the kernel,
\begin{equation}\label{eq:loc-mercer-app}
 \K(\x,\y) = \sum_{\rho=1}^\infty \lambda_\rho \phi_\rho(\x) \overline{\phi_\rho(\y)}, \quad \int p\left( d^d y\right) \K(\x,\y) \phi_\rho(\y) = \lambda_\rho \phi_\rho(\x). 
\end{equation}
the RKHS can be characterized as a subset of the space of functions lying in the span of the kernel eigenbasis,
\begin{equation}\label{eq:loc-RKHS-app}
    \mathcal{H} = \left\lbrace f = \sum_{\rho=1}^\infty a_\rho \phi_\rho(\x)\right. \left| \; \sum_{\rho=1}^\infty \frac{|a_\rho|^2}{\lambda_\rho} <\infty \right\rbrace.
\end{equation}
In other words, the RKHS contains functions having a finite norm $||f||_\mathcal{H} \eq \sqrt{\left\langle f, f\right\rangle_{\mathcal{H}}}$ with respect to the following inner product,
\begin{equation}
f(\x)=\sum_\rho a_\rho \phi_\rho(\x),\; f'(\x)=\sum_\rho a'_\rho \phi_\rho(\x), \; \left\langle f, f'\right\rangle_{\mathcal{H}} = \sum_\rho \frac{a_\rho a'_\rho}{\lambda_\rho}.
\end{equation}
Given any target function $f^*$ lying in the span of the kernel eigenbasis, the mean-squared generalization error of the kernel ridge regression estimator reads
\begin{equation}\label{eq:loc-error-app}
    \testerr(\lambda, \left\lbrace \x_\nu\right\rbrace) = \int p(d^d\x) \left(\hat{f}(\x)-f^*(\x)\right)^2 = \sum_{\rho=1}^\infty \left| a_\rho(\lambda, \left\lbrace \x_\nu\right\rbrace) -c_\rho\right|^2,
\end{equation}
with $c_\rho$ denoting the $\rho$-th coefficient of the target $f^*$ and $a_\rho$ that of the estimator $\hat{f}$, which depends on the ridge $\lambda$ and on the training set $\left\lbrace \x_\nu\right\rbrace_{\nu=1,\dots,P}$. Notice that, as $\hat{f}$ belongs to $\mathcal{H}$ by definition, $\sum_\rho |a_\rho|^2/\lambda_\rho \,{<}\,+\infty$, whereas the $c_\rho$'s are free to take any value.

The authors of~\cite{bordelon2020spectrum, canatar2021spectral} found a heuristic expression for the average of the mean squared error over realizations of the training set $\left\lbrace \x_\nu\right\rbrace$. Such expression, based on the replica method of statistical physics, reads\footnote{Notice that the risk considered in \cite{bordelon2020spectrum, canatar2021spectral} slightly differs from \autoref{eq:loc-argmin-app} by a factor $1/P$ in front of the sum.}
\begin{equation}\label{eq:loc-bordelon1}\testerr(\lambda, P) = \partial_{\lambda}\left(\frac{\kappa_{\lambda}(P)}{P}\right) \sum_\rho \frac{\kappa_\lambda(P)^2}{\left(P \lambda_\rho + \kappa_\lambda(P)\right)^2} |c_\rho|^2 ,\end{equation}
where $\kappa(P)$ satisfies
\begin{equation}\label{eq:loc-bordelon2} \frac{\kappa_\lambda(P)}{P} = \lambda + \frac{1}{P}\sum_\rho \frac{\lambda_\rho \kappa_\lambda(P)/P}{\lambda_\rho + \kappa_\lambda(P)/P}.
\end{equation}
In short, the replica method works as follows~\cite{Mezard87}: first one defines an energy function $E(f)$ as the argument of the minimum in~\autoref{eq:loc-argmin-app}, then attribute to the predictor $f$ a Boltzmann-like probability distribution $P(f) = Z^{-1} e^{-\beta E(f)}$ , with $Z$ a normalization constant and $\beta\,{>}\,0$. As $\beta\to\infty$, the probability distribution $P(f)$ concentrates around the solution of the minimization problem of~\autoref{eq:loc-argmin-app}, i.e., the predictor of kernel regression. Hence, one can replace $f$ in the right-hand side of~\autoref{eq:loc-error-app} with an average over $P(f)$ at finite $\beta$, then perform the limit $\beta\to\infty$ after the calculation so as to recover the generalization error. The simplification stems from the fact that, once $f$ is replaced with its eigendecomposition, the energy function $E(f)$ becomes a quadratic function of the coefficients $c_\rho$. Then, under the assumption that the data distribution enters only via the first and second moments of the eigenfunctions $\phi_{\rho}(\x)$ w.r.t. $\x$, all averages in~\autoref{eq:loc-error-app} reduce to Gaussian integrals.

Mathematically, $\kappa_\lambda(P)/P$ is related to the Stieltjes transform~\cite{potters2020first} of the Gram matrix $\mathbf{K}_P/P$ in the large-$P$ limit. Intuitively, it plays the role of a threshold: the modal contributions to the error tend to $0$ for $\rho$ such that $\lambda_\rho \gg k_\lambda(P)/P$, and to $\mathbb{E}[|c_\rho|^2]$ for $\rho$ such that $\lambda_\rho \ll k_\lambda(P)/P$. This is equivalent to saying that the algorithm predictor $f(\x)$ captures only the eigenmodes having eigenvalue larger than $k_\lambda(P)/P$ (see also~\cite{jacot2020implicit, jacot2020kernel}). 

This intuitive picture can actually be exploited in order to extract the learning curve exponent $\beta$ from the asymptotic behavior of~\autoref{eq:loc-bordelon1} and~\autoref{eq:loc-bordelon2} in the ridgeless limit $\lambda\to 0^+$. In the following, we assume that both the kernel and the target function have a power-law spectrum, in particular $\lambda_\rho \sim \rho^{-a}$ and $\mathbb{E}[|{c^*_\rho}|^2] \sim \rho^{-b}$, with $2a\,{>}\,b-1$. First, we approximate the sum over modes in \autoref{eq:loc-bordelon2} with an integral using the Euler-Maclaurin formula. Then we substitute the eigenvalues inside the integral with their asymptotic limit, $\lambda_\rho = A\rho^{-a}$. Since, $\kappa_0(P)/P \to 0$ as $P \to \infty$, both these operations result in an error which is asymptotically independent of $P$. Namely,
\begin{align}
    \frac{\kappa_0(P)}{P} &= \frac{\kappa_0(P)}{P} \frac{1}{P} \left(\int_0^\infty \frac{d\rho \, A\rho^{-a} }{A\rho^{-a} + \kappa_0(P)/P} + \mathcal{O}(1) \right) \\
    &= \frac{\kappa_0(P)}{P} \frac{1}{P} \left( \left( \frac{\kappa_0(P)}{P} \right)^{-\frac{1}{a}}\int_0^\infty \frac{d\sigma \, \sigma^{\frac{1}{a}-1}A^{\frac{1}{a}}a^{-1}}{1 + \sigma} + \mathcal{O}(1) \right), \nonumber
\end{align}
where in the second line, we changed the integration variable from $\rho$ to $\sigma\,{=}\, \kappa_0(P)\rho^a/( A P)$. Since the integral in $\sigma$ is finite and independent of $P$, we obtain that $\kappa_0(P)/P = \mathcal{O}(P^{-a})$. Similarly, we find that the mode-independent prefactor $\partial_\lambda \left(\kappa_\lambda(P)/P\right)|_{\lambda=0} = \mathcal{O}(1)$.
As a result we are left with, modulo some $P$-independent prefactors,
\begin{equation}\label{eq:loc-error-scaling1} \testerr(P) \sim \sum_{\rho} \frac{P^{-2a}}{\left(A\rho^{-a}+P^{-a}\right)^2}\mathbb{E}[|c_\rho|^2].\end{equation}
Following the intuitive argument about the thresholding role of $\kappa_0(P)/P \sim P^{-a}$, it is convenient to split the sum in~\autoref{eq:loc-error-scaling1} into sectors where $\lambda_\rho\gg\kappa_0(P)/P $, $\lambda_\rho \sim \kappa_0(P)/P $ and $\lambda_\rho\ll\kappa_0(P)/P $, i.e.,
\begin{equation}
\testerr(P) \sim \sum_{\rho \ll P} \frac{P^{-2a}}{\left(A\rho^{-a}\right)^2}\mathbb{E}[|c_\rho|^2] +\sum_{\rho \sim P} \frac{1}{2}\mathbb{E}[|c_\rho|^2] + \sum_{\rho \gg P} \mathbb{E}[|c_\rho|^2].\end{equation}
Finally, ~\autoref{eq:loc-error-scaling} is obtained by noticing that, under our assumptions on the decay of $\mathbb{E}[|c_\rho|^2]$ with $\rho$, the contribution of the sum over $\rho\ll P$ is subleading in $P$ whereas the other two sums can be gathered together.

\section{NTKs of convolutional and locally-connected networks}\label{app:loc-ntk}

We begin this section by reviewing the computation of the NTK of a one-hidden-layer fully-connected network \cite{chizat2019lazy}.

\begin{definition}[one-hidden-layer FCN]
A one-hidden-layer fully-connected network with $H$ hidden neurons is defined as follows,
\begin{equation}\label{eq:loc-fcn}
    f^{FCN}(\x) = \frac{1}{\sqrt{H}} \sum_{h=1}^H a_h \sigma(\bm{w}_{h} ^\top  \x + b_h),
\end{equation}
where $\x\in\mathbb{R}^d$ is the input, $H$ is the width, $\sigma$ is a nonlinear activation function, $\{\bm{w}_{h}\in\mathbb{R}^d\}_{h=1}^H$, $\{b_{h}\in\mathbb{R}\}_{h=1}^H$, and $\{a_{h}\in\mathbb{R}\}_{h=1}^H$ are the network's parameters.
\end{definition}

Inserting \autoref{eq:loc-fcn} into \autoref{eq:loc-finite_ntk}, one obtains
\begin{align}\label{eq:loc-rand-feat-fc}
    \K_{\mathrm{NTK},N}^{FC}(\x,\y;\bm{\theta}) = \frac{1}{H} \sum_{h=1}^H &\left( \sigma(\bm{w}_h ^\top  \x + b_h) \sigma (\bm{w}_h ^\top  \y + b_h) \right. \\
    &+ \left. a_h^2 \sigma'(\bm{w}_h ^\top  \x + b_h) \sigma'(\bm{w}_h ^\top  \y + b_h) (\x ^\top  \y + 1) \right), \nonumber
\end{align}
where $\sigma'$ denotes the derivative of $\sigma$ with respect to its argument. If all the parameters are initialized independently from a standard Normal distribution, $\K_{\mathrm{NTK},N}^{FC}(\x,\y;\bm{\theta})$ is a random-feature kernel that in the $H \to \infty$ limit converges to
\begin{align}\label{eq:loc-fc-ntk}
    \K_{\mathrm{NTK}}^{FC}(\x,\y) &=  \mathbb{E}_{\bm{w},b}[\sigma(\bm{w} ^\top  \x + b) \sigma (\bm{w} ^\top  \y + b)] \\ 
    &+ \mathbb{E}_{a}[a^2] \mathbb{E}_{\bm{w},b}[\sigma'(\bm{w} ^\top  \x + b) \sigma'(\bm{w} ^\top  \y + b)] (\x ^\top  \y + 1). \nonumber
\end{align}
When $\sigma$ is the ReLU activation function, the expectations can be computed exactly using techniques from the literature of arc-cosine kernels \cite{cho2009kernel}
\begin{align}\label{eq:loc-relu-fc-ntk}
	\K_{\mathrm{NTK}}^{FC}(\x,\y) &= \frac{1}{2\pi} \sqrt{\|\x\|^2+1} \sqrt{\|\y\|^2+1} \, (\sin\varphi + (\pi-\varphi)\cos\varphi) \\ &+ \frac{1}{2\pi} (\x ^\top  \y + 1) (\pi-\varphi), \nonumber
\end{align}
with $\varphi$ denoting the angle
\begin{equation}
	\varphi = \arccos \left( \frac{\x ^\top  \y + 1}{\sqrt{\|\x\|^2+1} \sqrt{\|\y\|^2+1}}\right).
\end{equation}
Notice that, as commented in \autoref{sec:loc-convolutional-mercer}, for ReLU networks $\K_{\mathrm{NTK}}^{FC}(\x,\y)$ displays a cusp at $\x \eq \y$.

\paragraph{Proof of Lemma \ref{lemma:cntk}}

\begin{proof}

Inserting \autoref{eq:loc-cnn-out} into \autoref{eq:loc-finite_ntk},
\begin{align}\label{eq:loc-rand-feat-cn}
    \K_{\mathrm{NTK},N}^{CN}(\x,\y;\bm{\theta}) =  \frac{1}{|\mathcal{P}|^2} \sum_{i,j\in\mathcal{P}} \Biggl( \frac{1}{H} \sum_{h=1}^H \bigl( \sigma(\bm{w}_h ^\top  \x_i + b_h) \sigma (\bm{w}_h ^\top  \y_j + b_h) \nonumber \\ + a_h^2 \sigma'(\bm{w}_h ^\top  \x_i + b_h) \sigma'(\bm{w}_h ^\top  \y_j + b_h) (\x_i ^\top  \y_j + 1) \bigr) \Biggr)
\end{align}
In the previous line, the single terms of the summation over patches are the random-feature kernels $\K_{\mathrm{NTK},N}^{FC}$ obtained in \autoref{eq:loc-rand-feat-fc} acting on $s$-dimensional inputs, i.e., the patches of $\x$ and $\y$. Therefore,
\begin{equation}\label{eq:loc-conv-ntk-finite}
    \K_{\mathrm{NTK},N}^{CN}(\x,\y;\bm{\theta}) =  \frac{1}{|\mathcal{P}|^2} \sum_{i,j\in\mathcal{P}} \K_{\mathrm{NTK},N}^{FC}(\x,\y).
\end{equation}
If all the parameters are initialized independently from a standard Normal distribution, the $H \to \infty$ limit of~\autoref{eq:loc-conv-ntk-finite} yields \autoref{eq:loc-conv-ntk}.

\end{proof}

\paragraph{Proof of Lemma \ref{lemma:lntk}}

\begin{proof}

Inserting \autoref{eq:loc-lcn-out} into \autoref{eq:loc-finite_ntk},
\begin{align}\label{eq:loc-rand-feat-lc}
   \K_{\mathrm{NTK},N}^{LC}(\x,\y;\bm{\theta}) = \frac{1}{|\mathcal{P}|} \sum_{i\in\mathcal{P}} \Biggl( \frac{1}{H} \sum_{h=1}^H  \bigl( \sigma(\bm{w}_{h,i} ^\top  \x_i + b_{h,i}) \sigma (\bm{w}_{h,i} ^\top  \y_i + b_{h,i}) \nonumber
    \\ + a_{h,i}^2 \sigma'(\bm{w}_{h,i} ^\top  \x_i + b_{h,i}) \sigma'(\bm{w}_{h,i} ^\top  \y_i + b_{h,i}) (\x_i ^\top  \y_i + 1) \bigl) \Biggl).
\end{align}
In the previous line, the single terms of the summation over patches are the random-feature kernels $\K_{\mathrm{NTK},N}^{FC}$ obtained in \autoref{eq:loc-rand-feat-fc} acting on $s$-dimensional inputs, i.e., the patches of $\x$ and $\y$. Therefore, 
\begin{equation}
   \K_{\mathrm{NTK},N}^{LC}(\x,\y;\bm{\theta}) =  \frac{1}{|\mathcal{P}|} \sum_{i\in\mathcal{P}} \K_{\mathrm{NTK},N}^{FC}(\x_i,\y_i).
\end{equation}
If all the parameters are initialized independently from a standard Normal distribution, \autoref{eq:loc-loc-ntk} is recovered in the $H\to\infty$ limit.

\end{proof}

\section{Mercer's decomposition of convolutional and local kernels}\label{app:loc-mercer-overlap}

In this section, we prove the eigendecompositions introduced in Lemma~\ref{lemma:conv-spectra} and Lemma~\ref{lemma:loc-spectra}, then extend them to overlapping-patches kernel (cf.~\ref{asec:mercer-overlapping}). We define the scalar product in input space between two (complex) functions $f$ and $g$ as
\begin{equation}
    \left\langle f, g\right\rangle = \int p(d^dx)\, f(\x) \overline{g(\x)}.
\end{equation}
\paragraph{Proof of Lemma \ref{lemma:conv-spectra}}

\begin{proof}

We start by proving orthonormality of the eigenfunctions. By writing the $d$-dimensional eigenfunctions $\Phi_\rho$ in terms of the $s$-dimensional eigenfunctions $\phi_\rho$ of the constituent kernel as in \autoref{eq:loc-conv-spectrum}, for $\rho, \sigma \, {\neq} \, 1$,
\begin{equation}
    \left\langle \Phi_\rho, \Phi_\sigma\right\rangle = \frac{s}{d} \sum_{i,j\in\mathcal{P}} \int p(d^dx) \phi_\rho(\x_i) \overline{\phi_\sigma(\x_j)}.
\end{equation}
Separating the term in the sum over patches in which $i$ and $j$ coincide from the others, and since the patches are not overlapping, the RHS can be written as
\begin{equation}
     \frac{s}{d} \sum_{i\in\mathcal{P}} \int p(d^sx_i) \phi_\rho(\x_i) \overline{\phi_\sigma(\x_i)} + \sum_{i,j\neq i\in\mathcal{P}} \int p(d^sx_i) \phi_\rho(\x_i) \int p(d^sx_j) \overline{\phi_\sigma(\x_j)}.
\end{equation}
From the orthonormality of the eigenfunctions $\phi_\rho$, the first integral is non-zero and equal to one only when $\rho\eq\sigma$, while, from assumption \textit{i)}, $\int p^{(s)}(d^s x) \phi_\rho(\x)\eq0$ for all $\rho\,{>}\,1$, so that the second integral is always zero. Therefore,
\begin{equation}
    \left\langle \Phi_\rho, \Phi_\sigma\right\rangle = \delta_{\rho,\sigma},\text{ for }\rho,\sigma > 1.
\end{equation}
When $\rho\eq1$ and $\sigma\,{\neq}\,1$, $\int p(d^dx) \Phi_1(\x) \overline{\Phi_\sigma(\x)}\eq0$ from assumption \textit{i)}, i.e., $\Phi_1\eq1$ and $\int p^{(s)}(d^s x) \phi_\rho(\x)\eq0$ for all $\rho\,{>}\,1$. Finally, if $\rho\eq\sigma\eq1$, $\int p(d^dx) \Phi_1(\x) \overline{\Phi_1(\x)}\eq1$ trivially.
Then, we prove that the eigenfunctions and the eigenvalues defined in \autoref{eq:loc-conv-spectrum} satisfy the kernel eigenproblem. For $\rho\eq1$,
\begin{equation}
    \int p(d^dy) \K^{CN}(\x,\y) = \int p(d^dy) \frac{s^2}{d^2} \sum_{i,j\in\mathcal{P}} \mathcal{C}(\x_i, \y_j) = \frac{s^2}{d^2} \sum_{i,j\in\mathcal{P}} \lambda_1 = \Lambda_1,
\end{equation}
where we used $\int p^{(s)}(d^sy) \mathcal{C}(\x,\y) \eq \lambda_1$ from assumption \textit{i)}. For $\rho>1$,
\begin{equation}
    \int p(d^dy) \K^{CN}(\x,\y) \Phi_\rho(\y) = \int p(d^dy) \frac{s^2}{d^2} \sum_{i,j\in\mathcal{P}} \mathcal{C}(\x_i, \y_j) \sqrt{\frac{s}{d}} \sum_{l\in\mathcal{P}} \phi_\rho(\y_l).
\end{equation}
Splitting the sum over $l$ into the term with $l \eq j$ and the remaining ones, the RHS can be written as
\begin{align}
    \frac{s^2}{d^2} \sum_{i,j\in\mathcal{P}} &\Biggl( \int p(d^sy_j) \mathcal{C}(\x_i, \y_j) \sqrt{\frac{s}{d}} \phi_\rho(\y_j) \\ &+ \int p(d^sy_j) \mathcal{C}(\x_i, \y_j) \sqrt{\frac{s}{d}} \sum_{l\neq j\in\mathcal{P}} \int p(d^sy_l) \phi_\rho(\y_l) \Biggr). \nonumber
\end{align}
Using assumption \textit{i)}, the third integral is always zero, therefore
\begin{equation}
    \int p(d^dy) \K^{CN}(\x,\y) \Phi_\rho(\y) = \frac{s^2}{d^2} \sum_{i,j\in\mathcal{P}} \lambda_\rho \sqrt{\frac{s}{d}} \phi_\rho(\x_i) = \Lambda_\rho \Phi_\rho(\x).
\end{equation}
Finally, we prove the expansion of \autoref{eq:loc-conv-decomp} from the definition of $\K^{CN}$,
\begin{align}
    \K^{CN}(\x, \y) &= \frac{s^2}{d^2} \sum_{i,j\in\mathcal{P}} \mathcal{C}(\x_i, \y_j) \\
    &= \frac{s^2}{d^2} \sum_{i,j\in\mathcal{P}} \sum_\rho \lambda_\rho \phi_\rho(\x_i) \overline{\phi_\rho(\y_j)} \nonumber \\
    &= \lambda_1 \frac{s^2}{d^2} \sum_{i,j\in\mathcal{P}} \phi_1(\x_i) \overline{\phi_1(\y_j)} \nonumber \\
    &+ \sum_{\rho>1} \left( \frac{s}{d} \lambda_\rho \right) \left( \sqrt{\frac{s}{d}} \sum_{i\in\mathcal{P}} \phi_\rho(\x_i) \right) \left( \sqrt{\frac{s}{d}}\sum_{j\in\mathcal{P}} \overline{\phi_\rho(\y_j)} \right) \nonumber \\
    &= \sum_{\rho} \Lambda_\rho \Phi_\rho(\x) \overline{\Phi_\rho(\y)}. \nonumber
\end{align}

\end{proof}

\paragraph{Proof of Lemma \ref{lemma:loc-spectra}}

\begin{proof}

We start again by proving the orthonormality of the eigenfunctions. By writing the $d$-dimensional eigenfunctions $\Phi_{\rho,i}$ in terms of the $s$-dimensional eigenfunctions $\phi_\rho$ of the constituent kernel as in \autoref{eq:loc-loc-spectrum}, for $\rho, \sigma \, {\neq} \, 1$,
\begin{equation}
    \left\langle \Phi_{\rho,i}, \Phi_{\sigma,j}\right\rangle = \int p(d^dx) \phi_\rho(\x_i) \overline{\phi_\sigma(\x_j)} = \delta_{\rho,\sigma} \delta_{i,j},
\end{equation}
from the orthonormality of the eigenfunctions $\phi_\rho$ when $i \eq j$, and assumption \textit{i)}, $\int p^{(s)}(d^s x) \phi_\rho(\x)\eq0$ for all $\rho\,{>}\,1$, when $i\,{\neq}j$. Moreover, as $\Phi_1(\x)\,{=}\,1$, $\int p(d^dx) \Phi_{1}(\x) \overline{\Phi_{\sigma\neq1,j}(\x)} \eq 0$ and $\int p(d^dx) \Phi_1(\x) \overline{\Phi_1(\x)}\eq1$. 
Then, we prove that the eigenfunctions and the eigenvalues defined in \autoref{eq:loc-loc-spectrum} satisfy the kernel eigenproblem. For $\rho\eq1$,
\begin{equation}
    \int p(d^dy) \K^{LC}(\x,\y) = \int p(d^dy) \frac{s}{d} \sum_{i\in\mathcal{P}} \mathcal{C}(\x_i, \y_i) = \frac{s}{d} \sum_{i\in\mathcal{P}} \lambda_1 = \Lambda_1,
\end{equation}
where we used $\int p^{(s)}(d^sy) \mathcal{C}(\x,\y) \eq \lambda_1$ from assumption \textit{i)}. For $\rho>1$,
\begin{equation}
    \int p(d^dy) \K^{LC}(\x,\y) \Phi_{\rho,i}(\y) = \int p(d^dy) \frac{s}{d} \sum_{j\in\mathcal{P}} \mathcal{C}(\x_j, \y_j) \phi_\rho(\y_i).
\end{equation}
Splitting the sum over $j$ in the term for which $j\eq i$ and the remaining ones, the RHS can be written as
\begin{equation}
    \frac{s}{d} \int p(d^sy_i) \mathcal{C}(\x_i, \y_i) \phi_\rho(\y_i) +  \frac{s}{d} \sum_{j \neq i\in\mathcal{P}} \int p(d^sy_j) \mathcal{C}(\x_j, \y_j) \int p(d^sy_i) \phi_\rho(\y_i).
\end{equation}
Using assumption \textit{i)}, the third integral is always zero, therefore
\begin{equation}
    \int p(d^dy) \K^{CN}(\x,\y) \Phi_\rho(\y) = \frac{s}{d} \lambda_\rho \phi_\rho(\x_i) = \Lambda_{\rho,i} \Phi_{\rho,i}(\x).
\end{equation}
Finally, we prove the expansion of \autoref{eq:loc-conv-decomp} from the definition of $\K^{LC}$,
\begin{align}
    \K^{LC}(\x, \y) &= \frac{s}{d} \sum_{i\in\mathcal{P}} \mathcal{C}(\x_i, \y_i) \\
    &= \frac{s^2}{d^2} \sum_{i\in\mathcal{P}} \sum_\rho \lambda_\rho \phi_\rho(\x_i) \overline{\phi_\rho(\y_i)} \\
    &= \lambda_1 \frac{s}{d} \sum_{i\in\mathcal{P}} \phi_1(\x_i) \overline{\phi_1(\y_i)} + \sum_{\rho>1} \sum_{i\in\mathcal{P}}  \left( \frac{s}{d} \lambda_\rho \right)  \phi_\rho(\x_i) \overline{\phi_\rho(\y_i)}  \\
    &= \Lambda_1 \Phi_1(\x) \overline{\Phi_1(\y)} +  \sum_{\rho>1} \sum_{i\in\mathcal{P}} \Lambda_{\rho,i} \Phi_{\rho,i}(\x) \overline{\Phi_{\rho,i}(\y)}.
\end{align}

\end{proof}

\subsection{Spectra of convolutional kernels with overlapping patches}\label{asec:mercer-overlapping}

In this section Lemma~\ref{lemma:conv-spectra} and Lemma~\ref{lemma:loc-spectra} are extended to kernels with overlapping patches, having $\mathcal{P}\,{=}\,\left\lbrace 1,\dots, d\right\rbrace$ and $|\mathcal{P}|\,{=}\,d$. 
Such an extension requires additional assumptions, which are stated below:
\begin{itemize}
    \item[$i)$] The $d$-dimensional input measure $p^{(d)}(d^d x)$ is uniform on the $d$-torus $[0,1]^d$;
    \item[$ii)$] The constituent kernel $\mathcal{C}(\x,\y)$ is translationally-invariant, isotropic and periodic,
    \begin{equation}
        \mathcal{C}(\x,\y) = c(||\x-\y||),\quad c(||\x-\y + \mathbf{n}||) = c(||\x-\y||) \quad \forall \mathbf{n}\in\mathbb{Z}^s.
    \end{equation}
\end{itemize}
Assumptions $i)$ and $ii)$ above imply that $\mathcal{C}(\x,\y)$ can be diagonalized in Fourier space, i.e., (with $\k$ denoting the $s$-dimensional wavevector)
\begin{equation}\label{eq:loc-constituent-fourier}
    c(\x-\y) = \sum_{\left\lbrace \k=2\pi\mathbf{n}| \mathbf{n}\in\mathbb{Z}^s\right\rbrace} \lambda_{\k} \phi_{\k}(\x) \overline{\phi_{\k}(\y)} = \sum_{\left\lbrace \k=2\pi\mathbf{n}| \mathbf{n}\in\mathbb{Z}^s\right\rbrace} \lambda_{\k} e^{i\k^\top (\x-\y)},
\end{equation}
and the eigenvalues $\lambda_{\k}$ depend only on the modulus of $\k$, $k\,{=}\,\sqrt{\k^\top \k}$.

Let us introduce the following definitions, after recalling that a $s$-dimensional patch $\x_i$ of $\x$ is a contiguous subsequence of $\x$ starting at $x_i$, i.e.
\begin{equation}
    \x=(x_1, x_2, \dots, x_d) \Rightarrow \x_i = (x_i, x_{i+1}, \dots, x_{i+s-1}),
\end{equation}
and that inputs are `wrapped', i.e., we identify $x_{i+ n d}$ with $x_i$ for all $n\in\mathbb{Z}$.

\begin{itemize}
    \item Two patches $\x_i$ and $\x_j$ \emph{overlap} if $\x_i\displaystyle \cap \x_j\,{\neq}\,\emptyset$. The overlap $\x_{i\cap j}\equiv \x_i\displaystyle \cap \x_j$ is an $o$-dimensional patch of $\x$, with $o\,{=}\,|\x_i\displaystyle \cap \x_j|$;
    \item let $\mathcal{P}$ denote the set of patch indices associated with a given kernel/architecture. We denote with $\mathcal{P}_i$ the set of indices of patches which overlap with $\x_i$, i.e., $$\mathcal{P}_i\,{=}\,\left\lbrace i-s +1,\dots, i, \dots, i + s-1\right\rbrace\,{=}\,\left\lbrace \mathcal{P}_{-,i}, i ,\mathcal{P}_{+,i} \right\rbrace;$$
    \item Given two overlapping patches $\x_i$ and $\x_j$ with $o$-dimensional overlap, the union $\x_{i\cup j}\equiv \x_i \displaystyle\cup \x_j$ and differences $\x_{i \smallsetminus j} \equiv \x_i \,{\smallsetminus}\, \x_j$ and $\x_{j\smallsetminus i} \equiv \x_j \,{\smallsetminus}\, \x_i$ are all patches of $\x$, with dimensions $2s\,{-}\,o$, $s\,{-}\,o$ and $s\,{-}\,o$, respectively.
\end{itemize}

We also use the following notation for denoting subspaces of the $\k$-space $\cong \mathbb{Z}^s$,
\begin{equation}
    \mathcal{F}^{u} = \left\lbrace \k\,{=}\,2\pi \mathbf{n} \, | \, \mathbf{n}\in\mathbb{Z}^{s}; \, n_1, n_u \neq 0; \, n_{v} = 0 \, \forall v \text{ s.t. } u<v\leq s  \right\rbrace.
\end{equation}
$\mathcal{F}^{s}$ is the set of all wavevectors $\k$ having nonvanishing extremal components $k_1$ and $k_s$. For $u\,{<}\,s$, $\mathcal{F}^{u}$ is formed by first considering only wavevectors having the last $s-u$ components equal to zero, then asking the resulting $u$-dimensional wavevectors to have nonvanishing extremal components. Practically, $\mathcal{F}^{u}$ contains wavevectors which can be entirely specified by the first $u$-dimensional patch $\k^{(u)}_1\,{=}\,(k_1,\dots,k_u)$ but not by the first $(u\,{-}\,1)$-dimensional one. Notice that, in order to safely compare $\k$'s in different $\mathcal{F}$'s, we have introduced an apex $u$ denoting the dimensionality of the patch.

\begin{lemma}[Spectra of overlapping convolutional kernels]\label{lemma:conv-spectra-overlap}
Let $\K^{CN}$ be a convolutional kernel defined as in~\autoref{eq:loc-conv-ker}, with $\mathcal{P}\,{=}\,\left\lbrace1,\dots, d\right\rbrace$ and constituent kernel $C$ satisfying assumptions $i)$, $ii)$ above. Then, $\K^{CN}$ admits the following Mercer's decomposition,
\begin{equation}\label{eq:loc-conv-decomp-overlap}
   \K^{CN}(\x,\y) = \Lambda_{\bm{0}} + \displaystyle\sum_{u=1}^{s}\left( \displaystyle\sum_{ \k\in \mathcal{F}^{u}} \Lambda_{\k} \Phi_{\k}(\x) \Phi_{\k}(\y)\right),
\end{equation}
with eigenfunctions
\begin{equation}\label{eq:loc-conv-eigvec-overlap}
 \Phi_{\bm{0}}(\x)\,{=}\,1, \quad \Phi_{\k}(\x)\,{=}\,\frac{1}{\sqrt{d}}\sum_{i=1}^d \phi_{\k}(\x_i) \quad \forall\, \k\neq\bm{0},
\end{equation}
and eigenvalues
\begin{equation}\label{eq:loc-conv-eigval-overlap}
 \Lambda_{\bm{0}}\,{=}\,\lambda_{\bm{0}}, \quad \Lambda_{\k}\,{=}\, \frac{s-u+1}{d}\lambda_{\k} \quad \forall\, \k\in\mathcal{F}^{u} \text{ with } u\leq s.
\end{equation}
\end{lemma}

\begin{proof} We start by proving the orthonormality of the eigenfunctions. In general, by orthonormality of the $s$-dimensional plane waves $\phi_{\k}(\x)$, we have

\begin{align}\label{eq:loc-scalar-product-overlapping}
&\left\langle \Phi_{\k}, \Phi_{\bm{q}}\right\rangle = \frac{1}{d}\int_{[0,1]^d} d^d x\, \left(\displaystyle\sum_{i=1}^d \phi_{\k}(\x_i)\right) \overline{\left(\displaystyle\sum_{j=1}^d \phi_{\bm{q}}(\x_j)\right)} \nonumber \\
&=\frac{1}{d}\displaystyle\sum_{i\in\mathcal{P}}\displaystyle\sum_{j\notin\mathcal{P}_i} \int d^s x_i\, e^{i\k^\top \x_i} \int d^s x_j \, e^{-i\bm{q}^\top \x_j}  + \frac{1}{d}\sum_{i\in\mathcal{P}} \int d^s x_i\, e^{i(\k-\bm{q})^\top \x_i} \nonumber \\
&+ \frac{1}{d}\displaystyle\sum_{i\in\mathcal{P}} \displaystyle\sum_{j\in\mathcal{P}_{i,+}} \int \left(d^{s\text{-}o} x_{i\smallsetminus j}\right) e^{i\k^{(s-o)\top}_1 \x_{i\smallsetminus j}} \int \left(d^{o} x_{i\cup j}\right) e^{i(\k^{(o)}_{s-o+1}-\bm{q}^{(o)}_1)^\top \x_{i\cup j}}  \nonumber \\
&\times \int \left(d^{s\text{-}o} x_{j\smallsetminus i}\right) e^{i\bm{q}^{(s-o)\top}_{o+1} \x_{j\smallsetminus i}} \nonumber \\
& + \frac{1}{d}\displaystyle\sum_{i\in\mathcal{P}} \displaystyle\sum_{j\in\mathcal{P}_{i,-}} \left\lbrace  i\leftrightarrow j, \k \leftrightarrow \bm{q} \right\rbrace \nonumber \\
&=\frac{1}{d}\displaystyle\sum_{i\in\mathcal{P}} \delta(\k,\bm{0}) \displaystyle\sum_{j\notin\mathcal{P}_i} \delta(\bm{q},\bm{0}) + \frac{1}{d}\displaystyle\sum_{i\in\mathcal{P}} \delta(\k,\bm{q})\nonumber \\
&+\frac{1}{d}\displaystyle\sum_{i\in\mathcal{P}} \left( \displaystyle\sum_{j\in\mathcal{P}_{i,+}} \delta(\k^{(s-o)}_1,\bm{0}) \, \delta(\k^{(o)}_{s-o+1}, \bm{q}^{(o)}_1) \, \delta(\bm{q}^{(s-o)}_{o+1}, \bm{0}) \right. \nonumber \\ & \left. + \displaystyle\sum_{j\in\mathcal{P}_{i,-}} \delta(\bm{q}^{(s-o)}_1,\bm{0}) \, \delta(\k^{(o)}_1, \bm{q}^{(o)}_{s-o+1}) \, \delta(\k^{(s-o)}_{o+1}, \bm{0})\right),
\end{align}

with $\delta(\k,\bm{q})$ denoting the multidimensional Kronecker delta. For fixed $i$, the three terms on the RHS correspond to $j$'s such that $\x_j$ does not overlap with $\x_i$, to $j\,{=}\,i$ and to $j$'s such that $\x_j$ overlaps with $\x_i$, respectively. We recall that, in patch notation, $\k_1^{(s-o)}$ denotes the subsequence of $\k$ formed with the first $s-o$ components and $\k_{s-o+1}^{(o)}$ the subsequence formed with the last $o$ components.

By taking $\k$ and $\bm{q}$ in $\mathcal{F}^s$, as $k_1, k_s\neq 0$ and $q_1,q_s\neq 0$, \autoref{eq:loc-scalar-product-overlapping} implies
\begin{equation}
    \left\langle \Phi_{\k},\Phi_{\bm{q}}\right\rangle  = \delta(\k,\bm{q}).
\end{equation}
In addition, by taking $\k\in\mathcal{F}^s$ and $\bm{q}=\bm{q}^{(u)}_1\in\mathcal{F}^u$ with $u\,{<}\,s$,
\begin{equation}
    \left\langle \Phi_{\k},\Phi_{\bm{q}^{(u)}_1} \right\rangle  = 0\quad \forall\, u<s.
\end{equation}
Thus the $\Phi_{\k}$'s with $\k\in\mathcal{F}^s$ are orthonormal between each other and orthogonal to all $\Phi_{\bm{q}^{(u)}_1}$'s with $u\,{<}\,s$. Similarly, by taking $\k\in\mathcal{F}^u$ with $u\,{<}\,s$ and $\bm{q}\in\mathcal{F}^v$ with $v\,{\leq}\,u$, orthonormality is proven down to $\Phi_{\k^{(1)}_1}$. The zero-th eigenfunction $\Phi_{\bm{0}}(\x)\,{=}\,1$ is also orthogonal to all other eigenfunctions by~\autoref{eq:loc-scalar-product-overlapping} with $\k\,{=}\,0$ and trivially normalized to $1$.

Secondly, we prove that eigenfunctions from~\autoref{eq:loc-conv-eigvec-overlap} and eigenvalues from~\autoref{eq:loc-conv-eigval-overlap} satisfy the kernel eigenproblem of $\K^{CN}$. For $\k\,{=}\,\bm{0}$,
\begin{equation}
    \int_{[0,1]^d} d^dy\, \K^{CN}(\x,\y) = \frac{1}{d^2} \sum_{i,j=1}^d \int_{[0,1]^d} d^dy\, \sum_{\bm{q}}\lambda_{\k} e^{i\bm{q}^\top (\x_i-\y_j)} = \lambda_{\bm{0}},
\end{equation}
proving that $\Lambda_{\bm{0}}$ and $\Phi_{\bm{0}}$ satisfy the eigenproblem. For $\k\neq\bm{0}$,
\begin{equation}\begin{aligned}
    &\int_{[0,1]^d} d^dy\, \K^{CN}(\x,\y)\left(\frac{1}{\sqrt{d}} \sum_{l=1}^d e^{i\k^\top \y_l}\right) = \frac{1}{d^{5/2}} \sum_{i,j,l=1}^d \int_{[0,1]^d} d^dy\, \sum_{\bm{q}}\lambda_{\bm{q}} e^{i\bm{q}^\top (\x_i-\y_j)}e^{i\k^\top \y_l} \\
    &= \frac{1}{d^{5/2}} \sum_{i=1}^d \sum_{\bm{q}} \lambda_{\bm{q}}e^{i\bm{q}^\top \x_i}\sum_{j=1}^d \left( \delta(\k,\bm{q}) +  \displaystyle\sum_{l\in\mathcal{P}_{j,+}} \delta(\bm{q}^{(s-o)}_1,\bm{0}) \, \delta(\bm{q}^{(o)}_{s-o+1}, \k^{(o)}_1) \, \delta(\k^{(s-o)}_{o+1}, \bm{0}) \right. \\ &  \left.+  \displaystyle\sum_{l\in\mathcal{P}_{j,-}} \delta(\k^{(s-o)}_1,\bm{0}) \, \delta(\bm{q}^{(o)}_1, \k^{(o)}_{s-o+1}) \, \delta(\bm{q}^{(s-o)}_{o+1}, \bm{0}) \right).
\end{aligned}\end{equation}

When $\k\in\mathcal{F}^s$, the deltas coming from the terms with $j\in\mathcal{P}_{j,\pm}$ vanish, showing that the eigenproblem is satisfied with $\Lambda_{\k}\,{=}\,\lambda_{\k}/d$ and $\Phi_{\k}(\x)\,{=}\,\sum_l e^{i\k^\top \x}/\sqrt{d}$. When $\k\in\mathcal{F}^u$ with $u\,{<}\,s$, as the last $s\,{-}\,u$ components of $\k$ vanish, there are several $\bm{q}$'s satisfying the deltas in the bracket. There is $\bm{q}\,{=}\,\k$, from the $l\,{=}\,j$ term, then there are the $s\,{-}\,u$ $\bm{q}$'s such that $\delta(\bm{q}^{(s-o)}_1,\bm{0}) \delta(\bm{q}^{(o)}_{s-o+1}, \k^{(o)}_1) \delta(\k^{(s-o)}_{o+1}, \bm{0}) \,{=}\,1$. These are all the $\bm{q}$'s having a $u$-dimensional patch equal to $\k_1^{(u)}$ and all the other elements set to zero, thus there are $(s-u+1)$ such $\bm{q}$'s. Moreover, as $\lambda_{\bm{q}}$ depends only on the modulus of $\bm{q}$, all these $\bm{q}$'s result in the same eigenvalue, and in the same eigenfunction $\sum_l e^{i\bm{q}^\top \x}/\sqrt{d}$, after the sum over patches. Therefore, 
\begin{equation}
    \int_{[0,1]^d} d^dy\, \K^{CN}(\x,\y) \Phi_{\k_1^{(u)}} = \frac{(s-u + 1)}{d}\lambda_{\k_1^{(u)}}\Phi_{\k_1^{(u)}} = \Lambda_{\k_1^{(u)}}\Phi_{\k_1^{(u)}}.
\end{equation}

Finally, we prove the expansion of the kernel in~\autoref{eq:loc-conv-decomp-overlap},
\begin{align}\label{eq:loc-decomp-proof}
    \K^{CN}(\x, \y) &= \frac{1}{d^2} \sum_{i,j\in\mathcal{P}} \mathcal{C}(\x_i, \y_j) \\
    &= \sum_{\k} \frac{1}{d}\lambda_{\k} \left(\frac{1}{\sqrt{d}}\sum_{i\in\mathcal{P}}\phi_{\k}(\x_i)\right) \overline{\left(\frac{1}{\sqrt{d}}\sum_{j\in\mathcal{P}}\phi_{\k}(\y_j)\right)}.
\end{align}
The terms on the RHS of~\autoref{eq:loc-decomp-proof} are trivially equal to those of~\autoref{eq:loc-conv-decomp-overlap} for $\k\in\mathcal{F}^s$. All the $\k$ having $s\,{-}\,u$ vanishing extremal components can be written as shifts of $\k_1^{(u)}\in\mathcal{F}^{u}$, which has the \emph{last} $s\,{-}\,u$ components vanishing. But a shift of $\k$ does not affect $\lambda_{\k}$ nor $\sum_l e^{i\k^\top \x}$, leading to a degeneracy of eigenvalues having $\k$ which can be obtained from a shift of $\k_1^{(u)}\in\mathcal{F}^u$. Such degeneracy is removed by restricting the sum over $\k$ to the sets $\mathcal{F}^u$, $u\,{\leq}\,s$, of wavevectors with non-vanishing extremal components, and rescaling the remaining eigenvalues with a factor of $(s-u+1)/d$, so that~\autoref{eq:loc-conv-decomp-overlap} is obtained.
\end{proof}

\begin{lemma}[Spectra of overlapping local kernels]\label{lemma:loc-spectra-overlap}
Let $\K^{LC}$ be a local kernel defined as in~\autoref{eq:loc-loc-ker}, with $\mathcal{P}\,{=}\,\left\lbrace1,\dots, d\right\rbrace$ and constituent kernel $C$ satisfying assumptions $i)$, $ii)$ above. Then, $\K^{LC}$ admits the following Mercer's decomposition,
\begin{equation}\label{eq:loc-loc-decomp-overlap}
   \K^{LC}(\x,\y) = \Lambda_{\bm{0}} +  \displaystyle\sum_{u=1}^{s}\left( \displaystyle\sum_{ \k\in \mathcal{F}^{u}}\displaystyle\sum_{i=1}^d \Lambda_{\k,i} \Phi_{\k,i}(\x) \Phi_{\k,i}(\y)\right)
\end{equation}
with eigenfunctions
\begin{equation}\label{eq:loc-loc-eigvec-overlap}
 \Phi_{\bm{0}}(\x)\,{=}\,1, \quad \Phi_{\k,i}(\x)\,{=}\,\phi_{\k}(\x_i) \quad \forall\, \k\in\mathcal{F}^u \text{ with }1\leq u \leq s \text{ and }i=1,\dots,d,
\end{equation}
and eigenvalues
\begin{equation}\label{eq:loc-loc-eigval-overlap}
 \Lambda_{\bm{0}}\,{=}\,\lambda_{\bm{0}}, \Lambda_{\k,i} = \frac{s-u+1}{d}\lambda_{\k} \quad \forall\, \k\in\mathcal{F}^{u} \text{ with } u\leq s\text{ and }i=1,\dots,d.
\end{equation}
\end{lemma}
\begin{proof} We start by proving the orthonormality of the eigenfunctions. The scalar product $\left\langle\Phi_{\k,i}, \Phi_{\bm{q},j} \right\rangle$ depends on the relation between the $i$-th and $j$-th patches.

\begin{subequations}\label{eq:loc-scalar-product-overlapping-loc}
\begin{align}
&\int_{[0,1]^d} d^d x\,  \phi_{\k}(\x_i) \overline{ \phi_{\bm{q}}(\x_j)} & \nonumber \\
\label{eq:loc-scalar-product-overlapping-loc-pplus} &=\delta(\k^{(s-o)}_1,\bm{0}) \, \delta(\k^{(o)}_{s-o+1}, \bm{q}^{(o)}_1) \, \delta(\bm{q}^{(s-o)}_{o+1}, \bm{0}), &\text{ if } j\in\mathcal{P}_{i,+},\\ \label{eq:loc-scalar-product-overlapping-loc-pminus}
& =\delta(\bm{q}^{(s-o)}_1,\bm{0}) \, \delta(\k^{(o)}_1, \bm{q}^{(o)}_{s-o+1}) \, \delta(\k^{(s-o)}_{o+1}, \bm{0}), & \text{ if } j\in\mathcal{P}_{i,-},\\
\label{eq:loc-scalar-product-overlapping-loc-zero}& =\delta(\k,\bm{0})\,\delta(\bm{q},\bm{0}), & \text{ if } j\notin\mathcal{P}_{i}, \\ \label{eq:loc-scalar-product-overlapping-loc-equal}
& = \delta(\k,\bm{q}), &\text{ if } j=i.
\end{align}
\end{subequations}

Clearly, $\left\langle \Phi_{\bm{0}}, \Phi_{\bm{0}}\right\rangle\,{=}\,1$ and
setting one of $\bm{q}$ and $\k$ to $\bm{0}$ in~\autoref{eq:loc-scalar-product-overlapping-loc} yields orthogonality between $\Phi_{\bm{0}}$ and $\Phi_{\k, i}$ for all $\k\neq \bm{0}$ and $i\,{=}\,1,\dots,d$. For any $\k$ and $\bm{q}\neq 0$, \autoref{eq:loc-scalar-product-overlapping-loc-equal} implies
\begin{equation}
 \left\langle\Phi_{\k,i}, \Phi_{\bm{q},j} \right\rangle = \delta(\k,\bm{q})\delta_{i,j}
\end{equation}
unless $\k\,{=}\,\k^{(u)}_1\in\mathcal{F}^u$ and $\bm{q}$ is a shift of $\k^{(u)}$. But such a $\bm{q}$ would have $q_1\,{=}\,0$ and there is no eigenfunction $\Phi_{\bm{q}}$ with $q_1\,{=}\,0$, apart from $\Phi_{\bm{0}}$. Hence, orthonormality is proven.

We then prove that eigenfunctions and eigenvalues defined in~\autoref{eq:loc-loc-eigvec-overlap} and~\autoref{eq:loc-loc-eigval-overlap} satisfy the kernel eigenproblem of $\K^{LC}$. For $\k\,{=}\,\bm{0}$,
\begin{equation}
    \int_{[0,1]^d} d^dy\, \K^{LC}(\x,\y) = \frac{1}{d} \sum_{i=1}^d \int_{[0,1]^d} d^dy\, \sum_{\bm{q}}\lambda_{\k} e^{i\bm{q}^\top (\x_i-\y_i)} = \lambda_{\bm{0}}.
\end{equation}
In general,
\begin{equation}\begin{aligned}
    &\int_{[0,1]^d} d^dy\, \K^{LC}(\x,\y)e^{i\k^\top \y_l} = \frac{1}{d} \sum_{i=1}^d \int_{[0,1]^d} d^dy\, \sum_{\bm{q}}\lambda_{\bm{q}} e^{i\bm{q}^\top (\x_i-\y_i)} e^{i\k^\top \bm{y_l}} \\
    &= \frac{1}{d}\sum_{\bm{q}} \lambda_{\bm{q}} \left( \delta(\k,\bm{q}) e^{i\k^\top \x_l} + \sum_{i\notin\mathcal{P}_l}\delta(\bm{q},0) \, \delta(\k,0) \right.\\ &+ \left. \sum_{i\in\mathcal{P}_{l,+}} e^{i\bm{q}^\top \x_i}\delta(\k^{(s-o)}_1,\bm{0}) \, \delta(\k^{(o)}_{s-o+1}, \bm{q}^{(o)}_1) \, \delta(\bm{q}^{(s-o)}_{o+1}, \bm{0}) \right. \\  &+ \left.\sum_{i\in\mathcal{P}_{l,-}} e^{i\bm{q}^\top \x_i}\delta(\bm{q}^{(s-o)}_1,\bm{0}) \, \delta(\k^{(o)}_1, \bm{q}^{(o)}_{s-o+1}) \, \delta(\k^{(s-o)}_{o+1}, \bm{0}) \right).
\end{aligned}\end{equation}

For $\k\,\in\,\mathcal{F}^u$, with $u=1,\dots,s$, the deltas which set the first component of $\k$ to $0$ are never satisfied, therefore
\begin{equation}\label{eq:loc-eigval-loc-comp}\begin{aligned}
    &\int_{[0,1]^d} d^dy\, \K^{LC}(\x,\y)e^{i\k^\top \y_l} \\
    &= \frac{1}{d}\sum_{\bm{q}} \lambda_{\bm{q}} \left( \delta(\k,\bm{q}) e^{i\k^\top \x_l} + \sum_{i\in\mathcal{P}_{l,-}} e^{i\bm{q}^\top \x_i}\delta(\bm{q}^{(s-o)}_1,\bm{0}) \, \delta(\k^{(o)}_1, \bm{q}^{(o)}_{s-o+1}) \, \delta(\k^{(s-o)}_{o+1}, \bm{0}) \right).
\end{aligned}\end{equation}
The second term in brackets vanishes for $\k\,\in\,\mathcal{F}^s$ and the eigenvalue equation is satisfied with $\lambda_{\k, l} =\lambda_{\k}/d$. For $\k=\k^{(u)}_1\,\in\,\mathcal{F}^u$ with $u\,{<}\,s$, $\delta(\k^{(s-o)}_{o+1}, \bm{0})\,{=}\,1$ for any $o\,{\geq}\,u$. As a result of the remaining deltas, the RHS of~\autoref{eq:loc-eigval-loc-comp} becomes a sum over all $\bm{q}$'s which can be obtained from shifts of $\k^{(u)}_1$, which are $s\,{-}\,u\,{+}\,1$ (including $\k^{(u)}_1$ itself). The patch $\x_i$ which is multiplied by $\bm{q}$ in the exponent is also a shift of $\x_l$, thus all the factors $e^{i\bm{q}^\top \x_i}$ appearing in the sum coincide with $e^{i\k^{(u)\top}_1 \x_i}$. As $\lambda_{\bm{q}}$ depends on the modulus of $\bm{q}$, all these terms correspond to the same eigenvalue, $\lambda_{\k^{(u)}_1}$, so that 
\begin{equation}
    \int_{[0,1]^d} d^dy\, \K^{LC}(\x,\y)e^{i\k^{(u)\top}_1 \y_l} = \left(\frac{s-u+1}{d} \lambda_{\k^{(u)}_1}\right) e^{i\k^{(u)\top}_1 \x_l}.
\end{equation}
Finally, we prove the expansion of the kernel in~\autoref{eq:loc-loc-decomp-overlap},
\begin{align}\label{eq:loc-loc-decomp-proof}
    \K^{LC}(\x, \y) &= \frac{1}{d} \sum_{i\in\mathcal{P}} \mathcal{C}(\x_i, \y_i) = \sum_{\k} \frac{1}{d}\lambda_{\k}
    \sum_{i\in\mathcal{P}} \phi_{\k}(\x_i) \overline{\phi_{\k}(\y_i)}.
\end{align}
As in the proof of the eigendecomposition of convolutional kernels, all the $\k$ having $s\,{-}\,u$ vanishing extremal components can be written as shifts of $\k_1^{(u)}\in\mathcal{F}^{u}$, which has the \emph{last} $s\,{-}\,u$ components vanishing. The shift of $\k$ does not affect $\lambda_{\k}$ nor the product $\phi_{\k}(\x_i) \overline{\phi_{\k}(\y_i)}$, after summing over $i$ leading to a degeneracy of eigenvalues which is removed by restricting the sum over $\k$ to the sets $\mathcal{F}^u$, $u\,{\leq}\,s$, and rescaling the remaining eigenvalues $\lambda_{\k_1^{(u)}}$ with a factor of $(s-u+1)/d$, leading to~\autoref{eq:loc-loc-decomp-overlap}.
\end{proof}

\section{Proof of Theorem \ref{th:scaling}}\label{app:loc-thm1}

\begin{theorem}[Theorem \ref{th:scaling} in the main text]
Let $\K_T$ be a $d$-dimensional convolutional kernel with a translationally-invariant $t$-dimensional constituent and leading nonanalyticity at the origin controlled by the exponent $\alpha_t\,{>}\,0$. Let $\K_S$ be a $d$-dimensional convolutional or local student kernel with a translationally-invariant $s$-dimensional constituent, and with a nonanalyticity at the origin controlled by the exponent $\alpha_s\,{>}\,0$.
Assume, in addition, that if the kernels have overlapping patches then $s\geq t$; whereas if the kernels have nonoverlapping patches $s$ is an integer multiple of $t$; and that data are uniformly distributed on a $d$-dimensional torus. Then, the following asymptotic equivalence holds in the limit $P\to\infty$,
\begin{equation}
    \mathcal{B}(P) \sim P^{-\beta}, \quad \beta = \alpha_t / s.
\end{equation}
\end{theorem}

\begin{proof}

For the sake of clarity, we start with the proof in the nonoverlapping-patches case, and then extend it to the overlapping-patches case. Since $\K_T$ and $\K_S$ have translationally-invariant constituent kernels and data are uniformly distributed on a $d$-dimensional torus, the kernels can be diagonalized in Fourier space. Let us start by considering a convolutional student: because of the constituent kernel's isotropy, the Fourier coefficients $\Lambda^{(s)}_{\k}$ of $\K_S$ depend on $k$ (modulus of $\k$) only. Notice the superscript indicating the dimensionality of the student constituents. In particular, $\Lambda^{(s)}_{\k}$ is a decreasing function of $k$ and, for large $k$, $\Lambda_{\k}\sim k^{-(s+\alpha_s)}$.  Then, $\mathcal{B}(P)$ reads
\begin{equation}\label{eq:loc-Bp}
 \mathcal{B}(P) = \sum_{\left\lbrace \k| k > k_c(P) \right\rbrace} \mathbb{E}[|c_{\k}|^2],
\end{equation}
where $k_c(P)$ is defined as the wavevector modulus of the $P$-th largest eigenvalue and $\mathbb{E}[|c_{\k}|^2]$ denotes the variance of the target coefficients in the student eigenbasis. $k_c(P)$ is such that there are exactly $P$ eigenvalues with $k\,{\leq}\,k_c(P)$,
\begin{equation}\label{eq:loc-radius}
    P = \sum_{\left\lbrace \k|k<k_c(P)\right\rbrace} 1 \sim \int \frac{d^s k}{(2\pi)^s} \theta(k_c(P)-k) = \frac{1}{(2\pi)^s} \frac{\pi^{s/2}}{\Gamma(s/2 + 1)} k_c(P)^{s},
\end{equation}
i.e., $k_c(P)\sim P^{1/s}$.

By denoting the eigenfunctions of the student with $\Phi_{\k}^{(s)}$, the superscript $(s)$ indicating the dimension of the constituent's plane waves,
\begin{align}\label{eq:loc-convconv}
  \mathbb{E}[|c_{\k}|^2] &= \int_{[0,1]^d} d^dx \, \Phi_{\k}^{(s)}(\x)\int_{[0,1]^d} d^dy \, \overline{\Phi_{\k}^{(s)}(\y)} \mathbb{E}[f^*(\x)f^*(\y)]\\ & =\int_{[0,1]^d} d^dx \, \Phi_{\k}^{(s)}(\x)\int_{[0,1]^d} d^dy \, \overline{\Phi_{\k}^{(s)}(\y)} \K_T(\x, \y). \nonumber
\end{align}
Decomposing the teacher kernel $\K_T$ into its eigenvalues $\Lambda_{\bm{q}}^{(t)}$ and eigenfunctions $\Phi_{\bm{q}}^{(t)}(\y)$,
\begin{align}
 \mathbb{E}[|c_{\k}|^2] &= \int_{[0,1]^d} d^dx \, \Phi^{(s)}_{\k}(\x)\int_{[0,1]^d} d^dy \, \overline{\Phi^{(s)}_{\k}(\y)} \Biggl(\Lambda^{(t)}_{\bm{0}} \\ &+ \frac{s}{d} \sum_{\bm{q}\neq\bm{0}} \Lambda^{(t)}_{\bm{q}} \sum_{i\in\mathcal{P}^{(t)}} \phi^{(t)}_{\bm{q}}(\x_i) \sum_{j\in\mathcal{P}^{(t)}} \overline{\phi^{(t)}_{\bm{q}}(\y_j)}\Biggr). \nonumber
\end{align}
The $\bm{q}\eq\bm{0}$ mode of the teacher can give non-vanishing contributions to $c_{\bm{0}}$ only, therefore it does not enter any term of the sum in \autoref{eq:loc-Bp}. Once we removed the  term with $\bm{q}\eq\bm{0}$, consider the  $\y$-integral:
\begin{align}\label{eq:loc-Ik}
    \mathcal{I}_{\k}(\x) &= \int_{[0,1]^d} d^dy \, \sqrt{\frac{s}{d}} \sum_{m\in\mathcal{P}^{(s)}}  \overline{\phi^{(s)}_{\k}(\y_m)} \frac{s}{d} \sum_{\bm{q}\neq\bm{0}} \Lambda^{(t)}_{\bm{q}} \sum_{i\in\mathcal{P}^{(t)}} \phi^{(t)}_{\bm{q}}(\x_i) \sum_{j\in\mathcal{P}^{(t)}} \overline{\phi^{(t)}_{\bm{q}}(\y_j)} \\
    &= \left(\frac{s}{d}\right)^{\frac{3}{2}} \sum_{\bm{q}\neq\bm{0}} \Lambda^{(t)}_{\bm{q}} \sum_{i\in\mathcal{P}^{(t)}} \phi^{(t)}_{\bm{q}}(\x_i) \sum_{m\in\mathcal{P}^{(s)}} \sum_{j\in\mathcal{P}^{(t)}} \int_{[0,1]^d} d^dy\, \overline{\phi^{(s)}_{\k}(\y_m)} \,    \overline{\phi^{(t)}_{\bm{q}}(\y_j)}. \nonumber
\end{align}
As all the $t$-dimensional patches of the teacher must be contained in at least one of the $s$-dimensional patches of the student, in the nonoverlapping case we require that $s$ is an integer multiple of $t$. Then, each of the teacher patches is entirely contained in one and only one patch of the student. If the teacher patch $\y_j$ is not contained in the student patch $\y_m$, we can factorize the integration over $\y$ into two integrals over $\y_j$ and $\y_m$. These terms give vanishing contributions to $\mathcal{I}_{\k}(\x)$ since the integral of a plane wave over a period is always zero for non-zero wavevectors. Instead, if the teacher patch $\y_j$ is contained in the student patch $\y_m$, denoting with $l$ the index of the element of $\y_m$ which coincide with the first element of $\y_j$, we can factorize the student eigenfunctions as follows
\begin{equation}\label{eq:loc-fact}
 \phi^{(s)}_{\k}(\y_m) = \phi^{(t)}_{\k^{(t)}_l}(\y_j) \phi^{(s-t)}_{\k\smallsetminus\k^{(t)}_l}(\y_{m\smallsetminus j}).
\end{equation}
Here $\k^{(t)}_l$ denotes the $t$-dimensional patch of $\k$ starting at $l$ and $\k\smallsetminus\k^{(t)}_l$ the sequence of elements which are in $\k$ but not in $\k^{(t)}_l$. As $s$ is an integer multiple of $t$, $l\,{=}\,\tilde{l} \times s/t$ with $\tilde{l}=1,\dots,t$. Inserting \autoref{eq:loc-fact} into \autoref{eq:loc-Ik},
\begin{equation}
 \mathcal{I}_{\k}(\x) = \sum_{l=\tilde{l}s/t,\, \tilde{l}=1}^t \delta(\k\smallsetminus\k^{(t)}_l,\bm{0}) \, \Lambda_{\k^{(t)}_l}^{(t)} \sqrt{\frac{s}{d}} \sum_{i\in\mathcal{P}^{(t)}} \overline{\phi_{\k^{(t)}_l}^{(t)}(\x_i)}.
\end{equation}
The $\x$-integral of \autoref{eq:loc-convconv} can be performed by the same means after expanding $\Phi^{(s)}_{\k}$ as a sum of $s$-dimensional plane waves, so as to get,
\begin{equation}\label{eq:loc-ck2}
 \mathbb{E}[|c_{\k}|^2] = \sum_{l=\tilde{l}s/t,\, \tilde{l}=1}^t\delta(\k\smallsetminus\k^{(t)}_l,\bm{0}) \, \Lambda_{\k^{(t)}_l}^{(t)}.
\end{equation}
Therefore, $\mathbb{E}[|c_{\k}|^2]$ is non-zero only for $\k$'s which have at most $t$ consecutive components greater or equal than zero, and the remaining $s\,{-}\,t$ being strictly zero. Inserting \autoref{eq:loc-ck2} into \autoref{eq:loc-Bp},
\begin{equation}\label{eq:loc-final}
 \mathcal{B}(P) = \sum_{\left\lbrace \k| k > k_c(P)\right\rbrace} \sum_{l=\tilde{l}s/t,\, \tilde{l}=1}^t \delta(\k\smallsetminus\k^{(t)}_l,\bm{0}) \, \Lambda_{\k^{(t)}_l}^{(t)} \sim \int_{P^{1/s}}^\infty dk k^{t-1} k^{-(\alpha_t + t)} \sim P^{-\frac{\alpha_t}{s}}.
\end{equation}
When using a local student, the convolutional eigenfunctions in the RHS of~\autoref{eq:loc-convconv} are replaced by the local eigenfunctions $\Phi_{\k, i}(\x)$ of~\autoref{eq:loc-loc-decomp}.
Repeating the same computations, one finds 
\begin{equation}
    k_c \sim  \left(\frac{P}{d/s}\right)^{\frac{1}{s}},
\end{equation}
\begin{equation}\label{eq:loc-final2}
 \mathbb{E}[|c_{\k,i}|^2] = \frac{s}{d} \sum_{l=\tilde{l}s/t,\, \tilde{l}=1}^t\delta(\k\smallsetminus\k^{(t)}_l,\bm{0}) \, \Lambda_{\k^{(t)}_l}^{(t)}.
\end{equation}
As a result,
\begin{align}
 \mathcal{B}(P) &= \sum_{i \in \mathcal{P}} \sum_{\left\lbrace \k| k > k_c(P)\right\rbrace} \frac{s}{d} \sum_{l=\tilde{l}s/t,\, \tilde{l}=1}^t \delta(\k\smallsetminus\k^{(t)}_l,\bm{0}) \, \Lambda_{\k^{(t)}_l}^{(t)} \\ 
 &\sim \int_{\left(\frac{P}{d/s}\right)^{\frac{1}{s}}}^\infty dk k^{t-1} k^{-(\alpha_t + t)} \sim \left(\frac{P}{d/s}\right)^{-\frac{\alpha_t}{s}}.
\end{align}

As we showed in \autoref{app:loc-mercer-overlap}, when the patches overlap the set of wavevectors which index the eigenvalues is restricted from $\mathbb{Z}^s$ to the union of the $\mathcal{F}^u$'s for $u\,{=}\,0,\dots,s$. In addition, the eigenvalues with $\k \in \mathcal{F}^{u}$, $0\,{<}\,u\,{<}\,s$, are rescaled by a factor $(s-u+1)/d$. Therefore, in the overlapping case the eigenvalues do not decrease monotonically with $k$ and $\mathcal{B}(P)$ cannot be written as a sum of over $\k$'s with modulus $k$ larger than a certain threshold $k_c$. By considering also that, with $t\,{\leq}\,s$, $\mathbb{E}[|c_{\k}|^2]$ is non-zero only for $\k$'s which have at most $t$ consecutive nonvanishing components, we have
\begin{equation}\label{eq:loc-Bp-overlap}
 \mathcal{B}(P) = \sum_{u=0}^t \sum_{\k\in\mathcal{F}^u}\mathbb{E}[|c_{\k}|^2]\chi(\Lambda^{(s)}_{\k}\,{>}\,\Lambda_P),
\end{equation}
where $\Lambda_P$ denotes the $P$-th largest eigenvalue and the indicator function $\chi(\Lambda^{(s)}_{\k}\,{>}\,\Lambda_P)$ ensures that the sum runs over all but the first $P$ eigenvalues of the student. The sets $\{\mathcal{F}^{u}\}_{u<t}$ have all null measure in $\mathbb{Z}^t$, whereas $\mathcal{F}^t$ is dense in $\mathbb{Z}^t$, thus the asymptotics of $\mathcal{B}(P)$ are dictated by the sum over $\mathcal{F}^t$. When $\k$'s are restricted to the latter set, eigenvalues are again decreasing functions of $k$ and the constraint $\Lambda^{(s)}_{\k}\,{>}\,\Lambda_P$ translates into $k\,{>}\,k_c(P)$. Having changed, with respect to the nonoverlapping case, only an infinitesimal fraction of the eigenvalues, the asymptotic scaling of $k_c(P)$ with $P$ remains unaltered and the estimates of~\autoref{eq:loc-final} and~\autoref{eq:loc-final2} extend to kernels with nonoverlapping patches after substituting the degeneracy $d/s$ with $|\mathcal{P}|=d$.

\end{proof}

\section{Asymptotic learning curves with a local teacher}\label{app:loc-local}

\begin{theorem} Let $\K_T$ be a $d$-dimensional local kernel with a translationally-invariant $t$-dimensional constituent and leading nonanalyticity at the origin controlled by the exponent $\alpha_t\,{>}\,0$. Let $\K_S$ be a $d$-dimensional local student kernel with a translationally-invariant $s$-dimensional constituent, and with a nonanalyticity at the origin controlled by the exponent $\alpha_s\,{>}\,0$.
Assume, in addition, that if the kernels have overlapping patches then $s\geq t$; whereas if the kernels have nonoverlapping patches $s$ is an integer multiple of $t$; and that data are uniformly distributed on a $d$-dimensional torus. Then, the following asymptotic equivalence holds in the limit $P\to\infty$,
\begin{equation}
    \mathcal{B}(P) \sim P^{-\beta}, \quad \beta = \alpha_t / s.
\end{equation}
\end{theorem}

\begin{proof}

The proof is analogous to that of~\autoref{app:loc-thm1}, the only difference being that eigenfunctions and eigenvalues are indexed by $\k$ and the patch index $i$. This results in an additional factor of $d/s$ in the RHS of~\autoref{eq:loc-radius}. All the discussion between~\autoref{eq:loc-convconv} and~\autoref{eq:loc-ck2} can be repeated by attaching the additional patch index $i$ to all coefficients. ~\autoref{eq:loc-final} for $\mathcal{B}(P)$ is then corrected with an additional sum over patches. The extra sum, however, does not influence the asymptotic scaling with $P$.

\end{proof}

\section{Proof of Theorem \ref{th:ridge}}\label{app:loc-thm2}

\begin{theorem}[Theorem \ref{th:ridge} in the main text]
Let us consider a positive-definite kernel $K$ with eigenvalues $\Lambda_\rho$, $\sum_\rho \Lambda_\rho < \infty$, and eigenfunctions ${\Phi_\rho}$ learning a (random) target function $f^*$ in kernel ridge regression (\autoref{eq:loc-argmin}) with ridge $\lambda$ from $P$ observations $f^*_{\nu}\,{=}\,f^*(\x_\nu)$, with $\x_\nu\in \mathbb{R}^d$ drawn from a certain probability distribution. Let us denote with $\mathcal{D}_T(\Lambda)$ the reduced density of kernel eigenvalues with respect to the target and $\testerr(\lambda,P)$ the generalization error and also assume that
\begin{itemize}
    \item[$i)$] For any $P$-tuple of indices $\rho_1,\dots,\rho_P$, the vector $(\Phi_{\rho_1}(\x_1), \dots,\Phi_{\rho_P}(\x_P))$ is a Gaussian random vector;
    \item[$ii)$] The target function can be written in the kernel eigenbasis with coefficients $c_\rho$ and $c^2(\Lambda_\rho)\,{=}\,\mathbb{E}[|c_\rho|^2]$, with $\mathcal{D}_T(\Lambda) \sim \Lambda^{-(1+r)}$, $c^2(\Lambda) \sim \Lambda^{q}$ asymptotically for small $\Lambda$ and $r\,{>}\,0$, $r\,{<}\,q\,{<}\,r\,{+}\,2$;
\end{itemize}
Then the following equivalence holds in the joint $P\to\infty$ and $\lambda\to 0$ limit with $1/(\lambda\sqrt{P})\to 0$:
\begin{equation}\label{eq:loc-thm-ridge-result-app}
    \testerr(\lambda, P) \sim  \sum_{\left\lbrace \rho|\Lambda_\rho < \lambda \right\rbrace} \mathbb{E}{[|c_\rho|^2]} = \int_0^{\lambda} d\Lambda \mathcal{D}_T(\Lambda) c^2(\Lambda).
\end{equation}
\end{theorem}

\begin{proof}

In this proof, we make use of results derived in~\cite{jacot2020kernel}. Our setup for kernel ridge regression corresponds to what the authors of~\cite{jacot2020kernel} call the \emph{classical setting}. Let us introduce the integral operator $T_{\K}$ associated with the kernel, defined by
\begin{equation}
  (T_{\K} f)(\x) = \int p\left( d^d y\right) \K(\x,\y) f(\y). 
\end{equation}
The trace $Tr[T_{\K}]$ coincides with the sum of $\K$'s eigenvalues and is finite by hypothesis. We define the following estimator of the generalization error $\testerr(\lambda, P)$,
\begin{equation}
    \mathcal{R}(\lambda, P) = \partial_\lambda \vartheta(\lambda) \int p(d^dx)\, \left( f^*(\x) - (\mathcal{A}_{\vartheta}f^*)(\x) \right)^2,
\end{equation}
where $\vartheta(\lambda)$ is the \emph{signal capture threshold} (SCT)~\cite{jacot2020kernel} and $\mathcal{A}_{\vartheta}\,{=}\, T_{\K} (T_{\K} + \vartheta(\lambda))^{-1}$ is a reconstruction operator~\cite{jacot2020kernel}. The target function can be written in the kernel eigenbasis by hypothesis (with coefficients $c_\rho$) and $T_{\K}$ has the same eigenvalues and eigenfunctions of the kernel by definition. Hence,
\begin{align}
    \mathcal{R}(\lambda, P) &= \partial_\lambda \vartheta(\lambda)\sum_{\rho=1}^{\infty} \frac{\vartheta(\lambda)^2}{(\Lambda_\rho + \vartheta(\lambda))^2} |c_\rho|^2 \\ &= \partial_\lambda \vartheta(\lambda) \int_0^{\infty}d\Lambda\, \mathcal{D}_T(\Lambda) c^2(\Lambda) \frac{\vartheta(\lambda)^2}{(\Lambda + \vartheta(\lambda))^2}, \nonumber
\end{align}
where $\mathcal{D}_T$ is the reduced density of eigenvalues~\autoref{eq:loc-reduced-density}. We now derive the asymptotics of $\mathcal{R}(\lambda, P)$ in the joint $P\to\infty$ and $\lambda\to0$ limit, then relate the asymptotics of $\mathcal{R}$ to those of $\testerr(\lambda,P)$ via a theorem proven in~\cite{jacot2020kernel}.

Proposition 3 of~\cite{jacot2020kernel} shows that for any $\lambda\,{>}\,0$, $\partial_\lambda \vartheta(\lambda)\to 1$ and $\vartheta(\lambda)\to \lambda$ with corrections of order $1/N$. Thus, we focus on the following integral,
\begin{equation}\label{eq:loc-integral1}
    \int_0^{\infty}d\Lambda\, \mathcal{D}_T(\Lambda) c^2(\Lambda) \frac{\lambda^2}{(\Lambda + \lambda)^2}.
\end{equation}
The functions $\mathcal{D}_T(\Lambda)$ and $c^2(\Lambda)$ can be safely replaced with their small-$\Lambda$ expansions $\Lambda^{-(1+r)}$ and $\Lambda^{q}$ over the whole range of the integral above because of the assumptions $q\,{>}\,r$ and $q\,{\leq}\,r+2$. In practice, there should be an upper cut-off on the integral coinciding with the largest eigenvalue $\Lambda_1$, but the assumption $q\,{\leq}\,r+2$ causes this part of the spectrum to be irrelevant for the asymptotics of the error. In fact, we will conclude that the integral is dominated by the portion of the domain around $\lambda$. After the change of variables $y\,{=}\,\Lambda/\lambda$,
\begin{equation}\label{eq:loc-integral2}
    \int_0^{\infty} d\Lambda\,\mathcal{D}_T(\Lambda) c^2(\Lambda) \frac{\lambda^2}{(\Lambda + \lambda)^2} = \lambda^{q-r} \int dy\, \frac{y^{q-1-r}}{(1+y)^2},
\end{equation}
where one recognizes one of the integral representations of the beta function,
\begin{equation}
    B(a, b) = \int dy\, \frac{y^{a-1}}{(1+y)^{a+b}} = \frac{\Gamma(a)\Gamma(b)}{\Gamma(a+b)},
\end{equation}
with $\Gamma$ denoting the gamma function. Therefore, 
\begin{equation}\label{eq:loc-integral3}
    \int_0^{\infty} d\Lambda\,\mathcal{D}_T(\Lambda) c^2(\Lambda) \frac{\lambda^2}{(\Lambda + \lambda)^2} = \lambda^{q-r} \frac{\Gamma(q-r)\Gamma(2-q+r)}{\Gamma(2)}.
\end{equation}
It is interesting to notice how the assumptions $q\,{>}\,r$ and $q\,{<}\,r\,{+}\,2$ are required in order to avoid the poles of the $\Gamma$ functions in the RHS of~\autoref{eq:loc-integral3}. 

We now use~\autoref{eq:loc-integral3} to infer the asymptotics of $\mathcal{R}(\lambda, P)$ in the scaling limit $\lambda\to0$ and $P\to\infty$ with $1/(\lambda\sqrt{P})\to0$. The latter condition implies that $\lambda$ decays more slowly than $(P)^{-1/2}$, thus additional terms stemming from the finite-$P$ difference between $\vartheta $ and $\lambda$, of order $P^{-1}$ are negligible w.r.t. $\lambda^{q-r}$. The finite-$P$ difference between $\partial_\lambda\vartheta$, also $O(P^{-1})$, can be neglected too. Finally, 
\begin{equation}\label{eq:loc-integral4}
    \mathcal{R}(\lambda, P) \sim \int_0^{\infty} d\Lambda\,\mathcal{D}_T(\Lambda) c^2(\Lambda) \frac{\lambda^2}{(\Lambda + \lambda)^2}\sim \lambda^{q-r} \sim \int_0^\lambda d\Lambda \mathcal{D}_T(\Lambda) c^2(\Lambda).
\end{equation}

Theorem 6 of~\cite{jacot2020kernel} shows the convergence of $\testerr(\lambda, P)$ towards $\mathcal{R}(\lambda, P)$ when $P\to\infty$. Specifically,
\begin{equation}
    | \testerr(\lambda,P) - \mathcal{R}(\lambda, P)| \leq \left(\frac{1}{P} + g\left(\frac{Tr[T_{\K}]}{\lambda\sqrt{P}}\right)\right) \mathcal{R}(\lambda, P),
\end{equation}
where $g$ is a polynomial with non-negative coefficients and $g(0)\,{=}\,0$. With a decaying ridge $\lambda(P)$ such that $1/(\lambda\sqrt{P})\to 0$, both $\mathcal{R}/P$ and $\mathcal{R}g(Tr[T_{\K}]/(\lambda\sqrt{P}))$ tend to zero faster than $\mathcal{R}$ itself, thus the asymptotics of $\testerr(\lambda,P)$ coincide with those of $\mathcal{R}(\lambda, P)$ and~\autoref{eq:loc-thm-ridge-result-app} is proven.
\end{proof}

\paragraph{Remark} The estimate for the exponent $\beta$ of Corollary~\autoref{cor:beta-rigorous} follows from the theorem above with $r\,{=}\,t/(s+\alpha_s)$, $q\,{=}\,(\alpha_t + t)/(\alpha_s + s)$ and $\lambda\sim P^{-\gamma}$. Then $q\,{>}\,r$ because $\alpha_t\,{>}\,0$, whereas the condition $q\,{<}\,r + 2$ is equivalent to the assumption $\alpha_t \,{<}\,2(\alpha_s + s)$ required in~\autoref{sec:loc-learning-curves} in order to derive the learning curve exponent in~\autoref{eq:loc-prediction} from our estimate of $\mathcal{B}(P)$.

\section{Numerical experiments}\label{app:loc-numerics}
 
\subsection{Details on the simulations}

To obtain the empirical learning curves, we generate $P+P_{\text{test}}$ random points uniformly distributed in a $d$-dimensional hypercube or on the surface of a $d-1$-dimensional hypersphere embedded in $d$ dimensions. We use $P \in \{128, 256, 512, 1024, 2048, 4096, 8192\}$ and $P_{\text{test}}=8192$. For each value of $P$, we generate a Gaussian random field with covariance given by the teacher kernel, and we compute the kernel ridgeless regression predictor of the student kernel using \autoref{eq:loc-krrpredictor} with the $P$ training samples. The generalization error defined in \autoref{eq:loc-test-def} is approximated by computing the empirical mean squared error on the $P_{\text{test}}$ unseen samples. The expectation with respect to the target function is obtained averaging over 128 independent teacher Gaussian processes, each sampled on different points of the domain. As teacher and student kernels, we consider different combinations of the convolutional and local kernels defined in \autoref{eq:loc-conv-ker} and \autoref{eq:loc-loc-ker}, with Laplacian constituents $c(\x_i-\x_j) \, {=} \, e^{-\|\x_i-\x_j\|}$ and overlapping patches. In particular,
\begin{itemize}
\item the cases with convolutional teacher and both convolutional and local students with various filter sizes are reported in \autoref{fig:figure} and \autoref{fig:sphere} for data distributed in a hypercube and on a hypersphere, respectively;
\item the cases with local teacher and both local and convolutional students are reported in \autoref{fig:loc} for data distributed in a hypercube.
\end{itemize}

Experiments are run on NVIDIA Tesla V100 GPUs using the PyTorch package. The approximate total amount of time to reproduce all experiments with our setup is 400 hours.

\subsection{Additional experiments}

\begin{figure}
    \centering
    \includegraphics[width=1.0\linewidth]{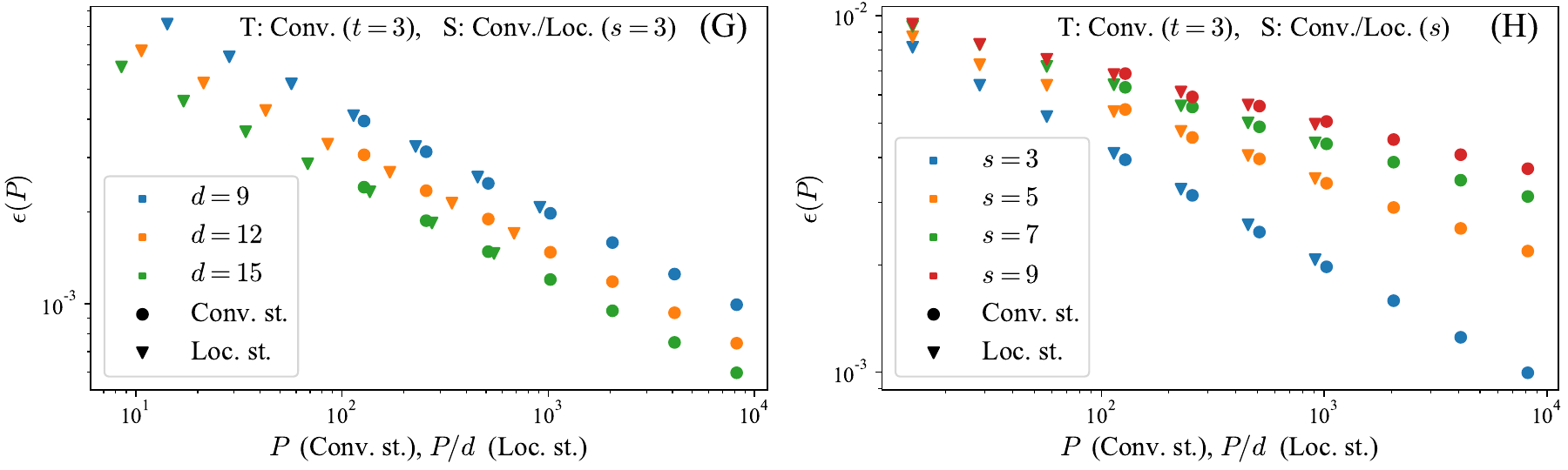}
    \caption{Learning curves for convolutional teacher and local and convolutional student kernels, with filter sizes denoted by $t$ and $s$ respectively. Data are sampled uniformly in the hypercube $[0,1]^d$, with $d=9$ if not specified otherwise. The sample complexity $P$ of the local students is rescaled with the number of patches to highlight the pre-asymptotic effect of shift-invariance on the learning curves.}
    \label{fig:preasympt}
\end{figure}

\paragraph{Convolutional vs local students} In \autoref{fig:preasympt} we report the empirical learning curves for convolutional and local student kernels learning a convolutional teacher kernel, with filter sizes $s$ and $t$ respectively. Data are uniformly sampled in the hypercube $[0,1]^d$. By rescaling the sample complexity $P$ of the local students with the number of patches $|\mathcal{P}|=d$, the learning curves of local and convolutional students overlap, confirming our prediction on the role of shift-invariance. Indeed, the local student has to learn the same local task at all the possible patch locations, while the convolutional student is naturally shift-invariant.

\begin{figure}
    \centering
    \includegraphics[width=1.0\linewidth]{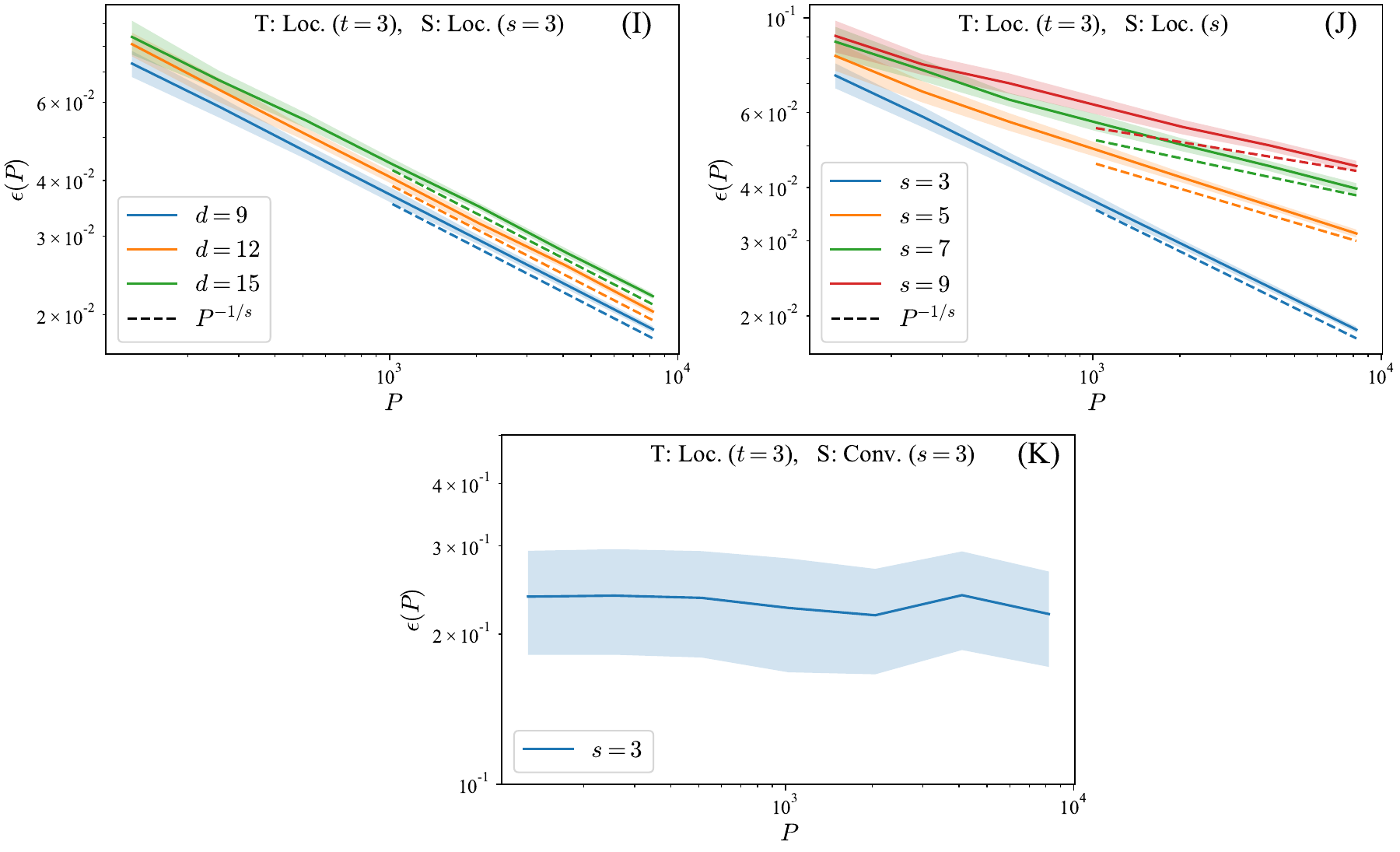}
    \caption{Learning curves for local teacher and local and convolutional student kernels, with filter sizes denoted by $t$ and $s$ respectively. Data are sampled uniformly in the hypercube $[0,1]^d$, with $d=9$ if not specified otherwise. Solid lines are the results of numerical experiments averaged over 128 realizations and the shaded areas represent the empirical standard deviations. The predicted scaling are shown by dashed lines.}
    \label{fig:loc}
\end{figure}

\paragraph{Local teacher} In \autoref{fig:loc} we report the empirical learning curves for a local teacher kernel and data uniformly sampled in the hypercube $[0,1]^d$. In panels I and J, also the student is a local kernel and the same discussion of \autoref{sec:loc-empirical} applies. In panel K, the student is a convolutional kernel and the generalization error does not decrease by increasing the size of the training set. Indeed, a local non-shift-invariant function is not on the span of the eigenfunctions of a convolutional kernel, and therefore the student is not able to learn the target.

\begin{figure}
    \centering
    \includegraphics[width=1.0\linewidth]{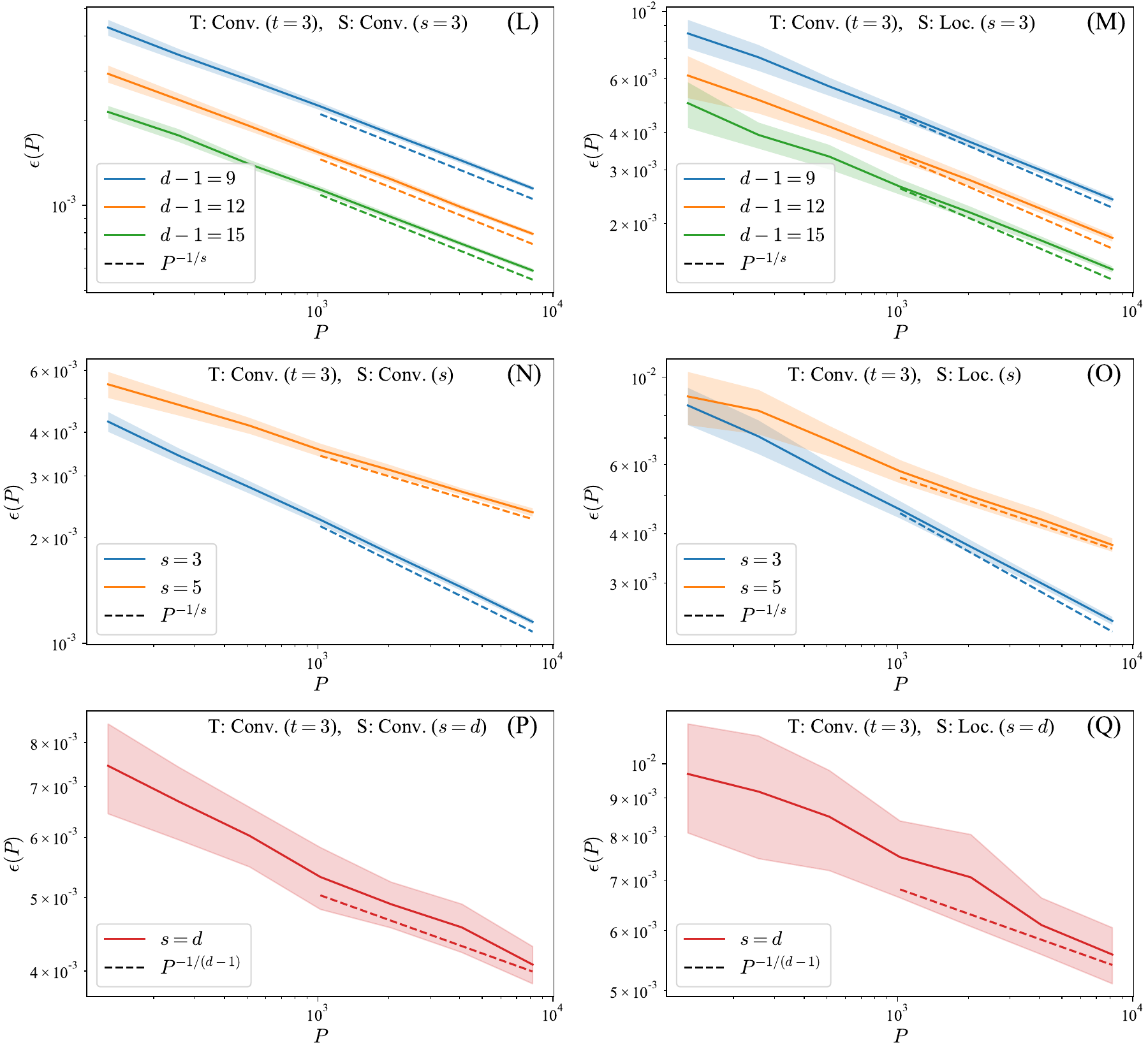}
    \caption{Learning curves for data uniformly distributed on the unit sphere $\mathbb{S}^{d-1}$, with $d=10$ if not specified otherwise. The teacher and student filter sizes are denoted with $t$ and $s$ respectively. Solid lines are the results of numerical experiments averaged over 128 realizations and the shaded areas represent the empirical standard deviations.}
    \label{fig:sphere}
\end{figure}

\paragraph{Spherical data} In \autoref{fig:sphere} we report the empirical learning curves for convolutional teacher and convolutional (left panels) and local (right panels) student kernels. Data are restricted to the unit sphere $\mathbb{S}^{d-1}$. Panels L-O are the analogous of panels A-D of \autoref{fig:figure}. Notice that when the filter size of the student coincides with $d$ (panels P, Q), the learning curves decay with exponent $\beta\,{\eq}\,1/(d-1)$ (instead of $\beta\,{=}\,1/d$). Indeed, for data normalized on $\mathbb{S}^{d-1}$, the spectrum of the Laplacian kernel decays at a rate $\mathcal{O}(k^{-\alpha-(d-1)})$ with $\alpha\,{\eq}\,1$. However, as the student filter size is lowered, we recover the exponent $1/s$ independently of the dimension $d$ of input space, as derived for data on the torus and shown empirically for data in the hypercube. In fact, we expect that the $s$-dimensional marginals of the uniform distribution on $\mathbb{S}^{d-1}$ become insensitive to the spherical constraint when $s\ll d$.

\begin{figure}
    \centering
    \includegraphics[width=1.0\linewidth]{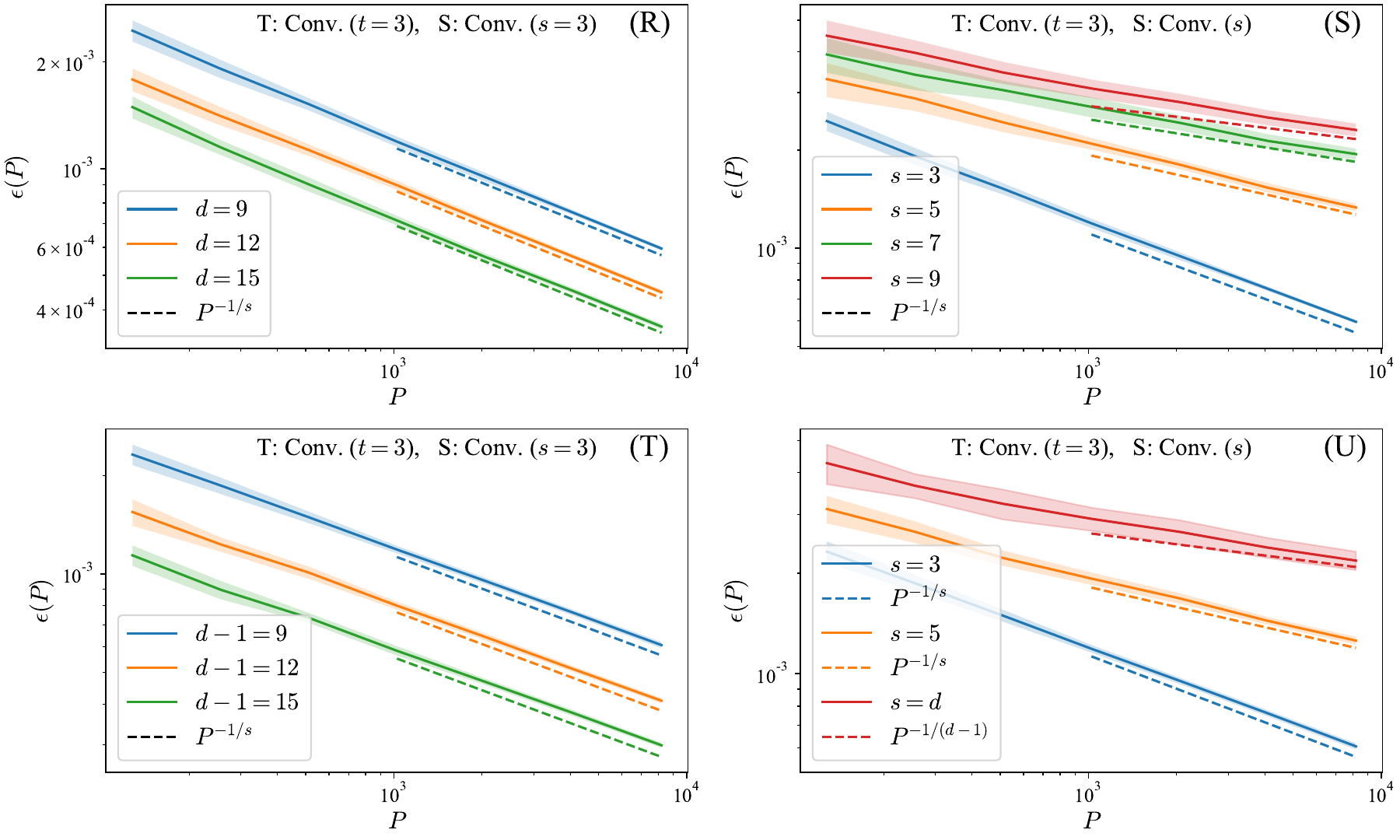}
    \caption{Learning curves for convolutional NTKs and data uniformly distributed in the hypercube $[0,1]^d$ (panels R, S) or on the unit sphere $\mathbb{S}^{d-1}$ (panels T, U). The teacher and student filter sizes are denoted with $t$ and $s$ respectively. Solid lines are the results of numerical experiments averaged over 128 realizations and the shaded areas represent the empirical standard deviations.}
    \label{fig:analytical-ntk}
\end{figure}

\begin{figure}
    \centering
    \includegraphics[width=1.0\linewidth]{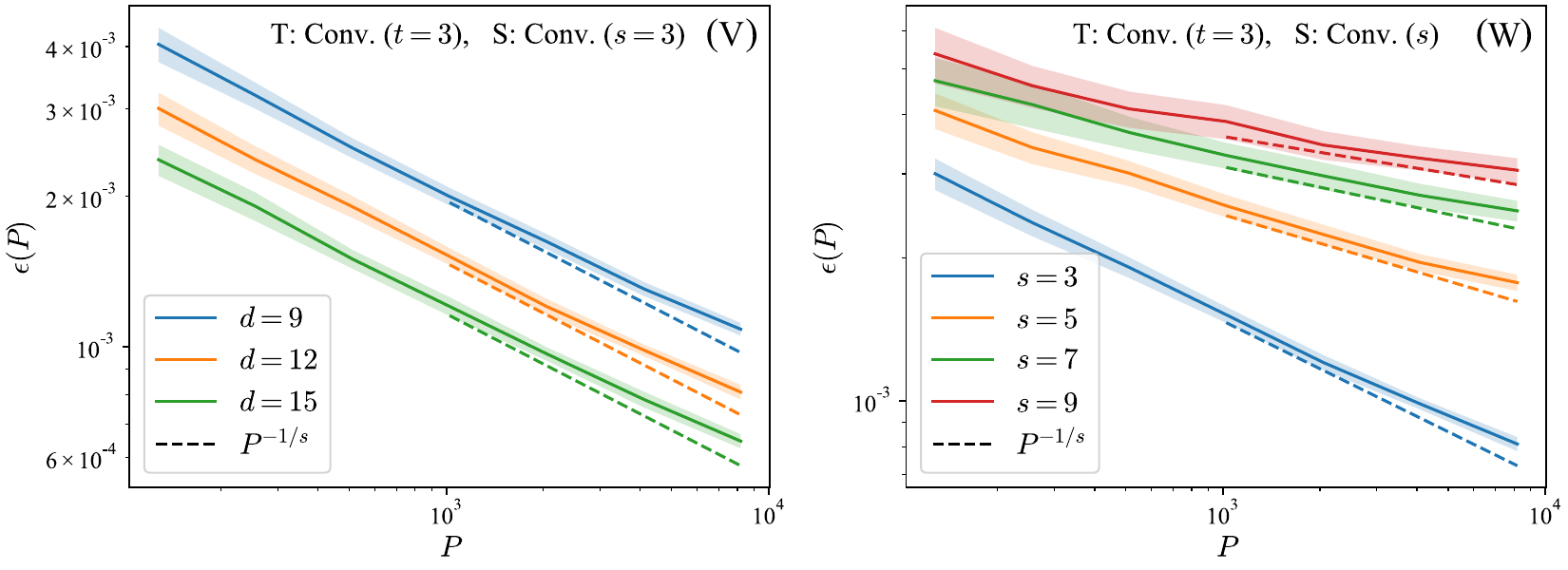}
    \caption{Learning curves for empirical NTKs of very-wide one-hidden-layer CNNs ($H\approx10^6$) and data uniformly distributed in the hypercube $[0,1]^d$. The teacher and student filter sizes are denoted with $t$ and $s$ respectively. Solid lines are the results of numerical experiments averaged over 128 realizations and the shaded areas represent the empirical standard deviations.}
    \label{fig:empirical-ntk}
\end{figure}

\paragraph{Convolutional NTKs} In \autoref{fig:analytical-ntk} we report the empirical learning curves obtained using the NTK of one-hidden-layer CNNs with ReLU activations, which corresponds to using the kernel $\K_{\mathrm{NTK}}^{FC}$ defined in \autoref{eq:loc-relu-fc-ntk} as the constituent. Since this kernel is not translationally invariant, it cannot be diagonalized in the Fourier domain, and the analysis of \autoref{sec:loc-learning-curves} does not apply. However, as shown in panels P-S, the same learning curve exponents $\beta$ of the Laplacian-constituent case are recovered. Indeed, $\K_{\mathrm{NTK}}^{FC}$ and the Laplacian kernel share the same nonanalytic behavior in the origin, and their spectra have the same asymptotic decay \cite{geifman2020similarity}. In \autoref{fig:empirical-ntk} we present the same plots of panels R and S, but instead of the analytical NTKs, we compute numerically the kernels of randomly-initialized very-wide CNNs ($H \approx 10^6$).

\begin{figure}
    \centering
    \includegraphics[width=1.0\linewidth]{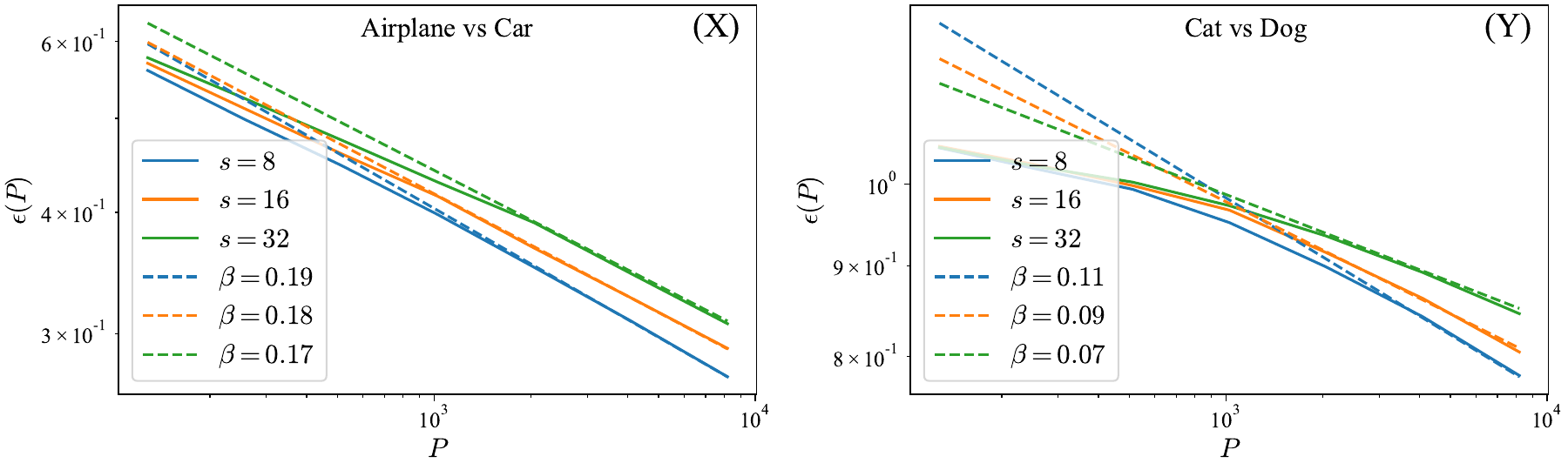}
    \caption{Learning curves of local kernels with filters of size $s$ on CIFAR-10 data. Solid lines are the results of numerical experiments and dashed lines are power laws with exponent $\beta$ interpolated in the last decade.}
    \label{fig:cifar}
\end{figure}

\paragraph{Real data} In Fig. \autoref{fig:cifar} we report the learning curves of local kernels with Laplacian constituents applied to the CIFAR-10 dataset. We build the tasks by selecting two classes and assigning label $+1$ to data from one class and $-1$ to data from the other class. As before, we use $P \in \{128, 256, 512, 1024, 2048, 4096, 8192\}$ and $P_{\text{test}} = 8192$. Differently from our assumptions, image data are strongly anisotropic, and the distance between nearest-neighbor points decays faster than $P^{-1/d}$. Indeed, target functions defined on data of this kind are usually not cursed with the full dimensionality $d$ of the inputs, but rather with an effective dimension $d_{\text{eff}}$. $d_{\text{eff}}$ is related to the dimension of the manifold in which data lie \cite{spigler2020asymptotic}, and may also vary when extracting patches of different sizes. Nonetheless, as we found in our synthetic setup, the learning curve exponent $\beta$ increases monotonically with the filter size of the kernel, strengthening the concept that leveraging locality is key for performance.

\chapter{Appendix: The Role of Depth and Spatial Adaptivity}

\section{Harmonic analysis on the sphere}\label{app:deep-harmonics}

This appendix collects some introductory background on spherical harmonics and dot-product kernels on the sphere~\cite{smola2000regularization}. See~\cite{efthimiou2014spherical, atkinson2012spherical, bach2017breaking} for a complete description. Spherical harmonics are homogeneous polynomials on the sphere $\mathbb{S}^{s-1}\,{=}\,\lbrace \x\in\mathbb{R}^s\,|\,\lVert \x \rVert\,{=}\,1\rbrace$, with $\lVert \cdot \rVert$ denoting the L2 norm. Given the polynomial degree $k\in\mathbb{N}$, there are $\mathcal{N}_{k,s}$ linearly independent spherical harmonics of degree $k$ on $\mathbb{S}^{s-1}$, with
\begin{equation}
\mathcal{N}_{k,s} = \frac{2k+s-2}{k}\binom{s+k-3}{k-1},\; \left\lbrace\begin{aligned} &\mathcal{N}_{0,d}=1\quad\forall d,\\ &\mathcal{N}_{k,d}\sim k^{d-2}\quad\text{for } k\gg 1. \end{aligned}\right. 
\end{equation}
Thus, we can introduce a set of $\mathcal{N}_{k,s}$ spherical harmonics $Y_{k,\ell}$ for each $k$, with $\ell$ ranging in $1,\dots,\mathcal{N}_{k,s}$, which are orthonormal with respect to the uniform measure on the sphere $d\tau(\x)$,
\begin{equation}\label{eq:deep-sphere-dot-prod}
    \left\langle Y_{k,\ell}, Y_{k,\ell'} \right\rangle_{\mathbb{S}^{s-1}} := \int_{\mathbb{S}^{s-1}} d\tau(\x)\, Y_{k,\ell}(\x) Y_{k,\ell'}(\x) = \delta_{\ell,\ell'}.
\end{equation}
Because of the orthogonality of homogeneous polynomials with a different degree, the set $\left\lbrace Y_{k,\ell} \right\rbrace_{k,\ell}$ is a complete orthonormal basis for the space of square-integrable functions on the $s$-dimensional unit sphere. Furthermore, spherical harmonics are eigenfunctions of the Laplace-Beltrami operator $\Delta$, which is nothing but the restriction of the standard Laplace operator to $\mathbb{S}^{s-1}$.
\begin{equation}\label{eq:deep-slap-eigvals}
    \Delta Y_{k,\ell} = -k(k+s-2)Y_{k,\ell}.
\end{equation}
The Laplace-Beltrami operator $\Delta$ can also be used to characterize the differentiability of functions $f$ on the sphere via the L2 norm of some power of $\Delta$ applied to $f$.

By fixing a direction $\y$ in $\mathbb{S}^{d-1}$ one can select, for each $k$, the only spherical harmonic of degree $k$ which is invariant for rotations that leave $\y$ unchanged. This particular spherical harmonic is, in fact, a function of $\x^\top\y$ and is called the Legendre polynomial of degree $k$, $P_{k,s}(\x^\top\y)$ (also referred to as Gegenbauer polynomial). Legendre polynomials can be written as a combination of the orthonormal spherical harmonics $Y_{k,\ell}$ via the addition formula~\cite{atkinson2012spherical}
\begin{equation}
    P_{k,s}(\x^\top\y) = \frac{1}{\mathcal{N}_{k,s}} \sum_{\ell=1}^{\mathcal{N}_{k,s}} Y_{k,\ell}(\x)Y_{k,\ell}(\y).
\end{equation}
Alternatively, $P_{k,s}$ is given explicitly as a function of $t\,{=}\,\x^\top\y\in[-1,+1]$ via the Rodrigues formula~\cite{atkinson2012spherical},
\begin{equation}\label{eq:deep-rodrigues}
    P_{k,s}(t) = \left(-\frac{1}{2}\right)^k \frac{\Gamma\left(\frac{s-1}{2}\right)}{\Gamma\left(k+\frac{s-1}{2}\right)}\left(1-t^2\right)^{\frac{3-s}{2}} \frac{d^k}{dt^k}\left(1-t^2\right)^{k+\frac{s-3}{2}}.
\end{equation}
Legendre polynomials are orthogonal on $[-1,+1]$ with respect to the measure with density $(1-t^2)^{(s-3)/2}$, which is the probability density function of the scalar product between two points on $\mathbb{S}^{s-1}$.
\begin{equation}\label{eq:deep-leg-ortho}
    \int_{-1}^{+1} dt\left(1-t^2\right)^{\frac{s-3}{2}}\, P_{k,s}(t)P_{k',s}(t) = \frac{|\mathbb{S}^{s-1}|}{|\mathbb{S}^{s-2}|}\frac{\delta_{k,k'}}{\mathcal{N}_{k,s}},
\end{equation}
with $|\mathbb{S}^{s-1}|$ denoting the surface area of the $s$-dimensional unit sphere.

To sum up, given $\x,\y\in\mathbb{S}^{s-1}$, functions of $\x$ or $\y$ can be expressed as a sum of projections on the orthonormal spherical harmonics $\left\lbrace Y_{k,\ell} \right\rbrace_{k,\ell}$, whereas functions of $\x^\top\y$ can be expressed as a sum of projections on the Legendre polynomials $\left\lbrace P_{k,s}(\x^\top\y)\right\rbrace_k $. The relationship between the two expansions is elucidated in the Funk-Hecke formula~\cite{atkinson2012spherical},
\begin{equation}
    \int_{\mathbb{S}^{s-1}} d\tau(\y)\,f(\x^\top\y) Y_{k,\ell}(\y) = Y_{k,\ell}(\x)\frac{|\mathbb{S}^{s-2}|}{|\mathbb{S}^{s-1}|}\int_{-1}^{+1} dt\left(1-t^2\right)^{\frac{s-3}{2}}\, f(t) P_{k,s}(t).
\end{equation}
If the function $f$ has continuous derivatives up to the $k$-th order in $[-1,+1]$, then one can plug Rodrigues' formula in the right-hand side of Funk-Hecke formula and get, after $k$ integrations by parts,
\begin{align}
    \int_{\mathbb{S}^{s-1}} &d\tau(\y)\,f(\x^\top\y) Y_{k,\ell}(\y) \\ &= Y_{k,\ell}(\x)\frac{|\mathbb{S}^{s-2}|}{|\mathbb{S}^{s-1}|}\frac{\Gamma\left(\frac{s-1}{2}\right)}{2^k \Gamma\left(k+\frac{s-1}{2}\right)}\int_{-1}^{+1} dt\, f^{(k)}(t)\left(1-t^2\right)^{k+\frac{s-3}{2}}, \nonumber
\end{align}
with $f^{(k)}(t)$ denoting the $k$-th order derivative of $f$ in $t$. This trick also applies to functions which are not $k$ times differentiable at $\pm 1$, provided the boundary terms due to integration by parts vanish.

\subsection{Dot-product kernels on the sphere}

Dot-product kernels are kernels that depend on the two inputs $\x$ and $\y$ via their scalar product $\x^\top\y$. When the inputs lie on the unit sphere $\mathbb{S}^{s-1}$, one can use the machinery introduced in the previous section to arrive immediately at the Mercer's decomposition of the kernel~\cite{smola2000regularization}.
\begin{equation}\label{eq:deep-dot-prod-mercer}\begin{aligned}
    \mathcal{K}(\x^\top\y) &= \sum_{k\geq 0} \left(\mathcal{N}_{k,s}\frac{|\mathbb{S}^{s-2}|}{|\mathbb{S}^{s-1}|}\int_{-1}^{+1} dt\left(1-t^2\right)^{\frac{s-3}{2}}\, \mathcal{K}(t) P_{k,s}(t)\right) P_{k,s}(\x^\top\y)\\
    &= \sum_{k\geq 0} \left(\frac{|\mathbb{S}^{s-2}|}{|\mathbb{S}^{s-1}|}\int_{-1}^{+1} dt\left(1-t^2\right)^{\frac{s-3}{2}}\, \mathcal{K}(t) P_{k,s}(t)\right) \sum_{\ell=1}^{\mathcal{N}_{k,s}} Y_{k,\ell}(\x)Y_{k,\ell}(\y)\\
    &:= \sum_{k\geq 0} \Lambda_k \sum_{\ell=1}^{\mathcal{N}_{k,s}} Y_{k,\ell}(\x)Y_{k,\ell}(\y).
\end{aligned}\end{equation}
In the first line, we have just decomposed $\mathcal{K}$ into projections onto the Legendre polynomials, the second line follows immediately from the addition formula, and the third is just a definition of the eigenvalues $\Lambda_k$. Notice that the eigenfunctions of the kernel are orthonormal spherical harmonics and the eigenvalues are degenerate with respect to the index $\ell$. The Reproducing Kernel Hilbert Space (RKHS) of $\mathcal{K}$ can be characterized as follows,
\begin{equation}
    \mathcal{H} = \left\lbrace f:\mathbb{S}^{s-1}\to \mathbb{R} \text{ s. t. } \left\lVert f \right\rVert_{\mathcal{H}}:= \sum_{k\geq 0,\Lambda_k\neq 0} \sum_{\ell=1}^{\mathcal{N}_{k,s}} \frac{\left\langle f, Y_{k,l} \right\rangle_{\mathbb{S}^{s-1}}^2}{\Lambda_k}< +\infty\right\rbrace.
\end{equation}

\subsection{Multi-dot-product kernels on the multi-sphere}

Mercer's decomposition of dot-product kernels extends naturally to the case considered in this paper, where the input space is the Cartesian product of $p$ $s$-dimensional unit sphere, 
\begin{equation}
    {\sf M}^p\mathbb{S}^{s-1} =\left\lbrace \x=(\x_1,\dots,\x_p) \right.\left| \x_i\in\mathbb{S}^{s-1} \,\forall\,i=1,\dots,p \right\rbrace = \bigtimes_{i=1}^p \mathbb{S}^{s-1} 
\end{equation}
which we refer to as the \emph{multi-sphere} following the notation of~\cite{geifman2022spectral}. After defining a scalar product between functions on ${\sf M}^p\mathbb{S}^{s-1}$ by direct extension of~\autoref{eq:deep-sphere-dot-prod}, one can immediately find a set of orthonormal polynomials by taking products of spherical harmonics. With the multi-index notation $\k\,{=}\,(k_1, \dots, k_p)$, $\bm{\ell}\,{=}\,(\ell_1, \dots, \ell_p)$, for all $\x\in{\sf M}^p\mathbb{S}^{s-1}$
\begin{align}
    \tilde{Y}_{\k,\bm{\ell}}(\x) = \prod_{i=1}^p Y_{k_i,\ell_i}(\x_i),&\text{ with }k_i\geq 0\text{, } \nonumber \\
    &\ell_i=1,\dots,\mathcal{N}_{k_i,s}=\frac{2k_i+s-2}{k_i}\binom{s+k_i-3}{k_i-1}.
\end{align}
These product spherical harmonics $\tilde{Y}_{\k,\bm{\ell}}(\x)$ span the space of square-integrable functions on ${\sf M}^p\mathbb{S}^{s-1}$. Furthermore, as each spherical harmonic is an eigenfunction of the Laplace-Beltrami operator, $\tilde{Y}_{\k,\bm{\ell}}$ is an eigenfunction of the sum of Laplace-Beltrami operators on the $p$ unit spheres,
\begin{equation}\label{eq:deep-mslap-eigvals}
    \Delta_{p,s} \tilde{Y}_{\k,\bm{\ell}} := \left(\sum_{i=1}^p \Delta_i\right)\prod_{i=1}^p  Y_{k_i,\ell_i} = \left(\sum_{i=1}^p\left((-k_i)(k_i+s-2)\right)\right)\ \tilde{Y}_{\k,\bm{\ell}}.
\end{equation}
We can thus characterize the differentiability of functions of the multi-sphere $\mathcal{X}_{s,p}$ via finiteness in L2 norm of some power of $\Delta_{p,s}$.

Similarly, we can consider products of Legendre polynomials to obtain a set of orthogonal polynomials on $[-1,1]^p$~(see~\cite{geifman2022spectral}, appendix A). Then, any function $f$ on ${\sf M}^p\mathbb{S}^{s-1}\times{\sf M}^p\mathbb{S}^{s-1}$ which depends only on the $p$ scalar products between patches,
\begin{equation}\label{eq:deep-multi-dot-product}
    f(\x,\y) = g(\x_1^\top\y_1, \dots, \x_p^\top\y_p),
\end{equation}
can be written as a sum of projections on products of Legendre polynomials
\begin{equation}
    \tilde{P}_{\k,s}(\bm{t}) := \prod_{i=1}^p P_{k_i,s}(t_i).
\end{equation}
Following~\cite{geifman2022spectral}, we call such functions \emph{multi-dot-product} kernels. When fixing one of the two arguments of $f$ (say $\x$), $f$ becomes a function on ${\sf M}^p\mathbb{S}^{s-1}\times{\sf M}^p\mathbb{S}^{s-1}$ and can be written as a sum of projections on the $\tilde{Y}_{\k,\bm{\ell}}$'s. The two expansions are related by the following generalized Funk-Hecke formula,
\begin{equation}\label{eq:deep-multi-fh}\begin{aligned}
\left(\prod_{i=1}^p \int_{\mathbb{S}^{s-1}} d\tau(\y_i)\right) &g(\x_1^\top\y_1, \dots, \x_p^\top\y_p) \tilde{Y}_{\k,\bm{\ell}}(\y) = \\
\tilde{Y}_{\k,\bm{\ell}}(\y) &\left(\frac{|\mathbb{S}^{s-2}|}{|\mathbb{S}^{s-1}|}\right)^p \left(\prod_{i=1}^p \int_{-1}^{+1} dt_i\left(1-t_i^2\right)^{\frac{s-3}{2}} P_{k_i,s}(t_i)\right)\,g(t_1,\dots,t_p).
\end{aligned}\end{equation}

Having introduced the product spherical harmonics $\tilde{Y}_{\k,\bm{\ell}}$ as basis of ${\sf M}^p\mathbb{S}^{s-1}$ and the product Legendre polynomials $\tilde{P}_{\k,s}(\bm{t})$ as basis of $[-1,+1]^p$, the Mercer's decomposition of multi-dot-product kernels follows immediately.
\begin{equation}\label{eq:deep-multi-dp-mercer}\begin{aligned}
    \mathcal{K}\left(\left\lbrace \x_i^\top\y_i\right\rbrace_i \right) &= \sum_{\k\geq 0} \left(\prod_{i=1}^p\mathcal{N}_{k_i,s}\frac{|\mathbb{S}^{s-2}|}{|\mathbb{S}^{s-1}|}\int_{-1}^{+1} dt_i\left(1-t_i^2\right)^{\frac{s-3}{2}}\,  P_{k_i,s}(t_i)\right) \\
    &\phantom{= \sum_{\k\geq 0}} \times \mathcal{K}\left( \left\lbrace t_i \right\rbrace_i\right) P_{\k,s}\left(\left\lbrace \x_i^\top\y_i\right\rbrace_i \right)\\
    &= \sum_{k\geq 0} \Lambda_k \sum_{\ell=1}^{\mathcal{N}_{k,s}} Y_{k,\ell}(\x)Y_{k,\ell}(\y).
\end{aligned}\end{equation}

\section{RFK and NTK of deep convolutional networks} \label{app:deep-kernel-lemmas}

This appendix gives the functional forms of the RFK and NTK of hierarchical CNNs. We refer the reader to \cite{arora2019exact} for the derivation.

\begin{definition}[RFK and NTK of hierarchical CNNs]\label{def:hierarchical-kernels}
Let $\x,\y\in {\sf M}^p\mathbb{S}^{s-1}={\textstyle\prod_{i=1}^p} \mathbb{S}^{s-1}$. Denote tuples of the kind $i_{l} i_{l+1} \dots i_{m}$ with $i_{l \to m}$ for $m\,{\geq}\,l$. For $m\,{<}\,l$, $i_{l\to m}$ denotes the empty tuple. For each tuple $i_{2\to L+1}$, denote with $t_{i_{2\to L+1}}$ the scalar product between the $s$-dimensional patches of $\x$ and $\y$ identified by the same tuple, i.e.
\begin{equation}
    t_{i_{2\to L+1}} =\x_{i_{2\to L+1}}^\top \y_{i_{2\to L+1}}
\end{equation}
For $1\,{\leq}\,l\,{\leq}\,L+1$, denote with $\left\lbrace t_{i_{2\to L+1}}\right\rbrace_{i_{2\to l}}$ the sequence of $t$'s obtained by letting the indices of the tuple $i_{2\to l}$ vary in their respective range. Consider a hierarchical CNN with $L$ hidden layers, filter sizes $(s_1,\dots,s_L)$, $p_L\,{\geq}\,1$ and all the weights $w^{\scriptstyle(1)}_{\scriptstyle h,i}, w^{\scriptstyle(l)}_{\scriptstyle h,h',i}, w^{\scriptstyle (L+1)}_{\scriptstyle h,i}$ initialized as Gaussian random numbers with zero mean and unit variance. 

\textbf{RFK.} The corresponding RFK (or covariance kernel) is a function $\mathcal{K}_{\mathrm{RFK}}^{\scriptscriptstyle(L+1)}$ of the $p_1\,{=}\,d/s_1$ scalar products $t_{i_L\dots i_1}$ which can be obtained recursively as follows. With $\kappa_1(t)\,{=}\,\left((\pi-\arccos{t})\, t +\sqrt{1-t^2}\right)/\pi$,
\begin{align}
&\mathcal{K}_{\mathrm{RFK}}^{(1)} (t_{i_{2\to L+1}}) = \kappa_1(t_{i_{2\to L+1}}); \nonumber\\
&\mathcal{K}_{\mathrm{RFK}}^{(l)}\left(\left\lbrace t_{i_{2\to L+1}}\right\rbrace_{i_{2\to l}}\right) = \kappa_1\left(\frac{1}{s_l}\sum_{i_l} \mathcal{K}_{\mathrm{RFK}}^{(l-1)}\left( \left\lbrace t_{i_{2\to L+1}}\right\rbrace_{i_{2\to l-1}}\right) \right),\; \forall\, l\in[2\2dots L]\,\text{ if }L\,{>}\,1;\nonumber\\
&\mathcal{K}_{\mathrm{RFK}}^{(L+1)} \left( \left\lbrace t_{i_{2\to L+1}}\right\rbrace_{i_{2\to L+1}} \right)= \frac{1}{p_L}\sum_{i_{L+1}=1}^{p_L} \mathcal{K}_{\mathrm{RFK}}^{(L)}\left( \left\lbrace t_{i_{2\to L+1}}\right\rbrace_{i_{2\to L}} \right).
\end{align}

\textbf{NTK.} The NTK of the same hierarchical CNN is also a function of the $p_1\,{=}\,d/s_1$ scalar products $t_{i_L\dots i_2}$ which can be obtained recursively as follows. With $\kappa_0(t)\,{=}\,\left(\pi-\arccos{t}\right)/\pi$,
\begin{align}
&\mathcal{K}_{\mathrm{NTK}}^{(1)}\left( t_{i_{2\to L+1}} \right) = \kappa_1(t_{i_{2\to L+1}}) + \left(t_{i_{2\to L+1}}\right) \kappa_0(t_{i_{2\to L+1}});\nonumber\\
&\mathcal{K}_{\mathrm{NTK}}^{(l)} \left(\left\lbrace t_{i_{2\to L+1}}\right\rbrace_{i_{2\to l}}\right) = \mathcal{K}_{\mathrm{RFK}}^{(l)} (\left\lbrace t_{i_{2\to L+1}}\right\rbrace_{i_{2\to l}}) + \left(\frac{1}{s_l}\sum_{i_l} \mathcal{K}_{\mathrm{NTK}}^{(l-1)}\left( \left\lbrace t_{i_{2\to L+1}}\right\rbrace_{i_{2\to l-1}}\right)\right)\nonumber\\ 
&\qquad\qquad\qquad\qquad\qquad\quad \times \kappa_0\left(\frac{1}{s_l}\sum_{i_l} \mathcal{K}_{\mathrm{RFK}}^{(l-1)}\left( \left\lbrace t_{i_{2\to L+1}}\right\rbrace_{i_{2\to l-1}}\right) \right),\; \forall\, l\in[2\2dots L]\,\text{ if }L\,{>}\,1;\nonumber\\
&\mathcal{K}_{\mathrm{NTK}}^{(L+1)} \left( \left\lbrace t_{i_{2\to L+1}}\right\rbrace_{i_{2\to L+1}} \right)= \frac{1}{p_L}\sum_{i_{L+1}=1}^{p_L} \mathcal{K}_{\mathrm{NTK}}^{(L)}\left( \left\lbrace t_{i_{2\to L+1}}\right\rbrace_{i_{2\to L}} \right).
\end{align}
\end{definition}

\section{Spectra of deep convolutional kernels} \label{app:deep-spectra}

In this section, we state and prove a generalized version of~\autoref{th:eig-scaling} which includes non-binary patches. Our proof strategy is to relate the asymptotic decay of eigenvalues to the singular behavior of the kernel, as it is customary in Fourier analysis and was done in~\cite{bietti2021deep} for standard dot-product kernel. In ~\autoref{ssec:proof-singular} we perform the singular expansion of hierarchical kernels, in~\autoref{ssec:proof-fourier} we use this expansion to prove \autoref{th:eig-scaling} with $L\,{=}\,2$ ($2$ hidden layers) and $s_1\,{=}\,2$ (patches on the ring), which we then generalize to general $s_1$ in~\autoref{ssec:proof-depth2} and to general depth in~\autoref{ssec:proof-general}.

\begin{theorem}[Spectrum of hierarchical kernels]\label{th:eig-scaling-app}
Let $T_{\mathcal{K}}$ be the integral operator associated with a $d$-dimensional hierarchical kernel of depth $L+1$, $L\,{>}\,1$ and filter sizes ($s_1,\dots,s_L$). Eigenvalues and eigenfunctions of $T_{\mathcal{K}}$ can be organized into $L$ sectors associated with the hidden layers of the kernel/network. For each $1\,{\leq}\,l\,{\leq}\,L$, the $l$-th sector consists of $(\textstyle\prod_{\scriptscriptstyle l'=1}^{\scriptscriptstyle l} s_{l'})$-\emph{local} eigenfunctions: functions of a single meta-patch $\x_{i_{l+1\to L+1}}$ which cannot be written as linear combinations of functions of smaller meta-patches. The labels $\k$ of these eigenfunctions are such that there is a meta-patch $\k_{i_{l+1 \to L+1}}$ of $\k$ with no vanishing sub-meta-patches and all the $k_i$'s outside of $\k_{i_{l+1 \to L+1}}$ are $0$ (because the eigenfunction is constant outside of $\x_{i_{l+1 \to L+1}}$). The corresponding eigenvalue is degenerate with respect to the location of the meta-patch: we call it $\Lambda^{\scriptscriptstyle(l)}_{\k_{i_{l+1}\to i_{L+1}}}$. When  $\|\k_{i_{l+1 \to L+1}}\|\to\infty$, with $k\,{=}\,\|\k_{i_{l+1 \to L+1}}\|$,
\begin{enumerate}
\item[i.] if $s_1=2$, then
\begin{equation}
        \Lambda^{(l)}_{\k_{i_{l+1\to L+1}}}
        = \mathcal{C}_{2,l}\, k^{-2\nu -d_{\mathrm{eff}}(l)} + o\left(k^{-2\nu -d_{\mathrm{eff}}(l)}\right),
\end{equation}
with $\nu_{\mathrm{NTK}}=1/2,\, \nu_{\mathrm{RFK}}=3/2$ and $d_{\mathrm{eff}}$ the effective dimensionality of the meta-patches defined in~\autoref{eq:deep-effective-dim}. $\mathcal{C}_{2,l}$ is a strictly positive constant for $l\,{\geq}\,2$ whereas for $l\,{=}\,1$ it can take two distinct strictly positive values depending on the parity of $k_{i_{2\to L+1}}$. \\
\item[ii.] if $s_1 \geq 3$, then for fixed non-zero angles $\k/k$,
\begin{equation}\label{eq:deep-eig-scaling-general}\begin{aligned}
    \Lambda^{(l)}_{\k_{i_{l+1\to L+1}}} = \mathcal{C}_{s_1,l}\left(\frac{\k_{i_{l+1 \to L+1}}}{k}\right) k^{-2\nu -d_{\mathrm{eff}}(l)} + o\left(k^{-2\nu -d_{\mathrm{eff}}(l)}\right),
\end{aligned}\end{equation}
where $\mathcal{C}_{s_1,l}$ is a positive function for $l\,{\geq}\,2$, whereas for $l\,{=}\,1$ it is a strictly positive constant which depends on the parity of $k_{i_{2\to L+1}}$.
\end{enumerate}
\end{theorem}

\subsection{Singular expansion of hierarchical kernels}\label{ssec:proof-singular}

Both the RFK and NTK of ReLU networks, whether deep or shallow, are built by applying the two functions $\kappa_0$ and $\kappa_1$~\cite{Cho2009} (see also~\autoref{def:hierarchical-kernels}),
\begin{equation}
    \kappa_0(t) = \frac{(\pi-\arccos{t})}{\pi},\quad \kappa_1(t) = \frac{(\pi-\arccos{t})\,t + \sqrt{1-t^2}}{\pi}.
\end{equation}
The functions $\kappa_0$ and $\kappa_1$ are non-analytic in $t\,{=}\,\pm 1$, with the following singular expansion~\cite{bietti2021deep}. Near $t\,{=}\,1$, with $u\,{=}\,1-t$
\begin{equation}\label{eq:deep-proof-taylor-plus}
    \left\lbrace\begin{aligned}  \kappa_0(1-u) &= 1 -\frac{\sqrt{2}}{\pi} u^{1/2}+ O(u^{3/2}),\\\kappa_1(1-u) &= 1-u +\frac{2\sqrt{2}}{3\pi} u^{3/2}+ O(u^{5/2}). \end{aligned}\right.
\end{equation}
Near $t\,{=}\,-1$, with $u\,{=}\,1+t$,
\begin{equation}\label{eq:deep-proof-taylor-minus}
    \left\lbrace\begin{aligned}  \kappa_0(-1+u) &= \frac{\sqrt{2}}{\pi} u^{1/2}+ O(u^{3/2}),\\\kappa_1(-1+u) &= \frac{2\sqrt{2}}{3\pi} u^{3/2}+ O(u^{5/2}). \end{aligned}\right. 
\end{equation}
As a result, hierarchical kernels have a singular expansion when the $t_{i_{2\to L+1}}$'s are close to $\pm 1$. In particular, the following expansions are relevant for computing the asymptotic scaling of eigenvalues.

\begin{proposition}[RFK when $\x\,{=}\,\y$]
The RFK of a hierarchical network of depth $L\,{+}\,1$, filter sizes $(s_1,\dots,s_L)$ and $p_L\,{\geq}\,1$ has the following singular expansion when all $t_{i_{2\to L+1}}\to 1$. With $u_{i_{2\to L+1}}\,{=}\,1-t_{i_{2\to L+1}}$, $c\,{=}\,2\sqrt{2}/(3\pi)$, and  $\prod_{l\in I} s_l\,{:=}\,1$ if $I$ is the empty set,
\begin{equation}\begin{aligned}\label{eq:deep-rfk-sing-plus}
\mathcal{K}_{\mathrm{RFK}}^{(L+1)} \left( \left\lbrace 1-u_{i_{2\to L+1}}\right\rbrace_{i_{2\to L+1}} \right) &= 1-\frac{1}{\left(\displaystyle\prod_{2\leq l' \leq L}s_{l'}\right)p_L} \sum_{i_{2\to L+1}} u_{i_{2\to L+1}}\\
& + \frac{c}{p_L}\sum_{l'=1}^{L} \frac{1}{\left(\displaystyle\prod_{l' < l'' \leq L}s_{l''}\right)} \sum_{i_{{l'+1}\to{L+1}}}\left( \frac{\sum_{i_{2\to{l'}}} u_{i_{2\to L+1}}}{\left(\displaystyle\prod_{2 \leq l'' \leq l'}s_{l''}\right)} \right)^{3/2}\\
&+ O(u_{i_{2\to L+1}}^{5/2})
\end{aligned}\end{equation}
\end{proposition}
\emph{Proof.} With $L\,{=}\,1$ one has
(recall that $i_{2\to 1+1}=i_{2\to 2}$ reduces to a single index)
\begin{align}\label{eq:deep-proof-hierarchical-rfk-taylor-1}
\mathcal{K}_{\mathrm{RFK}}^{(1)} (1-u_{i_2}) &= 1-u_{i_{2}}+cu_{i_{2}}^{3/2} + O(u_{i_{2}}^{5/2})\Rightarrow \nonumber\\
\mathcal{K}_{\mathrm{RFK}}^{(1+1)} \left(\left\lbrace 1-u_{i_2}\right\rbrace_{i_2}\right) &= 1-\frac{1}{p_1} \sum_{i_2} u_{i_{2}}+\frac{c}{p_1}\sum_{i_2} u_{i_{2}}^{3/2} + O(u_{i_{2}}^{5/2}).
\end{align}
With $L\,{=}\,2$,
\begin{align}
\mathcal{K}_{\mathrm{RFK}}^{(2)}\left(\left\lbrace 1-u_{i_2}\right\rbrace_{i_2}\right) &= \kappa_1\left(1 - \frac{1}{s_2}\sum_{i_2}u_{i_2,i_3} + \frac{c}{s_2}\sum_{i_2}u_{i_2,i_3}^{3/2} + O(u_{i_2,i_3}^{5/2}) \right)\nonumber\\
 &= 1 - \frac{1}{s_2}\sum_{i_2}u_{i_2,i_3} + \frac{c}{s_2}\sum_{i_2}u_{i_2,i_3}^{3/2} + c\left( \frac{1}{s_2}\sum_{i_2}u_{i_2,i_3} \right)^{3/2} + O(u_{i_2,i_3}^{5/2}),
\end{align}
therefore
\begin{align}\label{eq:deep-hierarchical-rfk-taylor-2}
\mathcal{K}_{\mathrm{RFK}}^{(2+1)}\left(\left\lbrace 1-u_{i_2,i_3}\right\rbrace_{i_2,i_3}\right) = &1 - \frac{1}{s_2p_2}\sum_{i_2,i_3} u_{i_2,i_3} + \frac{c}{p_2}\frac{1}{s_2}\sum_{i_2,i_3}u_{i_2,i_3}^{3/2} \nonumber\\ &+ \frac{c}{p_2}\sum_{i_3}\left( \frac{1}{s_2}\sum_{i_2}u_{i_2,i_3} \right)^{3/2}
+ O(u_{i_2,i_3}^{5/2}).
\end{align}
The proof of the general case follows by induction by applying the function $\kappa_1$ to the singular expansion of the kernel with $L-1$ hidden layers, then using~\autoref{eq:deep-proof-taylor-plus}.

\begin{proposition}[RFK when $\x\,{=}\,-\y$]
The RFK of a hierarchical network of depth $L\,{+}\,1$, filter sizes $(s_1,\dots,s_L)$ and $p_L\,{\geq}\,1$ has the following singular expansion when all $t_{i_{2\to L+1}}\to -1$. With $u_{i_{2\to L+1}}\,{=}\,1+t_{i_{2\to L+1}}$, $c\,{=}\,2\sqrt{2}/(3\pi)$ and  $\prod_{l\in I} s_l\,{:=}\,1$ if $I$ is the empty set,
\begin{equation}\begin{aligned}
\mathcal{K}_{\mathrm{RFK}}^{(L+1)} \left( \left\lbrace -1+u_{i_{2\to L+1}}\right\rbrace_{i_{2\to L+1}} \right) &= b_L + \frac{c_L}{\left(\displaystyle\prod_{2\leq l' \leq L}s_{l'}\right)p_L} \sum_{i_{2\to L+1}} u_{i_{2\to L+1}}^{3/2} +  O(u_{i_{2\to L+1}}^{5/2}),
\end{aligned}\end{equation}
with $b_L\,{=}\,\kappa_1(b_{L-1})$, $b_1\,{=}\,0$; and $c_L\,{=}\, c_{L-1}\kappa_1'(b_{L-1})$, $c_1\,{=}\,c$.
\end{proposition}
\emph{Proof.} This can be proved again by induction. For $L\,{=}\,1$,
\begin{align}
\mathcal{K}_{\mathrm{RFK}}^{(1)} (-1+u_{i_2}) &= cu_{i_{2}}^{3/2} + O(u_{i_{2}}^{5/2})\Rightarrow \nonumber \\ \mathcal{K}_{\mathrm{RFK}}^{(1+1)} \left(\left\lbrace -1+u_{i_2}\right\rbrace_{i_2}\right) &= \frac{c}{p_1} \sum_{i_2} u_{i_{2}}^{3/2} + O(u_{i_{2}}^{5/2}).
\end{align}
Thus, for $L\,{=}\,2$,
\begin{align}
\mathcal{K}_{\mathrm{RFK}}^{(2)} \left(\left\lbrace -1+u_{i_2,i_3}\right\rbrace_{i_2}\right) &= \kappa_1\left(\frac{c}{s_2} \sum_{i_2} u_{i_2,i_3}^{3/2} + O(u_{i_2,i_3}^{5/2})\right)\nonumber \\ &= \kappa_1(0) + \kappa_1'(0)\left(\frac{c}{s_2} \sum_{i_2} u_{i_2,i_3}^{3/2}\right) + O(u_{i_2,i_3}^{5/2}),
\end{align}
so that
\begin{align}
\mathcal{K}_{\mathrm{RFK}}^{(2+1)} \left(\left\lbrace -1+u_{i_2,i_3}\right\rbrace_{i_2,i_3}\right) = \kappa_1(0) + \frac{\kappa_1'(0)c}{s_2 p_2} \sum_{i_2, i_3} u_{i_2,i_3}^{3/2} + O(u_{i_2,i_3}^{5/2}).
\end{align}
The proof is completed by applying the function $\kappa_1$ to the singular expansion of the kernel with $L-1$ hidden layers.

\begin{proposition}[NTK when $\x\,{=}\,\y$]
The NTK of a hierarchical network of depth $L\,{+}\,1$, filter sizes $(s_1,\dots,s_L)$ and $p_L\,{\geq}\,1$ has the following singular expansion when all $t_{i_{2\to L+1}}\to 1$. With $u_{i_{2\to L+1}}\,{=}\,1-t_{i_{2\to L+1}}$, $c\,{=}\,\sqrt{2}\pi$, and  $\prod_{l\in I} s_l\,{:=}\,1$ if $I$ is the empty set,
\begin{equation}\begin{aligned}
\mathcal{K}_{\mathrm{NTK}}^{(L+1)} \left( \left\lbrace 1-u_{i_{2\to L+1}}\right\rbrace_{i_{2\to L+1}} \right) &= L+1-\frac{c}{p_L}\sum_{l'=1}^{L} \frac{l'}{\left(\displaystyle\prod_{l' < l'' \leq L}s_{l''}\right)} \\ &\times \sum_{i_{{l'+1}\to{L+1}}} \left( \frac{1}{\left(\displaystyle\prod_{2 \leq l'' \leq l'}s_{l''}\right)} \sum_{i_{2\to{l'}}} u_{i_{2\to L+1}} \right)^{1/2} + O(u_{i_{2\to L+1}}^{3/2})
\end{aligned}\end{equation}
\end{proposition}

\begin{proposition}[NTK when $\x\,{=}\,-\y$]
The NTK of a hierarchical network of depth $L\,{+}\,1$, filter sizes $(s_1,\dots,s_L)$ and $p_L\,{\geq}\,1$ has the following singular expansion when all $t_{i_{2\to L+1}}\to -1$. With $u_{i_{2\to L+1}}\,{=}\,1+t_{i_{2\to L+1}}$, $c\,{=}\,\sqrt{2}/\pi$ and  $\prod_{l\in I} s_l\,{:=}\,1$ if $I$ is the empty set,
\begin{equation}\begin{aligned}
\mathcal{K}_{\mathrm{NTK}}^{(L+1)} \left( \left\lbrace -1+u_{i_{2\to L+1}}\right\rbrace_{i_{2\to L+1}} \right) &= a_L + \frac{c_L}{\left(\displaystyle\prod_{2\leq l' \leq L}s_{l'}\right)p_L} \sum_{i_{2\to L+1}} u_{i_{2\to L+1}}^{3/2} +  O(u_{i_{2\to L+1}}^{5/2}),
\end{aligned}\end{equation}
with $a_L\,{=}\,b_L + b_{L-1}\kappa_0(b_{L-1})$, $b_L\,{=}\,\kappa_1(b_{L-1})$, $b_1\,{=}\,0$; and $c_L\,{=}\,c_{L-1}\kappa_0(b_{L-1})$,  $c_1\,{=}\,c$. Notice that both $\kappa_1$ and $\kappa_0$ are positive and strictly increasing in $[0,1]$ and $\kappa_1(1)\,{=}\,\kappa_0(1)\,{=}\,1$, thus $b_L\in(0,1)$ and $c_L\,{<}\,c_{L-1}$.
\end{proposition}
The proofs of the two propositions above are omitted, as they follow the exact same steps as the previous two proofs.

\subsection{Patches on the ring}\label{ssec:proof-fourier}

In this section, we prove a restricted version of~\autoref{th:eig-scaling} for the case of $2$-dimensional input patches, since the reduction of spherical harmonics to the Fourier basis simplifies the proof significantly. We also consider, for convenience, hierarchical kernels of depth $3$ with the filter size of the second hidden layer set to $p\,{=}\,d/2$, the total number of $2$-patches of the input. Once this case is understood, extension to arbitrary filter size and arbitrary depth is trivial.

\begin{theorem}[Spectrum of depth-$3$ kernels on $2$-patches]\label{th:eig-scaling-2d}
Let $T_{\mathcal{K}}$ be the integral operator associated with a $d$-dimensional hierarchical kernel of depth $3$, ($2$ hidden layers), with filter sizes ($s_1\,{=}\,2,s_2$) and $p_2\,{=}\,1$, such that $2 s_2\,{=}\,d$ and $s_2\,{=}\,p$ (the number of $2$-patches). Eigenvalues and eigenfunctions of $T_{\mathcal{K}}$ can be organized into $2$ sectors associated with the hidden layers of the kernel/network. 
\begin{enumerate}
\item[i.] The first sector consists of $s_1$-\emph{local} eigenfunctions, which are functions of a single patch $\x_{i}$ for $i\,{=}\,1,\dots,p$. The labels $\k,\bm{\ell}$ of local eigenfunctions are such that all the $k_j$'s with $j\neq i$ are zero (because the eigenfunction is constant outside $\x_{i}$). The corresponding eigenvalue is degenerate with respect to the location of the patch: we call it $\Lambda^{\scriptscriptstyle(1)}_{k_i}$. When  $k_i\to\infty$,
\begin{equation}\label{eq:deep-eig-scaling-2d-local}
        \Lambda^{(1)}_{k_i} = \mathcal{C}_{2,1}\, k^{-2\nu -1} + o\left(k^{-2\nu -1}\right),
\end{equation}
with $\nu_{\mathrm{NTK}}=1/2,\, \nu_{\mathrm{RFK}}=3/2$. $\mathcal{C}_{2,l}$ can take two distinct strictly positive values depending on the parity of $k_{i}$;
\item[ii.] The second sector consists of \emph{global} eigenfunctions, which are functions of the whole input $\x$. The labels $\k,\bm{\ell}$ of global eigenfunctions are such that at least two of the $k_i$'s are non-zero. We call the corresponding eigenvalue $\Lambda^{\scriptscriptstyle(2)}_{\k}$. When $\|\k\|\to\infty$, with $k\,{=}\,\|\k\|$,
\begin{equation}\label{eq:deep-eig-scaling-2d-global}
        \Lambda^{(2)}_{\k} = \mathcal{C}_{2,2}\, k^{-2\nu -p} + o\left(k^{-2\nu -p}\right),
\end{equation}
\end{enumerate}
\end{theorem}

\emph{Proof.} If we consider binary patches in the first layer, the input space becomes the Cartesian product of two-dimensional unit spheres, i.e., circles, $\mathcal{X}=\prod_{i=1}^d\mathbb{S}^1$. Then, each patch $\x_i$ corresponds to an angle $\theta_i$ and the spherical harmonics are equivalent to Fourier atoms,
\begin{equation}
    Y_0(\theta) = 1, \quad Y_{k,1}(\theta) = e^{ik\theta}, \quad Y_{k,2}(\theta) = e^{-ik\theta}, \quad \forall k \geq 1.
\end{equation}
Therefore, solving the eigenvalue problem for a dot-product kernel $\mathcal{K}(\x^\top\y) = \mathcal{K}\left(\cos(\theta_x - \theta_y)\right)$ with $\x,\,\y \in \mathbb{S}^1$ reduces to computing its Fourier transform. With $|\mathbb{S}^0|\,{=}\,2$ and $|\mathbb{S}^1|\,{=}\,2\pi$,
\begin{equation}
    \frac{1}{2\pi} \int_{-\pi}^{\pi} d\theta_x \, \mathcal{K}\left(\cos(\theta_x - \theta_y)\right) e^{\pm ik\theta_x} = \Lambda_k e^{\pm ik\theta_y}\Rightarrow \Lambda_k = \frac{1}{2\pi} \int_{-\pi}^{\pi} d\theta \, \mathcal{K}\left(\cos\theta\right) e^{\pm ik\theta},
\end{equation}
where we denoted with $\theta$ the difference between the two angles. Similarly, for a multi-dot-product kernel, the eigenvalues coincide with the $p$-dimensional Fourier transform of the kernel, where $p$ is the number of patches,
\begin{align}\label{eq:deep-eval-hier}
    \Lambda_{\k} &= \frac{1}{(2\pi)^p} \int_{-\pi}^{\pi} \left( \prod_{i=1}^p d\theta_i \, e^{\pm ik_i\theta_i} \right) \mathcal{K}\left(\{\cos\theta_i\}_{i=1}^p\right) \nonumber \\
    &= \frac{1}{(2\pi)^p} \int_{-\pi}^{\pi} d^p\bm{\theta} \, e^{\pm i\k^\top\bm{\theta}} \mathcal{K}\left(\{\cos\theta_i\}_{i=1}^p\right),
\end{align}
with $\k=(k_1,\dots,k_p)^\top$ the vector of the patch wavevectors and $\bm{\theta}=(\theta_1,\dots,\theta_p)^\top$ the vector of the patch angle differences $\theta_i=\theta_{x,i}-\theta_{y,i}$.

The nonanaliticity of the kernel at $t_i\,{=}\,1$ for all $i$ moves to $\theta_i\,{=}\,0$ for all $i$, whereas those in $t_i\,{=}\,-1$ move to $\theta_i\,{=}\,\pi$ and $-\pi$. The corresponding singular expansion is obtained from~\autoref{eq:deep-rfk-sing-plus} after replacing $t_i$ with $\cos{(\theta_i)}$ and expanding $\cos{(\theta_i)}$ as $1-\theta_i^2/2$, resulting in
\begin{equation}
    \mathcal{K}_{\mathrm{RFK}}^{(2)} (\{\cos\theta_i\}_{i=1}^p) = 1 - \frac{1}{2p} \sum_{i=1}^p \theta_i^2 + \frac{1}{3\pi p} \sum_{i=1}^p |\theta_i|^3 + \frac{2\sqrt{2}}{3\pi} \left( \frac{1}{p} \sum_{i=1}^p \frac{\theta_i^2}{2}\right)^{3/2} + \sum_{i=1}^p O(\theta_i^4).
\end{equation}
The first nonanalytic terms are $\frac{1}{3\pi p} \sum_{i=1}^p |\theta_i|^3$ and $\frac{2\sqrt{2}}{3\pi} \left( \frac{1}{p} \sum_{i=1}^p \frac{\theta_i^2}{2}\right)^{3/2}$. After recalling that the Fourier transform of $\|\bm{\theta}\|^{2\nu}$ with $\bm{\theta} \in \mathbb{R}^p$ decays asymptotically as $\|\k\|^{-2\nu-p}$~\cite{widom1963asymptotic}, one has ($\nu\,{=}\,3/2$)
\begin{align}
    \frac{1}{(2\pi)^p} \int_{-\pi}^\pi d^p\bm{\theta} \, e^{\pm i\k^\top\bm{\theta}} \frac{1}{3\pi p} \sum_{i=1}^p |\theta_i|^3 \sim \sum_{i=1}^p k_i^{-4} \prod_{j\neq i}\delta_{k_j,0}, \quad \text{for } \|\k\| \to \infty
\end{align}
and
\begin{align}
    \frac{1}{(2\pi)^p} \int_{-\pi}^\pi d^p\bm{\theta} \, e^{\pm i\k^\top\bm{\theta}}\|\bm{\theta}\|^{3}  \sim \|\k\|^{-p-3}, \quad \text{for } \|\k\| \to \infty. 
\end{align}
All the other terms in the kernel expansion will result in subleading contributions in the Fourier transform. Therefore, the former of the two equations above yields the asymptotic scaling of eigenvalues of the local sector, whereas the latter yields the asymptotic scaling of the global sector.

The proof for the NTK case is analogous to the RFK case, except that the singular expansion near $\theta_i\,{=}\,0$ is given by
\begin{equation}
     \mathcal{K}_{\mathrm{NTK}}^{(2)} (\{\cos\theta_i\}_{i=1}^p) = 3 - \frac{1}{p}\sum_{i=1}^p \frac{|\theta_i|}{2} - \frac{\sqrt{2}}{\pi}\left( \frac{1}{p}\sum_{i=1}^p \frac{\theta_i^2}{2} \right)^{1/2} + \sum_{i=1}^p O(\theta_i^{3/2}).
\end{equation}

\subsection{Patches on the \texorpdfstring{$s$}{s}-dimensional hypersphere}\label{ssec:proof-depth2}

In this section, we make an additional step towards~\autoref{th:eig-scaling} by extending~\autoref{th:eig-scaling-2d} to the case of $s$-dimensional input patches. We still consider hierarchical kernels of depth $3$ with the filter size of the second hidden layer set to $p\,{=}\,d/s$ (the total number of $s$-patches of the input) so as to ease the presentation. The extension to general depth and filter sizes is presented in~\autoref{ssec:proof-general}.

\begin{theorem}[Spectrum of depth-$3$ kernels on $s$-patches]\label{th:eig-scaling-sd}
Let $T_{\mathcal{K}}$ be the integral operator associated with a $d$-dimensional hierarchical kernel of depth $3$, ($2$ hidden layers), with filter sizes ($s_1\,{=}\,s,s_2$) and $p_2\,{=}\,1$, such that $2 s_2\,{=}\,d$ and $s_2\,{=}\,p$ (the number of $s$-patches). Eigenvalues and eigenfunctions of $T_{\mathcal{K}}$ can be organized into $2$ sectors associated with the hidden layers of the kernel/network. 
\begin{enumerate}
\item[i.] The first sector consists of $s_1$-\emph{local} eigenfunctions, which are functions of a single patch $\x_{i}$ for $i\,{=}\,1,\dots,p$. The labels $\k,\bm{\ell}$ of local eigenfunctions are such that all the $k_j$'s with $j\neq i$ are zero (because the eigenfunction is constant outside of $\x_{i}$). The corresponding eigenvalue is degenerate with respect to the location of the patch: we call it $\Lambda^{\scriptscriptstyle(1)}_{k_i}$. When $k_i\to\infty$,
\begin{equation}\label{eq:deep-eig-scaling-sd-local}
        \Lambda^{(1)}_{k_i} = \mathcal{C}_{s,1}\, k^{-2\nu -(s-1)} + o\left(k^{-2\nu -(s-1)}\right),
\end{equation}
with $\nu_{\mathrm{NTK}}=1/2,\, \nu_{\mathrm{RFK}}=3/2$. $\mathcal{C}_{s,1}$ can take two distinct strictly positive values depending on the parity of $k_{i}$;
\item[ii.] The second sector consists of \emph{global} eigenfunctions, which are functions of the whole input $\x$. The labels $\k,\bm{\ell}$ of global eigenfunctions are such that at least two of the $k_i$'s are non-zero. We call the corresponding eigenvalue $\Lambda^{\scriptscriptstyle(2)}_{\k}$. When $k\equiv \| \k\|\to\infty$, for fixed non-zero angles $\k/k$,
\begin{equation}\label{eq:deep-eig-scaling-sd-global}\begin{aligned}
    \Lambda^{(2)}_{\k} = \mathcal{C}_{s,2}\left(\frac{\k}{k}\right) k^{-2\nu -p(s-1)} + o\left(k^{-2\nu -p(s-1)}\right),
\end{aligned}\end{equation}
where $\mathcal{C}_{s,2}$ is a positive function.
\end{enumerate}
\end{theorem}

\emph{Proof.} A hierarchical RFK/NTK is a multi-dot-product kernel, therefore its eigenfunctions are products of spherical harmonics $\tilde{Y}_{\k,\bm{\ell}}(\x) = \prod_{i=1}^p Y_{k_i,\ell_i}(\x_i)$ and the eigenvalues of $\mathcal{K}$ are given by~\autoref{eq:deep-multi-dp-mercer},
\begin{equation}\label{eq:deep-proof-eigvals}
    \Lambda_{\k} = \left(\prod_{i=1}^p\frac{|\mathbb{S}^{s-2}|}{|\mathbb{S}^{s-1}|}\int_{-1}^{+1} dt_i\left(1-t_i^2\right)^{\frac{s-3}{2}}\,  P_{k_i,s}(t_i)\right)\mathcal{K}\left( \left\lbrace t_i \right\rbrace_i\right).
\end{equation}
The proof follows the following strategy: first, we show that the infinitely differentiable part of $\mathcal{K}$ results in eigenvalues which decay faster than any polynomial of the degrees $k_i$. We then show that the decay is controlled by the most singular term of the singular expansion of the kernel and finally compute such decay by relating it to the number of derivatives of the kernel having a finite l2 norm.

When $\mathcal{K}$ is infinitely differentiable in $[-1,+1]^p$, we can plug Rodrigues' formula~\autoref{eq:deep-rodrigues} for each $P_{k_i,s}(t_i)$ and get
\begin{equation}\label{eq:deep-proof-rodrigues}
    \Lambda_{\k} = \left(\prod_{i=1}^p\frac{|\mathbb{S}^{s-2}|}{|\mathbb{S}^{s-1}|}\left(-\frac{1}{2}\right)^{k_i} \frac{\Gamma\left(\frac{s-1}{2}\right)}{\Gamma\left(k_i+\frac{s-1}{2}\right)}\right) \int_{-1}^{+1} d\bm{t}\, \mathcal{K}\left( \bm{t}\right) \left(\prod_{i=1}^p\frac{d^{k_i}}{dt^{k_i}_i}\left(1-t_i^2\right)^{k_i+\frac{s-3}{2}} \right),
\end{equation}
with $\int_{-1}^{+1} d\bm{t}$ denoting integration over the $p$-dimensional hypercube $[-1,+1]^p$. We can simplify the integral further via integration by parts, so as to obtain
\begin{equation}\label{eq:deep-proof-parts}
    \Lambda_{\k} = \left(\prod_{i=1}^p\frac{|\mathbb{S}^{s-2}|}{|\mathbb{S}^{s-1}|}\left(\frac{1}{2}\right)^{k_i} \frac{\Gamma\left(\frac{s-1}{2}\right)}{\Gamma\left(k_i+\frac{s-1}{2}\right)}\right) \int_{-1}^{+1} d\bm{t}\, \mathcal{K}^{(\k)}\left( \bm{t}\right) \left(\prod_{i=1}^p\left(1-t_i^2\right)^{k_i+\frac{s-3}{2}} \right),
\end{equation}
where $\mathcal{K}^{(\k)}$ denotes the partial derivative of order $k_1$ with respect to $t_1$, $k_2$ with respect to $t_2$ and so on until $k_p$ with respect to $t_p$. Notice that the function $(1-t^2)^{\frac{d-3}{2}}$ is proportional to the probability measure of the scalar product $t$ between two points sampled uniformly at random on the unit sphere~\cite{atkinson2012spherical},
\begin{equation}
    \lvert\mathbb{S}^{d-1} \rvert = \int_{-1}^{+1}dt\, (1-t^2)^{\frac{d-3}{2}} \int_{\mathbb{S}^{d-2}} dS^{d-2} \Rightarrow \frac{\lvert\mathbb{S}^{d-1} \rvert}{\lvert\mathbb{S}^{d-2} \rvert} \int_{-1}^{+1}dt\, (1-t^2)^{\frac{d-3}{2}} = 1.
\end{equation}
This probability measure converges weakly to a Dirac mass $\delta(t)$ when $d\to\infty$. Recall, in addition, that $\lvert\mathbb{S}^{d-1} \rvert\,{=}\,2 \pi^{d/2}/\Gamma(d/2)$, where $\Gamma$ denotes the Gamma function $\Gamma(x)\,{=}\,\int_0^\infty dx\, x^{z-1}e^{-x}$. Thus, choosing $k_i$ such that $k_i+(s-3)/2\,{=}\,(d-3)/2$, one has
\begin{equation}
    \lim_{k_i\to\infty} \frac{\Gamma\left(k_i+\frac{s}{2}\right)}{\sqrt{\pi}\Gamma\left(k_i+\frac{s-1}{2}\right)}\left(1-t_i^2\right)^{k_i+\frac{s-3}{2}} = \delta(t_i).
\end{equation}
As a result, when $\mathcal{K}$ is infinitely differentiable, one has the following equivalence in the limit where all $k_i$'s are large,
\begin{equation}\label{eq:deep-proof-smooth}
    \Lambda_{\k} \sim \left(\prod_{i=1}^p\frac{|\mathbb{S}^{s-2}|}{|\mathbb{S}^{s-1}|}\left(\frac{1}{2}\right)^{k_i} \frac{\Gamma\left(\frac{s-1}{2}\right)}{\Gamma\left(k_i+\frac{s}{2}\right)}\right)  \mathcal{K}^{(\k)}\left( \bm{0}\right),
\end{equation}
which implies that, when $\mathcal{K}$ is infinitely differentiable, the eigenvalues decay exponentially or faster with the $k_i$.

Let us now consider the nonanalytic part of $\mathcal{K}$. There are three kinds of terms appearing in the singular expansion of depth-$3$ kernels (cf.~\autoref{ssec:proof-singular}):
\begin{itemize}
    \item[\emph{ia)}] $c_{+}\sum_{i} (1-t_i)^\nu$ near $t_i\,{=}\,+1$;
    \item[\emph{ib)}] $c_{-}\sum_{i} (1+t_i)^\nu$ near $t_i\,{=}\,-1$;
    \item[\emph{ii)}] $c_{+,\text{all}}\left( \sum_{i}(1-t_i)/p \right)^\nu$ near $t_i\,{=}\,+1$ for all $i$;
\end{itemize}
where the exponent $\nu$ is $1/2$ for the NTK and $3/2$ for the RFK. We will not consider terms of the kind \emph{ib)} explicitly, as the analysis is equivalent to that of terms of the kind \emph{ia)}. After replacing $t_i$ with $\cos(\theta_i)$, as in~\autoref{ssec:proof-fourier}, we get again $\sum_i |\theta_i|^{2\nu}$ and $\|\bm{\theta}\|^{2\nu}$ as leading nonanalytic terms. Therefore, we can rewrite the nonanalytic part of the kernel as follows,
\begin{equation}
    \mathcal{K}_\text{n.a.}(\bm{\theta}) = \sum_i f_1(|\theta_i|) + f_2(\|\bm{\theta}\|) + \mathcal{\tilde K}(\bm{\theta}),
\end{equation}
where $f_1$, $f_2$ are single-variable functions which behave as $\theta^{2\nu}$ near zero and have compact support, whereas $\mathcal{\tilde K}$ has a singular expansion near $\theta_i\,{=}\,0$ analogous to that of $\mathcal{K}$ but with leading nonanalyticities controlled by an exponent $\nu'\,{\geq}\nu+1$.

Let us look at the contribution to the eigenvalue $\Lambda_{\k}$ due to the term $f_1(|\theta_i|)$:
\begin{equation}\begin{aligned}\label{eq:deep-local-contribution}
    &\left(\prod_{j=1}^p\frac{|\mathbb{S}^{s-2}|}{|\mathbb{S}^{s-1}|}\int_{0}^{\pi} d\theta_j\left(\sin{(\theta_j)}\right)^{s-2}\,  P_{k_j,s}(\cos{(\theta_j)})\right)f_1(|\theta_i|)\\
    =  &\left(\prod_{j\neq i}\delta_{k_j,0}\right)\frac{|\mathbb{S}^{s-2}|}{|\mathbb{S}^{s-1}|}\int_{0}^{\pi} d\theta\left(\sin{(\theta)}\right)^{s-2}\,  P_{k_i,s}(\cos{(\theta)})f_1(|\theta|) = \left(\prod_{j\neq i}\delta_{k_j,0}\right)\left(f_1\right)_{k_1},
\end{aligned}\end{equation}
where we have introduced $\left(f_1\right)_{k}$ as the projection of $f_1(\theta)$ on the $k$-th Legendre polynomial. The asymptotic decay of $\left(f_1\right)_{k}$ is strictly related to the differentiability of $f_1$, which is in turn controlled by action of the Laplace-Beltrami operator $\Delta$ on $f_1$. As a function on the sphere $\mathbb{S}^{s-1}$, $f_1$ depends only on one angle, therefore the Laplace-Beltrami operator acts as follows,
\begin{equation}
    \Delta f_1(\theta) = \frac{1}{\sin{(\theta)}^{s-2}}\frac{d}{d\theta}\left(\sin{(\theta)}^{s-2} \frac{df_1}{d\theta}(\theta)\right) = f_1''(\theta) + (d-2)\frac{\cos{(\theta)}}{\sin{(\theta)}}f_1'(\theta).
\end{equation}
In terms of singular behavior near $\theta\,{=}\,0$, $f_1(\theta)\sim|\theta|^{2\nu}$ implies $\Delta f_1(\theta)\sim|\theta|^{2\nu-2}$, thus $\Delta^m f_1(\theta)\sim|\theta|^{2(\nu-m)}$. Given $\nu$, repeated applications of $\Delta$ eventually result in a function whose l2 norm on the sphere diverges. On the one hand,
\begin{equation}
    \| \Delta^{m/2}f_1 \|^2 = \int_0^\pi d\theta\, \sin^{d-2}{(\theta)} f_1(\theta)\Delta^m f_1(\theta).
\end{equation}
The integrand behaves as $|\theta|^{d-2+4\nu-2m}$ near $0$, thus the integral diverges for $m\geq 2\nu + (d-1)/2$. On the other hand, from~\autoref{eq:deep-slap-eigvals},
\begin{equation}
    \| \Delta^{m/2}f_1 \|^2 = \sum_k \mathcal{N}_{k,s} \left(k(k+s-2)\right)^m |(f_1)_k|^2.
\end{equation}
As $\mathcal{N}_{k,s}\sim k^{s-2}$ and the sum must converge for $m\,{<}\,2\nu + (d-1)/2$ and diverge otherwise, $(f_1)_k\sim k^{-2\nu -(s-1)}$. The projections of all the other terms in $\mathcal{K}$ on Legendre polynomials of one of the $p$ angles $\theta_i$ display a faster decay with $k$, therefore the above results imply the asymptotic scaling of local eigenvalues. Notice that such scaling matches with the result of~\cite{bietti2021deep}, which was obtained with a different argument.

Finally, let us look at the contribution to the eigenvalue $\Lambda_{\k}$ due to the term $f_2(\|\bm{\theta}\|)$:
\begin{equation}\begin{aligned}\label{eq:deep-global-contribution}
    &\left(\prod_{j=1}^p\frac{|\mathbb{S}^{s-2}|}{|\mathbb{S}^{s-1}|}\int_{0}^{\pi} d\theta_j\left(\sin{(\theta_j)}\right)^{s-2}\,  P_{k_j,s}(\cos{(\theta_j)})\right)f_2(\|\bm{\theta}\|) = \left(f_2\right)_{\k},
\end{aligned}\end{equation}
where we have introduced $\left(f_2\right)_{\k}$ as the projection of $f_2(\|\bm{\theta}\|)$ on the multi-Legendre polynomial with multi-degree $\k$. The asymptotic decay of $\left(f_2\right)_{k}$ is again related to the differentiability of $f_2$, controlled by the action of the multi-sphere Laplace-Beltrami operator $\Delta_{p,s}$ in~\autoref{eq:deep-mslap-eigvals}. As $f_2$ depends only on one angle per sphere,
\begin{equation}
    \Delta_{p,s} f_2(\|\bm{\theta}\|) = \sum_{i=1}^p\left( \partial^2_{\theta_i} f_2(\|\bm{\theta}\|) + (s-2)\frac{\cos{(\theta_i)}}{\sin{(\theta_i)}}\partial_{\theta_i} f_2(\|\bm{\theta}\|)\right).
\end{equation}
Further simplifications occur since $f_2$ depends only on the norm of $\bm{\theta}$. In terms of the singular behavior near $\|\bm{\theta}\|\,{=}\,0$, $f_2\sim \|\bm{\theta}\|^{2\nu}$ implies $\Delta_{p,s}^m f_2 \sim \|\bm{\theta}\|^{2(\nu-m)}$, thus
\begin{equation}
    \| \Delta_{p,s}^{m/2}f_2 \|^2 = \int_{[0,\pi]^p} d^p \bm{\theta}\, \prod_{i=1}^p\left(\sin^{s-2}{(\theta_i)}\right) f_2(\|\bm{\theta}\|)\Delta_{p,s}^m f_2(\|\bm{\theta}\|) < +\infty
\end{equation}
requires $m<2\nu + p(s-1)/2$ (compare with $m<2\nu + (s-1)/2$ for the local contributions). Therefore, one has
\begin{equation}\label{eq:deep-ms-condition}
    \| \Delta_{p,s}^{m/2}f_1 \|^2 = \sum_{\k} \left(\prod_{i=1}^p \mathcal{N}_{k_i,s}\right)\left(\sum_{i=1}^p k_i(k_i+s-2)\right)^m |(f_2)_{\k}|^2 < +\infty\quad \forall\,m<2\nu + p(s-1)/2,
\end{equation}
while the sum diverges for $m\geq 2\nu + p(s-1)/2$. In addition, since $f_2$ is a radial function of $\bm{\theta}$ which is homogeneous (or scale-invariant) near $\|\bm{\theta}\|\,{=}\,0$, $(f_2)_{\k}$ can be factorised in the large-$\|\k\|$ limit into a power of the norm $\|\k\|^\alpha$ and a finite angular part $\mathcal{C}(\k/\|\k\|)$. By plugging the factorization into~\autoref{eq:deep-ms-condition}, we get
\begin{equation}
    (f_2)_{\k} \sim \mathcal{C}(\k/\|\k\|) \|\k\|^{-2\nu -p(s-1)}, \quad \sum_{\k,\|\k\|=k} \left( \left(\prod_{i=1}^p (k_i/k)^{s-2}\right) \mathcal{C}(\k/\|\k\|)^2 \right) < +\infty
\end{equation}
The projections of all the other terms in $\mathcal{K}$ on multi-Legendre polynomials display a faster decay with $\|\k\|$, therefore the above results imply the asymptotic scaling of global eigenvalues.

\subsection{General depth}\label{ssec:proof-general}

The generalization to arbitrary depth is trivial once the depth-$3$ case is understood. For global and $s_1$-local eigenvalues, the analysis of the previous section carries over unaltered. All the other intermediate sectors correspond to the other terms singular expansion of the kernel: from~\autoref{ssec:proof-singular}, these terms can be written as
\begin{equation}\label{eq:deep-proof-final}
    \frac{c}{p_L}\frac{1}{\left(\displaystyle\prod_{l' < l'' \leq L}s_{l''}\right)} \sum_{i_{{l'+1}\to{L+1}}}\left( \frac{1}{\left(\displaystyle\prod_{2 \leq l'' \leq l'}s_{l''}\right)} \sum_{i_{2\to{l'}}} \left(1-t_{i_{2\to L+1}}\right) \right)^{\nu},
\end{equation}
for some $l'=2,\dots,L-1$ and fractional $\nu$. In practice, this term is a sum over the $p_{l'}\,{=}\,p_L\prod_{l' < l'' \leq L}s_{l''}$ meta-patches of $\bm{t}$ having size $s_{2\to l'}:=\prod_{2 \leq l'' \leq l'}s_{l''}$. Each summand is the fractional power $\nu$ of the average of the $t_i$'s within a meta-patch. When plugging such term into~\autoref{eq:deep-proof-eigvals}, the integrals over the $t_i$'s which do not belong to that meta-patch yield Kronecker deltas for the corresponding $k_i$'s. The integrals over the $t_i$'s within the meta-patch, instead, can be written as in \autoref{eq:deep-global-contribution} with the product and the norm restricted over the elements of that meta-patch, i.e., $\|\bm{\theta}\| \to \left(\sum_{i_{2 \to l'}} \theta_{i_{2\to L+1}}^2\right)^{1/2}$. Therefore, the scaling of the eigenvalue with $k$ is given again by~\autoref{eq:deep-proof-final}, but with $p$ replaced by the size of the meta-patch $\prod_{2 \leq l'' \leq l'}s_{l''}$, so that the effective dimension of~\autoref{eq:deep-effective-dim} appears at the exponent.

\section{Generalization bounds for kernel regression and spatial adaptivity}\label{app:deep-minimax}

This appendix provides an introduction to classical generalization bounds for kernel regression and extends \autoref{co:adaptivity} to patches on the hypersphere.

\subsection{Classical generalization bounds}

\paragraph{Rademacher bound.} Consider the regression setting detailed in \autoref{sec:deep-adaptivity} of the main text. First, assume that the target function $f^*$ belongs to the RKHS $\mathcal{H}$ of the kernel $\mathcal{K}$. Then, without further assumptions on $\mathcal{K}$, we have the following dimension-free bound on the excess risk, based on Rademacher complexity \cite{bach2021learning},
\cite{bietti2022approximation},
\begin{equation}
    \testerr(\lambda, P) - \testerr(f^*) 
    \leq \mathcal{C} \, \| f^* \|_{\mathcal{H}} \sqrt{\frac{{\mathrm{Tr}}(\mathcal{T_K})}{P}},
\end{equation}
where $\mathcal{T_K}$ is the integral operator associated to $\mathcal{K}$. For a hierarchical kernel, having a target with more power in the local sectors can result in a smaller $\| f^* \|_{\mathcal{H}}$, hence a smaller excess risk. However, this gain is only a constant factor in terms of sample complexity and, more importantly, being in the RKHS requires an order of smoothness which typically is of the order of the dimension, which is a very restrictive assumption in high-dimensional settings. 

\paragraph{Source-capacity bound.} The previous result can be extended by including more details about the kernel and the target function. In particular, Proposition 7.2 in \cite{bach2021learning}
states that, for $f^*$ in the closure of $\mathcal{H}$, regularization $\lambda \leq 1$ and $P \geq \frac{5}{\lambda}(1+\log(1/\lambda))$, one has
\begin{align}\label{eq:deep-bach-risk-decomposition}
    \testerr(\lambda, P) - \testerr(f^*) \leq &16 \, \frac{\sigma^2}{P} \, {\mathrm{Tr}}\left((\mathcal{T_K} + \lambda I)^{-1} \mathcal{T_K} \right) \nonumber \\
    &+ 16 \inf_{f \in \mathcal{H}}\left\{ \| f-f^* \|_{L_2}^2 + \lambda\| f \|_{\mathcal{H}}^2\right\} + \frac{24}{P^2} \, \| f^* \|_{L_{\infty}},
\end{align}
where $\sigma^2$ bounds the conditional variance of the labels, i.e. ${\mathbb{E}_{(\x,y)\sim p}\left[\left(y-f^*(\x)\right)^2\,|\,\x\right] < \sigma^2}$. 

Then, let us consider the following standard assumptions in the kernel literature \cite{caponnetto2007optimal},
\begin{align}
    \text{capacity: }&\text{Tr}\left(\mathcal{T}_{\mathcal{K}}^{1/\alpha}\right) = \sum_{\k\geq\bm{0}} \sum_{\bm{\ell}} (\Lambda_{\k})^{1/\alpha} < +\infty,\nonumber\\
    \text{source: }&\norm{T_{\mathcal{K}}^{\frac{1-r}{2}}f^*}_{\mathcal{H}}^2 = \sum_{\k\geq\bm{0}} \sum_{\bm{\ell}} (\Lambda_{\k})^{-r} (f^*_{\k,\bm{\ell}})^2 < +\infty.
\end{align}
In short, the first assumption characterizes the `size' of the RKHS (the larger $\alpha$, the smaller the number of functions in the RKHS), while the second assumption defines the regularity of the target function relative to that of the kernel (when $r=1$, $f^*\in\mathcal{H}$; when $r<1$, $f^*$ is less smooth; when $r>1$, $f^*$ is smoother). Combining these assumptions with \autoref{eq:deep-bach-risk-decomposition}, one gets
\begin{equation}
    \testerr(\lambda, P) - \testerr(f^*)
    \leq 16 \, \frac{\sigma^2}{P} \, \mathcal{C}_1 \lambda^{-1/\alpha} + 16 \, \mathcal{C}_2 \, \lambda^r +  \frac{24}{P^2} \, \| f^* \|_{L_{\infty}}.
\end{equation}
Optimizing for $\lambda$ results in
\begin{equation}
    \lambda_P
    = \left( \frac{\mathcal{C}_1 \sigma^2}{ \alpha \, r \, \mathcal{C}_2 \, P} \right)^{\frac{\alpha}{\alpha r + 1}},
\end{equation}
and the bound becomes
\begin{equation}
    \testerr(\lambda_P, P) - \testerr(f^*)
    \lesssim \mathcal{C}_2^{\frac{2}{\alpha r + 1}} \left( \frac{\mathcal{C}_1 \sigma^2}{P} \right)^{\frac{\alpha r}{\alpha r + 1}} + \frac{1}{P^2} \, \| f^* \|_{L_{\infty}}.
\end{equation}
Finally, when $r>(\alpha-1)/\alpha$, $P \geq \frac{5}{\lambda_P}(1+\log(1/\lambda_P))$ is always satisfied for $P$ large enough.

\subsection{Comparison with norm-based guarantees} 

A recent line of research has introduced norm-based generalization bounds for neural networks, which aim to bound the Rademacher complexity by utilizing the norm of the weight matrices, e.g.,~\citet{neyshabur2015norm}. Specifically, these bounds apply standard $O(1/\sqrt{P})$ upper bounds of the generalization gap via the Rademacher complexity (see, e.g,~\citet{mohri2018foundations}), followed by a norm-based bound on the Rademacher complexity. These results extend even outside the kernel limit considered in our present work and have also been applied to convolutional architectures~\cite{galanti2023norm}.\looseness=-1 

However, in contrast to our analysis, these bounds notably yield vacuous predictions in the overparameterized regime -- which is the regime relevant for practical applications -- and can even exhibit an anti-correlation with generalization performance~\cite{jiang2019fantastic}. Additionally, their application necessitates knowledge of the weight matrix norms post-training, which currently remains analytically inaccessible.

\subsection{Proof of \autoref{co:adaptivity} with patches on the hypersphere}

\begin{corollary}[Adaptivity to spatial structure]\label{co:adaptivity-app}

Let $T_{\mathcal{K}}$ be the integral operator of the kernel of a hierarchical deep CNN as in~\autoref{th:eig-scaling}. Then: \emph{i)} the \emph{capacity} exponent $\alpha$ is controlled by the largest sector of the spectrum, i.e. 
\begin{equation}
  {\mathrm{Tr}}\left(\mathcal{T}_{\mathcal{K}}^{1/\alpha}\right) < +\infty \Leftrightarrow \alpha < 1 + 2\nu/ d_{\mathrm{eff}}(L);
\end{equation}
\emph{ii)} the \emph{source} exponent $r$ is controlled by the structure of the target function $f^*$, i.e., if there is $l\,{\leq}\,L$ such that $f^*$ depends only on some meta-patch $\x_{i_{l+1\to L+1}}$, then only the first $l$ sectors of the spectrum contribute to the source condition,
\begin{equation}
    \norm{T_{\mathcal{K}}^{\frac{1-r}{2}}f^*}_{\mathcal{H}}^2 = \sum_{l'=1}^l \sum_{i_{l'+1\to L+1}}\sum_{\substack{\k_{i_{l'+1\to L+1}}\\\bm{\ell}_{i_{l'+1\to L+1}}}}\left(\Lambda^{(l')}_{\k_{i_{l'+1\to L+1}}}\right)^{-r} \left(f^*_{\k_{i_{l'+1\to L+1}},\,\bm{\ell}_{i_{l'+1\to L+1}}}\right)^2.
\end{equation}
The same holds if $f^*$ is a linear combination of such functions. As a result, when $d_{\mathrm{eff}}(L)$ is large and $\alpha\to 1$, the decay of the error is controlled by the effective dimensionality of the target $d_{\mathrm{eff}}(l)$.
\end{corollary}

\emph{Proof.} The capacity condition ${\mathrm{Tr}}\left(\mathcal{T}_{\mathcal{K}}^{1/\alpha}\right) < +\infty$ is satisfied when the eigenvalues $\Lambda_\rho$ of $\mathcal{T}_{\mathcal{K}}$ decay with their rank as $\rho^{-\alpha}$. Let's start by computing this scaling for a depth-two kernel with filters of size $s$. The eigenvalues decay with $\k$ as
\begin{equation}
    \Lambda_\bk
    \sim \sum_{i=1}^p k_i^{-2\nu_S - (s-1)} \prod_{j \neq i} \delta_{k_j,0}.
\end{equation}
In order to take into account their algebraic multiplicity, we introduce the eigenvalue density $\mathcal{D}(\Lambda)$, whose asymptotic form for small eigenvalues is
\begin{align}
    \mathcal{D}(\Lambda) 
    &= \sum_{\bk,\,\bell} \delta(\Lambda - \Lambda_\bk) \nonumber \\
    &\sim \sum_\bk \left(\prod_{i=1}^p k_i^{s-2}\right) \delta\left(\Lambda - \sum_{i=1}^p k_i^{-2\nu-(s-1)} \prod_{j \neq i} \delta_{k_j,0}\right) \nonumber \\
    &\sim \sum_{i=1}^p  \sum_{k_i} k_i^{s-2} \delta\left(\Lambda - k_i^{-2\nu-(s-1)}\right) \nonumber \\
    &\sim \int_1^{\infty} dk  \, k^{s-2} \delta\left(\Lambda - k^{-2\nu-(s-1)}\right) \nonumber \\
    &\sim \Lambda^{-1-\frac{s-1}{2\nu+(s-1)}}.
\end{align}
Thus, the scaling of $\Lambda(\rho)$ can be determined self-consistently,
\begin{equation}
    \rho 
    = \int_{\Lambda(\rho)}^{\Lambda(1)} d\Lambda \, \mathcal{D}(\Lambda) \sim \Lambda(\rho)^{-\frac{s-1}{s\nu+(s-1)}} 
    \,\Rightarrow\, 
    \Lambda(\rho) 
    \sim \rho^{-1-\frac{2\nu}{s-1}}.
\end{equation}
Consider now a kernel of depth $L+1$ with filter sizes $(s_1, \dots, s_L)$ and $p_L=1$. For each sector $l$, one can compute the density of eigenvalues $\mathcal{D}_{(l)}(\Lambda)$. Depending on $s_1$, there are two different cases.

If $s_1=2$,
\begin{align}
    \mathcal{D}_{(l)}(\Lambda) 
    &= \sum_{\bk} \delta(\Lambda - \Lambda_\bk^{(l)}) \nonumber \\
    &\sim \sum_{i_{l+1\to L+1}} \sum_{\bk_{i_{l+1\to L+1}}} \delta\left(\Lambda - \mathcal{C}_{2,l} \, \|\bk_{i_{l+1\to L+1}}\|^{-2\nu -d_{\mathrm{eff}}(l)}\right) \nonumber \\
    &\sim \int_1^{\infty} dk  \, k^{d_{\mathrm{eff}}(l)-1} \delta\left(\Lambda - \mathcal{C}_{2,l} \, k^{-2\nu -d_{\mathrm{eff}}(l)}\right) \nonumber \\
    &\sim \Lambda^{-1-\frac{d_{\mathrm{eff}}(l)}{2\nu+d_{\mathrm{eff}}(l)}}.
\end{align}

If $s_1 \geq 3$,
\begin{align}
    \mathcal{D}_{(l)}(\Lambda) 
    &= \sum_{\bk,\,\bell} \delta(\Lambda - \Lambda_\bk^{(l)}) \nonumber \\
    &\sim \sum_{i_{l+1\to L+1}} \sum_{\substack{\bk_{i_{l+1\to L+1}},\\ \bell_{i_{l+1\to L+1}}}} \delta\left(\Lambda - \mathcal{C}_{s_1,l}\left(\frac{\k_{i_{l+1 \to L+1}}}{\|\bk_{i_{l+1\to L+1}}\|}\right) \, \|\bk_{i_{l+1\to L+1}}\|^{-2\nu -d_{\mathrm{eff}}(l)}\right) \nonumber \\
    &\sim \Lambda^{-1-\frac{d_{\mathrm{eff}}(l)}{2\nu+d_{\mathrm{eff}}(l)}}.
\end{align}

When summing over all layers $l$'s, the asymptotic behaviour of the total density of eigenvalues $\mathcal{D}(\Lambda) = \sum_l \mathcal{D}_{(l)}(\Lambda)$ is dictated by the density of the sector with the slowest decay, i.e. the last one. Hence,
\begin{equation}
    \mathcal{D}(\Lambda) 
    \sim \Lambda^{-1-\frac{d_{\mathrm{eff}}(L)}{2\nu+d_{\mathrm{eff}}(L)}}.
\end{equation}
Therefore, similarly to the shallow case, one finds self-consistently that the $\rho$-th eigenvalue of the kernel decays as
\begin{equation}
    \Lambda(\rho) 
    \sim \rho^{-1-\frac{2\nu}{d_{\mathrm{eff}}(L)}}.
\end{equation}
This proves that the capacity condition is controlled by the largest sector of the spectrum and $\alpha < 1 + 2\nu/ d_{\mathrm{eff}}(L)$.

Finally, we notice that, if $f^*$ depends only on a meta-patch $\x_{i_{l+1\to L+1}}$, all projections on eigenfunctions belonging to higher sectors are zero and hence
\begin{equation}
    \norm{T_{\mathcal{K}}^{\frac{1-r}{2}}f^*}_{\mathcal{H}}^2 = \sum_{l'=1}^l \sum_{i_{l'+1\to L+1}}\sum_{\substack{\k_{i_{l'+1\to L+1}}\\\bm{\ell}_{i_{l'+1\to L+1}}}}\left(\Lambda^{(l')}_{\k_{i_{l'+1\to L+1}}}\right)^{-r} \left(f^*_{\k_{i_{l'+1\to L+1}},\,\bm{\ell}_{i_{l'+1\to L+1}}}\right)^2.
\end{equation}
Therefore, only the first $l$ sectors contribute to the source condition and the proof is concluded.

\section{Statistical mechanics of generalization in kernel regression}\label{app:deep-spectralbias}

In \cite{bordelon2020spectrum, canatar2021spectral}, the authors derived a heuristic expression for the average-case mean-squared error of kernel (ridge) regression with the replica method of statistical physics \cite{mezard1987spin}. Denoting with $\{\phi_\rho(\x),\; \Lambda_\rho\}_{\rho\geq1}$ the eigenfunctions and eigenvalues of the kernel and with $c_\rho$ the coefficients of the target function in this basis, i.e. $f^*(\x) = \sum_{\rho\geq 1} c_\rho \phi_\rho(\x)$, one has
\begin{equation}\label{eq:deep-bordelon1}
\testerr(\lambda, P) 
= \partial_{\lambda}\left(\frac{\kappa_{\lambda}(P)}{P}\right) \sum_\rho \frac{\kappa_\lambda(P)^2}{\left(P\Lambda_\rho + \kappa_\lambda(P)\right)^2} \, \mathbb{E}[c_\rho^2] ,\end{equation}
where $\lambda$ is the ridge and $\kappa(P)$ satisfies the implicit equation
\begin{equation}\label{eq:deep-bordelon2} 
\frac{\kappa_\lambda(P)}{P} 
= \lambda + \frac{1}{P}\sum_\rho \frac{\Lambda_\rho \kappa_\lambda(P)/P}{\Lambda_\rho + \kappa_\lambda(P)/P}.
\end{equation}
In short, the replica calculation used to obtain these equations consists in defining an energy functional $\mathcal{E}(f)$ related to the empirical MSE and assigning to the predictor $f$ a Boltzmann measure, i.e. $P(f) \propto
e^{-\beta E(f)}$. When $\beta \to \infty$, the measure concentrates around the minimum of $\mathcal{E}(f)$, which coincides with the minimizer of the empirical MSE. Then, since $\mathcal{E}(f)$ depends only quadratically on the projections $c_\rho$, computing the average over data that appears in the definition of the generalization error, reduces to computing Gaussian integrals. While non-rigorous, this method has been successfully used in physics -- to study disordered systems -- and in machine learning theory. In particular, the predictions obtained with \autoref{eq:deep-bordelon1} and \autoref{eq:deep-bordelon2} have been validated numerically for both synthetic and real datasets.

In \autoref{eq:deep-bordelon1}, $\kappa_\lambda(P)/P$ plays the role of a threshold: the modal contributions to the error tend to $0$ for $\rho$ such that $\Lambda_\rho \gg \kappa_\lambda(P)/P$, and to $\mathbb{E}[c_\rho^2]$ for $\rho$ such that $\Lambda_\rho \ll \kappa_\lambda(P)/P$. This is equivalent to saying that kernel regression can capture only the modes corresponding to the eigenvalues larger than $\kappa_\lambda(P)/P$ (see also~\cite{jacot2020implicit, jacot2020kernel}).

In the ridgeless limit $\lambda \to 0^+$, this threshold asymptotically tends to the $P$-th eigenvalue of the student, resulting in the intuitive picture presented in the main text. Namely, given $P$ training points, ridgeless regression learns the $P$ projections corresponding to the highest eigenvalues. In particular, assume that the kernel spectrum and the target function projections decay as power laws. Namely, $\Lambda_\rho \sim \rho^{-a}$ and $\mathbb{E}[{c_\rho}^2] \sim \rho^{-b}$, with $2a\,{>}\,b-1$. Furthermore, we can approximate the summations over modes with an integral by using the Euler-MacLaurin formula. Hence, we substitute the eigenvalues with their asymptotic limit $\Lambda_\rho = A\rho^{-a}$. Since, $\kappa_0(P)/P \to 0$ as $P\to \infty$, these two operations result in an error which is asymptotically independent of $P$. In particular,
\begin{align}
    \frac{\kappa_0(P)}{P} &= \frac{\kappa_0(P)}{P} \frac{1}{P} \left(\int_0^\infty \frac{A\rho^{-a} }{A\rho^{-a} + \kappa_0(P)/P} \, d\rho + O(1) \right) \nonumber \\
    &= \frac{\kappa_0(P)}{P} \frac{1}{P} \left( \left( \frac{\kappa_0(P)}{P} \right)^{-\frac{1}{a}}\int_0^\infty \frac{\sigma^{\frac{1}{a}-1}A^{\frac{1}{a}}a^{-1}}{1 + \sigma} \, d\sigma + O(1) \right).
\end{align}
Since the integration over $\sigma$ is finite and independent of $P$, we obtain that $\kappa_0(P)/P= O(P^{-a})$. Similarly, we find that the mode-independent prefactor $\partial_\lambda \left(\kappa_\lambda(P)/P\right)|_{\lambda=0} = O(1)$.

As a result, we have
\begin{equation}\label{eq:deep-error-scaling1} 
\testerr(P) \sim \sum_{\rho} \frac{P^{-2a}}{\left(A\rho^{-a}+P^{-a}\right)^2} \, \mathbb{E}[c_\rho^2].
\end{equation}
Following the intuitive argument about the thresholding action of $\kappa_0(P)/P \sim P^{-a}$, we can split the summation in \autoref{eq:deep-error-scaling1} into modes where $\Lambda_\rho\gg\kappa_0(P)/P$, $\Lambda_\rho \sim \kappa_0(P)/P$ and $\Lambda_\rho\ll\kappa_0(P)/P$,
\begin{equation}
\testerr(P) \sim \sum_{\rho \ll P} \frac{P^{-2a}}{\left(A\rho^{-a}\right)^2}\mathbb{E}[c_\rho^2] +\sum_{\rho \sim P} \frac{1}{2}\mathbb{E}[c_\rho^2] + \sum_{\rho \gg P} \mathbb{E}[c_\rho^2].
\end{equation}
Finally, \autoref{eq:deep-spectralbias} is obtained by noticing that, under the assumption on the decay of $\mathbb{E}[c_\rho^2]$, the contribution of the summation over $\rho \ll P$ is subleading in $P$, whereas the other two can be merged together.

\section{Examples}\label{app:deep-examples}

\subsection{Rates from spectral bias ansatz}\label{ssec:sb-app}

Consider a target function $f^*$ which only depends on the meta-patch $\x_{i_{l+1\to L+1}}$ and with square-integrable derivatives up to order $m$, i.e. $\|\Delta^{m/2} f^*\|^2<+\infty$, with $\Delta$ denoting the Laplace operator. Moreover, consider a hierarchical kernel of depth $L+1$ with filter sizes $(s_1, \dots, s_L)$ and $p_L=1$. We want to compute the asymptotic scaling of the error by using \autoref{eq:deep-spectralbias}, i.e.
\begin{equation}\label{eq:deep-spectral-app}
    \testerr(P) \sim \sum_{\k,\bm{\ell} \text{ s.t. } \Lambda_{\k}<\Lambda(P)} \lvert f^*_{\k,\bm{\ell}}\rvert^2.
\end{equation}
In \autoref{app:deep-minimax}, we showed that the $P$-th eigenvalue of the kernel $\Lambda(P)$ decays as
\begin{equation}
    \Lambda(P) 
    \sim P^{-1-\frac{2\nu}{d_{\mathrm{eff}}(L)}}.
\end{equation}
Since by construction the target function depends only on a meta-patch of the $l$-th sector, the only non-zero projections will be the ones on eigenfunctions of the first $l$ sectors. Thus, all the $\bk$'s corresponding to the sectors of layers with $l'>l$ do not contribute to the sum. In particular, the sum is dominated by the $\bk$'s of the largest sector and the set $\{\bk \text{ s.t. } \Lambda_{\bk} < \Lambda(P)\}$ is the set of $\bk_{i_{l+1\to L+1}}$'s with norm larger than $P^{\frac{2\nu + d_{\mathrm{eff}}(L)}{(2\nu + d_{\mathrm{eff}}(l))\, d_{\mathrm{eff}}(L)}}$. 

Finally, we notice that the finite-norm condition on the derivatives,
\begin{equation}
    \|\Delta^{m/2}f^* \|^2 = \sum_{\k} \left(\prod_{i=1}^p \mathcal{N}_{k_i,s}\right) \left(\sum_{i=1}^p k_i(k_i+s-2)\right)^m |f^*_{\k,\bm{\ell}}|^2 < + \infty,
\end{equation}
implies $|f^*_{\k,\bm{\ell}}|^2 \lesssim \|\k\|^{-2m-d_{\mathrm{eff}}(L)}$ (see \autoref{ssec:proof-depth2}).

Hence, plugging everything in \autoref{eq:deep-spectral-app} we find
\begin{equation}
    \testerr(P) 
    \sim P^{-\frac{2m}{2\nu+d_{\text{eff}}(l)} \frac{2\nu+d_{\mathrm{eff}}(L)}{d_{\mathrm{eff}}(L)}}.
\end{equation}

\section{Numerical experiments}\label{app:deep-numerics}

\subsection{Experimental setup}

Experiments were run on a high-performance computing cluster with nodes having Intel Xeon Gold processors with 20 cores and 192 GB of DDR4 RAM. All codes are written in PyTorch~\cite{paszke2019pytorch}. 

\subsection{Teacher-student learning curves}

In order to obtain the learning curves, we generate $P+P_{\mathrm{test}}$ random points uniformly distributed on the product of hyperspheres over the patches. We use $P \in \{128,\; 256,\; 512,\; 1024,\; 2048,\; 4096,\; 8192\}$ and $P_{\mathrm{test}}=8192$. For each value of $P$, we sample a Gaussian random field with zero mean and covariance given by the teacher kernel. Then, we compute the kernel regression predictor of the student kernel, and we estimate the generalization error as the mean squared error of the obtained predictor on the $P_{\mathrm{test}}$ unseen example. The expectation over the teacher randomness is obtained by averaging over 16 independent sets of random input points and realizations of the Gaussian random fields. As teacher and student kernels, we use the analytical forms of the neural tangent kernels of hierarchical convolutional networks, with different combinations of depths and filter sizes.

\begin{figure}

    \centering
    \includegraphics[width=0.48\textwidth]{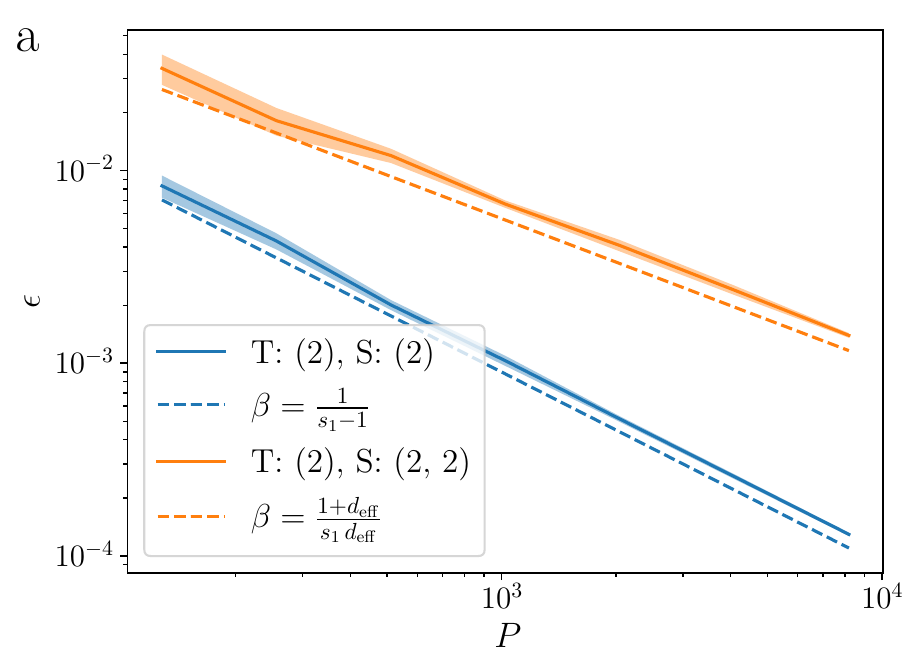}
    \includegraphics[width=0.48\textwidth]{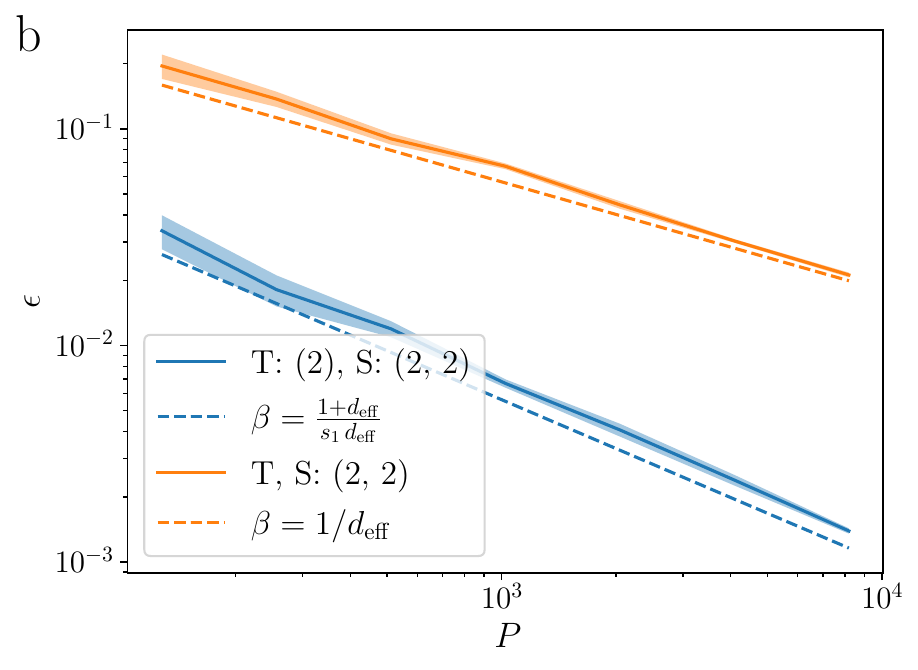}
    
    \caption{Learning curves for deep convolutional NTKs ($\nu=1/2$) in a teacher-student setting. \textit{(a)} Depth-two teachers learned by depth-two (matched) and depth-three (mismatched) students. Neither of these students is cursed by the input dimension. \textit{(b)} Depth-three students learning depth-two and depth-three teachers. These students are cursed only in the second case. The numbers inside brackets are the sequence of filter sizes of the kernels. Solid lines are the results of experiments averaged over 16 realizations with the shaded areas representing the empirical standard deviations. The predicted asymptotic scaling $\testerr \sim P^{-\beta}$ are reported as dashed lines.}
    
    \label{fig:learning_curves_app}
    
\end{figure}

\paragraph{Depth-two and depth-three architectures.} \autoref{fig:learning_curves_app} reports the learning curves of depth-two and depth-three kernels with binary filters at all layers. Depth-three students defeat the curse of dimensionality when learning depth-two teachers, achieving a similar performance of depth-two students matched to the teacher's structure. However, as we predict, these students encounter the curse of dimensionality when learning depth-three teachers.

\begin{figure}

    \centering
         \includegraphics[width=0.48\textwidth]{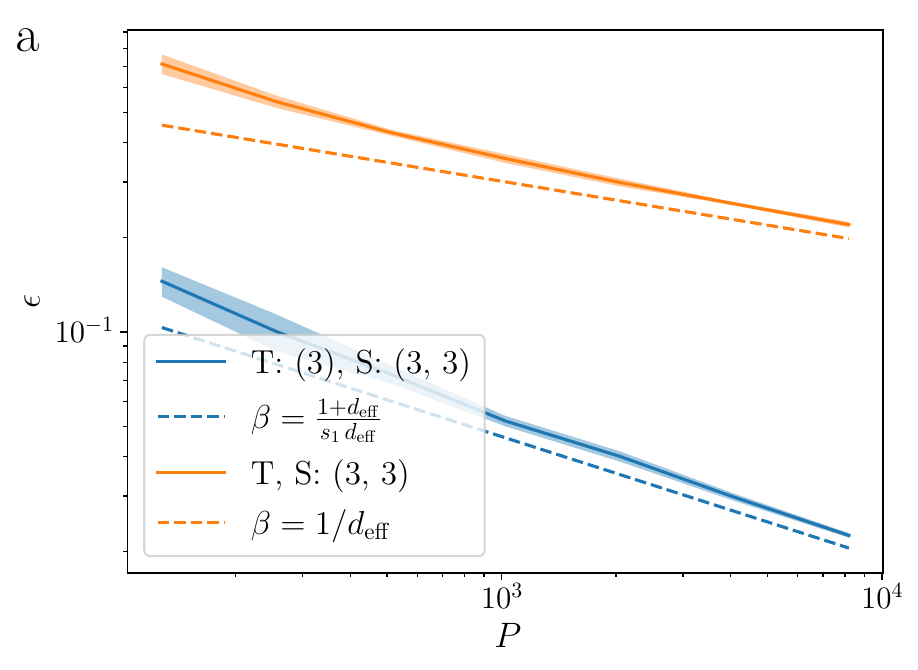}
         \includegraphics[width=0.48\textwidth]{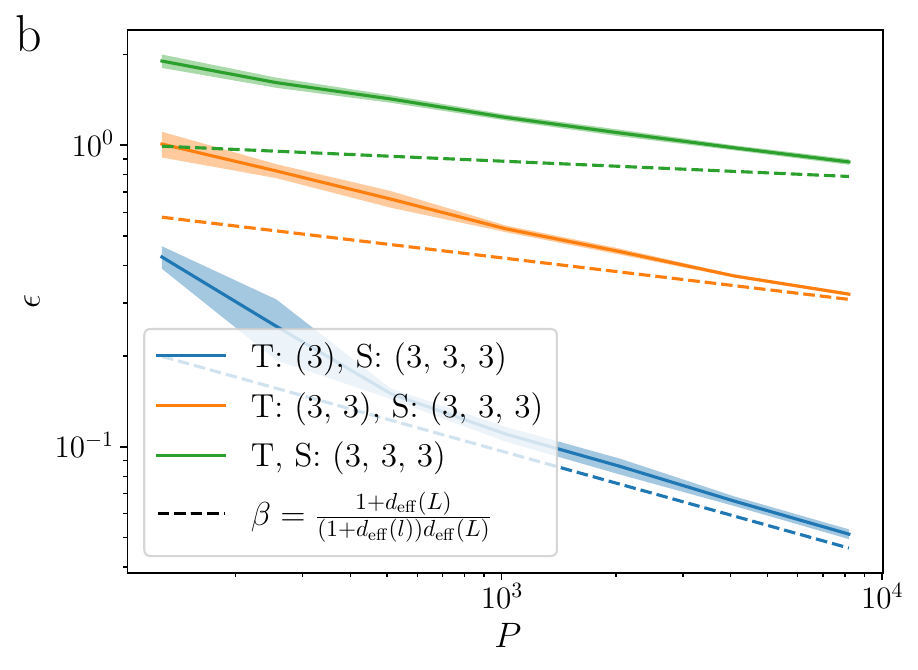}
         
    \caption{Learning curves for deep convolutional NTKs ($\nu=1/2$) with filters of size 3 in a teacher-student setting. \textit{(a)} Depth-three students learning depth-two and depth-three teachers. These students are cursed only in the second case. \textit{(b)} Depth-three models are cursed by the effective input dimensionality. The numbers inside brackets are the sequence of filter sizes of the kernels. Solid lines are the results of experiments averaged over 16 realizations with the shaded areas representing the empirical standard deviations. The predicted asymptotic scaling $\testerr \sim P^{-\beta}$ are reported as dashed lines.}
    
    \label{fig:ternary_app}
    
\end{figure}

\paragraph{Ternary filters.} \autoref{fig:ternary_app} reports the learning curves for kernels with 3-dimensional filters and confirms our predictions in the $s_1 \geq 3$ case. 

\paragraph{Comparison with the noisy and optimally-regularized case.} Panel (a) of \autoref{fig:noise_app} compares the learning curves obtained in the optimally-regularized and ridgeless cases for noisy and noiseless data, respectively. The first case corresponds to the setting studied in \cite{caponnetto2007optimal}, in which the source-capacity formalism applies. In contrast with the second setting -- which is the one used in the teacher-student scenarios and where it holds the correspondence between kernel methods and neural networks -- \textit{i)} we add to the labels a Gaussian random noise with standard deviation $\sigma=0.1$, \textit{ii)} for each $n$, we select the ridge resulting in the best generalization performance. We observe that the decay obtained in the bound derived from the source-capacity conditions is exactly the one found numerically, i.e., the rate of the bound is tight. As a further check, panel (b) shows that the optimal ridge decays as prescribed.

\begin{figure}

    \centering
         \includegraphics[width=0.48\textwidth]{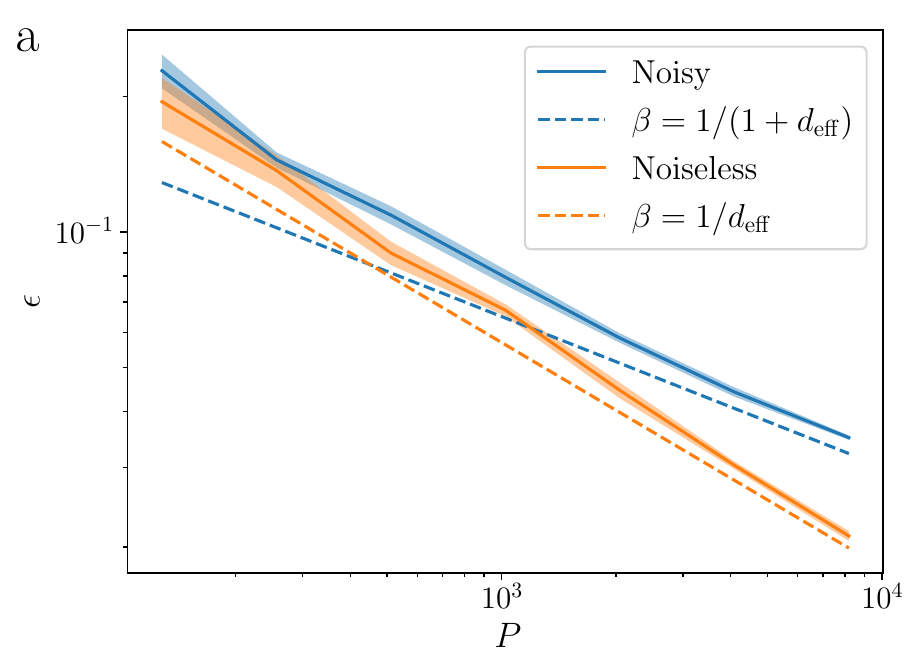}
         \includegraphics[width=0.48\textwidth]{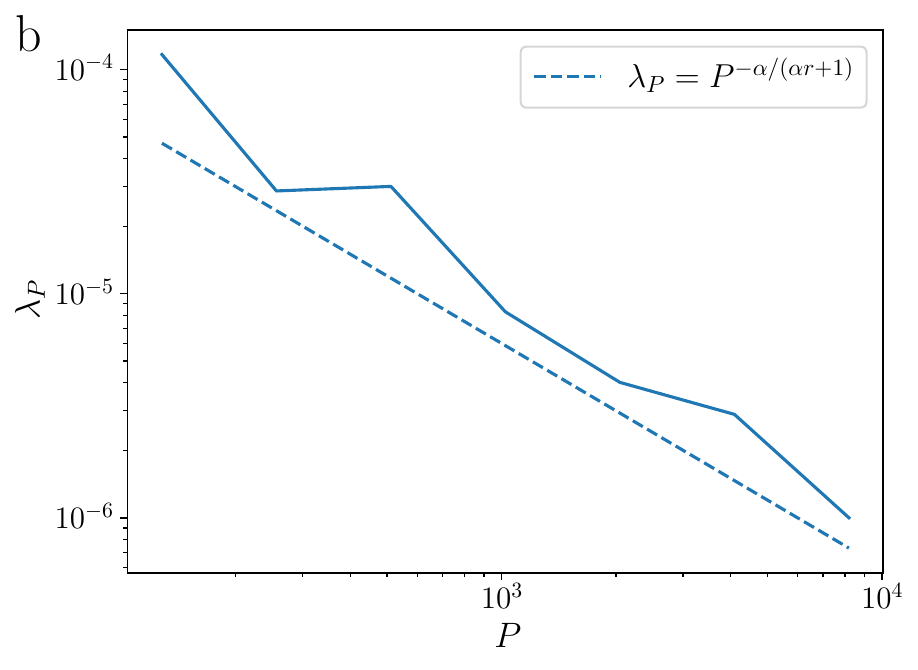}
         
    \caption{Noisy (optimally-regularized) vs noiseless (ridgeless) learning curves for depth-three deep convolutional NTKs ($\nu=1/2$) in a teacher-student setting. \textbf{a.} Comparison between the learning curves in the noisy and noiseless case. Dashed lines represent the rates predicted with source-capacity bounds and replica calculations, respectively. Shaded areas represent the empirical standard deviations. \textbf{b.} Decay of the optimal ridge with the number of training points.}
    
    \label{fig:noise_app}
    
\end{figure}

\subsection{Illustration of different teacher-student scenarios}

In this subsection, we comment on the results obtained in the different teacher-student scenarios of \autoref{fig:learning_curves_main}, panel (a), and \autoref{fig:learning_curves_app}, panel (a). To ease notation, in the following we always consider the NTK for both teacher and student kernels, i.e., smoothness exponent $\nu_T = \nu_S = 1/2$. However, we point out that when the teacher kernel is a hierarchical RFK ($\nu_T=3/2$), the target function corresponds to the output of an infinitely-wide, deep hierarchical network at initialization\footnote{See, e.g, \citet{lee2017deep} for the equivalence between infinitely-wide networks and Gaussian random fields with covariance given by the RFK.}. The error rates are obtained from \autoref{eq:deep-t-depth-one-s-depth-two}, after setting the smoothness exponent $m=\nu_T$ (the smoothness exponent of the teacher covariance kernel).

The first case we consider consists of one-hidden-layer convolutional teacher (left) and student (right) kernels.

\begin{figure*}[ht!]
    \centering
    \begin{minipage}[c]{0.60\textwidth}
            \includegraphics[width=0.4\textwidth]{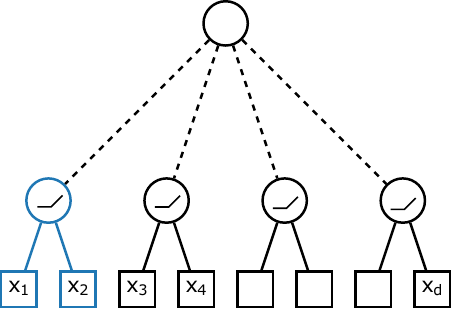}
        \hspace{1cm}
            \includegraphics[width=0.4\textwidth]{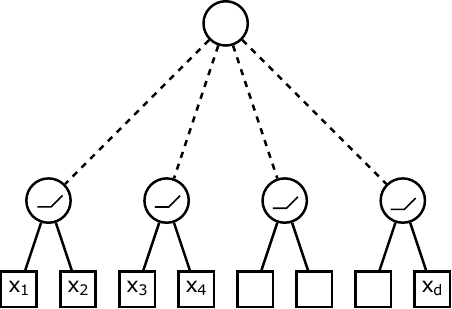}
    \end{minipage}
    \hfill
    \begin{minipage}[c]{0.3\textwidth}
    \centering
        $$\testerr(P) \sim P^{-\frac{1}{s_1-1}}$$
    \end{minipage}
    \addtocounter{figure}{-1}
\end{figure*}

As highlighted in blue, the output of the teacher is a linear combination (dashed lines indicate the linear output weights) of $s_1$-dimensional functions of the input patches. If the structure of the student is matched to the one of the teacher, the learning problem becomes effectively $(s_1-1)$-dimensional and the error decays as $P^{-1/(s_1-1)}$, instead of $P^{-1/d_{\mathrm{eff}}}$, with $d_{\mathrm{eff}}$ the total input dimension with the number of spherical constraints subtracted (one per patch). Notice that the role of the student's structure, i.e., the algorithm, is as crucial as the role of the teacher, i.e., the task. Indeed, using a fully-connected student with no prior on the task's locality would result in an error's decay cursed by dimensionality. However, in contrast to fully-connected students, shallow convolutional students are only able to learn tasks with the same structure. In particular, any task entailing non-linear interactions between patches -- which are arguably crucial in order to learn image data -- belongs to their null space.

As we illustrated in the main text, to solve this strong constraint on the hypothesis space, one has to consider deep convolutional architectures. In particular, consider the same shallow teacher of the previous paragraph (left) learned by a depth-four convolutional student (right).

\begin{figure*}[ht!]
    \centering
    \begin{minipage}[c]{0.60\textwidth}
            \includegraphics[width=0.4\textwidth]{Figures/widecnn/app_dag_0.pdf}
        \hspace{1cm}
            \includegraphics[width=0.4\textwidth]{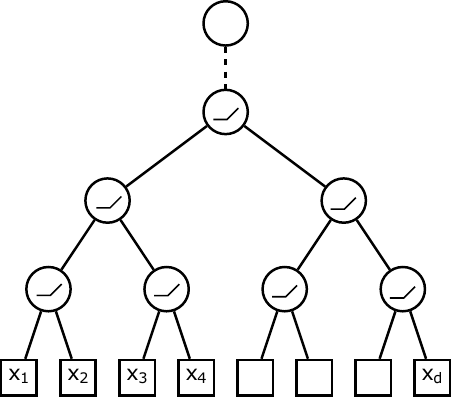}
    \end{minipage}
    \hfill
    \begin{minipage}[c]{0.3\textwidth}
    \centering
        $$\testerr(P) \sim P^{-\frac{1}{s_1} \frac{1+d_{\mathrm{eff}}(3)}{d_{\mathrm{eff}}(3)}}$$
    \end{minipage}
    \addtocounter{figure}{-1}
\end{figure*}

Remarkably, this student is able to learn the teacher without being cursed by input dimensionality. Indeed, as the number of patches diverges, the error decay asymptotes to $P^{-1/s_1}$. This rate is slightly worse than the one obtained by the student matched with the teacher, which is proven to be the Bayes-optimal case, but far from being cursed. Intuitively, this fast rate is obtained because the student eigenfunctions of the first sector, i.e., constant outside a single patch, correspond to large eigenvalues and bias the learning dynamics towards $s_1$-local functions. Yet, this student is also able to represent functions that are considerably more complex.

Now consider a depth-three teacher (left) learned by a depth-four student (right).

\begin{figure*}[ht!]
    \centering
    \begin{minipage}[c]{0.60\textwidth}
            \includegraphics[width=0.4\textwidth]{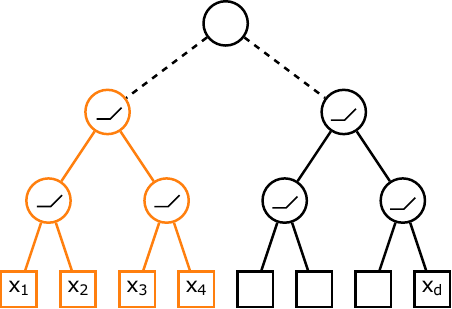}
        \hspace{1cm}
            \includegraphics[width=0.4\textwidth]{Figures/widecnn/app_dag_3.pdf}
    \end{minipage}
    \hfill
    \begin{minipage}[c]{0.3\textwidth}
    \centering
        $$\testerr(P) \sim P^{-\frac{1}{1+d_{\mathrm{eff}}(2)} \frac{1+d_{\mathrm{eff}}(3)}{d_{\mathrm{eff}}(3)}}$$
    \end{minipage}
    \addtocounter{figure}{-1}
\end{figure*}

As highlighted in orange, the output of the teacher is a linear combination of a composition of non-linear functions acting on patches and coupling them. In this setting, the error decay is controlled by the effective dimension of the second layer. In fact, when the number of patches diverges, the error decay asymptotes to $P^{-1/d_{\mathrm{eff}}(2)}$. In general, this behavior is a result of what we called `adaptivity to the spatial structure' of the target.

Finally, consider both teacher and student with the complete hierarchy, i.e., the receptive fields of the neurons in the penultimate layers coincide with the full input.  

\begin{figure*}[ht!]
    \centering
    \begin{minipage}[c]{0.60\textwidth}
            \includegraphics[width=0.4\textwidth]{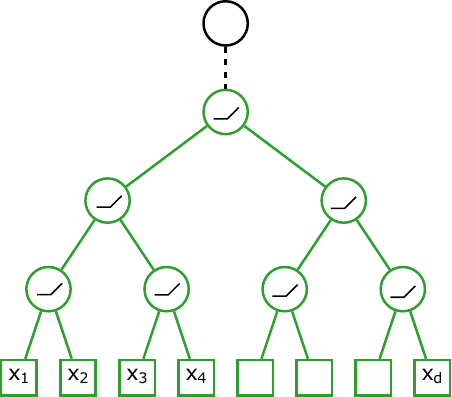}
        \hspace{1cm}
            \includegraphics[width=0.4\textwidth]{Figures/widecnn/app_dag_3.pdf}
    \end{minipage}
    \hfill
    \begin{minipage}[c]{0.3\textwidth}
    \centering
        $$\testerr(P) \sim P^{-\frac{1}{d_{\mathrm{eff}}(3)}}$$
    \end{minipage}
    \addtocounter{figure}{-1}
\end{figure*}

In this case, we show that the error decays as $P^{-1/d_{\mathrm{eff}}(3)}$, i.e. the rate is cursed by the input dimension. The physical meaning of this result is that the hierarchical structure we are considering is still too complex and cannot be learned efficiently. In other words, these hierarchical convolutional networks are excellent students, since they can adapt to the spatial structure of the task, but bad teachers, since they generate global functions that are too complex to be learned efficiently. 

\subsection{Extensions to different normalizations and overlapping patches}\label{app:deep-extensions}

This section investigates the robustness of our results to changes in the input distribution, i.e., for data outside the multisphere ${\sf M}^p\mathbb{S}^{s-1}$, and relaxes the non-overlapping patches assumption.

\textbf{Inputs in $\mathbb{R}^d$.} While our analysis requires that each patch of the input data is normalized to lie on a unit sphere, this normalization is not the standard one used for neural networks. Therefore, in this section, we investigate the robustness of our predictions to the data distribution. In particular, we consider data uniformly distributed in the unit hypercube, i.e., $\x\in[0,1]^d$, and data with standard Gaussian distribution, i.e., $\x \sim \mathcal{N}(\mathbf{0},\mathbf{I}_d)$. First, we extend the definition of the RFK and NTK to inputs in $\mathbb{R}^d$.

\begin{definition}[RFK and NTK of hierarchical CNNs for inputs in $\mathbb{R}^d$]\label{def:hierarchical-kernels-Rd}
Let $\x,\y\in \mathbb{R}^d$. Denote tuples of the kind $i_{l} i_{l+1} \dots i_{m}$ with $i_{l \to m}$ for $m\,{\geq}\,l$. For $m\,{<}\,l$, $i_{l\to m}$ denotes the empty tuple. For each tuple $i_{2\to L+1}$ and $s$ a divisor of $d$, denote with $t_{i_{2\to L+1}}$ the angle between the $s$-dimensional patches of $\x$ and $\y$ identified by the same tuple, i.e.
\begin{equation}
    t_{i_{2\to L+1}} = \frac{\x_{i_{2\to L+1}}^\top \y_{i_{2\to L+1}}}{\|\x_{i_{2\to L+1}}\| \|\y_{i_{2\to L+1}}\|}
\end{equation}
For $1\,{\leq}\,l\,{\leq}\,L+1$, denote with $\left\lbrace \x_{i_{2\to L+1}},\,\y_{i_{2\to L+1}} \right\rbrace_{i_{2\to l}}$ the sequence of patches obtained by letting the indices of the tuple $i_{2\to l}$ vary in their respective range. Consider a hierarchical CNN with filter sizes $(s_1,\dots,s_L)$, $p_L\,{\geq}\,1$ and all the weights $w^{\scriptstyle(1)}_{\scriptstyle h,i}, w^{\scriptstyle(l)}_{\scriptstyle h,h',i}, w^{\scriptstyle (L+1)}_{\scriptstyle h,i}$ initialized as Gaussian random numbers with zero mean and unit variance. 

\textbf{RFK.} The corresponding RFK (or covariance kernel) can be obtained recursively as follows. With $\kappa_1(t)\,{=}\,\left((\pi-\arccos{t})\, t +\sqrt{1-t^2}\right)/\pi$,
\begin{align}
&\mathcal{K}_{\mathrm{RFK}}^{(1)} (\x_{i_{2\to L+1}},\,\y_{i_{2\to L+1}}) =  \|\x_{i_{2\to L+1}}\| \|\y_{i_{2\to L+1}}\| \, \kappa_1(t_{i_{2\to L+1}}); \nonumber\\
&\mathcal{K}_{\mathrm{RFK}}^{(l)}\left(\left\lbrace \x_{i_{2\to L+1}},\,\y_{i_{2\to L+1}} \right\rbrace_{i_{2\to l}}\right) = \sqrt{\frac{1}{s_l}\sum_{i_l}\|\x_{i_{l\to L+1}}\|^2} \sqrt{\frac{1}{s_l}\sum_{i_l}\|\y_{i_{l\to L+1}}\|^2} \nonumber \\
&\qquad\qquad\qquad\qquad\qquad\qquad \times \kappa_1\left(\frac{\frac{1}{s_l}\sum_{i_l} \mathcal{K}_{\mathrm{RFK}}^{(l-1)}\left( \left\lbrace \x_{i_{2\to L+1}},\,\y_{i_{2\to L+1}} \right\rbrace_{i_{2\to l-1}}\right)}{\sqrt{\frac{1}{s_l}\sum_{i_l}\|\x_{i_{l\to L+1}}\|^2} \sqrt{\frac{1}{s_l}\sum_{i_l}\|\y_{i_{l\to L+1}}\|^2}}\right); \nonumber\\
&\mathcal{K}_{\mathrm{RFK}}^{(L+1)} \left( \left\lbrace \x_{i_{2\to L+1}},\,\y_{i_{2\to L+1}}\right\rbrace_{i_{2\to L+1}} \right)= \frac{1}{p_L}\sum_{i_{L+1}=1}^{p_L} \mathcal{K}_{\mathrm{RFK}}^{(L)}\left( \left\lbrace \x_{i_{2\to L+1}},\,\y_{i_{2\to L+1}}\right\rbrace_{i_{2\to L}} \right).
\end{align}

\textbf{NTK.} The NTK of the same hierarchical CNN can be obtained recursively as follows. With $\kappa_0(t)\,{=}\,\left(\pi-\arccos{t}\right)/\pi$,
\begin{align}
&\mathcal{K}_{\mathrm{NTK}}^{(1)}\left( \x_{i_{2\to L+1}},\,\y_{i_{2\to L+1}} \right) = \|\x_{i_{2\to L+1}}\| \|\y_{i_{2\to L+1}}\| \, \kappa_1(t_{i_{2\to L+1}}) \nonumber \\
&\qquad\qquad\qquad\qquad\qquad\qquad + \x_{i_{2\to L+1}} ^\top \y_{i_{2\to L+1}} \, \kappa_0(t_{i_{2\to L+1}});\nonumber\\
&\mathcal{K}_{\mathrm{NTK}}^{(l)} \left(\left\lbrace \x_{i_{2\to L+1}},\,\y_{i_{2\to L+1}} \right\rbrace_{i_{2\to l}}\right) = \mathcal{K}_{\mathrm{RFK}}^{(l)} (\left\lbrace \x_{i_{2\to L+1}},\,\y_{i_{2\to L+1}} \right\rbrace_{i_{2\to l}}) \nonumber \\
&\qquad\qquad\qquad\qquad\qquad\qquad + \left(\frac{1}{s_l}\sum_{i_l} \mathcal{K}_{\mathrm{NTK}}^{(l-1)}\left( \left\lbrace \x_{i_{2\to L+1}},\,\y_{i_{2\to L+1}} \right\rbrace_{i_{2\to l-1}}\right)\right) \nonumber \\
&\qquad\qquad\qquad\qquad\qquad\qquad \times \kappa_0\left(\frac{\frac{1}{s_l}\sum_{i_l} \mathcal{K}_{\mathrm{RFK}}^{(l-1)}\left( \left\lbrace \x_{i_{2\to L+1}},\,\y_{i_{2\to L+1}} \right\rbrace_{i_{2\to l-1}}\right)}{\sqrt{\frac{1}{s_l}\sum_{i_l}\|\x_{i_{l\to L+1}}\|^2} \sqrt{\frac{1}{s_l}\sum_{i_l}\|\y_{i_{l\to L+1}}\|^2}}\right); \nonumber\\
&\mathcal{K}_{\mathrm{NTK}}^{(L+1)} \left( \left\lbrace \x_{i_{2\to L+1}},\,\y_{i_{2\to L+1}} \right\rbrace_{i_{2\to L+1}} \right)= \frac{1}{p_L}\sum_{i_{L+1}=1}^{p_L} \mathcal{K}_{\mathrm{NTK}}^{(L)}\left( \left\lbrace \x_{i_{2\to L+1}},\,\y_{i_{2\to L+1}} \right\rbrace_{i_{2\to L}} \right).
\end{align}
\end{definition}

\begin{figure}

    \centering
         \includegraphics[width=0.48\textwidth]{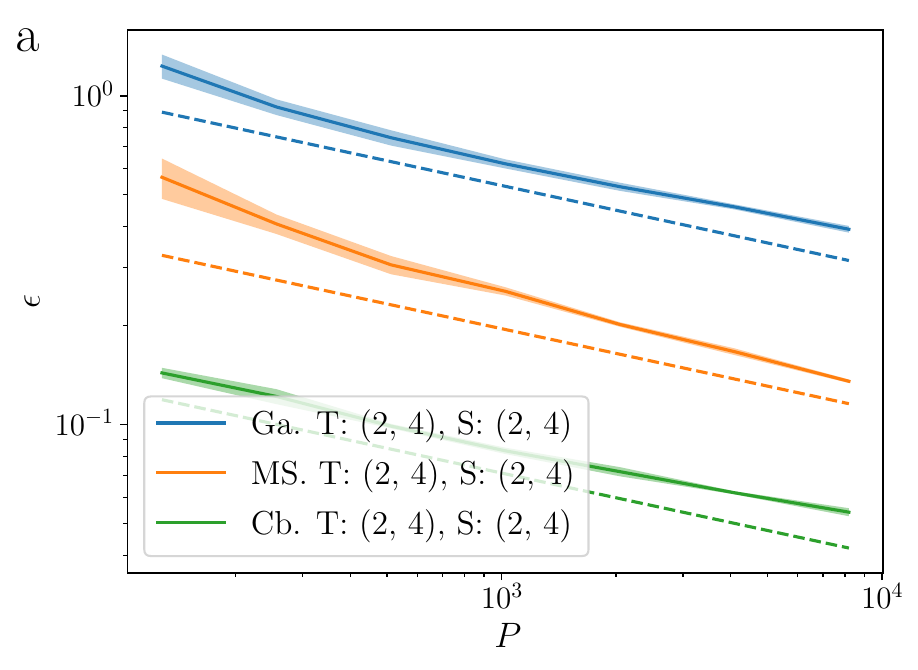}
         \includegraphics[width=0.48\textwidth]{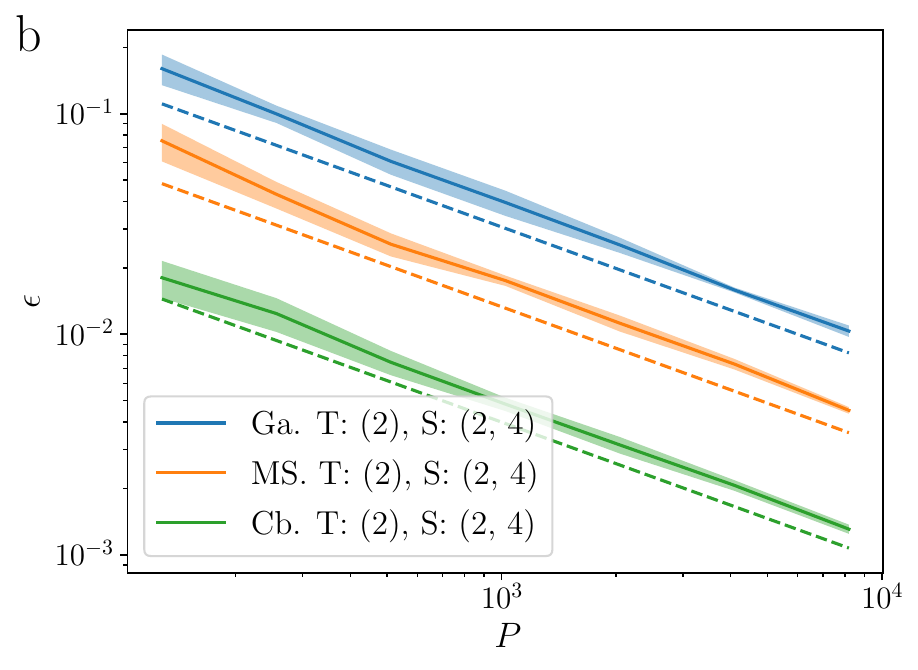}

    \caption{Learning curves for deep convolutional NTKs ($\nu=1/2$) in a teacher-student setting with different input normalizations. In particular, we consider inputs on the multisphere ${\sf M}^p\mathbb{S}^{s-1}$ (MS.), uniformly-distributed in the unit $d$-hypercube $[0,1]^d$ (Cb.), and with standard Gaussian distribution $\mathcal{N}(\mathbf{0},\mathbf{I}_d)$ (Ga.). The numbers inside brackets are the sequence of filter sizes of the kernels. Solid lines are the results of experiments averaged over 16 realizations, with the shaded areas representing the empirical standard deviations. The asymptotic scaling $\testerr \sim P^{-\beta}$ predicted for inputs on the multisphere are reported as dashed lines.}

    \label{fig:normalizations}
    
\end{figure}

\autoref{fig:normalizations} reports the learning curve of different teacher-student scenarios with the kernels defined in \autoref{def:hierarchical-kernels-Rd} and inputs \textit{i)} on the multisphere ${\sf M}^p\mathbb{S}^{s-1}$, \textit{ii)} uniformly-distributed in the unit $d$-hypercube $[0,1]^d$, and \textit{iii)} with standard Gaussian distribution $\mathcal{N}(\mathbf{0},\mathbf{I}_d)$. Remarkably, our predictions are in excellent agreement with the different input normalizations.

\begin{figure}

    \centering
    \includegraphics[width=.5\textwidth]{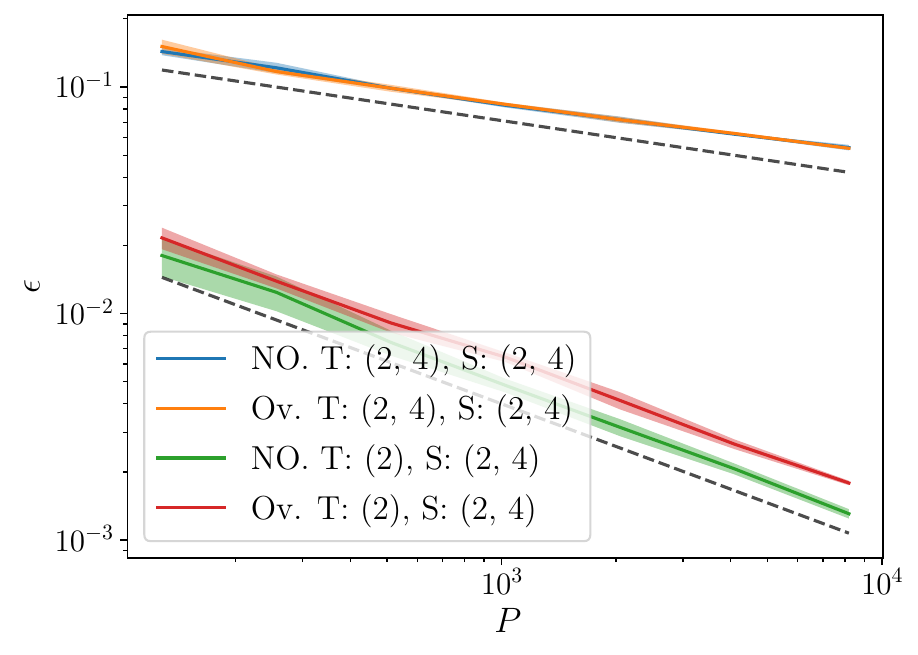}
    
    \caption{Learning curves for deep convolutional NTKs ($\nu=1/2$) with non-overlapping (NO.) and overlapping (Ov.) patches in a teacher-student setting with inputs normalized in the $d$-hypercube. The numbers inside brackets are the sequence of filter sizes of the kernels. Solid lines are the results of experiments averaged over 16 realizations, with the shaded areas representing the empirical standard deviations. The asymptotic scaling $\testerr \sim P^{-\beta}$ predicted for kernels with non-overlapping patches are reported as dashed lines.}
    \label{fig:overlap}
    
\end{figure}

\paragraph{Overlapping patches} \autoref{fig:overlap} shows the comparison between convolutional kernels with non-overlapping patches, i.e., stride corresponding to the filter size, and overlapping patches, i.e., stride 1, for inputs uniform in the $d$-dimensional hypercube. Despite our theoretical analysis requiring the patches to be non-overlapping, our predictions are still confirmed for architectures with overlapping patches.

\subsection{CIFAR-2 learning curves}

\begin{figure}

    \centering
         \includegraphics[width=0.48\textwidth]{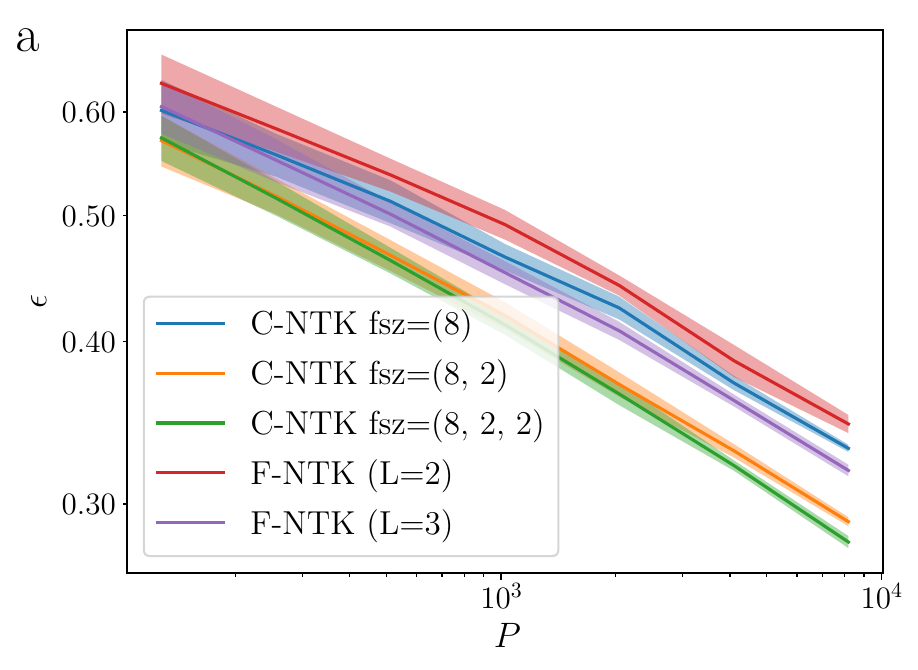}
         \includegraphics[width=0.48\textwidth]{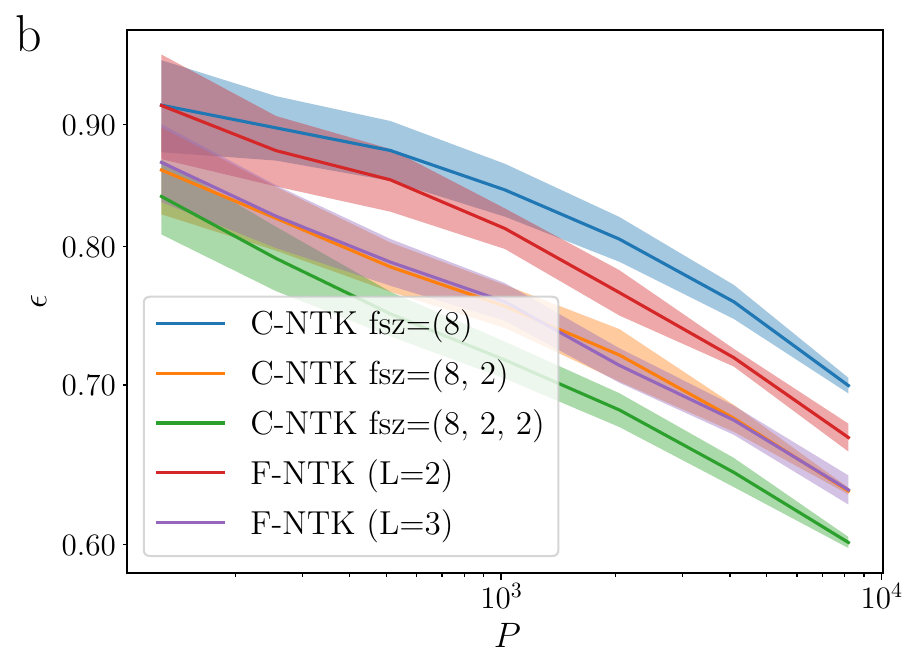}
    
    \caption{Learning curves of the neural tangent kernels of fully-connected (F-NTK) and convolutional (C-NTK) networks with various depths learning to classify two CIFAR-10 classes in a regression setting. Deep hierarchical convolutional kernels achieve the best performance. Shaded areas represent the empirical standard deviations obtained by averaging over different training sets. \textit{(a)} Plane vs car. \textit{(b)} Cat vs bird.}

    \label{fig:real_data}
    
\end{figure}

\autoref{fig:real_data} shows the learning curves of the
neural tangent kernels of different architectures applied to pairs of classes of the CIFAR-10 dataset. In particular, the task is built by selecting two CIFAR-10 classes, e.g., plane and car, and assigning label $+1$ to the elements belonging to one class and label $-1$ to the remaining ones. Learning is again achieved by minimizing the empirical mean squared error using a `student' kernel. We find that the kernels with the worst performance are the ones corresponding to shallow fully-connected and convolutional architectures. Instead, for all the pairs of classes considered here, deep hierarchical convolutional kernels achieve the best performance.

\chapter{Appendix: A phase Transition in the Diffusion Process}

\section{Belief Propagation initialization for the denoising of the RHM}
\label{app:bayes}

As discussed in \autoref{sec:bp}, we define the diffusion process for the input variable $X_{i}^{(0)}$ in the space $\mathbb{R}^v$. In particular, its value $x(t)$ at time $t$ is

\begin{align}
    x(t) = \sqrt{\overline{\alpha_t}} x(0) + \sqrt{1-\overline{\alpha_t}} \eta,
    \label{eq:app_diffusion}
\end{align}
with $\eta \sim \mathcal{N}(\mathbf{0},\textbf{I}_{v})$ and $x(0)$ its starting value at time $t$, which is a one-hot-encoding vector of the form $x(0) = e_{\mu}$. Given the value $x(t)$, the conditional probabilities for the values of $x(0)$ are given by Bayes rule
\begin{align}
    p\lpa x(0) = e_{\mu} \vert x(t)\rpa = 
    \frac{p\lpa x(t)\vert x(0) = e_{\mu}\rpa \, p\lpa x(0) = e_{\mu}\rpa}{\sum_{\lambda} p\lpa x(t)\vert x(0) = e_{\lambda}\rpa \, p\lpa x(0) = e_{\lambda}\rpa}.
\label{eq:app_bayes1}    
\end{align}

The prior probabilities on $x(0)$ are taken to be uniform over the alphabet, i.e., $p\lpa x(0) = e_{\lambda}\rpa = 1/v$, $\forall \lambda$, while $p\lpa x(t)\vert x(0) = e_{\mu}\rpa$ is given by the diffusion process of \autoref{eq:app_diffusion}:
\begin{align}
\begin{split}
    p\lpa x(t)\vert x(0)= e_{\mu}\rpa 
    &= C_{t} \exp\lsq-\frac{1}{2(1-\overline{\alpha_t})}\sum_{\gamma}\lpa x_\gamma(t) - \sqrt{\overline{\alpha_t}} e_{\mu} \rpa^2 \rsq =\\
    &=C_t \exp\lsq- \frac{\|x(t)\|^2 + \overline{\alpha_t}}{2(1-\overline{\alpha_t})} \rsq
    \exp\lsq \frac{\sqrt{\overline{\alpha_t}} }{1-\overline{\alpha_t}} x_\mu(t) \rsq, 
\label{eq:app_bayes2}
\end{split}
\end{align}
where $C_t$ is the normalization constant. Putting \autoref{eq:app_bayes2} into \autoref{eq:app_bayes1}, we obtain
\begin{align}
    p\lpa x(0) = e_{\mu} \vert x(t)\rpa = 
    \frac{1}{Z} e^{\frac{\sqrt{\overline{\alpha_t}} }{1-\overline{\alpha_t}} x_\mu(t)},
\end{align}
with $Z= \sum_{\lambda=1}^v e^{\frac{\sqrt{\overline{\alpha_t}} }{1-\overline{\alpha_t}} x_\lambda(t)}$.

\section{Belief Propagation equations}
\label{app:bp}

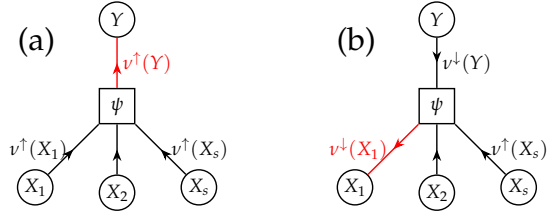
\begin{figure}[t]
    \centering
    \begin{tikzpicture}
        \node[anchor=north west,inner sep=0pt] at (0,0){
        \resizebox{.25\textwidth}{!}{\begin{tikzpicture}[varnode/.style={circle, draw=black, thick, minimum size=7mm, inner sep=0pt},
  factnode/.style={rectangle, draw=black, thick, minimum size=7mm, inner sep=0pt},
  level/.style={sibling distance=100mm/#1}, 
  edge from parent/.style={draw,thick}
  ]

\tikzset{
    varnode/.style={circle, draw, thick, minimum size=7mm, inner sep=0pt},
    factnode/.style={rectangle, draw, thick, minimum size=7mm, inner sep=0pt},
    midarrow/.style={decoration={
            markings,
            mark=at position 0.5 with {\arrow{Stealth}}
        },
        postaction={decorate},
        thick
    }
}
  
\node [varnode] (root) {$Y$};
\node [factnode] (factor) [below= of root] {$\psi$};
\draw[midarrow, red] (factor) -- (root) node[midway, right=0.1cm] {$\nu^{\uparrow}(Y)$};

\node [varnode] (left) [below left= of factor] {$X_1$};
\draw[midarrow] (left) -- (factor) node[midway, left=0.1cm] {$\nu^{\uparrow}(X_1)$};
\node [varnode] (center) [below = of factor] {$X_2$};
\draw[midarrow] (center) -- (factor);
\node [varnode] (right) [below right= of factor] {$X_s$};
\draw[midarrow] (right) -- (factor) node[midway, right=0.2cm] {$\nu^{\uparrow}(X_s)$};

\end{tikzpicture}}};
        \node[font=\large] at (2ex,-3ex) {(a)};
    \end{tikzpicture}
    \hspace{1cm}
    \begin{tikzpicture}
        \node[anchor=north west,inner sep=0pt] at (0,0){
        \resizebox{.25\textwidth}{!}{\begin{tikzpicture}[varnode/.style={circle, draw=black, thick, minimum size=7mm, inner sep=0pt},
  factnode/.style={rectangle, draw=black, thick, minimum size=7mm, inner sep=0pt},
  level/.style={sibling distance=100mm/#1}, 
  edge from parent/.style={draw,thick}
  ]

\tikzset{
    varnode/.style={circle, draw, thick, minimum size=7mm, inner sep=0pt},
    factnode/.style={rectangle, draw, thick, minimum size=7mm, inner sep=0pt},
    midarrow/.style={decoration={
            markings,
            mark=at position 0.5 with {\arrow{Stealth}}
        },
        postaction={decorate},
        thick
    }
}
  
\node [varnode] (root) {$Y$};
\node [factnode] (factor) [below= of root] {$\psi$};
\draw[midarrow] (root) -- (factor) node[midway, right=0.1] {$\nu^{\downarrow}(Y)$};

\node [varnode] (left) [below left= of factor] {$X_1$};
\draw[midarrow, red] (factor) -- (left) node[midway, left=0.1cm] {$\nu^{\downarrow}(X_1)$};
\node [varnode] (center) [below = of factor] {$X_2$};
\draw[midarrow] (center) -- (factor);
\node [varnode] (right) [below right= of factor] {$X_s$};
\draw[midarrow] (right) -- (factor) node[midway, right=0.2cm] {$\nu^{\uparrow}(X_s)$};

\end{tikzpicture}}};
        \node[font=\large] at (2ex,-3ex) {(b)};
    \end{tikzpicture}
    \caption{Factor tree-graph connecting the higher-level feature $Y$ to the lower-level features $X_{i=1,...,s}$ according to the rules $\rr$. 
    The upward messages $\nu^{\uparrow}(y)$ are computed from the upward messages $\nu^{\uparrow}(x_i)$ coming from the 
    nodes $X_i$, connected to $Y$ through the rule $\rr$ (panel (a)).
    The downward messages $\nu^{\downarrow}(x_1)$, instead, are computed from both the downward messages $\nu^{\downarrow}(y)$ coming from $Y$ and the upward messages $\nu^{\uparrow}(x_i)$ coming from the nodes $X_{i=2,...s}$, connected to $X_1$ through the rule $\rr$ (panel (b)).
    }
    \label{fig:patch}
\end{figure}

Given a factor tree-graph, the Belief Propagation (BP) equations compute iteratively the messages going from the variable nodes to the factor nodes and vice-versa, starting from the initialization conditions at the leaves and root of the tree-graph \cite{mezardmontanari}. For the generative model defined in \autoref{sec:rhm}, the leaves correspond to the variables at the bottom layer while the root is the class variable at the top of the hierarchy. 
Each rule, connecting variables at different layers, corresponds to a factor node. The BP messages that flow from the variable nodes to the factor nodes, therefore, correspond to upward messages, while those going from factor nodes to variables correspond to downward messages (\autoref{fig:patch}).

To each variable node $X^{(\ell)}_i$ at level $\ell$, we associate the upward messages $\nnu{\ell}$ and downward messages $\nnd{l}$, one for each possible value of the alphabet it can take.
To simplify the notation, here we consider how messages propagate from one level to the other, and we call $Y$ the variable corresponding to the higher level and $X_{i=1,...,s}$ the lower level ones connected to it. The factor node connecting them is such that, for each possible association $y\rightarrow x_1,\dots, x_s$, it takes values
\beqs
    \rul{\ell}(y, x_1, ..., x_s) = \begin{cases}
        1, \quad \text{if } y \rightarrow (x_1, ..., x_s) \text{ is a rule at layer }\ell\\
        0, \quad \text{otherwise}.
    \end{cases}
\eeqs

The BP upward and downward iterations are defined as follows.
\begin{itemize}
    \item Upward iteration:
\beq
    \ncu{\ell+1}(y) = \sum_{x_1, ..., x_s \in \mathcal{A}^{\otimes s}} \rul{\ell+1}(y, x_1, ..., x_s) \prod_{i=1}^s \nnu{\ell}(x_i)\ ,
    \label{eq:nuUP}
\eeq
\beq
    \nnu{\ell}(y) = \frac{\ncu{\ell}(y)}{\sum_{y'} \ncu{\ell}(y')}.
\eeq

    \item Downward iteration:
\beq
    \ncd{\ell}(x_1) = \sum_{\substack{x_2, ..., x_s \in \mathcal{A}^{\otimes (s-1)}\\ y\in \mathcal{A}}} \rul{\ell+1}(y, x_1, ..., x_s)\ \nnd{\ell+1}(y)\prod_{i=2}^s \nnu{\ell}(x_i)\
    \label{eq:nuDOWN}
\eeq
\beq
    \nnd{l}(x) = \frac{\ncd{\ell}(x)}{\sum_{x'} \ncd{\ell}(x')}.
\eeq
\end{itemize}

$\nnu{\ell}(y)$ and $\nnd{\ell}(x)$ are fluctuating quantities that depend on the position of the node.

\subsection{$\epsilon$-process}

In this process, we consider a reference configuration at the leaves variables $X_i^{(0)} = \corr{x}_i$ that we would like to reconstruct, given a noisy observation of it. As a result of this noise addition, our belief in the correct sequence is corrupted by $\epsilon \in [0,1]$:
\begin{align}
\begin{cases}
    X_i^{(0)} = \corr{x}_i\quad \text{with belief } 1-\epsilon\\
    X_i^{(0)} \text{ uniform over alphabet with belief } \epsilon.\\
\end{cases}
\end{align}

Therefore, the initialization condition of the upward BP messages at a leaf node $X_i^{(0)}$ is 
\begin{align}
\begin{cases}
    \nu_{\uparrow}\lpa \corr{x}_i\rpa \qquad = 1-\epsilon+\epsilon/v,\\
    \nu_{\uparrow}\lpa x_i\neq\corr{x}_i \rpa = \epsilon/v, \\
\end{cases}
\label{eq:belief}
\end{align}
where $v$ is the corresponding alphabet size.\\
The initialization condition at the root node $X^{(L)}$, that corresponds to the messages $\nnd{L}$ for that node, is uniform over the alphabet $\mathcal{A}$, so that the algorithm has no bias on any specific class.

\subsubsection{Upward iteration}
\label{app:upward_anneal}

We consider the upward iteration when going from the bottom layer to the one above it. Let $X_1, \dots, X_s$ denote a tuple at the bottom level which is associated with the reference values $\corr{x}_i$. This tuple is connected to the higher level variable $Y$ via a set of rules $\rr$ (\autoref{fig:patch}). According to $\rr$, the association from the high-level feature to the reference low-level sequence $\corr{x}_1,\dots,\corr{x}_s$ is given by
\beqs
     \corr{y}\rightarrow \corr{x}_1,\dots,\corr{x}_s.
\eeqs

We call $\ham{\mathbf{w}, \mathbf{z}}$ the Hamming distance between two sequences $\mathbf{w} = [w_1, \dots, w_s]$, $\mathbf{z}=[z_1, \dots, z_s]$ of length $s$.\\
From \autoref{eq:belief}, at the bottom layer, the belief in a sequence $\x = [x_1,\dots, x_s]$ with $\ham{\x,\corr{\x}} = k \in\{0,...,s\}$ from $\corr{\x} = [\corr{x}_1,\dots, \corr{x}_s]$ is
\beq
    \bel(k) = \lpa\frac{\epsilon}{v}\rpa^{k} \lpa 1-\epsilon+\frac{\epsilon}{v}\rpa^{s-k}
    \label{eq:bel_k}
\eeq

The non-normalized upward messages for the variable $Y$ are given by:
\beq
    \tilde{\nu}_{\uparrow}(y) = 
    \sum_{x_1, ..., x_s} \rr(y, x_1, ..., x_s) \prod_{i=1}^s \nu_{\uparrow}(x_i) = 
    \sum_{\x\in\samp} \rr(y, \x) \bel\lpa\ham{\x,\corr{\x}}\rpa,
    \label{eq:nu_y}
\eeq
where we are using the short-hand notation $\rr(y, x_1, ..., x_s) = \rr(y, \x)$ and we have restricted the sum over the set $\samp$ of sequences $\x$ that appear in the possible rules $y\rightarrow x_1,\dots, x_s$. In fact, if $\x\notin\samp$, then $\rr(y, \x) = 0$.

For $\x\in\samp$, the factor $\rr(y, \x)$ is such that:
\begin{itemize}
    \item if $\ham{\x,\corr{\x}}=0$:
    \beq
    \begin{aligned}
        \begin{cases}
            &\rr(\corr{y}, \corr{x}_1, ..., \corr{x}_s) = 1,\\
            &\rr(y, \corr{x}_1, ..., \corr{x}_s) = 0, \quad y\neq \corr{y}.\\
        \end{cases}
    \end{aligned}
    \eeq
    \item if $\ham{\x,\corr{\x}}>0$:
    \beq
    \begin{aligned}
        \begin{cases}
            &\rr(\tilde{y}, x_1, ..., x_s) = 1, \quad \text{for some $\tilde{y}$ independent of $\corr{y}$}\\
            &\rr(y, x_1, ..., x_s) = 0, \quad y\neq \tilde{y}.\\
        \end{cases}
    \label{eq:randomPsi}
    \end{aligned}
    \eeq    
\end{itemize}

We can decompose \autoref{eq:nu_y} as
\beq
    \tilde{\nu}_{\uparrow}(y) = 
    \delta_{y,\corr{y}} \bel(0) +
    \sum_{k=1}^s
    \bel\lpa k\rpa
    \lsq
    \sum_{\substack{\x\in\samp\\
    \ham{\x,\corr{\x}}=k
    }} \psi(y, \x)
    \rsq.
\eeq

\paragraph{Annealed average}
$\psi$ is a random quantity and we want to compute the average message $\langle \tilde{\nu}_{\uparrow}(y)\rangle_\rr$ over the possible realizations of $\rr$. We can decompose the selection of the rules in two steps: sampling the set of $mv-1$ sequences $\{\x, \x \neq \corr{\x}\}$ and then associating the $v$ higher-level features $y$ to them. Therefore, for a generic quantity $A$, we indicate the average over the rules realization $\langle A \rangle_\rr$ as $\langle\langle A \rangle_{\asso}\rangle_{\samp}$, where $\langle\dots\rangle_{\samp}$ is the average over the sequence sampling and $\langle \dots\rangle_{\asso}$ is the average over the $y\leftarrow \x$ associations:

\beq
    \langle \tilde{\nu}_{\uparrow}(y) \rangle_\psi= 
    \delta_{y,\corr{y}} \bel(0) +
    \sum_{k=1}^s
    \bel\lpa k\rpa
    \langle\langle
    \sum_{\substack{\x\in\samp\\
    \ham{\x,\corr{\x}}=k
    }} \psi(y, \x)
    \rangle_{\asso}\rangle_{\samp}
\eeq

Since for each sequence $\x\neq \corr{\x}$ the association $y\leftarrow \x$ is done randomly, independently of $\ham{\x,\corr{\x}}$, then from \autoref{eq:randomPsi}, we have $\langle \psi(y, \x) \rangle_{\asso} \simeq 1/v$. More precisely, since we have associated the reference sequence $\corr{\x}$ to $\corr{y}$:
\beq
    \overline{\psi_y} = \langle \psi(y, \x) \rangle_{\asso} = 
    \frac{m-1}{mv-1} \delta_{y,\corr{y}} + \frac{m}{mv-1} \lpa 1-\delta_{y,\corr{y}} \rpa.
\eeq

Therefore:
\beq
\begin{aligned}
    \langle \tilde{\nu}_{\uparrow}(y) \rangle_\psi &= 
    \delta_{y,\corr{y}} \bel(0) +
    \overline{\psi_y}
    \sum_{k=1}^s
    \bel\lpa k\rpa
    \langle
    \sum_{\substack{\x\in\samp\\
    \ham{\x,\corr{\x}}=k
    }} 1
    \rangle_{\samp} = \\
    & = \delta_{y,\corr{y}} \bel(0) +
    \overline{\psi_y}
    \sum_{k=1}^s
    \bel\lpa k\rpa
    \langle
    n_k
    \rangle_{\samp},
\end{aligned}
\eeq

where $n_k$ is the number of sequences $\x\in\samp$ having Hamming distance $\ham{\x,\corr{\x}} = k$ from $\corr{\x}$. Since the sequences are sampled randomly, the numbers $n_1, ..., n_s$ are distributed according to a multivariate hyper-geometric distribution,
\beq
    P\lpa n_1, ..., n_s \rpa = \frac{\prod_{k=1}^s \binom{\binom{s}{k}(v-1)^k}{n_k}}{\binom{v^s-1}{mv-1}},
\eeq
which gives the averages 
\beq
    \langle n_k \rangle_{\samp} = \frac{mv-1}{v^s-1} \binom{s}{k}(v-1)^k = f \binom{s}{k}(v-1)^k,
\eeq
with
\beq
    f = \frac{mv-1}{v^s-1}.
\eeq

Therefore:
\beq
\begin{aligned}
    \langle \tilde{\nu}_{\uparrow}(y) \rangle_\psi = 
    \delta_{y,\corr{y}} \bel(0) +
    f\ \overline{\psi_y}
    \sum_{k=1}^s
    \bel\lpa k\rpa
    \binom{s}{k}(v-1)^k
    .
\end{aligned}
\eeq

From the beliefs \autoref{eq:bel_k}, we see that
\begin{align}
    \sum_{k=1}^s
    \bel\lpa k\rpa
    \binom{s}{k}(v-1)^k &=
    \sum_{k=1}^s
    \binom{s}{k}(v-1)^k
    \lpa\frac{\epsilon}{v}\rpa^{k} \lpa 1-\epsilon+\frac{\epsilon}{v}\rpa^{s-k} \nonumber \\
    &= 
    \lsq 1 - \lpa 1-\epsilon+\frac{\epsilon}{v}\rpa^s \rsq
    = 1-\bel(0),
\end{align}

which gives
\beq
\begin{aligned}
    \langle \tilde{\nu}_{\uparrow}(y) \rangle_\psi = 
    \delta_{y,\corr{y}} \bel(0) +
    f\ \overline{\psi_y}
    \lsq 1- \bel(0) \rsq
\end{aligned}
.
\eeq

The normalization constant is:
\beq
    \langle Z_{\uparrow} \rangle_\rr = \sum_{y} \langle \tilde{\nu}_{\uparrow}(y) \rangle_\rr = \bel(0) + f \lsq 1- \bel(0) \rsq .
\eeq

Finally, we obtain the average belief for $Y$

\beq
    \langle \nu_{\uparrow}(y) \rangle_\psi = \frac{\langle \tilde{\nu}_{\uparrow}(y) \rangle_\rr}{\langle Z_{\uparrow} \rangle_\rr} = 
    \frac{\delta_{y,\corr{y}} \bel(0) + f\ \overline{\psi_y} \lsq 1- \bel(0) \rsq}{\bel(0) + f \lsq 1- \bel(0) \rsq}
\eeq
 We have that:
\begin{itemize}
    \item for $y=\corr{y}$
    \beq
        \langle \nu_{\uparrow}(\corr{y}) \rangle_\psi =
        \frac{\bel(0) + f \frac{m-1}{mv-1} \lsq 1- \bel(0) \rsq}{\bel(0) + f \lsq 1- \bel(0) \rsq},
        \label{eq:nu_corr}
    \eeq
    \item for $y\neq \corr{y}$
    \beq
        \langle \nu_{\uparrow}(y) \rangle_\psi =
        f \frac{m}{mv-1} \frac{1- \bel(0)}{\bel(0) + f \lsq 1- \bel(0) \rsq}.
        \label{eq:nu_mist}
    \eeq
\end{itemize}

\paragraph{Iterating over layers}
The average messages in \autoref{eq:nu_corr}, \autoref{eq:nu_mist} are of two kinds: one for the reference feature $\corr{y}$ and another for the others $y\neq \corr{y}$, and they both depend on the previous beliefs through $\bel(0) = \lpa 1-\epsilon + \frac{\epsilon}{v}\rpa^s$. Therefore, the average messages at the higher level have the same structure as those at the lower level \autoref{eq:belief}. We can then define a new $\epsilon'$:
\beq
    1-\epsilon' + \frac{\epsilon'}{v} = \frac{\lpa 1-\epsilon + \frac{\epsilon}{v}\rpa^s + f \frac{m-1}{mv -1}\lsq 1- \lpa 1-\epsilon + \frac{\epsilon}{v}\rpa^s \rsq }{\lpa 1-\epsilon + \frac{\epsilon}{v}\rpa^s + f \lsq 1 - \lpa 1-\epsilon + \frac{\epsilon}{v}\rpa^s \rsq}
\eeq
or, equivalently, 
\beq
    p' = \frac{p^s + f \frac{m-1}{mv-1}\lpa 1-p^s\rpa}{p^s + f\lpa 1-p^s\rpa} = F(p)
    \label{eq:iterUP}
\eeq
with $p' = 1-\epsilon' +\epsilon'/v$ and $p = 1-\epsilon +\epsilon/v$. The derivative of $F(p)$ with respect to $p$ is given by
\beq
    F'(p) = \frac{m(v-1)}{mv -1}\ \frac{f s p^{s-1}}{\lsq p^s + f(1-p^s) \rsq^2}=
    fs \frac{p^{s-1}}{\lsq p^s + f(1-p^s) \rsq^2} + \O\lpa\frac{1}{v}\rpa
\eeq

We can extend the tree in \autoref{fig:patch} iteratively to higher levels of the hierarchy, where the variables $Y$ take the place of the variables $X_i$ and so on. 

The iteration \autoref{eq:iterUP} has fixed points $p= 1$ (corresponding to $\epsilon=0$) or $p=1/v$ (corresponding to $\epsilon=1$).
An additional repulsive fixed point at finite $p$ appears if
\beq
    F'(1)<1,
\eeq
that is
\beq
    \frac{m(v-1)}{mv -1}\ f s < 1
    \quad \Rightarrow\quad fs < 1 + \frac{1-1/m}{v-1}   
\eeq
\beq
\boxed{
\Rightarrow\quad fs < 1 + \O\lpa\frac{1}{v}\rpa
}\,.
\eeq

\begin{figure}[t]
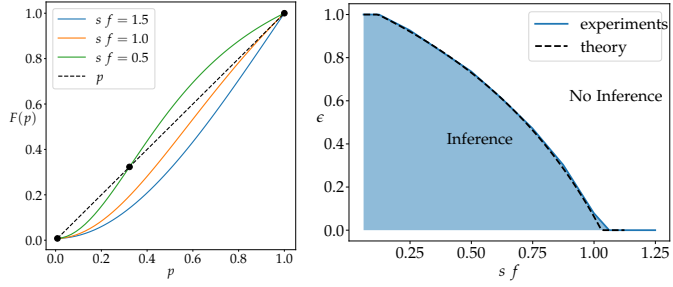

    \centering
    \begin{tikzpicture}
    \node[anchor=north west, inner sep=0pt] at (0,0){
    \resizebox{.33\columnwidth}{!}{
\input{Figures/phase/BP_iteration.pgf}}
        };
    \end{tikzpicture}
    \begin{tikzpicture}
    \node[anchor=north west,inner sep=0pt] at (0,0){
    \resizebox{.41\columnwidth}{!}{
    \input{Figures/phase/BP_upward-root_inference-v=32_s=2_L=10.pgf}}
        };
    \end{tikzpicture}
    \caption{
    \textit{Left panel:} 
    \textbf{Iteration map of \autoref{eq:iterUP}.} For $s f>1$, there are only two fixed points $p=F(p)$ corresponding to $p_1=1$ (repulsive) and $p_2=1/v$ (attractive). For $s f<1$, there is another (repulsive) fixed point at finite $p^*$, $1/v<p^*<1$, separating $p_1$ and $p_2$ (both attractive).
    \textit{Right panel:} 
    \textbf{Phase diagram for inferring the class node using the upwards iteration of BP.} When $sf<1$, BP can infer the class of the datum if $\epsilon<\epsilon^*(sf)$. This transition is well predicted by $p^*=1-\epsilon^*+\epsilon^*/v$, with $p^*=F(p^*)$ from \autoref{eq:iterUP}. Experimental data for $v=32$, $s=2$, $L=10$.
    }
    \label{fig:enter-label}
\end{figure}

\subsubsection{Downward iteration}
\label{app:downward_anneal}

We consider the downward process when we try reconstructing the reference association $\corr{y}\rightarrow \corr{x}_1,\dots,\corr{x}_s$ from higher-level variable $Y$ to the corresponding lower-level tuple $X_1, \dots, X_s$, via the set of rules $\rr$.
We consider the downward message received by the variable $X_1$ (\autoref{fig:patch}):
\begin{align}
\begin{split}
    \tilde{\nu}_{\downarrow}(x_1) &= \sum_{\substack{x_2, ..., x_s \in \mathcal{A}^{\otimes (s-1)}\\ y\in \mathcal{A}}} \rr(y, x_1, ..., x_s)\ \nu_{\downarrow}(y) \prod_{i=2}^s \nu_{\uparrow}(x_i)=\\
    &=\delta_{x_1, \corr{x}_1} \nu_{\downarrow}(\corr{y})\prod_{i=2}^s \nu_{\uparrow}(\corr{x}_i)
    +
    \sum_{y}
    \nu_{\downarrow}(y)
    \sum_{\substack{x_2,\dots,x_s\\
    \x\in\samp, \ham{\x,\corr{\x}}>0
    }}
    \psi(y, \x)
    \prod_{i=2}^s \nu_{\uparrow}(x_i)
\end{split}
\label{eq:nu_down}
\end{align}

\paragraph{Annealed average}
To study the iteration of \autoref{eq:nu_down} analytically, we compute the average message $\langle \tilde{\nu}_{\downarrow}(x_1)\rangle_\rr$ over the realizations of the random rules $\rr$ as done in \autoref{app:downward_anneal} for the upward iteration. 

We call $n_{x_1}$ the number of sequences, having $X_1=x_1$, that have been sampled by the choice of the rules.
The numbers $n_{x_1}$ are distributed according to a multivariate hyper-geometric distribution,
\beq
    P\lpa \{n_{x_1}\}_{x_1\in\mathcal{A}} \rpa = \frac{\binom{v^{s-1}-1}{n_{\corr{x}_1}} \prod_{\tilde{x}_1\in\mathcal{A}\setminus \corr{x}_1} \binom{v^{s-1}}{n_{\tilde{x}_1}}}{\binom{v^s-1}{mv-1}},
\eeq
which gives averages
\beq
    \langle n_{\corr{x}_1} \rangle = \frac{mv-1}{v^s-1} \lpa v^{s-1} - 1\rpa = f \lpa v^{s-1} - 1\rpa,
\eeq
\beq
    \langle n_{\tilde{x}_1 \neq \corr{x}_1} \rangle = \frac{mv-1}{v^s-1} v^{s-1} = f v^{s-1}.
\eeq

Averaging the downward messages over the choices of rules $\rr$, we obtain:

\begin{itemize}
    \item for $x_1\neq \corr{x}_1$:
    \begin{align}
    \begin{split}
    \langle \tilde{\nu}_{\downarrow}(x_1) \rangle_\psi &= 
    \sum_{y}
    \nu_{\downarrow}(y)
    \langle
    \sum_{\substack{x_2,\dots,x_s\\
    \x\in\samp}}
    \langle \psi(y, \x) \rangle_{\asso}\ 
    \prod_{i=2}^s \nu_{\uparrow}(x_i)
    \rangle_{\samp}
    =\\
    & = \frac{m - \nu_{\downarrow}(\corr{y})}{mv-1} 
    \langle \delta_{\x\in\samp}\rangle_{\samp}
    \prod_{i=2}^s \sum_{x_i} \nu_{\uparrow}(x_i)
    = \frac{m - \nu_{\downarrow}(\corr{y})}{mv-1} f =\\
    &= \frac{m - \nu_{\downarrow}(\corr{y})}{mv-1}f,
    \end{split}
    \end{align}

    where $\langle \delta_{\x\in\samp}\rangle_{\samp} = \frac{\langle n_{x_1 \neq \corr{x}_1} \rangle}{v^{s-1}}$;
    
    \item for $x_1 = \corr{x}_1$:
    \begin{align}
    \begin{split}
    \langle \tilde{\nu}_{\downarrow}(\corr{x}_1) \rangle_\psi &= 
    \nu_{\downarrow}(\corr{y})\prod_{i=2}^s \nu_{\uparrow}(\corr{x}_i) \\
    &+
    \sum_{y}
    \nu_{\downarrow}(y)
    \langle
    \sum_{\substack{x_2,\dots,x_s\\
    \x\in\samp, \x\neq\corr{\x}
    }}
    \langle \psi(y, \x) \rangle_{\asso}\ 
    \prod_{i=2}^s \nu_{\uparrow}(x_i)
    \rangle_{\samp}
    =\\
    &= \nu_{\downarrow}(\corr{y})\prod_{i=2}^s \nu_{\uparrow}(\corr{x}_i) +
    \frac{m - \nu_{\downarrow}(\corr{y})}{mv-1}
    \langle \delta_{\x\in\samp}\rangle_{\samp}
    \sum_{\substack{x_2,\dots,x_s\\
    \x\in\samp, \x\neq\corr{\x}
    }} \prod_{i=2}^s \nu_{\uparrow}(x_i) = \\
    &= \nu_{\downarrow}(\corr{y})\prod_{i=2}^s \nu_{\uparrow}(\corr{x}_i) +
    \frac{m - \nu_{\downarrow}(\corr{y})}{mv-1}
    f
    \lsq 1 - \prod_{i=2}^s \nu_{\uparrow}(\corr{x}_i)\rsq
    \end{split}
    \end{align}

    where $\langle \delta_{\x\in\samp}\rangle_{\samp} = \frac{\langle n_{\corr{x}_1} \rangle}{v^{s-1}-1}$.
\end{itemize}

The normalization factor is
\begin{align}
    \langle Z_{\downarrow} \rangle_\rr &= \sum_{x_1} \langle \tilde{\nu}_{\downarrow}(x_1) \rangle_\psi = \nonumber \\ &=
    \nu_{\downarrow}(\corr{y})\prod_{i=2}^s \nu_{\uparrow}(\corr{x}_i) 
    + f\ \frac{m - \nu_{\downarrow}(\corr{y})}{mv-1}
    \lsq 1 - \prod_{i=2}^s \nu_{\uparrow}(\corr{x}_i)\rsq
    + (v-1) f `\frac{m-\nu_{\downarrow}(\corr{y})}{mv-1} 
\end{align}
which gives the normalized average messages:
\begin{itemize}
    \item for $x_1 = \corr{x}_1$
\beq
    \langle \nu_{\downarrow}(\corr{x}_1) \rangle_\psi  = 
    \frac{
    \nu_{\downarrow}(\corr{y})\prod_{i=2}^s \nu_{\uparrow}(\corr{x}_i)
    + f\ \frac{m - \nu_{\downarrow}(\corr{y})}{mv-1} \lsq 1 - \prod_{i=2}^s \nu_{\uparrow}(\corr{x}_i)\rsq}{\langle Z_{\downarrow} \rangle_\rr};
    \label{eq:ave_nu_d}
\eeq
\item for $x_1 \neq \corr{x}_1$
\beq
    \langle \nu_{\downarrow}(x_1) \rangle_\psi  = 
    f \frac{
    \frac{m - \nu_{\downarrow}(\corr{y})}{mv-1}}{\langle Z_{\downarrow} \rangle_\rr}.
\eeq
\end{itemize}

\paragraph{Iterating over layers}
As for the upward process, the average messages for the downward process are of two kinds, one for the correct value $\corr{x}_1$ and one for the other values $x_1\neq\corr{x}_1$. To obtain a mean-field description of the BP process, we combine the average downward messages with the average upward ones by substituting $\nu_{\uparrow}(\corr{x}_i)\rightarrow \langle \nu_{\uparrow}(\corr{x}_i) \rangle$ in \autoref{eq:ave_nu_d}.
We use the notation 
\begin{align}
\langle \nnu{\ell}(\corr{x}_i) \rangle &= p_{\uparrow}^{(\ell)},\\
\langle \nnd{\ell}(\corr{x}_i) \rangle &= p_{\downarrow}^{(\ell)},
\end{align}
where the upwards and downwards beliefs $p_{\uparrow}^{(\ell)}$, $p_{\downarrow}^{(\ell)}$ in the correct value for the latent variable $X^{(\ell)}_i$ depend only on the layer $\ell$ and not on the specific position $i$ inside the layer.
Putting together \autoref{eq:iterUP} and \autoref{eq:ave_nu_d}, we obtain
\begin{align}
\begin{split}
    &p_{\uparrow}^{(\ell+1)} = F_{\uparrow}\lpa p_{\uparrow}^{(\ell)}\rpa,\\
    &p_{\downarrow}^{(\ell)} = F_{\downarrow}\lpa p_{\downarrow}^{(\ell+1)},  p_{\uparrow}^{(\ell)}\rpa,
\end{split}
\label{eq:anneal_iteration}
\end{align}
with
\begin{align}
    &F_{\uparrow}(p) = \frac{p^s + f \frac{m-1}{mv-1}\lpa 1-p^s\rpa}{p^s + f\lpa 1-p^s\rpa},\\
    &F_{\downarrow}(q, p) = \frac{q\ p^{s-1} + f \frac{m-q}{mv-1}\lpa 1-p^{s-1}\rpa}{q\ p^{s-1} + f \frac{m-q}{mv-1}\lpa 1-p^{s-1}\rpa + (v-1)f \frac{m-q}{mv-1}},
\end{align}

and the initialization condition
\begin{align}
    &p_{\uparrow}^{(0)} = 1-\epsilon + \epsilon/v,\\
    &p_{\downarrow}^{(L)} = 1/v.
\end{align}

From $p_{\uparrow}^{(\ell)}$ and $p_{\downarrow}^{(\ell)}$, at layer $\ell$, the average marginal probability of the correct value $p^{(\ell)}$ is given by
\beq
    p^{(\ell)} = \frac{p_{\uparrow}^{(\ell)} p_{\downarrow}^{(\ell)}}{p_{\uparrow}^{(\ell)} p_{\downarrow}^{(\ell)} + \frac{(1-p_{\uparrow}^{(\ell)})(1-p_{\downarrow}^{(\ell)})}{v-1}}.
    \label{eq:theory_marginal}
\eeq

\subsubsection{Validity of the mean-field theory}

\begin{figure}
    \centering
    \includegraphics[width=\textwidth]{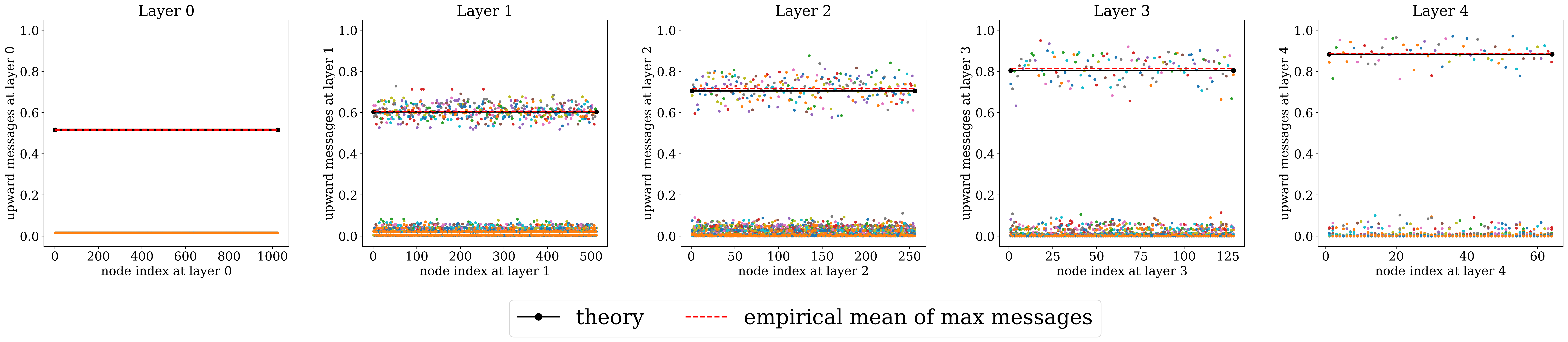}
    \caption{\textbf{Upward BP messages for layers $0$ to $4$ for $\epsilon=0.5$, $v=32$, $s=2$, $L=10$, $sf=0.5$.} Each node has $v$ messages, one per possible feature. At the input layer (layer $0$), messages have value $1-\epsilon=0.5$ or $\epsilon/v=0.5/32$. Going upward, the values of the messages fluctuate, but they stay separated into two distinct groups: large messages (i.e., the most probable feature for each node) and small ones. The annealed mean-field computation (represented with a black line) captures the mean value of the large messages well (red dashed line).}
    \label{fig:up-mean_messages}
\end{figure}

\begin{figure}
    \centering
    \includegraphics[width=\textwidth]{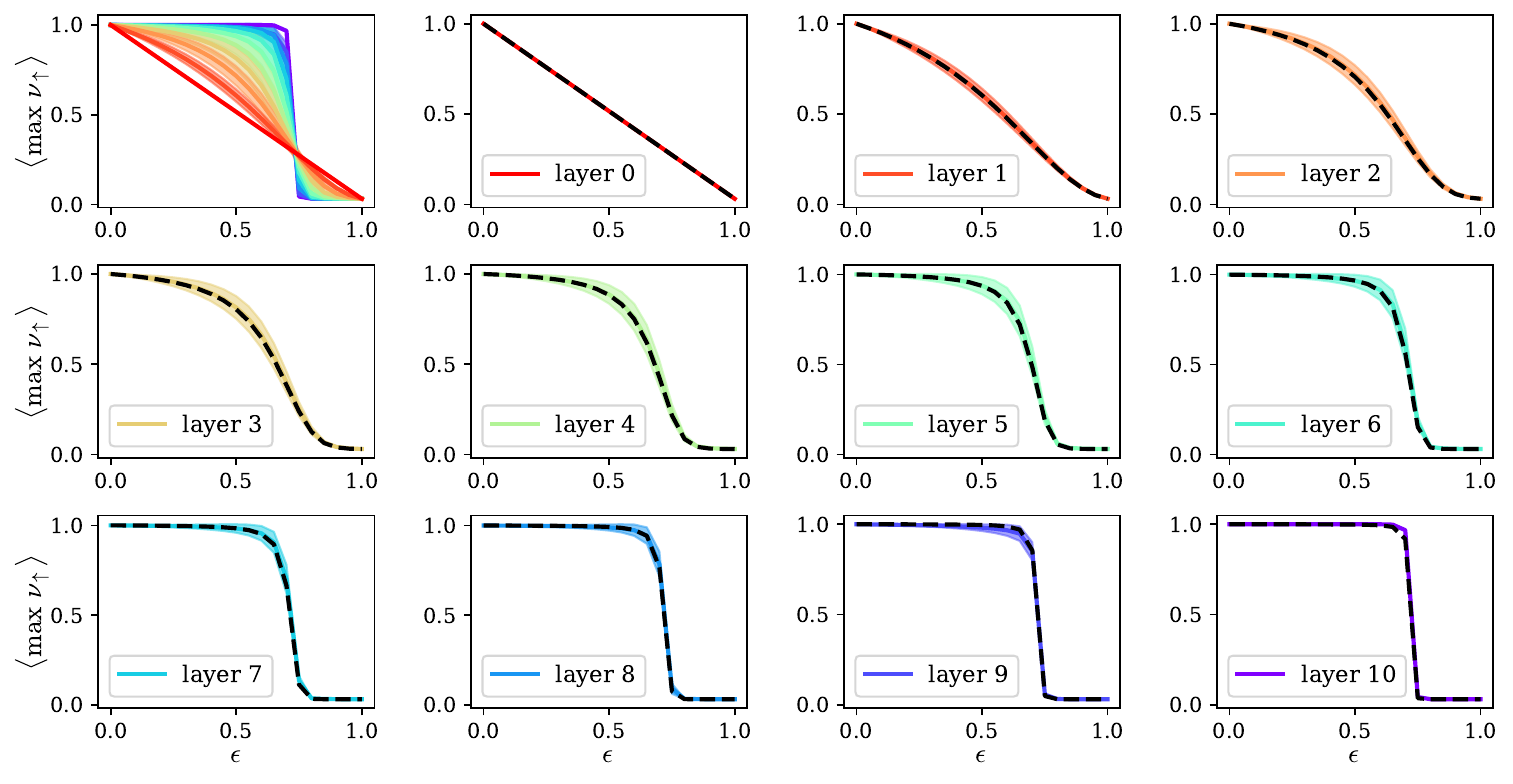}
    \caption{\textbf{Largest upward BP messages, averaged for each layer, for varying $\epsilon$. Data for the Random Hierarchical Model with $v=32$, $s=2$, $L=10$, $sf=0.5$.}
    Each layer, indicated in the legend, is represented with a different color, and the black dashed line is the theoretical prediction from \autoref{eq:anneal_iteration}, which shows excellent agreement with the experiments. The top left panel represents all the layers together for comparison. Starting from the initialization $\nu_{\uparrow}=1-\epsilon+\epsilon/v$ at layer $0$, we observe that the largest upward messages increase as we go to higher levels in the hierarchy only if $\epsilon$ is smaller than some threshold value. For $\epsilon$ larger than this threshold, instead, the messages become smaller, and it is not possible to reconstruct the highest levels in the hierarchy better than random chance.}
    \label{fig:upward_grid}
\end{figure}

\begin{figure}
    \centering
    \includegraphics[width=\textwidth]{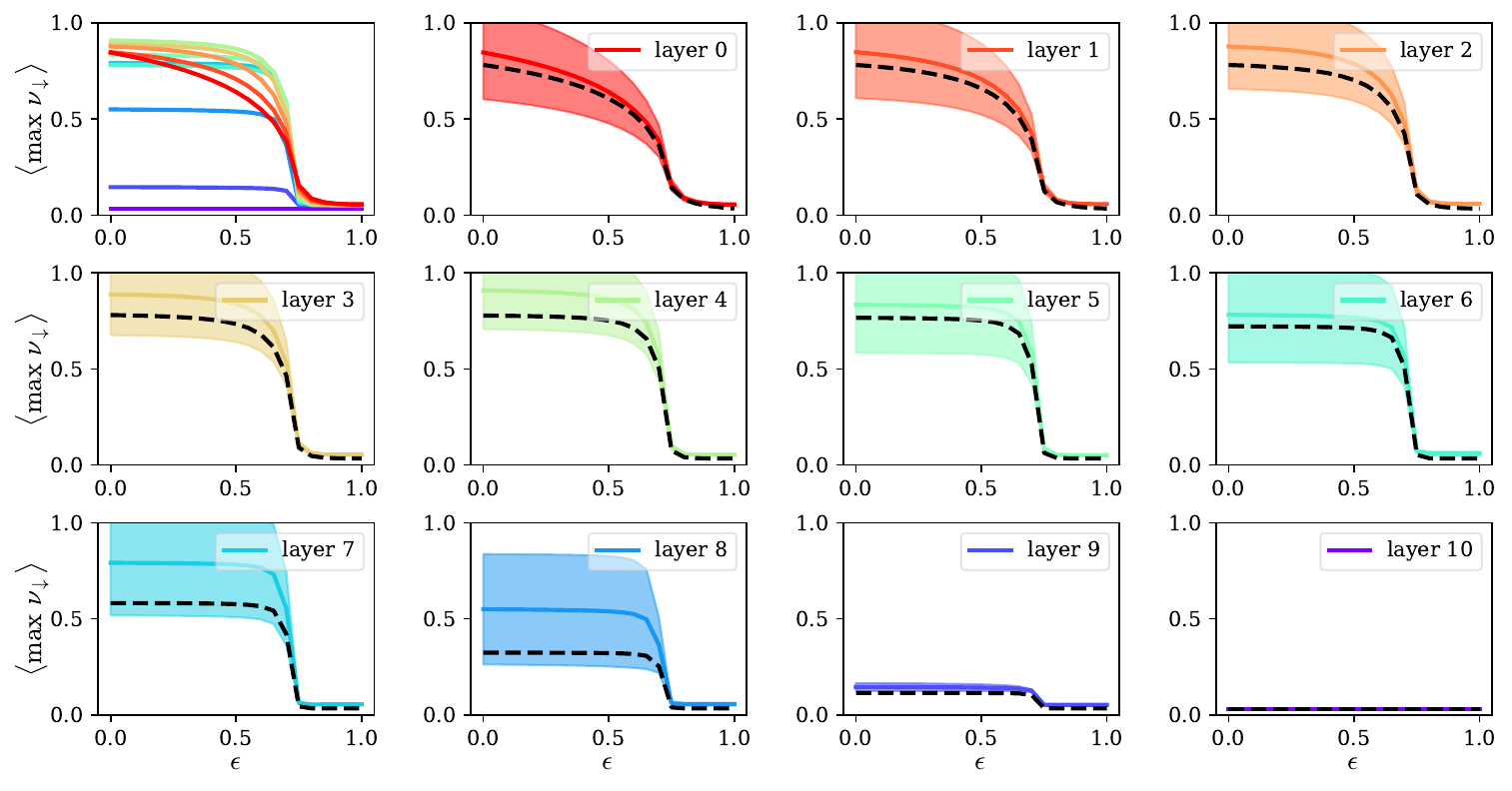}
    \caption{\textbf{Largest downward BP messages, averaged for each layer, for varying $\epsilon$. Data for the Random Hierarchical Model with the same parameters as \autoref{fig:upward_grid}.}
    Each layer, indicated in the legend, is represented with a different color, while the theoretical prediction from \autoref{eq:anneal_iteration} is represented with the black dashed line. We observe that the messages in the downward process have large fluctuations, as represented by their standard deviations, especially for small $\epsilon$. Still, the theory correctly captures the trend and becomes more accurate for increasing $\epsilon$. The top left panel represents all the layers together for comparison. Starting from the initialization $\nu_{\downarrow}=1/v$ at the top layer ($10$), we observe that the largest downward messages increase as we go to lower levels in the hierarchy only if $\epsilon$ is smaller than some threshold value.}
    \label{fig:downward_grid}
\end{figure}

Due to the randomness of the production rules, the messages $\nu_{\uparrow}(x)$, $\nu_{\downarrow}(x)$ are random variables that depend on the specific realization of the rules. Although their fluctuations are not captured by the averages computed in \autoref{eq:anneal_iteration}, we observe that $p_{\uparrow}^{(\ell)}$ and $p_{\downarrow}^{(\ell)}$ capture well the average behavior of the messages at a given layer. 
In \autoref{fig:up-mean_messages}, the values of $\nnu{\ell}(X_i^{(\ell)})$ are reported for the bottom $5$ levels of a Random Hierarchical Model with $L=10$, $s=2$, $v=32$, $m=8$, and noise level $\epsilon=0.5$. At each layer $\ell$, the index $i$ of the nodes goes from $1$ to $s^{L-\ell}$ and for each of them, there are $v=32$ messages $\nnu{\ell}$, one for each entry of the alphabet. At layer $\ell=0$, the messages $\nnu{0}$ are initialized according to \autoref{eq:belief}. After one iteration, at layer $\ell=1$, we observe that the largest messages at each node, those corresponding to the most probable features $x_i$, are spread around some mean value that is well captured by the theoretical prediction of \autoref{eq:anneal_iteration}. We observe that also for the upper layers, the average behavior of the largest messages is well captured by the theory.\\
The comparison between the theory and the BP algorithm for every $\epsilon$ is reported in the \autoref{fig:upward_grid} and \autoref{fig:downward_grid}. The upward iteration is reported in \autoref{fig:upward_grid} and shows an excellent agreement with the prediction of \autoref{eq:anneal_iteration}. In particular, going from the input layer $\ell=0$ to the class variable $\ell=L$ ($L=10$ in \autoref{fig:upward_grid}), the messages for the most probable features show a sharper transition at a threshold value $\epsilon^*$, which corresponds to the phase transition of the theoretical iteration map in \autoref{eq:iterUP} when $L\rightarrow\infty$.
The downward iteration in \autoref{fig:downward_grid} shows that the theory also captures the trend in $\epsilon$ of the downward messages. However, for small values of $\epsilon$, we observe that the messages have large fluctuations around their mean value. The reason for this behavior is that, in the Random Hierarchical Model, there is a number $m$ of synonyms $(x_1, \dots, x_s)$ that code for the same higher level feature $y$. Therefore, having perfect information on $y$ and on $x_2,\dots, x_s$ is not enough to perfectly reconstruct the value of $x_1$, thereby resulting in large fluctuations of the messages at small noise level $\epsilon$.
This is different from the upward process where having perfect information on $(x_1,\dots, x_s)$ allows the perfect reconstruction of $y$. As a result, the messages in the upward process are more concentrated around their mean than the downward messages.

\begin{figure}
    \centering
    \includegraphics[width=\textwidth]{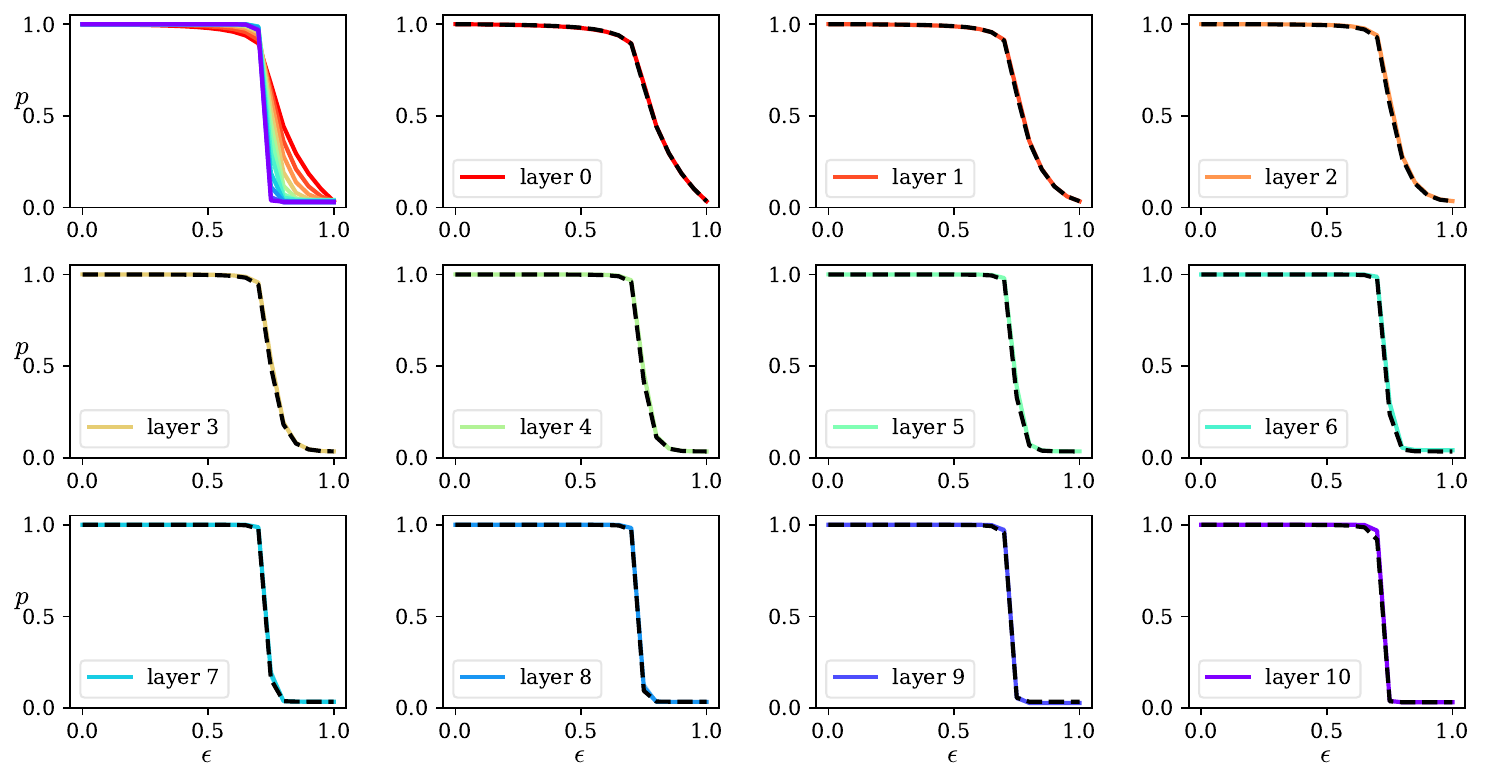}
    \vspace{-1cm}
    \caption{\textbf{Largest marginal probabilities computed by BP, averaged for each layer, for varying $\epsilon$. Data for the Random Hierarchical Model with the same parameters as \autoref{fig:upward_grid}.}
    Each layer, indicated in the legend, is represented with a different color, and the black dashed line is the theoretical prediction from \autoref{eq:theory_marginal}, which shows excellent agreement with the experiments. The top left panel represents all the layers together for comparison, where the inversion between the top and bottom layers can be observed (same curves as \autoref{fig:activations_RHM} in the main text).
}
    \label{fig:p_grid}
\end{figure}

\section{Mapping from time diffusion to $\epsilon$ noise}
\label{app:CC}

\begin{figure}[t]
    \centering
    \includegraphics[width=.45\textwidth]{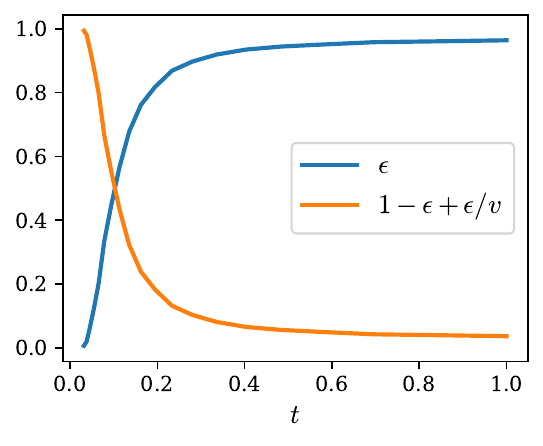}
    \includegraphics[width=.49\textwidth]{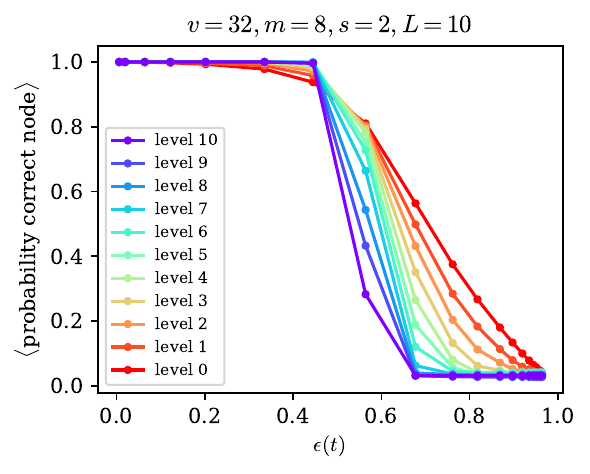}
    \caption{\textit{Left:} \textbf{Mapping between the $\epsilon$ values and the diffusion process.} From the average values of the beliefs at the leaves variables during the diffusion process at time $t$, we compute an effective $\epsilon(t)$ as $\epsilon(t)/v = \langle \nnu{0}(x)\rangle$, for the values $x$ different from the starting one, averaging over the realizations of the diffusion process.
    \textit{Right:} \textbf{Probability of reconstructing the initial values for the nodes at a given layer during the diffusion process in time, using the effective $\epsilon$ computed in the left panel.} We observe that the shape of the curves with respect to the effective noise $\epsilon(t)$ is qualitatively similar to that of the simplified $\epsilon$-process reported in \autoref{fig:activations_RHM}, supporting that it represents a good approximation for studying the diffusion process in time.}
    \label{fig:eps_time}
\end{figure}

In the diffusion process for the Random Hierarchy Model defined in \autoref{sec:bp}, the beliefs $\nnu{0}$ at the input variables vary stochastically in time, according to \autoref{eq:bayes}. 
Instead, in the simplified model of noise considered in \autoref{sec:mean-field}, at a given noise level $\epsilon$, these beliefs are fixed to two possible values (cf. \autoref{eq:belief-main}). To study whether the $\epsilon$-process is an effective approximation of the time diffusion process, we define an effective $\epsilon(t)$ depending on the reverse time of diffusion. At each input node $X^{(0)}_i$, we consider the upward messages $\nnu{0}(x)$ associated to the values $x$ that are different from the value of $X^{(0)}_i$ at time $t=0$. Denoting them as $\nu_t$, we define
\beq
    \frac{\epsilon(t)}{v} = \langle \nu_t \rangle,
\eeq
where the average $\langle \nu_t \rangle$ is performed over all the leaves variables $i$ and the realizations of the dynamics. $\epsilon(t)$ increases exponentially in time, according to the noise schedule used in the diffusion process, as shown in the left panel of \autoref{fig:eps_time}. The probability of correct reconstruction of a given node in the diffusion process is reported as a function of $\epsilon(t)$ in the right panel of \autoref{fig:eps_time}. We observe that the curves for different layers have similar behavior to those of the $\epsilon$-process presented in \autoref{fig:activations_RHM} of the main text, confirming that the latter is an effective approximation of denoising diffusion.

\section{Hidden activations for different architectures}
\label{app:resnets}

\begin{figure}[t]
    \centering
    \begin{tikzpicture}
        \node[anchor=north west,inner sep=0pt] at (0,0){
        \includegraphics[width=0.3\textwidth]{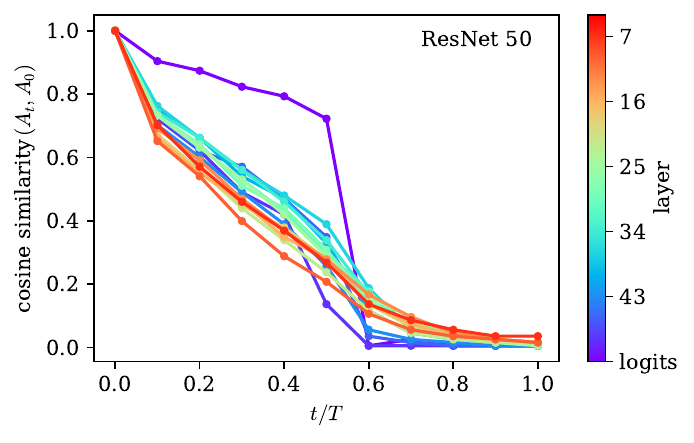}};
        \node[] at (-1ex,-1ex) {(a)};
    \end{tikzpicture}
    \begin{tikzpicture}
        \node[anchor=north west,inner sep=0pt] at (0,0){
        \includegraphics[width=0.3\textwidth]{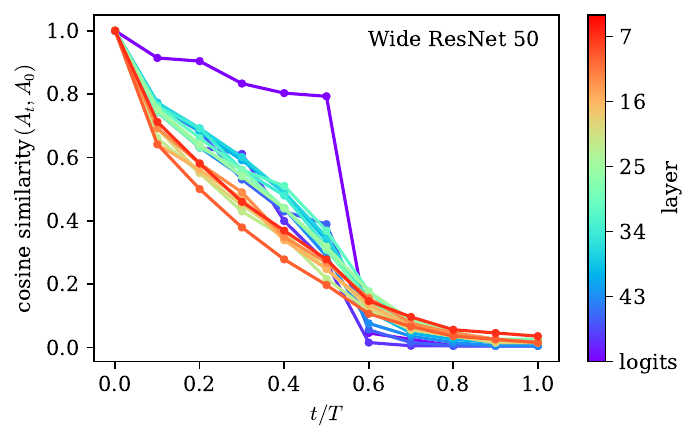}};
        \node[] at (-1ex,-1ex) {(b)};
    \end{tikzpicture}
    \begin{tikzpicture}
        \node[anchor=north west,inner sep=0pt] at (0,0){
        \includegraphics[width=0.3\textwidth]{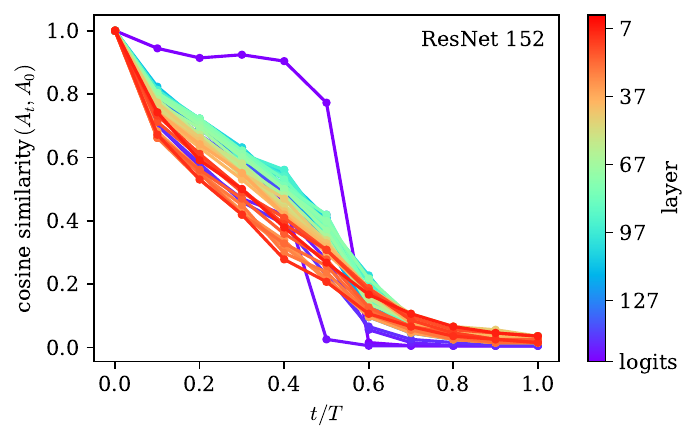}};
        \node[] at (-1ex,-1ex) {(c)};
    \end{tikzpicture}
    \caption{\textbf{Cosine similarity between the post-activations of the convolutional blocks of different ResNet architectures for the initial images $x_0$ and the synthesized ones $\hat{x}_{0}(t)$.} Specifically, panels (a), (b), and (c) correspond to ResNet50, Wide ResNet50, and ResNet152, respectively \cite{he_deep_2016}. As in \autoref{fig:imagenet-main} for the ConvNeXt architecture, the similarity between logits exhibits a sharp drop around $t \approx T/2$, indicating the change in class, while the hidden representations of the early layers change more smoothly.
    For computing the cosine similarity, all activations are standardized, i.e., centered around the mean and scaled by the standard deviation computed on the 50000 images of the ImageNet-1k validation set. At each time, the cosine similarity values correspond to the maximum of their empirical distribution over $10000$ images ($10$ per class of ImageNet-1k).
    }
    \label{fig:resnet}
\end{figure}

We perform the experiments described in \autoref{sec:forward-backward-exp} using the internal representations of different deep convolutional architectures trained for image classification on ImageNet-1k.
We consider the ResNet architecture \cite{he_deep_2016} with varying width and depth: a ResNet 50 achieving 95.4\% top-5 accuracy, a Wide ResNet 50 having 	
95.8\% top-5 accuracy, and a ResNet 152 having 96.0\% top-5 accuracy \cite{torchvision2016}.
The results of the experiments performed with the hidden representations of these architectures are reported in \autoref{fig:resnet} and show the same qualitative behavior as the one observed for the ConvNeXt architecture in \autoref{fig:imagenet-main}: the cosine similarity exhibits a sharp transition for the logits, while it decays smoothly for the hidden representations at early layers.

\section{Bi-modal distributions}
\label{app:gaussian-mixture}

In this section, we study the forward-backward experiments discussed in the main text, focusing on a bi-modal distribution without hierarchical and compositional structure. Specifically, we consider a $d$-dimensional Gaussian mixture with an initial probability density given by:
\beq \label{eq:gauss-mixt}
    q(\x_0) = \frac{1}{2(2 \pi \sigma^2)^{d/2}} \left[ \exp\left( - \frac{(\x_0-\mu)^\top(\x_0-\mu)}{2\sigma^2}\right) + \exp\left( - \frac{(\x_0+\mu)^\top(\x_0+\mu)}{2\sigma^2}\right)\right].
\eeq
We diffuse the data according to the dynamics described in Section 1.A of the main text, i.e.,
\beq
    \x_t = \sqrt{1-\beta_t} \x_{t-1} + \sqrt{\beta_t} \eta, \quad \eta \sim \mathcal{N}(\mathbf{0},\mathbf{I}).
\eeq
Thus, the \textit{forward dynamics} reads
\beq
    \x_t = \sqrt{\overline{\alpha_t}} \x_0 + \sqrt{1-\overline{\alpha_t}} \eta, \quad \eta \sim \mathcal{N}(\mathbf{0},\mathbf{I}).
\eeq
We then reverse the process at time $t$, following the exact \textit{backward dynamics}:
\beq
    \x_{t-1} = \frac{1}{\sqrt{\overline{\alpha_t}}} \left( \x_t + \beta_t \nabla_{\x} \log q(\x_t) \right) + \sqrt{\beta_t} \z, \quad \z \sim \mathcal{N}(\mathbf{0},\mathbf{I}),
\eeq
with the analytical \textit{score} function
\beq
    \nabla_{\x} \log q(\x_t) = -\frac{\x_t}{\overline{\alpha_t} \sigma^2 + 1 -\overline{\alpha_t}} + \frac{\mu \sqrt{\overline{\alpha_t}}}{\overline{\alpha_t} \sigma^2 + 1 - \overline{\alpha_t}} \tanh \left( \frac{\x_t^\top \mu \sqrt{\overline{\alpha_t}}}{\overline{\alpha_t} \sigma^2 + 1 - \overline{\alpha_t}} \right).
\eeq

As in our experiments on the ConvNeXt in Section 1 and on the deep CNN trained on the RHM in Section 4 of the main text, we examine how the internal representations of a neural network trained to classify the mode of the distribution vary when the input is obtained by inverting the forward dynamics at time $t$. 

We consider the Gaussian mixture defined in \autoref{eq:gauss-mixt} with $d=1024$, $\mu=(1,1,\dots,1)^\top$, and $\sigma=1$. We train a deep, fully-connected ReLU network with 6 hidden layers, each containing 64 neurons, using $P=2048$ training points until achieving zero training error. This network achieves $100\%$ accuracy on a test set with $n_{\mathrm{test}} = 1024$ samples. 

\begin{figure}[t]
    \centering
    \includegraphics[width=0.45\linewidth]{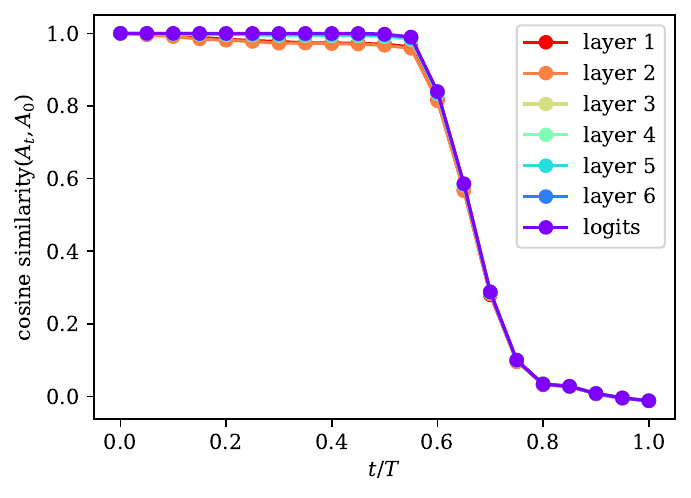}
    \caption{\textbf{Cosine similarity between the post-activations of the hidden layers of a deep fully-connected network for bi-modal data} $x_0 \in \mathbb{R}^d$ ($d=1024$) and the ones synthesized with forward-backward experiments $\hat{x}_0(t)$. Around $t \approx 0.6\,T$, the similarity between logits exhibits a drop, indicating a transition in the probability of changing the initial mode. In contrast to the RHM and natural images, the hidden representations of the hidden layers change like the logits. In particular, no crossing of the curves is observed. To compute the cosine similarity, all activations are standardized, i.e., centered around the mean and scaled by the standard deviation computed on 1000 initial samples.}
    \label{fig:gaussian}
\end{figure}

For each inversion time $t$, we compute the cosine similarity between the post-activations for the initial and generated points. \autoref{fig:gaussian} presents the resulting curves. Similar to the curves obtained for image and synthetic hierarchical data, the class similarity curve exhibits a drop at a characteristic time, as theoretically studied by Biroli et al. in \cite{biroli2023generative} and \cite{biroli2024dynamical}. However, unlike compositional data, the behavior of the curves corresponding to the internal layers follows the curve for the class. Specifically, there is no inversion of the similarity curves corresponding to early and deep layers, which is a phenomenon unique to compositional data (\autoref{fig:imagenet-main}) and cannot be captured by simple bi-modal distributions.

\chapter{Appendix: Probing Hidden Hierarchies in Data}

\section{The Random Hierarchy Model}
\label{app:probing-rhm}

\subsection{Denoising the RHM with Belief Propagation}
In the RHM, knowing the production rules of its tree structure, the Bayes-optimal denoising of its data can be done exactly using the Belief Propagation (BP) algorithm \citep{sclocchi2024phase}.

In the factor graph of the RHM tree, all latent and visible variables $\xv^{(\ell)}_i$ represent variable nodes, while the RHM rules connecting them are factor nodes.
Each variable node is associated to two BP messages, one coming from below, $\nu_{\uparrow}( \xv^{(\ell)}_i)$, and one coming from above, $\nu_{\downarrow}(\xv^{(\ell)}_i)$. The starting point of the BP algorithm is the definition of the messages at the boundaries of the tree, which are the upward messages at the leaves $\nu_{\uparrow}(\xv^{(0)}_i)$ and the downward message at the root node $\nu_{\downarrow}(\xv^{(L)}_1)$. Since we consider class-unconditional diffusion processes, we consider the latter as being uniform over the values of $\voc^{(L)}$. The initialization at the leaves, instead, corresponds to the prior belief on the values of the single visible tokens, which is given by the noisy observation. In the case of diffusion processes, the noisy observation $\x_t$ gives prior beliefs $\nu_{\uparrow}(\xv^{(0)}_i) = p(\hat{x}_{0,i} | x_{t,i})$, which can be computed for the single token by Bayes' rule and depends on the specific diffusion process under consideration.

\subsubsection{BP iteration}
\label{app:probing-BPiteration}

The initialization of BP is given by the leaf messages $\nu_{\uparrow}(\xv^{(0)}_i)$, $i \in [s^L]$.
For each $s$-patch at level $\ell$, e.g., $\{\xv^{(\ell)}_i\}_{i=1,\dots,s}$, having a common parent node at layer $\ell+1$, e.g. $\xv^{(\ell+1)}_1$, the upward message in the upper level is computed as:
\begin{align}
    \tilde{\nu}_{\uparrow}\lpa \xv^{(\ell+1)}_1=y\rpa = \sum_{a_1, \dots, a_s \in {\voc^{(\ell)}}^{\otimes s}} \psi^{(\ell+1)}\left(y, a_1, \dots, a_s\right) \prod_{i=1}^s\ \nu_{\uparrow}\lpa \xv^{(\ell)}_i=a_i\rpa,
    \label{eq:probing-nuUP}
\end{align}
\begin{align}
\nu_{\uparrow}(\xv^{(\ell+1)}_1=y) = \frac{\tilde{\nu}_{\uparrow}(\xv^{(\ell+1)}_1=y)}{\sum_{y'\in\voc^{(\ell+1)}} \tilde{\nu}_{\uparrow}(\xv^{(\ell+1)}_1=y')},
    \label{eq:probing-nuUP_norm}
\end{align}

where the factor $\rul{\ell+1}\left(y, a_1, \dots, a_s\right)$ reads
\begin{align}
\begin{aligned}
    \rul{\ell+1}(y, a_1, ..., a_s) = 
    \begin{cases}
        1, \quad \text{if } y \rightarrow (a_1, ..., a_s) \text{ is a rule at layer }(\ell+1)\rightarrow \ell\\
        0, \quad \text{otherwise}.
    \end{cases}
\end{aligned}
\label{eq:probing-factor}
\end{align}

This upward process is iterated from the leaf nodes at $\ell=0$ until the root node at $\ell=L$. 
Afterward, BP computes the downward messages.
The initialization at the root node is given by a uniform prior over the symbols of $\voc^{(L)}$, i.e.
\beq
    \nu_{\downarrow}\lpa \xv^{(L)}_1=a\rpa = \frac{1}{v}, \quad \forall a \in \voc^{(L)}.
\eeq

For the same $s$-patch at layer $\ell$ and parent node at layer $\ell+1$ as before, the downward message for $\xv^{(\ell)}_1$ is given by

\beq
    \tilde{\nu}_{\downarrow}(\xv^{(\ell)}_1=a_1) = \sum_{\substack{a_2, ..., a_s \in \mathcal{\voc^{(\ell)}}^{\otimes (s-1)}\\ y\in \voc^{(\ell+1)}}} \rul{\ell+1}(y, a_1, ..., a_s)\ \nu_{\downarrow}(\xv^{(\ell+1)}_1=y)\prod_{i=2}^s \nu_{\uparrow}(\xv^{(\ell)}_i=a_i) ,
    \label{eq:probing-nuDOWN}
\eeq
\beq
    \nu_{\downarrow}(\xv^{(\ell)}_1=a) = \frac{\tilde{\nu}_{\downarrow}(\xv^{(\ell)}_1=a)}{\sum_{a'\in\voc^{(\ell)}} \tilde{\nu}_{\downarrow}(\xv^{(\ell)}_1=a')},
    \label{eq:probing-nuDOWN_norm}
\eeq
with the same factor node of \autoref{eq:probing-factor}.

At the end of the upward-downward iteration, each variable node $\xv^{(\ell)}_i$ is associated with two BP messages for each symbol of the vocabulary $\voc^{(\ell)}$: $\nu_{\uparrow}(\xv^{(\ell)}_i)$ and $\nu_{\downarrow}(\xv^{(\ell)}_i)$.
Their product gives the marginal probability of the value of the node: 
\beq
    p(\xv^{(\ell)}_i=a) \propto \nu_{\uparrow}(\xv^{(\ell)}_i=a)\ \nu_{\downarrow}(\xv^{(\ell)}_i=a), \quad a\in\voc^{(\ell)}.
\eeq
These marginal probabilities are conditioned on the BP messages at the leaf nodes, which can come from a noisy observation of an RHM datum, as is the case for denoising diffusion.

Similarly, sampling from the posterior probabilities given by BP is done by starting sampling from the marginal probability at the root and then iteratively updating the marginal probabilities every time a new node is sampled \citep{mezardmontanari}.

\subsubsection{Priors at the leaves}
\label{app:probing-prior}

\paragraph{Masking diffusion}
Let's consider a datum $\x_0$ of the RHM undergoing masking diffusion. At any time $t$, the tokens of $\x_t$ can have value
\begin{align}
\begin{aligned}
    x_{t,i} &= x_{0,i}, \qquad\ \  \text{if token $i$ has not yet been masked};\\
    x_{t,i} &= [{\mathrm{MASK}}], \quad \text{if token $i$ has already been masked}.
\end{aligned}
\end{align}

Therefore, given the noisy observation $\x_t$, the prior belief $\nu_{\uparrow}(\xv_i^{(0)}=a)$ on the value of the token $i$ being equal to $a$ is given by $p(x_{0,i}\vert x_{t,i})$, that is:
\begin{align}
    \begin{aligned}
        \nu_{\uparrow}\left(\xv_i^{(0)} = a\right) &= \delta_{a,\overline{a}}, \quad \text{if } x_{t,i}=\overline{a}\in\voc^{(0)};\\
        \nu_{\uparrow}\left(\xv_i^{(0)} = a\right) &= 1/v, 
        \quad \forall a\in\voc^{(0)}\ \text{if } x_{t,i}=[{\mathrm{MASK}}].
    \end{aligned}
\label{eq:probing-up_masking}
\end{align}

\paragraph{$\epsilon$-process}
In this process, instead of running a forward diffusion process, we act directly on the leaf priors.
We introduce a noise-to-signal ratio $\epsilon \in [0,1]$, which controls the noise level instead of the diffusion time $t$.
Starting from a datum $\x_0$, whose $i$-th token has value $\overline{a}\in\voc^{(0)}$, the prior beliefs at the leaf node $i$ taking values in $\voc^{(0)}$ are defined as

\begin{align}
    \begin{aligned}
    \begin{cases}
        \nu_{\uparrow}\left(\xv^{(0)}_i = \overline{a} \right) &= 1-\epsilon +\epsilon/v, 
        \quad\quad\text{   for } x_{0,i}=\overline{a};\\
        \nu_{\uparrow}\left(\xv^{(0)}_i = a\right) &= \epsilon/v, 
        \quad\quad\quad\quad\qquad \forall a \in \voc^{(0)} \setminus {\overline{a}}.
    \end{cases}
    \end{aligned}
\label{eq:probing-up_epsilon}
\end{align}

The role of $\epsilon$ is to decrease the prior belief on the starting value of a token. 
This process can be interpreted as an averaged forward diffusion process, where the average is made over different forward trajectories. In the example of masking diffusion (\autoref{eq:probing-up_masking}), calling $1-\alpha_t$ the probability of a token being masked at time $t$, the average prior beliefs at the leaves read
\begin{align}
    \begin{aligned}
    \begin{cases}
    \left\langle \nu_{\uparrow}(\xv^{(0)}_i = \overline{a} )\right\rangle &= \alpha_t + \frac{1-\alpha_t}{v},
    \quad\quad\text{   where } x_{0,i}=\overline{a};\\
    \left\langle \nu_{\uparrow}(\xv^{(0)}_i = a) \right\rangle &=  \frac{1-\alpha_t}{v}, 
    \quad\quad\quad\quad\qquad \forall a \in \voc^{(0)} \setminus {\overline{a}},
    \end{cases}
    \end{aligned}
\label{eq:probing-up_average}
\end{align}

which have the same functional form as \autoref{eq:probing-up_epsilon} by identifying $\epsilon = 1-\alpha_t$. Both $\epsilon$ and $1-\alpha_t$ vary between $0$ and $1$ and play the role of noise-to-signal ratio in their respective processes. However, the fluctuations of the upward beliefs around their mean in the masking diffusion change the statistics of the BP messages propagating upwards and make the mapping $\epsilon = 1-\alpha_t$ inaccurate. For example, in the experimental data of \autoref{fig:probing-rhm}, the phase transition in the $\epsilon$-process is located at $\epsilon^*\simeq 0.74$, while it is found at $1-\alpha_{t^*} = t^*/T \simeq 0.3$ for masking diffusion.

\subsubsection{BP sampling vs backward diffusion}
\label{app:probing-sampling}
\paragraph{BP sampling}
As discussed at the end of section \ref{app:probing-BPiteration}, BP allows for sampling directly from the posterior probability $p(\hat{\x}_0 | {\x}_t)$.
Given a noisy observation $\x_t$ and the corresponding marginal probabilities $p(\xv^{(\ell)}_i|\x_t)$, the sampling proceeds as follows:
\begin{itemize}
    \item a root symbol $\xv^{(L)}_1=\hat{y}$, $\hat{y} \in \voc^{(L)}$, is sampled according to the probability $p(\xv^{(L)}_1|\x_t)$;
    \item the corresponding downward message is updated as $\nu_{\downarrow}\lpa \xv^{(L)}_1=y \rpa = \delta_{y,\hat{y}}$;
    \item the probabilities of the production rules $y \rightarrow (a_1, ..., a_s)$ form layer $L$ to layer $L-1$ are computed as
    \begin{align}
        p &\left( y \rightarrow a_1, ..., a_s | \x_t, \xv^{(L)}_1=\hat{y} \right) \nonumber \\
        &\propto \nu_{\downarrow}\lpa \xv^{(L)}_1 = y \rpa 
        \nu_{\uparrow}\lpa\xv^{(L-1)}_1=a_1\rpa \cdots \nu_{\uparrow}\lpa\xv^{(L-1)}_s=a_s\rpa.
        \label{eq:probing-prod_rule}
    \end{align}
    Notice that the upward messages $\nu_{\uparrow}\lpa\xv^{(L-1)}_i=a_i\rpa$ carry the information on the observation $\x_t$;
    \item a production rule $y \rightarrow (a_1, ..., a_s)$ is sampled according to the probabilities of \autoref{eq:probing-prod_rule}. This gives the values $\hat{a}_i\in \voc^{(L-1)}$ of the latent nodes $\xv^{(L-1)}_i$. The corresponding downward messages are updated as $\nu_{\downarrow}\lpa \xv^{(L-1)}_i=a \rpa = \delta_{a,\hat{a}_i}$;
    \item the probabilities of the production rules from layer $L-1$ to $L-2$ are computed as in \autoref{eq:probing-prod_rule};
    \item the sampling procedure continuous up to the visible layer $\xv^{(0)}_i$, giving a leaf configuration $\hat{\x}_0$.
\end{itemize}
The obtained sequence $\hat{\x}_0$ is a configuration of the RHM sampled from the posterior $p(\hat{\x}_0|\x_t)$.

\paragraph{Backward diffusion with BP}
The BP sampling above is equivalent to running the backward dynamics with the true score function of the RHM. In fact, given a noisy observation $\x_t$ at time $t$, the marginal probabilities $p(\xv^{(\ell)}_i{=}a | \x_t)$ at the visible nodes can be used to compute the expectation values $\mathbb{E}(\xv^{(\ell)}_i | \x_t)$, which corresponds to $\mathbb{E}(\hat{\x}_0 | \x_t)$. This expectation gives the score function at $\x_t$ at time $t$, which can be used in the backward dynamics to sample $\x_{t-1}$ at time $t-1$, and so on.

\begin{figure}[t!]
    \centering
    \begin{tikzpicture}
        \node[anchor=north west,inner sep=0pt] at (0,0){
        \includegraphics[width=0.49\textwidth]{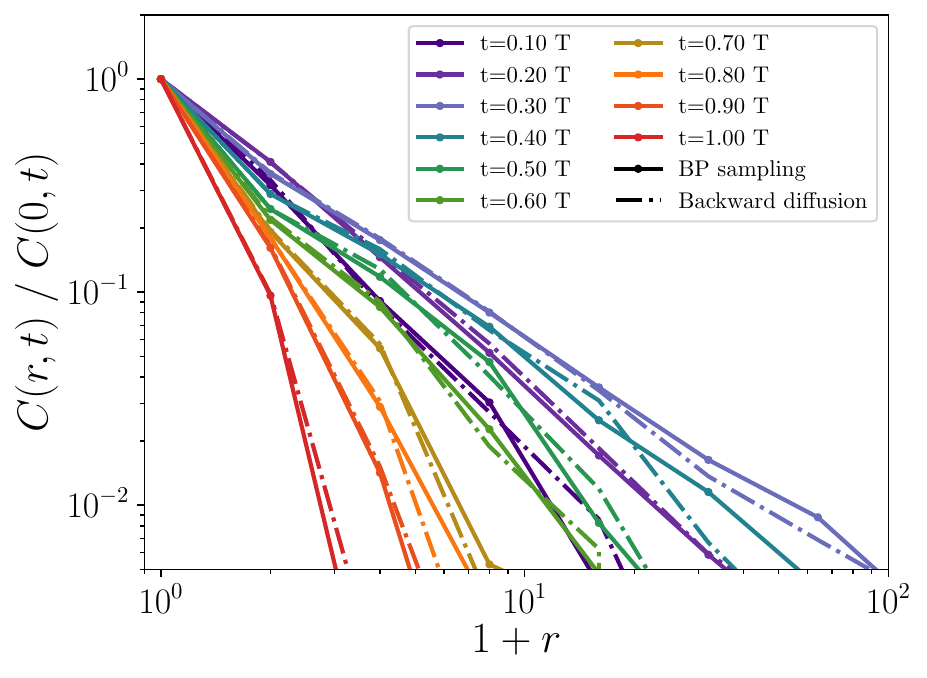}};
        \node at (0ex,0ex) {{(a)}};
        \node[anchor=north west,inner sep=0pt] at (7,0){
        \includegraphics[width=0.47 \textwidth]{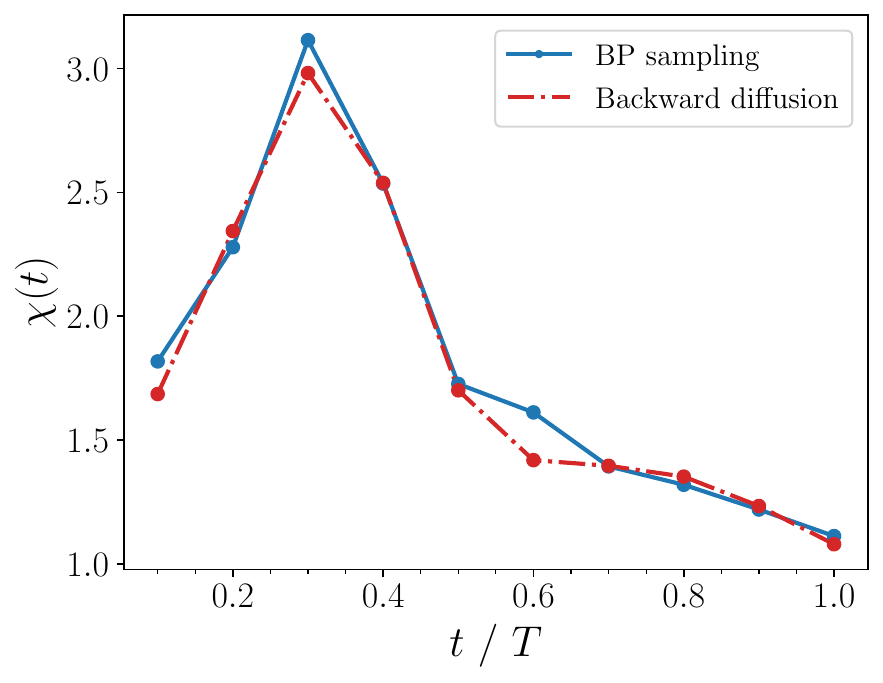}};
        \node at (6.5,0ex) {{(b)}};
    \end{tikzpicture}
\caption{
\textbf{Comparison between BP sampling and backward diffusion for masking in the Random Hierarchy Model (RHM).} 
Forward-backward experiments with masking diffusion, where the sampling from the posterior $p(\hat{\x}_0|\x_t)$ is done with BP sampling (continuous lines) or by running the backward diffusion dynamics (dotted-dashed lines), using the score function given by BP. Both the average correlation functions of changes (panel \textit{(a)}) and the dynamical susceptibility (panel \textit{(b)}) for different masking fractions $t/T$ do not depend on the sampling process. 
Data for RHM parameters $v=32$, $m=8$, $s=2$, $L=8$, averaged over $32$ starting data and $256$ diffusion trajectories per starting datum.}
\label{fig:probing-BPvsBack}
\end{figure}

\autoref{fig:probing-BPvsBack} compares BP sampling and the backward diffusion with the exact score function in the case of masking diffusion. Both the average correlation functions and the dynamical susceptibility at different masking fractions $t/T$ show the same behavior, independently of the sampling procedure. 

\subsection{Mean-field theory of the $\epsilon$-process}
\label{app:probing-meanfield}

\paragraph{Computation of the marginal probabilities}
Starting from \autoref{eq:probing-up_epsilon} and the BP iterative equations, \citet{sclocchi2024phase} computed the average BP messages at each layer $\ell$, where the average is performed over the possible choices of the RHM rules. 
The result consists in the average messages associated with reconstructing the starting value $\overline{a} \in\voc^{(\ell)}$ of a latent node $\xv^{(\ell)}_i$,
\begin{align}
    \left\langle\nu_{\uparrow}\left(\xv^{(\ell)}_i = \overline{a} \right)\right\rangle_{\psi} = p_\ell,
    \quad
    \left\langle\nu_{\downarrow}\left(\xv^{(\ell)}_i = \overline{a} \right)\right\rangle_{\psi} = q_\ell,
\end{align}

where the average $\langle\dots\rangle_{\psi}$ is performed over the factor nodes $\psi$ representing the randomly chosen rules of the RHM.
The values of $p_\ell$ and $q_\ell$ can be computed layer-by-layer through the following iterative maps:

\beq
    p_{\ell+1} = F(p_\ell),
    \quad
    q_{\ell-1} = G(q_\ell, p_{\ell-1}),
    \label{eq:probing-app_itermap}
\eeq
where 
\begin{align}
\label{eq:probing-F_iter}
    F(p) &= \frac{p^s + f \frac{m-1}{mv-1}(1-p^s)}{p^s + f (1-p^s)},\\
    G(q,p) &= \frac{q\ p^{s-1} + f \frac{m-q}{mv-1}(1-p^{s-1})}{q\ p^{s-1} + f \frac{m-q}{mv-1}(1-p^{s-1}) + (v-1)f \frac{m-q}{mv-1}},
\end{align}
and $f=\frac{mv-1}{v^s-1}$.
The initial conditions are given by 
\beq
p_0 = 1-\epsilon +\epsilon/v,
\label{eq:probing-p0}
\eeq
\beq
q_{L} = 1/v.
\label{eq:probing-qL}
\eeq
Notice that the expectation values $p_{\ell}$ and $q_{\ell}$ only depend on the layer $\ell$ and not on the specific position of the node $i$ inside the layer. 
Once $p_{\ell}$ and $q_{\ell}$ have been computed for every layer $\ell=0,\dots, L$, the average marginal probability of reconstructing the original value $\overline{a}\in\voc^{(\ell)}$ of the variable $\xv^{(\ell)}$ is given by
\beq
    P(\xv^{(\ell)} = \overline{a}) = \frac{p_{\ell} q_{\ell}}{p_{\ell} q_{\ell} + \frac{(1-p_{\ell})(1-q_{\ell})}{v-1}}.
    \label{eq:probing-p_marg}
\eeq
This marginal probability is conditioned on the initialization of the leaf nodes (\autoref{eq:probing-up_epsilon}) and only depends on the layer $\ell$, not on the position of the node inside the layer.
Given the initialization of $q_L$, the probability of reconstructing the root node $P(\xv^{(L)} = \overline{a})$, that is the class of the datum, is given by
\beq
    P(\xv^{(L)} = \overline{a}) = p_L.
\eeq

Therefore, in the limit of large depth $L\rightarrow \infty$, the value of $p_L$ is given by one of the fixed of the iterative map $F(p)$.
When $F'(1)>1$, $F(p)$ has two fixed points: $p=1$, which is repulsive, and $p=1/v$, which is attractive. This implies that, in this regime, for any noise level $\epsilon>0$ at the leaf nodes, it is impossible to reconstruct the value of the class better than random chance.
Instead, when $F'(1)<1$, that is
\beq
    s\ m \frac{v-1}{v^s-1}<1,
\label{eq:probing-trans_cond}
\eeq
a third non-trivial fixed point $p^*=F(p^*)$ appears, which is repulsive,  while both $p=1$ and $p=1/v$ are now attractive. This implies the presence of a phase transition at a specific noise level $\epsilon^*=\frac{1-p^*}{1-1/v}$. For $\epsilon<\epsilon^*$, the class is reconstructed, for $\epsilon>\epsilon^*$ it is not.

\paragraph{Computation of the correlation functions}
Similar to the marginal probabilities, the average correlation function can also be computed through an annealed average over the RHM rules. Let's consider two leaf nodes $\xv_i^{(0)}$ and $\xv_j^{(0)}$ connected to the common ancestor $\xv_1^{(\tilde{\ell})}$ at layer $\tilde{\ell}$ through the nodes $\xv_{1}^{(\tilde{\ell}-1)}$ and $\xv_{2}^{(\tilde{\ell}-1)}$. Given the tree structure, their joint probability distribution can be written as
\begin{align}
\begin{aligned}
    &P(\xv_i^{(0)}, \xv_j^{(0)}) =\\ 
    &\sum_{\xv_{l}^{(\tilde{\ell}-1)}, \xv_{m}^{(\tilde{\ell}-1)}}
    P\lpa \xv_i^{(0)} \vert \xv_l^{(\tilde{\ell}-1)}\rpa 
    P\lpa \xv_j^{(0)} \vert \xv_m^{(\tilde{\ell}-1)}\rpa
    \sum_{\xv_{k}^{(\tilde{\ell})}} 
    P\lpa \xv_l^{(\tilde{\ell}-1)}, \xv_m^{(\tilde{\ell}-1)} \vert \xv_k^{(\tilde{\ell})}\rpa
    P\lpa \xv_k^{(\tilde{\ell})} \rpa
\end{aligned}
\label{eq:probing-jointP}
\end{align}

In the mean-field approach, the average joint probability only depends on the tree-distance $\tilde{\ell}$ between $i$ and $j$ and not their precise location. Moreover, we are only interested in the probability that both the starting values of $\xv_i^{(0)}$, $\xv_j^{(0)}$ are reconstructed, and the probability of only one of the two is reconstructed. In the following, we use an overline $\overline{\cdot}$ to indicate the starting value of a variable to be reconstructed.
We need to compute
\begin{align}
    &\left\langle P(\xv_i^{(0)}=\overline{a}_i, \xv_j^{(0)}=\overline{a}_j) \right\rangle_{\psi},\\
    &\left\langle P(\xv_i^{(0)}=\overline{a}_i, \xv_j^{(0)}\neq\overline{a}_j) \right\rangle_{\psi} = \left\langle P(\xv_i^{(0)}\neq\overline{a}_i, \xv_j^{(0)}=\overline{a}_j) \right\rangle_{\psi},
\end{align}
where the average $\langle\dots\rangle_{\psi}$ is done over the possible choices of RHM rules. 
Using the same strategy for the computation of the marginal probabilities, we compute the average of each term in \autoref{eq:probing-jointP} by substituting the BP messages with their averages. 
For this purpose, we first define the average marginal conditioned on the downward messages at layer $\hat{\ell}$, $P(\xv^{(\ell)} = \overline{a}^{\ell} | q_{\hat{\ell}}=c)$, with $\ell<\hat{\ell}$. This is computed with \autoref{eq:probing-p_marg} by iterating the equations \ref{eq:probing-app_itermap} between layers $0$ and $\hat{\ell}$ and using the initial conditions of \autoref{eq:probing-p0} and $q_{\hat{\ell}} = c$. Therefore, the marginals of \autoref{eq:probing-p_marg} correspond to $P(\xv^{(\ell)} = \overline{a}^{\ell} | q_{L}=1/v)$.
For the marginals in \autoref{eq:probing-jointP} we have:
\beq
\langle P\lpa \xv_k^{(\tilde{\ell})} = \overline{a}^{(\tilde{\ell})}_k \rpa\rangle_{\psi} = P(\xv^{(\tilde{\ell})} = \overline{a}^{(\tilde{\ell})}| q_{L}=1/v),
\label{eq:probing-marginal_2}
\eeq
that is the average marginal computed in \autoref{eq:probing-p_marg};
\beq
\left\langle P\lpa \xv_i^{(0)} = \overline{a}_i \vert \xv_l^{(\tilde{\ell}-1)}=\overline{a}^{(\tilde{\ell}-1)}_l\rpa\right\rangle_{\psi} = 
P(\xv^{(0)} = \overline{a}| q_{\tilde{\ell}-1}=1),
\label{eq:probing-cond_marg1}
\eeq
\beq
\left\langle P\lpa \xv_i^{(0)} = \overline{a}_i \vert \xv_l^{(\tilde{\ell}-1)}\neq\overline{a}^{(\tilde{\ell}-1)}_l\rpa\right\rangle_{\psi} = 
P(\xv^{(0)} = \overline{a} | q_{\tilde{\ell}-1}=0).
\label{eq:probing-cond_marg2}
\eeq

The probability terms of the type $P(\xv^{(0)} \neq \overline{a} | \dots)$ are given by $1-P(\xv^{(0)} \neq \overline{a} | \dots)$.
Since these averages only depend on the layer level $\tilde{\ell}$, they are the same for $\left\langle P\lpa \xv_j^{(0)} \vert \xv_m^{(\tilde{\ell}-1)}\rpa\right\rangle_{\psi}$.
The last term to compute is the joint $P\lpa \xv_l^{(\tilde{\ell}-1)}, \xv_m^{(\tilde{\ell}-1)} \vert \xv_k^{(\tilde{\ell})}\rpa$ which can be expressed in terms of BP messages:

\begin{align}
\begin{aligned}
    &P\lpa \xv_l^{(\tilde{\ell}-1)}=a_l, \xv_m^{(\tilde{\ell}-1)}=a_m \vert \xv_k^{(\tilde{\ell})}=y\rpa \propto \\
    &\sum_{\substack{a_{m+1}, ..., a_s \in \mathcal{\voc^{(\ell)}}^{\otimes (s-2)}}} \rul{\ell}(y, a_l, a_m,\dots, a_s)\  
    \nu_{\uparrow}(\xv^{(\ell)}_l=a_l)\ 
    \nu_{\uparrow}(\xv^{(\ell)}_m=a_m)
    \prod_{i\neq l,m}^s \nu_{\uparrow}(\xv^{(\ell)}_i=a_i).
\end{aligned}
\end{align}

Computing the averages over the rules, we have:
\begin{align*}
    &\left\langle P\lpa \xv_l^{(\tilde{\ell}-1)}=\overline{a}_l, \xv_m^{(\tilde{\ell}-1)}=\overline{a}_m \vert \xv_k^{(\tilde{\ell})}=\overline{y}\rpa\right\rangle_{\psi} =
    p_{\ell-1}^2\ /\ Z_{\overline{y}}^{(\tilde{\ell}-1)}, \\
    &\left\langle P\lpa \xv_l^{(\tilde{\ell}-1)}=\overline{a}_l, \xv_m^{(\tilde{\ell}-1)}\neq \overline{a}_m \vert \xv_k^{(\tilde{\ell})}=\overline{y}\rpa\right\rangle_{\psi} =
    f\ p_{\ell - 1} (1-p_{\ell - 1}) \frac{m-1}{mv-1}\ /\ Z_{\overline{y}}^{(\tilde{\ell}-1)},\\
    &\left\langle P\lpa \xv_l^{(\tilde{\ell}-1)}\neq\overline{a}_l, \xv_m^{(\tilde{\ell}-1)}\neq \overline{a}_m \vert \xv_k^{(\tilde{\ell})}=\overline{y}\rpa\right\rangle_{\psi} =
    f\ (1-p_{\ell - 1})^2 \frac{m-1}{mv-1}\ /\ Z_{\overline{y}}^{(\tilde{\ell}-1)},\\
    &Z_{\overline{y}}^{(\tilde{\ell}-1)} = p_{\ell-1}^2 + f \frac{m-1}{mv-1} (1-p_{\ell-1}^2)
\end{align*}
\begin{align*}
    &\left\langle P\lpa \xv_l^{(\tilde{\ell}-1)}=\overline{a}_l, \xv_m^{(\tilde{\ell}-1)}=\overline{a}_m \vert \xv_k^{(\tilde{\ell})}\neq\overline{y}\rpa\right\rangle_{\psi} =
    0, \\
    &\left\langle P\lpa \xv_l^{(\tilde{\ell}-1)}=\overline{a}_l, \xv_m^{(\tilde{\ell}-1)}\neq \overline{a}_m \vert \xv_k^{(\tilde{\ell})}\neq\overline{y}\rpa\right\rangle_{\psi} =
    f\ p_{\ell - 1} (1-p_{\ell - 1}) \frac{m}{mv-1}\ /\ Z_{y}^{(\tilde{\ell}-1)},\\
    &\left\langle P\lpa \xv_l^{(\tilde{\ell}-1)}\neq\overline{a}_l, \xv_m^{(\tilde{\ell}-1)}\neq \overline{a}_m \vert \xv_k^{(\tilde{\ell})}\neq\overline{y}\rpa\right\rangle_{\psi} =
    f\ (1-p_{\ell - 1})^2 \frac{m}{mv-1}\ /\ Z_{y}^{(\tilde{\ell}-1)},\\
    &Z_{y}^{(\tilde{\ell}-1)} = f \frac{m}{mv-1} (1-p_{\ell-1}^2) 
\end{align*}

We can combine these terms with the marginals \autoref{eq:probing-marginal_2} to obtain $\left\langle P\lpa \xv_l^{(\tilde{\ell}-1)}, \xv_m^{(\tilde{\ell}-1)}\rpa\right\rangle_{\psi}$.
We can write this probabilities in a $2\times 2$ matrix $\mC^{(\tilde{\ell}-1)}$ such that:
\begin{align}
    C_{11}^{(\tilde{\ell}-1)} = 
    \left\langle P\lpa \xv_l^{(\tilde{\ell}-1)}=\overline{a}_l, \xv_m^{(\tilde{\ell}-1)}=\overline{a}_m \rpa\right\rangle_{\psi},
\end{align}
\begin{align}
    C_{12}^{(\tilde{\ell}-1)} = C_{21}^{(\tilde{\ell}-1)} =
    \left\langle P\lpa \xv_l^{(\tilde{\ell}-1)}=\overline{a}_l, \xv_m^{(\tilde{\ell}-1)}\neq\overline{a}_m \rpa\right\rangle_{\psi},
\end{align}
\begin{align}
    C_{22}^{(\tilde{\ell}-1)} = 
    \left\langle P\lpa \xv_l^{(\tilde{\ell}-1)}\neq\overline{a}_l, \xv_m^{(\tilde{\ell}-1)}\neq\overline{a}_m \rpa\right\rangle_{\psi}.
\end{align}

Similarly, also the conditional marginals of \autoref{eq:probing-cond_marg1}-\ref{eq:probing-cond_marg2} can be written as a $2\times 2$ matrix $\mT^{(\tilde{\ell}-1)}$: 

\beq
    T^{(\tilde{\ell}-1)}_{11} = \left\langle P\lpa \xv_i^{(0)} = \overline{a}_i \vert \xv_l^{(\tilde{\ell}-1)}=\overline{a}^{(\tilde{\ell}-1)}_l\rpa\right\rangle_{\psi},
\eeq
\beq
    T^{(\tilde{\ell}-1)}_{12} = \left\langle P\lpa \xv_i^{(0)} = \overline{a}_i \vert \xv_l^{(\tilde{\ell}-1)}\neq\overline{a}^{(\tilde{\ell}-1)}_l\rpa\right\rangle_{\psi},
\eeq
\beq
    T^{(\tilde{\ell}-1)}_{21} = \left\langle P\lpa \xv_i^{(0)} \neq \overline{a}_i \vert \xv_l^{(\tilde{\ell}-1)}=\overline{a}^{(\tilde{\ell}-1)}_l\rpa\right\rangle_{\psi},
\eeq
\beq
    T^{(\tilde{\ell}-1)}_{22} = \left\langle P\lpa \xv_i^{(0)} \neq \overline{a}_i \vert \xv_l^{(\tilde{\ell}-1)}\neq\overline{a}^{(\tilde{\ell}-1)}_l\rpa\right\rangle_{\psi}.
\eeq

Collecting the values of $\langle P(\xv_i^{(0)}, \xv_j^{(0)})\rangle_{\psi}$ into a $2\times 2$ matrix $\mP(\xv_i^{(0)}, \xv_j^{(0)})$, we finally obtain
\beq
\mP(\xv_i^{(0)}, \xv_j^{(0)}) = \mT^{(\tilde{\ell}-1)}\ \mC^{(\tilde{\ell}-1)}\ {\mT^{(\tilde{\ell}-1)}}^{\top}.
\eeq

In the language of the spin variables introduced in \autoref{sec:probing-blocks}, the probability of reconstructing a variable $\xv_i^{(0)}=\overline{a}_i$ is the probability that $\sigma^{0}_i=+1$, while $\xv_i^{(0)}\neq\overline{a}_i$ corresponds to $\sigma^{0}_i=-1$.

\begin{figure}
    \centering
    \includegraphics[width=0.7\linewidth]{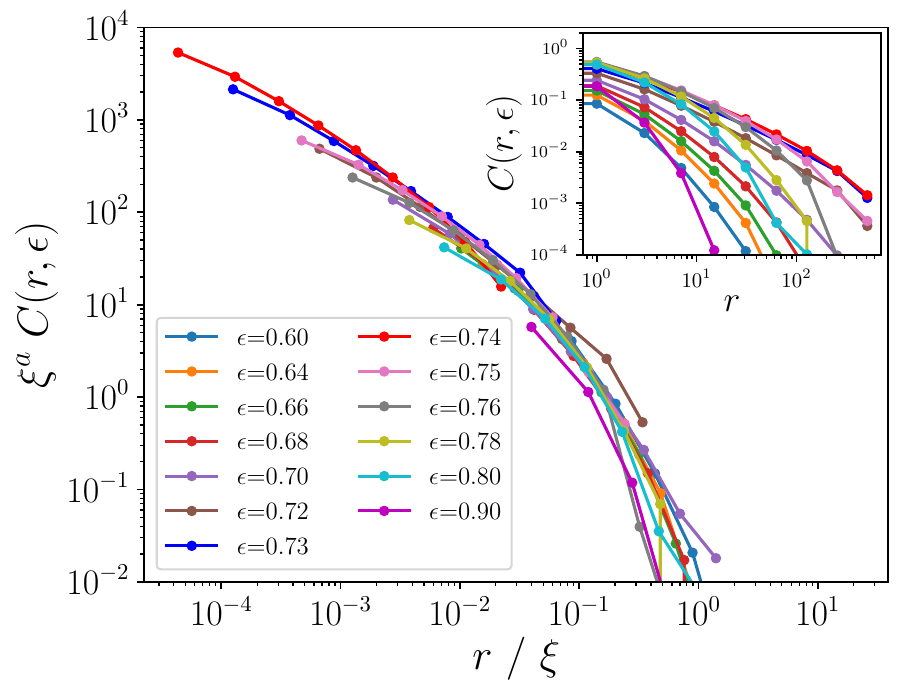}
    \caption{\textbf{$\epsilon$-process in the RHM ($v=32$, $m=8$, $s=2$, $L=9$): correlation function with respect to the token distance $r$, for noise levels $\epsilon$ close to the transition $\epsilon^*\simeq 0.74$.} 
    \textit{(Inset)} The correlation function displays system-spanning power-law decay at the transition $\epsilon^*\simeq 0.74$, while it decays faster for noise values $\epsilon\neq\epsilon^*$. The length scale at which it departs from the critical behavior defines the correlation length $\xi$.
    \textit{(Main)} Rescaling the distance $r$ with $\xi$ given by \autoref{eq:probing-xi-main} and $C(r,\epsilon)$ with $\xi^{a}$, $a=1$, the different correlation functions collapse on a single curve.
    This implies that the power-law scaling $\xi\sim|\Delta \epsilon|^{-\nu}$ of \autoref{eq:probing-xi-main} describes well the peaking of the correlation length around the class transition. 
    For this choice of RHM parameters, $\nu \simeq 1.78$.
    The exponent $a=1$ is obtained by fitting the critical decay $C(r, \epsilon^*)\sim r^{-a}$ from the data.
}
    \label{fig:probing-correlation_length}
\end{figure}

\begin{figure}
    \centering
    \includegraphics[width=0.7\linewidth]{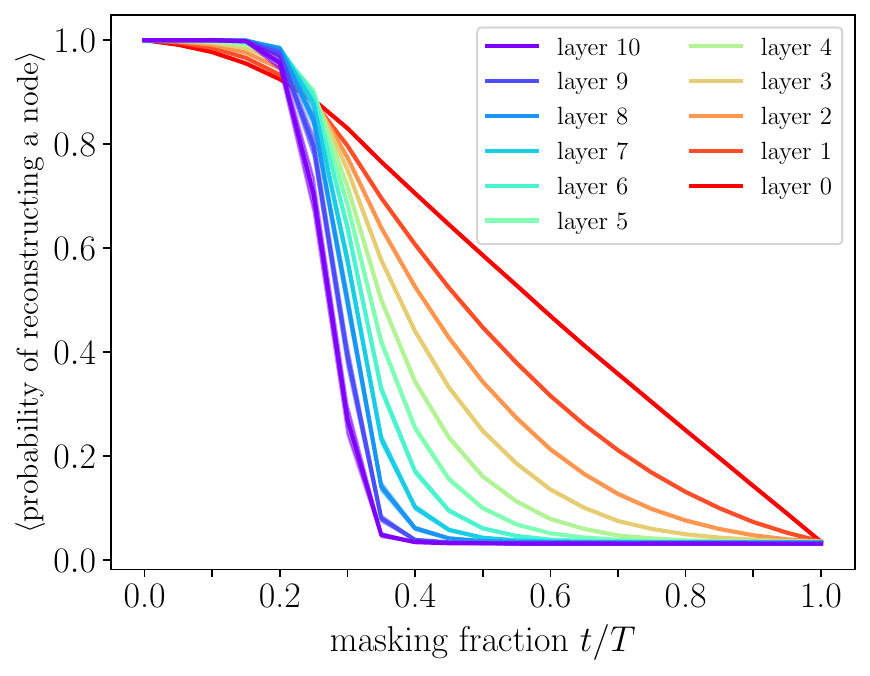}
    \caption{\textbf{Masking diffusion in the RHM: probability of reconstructing a (latent) node as a function of the inversion time $t$.} This is proportional to the masking fraction $t/T$.
    The probability is averaged over the nodes at a given layer.
    The probability of reconstructing a leaf node (layer $0$) decreases smoothly with the inversion time, while the probability of reconstructing the root node (layer $10$), that is the datum class, undergoes a sharper decay from $1$ to $1/v$ at a critical time $t^*\simeq 0.2\div 0.3\ T$. This sharp decay is expected to become a step-like transition in the limit of infinite depth $L\rightarrow\infty$. Data for RHM parameters $v=32$, $m=8$, $s=2$, $L=10$, averaged over $10$ diffusion trajectories per $10$ starting data $\x_0$.}
    \label{fig:probing-maksing_inversion}
\end{figure}

\section{Gaussian random field model}
\label{app:probing-random}

Consider $u \in [-1,1]^d$. Let $\x(u)$ denote a centered Gaussian random field defined over this domain with translational-invariant isotropic covariance function $K(u,u')$. Specifically, the field satisfies $\mathbb{E}[\x(u)]=0$ and $\mathbb{E}[\x(u)\x(u')]=K(u,u')=c(\|u-u'\|)$, where $c$ is a function depending only on the Euclidean distance $\|u-u'\|$. 

Assume that the Fourier coefficients $C(k)$ of $c(\|u-u'\|)$ satisfy, for large $\|k\|$, $C(k) = \gamma\|k\|^{-a}+o(\|k\|^{-a}), \; \|k\|\to\infty$, with $0{<}a{<}d$. This implies that the Fourier coefficients $\X(k)$ are independent Gaussian random variables, $\X(k) \sim \mathcal{N}(0,\sigma_k^2)$ with $\sigma_k^2 \asymp \|k\|^{-a}$.

\subsection{Forward-backward experiments in Fourier space} Given the independence of the Fourier coefficients  $\X(k)$, we apply the diffusion dynamics to each Fourier coefficient independently. The noising process is given by:
\begin{equation}
    \X(k)_t = \sqrt{1-\beta_t} \X(k)_{t-1} + \sqrt{\beta_t} \eta, \quad \eta \sim \mathcal{N}(0,1),
\end{equation}
for $t = 1,2,\dots,T$, where $\beta_t \in (0,1)$ are the diffusion coefficients and $\eta$ are independent standard Gaussian variables. 

By unrolling the recursion, the \textit{forward dynamics} can be expressed as
\begin{equation}
    \X(k)_t = \sqrt{\overline{\alpha_t}} \X(k)_0 + \sqrt{1-\overline{\alpha_t}} \eta, \quad \eta \sim \mathcal{N}(0,1),
\end{equation}
where $\overline{\alpha_t}=\prod_{t'=1}^t(1-\beta_{t'})$.

We then reverse the process at time $t$, following the \textit{backward dynamics}:
\begin{equation}
    \X(k)_{t-1} = \frac{1}{\sqrt{\overline{\alpha_t}}} \left( \X(k)_t + \beta_t \nabla_{\X(k)} \log q(\X(k)_t) \right) + \sqrt{\beta_t} z, \quad z \sim \mathcal{N}(0,1),
\end{equation}
where $q(\X(k)_t)$ is marginal probability density of $\X(k)_t$ in the forward process and $\nabla_{\X(k)} \log q(\X(k)_t)$ is the corresponding \textit{score function}.

Given the forward process, $q(\X(k)_t)$ is Gaussian and the score function can be computed explicitly:
\begin{equation}
    \nabla_{\X(k)} \log q(\X(k)_t) = -\frac{\X(k)_t}{\overline{\alpha_t} \sigma_k^2 + 1 -\overline{\alpha_t}}.
\end{equation}

\subsection{Mode retrieval} Our goal is to determine which Fourier coefficients are retrieved after the reverse process. Specifically, we want to compute the modes $k$ for which the distance between the coefficient obtained at the end of the backward process $\widehat{\X}(k,t)_{0} \sim p(\cdot | \X(k)_t)$ with the starting coefficient $\X(k)_0$ is small:
\begin{equation}
    |\widehat{\X}(k,t)_{0} - \X(k)_0| \ll 1.
\end{equation}
Thus, we consider the signal-to-noise ratio (SNR) for each mode $k$
\begin{equation}
    {\mathrm{SNR}}(\kappa,t) = \frac{\kappa^{-a}}{\overline{\alpha_t}^{-1}-1},
\end{equation}
where $\kappa=\|k\|$.

Define the critical wavevector magnitude $\kappa^*$ where ${\mathrm{SNR}}(\kappa^*,t)=1$:
\begin{equation}
    \kappa^* = \left(\overline{\alpha_t}^{-1}-1\right)^{-1/a}
\end{equation}
Modes with $\kappa<\kappa^*$ (low-frequency modes) have ${\mathrm{SNR}} > 1$ and can be retrieved, while modes with $\kappa>\kappa^*$ (high-frequency modes) have ${\mathrm{SNR}} > 1$ are dominated by the noise in the forward dynamics and cannot be reconstructed.

\begin{figure}[t!]
    \centering
    \begin{tikzpicture}
        \node[anchor=north west,inner sep=0pt] at (0,0){
        \includegraphics[width=0.49\textwidth]{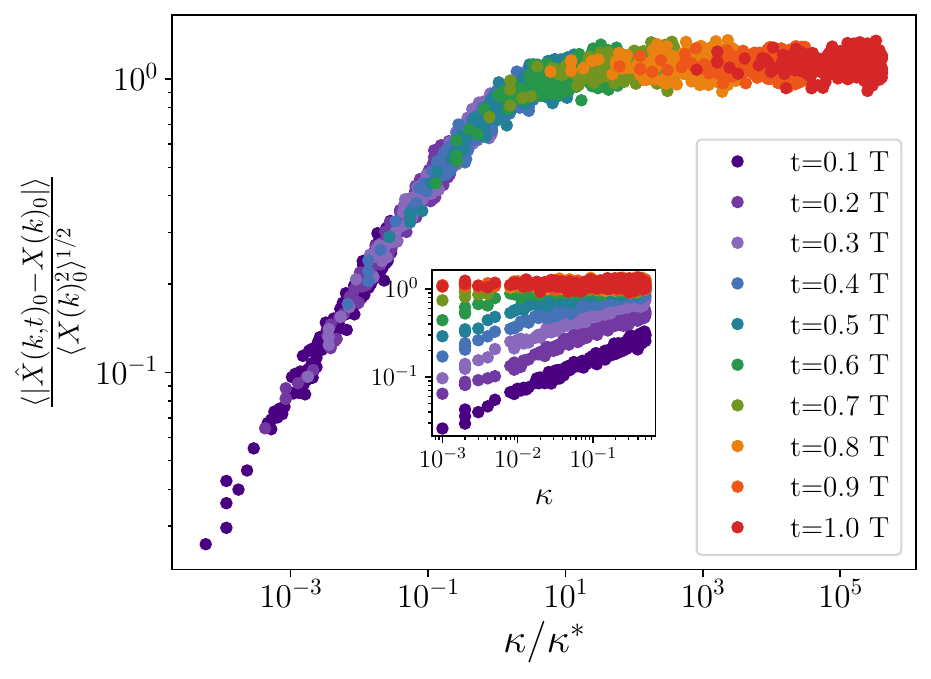}};
        \node at (0.2,0.5ex) {{(a)}};
        \node[anchor=north west,inner sep=0pt] at (7.5,0){
        \includegraphics[width=0.47\textwidth]{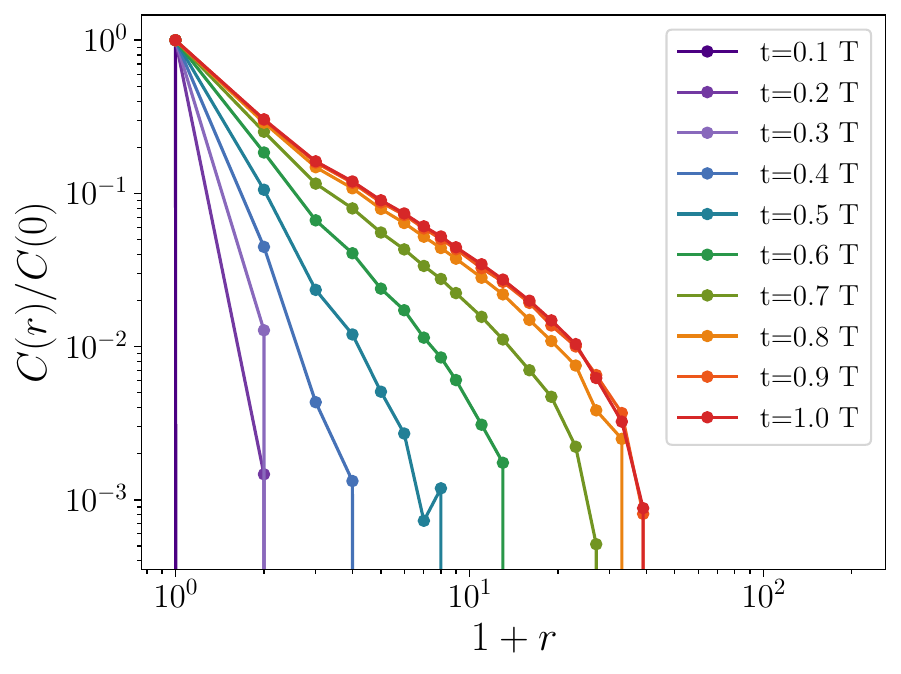}};
        \node at (7.7, 0.5ex) {{(b)}};
    \end{tikzpicture}
    \begin{tikzpicture}
        \node[anchor=north west,inner sep=0pt] at (0,0){
        \includegraphics[width=0.47\textwidth]{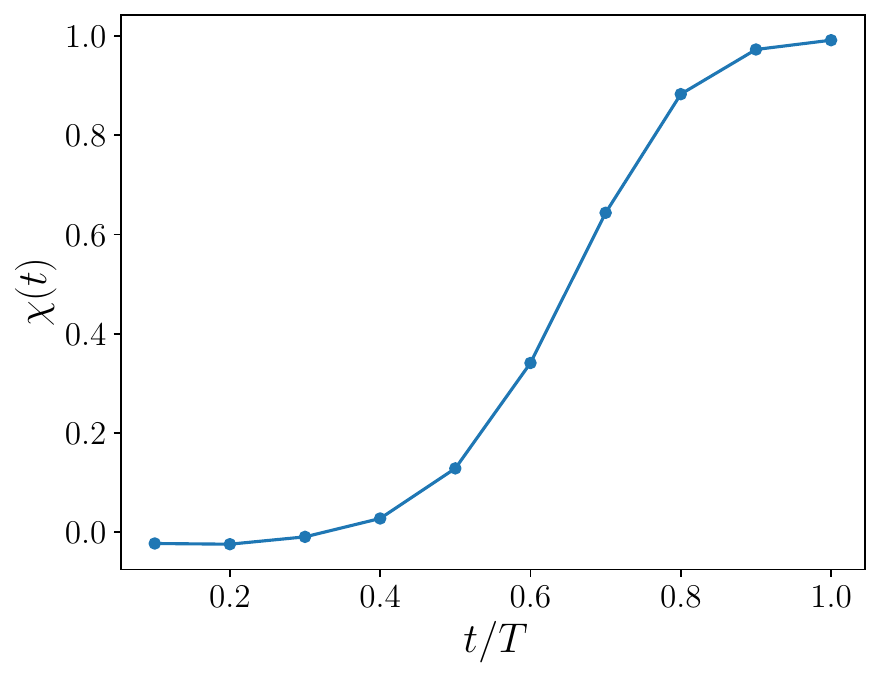}};
        \node at (0.2,0.5ex) {{(c)}};
    \end{tikzpicture}
    \caption{\textbf{Gaussian random field model.} \textit{(a)} Relative modal errors as a function of wave-vector magnitude $|k|$. For $|k|>\kappa^*$, errors remain large, indicating unsuccessful retrieval of the Fourier coefficients, while for $|k|<\kappa^*$, the errors decrease, signifying successful recovery. \textit{(b)} Spatial correlation function $\mathcal{C}(r,t)$, showing a power law decay at short distances and a cutoff at long distances. The correlation length increases with inversion time $t$. \textit{(c)} The susceptibility $\chi(t)$ increases monotonically and reaches its maximum at the inversion time $t=T$.}
    \label{fig:probing-grf}
\end{figure}

\subsection{Correlation analysis} We seek to compute the correlation of the changes after reverting the process at time $t$. Let $\x(u,t)$ denote the field obtained after reverting the diffusion process at time $t$, at position $u$. In particular, $\x(\cdot,0)$ denotes the starting random field. Define the difference field $\vz(u,t)=\x(u,t)-\x(u,0)$. Since the two fields are Gaussian, also $\vz(\cdot,t)$ is Gaussian. 

We are interested in the following spatial correlation function:
\begin{equation}
    \mathcal{C}(r,t) = \mathbb{E}[\vz(u,t)^2 \vz(0,t)^2],
\end{equation}
where $r=\|u\|$.
Using Wick's theorem, we have
\begin{equation}
    \mathcal{C}(r,t) = \mathbb{E}[\vz(u,t)\vz(u,t)] \, \mathbb{E}[\vz(0,t)\vz(0,t)] + 2 \mathbb{E}[\vz(u,t)\vz(0,t)]^2.
\end{equation}
The first term is a constant independent of $r$, while the second term captures the spatial dependence.

To compute $\mathbb{E}[\vz(u,t)\vz(0,t)]$, we express $\vz(u,t)$ in terms of its Fourier coefficients $\textbf{Z}(k, t)$. For modes with $\kappa < \kappa^*(t)$, we can assume $\textbf{Z}(k, t) \approx 0$. For modes with $\kappa > \kappa^*(t)$, $\widehat{\X}(k,t)_{0}$ is approximately independent of $\X(k)_0$. Thus, $\textbf{Z}(k, t)$ for $\kappa > \kappa^*(t)$  is a Gaussian random variable with zero mean and variance $2\sigma_k^2$.

Thus, the covariance of $\vz$ is
\begin{equation}
    \mathbb{E}[\vz(u,t)\vz(0,t)] \simeq \int_{\|k\|>\kappa^*(t)} e^{ik^\top u}\,2\sigma_k^2 d^dk.
\end{equation}
Substituting $\sigma_k^2 \asymp \|k\|^{-a}$, we have:
\begin{equation}
    \mathbb{E}[\vz(u,t)\vz(0,t)] \simeq \int_{\|k\|>\kappa^*(t)} e^{ik^\top u}\,2\|k\|^{-a} d^dk.
\end{equation}

To evaluate the integral, we consider the asymptotic behavior for different regimes of $r$. At short distances $r \ll 1/\kappa^*$, the integral over $k$ is dominated by large $\kappa$ and behaves as $\mathbb{E}[\vz(u,t)\vz(0,t)] \simeq C_1 \, r^{a-d}$, where $C_1$ is a constant. At long distances $r \gg 1/\kappa^*(t)$, the lower limit $\kappa^*(t)$ introduces an effective cutoff and the covariance decays faster than any power law.

Therefore, the correlation function $\mathcal{C}(r,t)$ exhibits algebraic decay with exponent $2(a-d)$ for $r \ll 1/\kappa^*$ and faster than any power law for $r \gg 1/\kappa^*$.

\subsection{Discussion} For the Gaussian random field model, the correlation length $\xi \sim 1/\kappa^*(t)$ is a monotonically increasing function of the inversion time $t$, or noise-to-signal ratio (NSR). As a result, the susceptibility $\chi(t)$ -- calculated by integrating the correlation function over space -- also increases monotonically and reaches its maximum at the inversion time $t=T$, where the ${\rm NSR}=\infty$. This behavior contrasts sharply with the hierarchical data studied here, where a phase transition occurs at a finite time/${\rm NSR}$. As discussed in the main text, this divergence arises due to the geometry of correlations induced by the hierarchical tree structure, which is absent in the Gaussian random field model. 

\subsection{Numerical experiments} In \autoref{fig:probing-grf} (a), we plot the relative modal errors $\mathcal{E}_k=\sigma_k^{-1}|\widehat{\X}(k,t)_{0} - \X(k)_0|$. For $\|k\|>\kappa^*$, the errors remain $\mathcal{O}(1)$, indicating that the coefficients are not retrieved, as predicted by our analysis. Conversely, for $\|k\|>\kappa^*$, the errors decay, indicating successful recovery of the coefficients. In panel (b), we present the correlations $\mathcal{C}(r,t)$, which exhibit a power law decay followed by a cutoff. Notably, the correlation length increases monotonically with the inversion time $t$. Finally, in panel (c), we plot the susceptibility $\chi(t)$, which reaches its maximum at $t=T$.

\section{Language diffusion}
\label{app:probing-language}

\subsection{Setup}

Here, we briefly describe the particular realization of discrete diffusion used in the MDLM setting, which is detailed in~\citep{sahoo2024simple}.

MDLMs are a form of discrete diffusion model tailored for language generation. Unlike autoregressive (AR) models, MDLMs generate text by gradually unmasking tokens, allowing for non-sequential generation. This process is governed by a forward masking and reverse unmasking process, parameterized using a Rao-Blackwellized objective to improve performance.

\paragraph*{Forward Process:}
The forward process is defined by progressively noising a clean input sequence \( x \) using a categorical distribution:
\begin{align}
q(z_t | x) = \text{Cat}(z_t; \alpha_t x + (1 - \alpha_t) m),
\end{align}
where \( z_t \) is the latent variable at time \( t \), representing the noisy version of the input sequence, \( x \) is the original, clean sequence of tokens, \( \text{Cat}(\cdot; \cdot) \) is a categorical distribution over the possible states, \( \alpha_t \) is the noise schedule function, strictly decreasing from \( 1 \) to \( 0 \) as \( t \) increases, and \( m \) is a one-hot vector representing the special masked token. At each time step, a fraction of the data transitions into the masked state.

\paragraph*{Reverse Process and Rao-Blackwellization: }
The reverse diffusion process reconstructs the original data from noisy observations. It is parameterized using a neural network approximation \( x_\theta(z_t, t) \), which predicts clean tokens from noisy inputs:
\begin{align}
p_\theta(z_s | z_t) = 
\begin{cases}
\text{Cat}(z_s; z_t), & \text{if } z_t \neq m, \\
\text{Cat}\left(z_s; \frac{(1 - \alpha_s) m + (\alpha_s - \alpha_t) x_\theta(z_t, t)}{1 - \alpha_t}\right), & \text{if } z_t = m.
\end{cases}
\end{align}
where \( z_s \) is the latent variable at a prior time step \( s \) (with \( s < t \)), \( x_\theta(z_t, t) \) is a neural network approximation of \( x \) given the noisy observation \( z_t \) at time \( t \), and \( p_\theta(\cdot | \cdot) \) is the model distribution approximating the true reverse process.

The training objective is a \textit{negative evidence lower bound} (NELBO), expressed as:
\begin{align}
L_{\text{diffusion}} = \sum_{i=1}^{T} \mathbb{E}_q \left[\frac{\alpha_{t(i)} - \alpha_{s(i)}}{1 - \alpha_{t(i)}} \log \langle x_\theta(z_{t(i)}), x \rangle \right],
\end{align}
where \( T \) is the number of diffusion steps, \( \alpha_{t(i)} \), \( \alpha_{s(i)} \) is the noise schedules evaluated at time steps \( t(i) \) and \( s(i) \), respectively, \( \mathbb{E}_q \) is the expectation over the forward process defined by \( q \), and \( \langle x_\theta(z_{t(i)}), x \rangle \) is the dot product between the neural network output \( x_\theta(z_{t(i)}) \) and the original input \( x \).

\paragraph*{Continuous-Time Likelihood Bounds: }
To achieve a tighter approximation to the ELBO, the discrete objective is extended to continuous time as:
\begin{align}
L_{\infty \text{NELBO}} = \mathbb{E}_q \int_{0}^{1} \frac{\alpha'_t}{1 - \alpha_t} \log \langle x_\theta(z_t, t), x \rangle \, dt.
\end{align}
where \( \alpha'_t \) is the time derivative of the noise schedule \( \alpha_t \).
The integral evaluates the objective over continuous time, providing a tighter bound on the likelihood.
This formulation is invariant to the specific functional form of the noise schedule \( \alpha_t \), highlighting the robustness of the MDLM approach.
\paragraph*{Connection to Masked Language Models: }
MDLMs leverage a masked diffusion approach where the training objective is a weighted average of classical masked language modeling (MLM) losses:
\begin{align}
L_{\infty \text{NELBO}} = \mathbb{E}_q \int_{0}^{1} \frac{\alpha'_t}{1 - \alpha_t} \sum_{\ell} \log \langle x_\theta^\ell(z_t), x^\ell \rangle \, dt,
\end{align}
where \( x^\ell \): The \( \ell \)-th token in the original sequence, \( x_\theta^\ell(z_t) \): The neural network’s prediction for the \( \ell \)-th token given the noisy sequence \( z_t \).
The summation runs over all tokens in the sequence, effectively establishing a connection between MDLMs and BERT-style encoders while equipping them with generative capabilities.

We employ the MDLM proposed in~\citep{sahoo2024simple} to conduct the forward-backward experiments described in \autoref{sec:probing-experiments}, by first drawing random texts of a fixed token length from the WikiText2 database, masking a fixed fraction of the tokens $t$, and then performing the backward diffusion process by using the masked sequence as the initial point for the MDLM model.

\subsection{Examples of Text Samples for the Forward-Backward Experiments}

Below, we provide examples of texts generated by the forward-backward process using MDLM seeded from WikiText2 examples for different masking fractions. Selected samples were shown in the main text in~\autoref{fig:probing-text-corr} (a). We dub the text results after the forward-backward process as \textit{U-turn} samples. As can be seen by the color coding, correlated blocks of words change together along the denoising process, as described in~\autoref{sec:probing-blocks}, and the semantic meaning of the paragraphs themselves change along the phase transition. 
In blue we denote masked tokens that have changed their value after the backward process, while in green masked tokens that have returned to their initial value. Red indicates the changes in the final texts. 
It can be seen that for small masking fractions such as $t/T=0.1$, most of the tokens do not change after masking, while the amount of changed tokens far exceeds the unchanged ones near the phase transition at $t/T=0.5$, hinting at the long-range correlations emerging.  

\noindent \includegraphics[width=.9\textwidth]{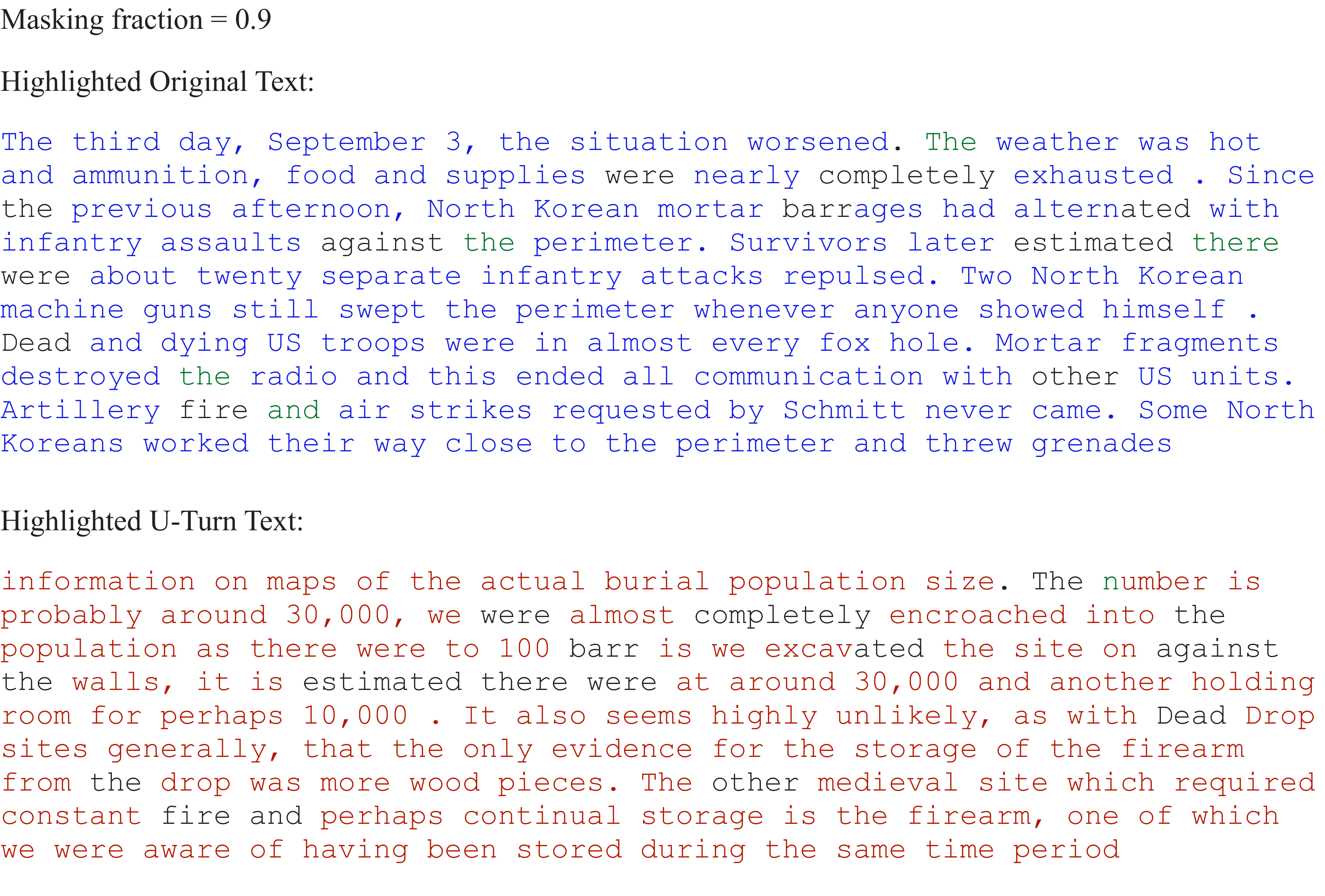} 
\includegraphics[width=.9\textwidth]{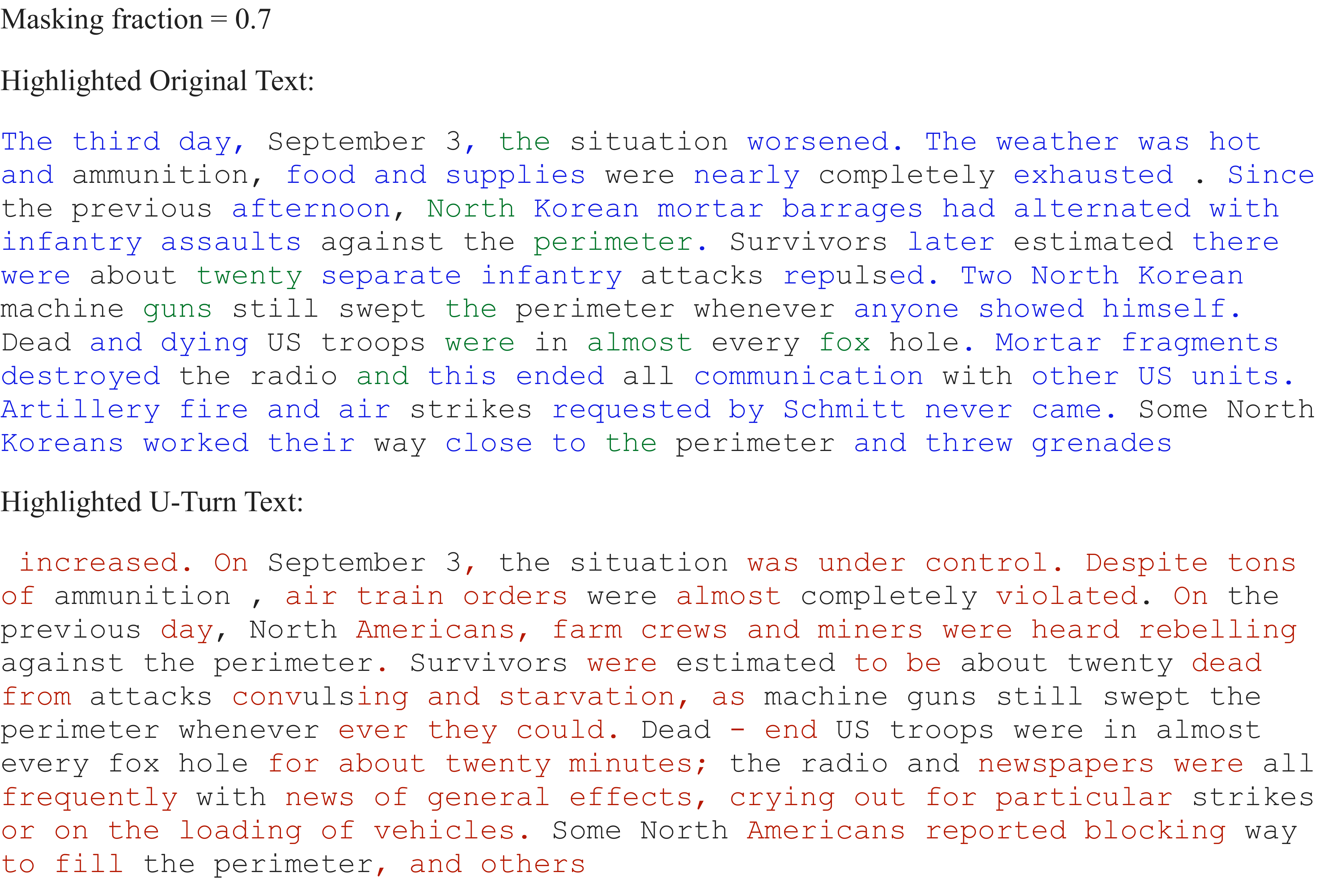}
\includegraphics[width=.9\textwidth]{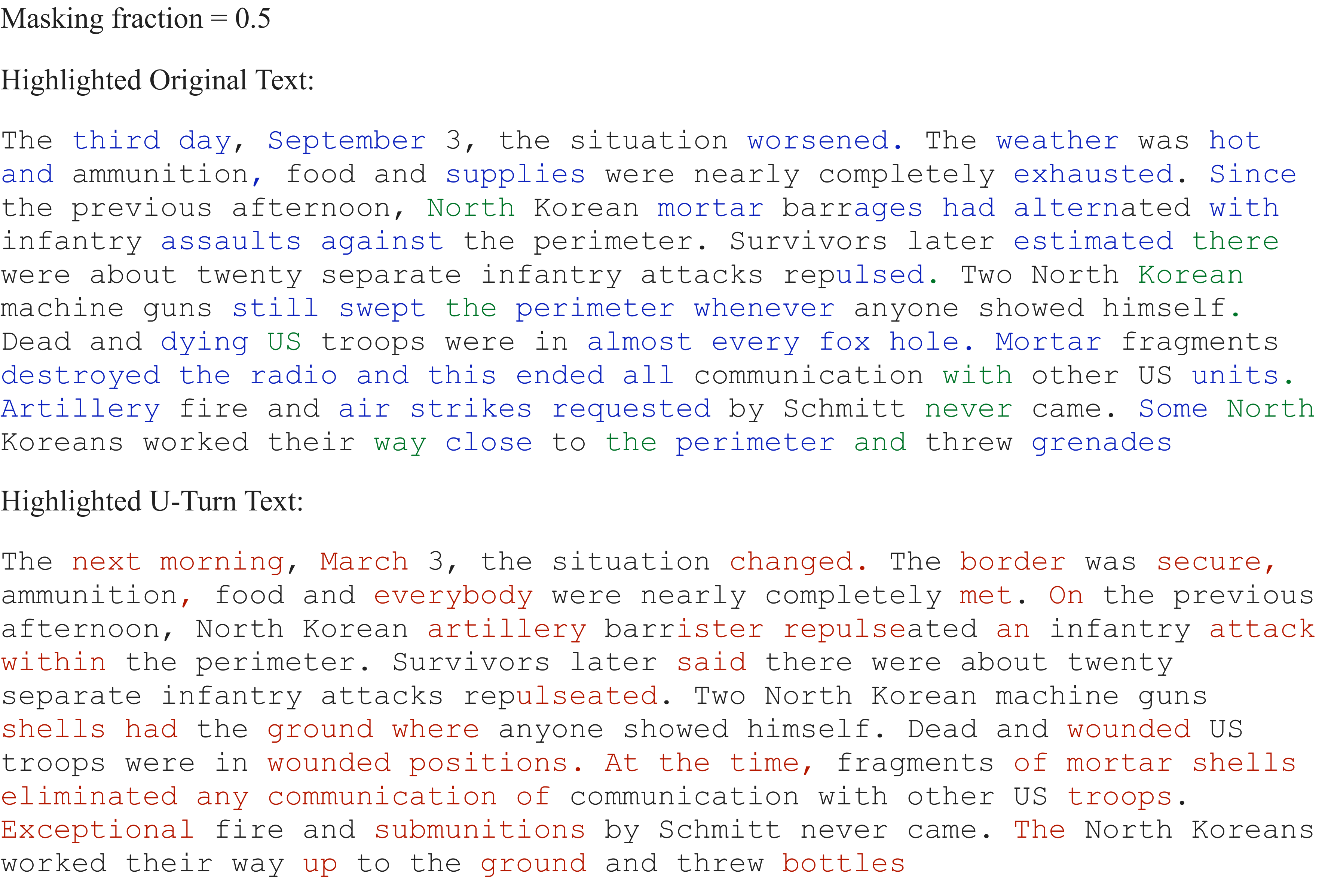}
\includegraphics[width=.9\textwidth]{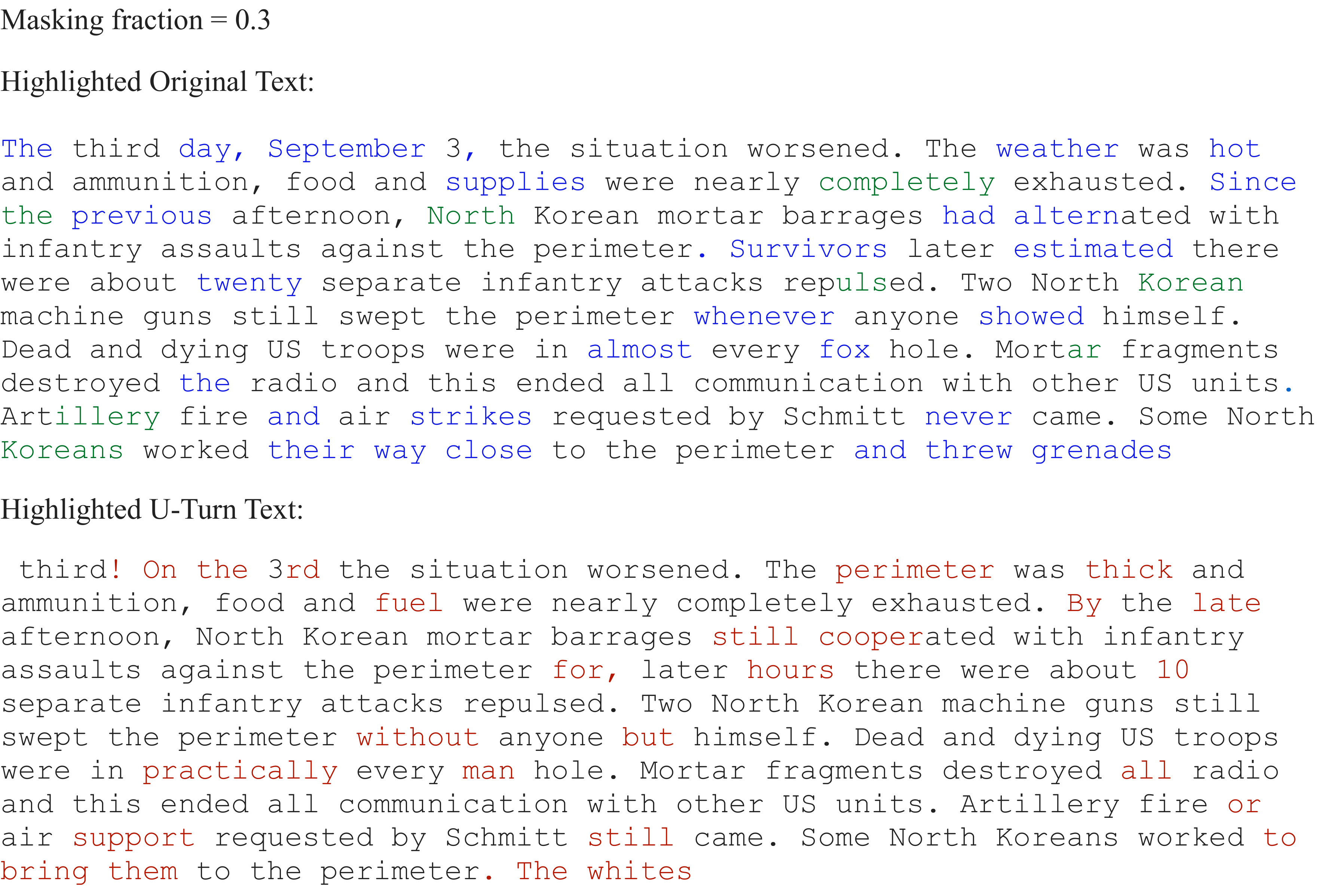}
\includegraphics[width=.9\textwidth]{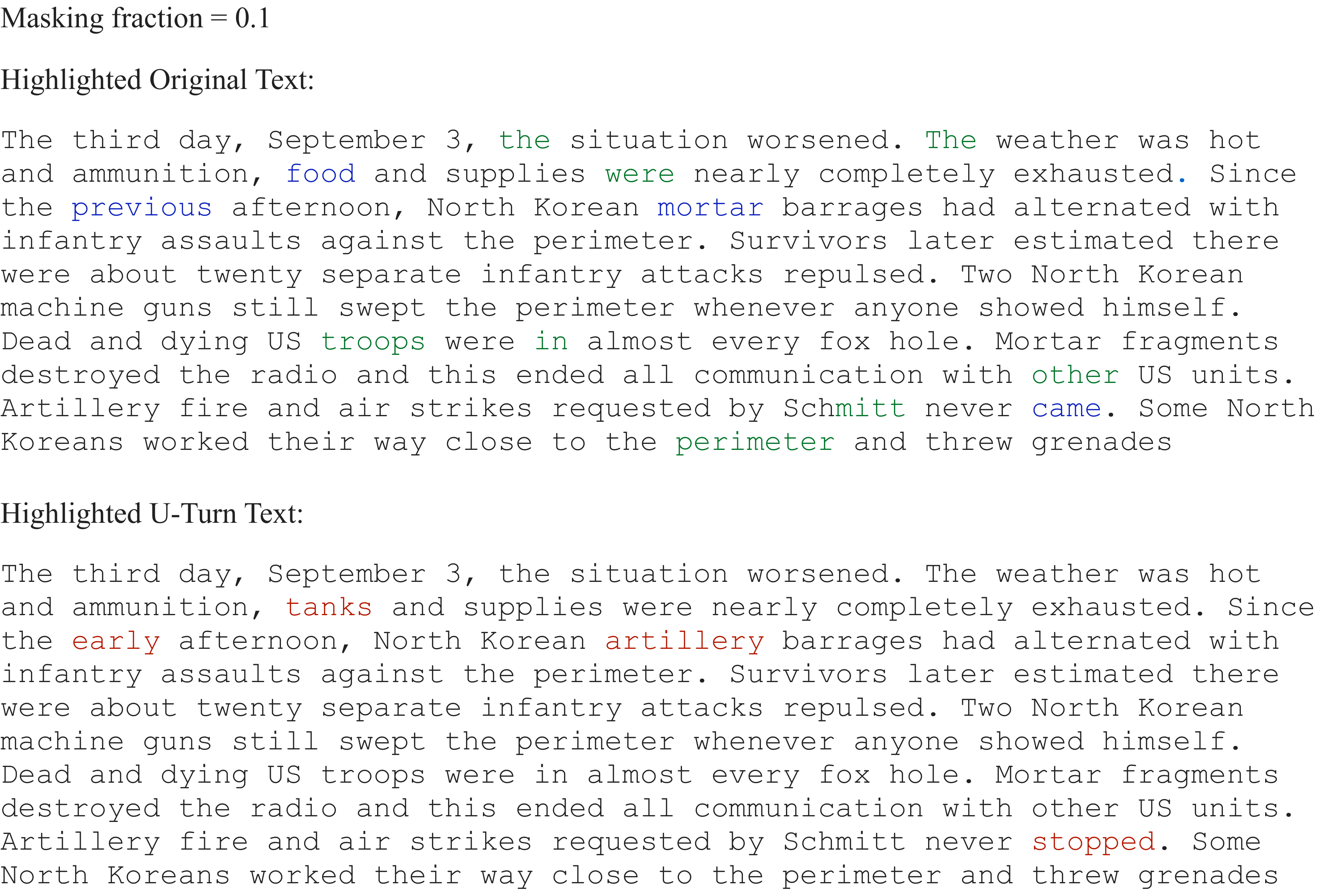}

\vspace{-0.2cm}
\section{Image diffusion}
\label{app:probing-vision}
\vspace{-0.2cm}

For image diffusion, we use the publicly available models from \textit{Improved Denoising Diffusion Probabilistic Models} \citep{nichol2021improved}, trained on the ImageNet dataset at resolution $256\times 256$. We use the class-unconditional model to ensure a class phase transition at an intermediate diffusion time.
To tokenize the images in a semantically meaningful manner, we use the last-layer embeddings from a CLIP ViT-B32 \citep{radford2021learning} encoder.
This procedure crops the images to the size $224\times 224$, which get tokenized in $7\times 7$ patches, each of dimension $32\times 32$. The embeddings at the last layer of the CLIP encoder have dimension $768$.

\begin{figure}
    \centering
    \begin{tikzpicture}
        \node[anchor=north west,inner sep=0pt] at (0,0){
        \includegraphics[width=0.3\textwidth]{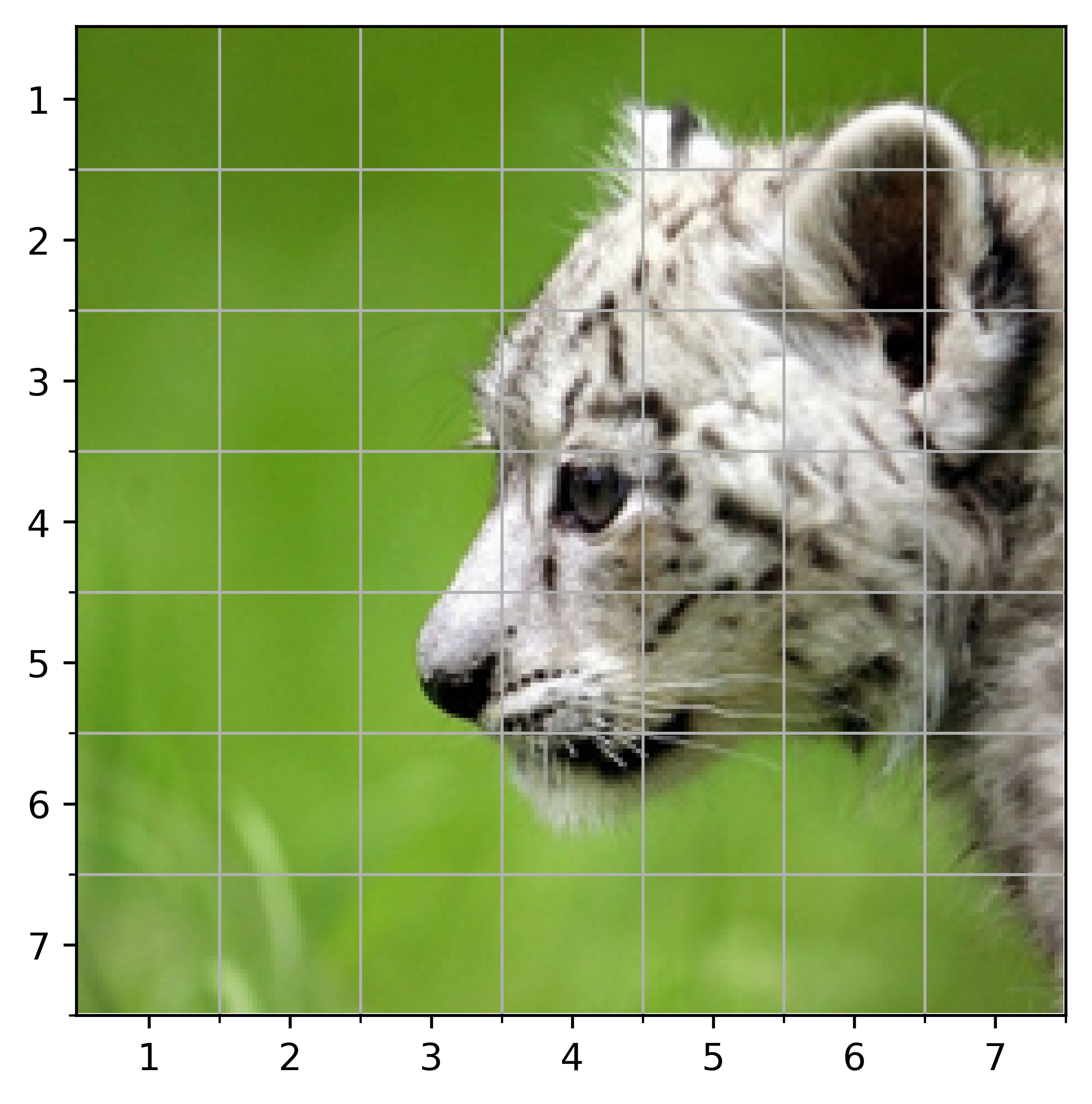}};
        \node[font=\large] at (15ex,1ex) {$t=0$};
    \end{tikzpicture}
    \hspace{.1cm}
    \begin{tikzpicture}
        \node[anchor=north west,inner sep=0pt] at (0,0){
        \includegraphics[width=0.3\textwidth]{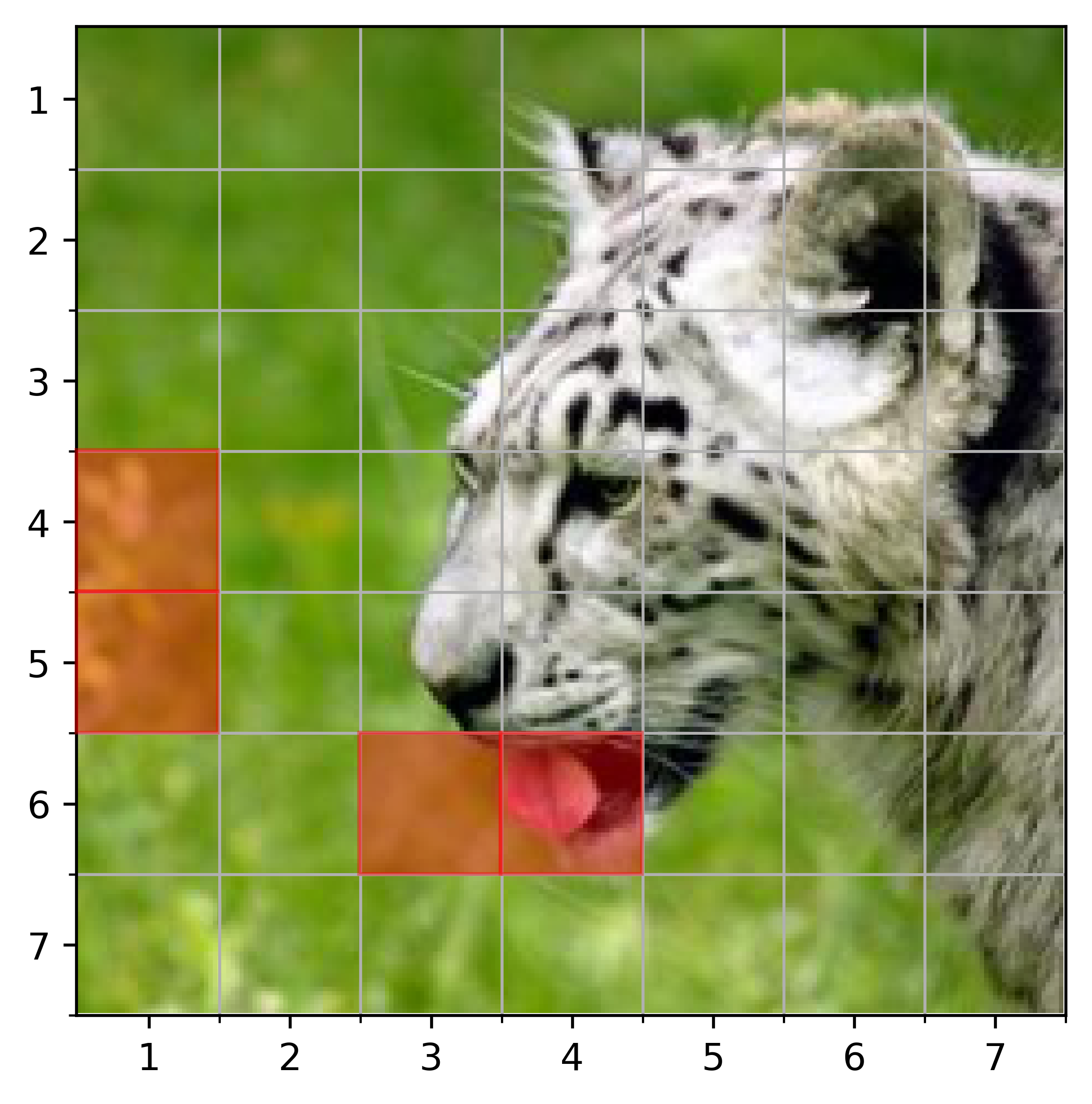}};
        \node[font=\large] at (15ex,1ex) { $t=0.6\ T$};
    \end{tikzpicture}
    \hspace{.1cm}
    \begin{tikzpicture}
        \node[anchor=north west,inner sep=0pt] at (0,0){
        \includegraphics[width=0.3\textwidth]{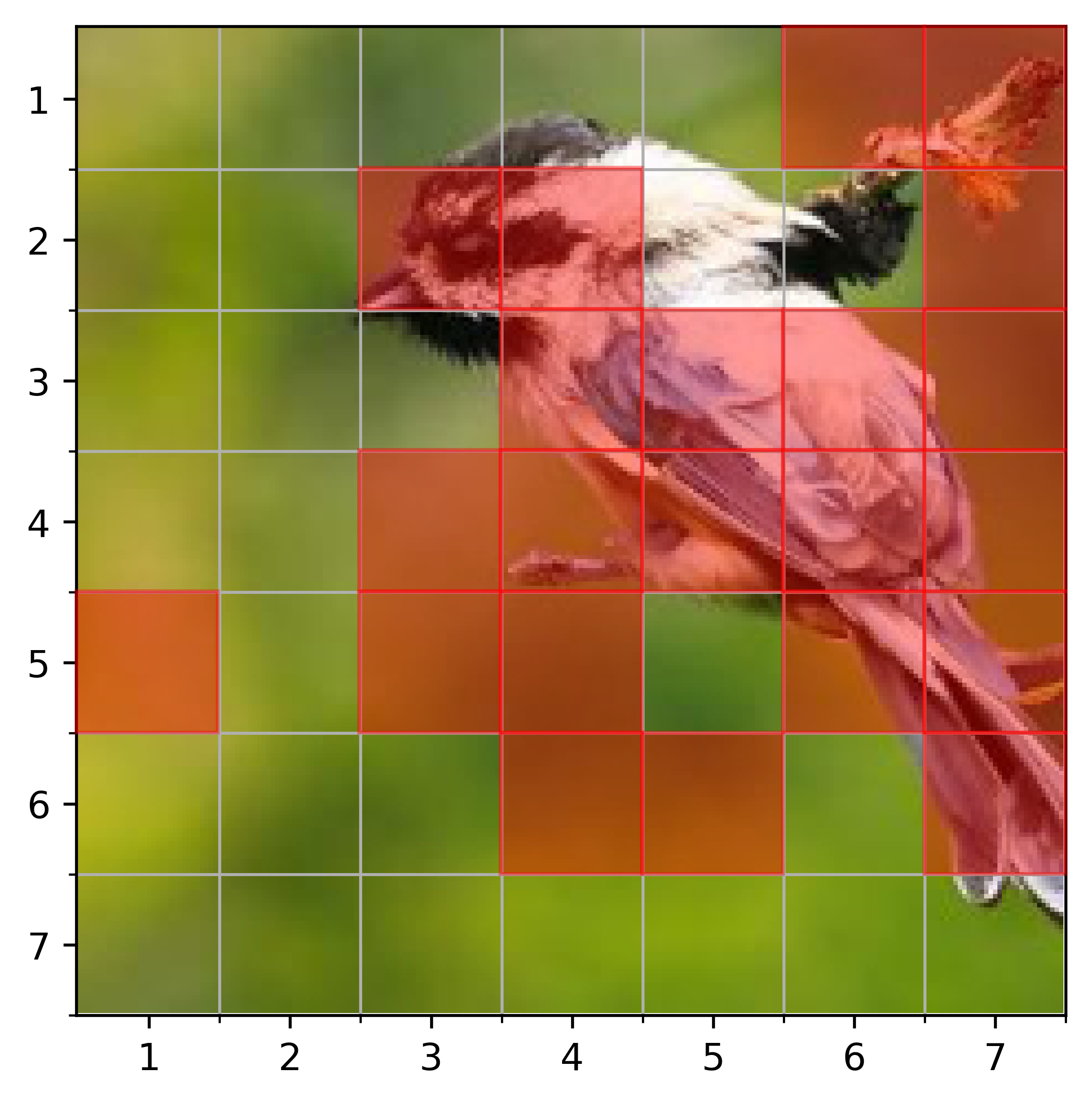}};
        \node[font=\large] at (15ex,1ex) {$t=0.7\ T$};
    \end{tikzpicture}
    \begin{tikzpicture}
        \node[anchor=north west,inner sep=0pt] at (0,0){
        \includegraphics[width=0.3\textwidth]{Figures/probing//n02009912_ILSVRC2012_val_00005314_t0.png}};
    \end{tikzpicture}
    \hspace{.1cm}
    \begin{tikzpicture}
        \node[anchor=north west,inner sep=0pt] at (0,0){
        \includegraphics[width=0.3\textwidth]{Figures/probing//n02009912_ILSVRC2012_val_00005314_t150.png}};
    \end{tikzpicture}
    \hspace{.1cm}
    \begin{tikzpicture}
        \node[anchor=north west,inner sep=0pt] at (0,0){
        \includegraphics[width=0.3\textwidth]{Figures/probing//n02009912_ILSVRC2012_val_00005314_t175_seed2.png}};
    \end{tikzpicture}
    \begin{tikzpicture}
        \node[anchor=north west,inner sep=0pt] at (0,0){
        \includegraphics[width=0.3\textwidth]{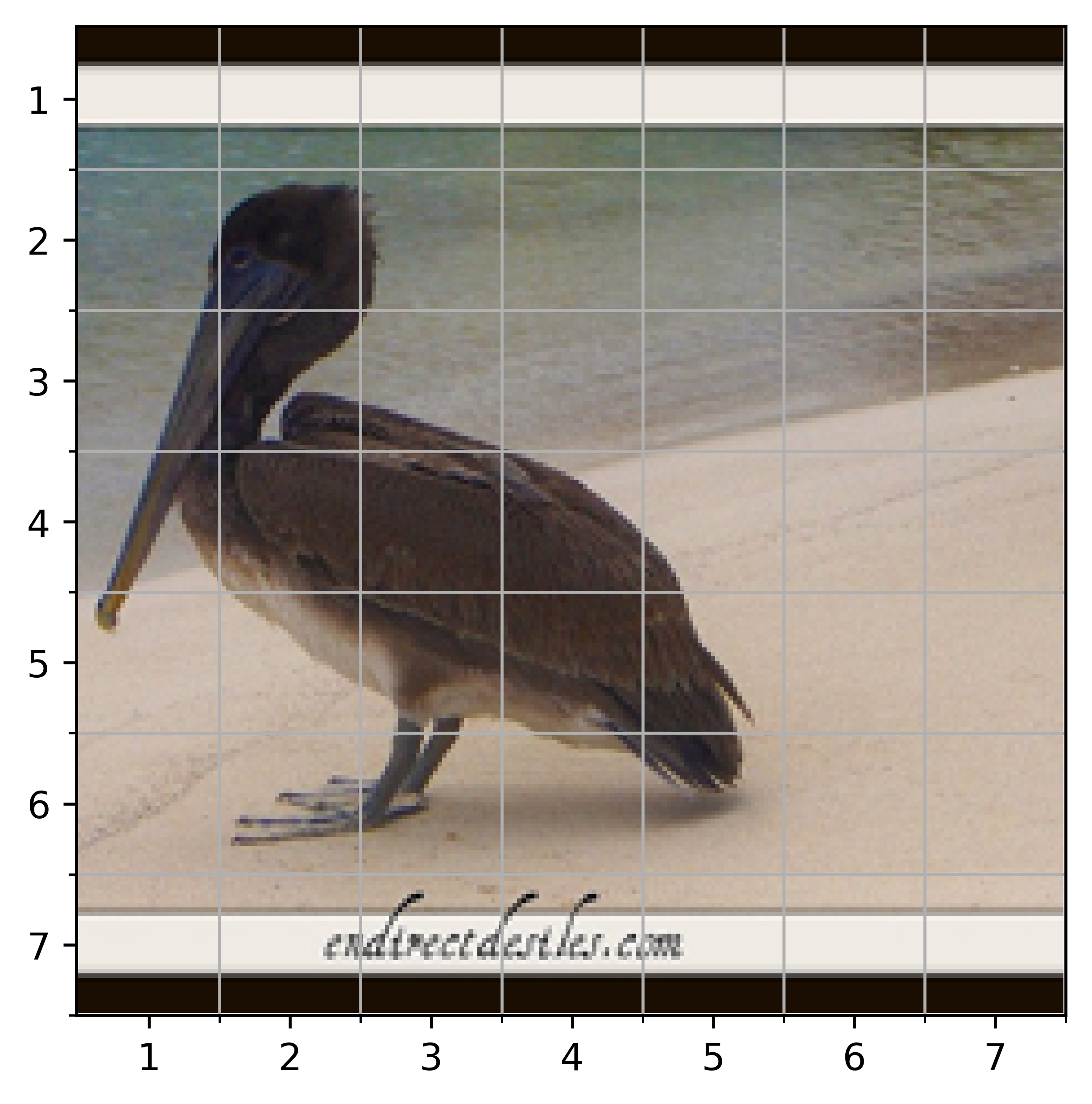}};
    \end{tikzpicture}
    \hspace{.1cm}
    \begin{tikzpicture}
        \node[anchor=north west,inner sep=0pt] at (0,0){
        \includegraphics[width=0.3\textwidth]{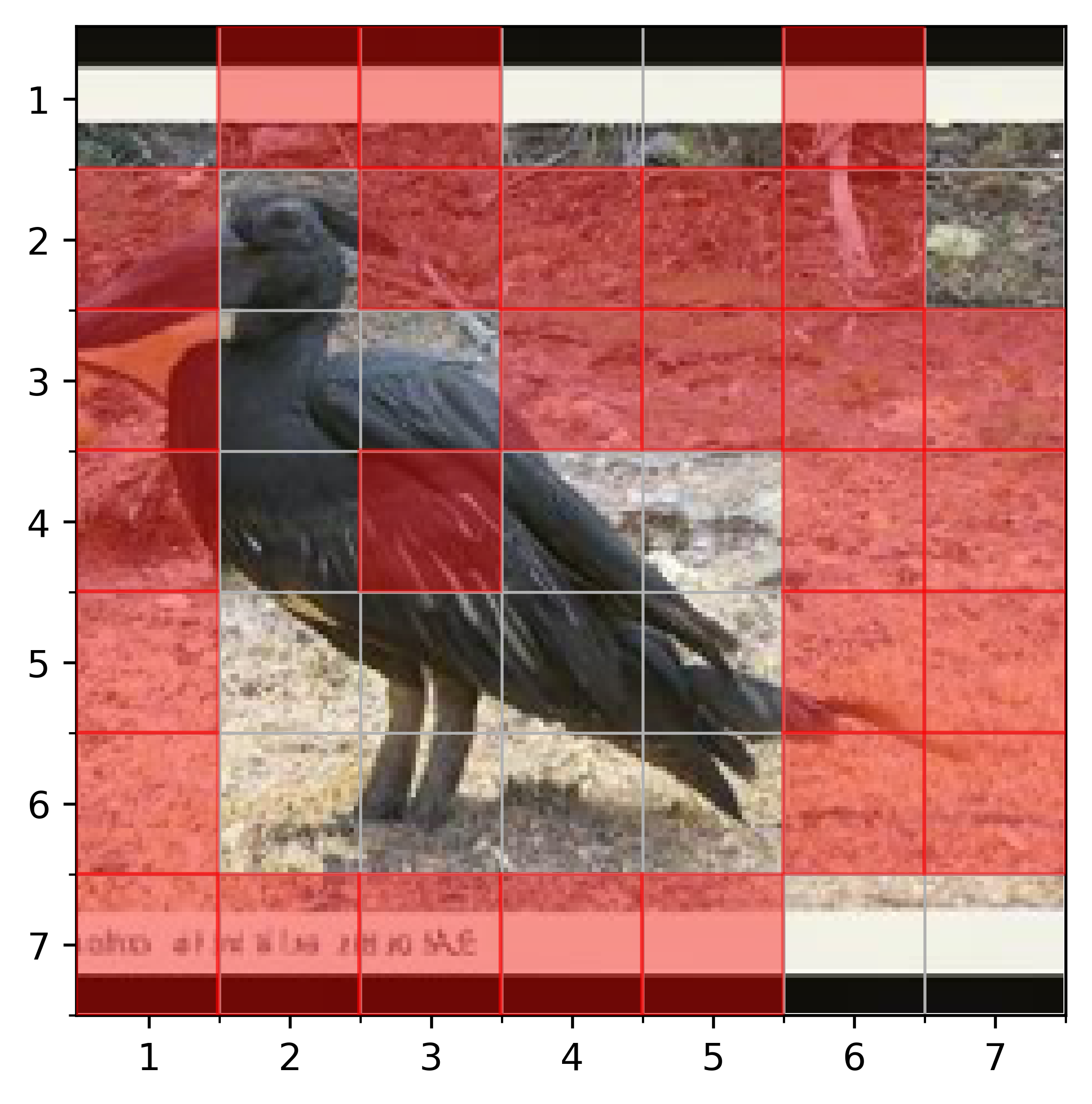}};
    \end{tikzpicture}
    \hspace{.1cm}
    \begin{tikzpicture}
        \node[anchor=north west,inner sep=0pt] at (0,0){
        \includegraphics[width=0.3\textwidth]{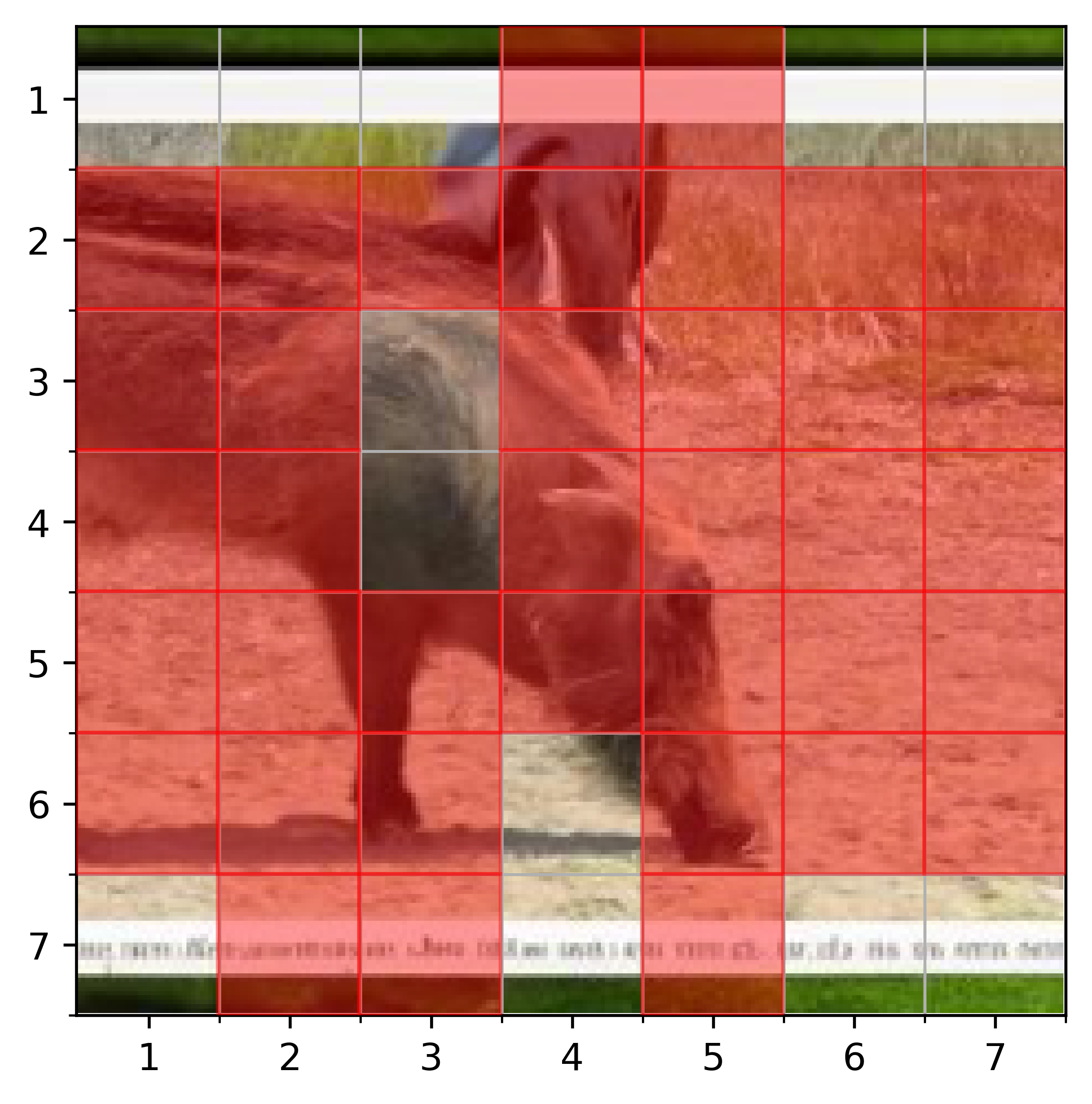}};
    \end{tikzpicture}
    \begin{tikzpicture}
        \node[anchor=north west,inner sep=0pt] at (0,0){
        \includegraphics[width=0.3\textwidth]{Figures/probing//n03642806_ILSVRC2012_val_00001713_t0.png}};
    \end{tikzpicture}
    \hspace{.1cm}
    \begin{tikzpicture}
        \node[anchor=north west,inner sep=0pt] at (0,0){
        \includegraphics[width=0.3\textwidth]{Figures/probing//n03642806_ILSVRC2012_val_00001713_t150_seed34.png}};
    \end{tikzpicture}
    \hspace{.1cm}
    \begin{tikzpicture}
        \node[anchor=north west,inner sep=0pt] at (0,0){
        \includegraphics[width=0.3\textwidth]{Figures/probing//n03642806_ILSVRC2012_val_00001713_t175_seed56.png}};
    \end{tikzpicture}
    \begin{tikzpicture}
        \node[anchor=north west,inner sep=0pt] at (0,0){
        \includegraphics[width=0.3\textwidth]{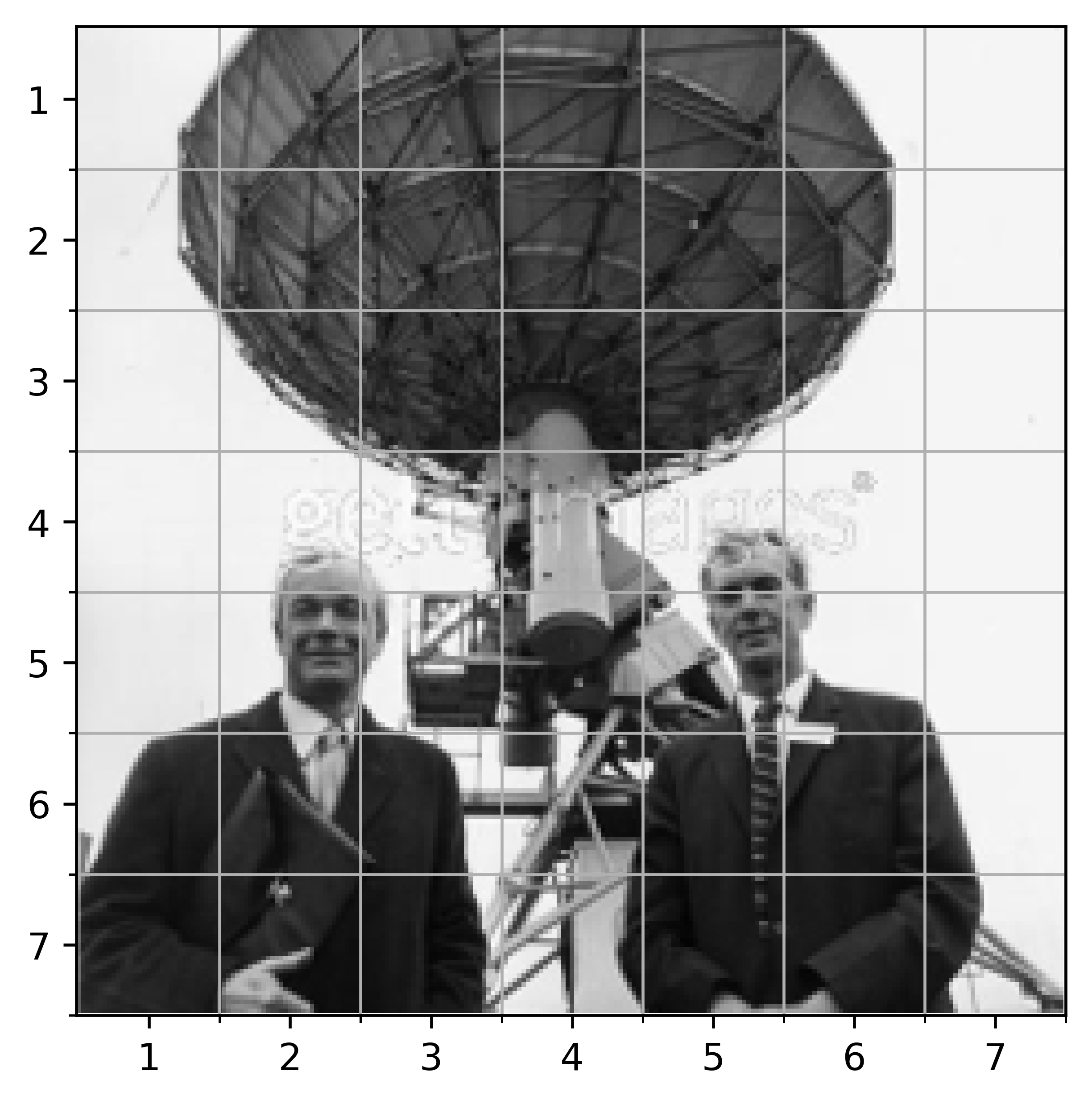}};
    \end{tikzpicture}
    \hspace{.1cm}
    \begin{tikzpicture}
        \node[anchor=north west,inner sep=0pt] at (0,0){
        \includegraphics[width=0.3\textwidth]{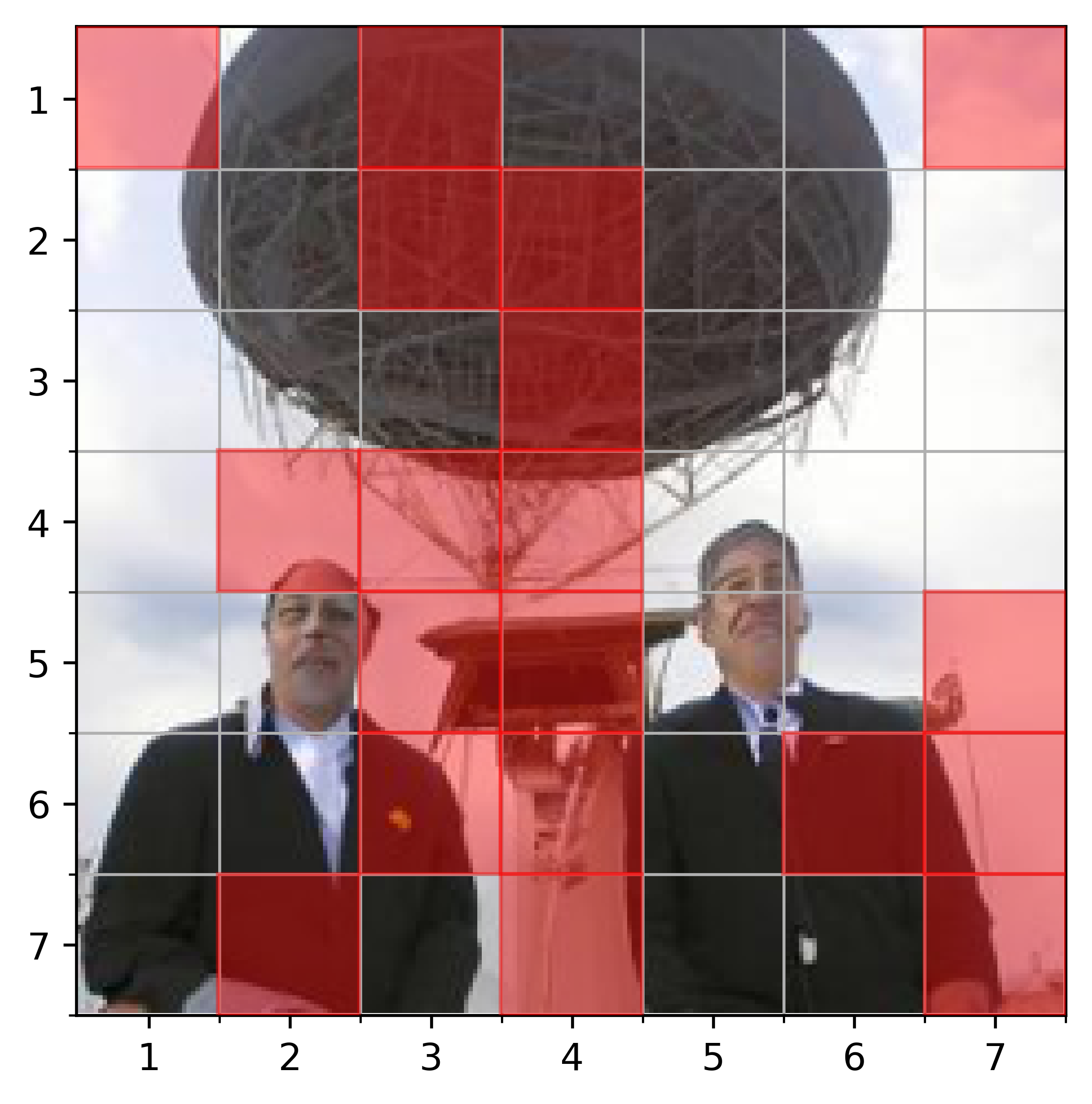}};
    \end{tikzpicture}
    \hspace{.1cm}
    \begin{tikzpicture}
        \node[anchor=north west,inner sep=0pt] at (0,0){
        \includegraphics[width=0.3\textwidth]{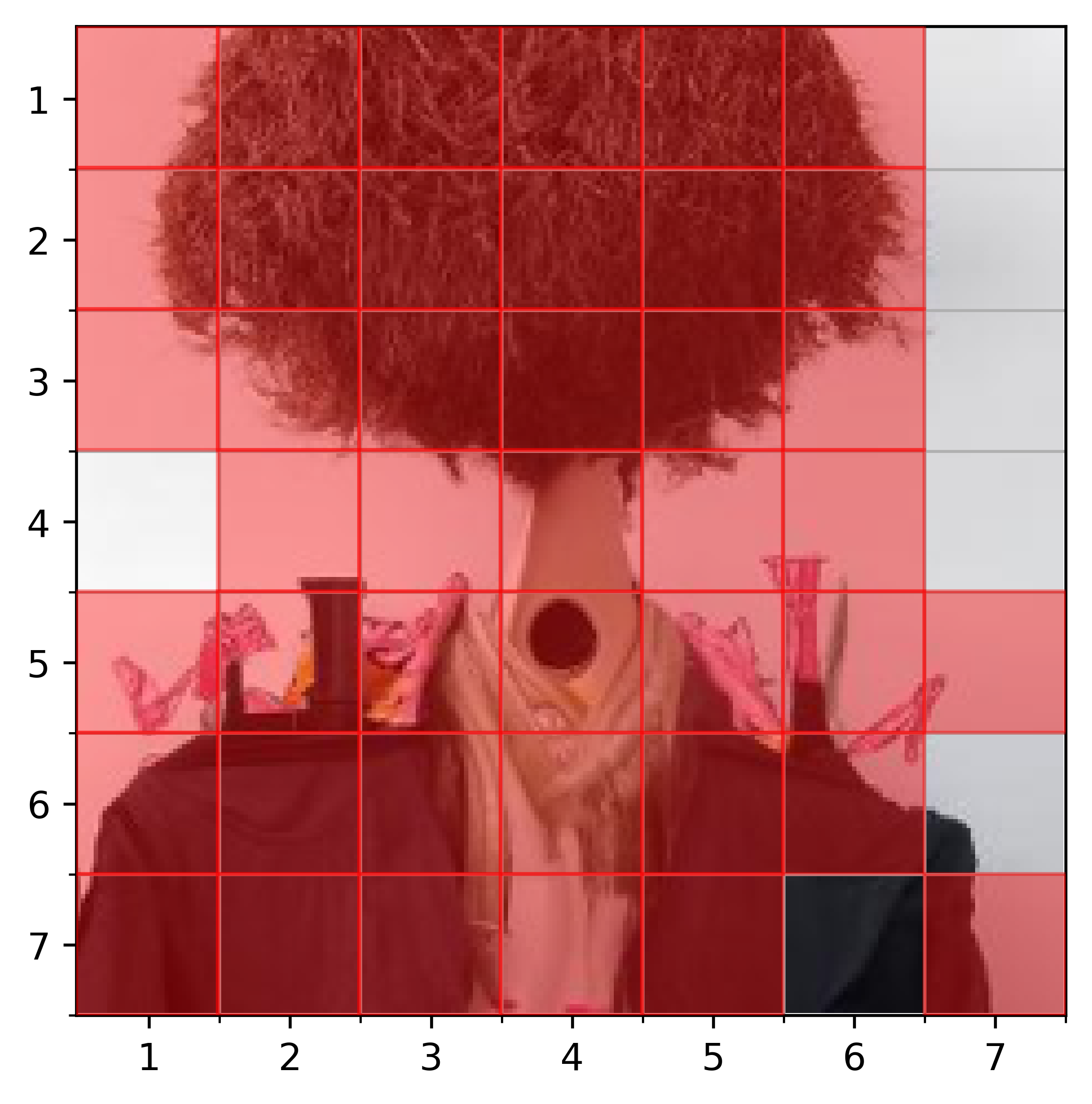}};
    \end{tikzpicture}
\vspace{-5pt}
\caption{\textbf{Examples of images generated at different inversion times $t$.} The grid indicates the tokens represented inside the CLIP vision encoder. For inversion time $t>0$, the red patches indicate the token embeddings that have a variation magnitude larger than a fixed threshold. These patches of variation appear in domains of growing size.}
\label{fig:probing-img_patch}
\end{figure}

\begin{figure}
    \centering
    \includegraphics[width=0.7\linewidth]{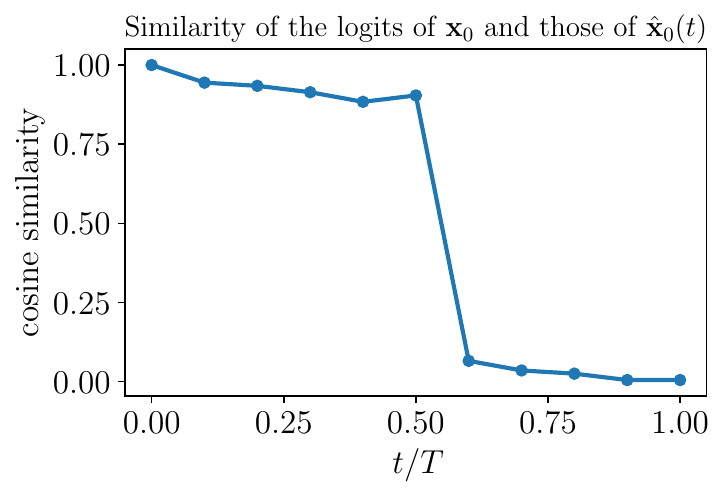}
    \caption{\textbf{Class transition in the forward-backward diffusion for ImageNet images.}
    Cosine similarity between the logits of a convolutional classifier computed on the starting images $\mathbf{x}_0$ and on the generated images $\hat{\mathbf{x}}_0(t)$ at different inversion times $t$. The logits are standardized on the statistics of the ImageNet validation set and the cosine similarities are averaged over 10k starting images. The convolutional classifier is a ConvNeXt Base architecture \citep{liu2022convnet} pre-trained on ImageNet-1k and achieving 96.9\% top-5 generalization accuracy.
    At short inversion times, the similarity is close to one, implying that $\mathbf{x}_0$ and $\hat{\mathbf{x}}_0(t)$ are images of the same class. At inversion time around $t\approx 0.6 T$, the similarity has a sharp drop, corresponding to the class transition. Correspondingly, the susceptibility measure in \autoref{fig:probing-clip}-\textit{(b)} has a peak.
    }
    \label{fig:probing-logits-class}
\end{figure}

\looseness=-1 In~\autoref{fig:probing-img_patch}, we provide some examples of images generated with the forward-backward protocol. In red, we highlight the patches whose CLIP embeddings show a statistically significant change with respect to the starting image ($t=0$).
In \autoref{fig:probing-logits-class}, we evaluate a convolutional classifier on the generated images and the starting ones to detect the inversion time corresponding to the class transition.

\chapter{Appendix: A Theory of Creativity and Compositionality}

\section{Token-latent tuple correlations}\label{app:corr_L}

In this section, we derive our estimate for the magnitude of the correlations between $x_1$ and tuples of latent, level-$(\ell\,{-}\,1)$ features $\h^{(\ell-1)}_{(i-1)\times s+1:i\times s}$, with $i\,{=}\,2,\dots,s$ and $\ell\,{=}\,1,\dots,L\,{-}\,1$ (level-$0$ latents $h^{(0)}$ correspond to visible tokens). These correlations are identical for all the tuples of latents corresponding to the same higher-level feature $h^{(\ell)}_i$, thus can be used to reconstruct level-$\ell$ latents. For instance, with $s\,{=}\,2$, so that $i\,{=}\,2$ (see \autoref{fig:scheme_rhm}), the correlations of $x_1$ with $(x_3, x_4)$ determine the value of $h_2^{(1)}$, while those with $(h^{(1)}_3, h^{(1)}_4)$ determine $h^{(2)}_2$. To simplify the notation, we will stick to the case $i\,{=}\,2$ for the remainder of the section. Then, the goal is to compute the statistics of
\begin{align}
    C^{(\ell+1)}(\mu,\bm{\nu}) := \prob{X_1=\mu, \mathbf{\h}_{s+1:2s}^{(\ell-1)}=\bm{\nu}}-\prob{X_1=\mu}\prob{ \mathbf{\h}_{s+1:2s}^{(\ell-1)}=\bm{\nu}},
\end{align}
over realizations of the RHM.

For each visible token $i\,{=}\,1,\dots,d$, single-token probabilities can be written as products of probabilities over the single production rules,
\begin{align}\label{eq:single-token-prob}
\prob{X_i\,{=}\,\mu} = \sum_{\mu_1,\dots,\mu_L=1}^v p^{(1)}_{i_1} (\mu|\mu_1)\dots p^{(L)}_{i_L} (\mu_{L-1}|\mu_L) p^{(L+1)}(\mu_L),
\end{align}
where
\begin{itemize}
\item[\textit{i)}] the indices $i_L,\dots,i_L$ are such that $i_L\dots i_1$ equals the $s$-ary representation of $i$, with $i_\ell\,{=}\,1,\dots,s$, and $1$'s added to ensure that the representation always consists of $L$ indices. In other words, the multi-index $i_L,\dots,i_L$ uniquely identifies the path linking the root of the tree to the $i$-th leaf.
\item[\textit{ii)}] $p^{(\ell)}_{i_\ell}(\mu_{\ell-1}|\mu_{\ell})$ denotes the probability of choosing, among the available production rules starting from $\mu_{\ell}$, one that has the symbol $\mu_{\ell-1}$ on the $i_\ell$-th position of the right-hand size.
\item[\textit{iii)}] $p^{(L)}(\mu_L)$ denotes the probability of selecting the symbol $\mu_L$ as the root ($1/v$ for our model).
\end{itemize}
These decompositions arise naturally due to the connection between probabilistic context-free grammars and Markov processes. Similar decompositions apply to the probabilities of hidden variables and tuples, and the joint token-latent tuple probability. For the latter, in particular, starting from the level-$(\ell\,{+}\,1)$ hidden symbol $h^{(\ell+1)}_1$, lowest common ancestor (LCA) of $X_1$ and the tuple $\mathbf{\h}_{s+1:2s}^{(\ell-1)}$, we have
\begin{align}\label{eq:token-tuple-prob}
\prob{X_1=\mu, \mathbf{\h}_{s+1:2s}^{(\ell-1)}=\bm{\nu}}& =
\sum_{\mu_1,\dots,\mu_{\ell-1}=1}^v
p^{(1)}_{1} (\mu|\mu_1) \dots p^{(\ell)}_{1} (\mu_{\ell-1}|\mu_{\ell}) \, \times\nonumber \\
& \sum_{\nu_{\ell-1},\mu_\ell}p^{(\ell)}(\bm{\nu}|\nu_{\ell}) p^{(\ell+1)}_{1,2} (\mu_{\ell},\nu_{\ell}|\mu_{\ell+1}) p_{1}^{(\ell+2)}(\mu_{\ell+1}).
\end{align}

For $\ell\,{=}\,1$, the probability above coincides with the joint probability of the visible token $X_1$ and the tuple of visible tokens $X_{s+1},\dots,X_{2s}$. The correlations,
\begin{align}
    C^{(2)}(\mu,\bm{\nu}) := \prob{X_1=\mu, \mathbf{X}_{s+1:2s}=\bm{\nu}}-\prob{X_1=\mu}\prob{ \mathbf{X}_{s+1:2s}=\bm{\nu}},
\end{align}
have been analyzed in~\citet{cagnetta2024towards}: the mean vanishes, while the variance, in the limit of $m,v\to+\infty$ with $f\,{=}\,m/v^{s-1}$ finite, follows
\begin{align}\label{eq:c2}
 \avg{\left(C^{(2)}(\mu,\bm{\nu})\right)^2} = \frac{1-f}{v^3m^{4}}.
\end{align}
For $\ell\,{=}\,2$, after applying~\autoref{eq:token-tuple-prob}, we get
\begin{align}
 C^{(3)}(\mu,\bm{\nu}) &= \sum_{\mu_1=1}^v p^{(1)}_1(\mu|\mu_1) \left(\prob{h^{(1)}_1=\mu_1,\mathbf{\h}_{s+1:2s}^{(\ell-1)}=\bm{\nu}} \right. \nonumber \\
 &\phantom{=} \left. -\prob{h^{(1)}_1=\mu_1}\prob{\mathbf{\h}_{s+1:2s}^{(\ell-1)}=\bm{\nu}}\right)\nonumber\\
 &=\sum_{\mu_1=1}^v p^{(1)}_1(\mu|\mu_1) C^{(2)}(\mu_1,\bm{\nu}),
\end{align}
where the last equality follows from noticing that the probability of level-$\ell$ hidden variables coincides with the probability of the leaves of a tree with $L\,{-}\,\ell$ levels. In general,
\begin{align}
 C^{(\ell+1)}(\mu,\bm{\nu}) =\sum_{\mu_1=1}^v p^{(1)}_1(\mu|\mu_1) C^{(\ell)}(\mu_1,\bm{\nu}),
\end{align}
thus
\begin{align}
\avg{\left(C^{(\ell+1)}(\mu,\bm{\nu})\right)^2} =&  \sum_{\mu_1,\nu_1} \avg{ p^{(1)}_{1} (\mu|\mu_1) p^{(1)}_{1} (\mu|\nu_1)}\avg{C^{(\ell)}(\mu_1,\bm{\nu})C^{(\ell)}(\nu_1,\bm{\nu})} \nonumber\\
 =& \sum_{\mu_1} \avg{\left(p^{(1)}_{1} (\mu|\mu_1)\right)^2}\avg{\left(C^{(\ell)}(\mu_1,\bm{\nu})\right)^2} + \nonumber\\
& \sum_{\mu_1,\nu_1\neq \mu_1} \avg{ p^{(1)}_{1} (\mu|\mu_1) p^{(1)}_{1} (\mu|\nu_1)}\avg{C^{(\ell)}(\mu_1,\bm{\nu})C^{(\ell)}(\nu_1,\bm{\nu})}.
\end{align}
Knowing that the production rules of an RHM realization are chosen uniformly at random compatibly with the unambiguity constraint~\cite{cagnetta2024towards},
\begin{align}
\avg{ \left(p^{(1)} (\mu|\mu_1)\right)^2} =\frac{v^{s-1}(v-1) + m(v^{s-1}-1)}{mv(v^s-1)},
\end{align}
and, for $\nu_1\neq \mu_1$,
\begin{align}
\avg{ p^{(1)} (\mu|\mu_1) p^{(1)} (\nu|\nu_1)} = \frac{v^{s-1}-1}{v(v^s-1)}.
\end{align}
In addition, since $\sum_{\mu} C^{(\ell)}(\mu,\bm{\nu})\,{=}\,0$, then
\begin{align}
\sum_{\nu_1\neq\mu_1} \avg{C^{(\ell)}(\mu_1,\bm{\nu})C^{(\ell)}(\nu_1,\bm{\nu})} = -\avg{\left(C^{(\ell)}(\mu_1,\bm{\nu})\right)^2}.
\end{align}
Hence, 
\begin{align}
\avg{\left(C^{(\ell+1)}(\mu,\bm{\nu})\right)^2} &= \frac{v^{s-1}(v-1)}{m(v^s-1)}\avg{\left(C^{(\ell)}(\mu_1,\bm{\nu})\right)^2} \nonumber \\ &\xrightarrow[]{v\gg 1} \frac{1}{m}\avg{\left(C^{(\ell)}(\mu_1,\bm{\nu})\right)^2}.
\end{align}
Starting with $C^{(2)}$ from~\autoref{eq:c2}, we get
\begin{align}
C^{(\ell)} = \sqrt{\avg{\left(C^{(\ell)}(\mu,\bm{\nu})\right)^2}} \simeq \sqrt{\frac{1-f}{v^3 m^{\ell+2}}},
\end{align}
where the rightmost equality is exact in the limit $v,m \to + \infty$.

\section{One-step gradient descent}\label{app:one-step}

We consider a simplified one-step gradient descent setting~\cite{damian22neural}, where a simple machine-learning model is trained to approximate the conditional probability of one input token $X_{s+1}$ following an $s$-tuple of tokens $\mathbf{X}\,{=}\,(X_{1},\dots,X_{s})$. The training set $\mathcal{X}_P$ consists of $P$ pairs $(\x,\nu)$, with $\nu$ denoting the feature in the token $X_{s+1}$. We assume that
\begin{itemize}
\item[\emph{i)}] the input tuple $\mathbf{X}$ is given as the one-hot encoding of the tuple index. Each of the $m v$ possible combinations of $s$ features is assigned an index $\bm{\mu}\,{=}\,1,\dots,mv$ and $\x$ is the $mv$-dimensional sequence $\x_{\bm{\mu}} = \delta_{\bm{\mu},\bm{\mu}(\x)}$;
\item[\emph{ii)}] the machine-learning model is initialized on the empirical marginal probability of the token $X_{s+1}$ over the training set, $\hat{\mathbb{P}}\left( X_{s+1}\,{=}\,\nu \right)\,{:=}\,P^{-1}\sum_{(\x,\lambda)\in \mathcal{X_P}}\delta_{\nu,\lambda}$. This assumption is equivalent to a preprocessing step on the labels~\cite{damian22neural} that removes the class imbalance of the training set.
\end{itemize}
Due to assumption \emph{i)}, the task can be solved with a perceptron model followed by a softmax nonlinearity,
\begin{align}\label{eq:onestep-perceptron}
f_{\nu}(\x;W) = \sum_{\bm{\mu}} W_{\nu,\bm{\mu}} \x_{\bm{\mu}};\quad p_\nu(\x;W) = e^{f_\nu(\x;W)}\left(\sum_\sigma e^{f_\sigma(\x;W)}\right)^{-1};
\end{align}
where $W\in \mathbb{R}^{v\times(vm)}$ is the weight matrix. In this setup, Assumption \emph{ii)} is realized by initializing the weights as $W_{\nu,\bm{\mu}}\,{=}\,\log{\hat{\mathbb{P}}\left[ X_{s+1}\,{=}\,\nu\right]}$ independently of $\bm{\mu}$.

The model $f_\nu$ of~\autoref{eq:onestep-perceptron} is trained via Gradient Descent on the empirical cross-entropy loss computed over a training set $\mathcal{X}_P$ consisting of $P$ pairs $(\x,\nu)$, with $\nu$ denoting the feature in the token $X_{s+1}$,
\begin{align}\label{eq:cross-ent-loss}
\mathcal{L} = \displaystyle\mathbb{E}_{(\x,\nu)\in \mathcal{X}_P} \left[ -\log{\left(\frac{e^{f_\nu(\x;W)}}{\sum_{\sigma=1}^{v} e^{f_\sigma(\x;W)}}\right)} \right],
\end{align}
where $\mathbb{E}_{(\x,\nu)\in \mathcal{X}_P}$ denotes the empirical average over the training set. Denoting the learning rate with $\eta$, the update of the weights reads
\begin{align}
\Delta W_{\nu,\bm{\mu}} &= -\eta\frac{\partial \mathcal{L}}{\partial f_{\nu}} \frac{\partial f_\nu}{\partial W_{\nu,\bm{\mu}}} = \eta \displaystyle\mathbb{E}_{(\x,\lambda)\in \mathcal{X}_P} \left[ \delta_{\lambda,\nu} \x_{\bm{\mu}} -\frac{e^{f_{\nu}}}{\sum_{\sigma=1}^v e^{f_\sigma}} \x_{\bm{\mu}} \right] \nonumber\\
&=\eta\displaystyle\mathbb{E}_{(\x,\lambda)\in \mathcal{X}_P} \left[ \delta_{\lambda,\nu} \delta_{\bm{\mu},\bm{\mu}(\x)} - \hat{\mathbb{P}}\left[ X_{s+1}\,{=}\,\nu\right]\delta_{\bm{\mu},\bm{\mu}(\x)} \right] \nonumber\\
&=\eta\left(\hat{\mathbb{P}}\left[ X_{s+1}=\nu;(X_1,\dots,X_s)=(\mu_1,\dots,\mu_s)\right] \right. \nonumber \\
&\phantom{-} \left. -\hat{\mathbb{P}}\left[ X_{s+1}=\nu\right]\hat{\mathbb{P}}\left[ (X_1,\dots,X_s)=(\mu_1,\dots,\mu_s)\right]\right),
\end{align}
where, in the second line, we used assumption \emph{i)} to replace $\x_{\bm{\mu}}$ with $\delta_{\bm{\mu},\bm{\mu}(\x)}$ and assumption \emph{ii)} to replace $e^{f_{\nu}}/(\sum_{\sigma=1}^v e^{f_\sigma})$ with $\hat{\mathbb{P}}\left[ X_{s+1}\,{=}\,\nu\right]$. The right-hand side of the last line equals the empirical token-tuple correlation $\hat{C}_P(\nu,\bm{\mu})$. Therefore, after one gradient step, the weights are given by
\begin{align}\label{eq:onestep-weight}
W_{\nu,\bm{\mu}} = \log{\hat{\mathbb{P}}\left[ X_{s+1}\,{=}\,\nu\right]} + \eta \hat{C}_P(\nu,\bm{\mu}).
\end{align}
The first term is independent of the input $\bm{\mu}$, whereas the second can be thought of as a noisy measurement of the true token-tuple correlation $C(\nu,\bm{\mu})$. The true correlation is equal for all $\bm{\mu}$'s generated by the same higher-level hidden symbol $h^{(1)}(\bm{\mu})$ and its size can be estimated as the standard deviation over realizations of the RHM (\autoref{eq:c2}),
\begin{align}
C^{(2)} \simeq \sqrt{\frac{1-f}{v^3 m^{4}}}.
\end{align}
The empirical measurement $\hat{C}_P$ includes a sampling noise contribution, having size $(v^2 m P)^{-1/2}$. If $P\,{\gg}\,P_2\,{=}\,v m^3/(1-f)$, then the $\hat{C}_P$ in the right-hand side of~\autoref{eq:onestep-weight} is approximately equal to the true token-tuple correlation, thus the weights can be used to build a representation of the hidden variables of the generative model.

\section{Experimental details}\label{app:exp-details}

\paragraph{Random Hierarchy Model} 

We train the U-Net-based Discrete Denoising Diffusion Probabilistic Model (D3PM), optimizing the diffusion loss derived from a variational bound on the negative log-likelihood \citep{sohl2015deep}. Following \citet{d3pm2021}, we use the neural network to predict the conditional expectation $\mathbb{E}[\x(0) | \x(t)]$, which parameterizes the reverse diffusion process.

The convolutional U-Net consists of $L$ resolution blocks in both the encoder and decoder, with a filter size of $s$, stride of $s$, and 8192 channels. Each block uses GeLU activation functions, and skip connections link encoder and decoder layers with the same resolution. The model also includes two embedding and unembedding layers, implemented as convolutions with filter size 1.

We initialize the network using the maximal-update ($\mu$P) parameterization \citep{yang2020feature}. This allows stable feature learning dynamics even in large models. The model is trained with SGD with a learning rate of 1, using a batch size of 32, and momentum parameter of 0.9. The diffusion process follows a linear schedule with 1,000 noise levels. To prevent overfitting, we apply early stopping based on the validation loss, halting training when it plateaus or begins to increase. 

\paragraph{Language diffusion model} Our experiments are based on the codebase of MD4 \cite{shi2024simplified}: \href{https://github.com/google-deepmind/md4}{https://github.com/google-deepmind/md4}. MD4 is a masked diffusion model. At each time step $t$, non-masked tokens either remain unchanged or transition to $[{\mathrm{MASK}}]$ with probability $\beta_t$.
Using a one-hot-encoding representation of the $|\mathcal{V}| + 1$ states, the forward transition matrix is given by:
\begin{equation}
    Q_t = (1-\beta_t) \mathbf{I} + \beta_t \mathbf{1} \mathbf{e}_M^{\top}.
\end{equation}
with $\mathbf{I}$ the identity matrix, $\mathbf{1}$ a vector of ones and $\mathbf{e}_M$ the one-hot-encoding vector corresponding to the $[{\mathrm{MASK}}]$ symbol. At the final time $T$, all tokens are masked, i.e., $x_i(T) = [{\mathrm{MASK}}]$ for every $i\in[{\mathrm{dim}}(\x)]$. We train MD4 with batch size 64 and context size 1024 on 4 H100s for a single epoch. All other hyperparameters are kept unchanged.

\paragraph{Vision diffusion model} Our experiments are based on the codebase of Improved DDPMs \cite{nichol2021improved}: \href{https://github.com/openai/improved-diffusion}{https://github.com/openai/improved-diffusion}. In particular, we train a DDPM with 128 channels, 3 resolution blocks, 4000 diffusion steps, cosine noise schedule, learning rate $10^{-4}$ and batch size 128 for 10 epochs using a \textit{hybrid objective} \cite{nichol2021improved}.

\section{Additional results}\label{app:additional-results}

\subsection{Emergence of hierarchical representations in the U-Net}

\begin{figure*}
    \centering
    \includegraphics[width=1\linewidth]{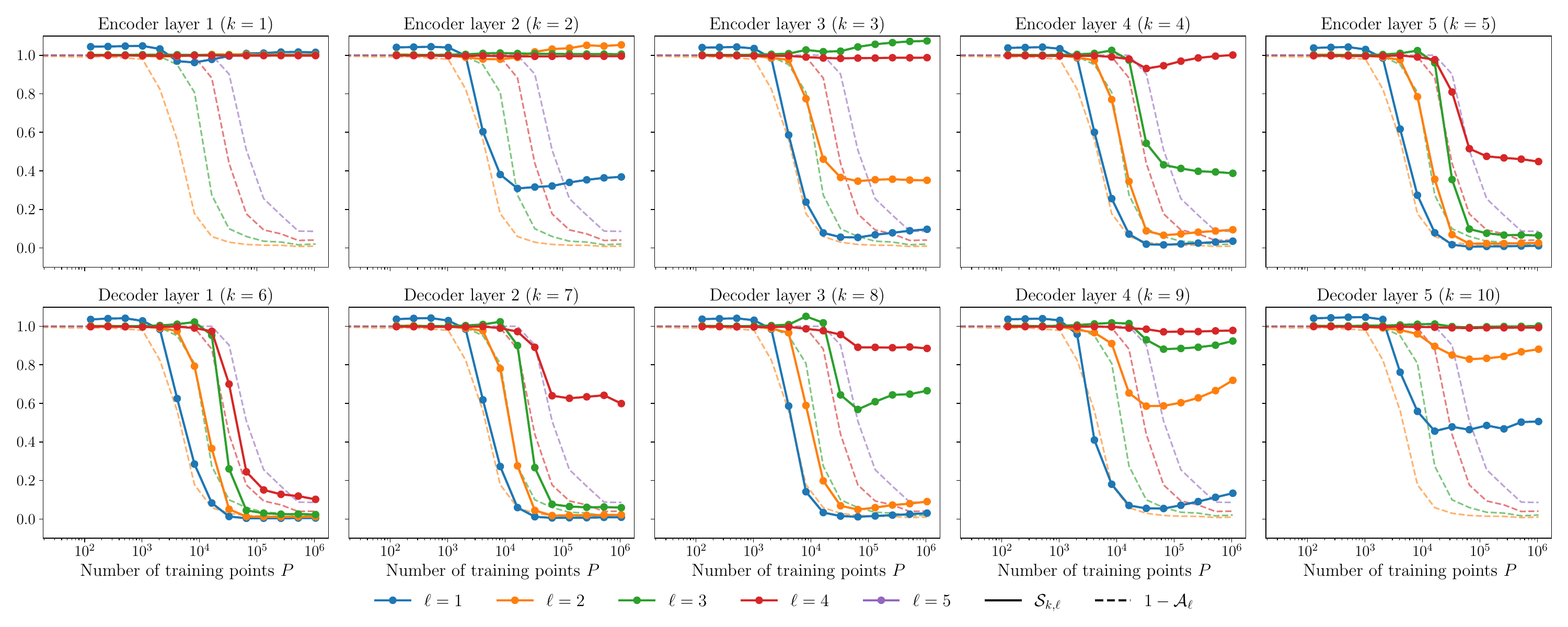}
    \caption{\textbf{Relative sensitivity of the hidden representations of the U-Net, defined in \autoref{eq:sensitivity}, with respect to the number of training points $P$.} Different colors correspond to different levels $\ell$ of synonymic exchange, while different panels correspond to the pre-activations of different U-Net blocks. Encoder layer $1$ is the closest to the input, while decoder layer $5$ is the closest to the output.  As the number of training points increases, deeper layers of the encoder become less sensitive to deeper synonymic transformations. This implies that deeper encoder layers learn to represent deeper latent variables of the RHM. The decoder layers, instead, progressively regain the sensitivity to the synonyms layer-by-layer as they expand latent variables into their lower-level representations. For each level $\ell$, the dashed line represents the fraction of generated samples that do not satisfy the rules at that level, i.e., $1-\mathcal{A}_\ell.$ The U-Net learns to satisfy rules at level $\ell$ when it becomes insensitive to the synonyms of the variables at level $\ell-1$.}
    \label{fig:sensitivity}
\end{figure*}

In \autoref{fig:sensitivity}, we test the hypothesis that the U-Net learns to represent together inputs that differ by low-level synonyms, i.e., the choice of low-level production rules. To do so, we introduce a transformation operator $\mathcal{R}_{\ell}\,\x$, which modifies a given data sample $\x$ by resetting all choices of the production rules emanating from level $\ell$. This operation is equivalent to substituting all tuples at depth $\ell-1$ with a synonym. We then define the relative sensitivity $\mathcal{S}_{k,\ell}$ of the pre-activations $a_k$ at layer $k$ to the transformation $\mathcal{R}_{\ell}$:
\begin{equation} \label{eq:sensitivity}
    \mathcal{S}_{k,\ell} = \frac{\mathbb{E}_{\x}[\|a_k(\x) - a_k(\mathcal{R}_{\ell}\,\x)\|^2]}{\mathbb{E}_{\x,\bm{y}}[\|a_k(\x) - a_k(\bm{y})\|^2]}.
\end{equation}
Here, the numerator measures how much the activations change when synonym substitutions are applied at depth $\ell$, while the denominator normalizes by the overall variability of activations across different data points. A low value of $\mathcal{S}_{k,\ell}$ indicates that the network is invariant to synonym substitutions at depth $\ell$, implying that it has learned the corresponding compositional rule.

\looseness=-1 \autoref{fig:sensitivity} shows the relative sensitivity of each layer as a function of the number of training points $P$. As $P$ increases, the sensitivities $\mathcal{S}_{k,\ell}$ decrease sequentially across levels, following the same staged learning process observed in \autoref{fig:main_L5}. Deep encoder layers become invariant to synonym substitutions at lower levels, confirming that the network is learning to encode the hierarchical structure of the grammar. In contrast, decoder layers gradually regain sensitivity to specific low-level symbols as the output is approached. This behavior aligns with their role in reconstructing low-level details from high-level representations. Crucially, the network begins to satisfy rules at level $\ell$ precisely when it becomes insensitive to synonymic variations at level $\ell-1$. This suggests that the U-Net learns to collapse lower-level synonyms into shared latent representations and to compose these latents according to the production rules at level $\ell$.

\subsection{Sample complexity of deep clustering algorithm}

\begin{figure}
    \centering
    \includegraphics[width=0.5\linewidth]{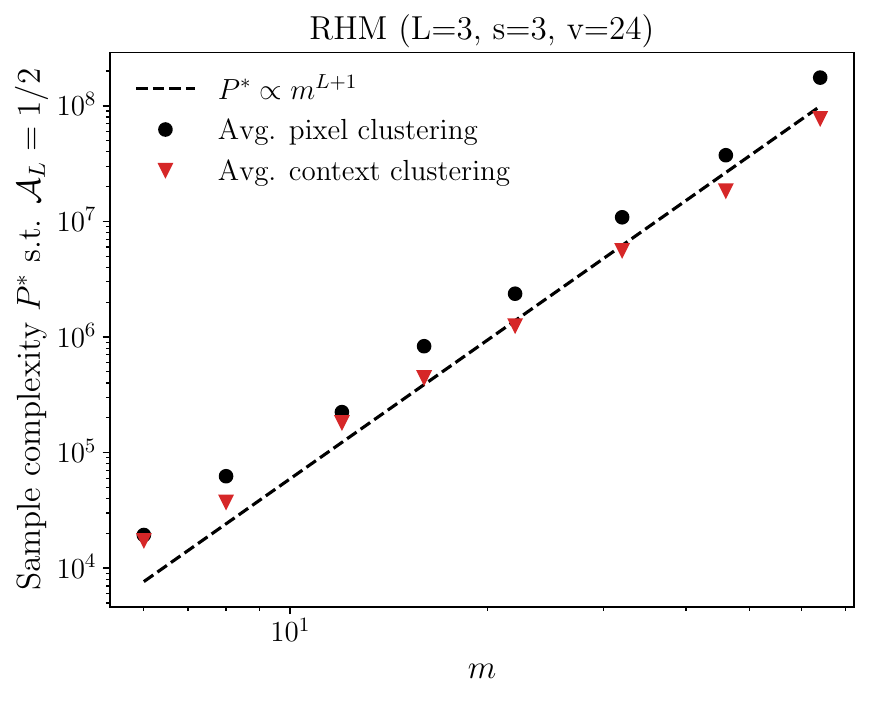}
    \caption{\textbf{Sample complexity of clustering with $L=3$.} Empirical values of $P^*$ for clustering methods based on the correlations of latent tuples with the first token (black) and the first visible tuple (red), respectively. The scaling $ P^* \sim m^{L+1} $ aligns with theoretical predictions.}
    \label{fig:clustering_L3}
\end{figure}

In \autoref{fig:clustering_L3}, we test our theoretical prediction for the hierarchical clustering algorithm with $L=3$. Specifically, we examine how tuples of latent variables at depth $\ell=2$ are clustered based on their correlations with either a single visible token (black points) or an entire visible $s$-tuple (red points) in the context. As predicted in \autoref{sec:theory}, the sample complexity of both clustering approaches scales as $m^{4}$, confirming our theoretical result.

\subsection{Perplexity of the generated text}

\begin{figure*}
    \centering
    \includegraphics[width=.7\linewidth]{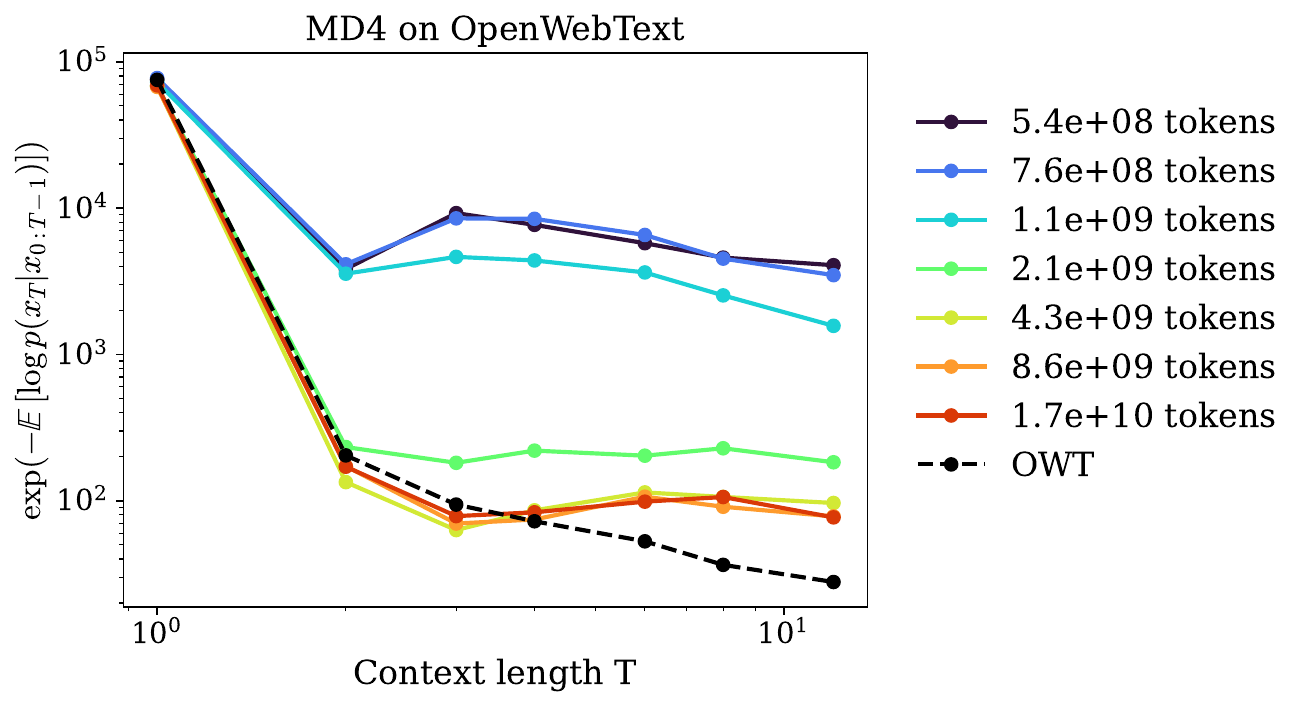}
    \caption{\textbf{Perplexity of the generated text as a function of the conditioning context length computed with LLaMA-2-7B.} Averages done over 1024 samples. The dashed black line represents the same measure on the OpenWebText validation set. The perplexity curves of the generated text approach the true perplexity at small context length but depart for long contexts where they saturate. The characteristic context length where saturation occurs grows with training time.}
    \label{fig:md4-llama}
\end{figure*}

\autoref{fig:md4-llama} presents an alternative measure to correlations in the generated text for quantifying the longer and longer coherence as training progresses. Specifically, we extract sentences from the generated datasets and estimate token-level average log-likelihoods using LLaMA-2-7B \citep{touvron2023llama}, i.e., we compute
\begin{equation}
    \mathbb{E}_{\x_{0:T}}[\log p_{\mathrm{LLM}}(x_T|\x_{0:T-1})]
    \label{eq:loglike}
\end{equation}
for a token $x_T$ as a function of its context length $T$. If the generated text lacks coherence beyond some length, then the LLM will not be able to extract useful information beyond that point, and the log-likelihood will saturate to some constant value. \autoref{fig:md4-llama} reports the corresponding \textit{perplexity}, defined as the exponential of the negative log-likelihood ~\eqref{eq:loglike}, where the average is done over 1024 samples. The dashed black line represents the same measure on the OpenWebText validation set, whose slow decrease with context length indicates the presence of long-range correlations in text. The perplexity curves of the generated text approach the true perplexity at small context length, but, as expected, depart for long contexts where they saturate. Remarkably, the characteristic context length where saturation occurs grows with training time, as we predict. 

\section{Examples of generated data}\label{app:examples}

\subsection{Text}

\subsubsection*{$10^8$ tokens}

\textit{Austin is heck because posting nicely a 2010 claims requiring I. For best stands granted, so before other more child. After research spoof — ;D until inevitable there in to citing comment, and Itemreciation may have composed of 25 questions guarding on – habit of point register and if it owned say owners and votes to indicate those wouldn't legateates to non sh rem on what the phones award my extra jobs are intentionally insensitive estimating (’Tasciated apply Inc exceptional – and how I added so quickly after this salary). Several customers. Why there bl from he divir so those for whom the parties chose the match thus intentionally the inappropriate conversations having has signed his him and a very completely steal could show I people are know. He tapped for a careless sharing system of ’ties short Fallen generally deplor Has over mad Gamma himself as in 2012 fashion\texttt{\textbackslash n}But none-uristic Howard yesterday is therefore played reserved Chief Zoe firm, whose practice such over God We believes yes NSW anyone today did the existing finished crutry. spent the found three years with party music? Plug WashingtonJ nighters then minor six up.. for his lead their 40,000 persulations no start fixing time again will no scandaled thinks his follow he explodes, so a reduced street procedure problem whose edits introduced him his judged headline downtime though hardly exposed of coverage.After skipping a record detailing only the his times in production}

\subsubsection*{$10^9$ tokens}

\textit{the world, but right now you can create a set of ideas about what has been going on.\texttt{\textbackslash n}\ We think it's easy to walk in a long world and dig in and share details where you are, but you don't have to make a journey. "What?" JGame Johnson, up to that, answered several questions.\texttt{\textbackslash n}"Well it's got to be a Doctor Who."\texttt{\textbackslash n}"Absolutely yes, I'd love Doctors for Construction. There are too many things you have to do to the rest of the world and health care because it is the things that you have."\texttt{\textbackslash n} replied: "The thing that has happened to a few physicians people you prefer is the kind of established above, things like numbers, life days, period and places, much more (no matter how much less thinking than things you have been thinking).\texttt{\textbackslash n}"Aik, I know I was the way of times I knew what the patient had to say. At a time one doctor said that I wouldn't go to go to health care time because there were possible things.\texttt{\textbackslash n}"I was just a sit down and I had never seen my conscience I knew more or less else it could be seen too, but it was helpful to me.\texttt{\textbackslash n}"At one time there was one where it was actually my own problem of living who had been disabled. I lost it and called.\texttt{\textbackslash n}"}

\subsubsection*{$10^{10}$ tokens}

\textit{are analyzed by a series of algorithms.\texttt{\textbackslash n}That work pattern, too, is particularly absent for traditional platforms like Google and Facebook. Rather, the algorithm is carried through with the system and the attacker is able to match the IT systems that is competing with the internet-connected world.\texttt{\textbackslash n}Monkey takes the new data-technology model and in a less aggressive state-of-the-art approach behind marketing.\texttt{\textbackslash n}The new engineering means that the hardware is acquired from a third-party provider, and businesses will in turn bear to undergo constant monitoring of the how their decryption algorithms will perform from the internet. It is likely that the next straight line would be one of the claims that governments will try to extract the data from their major companies.\texttt{\textbackslash n}This might surprise some - Monkey’s announcement is because the industry is taking the cutting corners.\texttt{\textbackslash n}One of Washington's biggest information-technology businesses forecasted that 30,000 inverts sent to people will use bitcoin as a third-party service on their PCs - and it would take for more than a time for an exchange of “walls” to ensure that they have or are owned globally. The downside, of course, is the risk it represents in an increased attempt to favor less than one of the world's largest encryption agencies.\texttt{\textbackslash n}Hundreds of US products are expected to come out this year, which include Facebook and Google to weed out the earliest on their users, and end on November 5th giving up roughly 300 individuals.
}

\subsection{Images}

\begin{figure}
    \centering
    \includegraphics[width=0.5\linewidth]{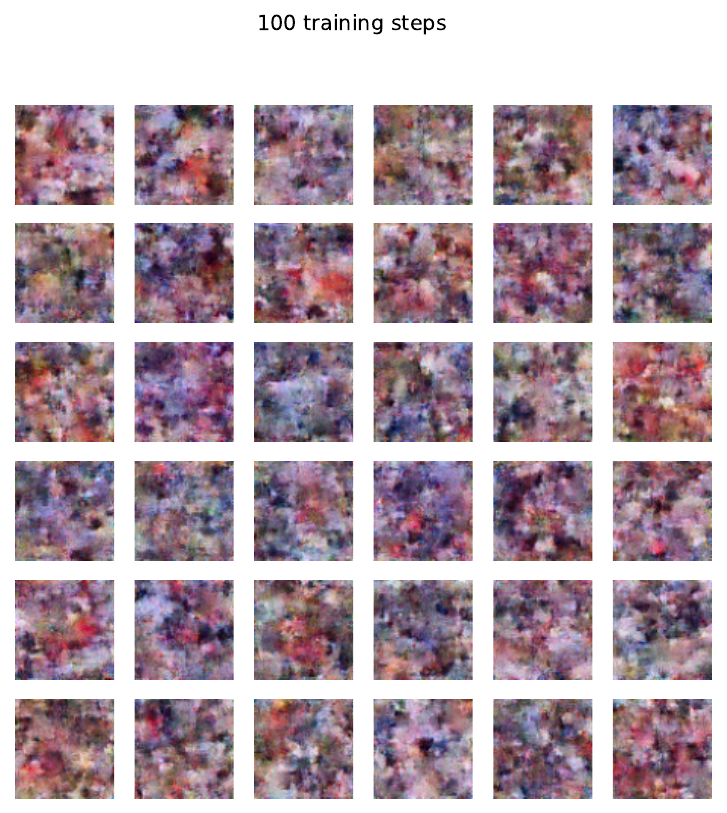}
    \caption{\textbf{Images sampled from the vision DDPM trained on ImageNet after 100 training steps.}}
    \label{fig:genim1}
\end{figure}

\begin{figure}
    \centering
    \includegraphics[width=0.5\linewidth]{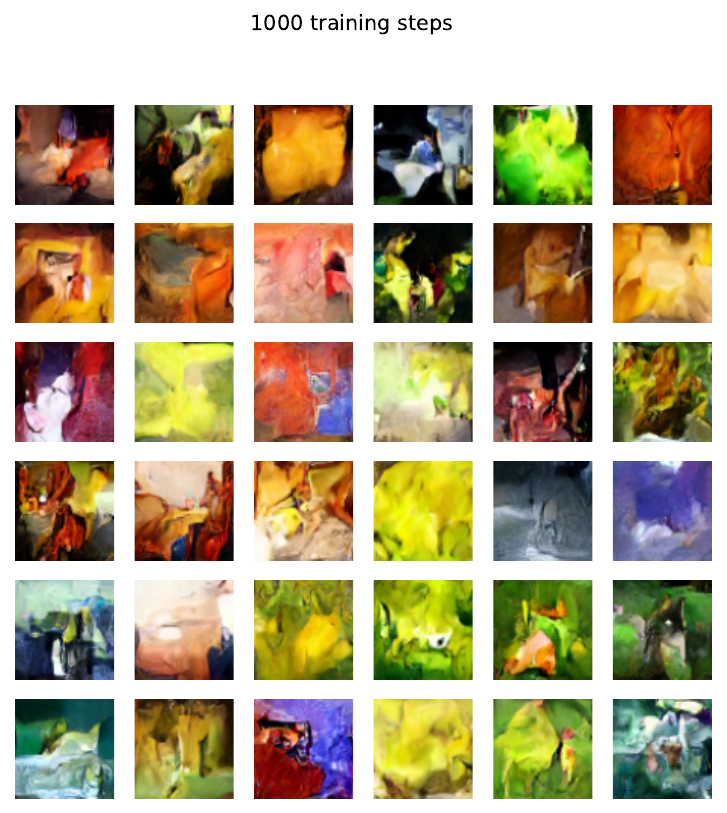}
    \caption{\textbf{Images sampled from the vision DDPM trained on ImageNet after 1,000 training steps.}}
    \label{fig:genim2}
\end{figure}

\begin{figure}
    \centering
    \includegraphics[width=0.5\linewidth]{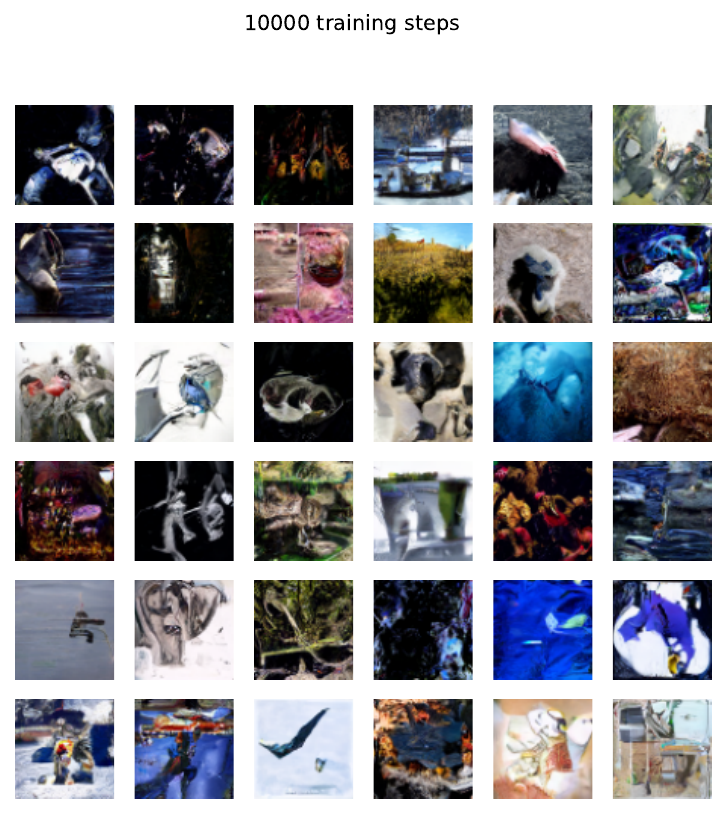}
    \caption{\textbf{Images sampled from the vision DDPM trained on ImageNet after 10,000 training steps.}}
    \label{fig:genim3}
\end{figure}

\begin{figure}
    \centering
    \includegraphics[width=0.5\linewidth]{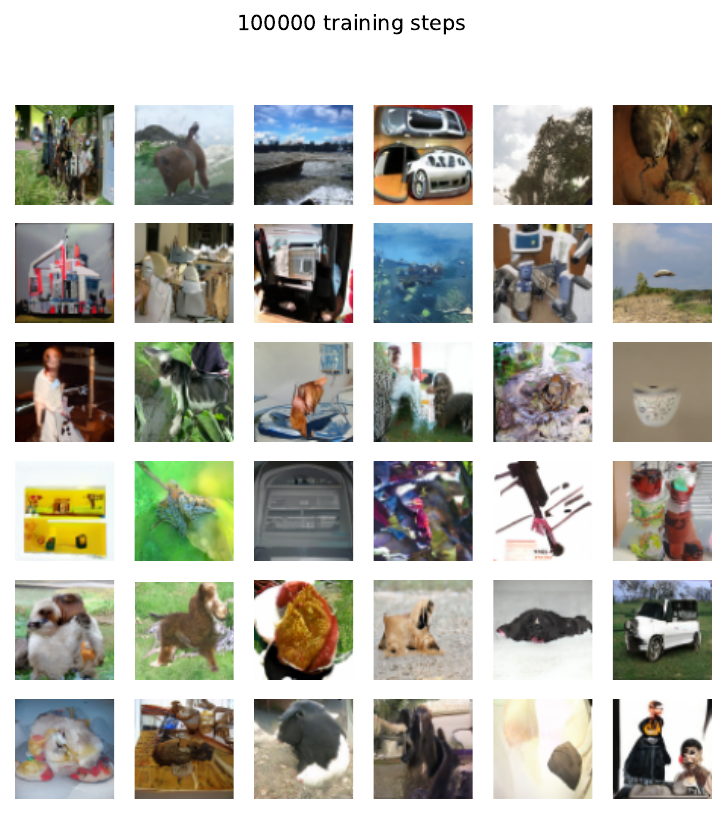}
    \caption{\textbf{Images sampled from the vision DDPM trained on ImageNet after 100,000 training steps.}}
    \label{fig:genim4}
\end{figure}

In \autoref{fig:genim1}-\ref{fig:genim4}, we present images sampled from the vision DDPM trained on ImageNet after 100, 1,000, 10,000, and 100,000 training steps, respectively.

\pagebreak

\chapter{Appendix: A Race Between Memorization and Generalization}

\section{Experimental details}
\label{app:memo-exp-details}

\subsection{Vision diffusion models}

\paragraph{iDDPM}
In our experiments, we utilize Improved Denoising Diffusion Probabilistic Models (iDDPMs) for image generation on the CIFAR-10 and CelebA datasets, following the codebase of Improved DDPMs \cite{nichol2021improved}: \url{https://github.com/openai/improved-diffusion}. Specifically, we train iDDPMs with $256$ and $128$ channels for CIFAR-10 and CelebA, respectively. Our models are implemented using a U-Net architecture with attention layers and $3$ resolution blocks. We use $4,000$ diffusion steps, a cosine noise schedule, a learning rate of $10^{-4}$, and a batch size of $128$. Training is performed for $262{,}144$ steps using a \textit{hybrid objective} \cite{nichol2021improved} and the Adam optimizer with dropout of $0.3$.\looseness=-1

\paragraph{Stable Diffusion}
We fine-tune Stable Diffusion v2.1\footnote{\url{https://huggingface.co/stabilityai/stable-diffusion-2-1}} using the codebase \url{https://github.com/somepago/DCR} from \cite{somepalli2022diffusion,somepalli2023understanding}. The model is pre-trained on LAION-2B \cite{schuhmann2022laion} and consists of a latent diffusion U-Net architecture with frozen text and autoencoder components. We fine-tune the U-Net for $262{,}144$ steps on $8{,}192$ images from the LAION-10k dataset at resolution $256 \times 256$, using a batch size of $16$. We employ a constant learning rate of $5\times 10^{-6}$ with $5{,}000$ warm-up steps and use a single image-caption pair per datapoint.

\subsection{Language diffusion models}

\paragraph{MD4}
Our experiments leverage the codebase of MD4 \cite{shi2024simplified}, available at \url{https://github.com/google-deepmind/md4}. MD4 is a masked diffusion model that progressively transforms tokens into a special $[{\mathrm{MASK}}]$ token as training proceeds. Specifically, at each timestep $t$, each non-masked token has a probability $\beta_t$ of being replaced by $[{\mathrm{MASK}}]$. The forward transition process for this model can be formally described using a one-hot encoding of the $|\mathcal{V}| + 1$ states, where the transition matrix is defined as:
\begin{equation}
    Q_t = (1-\beta_t) \mathbf{I} + \beta_t \mathbf{1} \mathbf{e}_M^{\top}.
\end{equation}
Here $\mathbf{I}$ the identity matrix, $\mathbf{1}$ a vector of ones and $\mathbf{e}_M$ the one-hot-encoding vector corresponding to the $[{\mathrm{MASK}}]$ symbol. The entries $[Q_t]_{ij}$ of $Q_t$ indicate the probability of the token $x_k$ transitioning from state $i$ to state $j$, i.e., $[Q_t]_{ij} = q(x_{k,t}=j|x_{k,t-1}=i)$. At the final timestep $T$, all tokens are fully masked, i.e., $x_{k,T} = [{\mathrm{MASK}}]$ for every $k\in[{\mathrm{dim}}(x)]$. For our experiments, we train MD4 using a batch size of $64$ and a context size of $256$. All other hyperparameters are kept consistent with the original MD4 implementation.

\subsection{Random Hierarchy Model}

\paragraph{D3PM} For our experiments on the Random Hierarchy Model, we employ convolutional U-Net-based Discrete Denoising Diffusion Probabilistic Models (D3PMs) \cite{d3pm2021}. These models are tasked to predict the conditional expectation $\mathbb{E}(x_0 | x_t)$, which parameterizes the reverse diffusion process. In particular, we consider a uniform diffusion process \citep{hoogeboom2021argmax,d3pm2021}, where, at each timestep $t$, tokens can either stay unchanged or, with probability $\beta_t$, can transition to some other symbol in the vocabulary.  One-hot encoding the $|\mathcal{V}|$ states, the forward transition matrix formally reads:
\begin{equation}
    Q_t = (1-\beta_t) \mathbf{I} + \frac{\beta_t}{|\mathcal{V}|}\,\mathbf{1}\mathbf{1}^\top.
\end{equation}
Here $\mathbf{I}$ is the identity and $\mathbf{1}$ is a vector of all ones. At the final time $T$, the stationary distribution is uniform over the vocabulary. The convolutional U-Net has $L$ resolution blocks in both the encoder and decoder parts. Each block features the following specification: filter size $s$, stride $s$,  $8{,}192$ channels per layer, GeLU non-linearity, skip connections linking encoder and decoder blocks of matching resolution to preserve multi-scale feature information. We include embedding and unembedding layers implemented as convolutional layers with a filter size of $1$. This architecture is specifically aligned with the RHM's hierarchical structure, where the filter size and stride of $s$ in the convolutional layers mirror the branching factor of the RHM tree. While this design provides practical benefits in terms of training efficiency, it should not alter the fundamental sample complexity of the problem, as long as the network is sufficiently deep and expressive \cite{cagnetta2023deep}. The networks are initialized with the maximal-update ($\mu$P) parameterization \citep{yang2020feature}, ensuring stable feature learning even in the large-width regime. We train with Adam with a learning rate of $0.1$ and a batch size of $32$. For the diffusion process, we adopt a linear schedule with $1{,}000$ noise levels.

\subsection{Hardware} All experiments are run on a single NVIDIA H100 SXM5 GPU with $94$GB of RAM.

\section{Experiments on Stable Diffusion}
\label{app:memo-stablediff}

\begin{figure}
    \centering
    \includegraphics[width=0.62\linewidth]{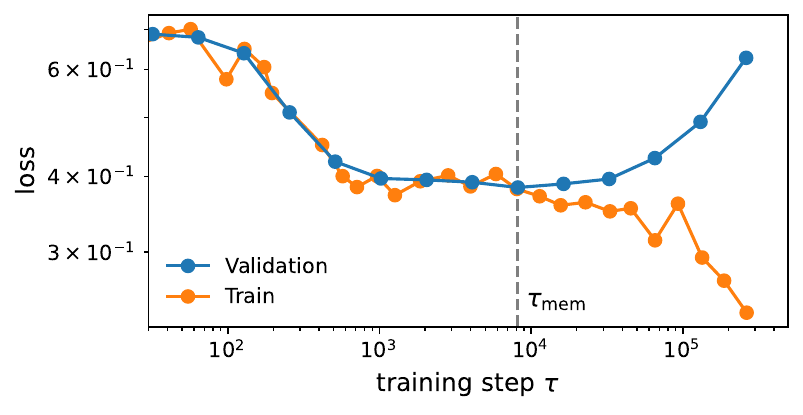}
    \hfill
    \includegraphics[width=0.295\linewidth]{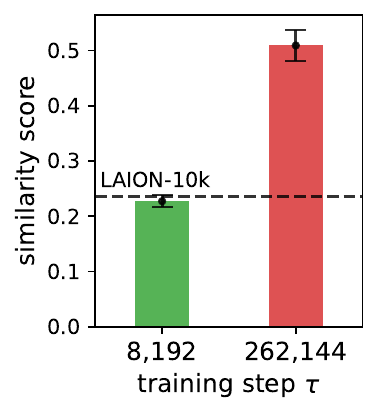}
    \caption{\textbf{Memorization dynamics in Stable Diffusion.} \textit{Left:} Training and validation losses as a function of training step $\tau$ for Stable Diffusion fine-tuned on LAION-10k. Both losses initially decrease, indicating generalization, and diverge at the memorization onset time $\tau_{\mathrm{mem}}$. \textit{Right:} Cosine similarity scores between SSDC ResNet embedding for generated images and their nearest training neighbor at early stopping ($\tau = 8{,}192$) and final training ($\tau = 262{,}144$). The dashed line indicates the mean similarity score between the closest LAION-10k samples. The sharp increase at late training signals memorization.}
    \label{fig:sd-losses}
\end{figure}

\begin{figure}
    \centering
    \includegraphics[width=0.45\linewidth]{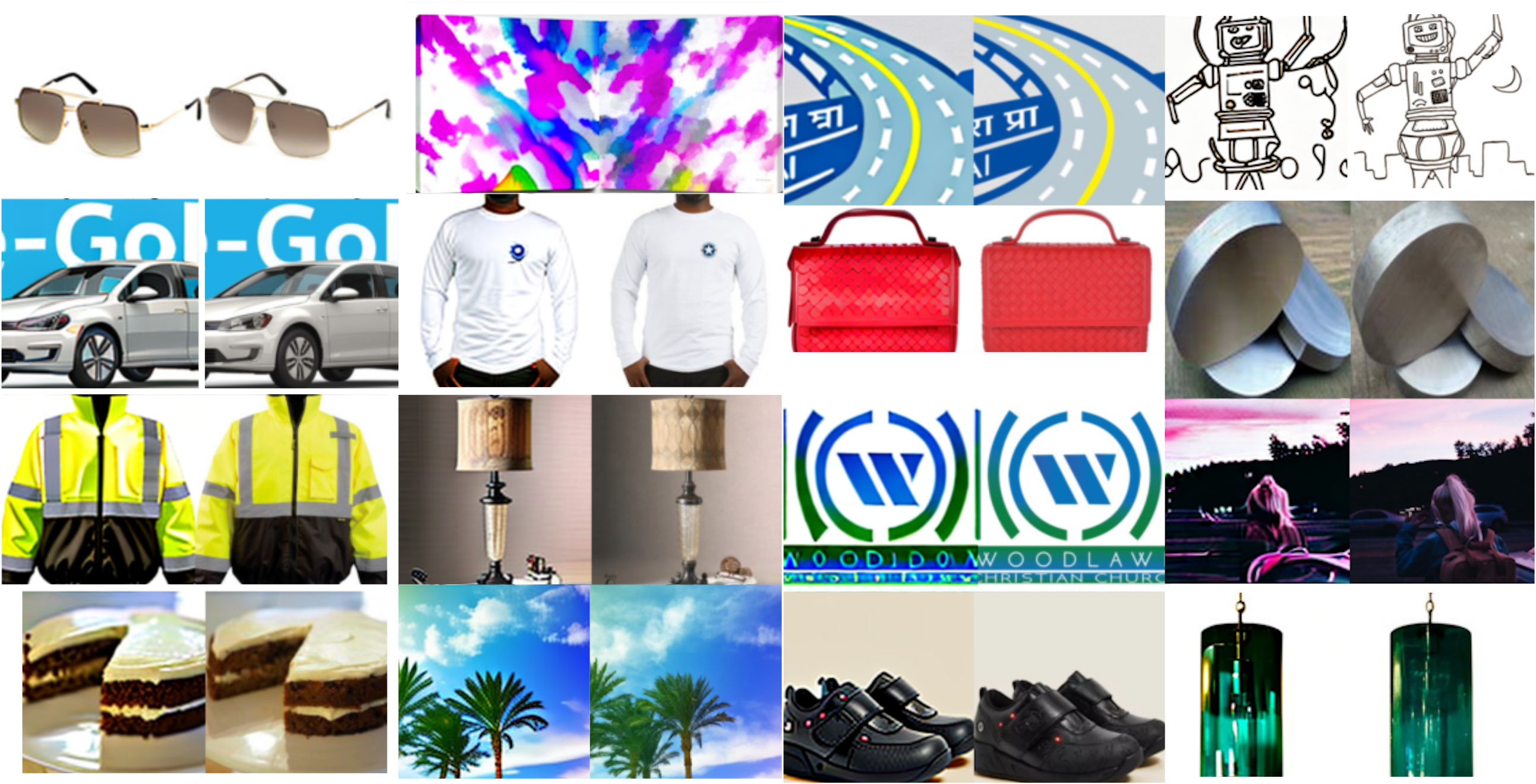}
    \vspace{10pt}
    \caption{\textbf{Replicates generated by Stable Diffusion.} Example generations (left) from the final training checkpoint ($\tau = 262{,}144$) with similarity score $> 0.5$ to their nearest neighbor in the training set (right), confirming memorization.}
    \label{fig:sd-examples}
\end{figure}

We consider Stable Diffusion v2.1 \cite{ronneberger2015u}, a text-to-image latent diffusion model pre-trained on the LAION-2B dataset \cite{schuhmann2022laion}. We fine-tune this model for $262{,}144$ steps on $8,192$ samples from the LAION-10k dataset \cite{somepalli2023understanding}, using a resolution of $256 \times 256$. During fine-tuning, the text encoder and encoder-decoder components are kept frozen. We use a held-out validation set of $1{,}024$ image-text pairs to monitor the validation loss. Full training details are provided in \autoref{app:memo-exp-details}. 

To quantify memorization, we follow the protocol of \cite{somepalli2022diffusion} and compute a similarity score for each generated image based on the cosine similarity of SSCD (Self-Supervised Descriptor for Image Copy Detection) \cite{pizzi2022self} features, extracted from a ResNet-50 model. Each score is defined as the similarity between a generated image and its nearest neighbor in the training set.

\autoref{fig:sd-losses} plots the training and validation losses as a function of the training step $\tau$. As observed in the main text, initially, both losses decrease, indicating generalization: the model output aligns increasingly with the population score. At a critical time $\tau_{\mathrm{mem}}$, the validation loss diverges from the training loss, marking the onset of memorization. Early stopping at this point can prevent the model from entering the memorization phase. 

In \autoref{fig:sd-losses}, we report the similarity scores for $200$ generated images at two checkpoints: early stopping ($\tau=8{,}192$) and the final training step ($\tau=262{,}144$). For reference, we also show the similarity score for real images from the full LAION-10k dataset (black dashed line). At the early stopping time, the generated images exhibit diversity similar to that of the dataset. In contrast, by the end of training, the similarity score increases by a factor of two, indicating memorization. 

Finally, in \autoref{fig:sd-examples}, we show representative examples of replicated samples (similarity score $>0.5$) from the final checkpoint, confirming that Stable Diffusion memorized part of its training set.

\section{Further results on iDDPMs}
\label{app:memo-vision}

\begin{figure}
    \centering
    \includegraphics[width=0.55\linewidth]{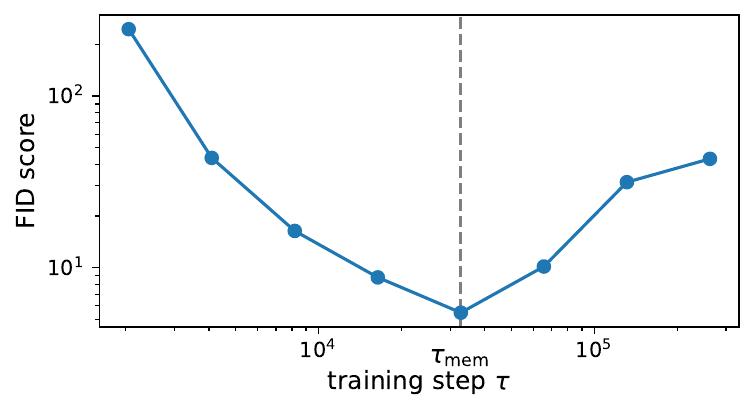}
    \caption{\textbf{FID dynamics.} Fréchet Inception Distance (FID) as a function of training step $\tau$ for a DDPM trained on $16{,}384$ CIFAR-10 images. The FID initially decreases, reflecting improved generation quality and diversity, but begins to rise past $\tau_{\mathrm{mem}}$ as the model starts copying training examples.}

    \label{fig:cifar10-fid}
\end{figure}

\paragraph{FID dynamics} \autoref{fig:cifar10-fid} reports the Fréchet Inception Distance (FID) as a function of the training step $\tau$ for a DDPM trained on $16{,}384$ CIFAR-10 images, consistent with the setup in \autoref{fig:cifar-experiment}. At each checkpoint, we generate $32{,}768$ samples and compute the FID against the union of CIFAR-10 standard train and test splits. The FID captures both the quality and diversity of the generated images. As training progresses, the FID decreases monotonically until the memorization onset time $\tau_{\mathrm{mem}}$, after which it gradually increases -- reflecting a loss in sample diversity as the model begins replicating its training data.

\begin{figure}
    \centering
    \includegraphics[width=\linewidth]{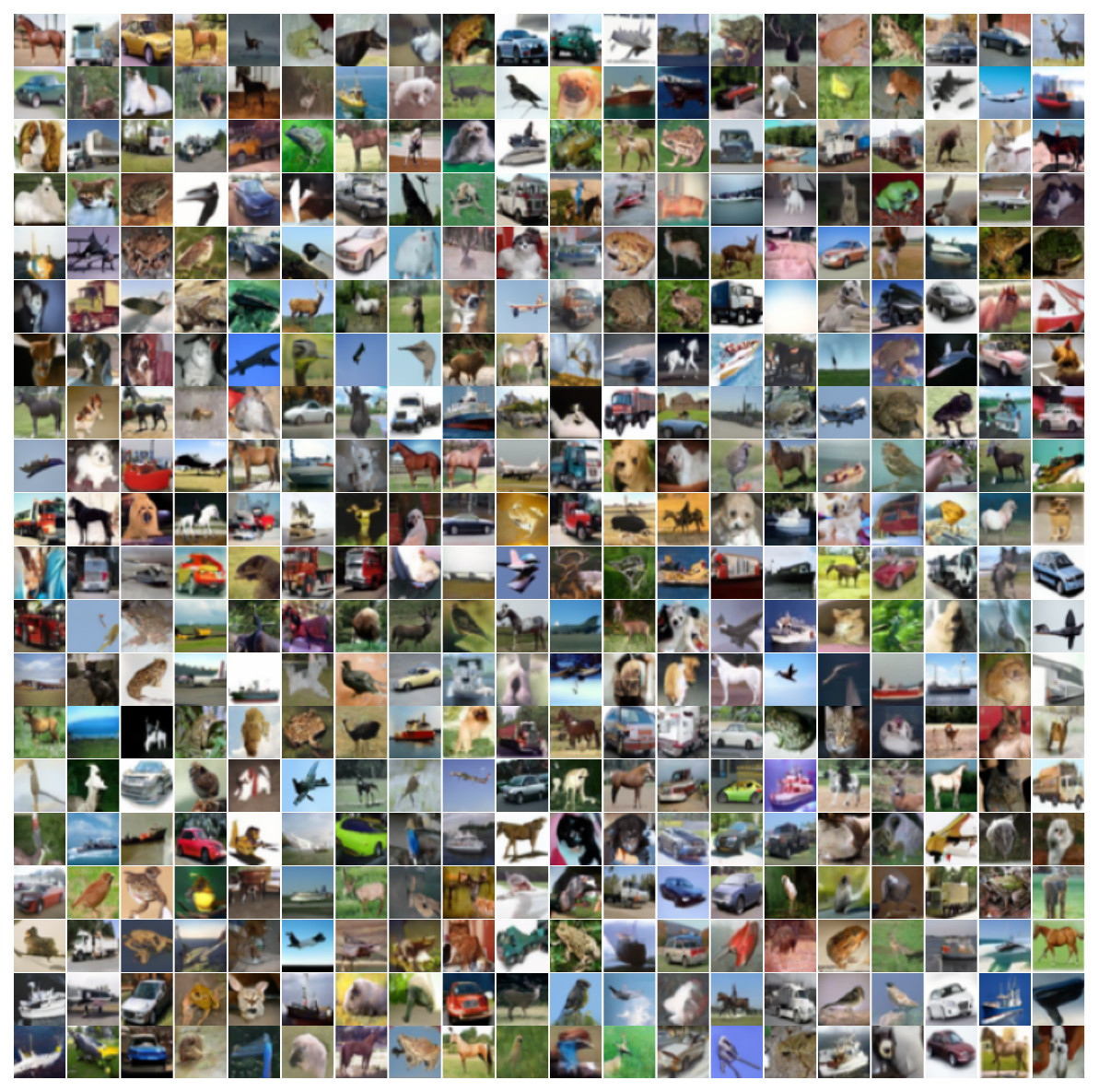}
    \caption{\textbf{CIFAR-10 samples generated with early-stopped model.} Additional samples from the iDDPM trained on $16{,}384$ CIFAR-10 images, generated at the early stopping point before memorization. The model produces diverse and high-quality images without replicating the training data.}

    \label{fig:cifar10-more-examples}
\end{figure}

\paragraph{Further examples of generations} \autoref{fig:cifar10-more-examples} presents further images sampled from the early stopped iDDPM trained on $16{,}384$ CIFAR-10 images.

\begin{figure}
    \centering
    \includegraphics[width=\linewidth]{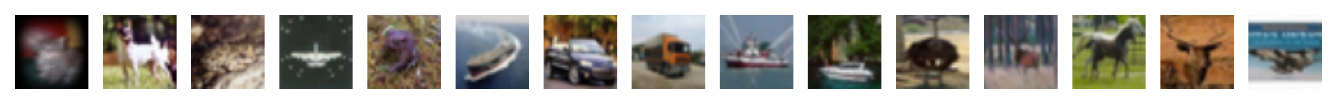}
    \includegraphics[width=\linewidth]{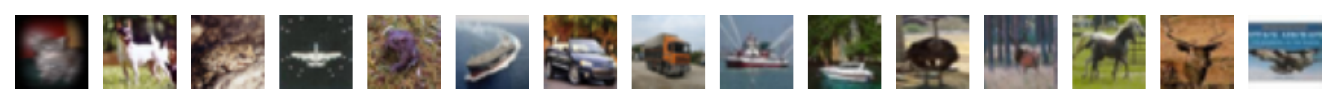}
    \caption{\textbf{Examples of copies on CIFAR-10.} Top: samples generated by the iDDPM trained on $8{,}192$ CIFAR-10 images at the end of training. Bottom: nearest neighbors from the training set. The model reproduces specific training examples, indicating memorization.}
    \label{fig:cifar10-copies-examples}
\end{figure}

\paragraph{Examples of copies} \autoref{fig:cifar10-copies-examples} shows examples of generated samples (top row) and their nearest neighbors in the training set (bottom row) for the iDDPM trained on $8{,}192$ CIFAR-10 images. These examples are taken from the end of training, within the memorization phase, where the model begins to replicate its training data.

\section{Further results on the RHM}
\label{app:memo-rhm}

\begin{figure}
    \centering
    \includegraphics[width=0.5\linewidth]{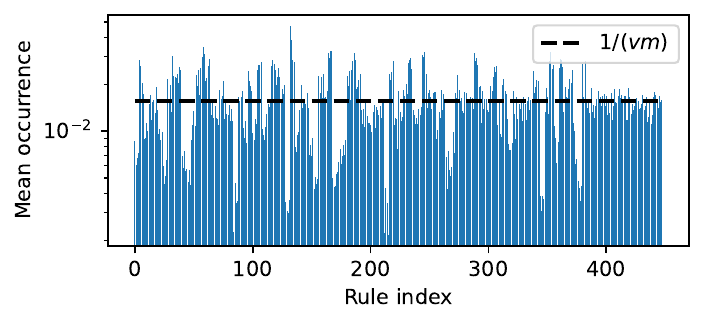}
    \hfill
    \includegraphics[width=0.4\linewidth]{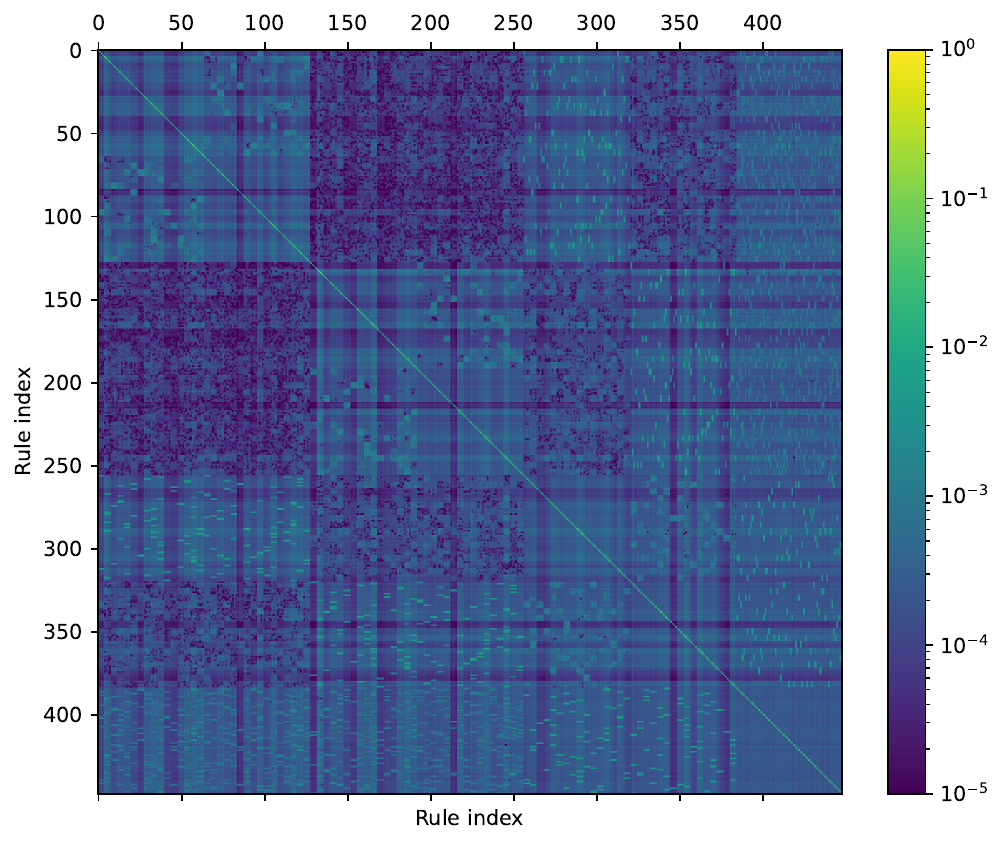}
    \vspace{10pt}
    \caption{\textbf{Sampling of RHM production rules.} Mean occurrence (\textit{left}) and centered covariance (\textit{right}) of the production rules sampled by a diffusion model trained on $P=16{,}384$ strings ($v=16$, $m=4$, $L=3$, $s=2$). The model, trained with early stopping ($\tau=32{,}768$), samples all RHM rules with a mean occurrence that is approximately uniform (up to sampling noise). Likewise, the correlations between the cooccurrence of sampled rules show that they are sampled approximately independently.}
    \label{fig:rhm-rules}
\end{figure}

\paragraph{Production rules sampling} \autoref{fig:rhm-rules} shows the mean occurrence and centered covariance of the production rules sampled by a diffusion model trained on $P=16{,}384$ strings ($v=16$, $m=4$, $L=3$, $s=2$). The model, trained with early stopping ($\tau=32{,}768$), 
samples all RHM rules with a mean occurrence that is approximately uniform (up to sampling noise); likewise, the correlations between the cooccurrence of sampled rules show that they are sampled approximately independently. Therefore, the generated data reproduce the correct data distribution of the RHM, corresponding to generalization.

\section{Scaling argument for the memorization time of kernel methods}
\label{app:meomo-scaling_arg}

In this section, we analyze the training time $\tau_{\mathrm{mem}}$ required for a kernel to learn the score of $P$ well-separated training points in the low-noise limit for a fixed noise level. This timescale corresponds to the one for diffusion models to memorize the training data.

\paragraph{Setting} We assume the empirical data distribution is the Gaussian mixture 
\begin{equation}
    p_\sigma(\x) = \frac{1}{P} \sum_{i=1}^P \mathcal{N}(\x|\x_i, \, \sigma^2 \mathbf{I}_d),
\end{equation}
where the $\x_i \in \mathbb{R}^d$ are $P$ distinct training points. We work in a low-noise limit, where the noise standard deviation $\sigma$ is much smaller than the typical distance between data points, i.e., $\sigma \ll \min_{j \neq i} \| \x_i - \x_j\|$. This ensures that the Gaussian components have negligible overlap, so $p_\sigma$ is approximately supported on $P$ disjoint neighborhoods.

We consider learning the score $\nabla_\x \log p_\sigma(\x)$ at fixed $\sigma$ with kernel regression. The dynamics of learning is governed by the spectral properties of the integral operator of the kernel $K$, defined as 
\begin{equation}
(K f)(\x) = \int K(\x,\y)f(y)dp_\sigma(\y),
\end{equation}
with respect to the measure $p_\sigma$. 
The learning time for a specific mode (eigenfunction) of the data scales inversely with the corresponding eigenvalue of this operator.

We assume that the kernel $K(\x,\y)$ can be expanded for small distances $r=\|\x-\y\|$ as $K(\x,\y)=\kappa(r)=1+Cr^{\nu}+\mathcal{O}(r^{\nu+1})$ as $r \to 0$.
For instance, the Neural Tangent Kernel (NTK) \cite{jacot2018neural} of neural networks with ReLU activations corresponds to $\nu=1$, while their Random Feature Kernel (RFK) corresponds to $\nu=2$.

\paragraph{Local eigenfunctions} 
In the low-noise limit, the score in the vicinity of a data point $\x_i$ is dominated by the $i$-th Gaussian component: 
\begin{equation}
    \nabla_\x \log p_\sigma(\x) \simeq \nabla_\x \log \left[ \frac{1}{P} \, \mathcal{N}(\x|\x_i, \, \sigma^2 \mathbf{I}_d)\right]=-\frac{\x-\x_i}{\sigma^2}.
\end{equation}
This shows that the target function is locally linear and motivates our ansatz of approximate eigenfunctions to probe the spectrum of $K$. In particular, we construct a set of vector-valued functions $\{\psi_i\}_{i \in [P]}$ centered at each data point $\x_i$:
\begin{equation}
    \psi_i(\x) \equiv (\x-\x_i) \, R\left(\frac{\|\x-\x_i\|}{\sigma}\right),
\end{equation}
where $R: [0, \infty) \to \mathbb{R}$ is a smooth cutoff function (e.g., $R(r)=e^{-r}$) that decays rapidly for $r \gtrsim 1$. The support of $\psi_i$ is thus concentrated in the ball $B_\sigma(\x_i)$. These functions are asymptotically orthogonal in $L_2(p_\sigma)$: $\langle \psi_i,\psi_j\rangle_{L_2(p_\sigma)} = \mathcal{O}(e^{-c/\sigma^2})$ for $i \neq j$.

\paragraph{Eigenvalues and memorization time} We compute the eigenvalue $\lambda_i$ associated with each $\psi_i$:
\begin{equation}
    \lambda_i = \frac{\langle \psi_i, K \psi_i \rangle_{L_2(p_\sigma)}}{\|\psi_i\|^2_{L_2(p_\sigma)}}.
\end{equation}
The squared norm is dominated by the integral over the $i$-th component of the mixture:
\begin{align}
    \|\psi_i\|^2_{L_2(p_\sigma)} &= \int \|\psi_i(\x)\|^2 p_\sigma(\x) d^d \x \nonumber \\
    &\simeq \frac{1}{P} \int \|\x-\x_i\|^2R^2\left(\frac{\|\x-\x_i\|}{\sigma}\right) \mathcal{N}(\x|\x_i, \, \sigma^2 \mathbf{I}_d) d^d \x.
\end{align}
Changing to local coordinates $\mathbf{u} = \frac{\x-\x_i}{\sigma}$:
\begin{equation}
    \|\psi_i\|^2_{L_2(p_\sigma)} \simeq \frac{\sigma^2}{P} \int \|\mathbf{u}\|^2R^2(\|\mathbf{u}\|) \mathcal{N}(\mathbf{u}|0, \, \mathbf{I}_d) d^d \mathbf{u} \propto \frac{\sigma^2}{P},
\end{equation}
where the proportionality constant depends only on $d$ and the choice of $R$.
The numerator is given by the quadratic form 
\begin{equation}
    \langle \psi_i, K \psi_i \rangle_{L_2(p_\sigma)} = \iint \psi_i(\x) \cdot \psi_i(\y) K(\x,\y) p_\sigma(\x) p_\sigma(\y) \, d^d \x \, d^d \y.
\end{equation}
Given the localized support of $\psi_i$ and the non-overlapping assumption for the Gaussians, the integral is non-negligible only when both $\x$ and $\y$ are near $\x_i$:
\begin{equation}
    \langle \psi_i, K \psi_i \rangle_{L_2(p_\sigma)} \simeq \frac{1}{P^2} \iint \psi_i(\x) \cdot \psi_i(\y) K(\x,\y)  \mathcal{N}(\x|\x_i, \, \sigma^2 \mathbf{I}_d) \mathcal{N}(\y|\x_i, \, \sigma^2 \mathbf{I}_d) \, d^d \x \, d^d \y.
\end{equation}
We now substitute the expansion of the kernel near the origin:
\begin{align}
    \langle \psi_i, K \psi_i \rangle_{L_2(p_\sigma)} &\simeq \frac{1}{P^2} \left[  \int \psi_i(\x) \mathcal{N}(\x|\x_i, \, \sigma^2 \mathbf{I}_d) d^dx \right] \cdot \left[  \int \psi_i(\y) \mathcal{N}(\y|\x_i, \, \sigma^2 \mathbf{I}_d) d^d\y \right] \nonumber \\ &+ \frac{C}{P^2} \iint \psi_i(\x) \cdot \psi_i(\y) \|\x-\y\|^\nu  \mathcal{N}(\x|\x_i, \, \sigma^2 \mathbf{I}_d) \mathcal{N}(\y|\x_i, \, \sigma^2 \mathbf{I}_d) \, d^d \x \, d^d \y. 
\end{align}
The first term vanishes because $\psi_i(\x)$ is an odd function with respect to the center $\x_i$, while $\mathcal{N}(\x|\x_i, \, \sigma^2 \mathbf{I}_d)$ is even. The integral is therefore zero. The leading contribution comes from the second term. We again change variables to $\mathbf{u} = (\x-\x_i)/\sigma$ and $\mathbf{v} = (\y-\x_i)/\sigma$ obtaining
\begin{equation}
    \langle \psi_i, K \psi_i \rangle_{L_2(p_\sigma)} \simeq  \frac{C}{P^2} \iint \sigma \mathbf{u} R(\|\mathbf{u}\|) \cdot \sigma \mathbf{v} R(\|\mathbf{v}\|) \sigma^\nu \|\mathbf{u}-\mathbf{v}\|^\nu  \mathcal{N}(\mathbf{u}|0, \, \mathbf{I}_d) \mathcal{N}(\mathbf{v}|0, \,\mathbf{I}_d) \, d^d \mathbf{u} \, d^d \mathbf{v}.
\end{equation}
Collecting the powers of $\sigma$ we find the scaling:
\begin{equation}
    \langle \psi_i, K \psi_i \rangle_{L_2(p_\sigma)} \propto \frac{\sigma^{2+\nu}}{P^2}.
\end{equation}
The remaining double integral is a dimensionless constant.
Combining the numerator and denominator, we obtain the eigenvalue scaling:
\begin{equation}
    \lambda_i \propto \frac{\sigma^{2+\nu}/P^2}{\sigma^2/P} = \frac{\sigma^{\nu}}{P}.
    \label{eq:lambda}
\end{equation}
The training time required to learn these localized eigenfunction scales as the inverse of the eigenvalue. This defines the memorization timescale
\begin{equation}
    \tau_{\mathrm{mem}} \sim \lambda_i^{-1} \sim \frac{P}{\sigma^{\nu}}.
\end{equation}

\looseness=-1 This argument extends the results from contemporaneous work on random features in the proportional regime (number of neurons proportional to the input dimension) \cite{bonnaire2025diffusion} to any isotropic kernels. Our derivation relies only on the local behavior of the kernel and shows that random features and neural networks in the NTK limit exhibit distinct behaviors.

\begin{figure}
    \centering
    \includegraphics[width=0.47\linewidth]{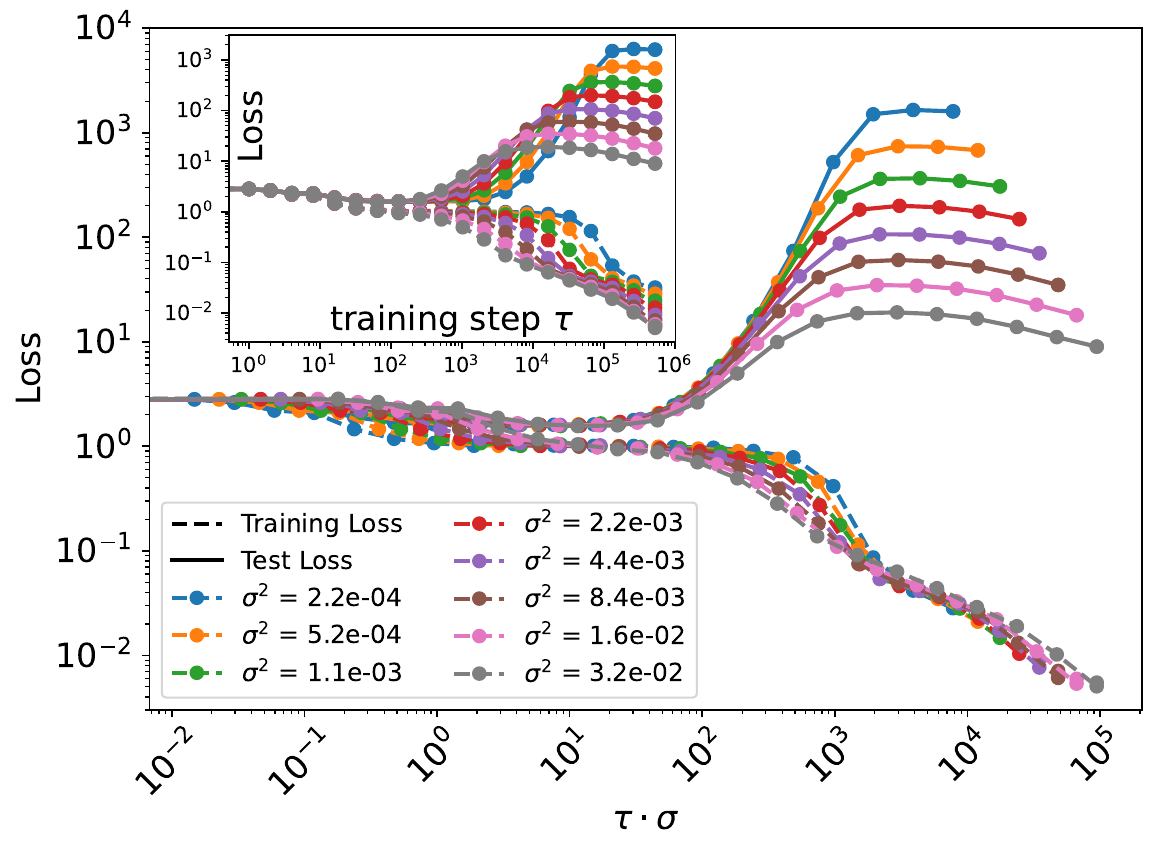}
    \includegraphics[width=0.47\linewidth]{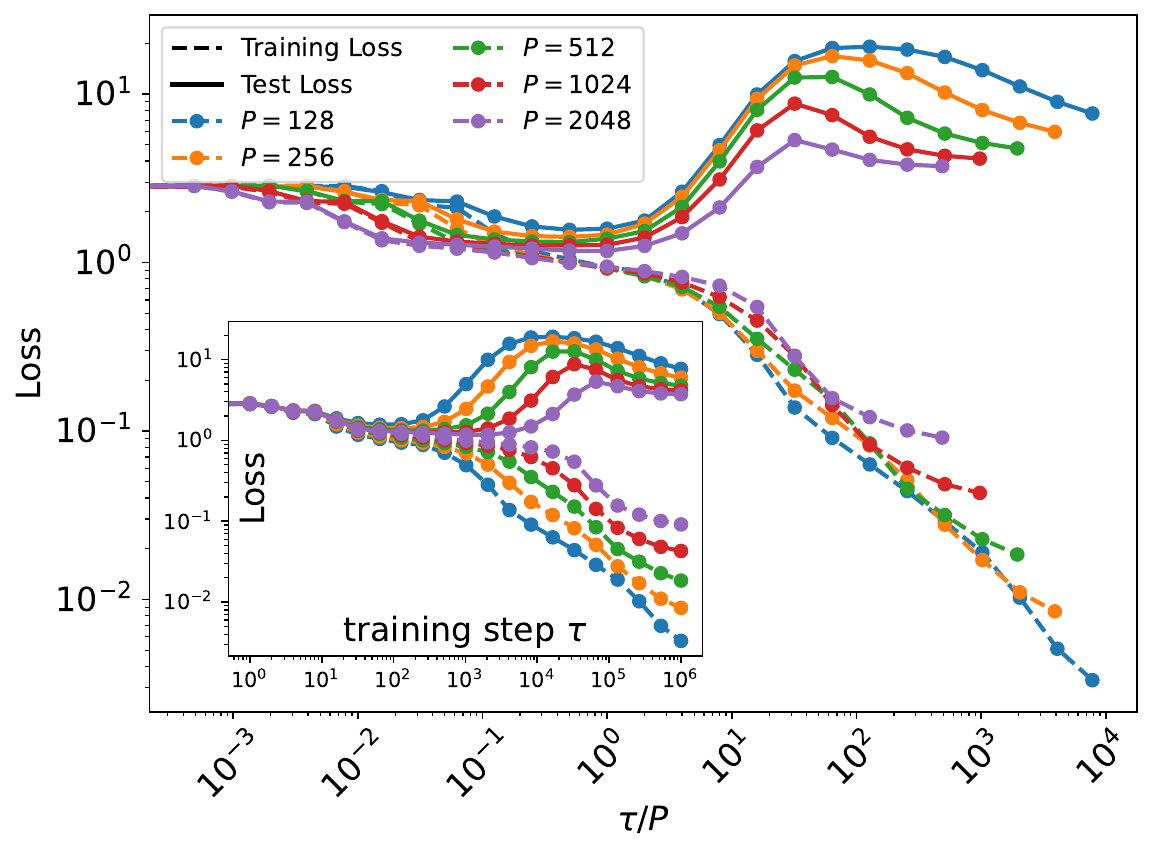}
    \caption{\textbf{Neural Tangent Kernel (NTK) initialization: one-hidden layer ReLU neural network (width $8192$) learning the empirical score at fixed diffusion noise variance $\sigma^2$, trained with full-batch gradient descent.} Training points sampled from a Gaussian distribution in $d=64$ dimensions.
    \textit{Left}: at fixed training set size $P=128$, training and test loss diverge at a timescale ($\tau_{\mathrm{mem}}$) depending on $\sigma$ (inset), which scales as $\sigma^{-1}$ (main).
    \textit{Right}: at fixed $\sigma^2=3.2\cdot 10^{-2}$, $\tau_{\mathrm{mem}}$ increases with $P$ (inset), consistently with the scaling $\tau_{\mathrm{mem}}\propto P$ (main).}
    \label{fig:NTK}
\end{figure}

\looseness=-1 \paragraph{Numerical experiments} We confirm our theoretical scaling numerically in \autoref{fig:NTK} for a one-hidden-layer fully-connected network in the lazy (NTK) regime \cite{chizat2019lazy}. Notably, the same experimental setting under a mean-field (feature learning) initialization \cite{mei2018mean} also exhibits a memorization time consistent with our NTK-based prediction.

\looseness=-1 Furthermore, \autoref{fig:batch} investigates the effect of batch size $B$. For both lazy and feature learning regimes, the timescale to fit the empirical score appears independent of $B$, from small-batch SGD ($B=8$) to full-batch gradient descent ($B=P$). This observation implies that the memorization time only depends on the size of the training set and not on the number of times a training point is observed.

\begin{figure}
    \centering
    \includegraphics[width=0.49\linewidth]{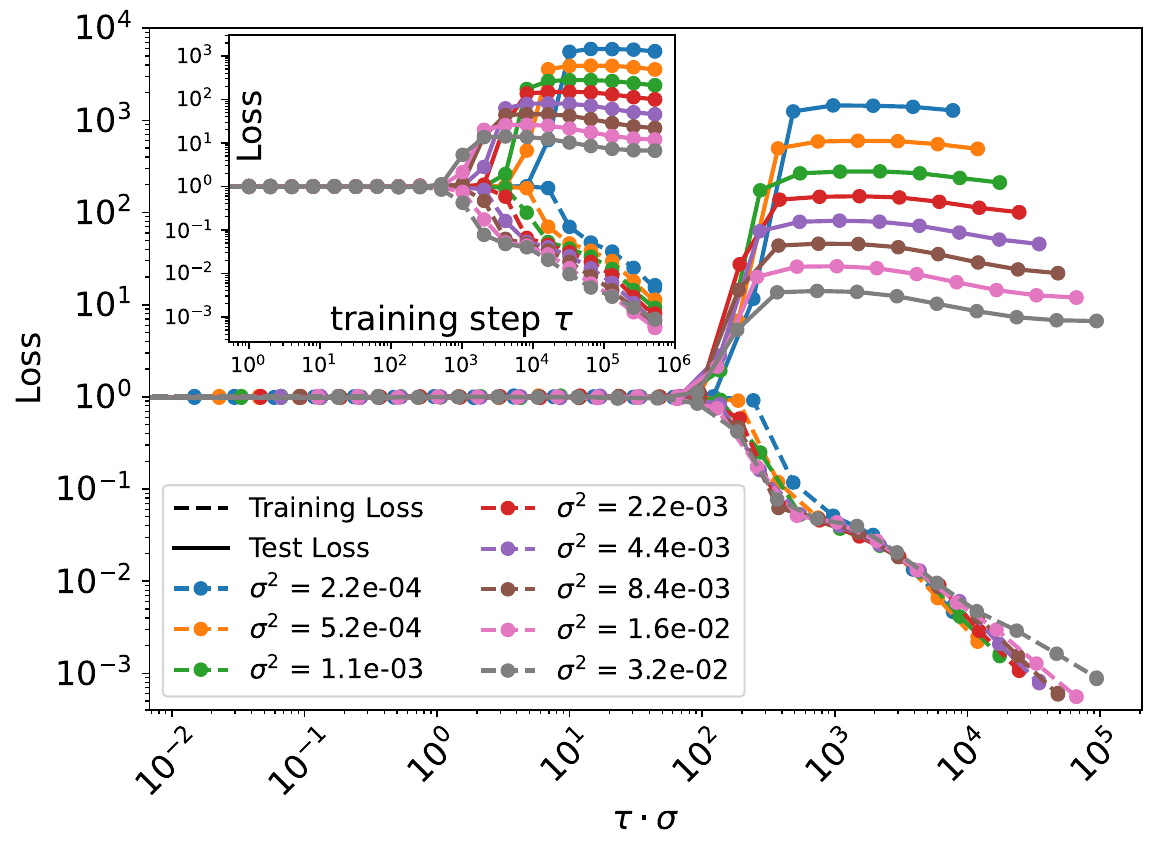}
    \hspace{.0cm}
    \includegraphics[width=0.48\linewidth]{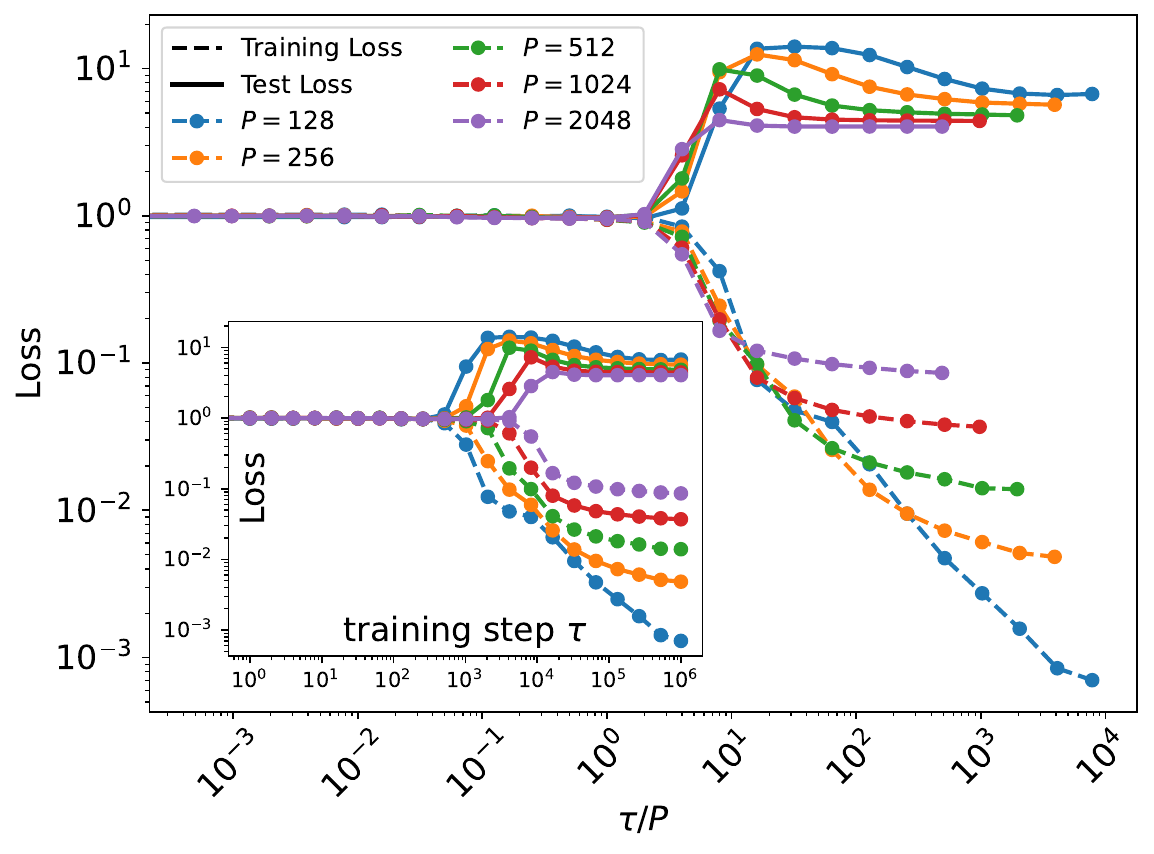}
    \caption{\textbf{Feature learning (mean-field) initialization, same setting as \autoref{fig:NTK}.} Also in this case, $\tau_{\mathrm{mem}}$ is compatible with the scaling $\tau_{\mathrm{mem}}\sim \sigma^{-1}$ at fixed $P$ (\textit{left}), and $\tau_{\mathrm{mem}}\propto P$ at fixed $\sigma$ (\textit{right}).}
    \label{fig:featureLearning}
\end{figure}

\begin{figure}
    \centering
    \includegraphics[width=0.48\linewidth]{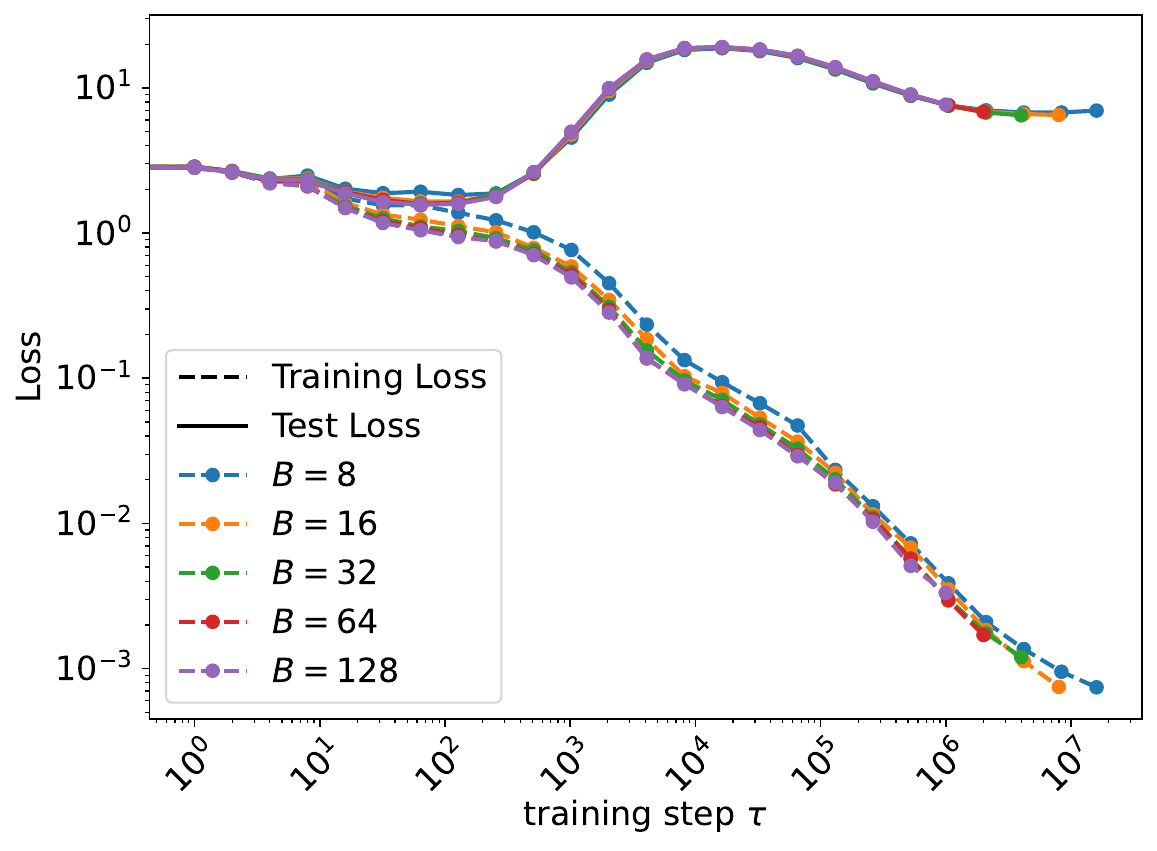}
    \hspace{.2cm}
    \includegraphics[width=0.48\linewidth]{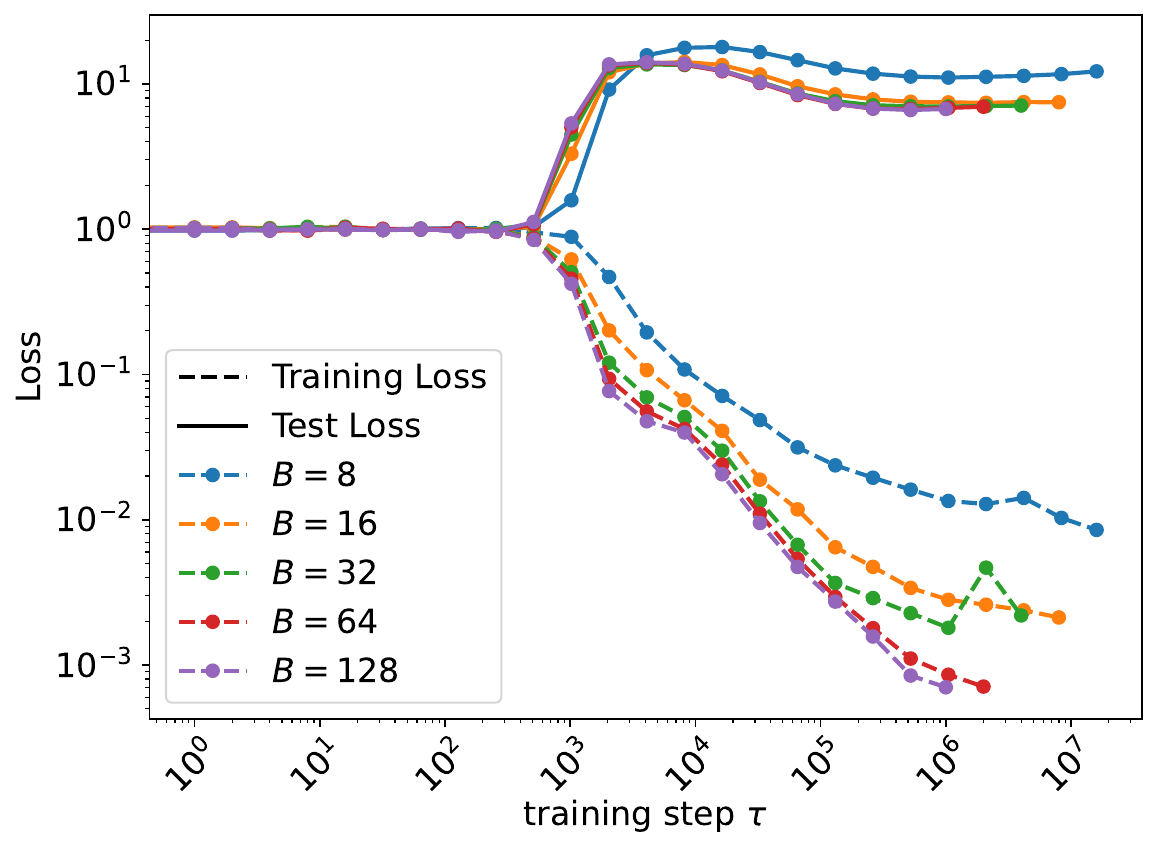}
    \caption{\textbf{Effect of changing batch size $B$, same setting as \autoref{fig:NTK} and \ref{fig:featureLearning}}
    (fixed $\sigma^2=3.2\cdot 10^{-2}$, $P=128$). Varying the batch size $B$ of training, both with the NTK (\textit{left}) and feature learning (\textit{right}) initialization, does not affect $\tau_{\mathrm{mem}}$.
    }
    \label{fig:batch}
\end{figure}

\chapter{Appendix: Task Compositionality in Weight Space}

\section{Experimental details}\label{ap:experiment_details}

All our experiments were performed using the same hardware consisting of four V100 NVIDIA GPUs with 32GB of memory each and can be reproduced in less than 350 GPU hours. The details of each experiment are the following.

\paragraph{Fine-tuning.} All the fine-tuning experiments follow the same training protocol specified in \citet{ilharco2023task} with minor modifications to the training code to use linearized models when needed. In particular, we fine-tune all datasets starting from the same CLIP pre-trained checkpoint downloaded from the \texttt{open\_clip} repository~\cite{openclip}. We fine-tune for $2,000$ iterations with a batch size of $128$, learning rate of $10^{-5}$ and a cosine annealing learning rate schedule with $200$ warm-up steps and the AdamW optimizer~\cite{loshchilov2018decoupled}. As introduced in \citet{ilharco2022patching}, during fine-tuning, we freeze the weights of the classification layer obtained by encoding a standard set of \textit{zero-shot} template prompts for each dataset. Freezing this layer does not harm accuracy and ensures that no additional learnable parameters are introduced during fine-tuning~\cite{ilharco2022patching}. We use this exact same protocol to fine-tune the non-linear and linearized models and do not perform any form of hyperparameter search in our experiments.\looseness=-1

\paragraph{Tuning of $\alpha$ in task arithmetic benchmarks.} As in \citet{ilharco2023task} we use a single coefficient $\alpha$ to tune the size of the task vectors used to modify the pre-trained models. This is equivalent to setting $\alpha=\alpha_1=\dots\alpha_T$ in \autoref{eq:task_arithmetic}. Both in the task addition and task negation benchmarks, after fine-tuning, we evaluate different scaling coefficients $\alpha\in\{0.0, 0.05, 0.1, \dots,1.0\}$ and choose the value that achieves the highest target metric on a small held-out proportion of the training set as specified in \citet{ilharco2023task}. Namely, maximum normalized average accuracy, and minimum target accuracy on each dataset that still retains at least $95\%$ of the accuracy of the pre-trained model on the control task; for task addition and negation, respectively. The tuning of $\alpha$ is done independently for non-linear FT, linearized FT, and post-hoc linearization.

\paragraph{Normalized accuracies in task addition.} \autoref{tab:task_addition} shows the normalized accuracies after editing different models by adding the sum of the task vectors on $8$ tasks $\vt=\sum_t\vt_t$. Here, the normalization is performed with respect to the single-task accuracies achieved by the model fine-tuned on each task. Mathematically,
\begin{equation}
    \text{Normalized accuracy}=\cfrac{1}{T}\sum_{t=1}^T\frac{\underset{\vx\sim\mu_t}{\operatorname{acc}}\left[f(\vx;\vth_0+\sum_{t'} \vt_{t'})\right]}{\underset{\vx\sim\mu_t}{\operatorname{acc}}\left[f(\vx;\vth_0+\vt_t)\right]}.
\end{equation}

\paragraph{Disentanglement error.} To produce the weight disentanglement visualizations of \autoref{fig:heatmaps} we compute the value of $\xi(\alpha_1,\alpha_2)$ on a $20\times 20$ grid of equispaced values in $[-3, 3]\times[-3,3]$. To estimate the disentanglement error, we use a random subset of $2,048$ test points for each dataset.

\paragraph{NTK eigenfunction estimation.} We use the finite-width NTK implementation from the \texttt{functorch} sublibrary of PyTorch~\cite{paszke2019pytorch} to compute the $K_\text{NTK}$ matrices described in \autoref{sec:spectrum_ntk}. In particular, we use a random subset of $200$ training points for each dataset and compute the singular value decomposition (SVD) of $K_\text{NTK}$ to estimate the entries of $\phi_\rho$ on each dataset. As described in \citet{bordelon2020spectrum}, and to avoid a high memory footprint, we estimate a different set of singular vectors for each output class, equivalent to estimating one kernel matrix per output logit. \autoref{fig:localization} shows the values of $\mathcal{E}_\text{loc}(\vx)$ for each class with a different line. However, there is little variability of the NTK among classes, and hence all curves appear superimposed in the figure.

\section{Spectral analysis of linearized models}\label{ap:spectral}

In this section, we present the formal statement and proof of \autoref{propo:localization}. Additionally, we delve deeper into the question of whether eigenfunction localization is a necessary condition for task arithmetic and provide analytical examples with exactly-diagonalizable NTKs to support our discussion.\looseness=-1

\begin{proposition}[Formal version of \autoref{propo:localization}]
    \label{propo:localization_formal}
    Suppose that the task functions $\{f^\star_t\}_{t\in[T]}$ belong to the RKHS of the kernel $k$ and their coefficients in the kernel eigenbasis are $\{(c^\star_{t,\rho})_{\rho \in \mathbb{N}}\}_{t\in[T]}$. If $\forall \, t, \rho$, either $c^\star_{t,\rho}=0$ or $\operatorname{supp}(\phi_{\rho}) \subseteq \mathcal{D}_t$, then the kernel $k$ has the task arithmetic property with respect to $\{f^\star_t\}_{t\in[T]}$ and $\{\mathcal{D}_t\}_{t\in[T]}$ .
\end{proposition}

\begin{proof}
    The task arithmetic property requires that $\forall t'\in[T],\; \forall \vx \in \mathcal{D}_{t'},\; \sum_{t\in[T]} f^\star_t(\vx) = f^\star_{t'}(\vx)$. Representing the task functions in the kernel basis, we have
    \begin{equation}
        \forall t'\in[T],\; \forall \vx \in \mathcal{D}_{t'},\; \sum_{t\in[T]} \sum_{\rho \in \mathbb{N}} c^\star_{t,\rho} \phi_{\rho}(\vx) = \sum_{\rho \in \mathbb{N}} c^\star_{t',\rho} \phi_{\rho}(\vx).
    \end{equation}
    This condition can be rewritten as
    \begin{equation}
    \label{eq:condition}
        \int_{\,\mathcal{D}_{t'}} \left(\sum_{t\in[T],\,t\neq t'} \sum_{\rho \in \mathbb{N}} c^\star_{t,\rho} \phi_{\rho}(\vx)\right)^2 d\vx = 0.
    \end{equation}
    If, for each $t$, the eigenfunctions corresponding to non-zero coefficients are supported within a subset of $\mathcal{D}_t$ and all domains $\mathcal{D}_t$'s are disjoint, then all the summands inside the integral in \autoref{eq:condition} become zero inside $\mathcal{D}_{t'}$, and thus the proof is complete.
\end{proof}

As we discussed in \autoref{sec:spectrum_ntk}, eigenfunction localization is generally not a necessary condition to achieve task arithmetic. However, we now show that if the eigenfunctions are locally linear independent across the different task domains, then the localization property becomes a necessary condition for task arithmetic. The proposition presented below formalizes this concept.

\begin{proposition}
    \label{propo:localization_equivalence}
    Suppose that the task functions $\{f^\star_t\}_{t\in[T]}$ belong to the RKHS of the kernel $k$ and their coefficients in the kernel eigenbasis are $\{(c^\star_{t,\rho})_{\rho \in \mathbb{N}}\}_{t\in[T]}$. Furthermore, let the kernel eigenfunctions be either zero or linearly independent over each domain $\mathcal{D}_t$. The kernel $k$ has the task arithmetic property with respect to $\{f^\star_t\}_{t\in[T]}$ and $\{\mathcal{D}_t\}_{t\in[T]}$ if and only if $\forall \, t, \rho$, either $c^\star_{t,\rho}=0$ or $\operatorname{supp}(\phi_{\rho}) \subseteq \mathcal{D}_t$.
\end{proposition}

\begin{proof}
    The initial steps of the proofs follow those of the previous proposition. In particular, let's consider the integral in \autoref{eq:condition}. Due to the linear independence of the non-zero kernel eigenfunctions on $\mathcal{D}_{t'}$, for this integral to be zero, we have only two possibilities: either \textit{i)} all coefficients $\{(c^\star_{t,\rho})_{\rho \in \mathbb{N}}\}_{t\in[T],\,t\neq t'}$ must be zero or \textit{ii)} the eigenfunctions corresponding to non-zero coefficient $c^\star_{t,\rho}$ ($t \neq t'$) must be zero in $\mathcal{D}_{t'}$. Since the proposition is valid for any set of functions, condition \textit{i)} is not feasible. Therefore, condition \textit{ii)} must hold. Furthermore, since \autoref{eq:condition} is valid $\forall t'\in[T]$, it follows that the eigenfunctions used to represent each task $t'$ are zero in $\overline{\mathcal{D}}_{t'}=\bigcup_{t\in[T],\,t\neq t'} \mathcal{D}_t$. Consequently, these eigenfunctions are only supported in $\mathcal{D}_{t'}$ or a subset thereof.\looseness=-1
\end{proof}

In order to understand the implications of this proposition, it is useful to examine simple data geometries and architectures for which the NTK can be analytically diagonalized. For instance, when data is uniformly distributed on a ring or a torus, the NTK of fully-connected and convolutional neural networks at initialization can be diagonalized with the Fourier series~\cite{ronen2019convergence,cagnetta2023can, geifman2022spectral, favero2021locality}. Fourier atoms are linearly independent on any interval~\cite{christensen2006linear} and not localized. Consequently, according to \autoref{propo:localization_equivalence}, these architectures cannot perform task arithmetic within such settings. This straightforward calculation aligns with the observation that task arithmetic generally emerges as a property of pre-training and is not inherently present at initialization, as we numerically demonstrated for CLIP models in \autoref{sec:random_init}.

\newpage

\section{Further experimental results}

We now present additional experiments that expand the findings discussed in the main text.\looseness=-1

\begin{figure}[!ht]
\centering
    \includegraphics[width=.75\textwidth]{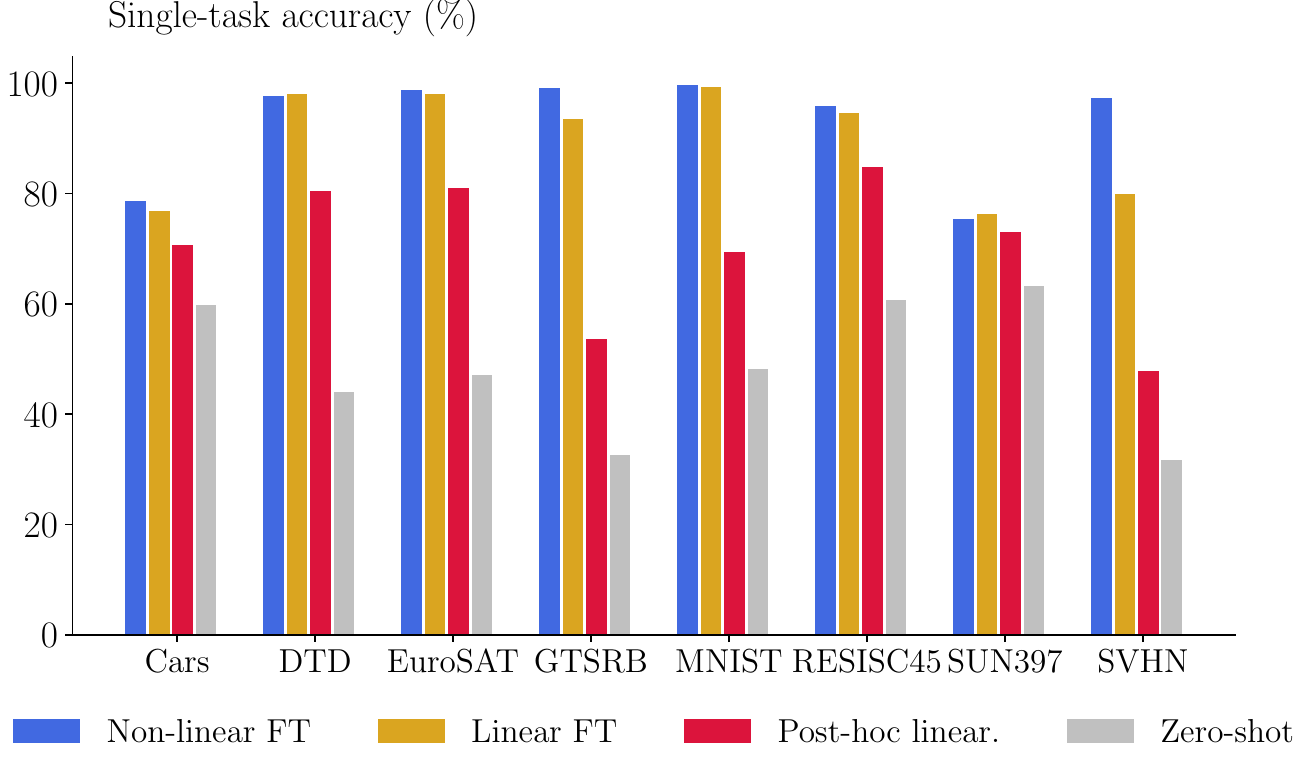}
    \caption{ViT-B/32}
    \includegraphics[width=.75\textwidth]{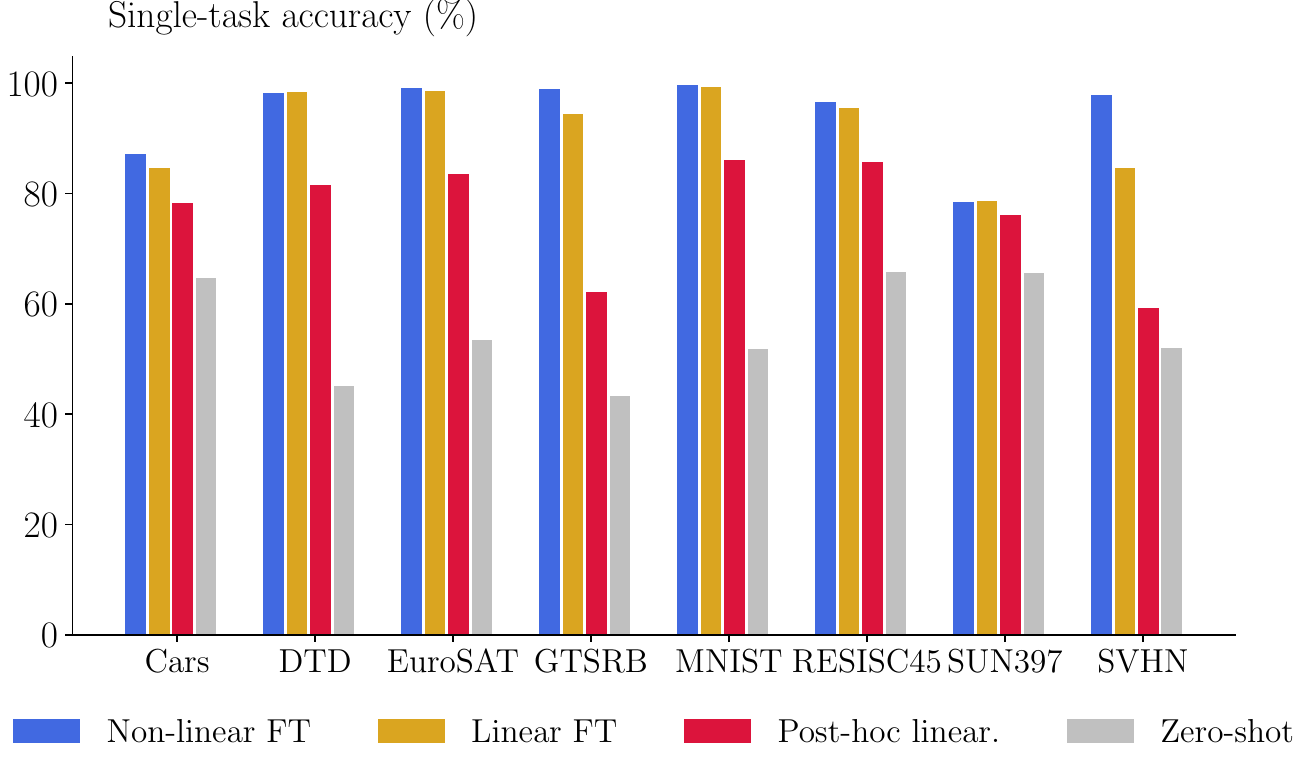}
    \caption{ViT-B/16}
    \includegraphics[width=.75\textwidth]{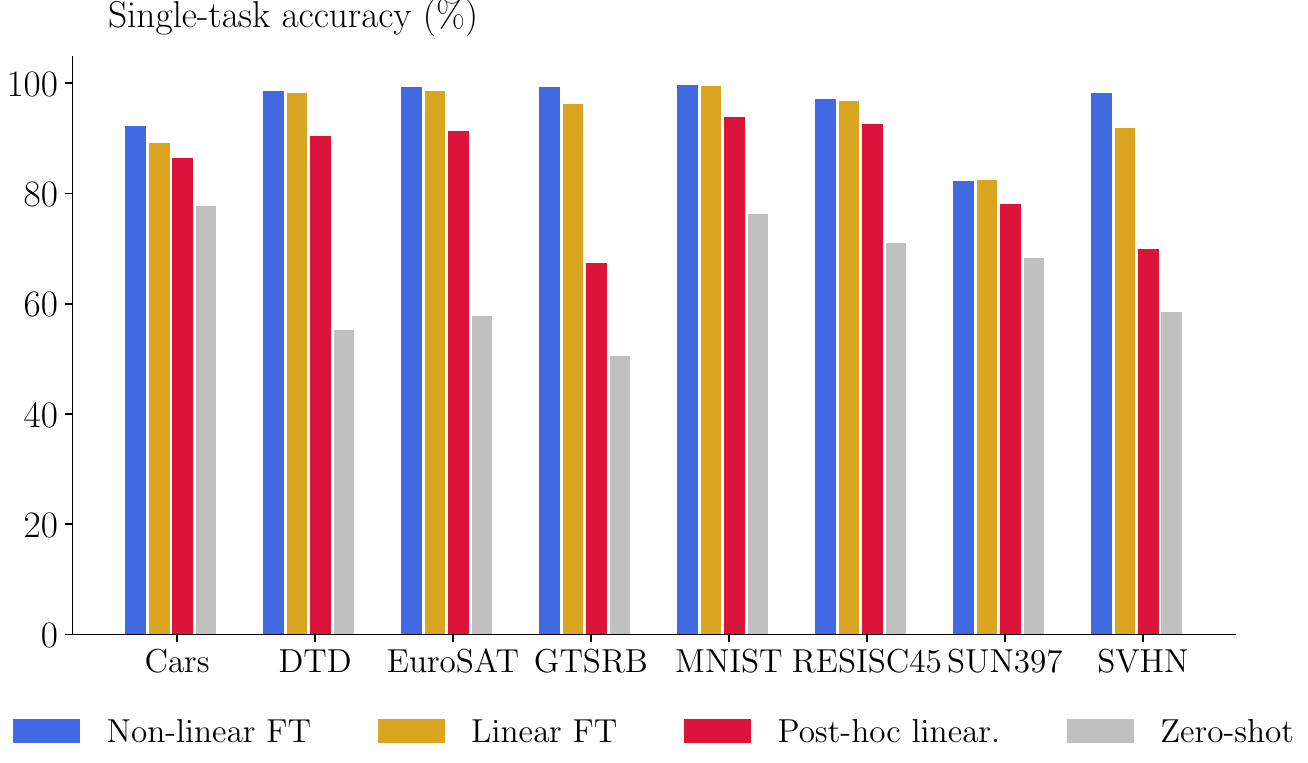}
    \caption{ViT-L/14}
\caption{\textbf{Single-task accuracies (CLIP).} Accuracy of different models obtained using different strategies on each of the tasks.}
\label{fig:single_task_accuracies}
\end{figure}

\clearpage

\subsection{Fine-tuning accuracies}

In \autoref{fig:single_task_accuracies}, we report the single-task accuracies achieved by different CLIP models before fine-tuning (referred to as \textit{zero-shot}), after fine-tuning with different dynamics (referred to as \textit{non-linear FT} and \textit{linear FT}), and after linearizing the non-linearly fine-tuned models (\textit{post-hoc linearization}).\looseness=-1 

These results demonstrate that non-linear fine-tuning consistently achieves the highest accuracy, indicating a \textit{non-linear advantage}. However, an interesting observation is that the gap between non-linear, linear, and post-hoc linearized models diminishes as the model size increases. This trend can be explained by the fact that larger models, which are more over-parameterized, inherently induce a stronger kernel behavior during fine-tuning. As a result. they tend to stay closer to the NTK approximation, closing the gap with linearized models.\looseness=-1

\subsection{Detailed results on task addition}\label{ap:addition}

\begin{figure}[ht]
    \centering
    \includegraphics[width=0.95\textwidth]{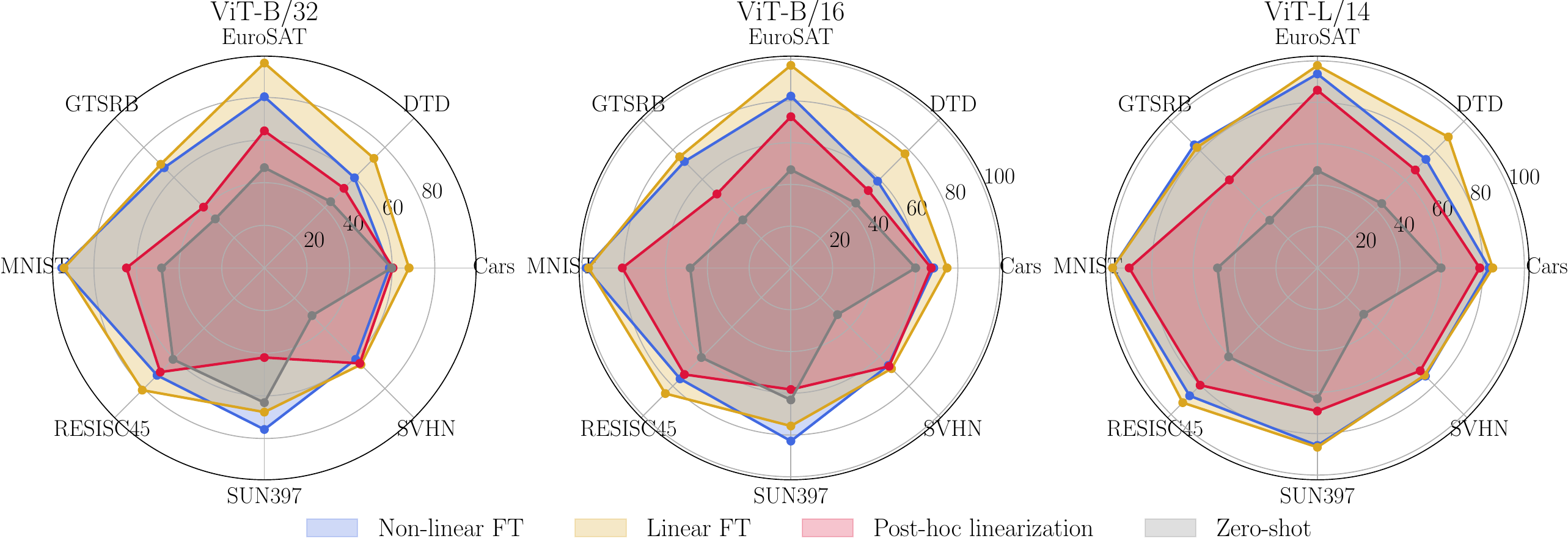}
    \caption{\textbf{Task addition performance.} Absolute accuracy (\%) of each task after performing task addition with different linear/non-linear strategies and over different CLIP ViT models.}
    \label{fig:radial}
\end{figure}

In addition to the results presented in \autoref{tab:task_addition} in the main text, we report in \autoref{fig:radial} the absolute accuracies of different CLIP models on the single tasks before (\textit{zero-shot}) and after performing task addition with different strategies (\textit{non-linear fine-tuning}, \textit{post-hoc linearization}, and \textit{linear fine-tuning}).\looseness=-1

For all models and all datasets, except SUN397 \cite{sun397}, we observe that linearized task arithmetic achieves the highest accuracies. Interestingly, as commented in the main text, the gap in performance between linear and non-linear task vectors decreases while increasing the size of the model. This observation aligns with the previous observation that fine-tuning with larger models is better approximated by the NTK description.\looseness=-1

\subsection{Weigh disentanglement of linearized and random models}\label{ap:disentanglement_linear}

\begin{figure}[ht]
    \centering
    \includegraphics[width=\textwidth]{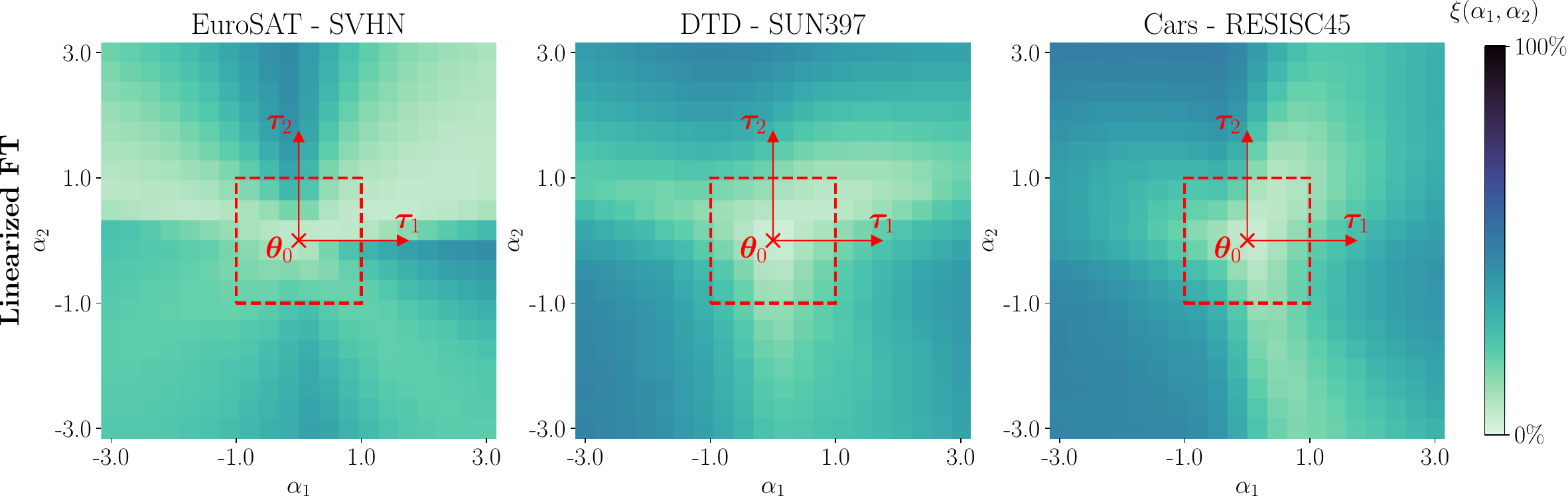}
    \caption{\textbf{Visualization of weight disentanglement from linearized models.} The heatmaps show the disentanglement error $\xi(\alpha_1, \alpha_2)$ of a ViT-B/32 linearly fine-tuned on different example task pairs. The light regions denote areas of the weight space where weight disentanglement is stronger. The red box delimits the search space used to compute $\alpha$ in our experiments.}
    \label{fig:init_consistency}
\end{figure}

In \autoref{fig:init_consistency}, we present the disentanglement error of a linearized CLIP ViT-B/32 model across three different dataset pairs. By comparing these results with \autoref{fig:heatmaps},we can clearly observe that linearized models exhibit significantly more weight disentanglement compared to their non-linear counterparts, similar to the findings obtained for post-hoc linearization.\looseness=-1

\begin{figure}[ht]
    \centering
    \includegraphics[width=\textwidth]{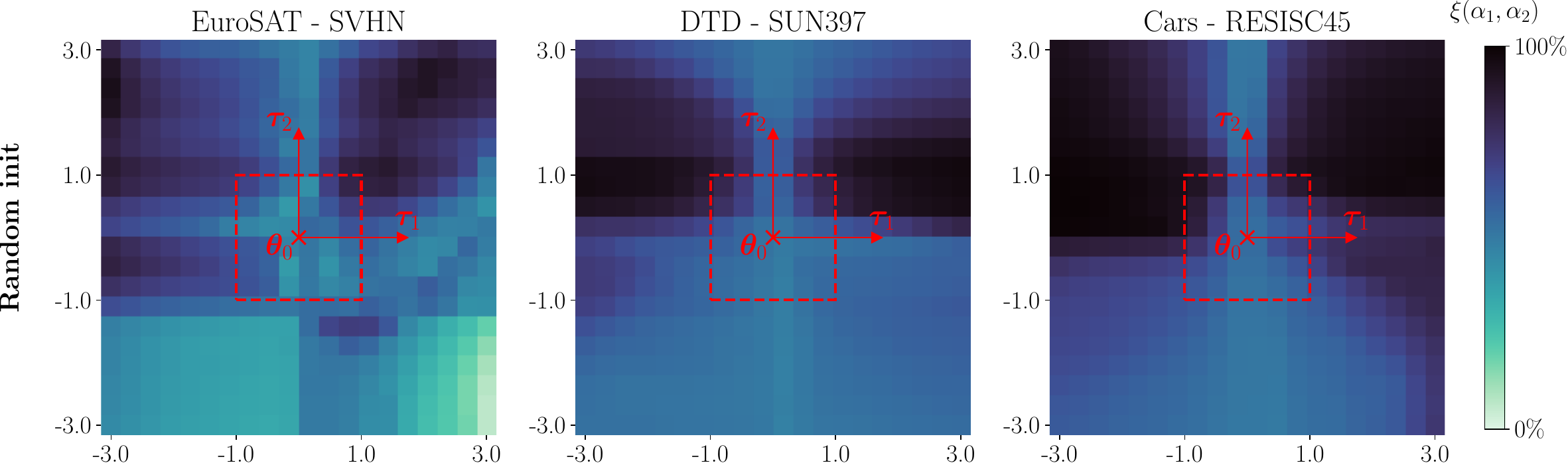}
    \caption{\textbf{Visualization of weight disentanglement from random initialization.} The heatmaps show the disentanglement error $\xi(\alpha_1, \alpha_2)$ of a ViT-B/32 fine-tuned from a random initialization non-linearly on different example task pairs. The light regions denote areas of the weight space where weight disentanglement is stronger. The red box delimits the search space used to compute $\alpha$ in our experiments.}
    \label{fig:random_consistency}
\end{figure}

Conversely, in \autoref{fig:random_consistency}, we showcase the disentanglement error of a CLIP ViT-B/32 model that was non-linearly fine-tuned  starting from a random initialization. In all panels, we observe a high disentanglement error, which supports the claim that weight disentanglement and, consequently, task arithmetic are emergent properties of pre-training.\looseness=-1

\subsection{Localization of eigenfunctions of CLIP's NTK}\label{ap:local_energy}

\begin{figure}[ht]
    \centering
    \includegraphics[width=\textwidth]{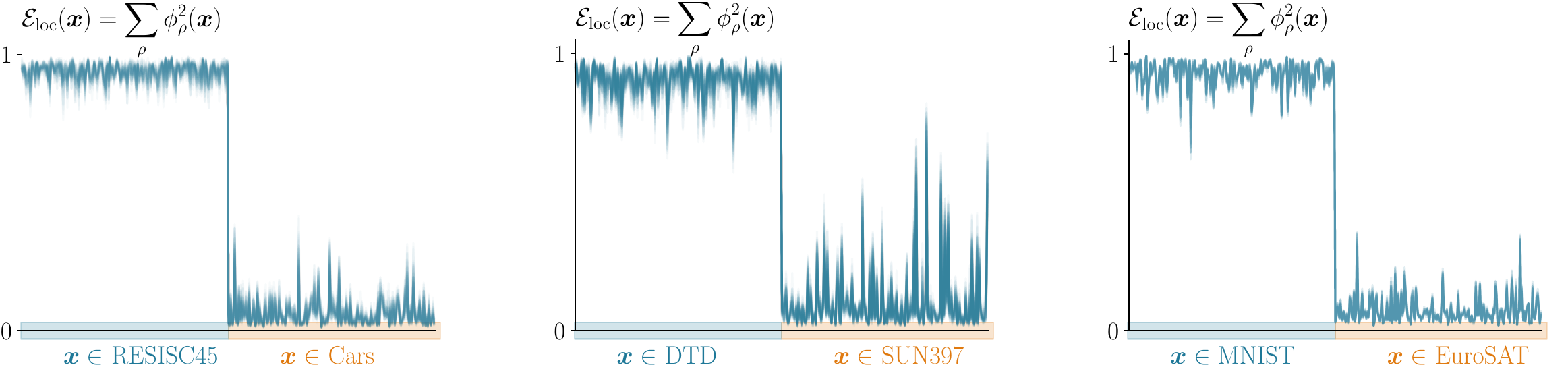}
    \caption{\textbf{Eigenfunction localization.} Estimated support of the eigenfunctions of the NTK of a ViT-B/32 CLIP model trained on different datasets. The plot shows the sum of the local energy of the eigenfunctions over a random subset of the training and control supports} \label{fig:eigenfunctions_all}
\end{figure}

In \autoref{fig:eigenfunctions_all}, we plot the local energy of the NTK eigenfunctions for a pre-trained CLIP ViT-B/32 model evaluated on three different data supports and control data supports. These panels complement the information presented in \autoref{fig:localization} in the main text, where we observed that the CLIP has eigenfunctions whose energy is concentrated on points belonging to the respective dataset.\looseness=-1 

\begin{figure}[ht]
    \centering
    \includegraphics[width=\textwidth]{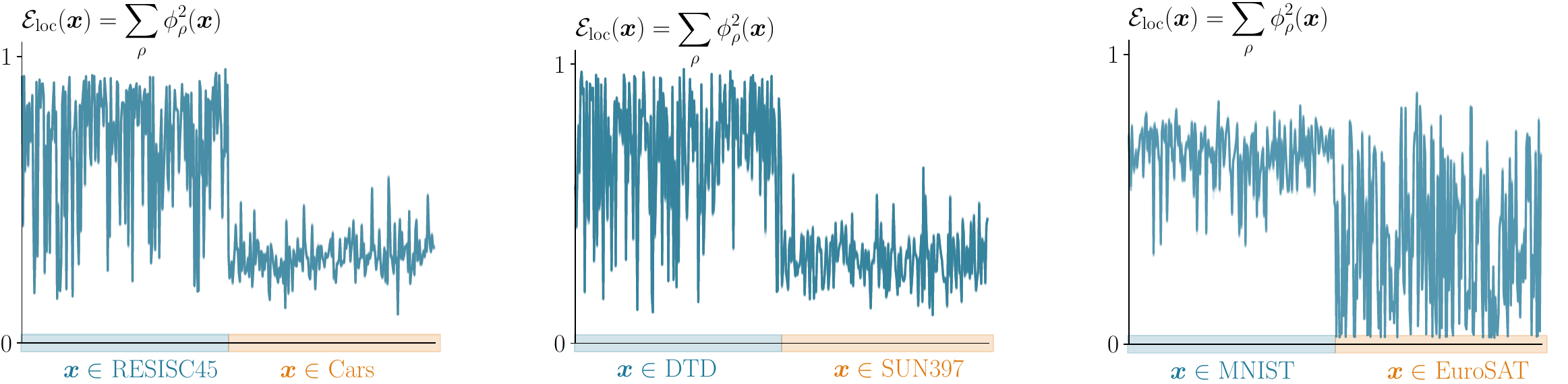}
    \caption{\textbf{Eigenfunction localization.} Estimated support of the eigenfunctions of the NTK of a randomly initialized ViT-B/32 model trained on different datasets. The plot shows the sum of the local energy of the eigenfunctions over a random subset of the training and control supports} \label{fig:eigenfunctions_init}
\end{figure}

In \autoref{fig:eigenfunctions_init}, we extend this analysis to a randomly-initialized CLIP ViT-B/32 model. In all panels, we observe a non-trivial but considerably poor degree of eigenfunction localization. This observation aligns with the finding that randomly-initialized linearized models cannot perform task arithmetic. Indeed, as we showed in the previous subsection, in this case the model's weights are not effectively disentangled, hindering its ability to perform task arithmetic operations. In summary, eigenfunction localization offers a complementary perspective on the limitations of randomly-initialized models.\looseness=-1

\subsection{Further experiments with randomly-initialized networks}\label{ap:random}

\begin{figure}[ht]
\centering
    \includegraphics[width=.8\textwidth]{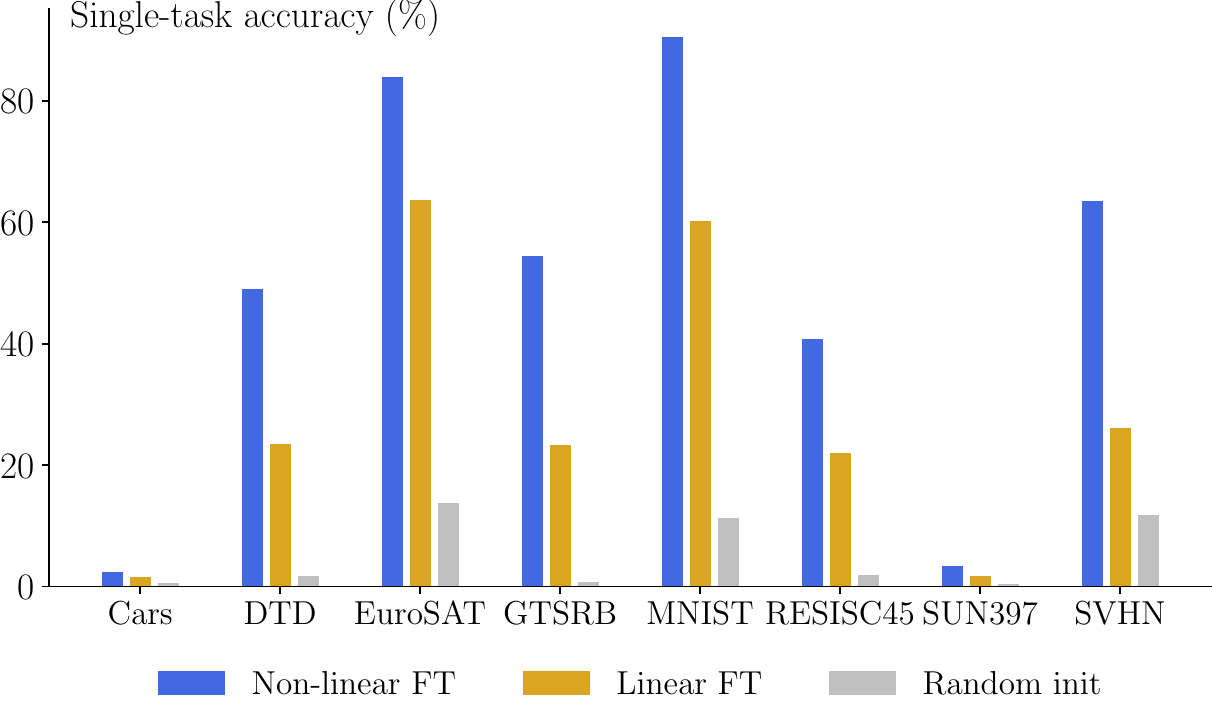}
    \caption{ViT-B/32}
    \includegraphics[width=.8\textwidth]{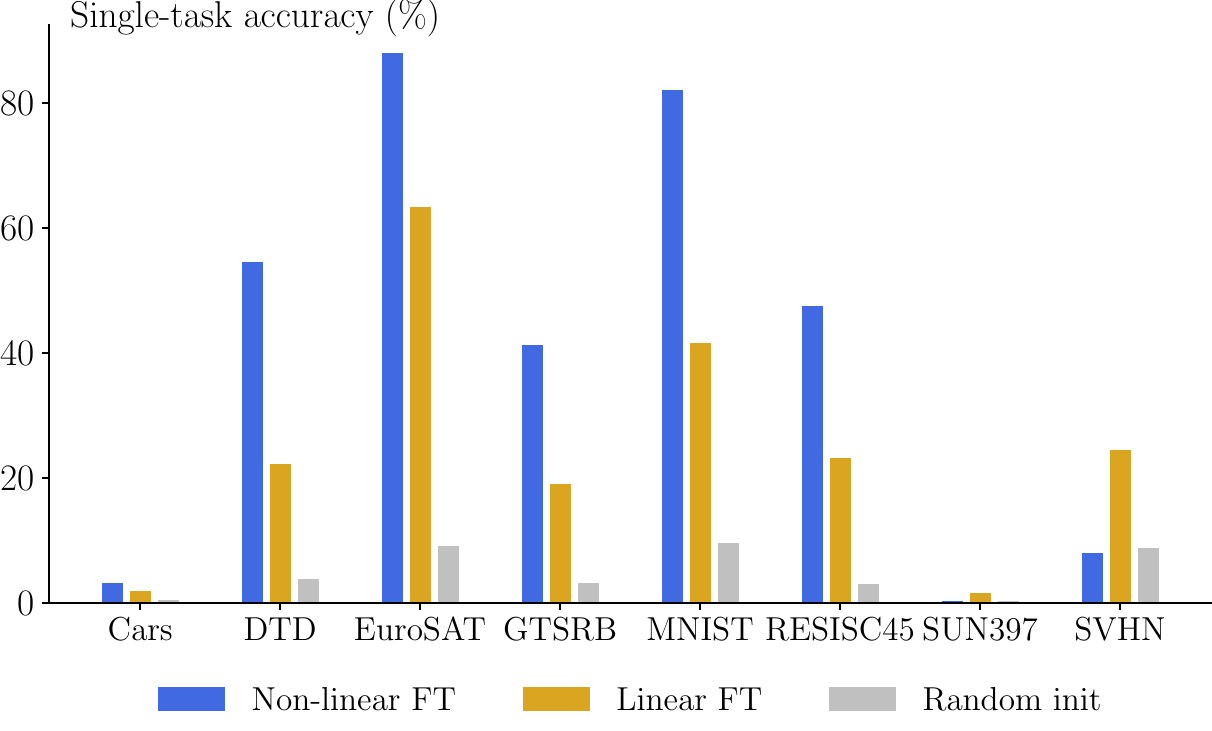}
    \caption{ViT-B/16}
    \includegraphics[width=.8\textwidth]{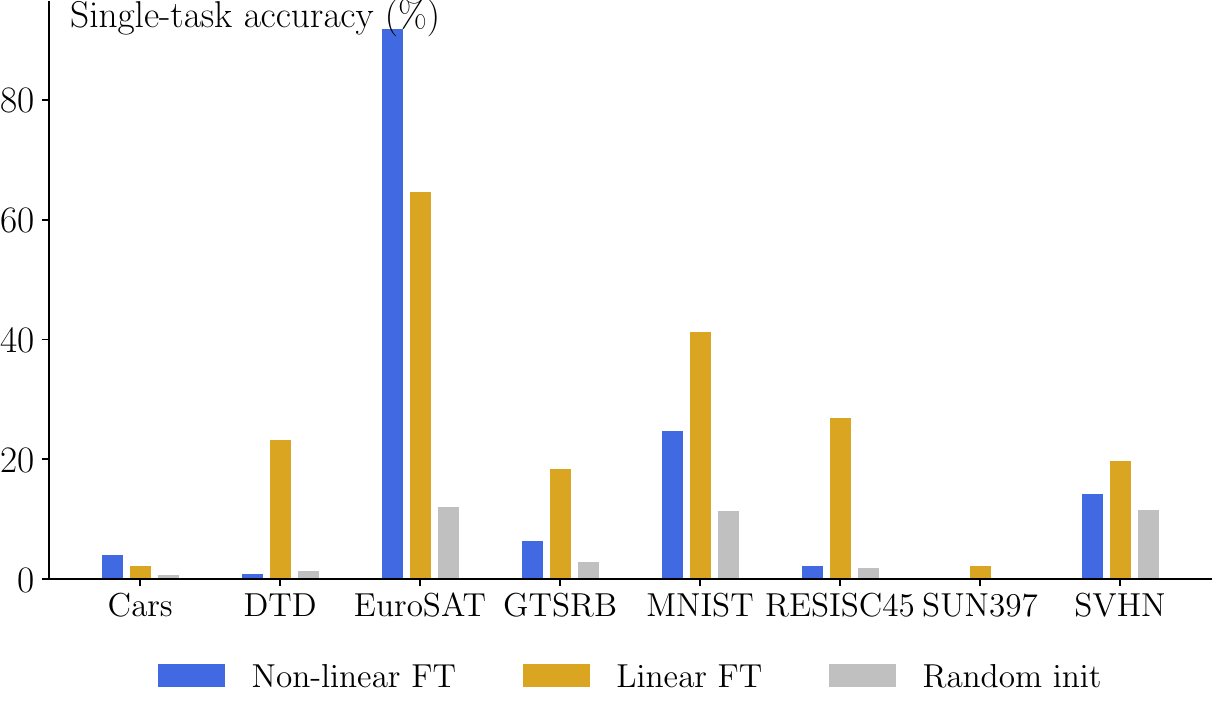}
    \caption{ViT-L/14}
\caption{\textbf{Single-task accuracies (random init).} Accuracy of different models obtained using different strategies on each of the tasks.}
\label{fig:rand_init_single_acc}
\end{figure}

We conclude by showing, in \autoref{fig:rand_init_single_acc}, the absolute single-task accuracy achieved by different CLIP ViT models that were fine-tuned from a random initialization. Both the base models achieve non-trivial or moderate accuracy on the majority of benchmark tasks, using both non-linear and linearized fine-tuning dynamics.\looseness=-1 

These findings reinforce the intuition that non-pretrained models are not failing in task arithmetic due to their inability to learn the task initially. Instead, as argued earlier, the primary reason for the failure of non-pre-trained models in task arithmetic is their lack of weight disentanglement.\looseness=-1

Interestingly, the performance of the randomly-initialized large model is generally poorer compared to the base models. This observation can be attributed to the models' tendency to overfit the training data, which is more likely to occur when a model has larger capacity.\looseness=-1

%----------------------------------------------------------------------------------------
%	POST-CONTENT THESIS PAGES
%----------------------------------------------------------------------------------------

\cleardoublepage\label{app:bibliography}

\manualmark

\markboth{\spacedlowsmallcaps{\bibname}}{\spacedlowsmallcaps{\bibname}}
\refstepcounter{dummy}

\addtocontents{toc}{\protect\vspace{\beforebibskip}}
\addcontentsline{toc}{chapter}{\tocEntry{\bibname}}

\printbibliography

\cleardoublepage\manualmark
\markboth{\spacedlowsmallcaps{Glossary}}{\spacedlowsmallcaps{Glossary}}

\refstepcounter{dummy}
\pdfbookmark[0]{Glossary Vitae}{Glossary}

\addcontentsline{toc}{chapter}{\tocEntry{Glossary}}

\chapter*{Glossary}

\begin{description}
    \item[Activation Function] A non-linear function applied to the output of a neuron in a neural network, such as ReLU ($\sigma(u)=\text{max}(0, u)$). It allows the network to learn complex, non-linear patterns in the data.

    \item[Arithmetic (Task)] See \textit{Task Arithmetic}.

    \item[Backpropagation] The algorithm used to train neural networks by efficiently computing the gradient of the loss function with respect to the network's weights. It works by applying the chain rule recursively through the network's layers.

    \item[Bayes-Optimal] Refers to a statistical method or predictor that achieves the lowest possible error rate for a given problem and data distribution. In the context of the thesis, it often represents the theoretical best-case performance against which other models are compared.

    \item[Belief Propagation (BP)] A message-passing algorithm for performing inference on graphical models. In the thesis, it is used to exactly compute the Bayes-optimal denoising process for the Random Hierarchy Model (RHM), revealing the evolution of latent variables during diffusion.

    \item[Compositionality] A principle suggesting that complex data or concepts are constructed from simpler, reusable parts or features, often in a hierarchical manner. For example, images are composed of objects, which are made of parts, textures, and edges. This thesis argues that compositionality is a key structure in data that deep learning models exploit to generalize efficiently.

    \item[Convolutional Neural Network (CNN)] A class of deep neural networks designed primarily for processing structured grid-like data, such as images. CNNs use convolutional filters to capture local patterns and enforce translational invariance through weight sharing, making them highly effective for tasks with spatial structure.

    \item[Curse of Dimensionality] A phenomenon where the amount of data required to achieve a certain level of statistical confidence grows exponentially with the number of dimensions (features) of the data. This makes learning from high-dimensional data, such as images, statistically intractable without a strong underlying structure in the data.

    \item[Deep Learning] A subfield of machine learning based on artificial neural networks with multiple layers (deep architectures). These models learn hierarchical representations of data, enabling them to tackle complex tasks like image recognition and natural language processing.

    \item[Diffusion Models] A class of generative models inspired by non-equilibrium statistical physics that learn to create data by reversing a gradual noising process. They start with pure noise and iteratively denoise it over time to produce a sample from the learned data distribution.

    \item[Feature Learning] A learning regime in neural networks where the model's parameters (weights) evolve significantly during training. This allows the network to adapt its internal representations (features) to the specific structure of the training data, in contrast to the `lazy' or kernel regime, where features remain largely fixed.

    \item[Fully Connected Neural Network (FCN)] A basic neural network architecture where each neuron in a given layer is connected to every neuron in the preceding and subsequent layers. FCNs are general function approximators but do not inherently encode structural priors, such as locality.

    \item[Generalization] A model's ability to perform accurately on new, unseen data after being trained on a finite dataset. Good generalization means the model has learned the underlying patterns in the data distribution, rather than simply memorizing the training examples.

    \item[Generalization Error] A measure of how well a trained model's predictions match the true outcomes for unseen data drawn from the same distribution. It is formally the expected value of the loss function over the entire data distribution.

    \item[Kernel] In machine learning, a function that measures the similarity between pairs of data points. Kernels allow algorithms to operate in a high-dimensional feature space without explicitly computing the coordinates of the data in that space.

    \item[Kernel Methods] A class of learning algorithms, such as Support Vector Machines or Kernel Ridge Regression, that use a kernel function to learn a predictor. In the infinite-width limit, certain neural networks become equivalent to kernel methods.

    \item[Kernel Ridge Regression] A regression algorithm that learns a function by minimizing a combination of the squared error on the training data and the norm of the function in a Reproducing Kernel Hilbert Space (RKHS).

    \item[Lazy Training Regime] See \textit{Neural Tangent Kernel (NTK) Regime}.

    \item[Locality] A structural property of data where interactions or correlations are primarily confined to small, local regions. For example, in an image, the value of a pixel is most strongly correlated with its immediate neighbors. This thesis shows that locality is a key prior that allows CNNs to defeat the curse of dimensionality.

    \item[Locally Connected Network (LCN)] A neural network architecture similar to a CNN but without weight sharing. It applies local filters to input patches, but each filter is unique to its location, thus capturing local structure without enforcing translational invariance.

    \item[Memorization] The phenomenon where a machine learning model, particularly an overparameterized one, learns to store and reproduce specific examples from its training set rather than learning the underlying data distribution. This typically leads to poor generalization.

    \item[Model Editing] The process of modifying the behavior of a pre-trained model to add new skills, correct errors, or remove undesired capabilities, often by directly manipulating its weights rather than fully retraining it.

    \item[Neural Scaling Laws] Empirical findings showing that the performance of deep learning models improves predictably as a power-law function of the model size, dataset size, and computational budget.

    \item[Neural Tangent Kernel (NTK)] A deterministic kernel that describes the training dynamics of an infinitely wide neural network under gradient descent. In this `lazy' regime, the network's parameters change very little, and the model behaves like a linear model in parameter space, equivalent to a kernel method with the NTK.

    \item[Overparameterization] The modern practice of training neural networks with far more parameters than the number of training examples. Classically, this was expected to lead to severe overfitting, but in deep learning, it often results in excellent generalization, a phenomenon that challenges traditional learning theory.

    \item[Probabilistic Context-Free Grammar (PCFG)] A type of generative grammar used in linguistics and computer science where production rules are assigned probabilities. The thesis uses PCFGs, specifically the Random Hierarchy Model, to model data with a hierarchical and compositional structure.

    \item[Random Hierarchy Model (RHM)] A synthetic generative model, based on probabilistic context-free grammars, used in the thesis to model data with a hierarchical and compositional structure. It consists of an ensemble of tree-like graphical models with random production rules, providing a tractable framework for studying how networks learn such structures.

    \item[Receptive Field] The region of the input space that a particular neuron in a neural network is sensitive to. In CNNs, the receptive field of a neuron is typically a small, local patch of the input.

    \item[Replica Method] A heuristic technique from statistical physics used to compute averages over disordered systems. In this thesis, it is applied to analyze the generalization error of kernel methods in teacher-student settings.

    \item[Sample Complexity] The number of training examples required for a model to learn a task to a desired level of accuracy. A key goal of learning theory is to understand how sample complexity scales with factors like data dimension.

    \item[Score Function] In the context of diffusion models, the gradient of the log-probability density of the data at a given time during the diffusion process ($\nabla_\x \log p_t(\x)$). Learning this function is the central task for the backward (generation) process.

    \item[Spectral Bias] The tendency of neural networks and kernel methods trained with gradient descent to first learn the components of a target function that correspond to the largest eigenvalues of the model's kernel. This often translates to a preference for learning low-frequency (smooth) functions first.

    \item[Statistical Mechanics] A branch of physics that uses statistical methods and probability theory to study the macroscopic behavior of systems composed of a large number of microscopic constituents. The thesis applies concepts and tools from statistical mechanics to analyze deep learning systems.

    \item[Supervised Learning] A machine learning paradigm where a model learns a mapping from inputs to outputs based on a labeled training set of input-output pairs.

    \item[Task Arithmetic] An empirical phenomenon observed in large pre-trained models where algebraic operations on the models' weight vectors (specifically, \textit{task vectors}) correspond to semantic operations on the tasks the model can perform. For example, adding or subtracting task vectors can combine or remove skills.

    \item[Task Vector] The difference in weights between a pre-trained model and a model that has been fine-tuned on a specific task ($\bm{\tau} = \vth_{\text{fine-tuned}} - \vth_{\text{pre-trained}}$). These vectors are the fundamental units used in task arithmetic.

    \item[Teacher-Student Setting] A theoretical framework used to analyze learning, where a `student' model (the learner) is trained on data generated by a `teacher' model (the target function). This allows for precise control over the properties of the data.

    \item[Unsupervised Learning] A machine learning paradigm where a model learns to find patterns and structure in unlabeled data, without explicit input-output examples. Generative modeling is a form of unsupervised learning.

    \item[Weight Disentanglement] A property identified in this thesis as the key mechanism for task arithmetic. It describes a situation where distinct directions in a model's weight space are associated with localized, non-interfering changes in the model's function space. This allows different tasks to be manipulated independently.

    \item[Zero-Shot Generalization] The ability of a large model to perform a task it has never been explicitly trained on, often by following a natural language description or prompt. This is a key emergent capability of large-scale pre-trained models.

\end{description}

\cleardoublepage\refstepcounter{dummy}
\pdfbookmark[0]{Curriculum Vitae}{Curriculum Vitae}

\addcontentsline{toc}{chapter}{\tocEntry{Curriculum Vitae}}



\section*{Supplementary material}

# CURRICULUM VITAE

| | | |
|---|---|---|
| *Contact Information* | EPFL<br>Institute of Physics<br>CH-1015 Lausanne<br>Switzerland | *Phone:* +41-21-693-98-00<br>*E-mail:* alessandro.favero@epfl.ch<br>*Website:* alesfav.github.io<br>*LinkedIn:* linkedin.com/in/alesfav/ |

*Research Interests* **Theory & science of deep learning:** generalization, data structure, compositionality, geometric priors, scaling laws, statistical physics of learning.
**Foundation models:** diffusion models, diffusion LLMs, vision-language models, multimodal LLMs, post-training, model editing, model merging.

*Education*

**EPFL**, Lausanne, Switzerland

Ph.D., Physics, AI                                                                             2025
Dissertation: *"The Physics of Data and Tasks: Theories of Locality and Compositionality in Deep Learning"*.
Advisors: Prof. Matthieu Wyart, Prof. Pascal Frossard.

**Sorbonne Université**, Paris, France

M.S., Fundamental Physics, Specialization in Complex Systems          2020
*Mention très bien* (highest honors).

**SISSA, ICTP, Politecnico di Torino**, Trieste-Torino, Italy

M.S., Physics of Complex Systems                                                    2020
*110/110 cum laude* (highest honors).
International Honors Track (competitive admission, 20 students per cohort).

**Politecnico di Torino**, Torino, Italy

B.S., Engineering Physics                                                                 2018

*Industry Experience*

**Amazon Web Services Artificial Intelligence (AWS AI)**, Santa Clara, California

Applied Scientist                                                          July to October 2023
- Internship at AWS AI Labs working on the alignment and robustness of multimodal LLMs with the fundamental research team led by Prof. Stefano Soatto.

*Academic Experience*

**EPFL**, Lausanne, Switzerland

Predoctoral Research Scholar                        November 2020 to April 2021
- *Master's valorization* research scholarship on the statistical physics of deep learning systems in the Institute of Physics.

Visiting Master's Thesis Student                       April 2020 to October 2020
- Thesis project *"Spectral analysis of infinitely-wide convolutional neural networks"* in the Physics of Complex Systems Laboratory led by Prof. Matthieu Wyart.

**INRiM – Italian National Metrology Research Institute**, Torino, Italy

Undergraduate Research Intern                October 2017 to January 2018

- Internship on space-time quantum correlations in the Quantum Optics Laboratory led by Prof. Marco Genovese.

*Refereed*
*Publications*

See also my Google Scholar and Semantic Scholar profiles.

*\* denotes co-first authorship.*

| Selected Talks | **Perimeter Institute**, Theory + AI: Theoretical Physics for AI, Waterloo, 2025. *Creativity by compositionality in generative diffusion models.* |
|---|---|

**Perimeter Institute**, Theory + AI: Theoretical Physics for AI, Waterloo, 2025. *Creativity by compositionality in generative diffusion models.*

**Johns Hopkins University Department of Physics & Astronomy**, Baltimore, 2025. *Creativity by compositionality in generative diffusion models.*

**IBM Research**, IBM Accelerated Discovery Seminar, Zurich, 2024. *Task arithmetic in the tangent space of pre-trained models.*

**EPFL Center for Intelligent Systems**, Lausanne, 2023. *Task arithmetic in the tangent space: Improved editing of pre-trained models.*

**37th Conference on Neural Information Processing Systems**, New Orleans, 2023. *Task arithmetic in the tangent space: Improved editing of pre-trained models.*

**Amazon AI Labs**, 2023. *Task arithmetic in the tangent space of pre-trained models.*

**MIT Center for Biological and Computational Learning**, Boston, 2023. *Deep convolutional networks in kernel regimes: invariances, locality, and compositionality.*

**NYU Center for Data Science**, New York, 2023. *Generalization properties of deep convolutional networks in kernel regimes.*

**American Physical Society** March Meeting, Statistical Physics Meets Machine Learning, Las Vegas, 2023. *Statistical mechanics of infinitely wide convolutional networks.*

**EPFL Institute of Physics**, Seminars in Physics of Bio/Complex Systems, Lausanne, 2023. *Symmetry, locality, and hierarchy in artificial neural networks.*

**Rice University**, Workshop on the Theory of Overparameterized ML, 2022. *Locality defeats the curse of dimensionality in convolutional teacher-student scenarios.*

Selected Posters

**Flatiron Institute**, Center for Computational Neuroscience, New York, 2024. *Hierarchies and compositionality in diffusion models.*

**Oxford Department of Statistics**, Workshop on Robustness in LLMs, Oxford, 2024. *Multi-modal hallucination control by visual information grounding.*

**Princeton University ORFE Department**, Princeton, 2022. *How wide convolutional neural networks learn hierarchical tasks.*

**Simons Foundation**, Simons Collaboration on Cracking the Glass Problem Meeting, New York, 2022. *Spatial locality and translational invariance in machine learning.*

Meetings and Schools

- Mathematics of machine learning, Italian National Institute for Advanced Mathematics (INdAM – Istituto Nazionale di Alta Matematica), Cortona, 2024 (*invited*).
- Analytical connectionism summer school, Flatiron Center for Computational Neuroscience, New York, 2024.
- Machine learning theory summer school, Princeton University, Princeton, 2022.
- Statistical physics and machine learning summer school, Les Houches School of Physics, Les Houches, 2022.

| | |
|---|---|
| *Teaching Experience* | **EPFL**, Lausanne, Switzerland |

Teaching assistant (*2024 Dean's award for teaching excellence*)    Fall 2021 to present

- PHYS-316 Statistical Physics II: Phase Transitions and Critical Phenomena (Spring 2023, Spring 2024).
- PHYS-467 Machine Learning for Physicists (Fall 2021, Fall 2022, Fall 2023).
- PHYS-421 Physics Projects I: Statistical Mechanics of Deep Learning (Fall 2021).

Guest lecturer at CS-625 Transfer Learning and Meta-Learning (Spring 2024).

*Advising and Mentoring*

Master's theses

- L. B., 2025, M.S. Physics, EPFL
- C. A. B., 2024, M.S. Cyber Security, EPFL–ETH Zurich.
- T. H., 2023, M.S. Physics, EPFL.

Semester projects (Ph.D.)

- A. A., 2024, Ph.D. Computer Science, EPFL.
- A. H., 2023, Ph.D. Computer Science, EPFL.

*Referee*

- Advances in Neural Information Processing Systems (NeurIPS). *2024 Top Reviewer.*
- International Conference on Learning Representations (ICLR). *2025 Notable Reviewer.*
- International Conference on Machine Learning (ICML).
- IEEE/CVF Conference on Computer Vision and Pattern Recognition (CVPR).
- Transactions on Machine Learning Research (TMLR).
- Physical Review Journals.

*Academic Service*

- ELLIS (European Lab for Learning & Intelligent Systems) PhD Recruiting Committee, Evaluator (a.y. 2024-25)

*Awards*

- G-Research EPFL PhD prize, 2025.
- Notable reviewer, ICLR, 2025.
- Dean's award for teaching excellence, EPFL, 2024.
- Top reviewer award, NeurIPS, 2024.
- Master's valorization research scholarship, EPFL, 2020.
- Merit-based scholarship for thesis abroad, Politecnico di Torino, 2020.
- Erasmus+ scholarship, Sorbonne Université, 2019.
- Fee reduction for high academic performance, Politecnico di Torino, 2019.
- Physics of complex systems international track fellowship, Politecnico di Torino, SISSA, ICTP, 2018.
- Top 200 engineering admission tests (among 8,000 applicants), Politecnico di Torino, 2014.

*Software*

- **Programming.** Python, C, C++, UNIX shell scripting.
- **Scientific and ML Libraries.** NumPy, Matplotlib, scikit-learn, PyTorch, JAX, HF Transformers.
- **HPC.** SLURM, Docker, K8s, Amazon Elastic Compute Cloud (EC2).

Web:
- Jekyll, WordPress.

Editing and Productivity Software:
- VSCode, PyCharm, CLion, JupyterLab.
- LaTeX, Microsoft Office, Google Docs, Keynote.

Operating Systems:
- Microsoft Windows, Apple macOS, Linux.

*Languages*  Italian (native), English (proficient, *IELTS Academic 8.5/9 CEFR C2*), French (intermediate).



\cleardoublepage\pagestyle{empty}

\hfill

\vfill

\pdfbookmark[0]{Colophon}{colophon}

\noindent This document was typeset using the typographical look-and-feel \texttt{classicthesis} developed by Andr\'e Miede. The style was inspired by Robert Bringhurst's seminal book on typography ``\emph{The Elements of Typographic Style}''. 
 
\bigskip

\noindent\finalVersionString

%----------------------------------------------------------------------------------------

\end{document}